\documentclass[a4paper, british]{memoir}

\usepackage{style}       
\usepackage{mnfrontpage} 
\usepackage{kantlipsum}  
\usepackage{mathtools}
\usepackage{subcaption}  
\usepackage[
    left = \flqq{},%
    right = \frqq{},%
    leftsub = \flq{},%
    rightsub = \frq{} %
]{dirtytalk}  

\usepackage{comment}

\setsecnumdepth{subsubsection}
\settocdepth{subsubsection} 

\usepackage{listings}
\usepackage[table,xcdraw]{xcolor}
\usepackage{tcolorbox} 

\usepackage{orcidlink}

\usepackage{algorithm}
\usepackage[noend]{algpseudocode}

\usepackage{multirow}
\usepackage{graphicx}

\usepackage{diagbox} 
\usepackage{pifont}
\newcommand{\cmark}{\textcolor{green}{\ding{51}} }%
\newcommand{\xmark}{\textcolor{red}{\ding{55}} }%

\usepackage{amsmath}
\usepackage{tikz}

\usepackage{makecell}
\usepackage{multirow}
\usepackage{wrapfig}

\usepackage{mathrsfs}

\newcommand{\defeq}{\coloneqq}   
\newcommand{\rdefeq}{\eqqcolon}  
\newcommand{\e}{\mathrm{e}}
\newcommand{\ee}{\mathrm{e}}
\newcommand{\dd}{\mathrm{d}}
\newcommand{\ii}{\mathrm{i}}
\newcommand{\ket}{\rangle}
\newcommand{\bra}{\langle}
\newcommand{\RR}{\mathbb{R}}

\newcommand{\pdf}{\mathrm{pdf}}

\newcommand{\sdf}{\mathrm{sdf}}
\newcommand{\EE}{\mathbb{E}}
\newcommand{\pd}{\partial}
\newcommand{\DNN}{\mathrm{DNN}}
\newcommand{\quotespace}{\vspace{1mm}}
\newcommand{\half}{\frac{1}{2}}
\newcommand{\SNR}{\mathrm{SNR}}
\newcommand{\std}{\mathrm{std}}
\newcommand{\LL}{\mathcal{L}}
\newcommand{\xx}{\mathbf{x}}
\newcommand{\yy}{\mathbf{y}}
\newcommand{\Tr}{\text{Tr}}
\newcommand{\TW}{\text{TW}}
\newcommand{\AAA}{\mathcal{A}}
\newcommand{\GG}{\mathcal{G}}
\newcommand{\UU}{\mathcal{U}}
\newcommand{\cU}{\mathcal{U}}
\newcommand{\cA}{\mathcal{A}}
\newcommand{\cG}{\mathcal{G}}
\newcommand{\cP}{\mathcal{P}}
\newcommand{\cQ}{\mathcal{Q}}
\newcommand{\cK}{\mathcal{K}}
\newcommand{\NN}{\mathsf{N}}
\newcommand{\FFT}{\mathrm{FFT}}
\newcommand{\Gtrue}{\mathcal{G}^{\dagger}}
\newcommand{\cvd}{\blacksquare}

\newtheorem{assumption}[theorem]{Hypotesis}

\newcommand{\note}[2][]{%
  \todo[linecolor=yellow!70!white, backgroundcolor=yellow!20!white, #1]{#2}%
}

\newtcolorbox[auto counter, number within=section]{recipe}[2][]{%
  colframe=blue, colback=white,
  fonttitle=\bfseries,
  title=Recipe~\thetcbcounter: #2,
  #1
}

\definecolor{darkblue}{RGB}{0,0,112}
\definecolor{ferrarired}{RGB}{255,40,0}
\definecolor{darkgreen}{RGB}{23,200,69}

\title{Lecture notes on Physics Informed Neural Networks, Neural Operators, and their applications}
\subtitle{From the PhD  course held at the Free University of Bozen-Bolzano, valid for the Ph.D. in Computer Science
}
\author{Alessandro Bombini\, \orcidlink{0000-0001-7225-3355}}
\kind{Istituto Nazionale di Fisica Nucleare, Sezione di Firenze \hfill\the\year} 

\begin{document}

    \frontmatter        

    \mnfrontpage

    \chapter{Abstract}
This is the set of lecture notes for the PhD course \href{https://www.unibz.it/en/faculties/engineering/phd-computer-science/study-course-offering/2025/36967}{\textit{Physics Informed Neural Network}, held at the University of Bozen/Bolzano} in the academic year 2025/2026. 

The goal of the course was to introduce the concept of Physics Informed Deep Neural Networks (PINN) and Neural Operators (NOs), discuss their implementation from scratch in PyTorch and using advanced ad-hoc developed open-source libraries such as NVIDia PhysicsNeMo to address real-world problems in various fields (engineering, physics, petroleum reservoir). We discuss recent topics such as Mixture-of-Models, Fourier Neural Operators, Physics-Informed Kolmogorov-Arnold Networks (PIKANs) and Fourier Neural Operators.

    \chapter{Acknowledgements}

I would like to thank Prof.~Oswald Lanz for inviting me to held a 20 hours PhD course on this subject at the \textit{Freie Universit\"at Bozen-Bolzano} (academic year 2025/2026), and for motivating me in start drafting this lecture notes. 

I would also like to thank Prof.~Stefano Berretti for granting me the opportunity to held this course for two years (a.y.~2023/2024, 2024/2025), in a smaller form factor. 

Furthermore, I would like to thank Dr.~Lucio Anderlini, for pointed me towards this topic, and for for the help in our joint work on the subject.

Finally, I would like to thank Dr.~Alessandro Rosa for having read the draft, and for all his invaluable comments, suggestions, and critiques to the draft, which really helped me improving the quality of this lecture notes. All errors, imprecisions, confusing points, etc., are all entirely my fault.


    \cleartorecto
    \microtypesetup{protrusion = false}
    \tableofcontents    
    \cleartorecto
    \listoffigures      
    \cleartorecto
    \listoftables       
    \microtypesetup{protrusion = true}

    \mainmatter         

    \part{Introductory Lectures}

    \chapter{Lecture 1: General Introduction to the course}\label{chap:1}

\section{A brief introduction: why should I care?}\label{sec:1:1}
The goal of the course is to introduce the concept of Physics Informed Deep Neural Networks (PINNs) and Neural Operators (NOs), discuss their implementation from scratch in PyTorch and using advanced ad-hoc developed open-source libraries such as nvidia-PhysicsNeMo to address real-world problems in various fields (engineering, physics, oil industry).

The goal is to have each lecture composed by two parts: 
\begin{enumerate}
    \item Theory   (frontal lectures)
    \item Practice (hands-on using JuPyTer notebooks)
\end{enumerate}

\noindent The code is available on GitHub\footnote{\url{https://github.com/androbomb/PINN_Course_2026}}. The latest version of the slides for the frontal lecture, and of the code for the hands-on lecture, as well as this lecture notes as pdf file, are also freely available \cite{bombini_2026_pinnunibz} (the live code is available in a dedicated GitHub repository \cite{Bombini_Code_repository_for_2026}). 

For the interested reader, \cite{GitHub2026LivingReview} contains an updated (as of \today) living review of paper on the subject. 


\section{A brief introduction: what is a Partial Differential Equation?} \label{sec:1:2}

Let us start by introducing a bit of notation. Let us be $f : [a,b]   \subset\mathbb{R} \to \mathbb{R}$ a function from a compact interval on the real numbers to the real numbers, and let it be continuous, i.e. $f\in C([a,b])$. The \textit{different quotient} $R_h [f](x)$ of $f$ in $x$ is 
\begin{equation}
    R_h [f](x) \coloneqq \frac{f(x+h) - f(x)}{h} \,.
\end{equation}
We say that $f$ is differentiable in $x$ if the limit 
\begin{equation}
    \lim_{h\to 0} R_h [f](x) = \lim_{h\to 0} \frac{f(x+h) - f(x)}{h} \,,
\end{equation}
exists. If $f$ is differentiable $\forall x \in [a,b]$, we say it is differentiable. The function obtained by this limit is the \textit{derivative} of $f$, and it is denoted with the Newton notation $f'(x)$, or with the Leibniz notation, $\frac{\dd f}{\dd x}$. If a function $f\in C([a,b])$ is differentiable, and its derivate is continuous $(a,b)$\note{We denote with curly braces the \textit{open} intervals, and with \textit{squared braces} the \textit{closed} intervals.}, we write it as $f\in C^1((a,b))$\footnote{Actually, to be rigorous, we need also that exists finite the right limit in $a$ and the left limit in $b$ of the different quotient. }. We can define derivates of higher order by repeating the process, 
\begin{equation}
    f^{(n)}(x) = \frac{\dd^n f(x)}{\dd x} \coloneqq \lim_{h\to 0} \frac{f^{(n-1)}(x+h) - f^{(n-1)}(x)}{h} \,,
\end{equation}
and if a function is continuous with up to $n$ continuous derivates in the interval $[a,b]$, we write $f\in C^n([a,b])$.

The interpretation of $f'(x)$ is that it defines the angular coefficient of the tangent line to $f$ in $x$. 

If, for example, $f\in C^3([a,b]),$ we can approximate the behaviour of $f$ in a small interval $h$ nearby any point $x\in[a,b]$ via its Taylor expansion
\begin{equation}
    f(x+h) = f(x) + h f'(x) + \frac{h^2}{2} \, f''(x) + O(h^3),
\end{equation}
where we have introduced the \textit{big-O} notation, where $O(h^2)$ describes the order of error (in powers of $h$) we commit by approximating $f$ with its second-order Taylor expansion. We can generalise it to higher powers of $n$; nevertheless, even if a function is $C^{\infty}([a,b])$, i.e. for every $n\in \mathbb{N}, f^{(n)}$ exists and it is continuous, it does not always admits a complete Taylor expansion. Such functions are $f\in C^\omega([a,b])$, and for those we have
\begin{equation}\label{eq:1:taylorexp}
    f(x+h) = \sum_{n=0}^{\infty} \frac{h^n}{n!} \, f^{(n)}(x) \,,
\end{equation}
which is an \textit{exact} relation, i.e. we can \textit{represent} $f\in C^\omega([a,b])$ with its Taylor expansion. If $0 \in [a,b]$, the Taylor expansion in $0$ of $f\in C^\omega([a,b])$ is called the MacLaurin series. Fortunately, the most famous functions (like $\sin, \cos, \exp, \log$) are $C^\omega$ in certain subsets of $\mathbb{R}$.

We can move to higher dimensional settings, e.g. $\mathbf{f}: \Omega \subset \mathbb{R}^m \to \mathbb{R}^n$. Let select an orthonormal basis for $\mathbb{R}^n$ as $\{\hat{\mathrm{e}}_i\}_{i=1, \ldots n}$, i.e. $\hat{\mathrm{e}}_i \cdot \hat{\mathrm{e}}_j = \delta_{ij}$ . We can then write without loss of generality  $\mathbf{f}(x) = f_i(x) \,\hat{\mathrm{e}}_i$, where $f_i$ are the function components  $f_i: \Omega \subset \mathbb{R}^m \to \mathbb{R}$. If $\Omega \subset\RR^m$ is open, we can define the \textit{directional derivative} of $\mathbf{f}$ along the vector $\mathbf{v} \in \mathbb{R}^m$ if the limit 
\begin{equation}
    \lim_{t\to 0} \frac{\mathbf{f}(\mathbf{x}  + t\mathbf{v} ) - \mathbf{f}(\mathbf{x} ) }{t  }
\end{equation}
exists. In that case, we write it as the \textit{directional derivative}
\begin{equation}
    D_{\mathbf v}\mathbf f(\mathbf x) \coloneqq \lim_{t\to 0} \frac{\mathbf{f}(\mathbf{x}  + t\mathbf{v} ) - \mathbf{f}(\mathbf{x} ) }{t } \,,
\end{equation}
and the matrix of partial derivatives is explicitly introduced as the Jacobian,
\begin{equation}
    D\mathbf f(\mathbf x)
 \coloneqq
 \left(\frac{\partial f_i(\mathbf x)}{\partial x_j}\right)
 \in\mathbb R^{n\times m},
\end{equation}
with
\[
D_{\mathbf v}\mathbf f(\mathbf x)
 = D\mathbf f(\mathbf x)\,\mathbf v.
\]

If $\mathbf{v}$ is one of the basis versor $\{\hat{\mathbf{u}}_j\}_{j=1, \ldots, m}$ of $\RR^m$, we have the \textit{partial derivative}\begin{equation}
    \frac{ \partial\mathbf{f}(\mathbf{x})  }{\partial x_j} \coloneqq \lim_{t\to 0} \frac{\mathbf{f}(\mathbf{x}  + t\mathbf{\hat{u}}_j ) - \mathbf{f}(\mathbf{x} ) }{t } \,.
\end{equation}
The vector of the derivatives along all directions is called\note{Notice that, for $n>1$, this is indeed a matrix. Nevertheless, sometimes, especially in Physics literature, it is still denoted with $\nabla$.} \textit{gradient} of $\mathbf{f}$
\begin{equation}
    D\mathbf{f}(\mathbf{x}) = \left( \frac{ \partial\mathbf{f}(\mathbf{x})  }{\partial x_1} \,, \frac{ \partial\mathbf{f}(\mathbf{x})  }{\partial x_2}, \ldots, \frac{ \partial\mathbf{f}(\mathbf{x})  }{\partial x_m}  \right).
\end{equation}
If $\nabla \mathbf{f}$ exists and it is continuous on $\Omega$, we can easily see that
\begin{equation}
    D_{\mathbf{v} } \mathbf{f}(\mathbf{x}) = \mathbf{v} \cdot \nabla \mathbf{f}(\mathbf{x}) .
\end{equation}
For $n=1$,
\begin{equation}
    \nabla_{\mathbf{v} } {f}(\mathbf{x}) = \mathbf{v} \cdot \nabla {f}(\mathbf{x}) .
\end{equation}

After this \textit{fast and dirty} recap of real analysis, we can introduce the concept of differential equation. While an equation is a relation between two functions, and comprises the standard operations (sum, multiplications, and composition), a \textit{differential equation} is a relation between functions \textit{and their derivates}. If it is a relation between univariate functions (i.e., functions of a single variable) and their derivatives, it is called an \textit{ordinary differential equation} (ODE). If it is a relation between multivariate functions - sometimes called \textit{fields} - it is a \textit{partial differential equation} (PDE). The order of the highest derivate appearing in the relation is the order of the differential equation. 

When dealing with differential problems, we need to furnish a set of additional constraints, called boundary conditions (BC), which describes the behaviour of the solution $u : \overline{\Omega} \subset  \mathbb{R}^m\to \mathbb{R}^n$\note{With $\overline{\Omega}$ we denote the closure of $\Omega$, i.e. $\overline{\Omega} \defeq \Omega\cup\partial\Omega$. We will later on simplify this notation, at the sole scope of enraging our mathematician friends, and to not put an \textsc{overline} latex command.} of the differential problem at the boundary of the space, i.e. $\left. u \right|_{\partial \Omega}$. If the function is defined also to have a time variable, i.e. $u : [0, T] \times \Omega \to \mathbb{R}^n$, we must furnish an initial condition (IC), describing the behaviour of the solution when $t=0$, i.e. $u(t=0, x)$.

In compact form, a PDE can be written as 
\begin{align}
    \mathcal{F}[u(x), x; \gamma] &= J(x) \,, \qquad x\in \Omega \,,\\
    \mathcal{B}[u(x), x; \gamma] &= g(x) \,, \qquad x\in \partial \Omega \,,
\end{align}
where $\mathcal{F}$ is the differential operator describing the evolution of the system, $J$ is the source term, $\mathcal{B}[u(x), x; \gamma]$ is the operator describing the boundary condition, and $g$ is the boundary value.

Usually, we will work with second-order linear PDE systems,  which are the most common in nature (yet, hPDEs outside this class may be relevant, e.g. the nonlinear second-order Allen–Cahn equation \cite{ALLEN19751017} or the third-order Korteweg–de Vries PDE \cite{Korteweg01051895}). Furthermore, (constant coefficients) second order linear PDE operator can be \textit{classified}: the general form of the second-order part is 
\begin{equation}
   \mathcal{F}[u] \supseteq  L^{ij} \partial_i\partial_j u \,,
\end{equation}
where $L_{ij}$ is the $m\times m$ constant-coefficient symmetric matrix\note{It is symmetric by the Schwarz theorem, by which $\partial_i \partial_j u = \partial_j \partial_i u$.} 

We can classify the second-order linear operators by studying the eigenvalues of the coefficient matrix $L_{ij}$, i.e., collecting its eigenvalues\note{Notice that, since $L_{ij}$ is real and symmetric, by the Spectral Theorem, it is \textit{diagonalisable}.}; we say that $L_{ij}$ is
\begin{enumerate}
    \item \textit{elliptic}, if all the eigenvalues have the same sign (and thus there are no zero eigenvalues);
    \item \textit{parabolic}, if one eigenvalue is zero and all the others have the same sign ;
    \item \textit{hyperbolic}, if one eigenvalue has a sign, and all the others have opposite sign;
    \item \textit{ultra-hyperbolic}, in all other cases (i.e., we have a multiplicity of eigenvalues with different signs).
\end{enumerate}

A notable \textit{elliptic} operator is the \textit{Laplacian}, $\Delta \defeq \nabla^2$; a notable \textit{parabolic} operator is the \textit{diffusion operator} $\partial_t - D^2\Delta$; a notable \textit{hyperbolic} operator is the \textit{wave} operator $\Box \defeq \partial_t^2 - c^2 \Delta$. 

Generally, the boundary conditions are
\begin{itemize}
    \item \textbf{Dirichlet}: $u(x) = g(x), x\in\partial \Omega$, i.e. the value of the function is \textit{fixed} at the boundary.
    \item \textbf{Neumann}: $\nabla_{\hat{n}} u = h(x), x\in\partial \Omega$, i.e. the value of the part of the gradient of the function normal to the boundary is fixed ($\hat{n}$ is the normal to $\partial \Omega$). This fixes the \textit{flux} through the boundary. 
    \item \textbf{Robin}: $u(x) + \alpha \nabla_{\hat{n}} u  = r(x), x\in\partial \Omega$, i.e. a linear combination of the Dirichlet and Neumann conditions.
\end{itemize}

\subsection{A first example: the Heat equation in 1+1 dimensions}

\subsubsection{A heuristic derivation of 1+1D heat equation as thermal diffusion} \label{subsubsec:1:2:1:1:DiffEqDer}

There are many way of deriving the diffusion equation; one example is starting from the continuity equation
\[
\partial_t u + \nabla \cdot \mathbf{j} = 0 \,,
\]
and assume that the flow $\mathbf{j}$ follows the Fick's Law, i.e., that it is somehow proportional to the spatial rate change of $u$, 
\[
\mathbf{j}(t,x) = - D^2\nabla u(t,x) \,.
\]

Another approach is to describe a random walk. In 1+1 (time+space) dimension (without loss of generality, WLOG), suppose we want to model a random walk of a particle that moves, at each time step $\tau$, of $h$ in a random direction, i.e. $x \to x\pm h$. Thus, the probability of finding the particle at time $t+\tau$ in $x$ is
\[
    p(t+\tau, x) = \half [p(t, x+h) + p(t, x-h)].
\]
Now, by Taylor expanding both side we see that
\begin{align*}
    p(t+\tau, x) &\approx p(t,x) + \tau \partial_t p(t,x) + O(\tau^2) \,, \\
    p(t, x\pm h)&\approx p(t,x) \pm h \partial_x p(t,x) + \frac{h^2}{2} \partial_x^2 p(t,x) + O(h^3) \,,
\end{align*}
so that we get\note{In the limit $h,\tau\to 0$ we need that \[\frac{h^2}{2\tau} \to D^2\,.\]}
\begin{align*}
    p(t,x) + \tau \partial_t p(t,x) &=  \half \left[  \left(  p(t,x) + h \partial_x p(t,x) + \frac{h^2}{2} \partial_x^2 p(t,x)\right) \right.  \\     
    & \hspace{20mm} \left. + \left( p(t,x) - h \partial_x p(t,x) + \frac{h^2}{2} \partial_x^2 p(t,x) \right) \right] \\
     \Rightarrow \partial_tp(t,x)& = \frac{h^2}{2\tau} \, \partial_x^2 p(t,x) \,.    
\end{align*}

Finally, a last way to obtain the heat equation, is to think about heat balancing, still in 1+1D. Suppose that we want to describe the diffusion in time of heat; we can model it as saying that the temperature $u$ at time $t+\tau$ in $x$ is 
\[
    u(t+ \tau, x) = u(t,x) + Q(t,x) , 
\]
where $Q$ is the heat flux. If we see the  We can suppose that such heat flux is proportional to the difference in temperature between the cell of size $h$ in $x$, and the two nearby ones, the one at $x+h$ and the one at $x-h$, i.e.\note{Notice that we can generalise that in higher dimensions and in the continuous limit, by integrating the (signed) difference $u(t, \mathbf{y}) - u(t, \mathbf{x})$ over a ball $B_d \subseteq\RR^d$ of radius $h$ centred around $x$.  }
\[
Q(t,x) \propto [ u(t, x+h) - u(t,x)] + [ u(t, x-h) - u(t,x)] \,,
\]
so that the flux is \textit{positive} if $u(t,x\pm h)>u(t,x)$, i.e. the cell at $x\pm h$ is at higher temperature (heat flows from higher temperature to lower ones). 

Now, again, by Taylor expanding, and calling $\alpha$ the proportionality constant, we get
\begin{align} 
    \Rightarrow\partial_t u(t,x) = \frac{\alpha h^2}{2\tau} \, \partial_x^2 u(t,x) \,.
\end{align}

\subsubsection{Heated bar thermalisation via diffusion}

Let us suppose we want to describe the thermalisation of a metallic bar of length $\ell=2$, We thus describe our system with the coordinates $(t,x) \in [0, T] \times [-1, +1]$. Let now suppose that we have heated the centre of the bar (with a cosinusoidal shape, i.e. $u(t=0,x)=\cos \frac{\pi x}{2}$, and suppose we put the boundary of the bar on a thermal bath, keeping it at $u(t, x=\pm 1)=0$. The system is described by the Cauchy problem
\begin{align}
    \partial_t u(t,x) &= D^2 \partial_x^2 u(t,x) \,, \quad (t,x)\in [0, T] \times [-1, +1] \\
    u(t=0, x)&= \cos \left( \frac{\pi x}{2}\right) ,  \\
    u(t, x) &= 0, \quad x\in \partial[-1, 1] = \{-1, +1\}.
\end{align}

This system can be easily solved by \textit{the separation of variables}: let us suppose that we can rewrite the solution as $u(t,x) = T(t) X(x)$, where $T,X$ are two univariate functions we need to find. In this case, the PDE becomes
\begin{equation}
    \dot{T}(t) X(x) = D^2 T(t) X''(x) \Rightarrow \frac{\dot T(t)}{T(t)} = D^2 \frac{X''(x)}{X(x)} \,,  
\end{equation}
which is possible only if
\begin{equation}
    \frac{\dot T(t)}{T(t)} = - \lambda^2 = D^2 \frac{X''(x)}{X(x)} \,, \quad \lambda \in \mathbb{R}  \,.
\end{equation}
This gives two simple ODEs, whose general solutions are
\begin{align}
    T(t) &= T_0 \exp \left(- \lambda^2 (t - t_0) \right) , \\
    X(x) &= X_0 \cos \left( \frac{\lambda}{D} (x-x_0)\right) + \tilde X_0 \sin \left( \frac{\lambda}{D} (x-x_0)\right) ,
\end{align}
where $T_0, X_0, \tilde X_0, x_0$ are constants to be determined. 

Now, imposing the boundary condition, we see that we can fix $t_0=0 = x_0$, $T_0 = X_0 = 1, \tilde X_0 = 0$. Finally, to satisfy the initial condition, we need that $2\lambda = \pi D$. Thus the solution is 
\begin{equation}
    u(t,x) = \e^{- \frac{\pi^2 D^2}{4} t} \cos \left( \frac{\pi x}{2}\right) .
\end{equation}

Let us now interpret the role of the \textit{Dirichlet boundary conditions}: we have imposed that the function has to have constant value at the boundary (being in contact with the thermal bath); thus, at the boundary, we have a non-zero \textit{outward flux}\note{The Flux definition contains a $-D^2$ factor because it can be seen as the integral over the domain of the variation in time of the field; then, applying the Gauss (de Rahm) theorem, we get the expression known as Fick's law. }
\begin{equation}
\begin{split}
    \left. \Phi \right|_{\partial[-1,+1]} &\defeq -D^2 \sum_{i=L,R} \hat{n}_i \cdot \nabla u \\
    &= -D^2\sum_{\pm 1} \mp \partial_x u  \\
    &=  -D^2\sum_{\pm 1} \pm \frac{\pi}{2} T(t) \left. \sin \left( \frac{\pi x}{2}\right) \right|_{x=\pm 1} \\
    &= -D^2\pi  \e^{- \frac{\pi D}{4} t} \,,
\end{split}
\end{equation}
which signals the outward flux, i.e. the loss of heat through the boundary. Indeed, the thermal bath we have inserted our bar in is \textit{thermalising} our object.

\subsection{The effect of Robin conditions: the loose end violin string} \label{subsec:1:2:1:Loose_String}

We want to describe the effect of a pluck on a violin string, which is fixed at the boundary \cite{Gustafson1998}; the system is 
\begin{equation}
    \begin{split}
        &\Box u(t,x) = 0 \,, \qquad \qquad  (t,x) \in [0,T]\times [0,1] \,, \\
        &\begin{cases}
            u(t,x=0) = 0 \,, \\
            u(t,x=1) = 0 \,, 
        \end{cases}\\
        &\begin{cases}
            u(t=0,x) = f(x) \,,  & \text{(pluck)} \\
            \partial_t u(t=0,x) = 0 \,, 
        \end{cases}\\
    \end{split}
\end{equation}
where $\Box \defeq \partial_t^2 - c^2 \partial_x^2$.

Let us now suppose that we are cheap, and we have a cheap violin of poor quality. Its peg\footnote{In Italian, \textit{il bischero}.} does not perfectly holds the string, which can thus move \textit{orthogonally}. This effects is modelled by a possible flux of $u$ from a side; 
\note{More precisely, we can model the string as an \textit{elastically restrained} object, like a spring. If we have a \textit{tension} $\mathcal{T}$ and a \textit{stiffness} $k$ at the end, we have \[\mathcal{T} u_x (t, 1) + k \, u(t,1)) =0\,.\]}
the problem this requires a \textit{Robin} \note{With $\alpha=0$ we recover the pure Dirichlet Cauchy problem.} boundary condition 
\begin{equation}
    \begin{split}
        &\Box u(t,x) = 0 \,, \qquad \qquad  (t,x) \in [0,T]\times [0,1] \,, \\
        &\begin{cases}
            u(t,x=0) = 0 \,, \\
            u(t,x=1) +  \alpha \partial_x u(t, x=1) = 0 \,,  
        \end{cases}\\
        &\begin{cases}
            u(t=0,x) = f(x) \,,  & \text{(pluck)} \\
            \partial_t u(t=0,x) = 0 \,.
        \end{cases}\\
    \end{split}
\end{equation}

Both these systems can be solved analytically, with small differences. \note{I.e., we set
\begin{equation}
\begin{split}
    &u(t,x) = \sum_{n} X_n(x) T_n(t) \\
    & \Rightarrow \frac{\ddot{T}}{T}= - \lambda_n =\frac{X''}{X} \\
    &\Rightarrow \begin{cases}
        X'' + \lambda_n X = 0, \\ \ddot{T} + c^2 \lambda_n T = 0
    \end{cases}    
\end{split}
\end{equation}
}Using again the separation of variables, \note{Indeed, we have: 
\begin{equation}
\begin{split}
    \partial_t^2 u &= - c^2\lambda_n  u\,,  \\
    \partial_x^2 u &= -   \lambda_n u \,.
\end{split}    
\end{equation}
} we see that we can write 
\begin{equation}
    u(t,x) = \sum_{n=0}^\infty c_n \sin \left( \sqrt{\lambda_n} \, x \right) \cos \left( \sqrt{\lambda_n} \, c t \right) \,,
\end{equation}
where $\lambda_n$ is a sequence of real numbers labelled by a natural number $n$. 
We can now show that we can express the coefficients $c_n$  in terms of the pluck $f(x)$ as
\begin{equation}
    c_n = \frac{\int_0^1 f(x)  \sin \left( \sqrt{\lambda_n} \, x \right) \dd x }{\int_0^1  \sin^2 \left( \sqrt{\lambda_n} \, x \right) \dd x} .
\end{equation}
Indeed, from the first initial condition, 
\begin{equation}
    \begin{split}
        f(x) &= u(t=0,x) =  \sum_{n=0}^\infty c_n \sin \left( \sqrt{\lambda_n} \, x \right)  \\
        &\Rightarrow  c_n = \frac{\int_0^1 f(x)  \sin \left( \sqrt{\lambda_n} \, x \right) \dd x }{ \int_0^1  \sin^2 \left( \sqrt{\lambda_n} \, x \right) \dd x } \, , \quad  \forall n=1,\ldots, \infty .
    \end{split}
\end{equation}
where we have used the \textit{orthogonality} property of the sinusoidal basis, i.e.
\begin{equation}
    \begin{split}
        \int_0^1  \sin \left( \sqrt{\lambda_n}  \, x \right) \sin \left( \sqrt{\lambda_m}  \, x \right) \dd x = \delta_{mn} \, \int_0^1  \sin^2 \left( \sqrt{\lambda_n} \, x \right) \dd x .
    \end{split}
\end{equation}

This relation holds for these $\lambda_n$ because of the \textit{orthogonality of the eigenfucntions of the wave operator} respecting the Dirichlet/Robin BC. 

Indeed, let us pick 2 random eigensolutions, $X_a, X_b$\note{I.e., $X''_{a,b} + \lambda_{a,b} X_{a,b} = 0.$}; now, let's multiply the equation satisfied by one for the other, and subtract them:
\begin{equation} \label{eq:1:2:2:Robin1}
    \begin{cases}
        X_b (X''_{a} + \lambda_{a} X_{a}) = 0 \\
        X_a (X''_{b} + \lambda_{b} X_{b}) = 0
    \end{cases} \Longrightarrow (X''_a X_b - X_a X''_b) + (\lambda_a - \lambda_b)X_aX_b = 0.
\end{equation}
The first term can be written as a total derivative
\begin{equation}
     (X''_a X_b - X_a X''_b) = \partial_x (X'_a X_b - X_a X'_b)
\end{equation}
so that, if we integrate \cref{eq:1:2:2:Robin1} in the whole domain the relation above, we get
\begin{equation}
    [X'_a X_b - X_a X'_b]_0^1 + (\lambda_a - \lambda_b) \int_0^1 X_a(x) X_b  (x) \dd x = 0.
\end{equation}
But, by the Dirichlet BC, $X_{a,b}(0) = 0$, while, by the Robin BC, $\alpha X_a'(1) = - X_a(1)$, and so, the left term cancels out! This means that, if $\lambda_a \neq \lambda_b$ the integral must be zero, and this 
\begin{equation}
     \int_0^1 X_a(x) X_b  (x) \dd x = \delta_{a,b}\int_0^1 X_a(x)^2 \dd x .
\end{equation}

While the above results is general, we can explicitly write $\lambda_n$. For the \textit{good quality violin}, i.e. pure Dirichlet boundary conditions, we need to impose that 
\begin{equation}
    (\text{pure Dirichlet}) : \quad  u(t, x=1) = 0 \Rightarrow \sin \left( \sqrt{\lambda_n}  \right) = 0 \Rightarrow \lambda_n = n^2 \pi^2 \, \forall n \in \mathbb{N}\,.
\end{equation}

\noindent Reintroducing the \textit{length} of the string $\ell$ (via the shift $x\mapsto x/\ell$), we have
\begin{equation}
    (\text{pure Dirichlet}) : \lambda_n = \frac{n^2 \pi^2}{\ell^2} \,.
\end{equation}

\begin{figure}
    \centering
    \includegraphics[width=0.5\linewidth]{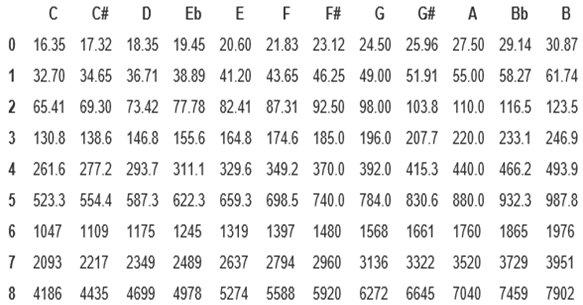}
    \caption{The tables of standard musical note frequencies. We move from column to column by changing $\ell$ of a certain amount. Instead, we move from row to row by increasing $n$. From \cite{FreqNotes}.}
    \label{fig:lect1:frequencies-of-notes}
\end{figure}

This gives the standard behaviour of \textit{tempered} notes while playing string instruments, like violins or guitars: moving the finger on the fretboard increases the frequency (since we reduce $\ell$), and a well-executed pluck creates all the harmonics of a certain \textit{note}. See \cref{fig:lect1:frequencies-of-notes}.

If we instead have the cheap violin (or we are describing us playing, and we are \textit{bad} musical players), we need to solve
\begin{equation}
\begin{split}
    (\text{Robin}) : \qquad &u(t,x=1) +  \alpha \partial_x u(t, x=1) = 0  \\
    &\Rightarrow  \sin \left( \sqrt{\lambda_n}  \right) + \alpha \sqrt{\lambda_n}  \cos \left(\sqrt{\lambda_n}  \right) = 0 \\
    & \Rightarrow  \tan \sqrt{\lambda_n} = - \alpha \sqrt{\lambda_n}  \,. \\
\end{split}
\end{equation}
$\forall n \in \mathbb{N}$, such that $\sqrt{\lambda_n} \neq \frac{\pi}{2} + k \pi, \forall k\in \mathbb{Z}, \forall n\in \mathbb{N}$, which is a \textit{transcendental} equation. It cannot be solved in closed form analytically, but we can obtain a \textit{numerical} approximation of it, via numerical methods (see \cref{lst:1:2:transcendental}).

\begin{lstlisting}[caption={Python function to solve the transcendental equation.}, label={lst:1:2:transcendental}, language=Python]
import numpy as np
from scipy.optimize import brentq

def solve_transcendental(a, N):
    """
    Solve   tan(sqrt(\lambda)) = a sqrt(\lambda) for the first N positive solutions.
    Returns array of x solutions.
    """
    roots = []
    n = 0
    while len(roots) < N:
        # Interval between asymptotes of tan(y) ; y = \sqrt{\lambda}
        left  = n * np.pi
        right = n * np.pi + np.pi/2 - 1e-6  # avoid asymptote
        # Define function f(y) = a*tan(y) + y
        f = lambda y:  np.tan(y) + a * y
        # Check if sign change exists in interval
        try:
            if np.sign(f(left + 1e-6)) != np.sign(f(right)):
                y_root = brentq(f, left + 1e-6, right)
                roots.append((y_root)**2)  # convert back to x
        except ValueError:
            pass
        n += 1
    return np.array(roots)
\end{lstlisting}

\begin{figure}[ht]
    \centering
    \includegraphics[width=0.85\linewidth]{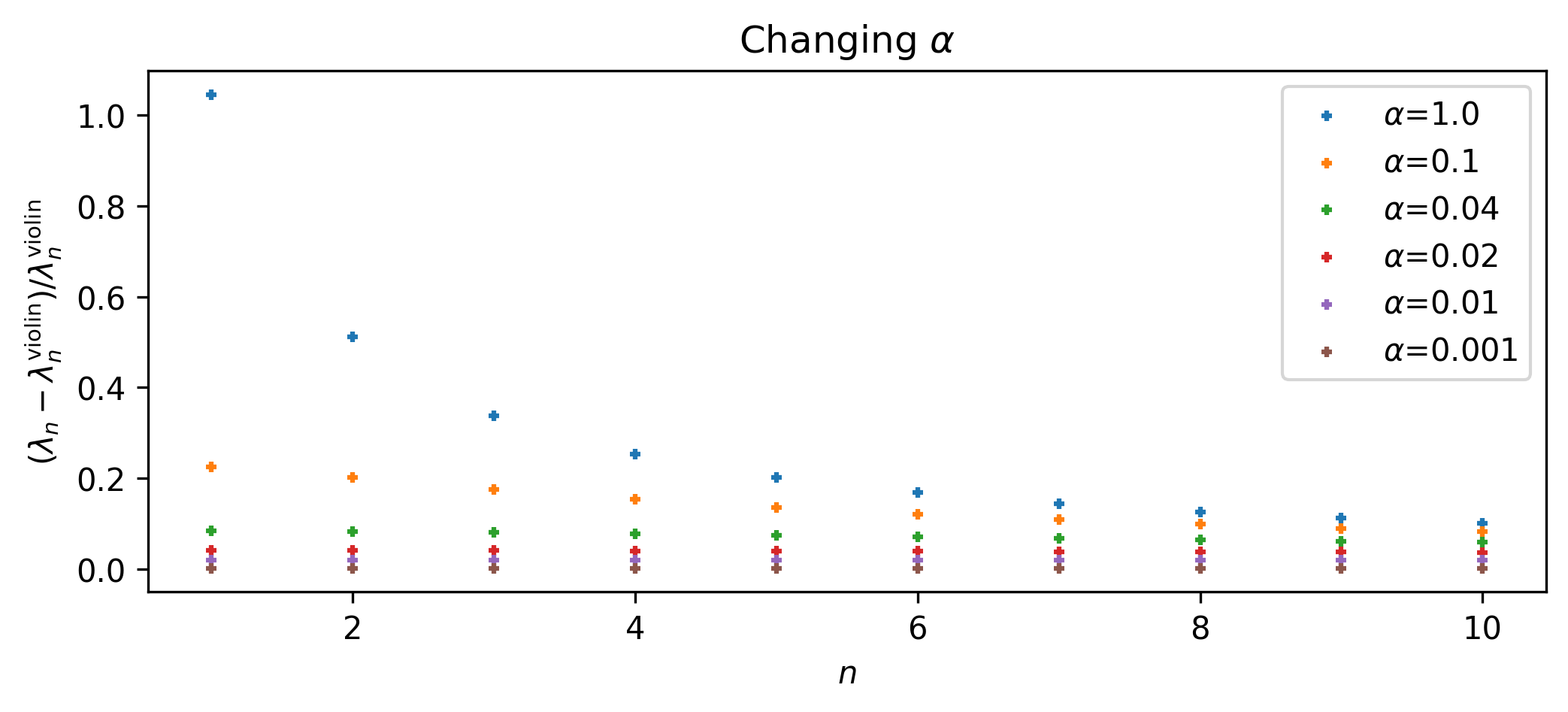}
    \caption{Numerical $\lambda_n$ for various $\alpha$.}
    \label{fig:1:2:violin_robin_1}
\end{figure}

If we numerically solve it for various $\alpha$, we see that the effect of the loose end (i.e., $\alpha>0$) is more relevant for \textit{lower} notes, as reported in \cref{fig:1:2:violin_robin_1}.

\section{Recap of Functional Analysis}\label{sec:1:3:func_analys}

To properly understand the approximation power of neural networks and the weak formulation of PDEs, we need a few notions from functional analysis. These concepts will be useful when we introduce PINNs in Lecture 3, where we'll see how neural networks approximate functions in Sobolev spaces and how the weak formulation enables residue-based training.

For a brief and formal introduction to the subject, see \cite{salsa2016partial}, chapter 7, or \cite{quarteroni2020numerical}, chapter 2, or the lecture notes \cite{RodriguezMIT}. 

\begin{definition}[Vector space]
A \textbf{vector space} $V$ over a field $\mathbb{K}$ (e.g., $\mathbb{R}$ or $\mathbb{C}$) is a set $V = \{v_i\}_{i=1, 2, \ldots}$ with two operations defined: an addition $+ : V\times V \to V$ and a scalar multiplication $\cdot : \mathbb{K}\times V \to V$, satisfying certain axioms: commutativity, associativity, null element, and inverse of addition, identity of multiplication, and distributivity.
\end{definition}

\begin{definition}[Finite-dimensional vector space] 
A vector space $ V $ is \textbf{finite-dimensional} if it has a finite spanning set (or equivalently, a finite basis). In other words, if $a=(a_1, \dots a_N) \in V$ for all sets \( E \subseteq V \) such that 
\[
    \sum_{i=1}^{N} a_i v_i = 0 \Longrightarrow a_1 = a_2 = \cdots = a_N = 0 \quad \forall v_1, \dots, v_N \in E, 
\] 
then \( E \) has a finite cardinality. \( V \) is \textbf{infinite-dimensional} if it is not finite-dimensional.
\end{definition}

\begin{definition}[Norm]
A \textbf{norm} on a vector space \( V \) is a function \( \|\cdot\| : V \rightarrow [0, \infty) \) satisfying the following three properties:
\begin{enumerate}
    \item (Definiteness) \( \|v\| = 0 \) if and only if \( v = 0 \),
    \item (Homogeneity) \( \|\lambda v\| = |\lambda| \|v\| \) for all \( v \in V \) and \( \lambda \in \mathbb{K} \),
    \item (Triangle inequality) \( \|v_1 + v_2\| \leq \|v_1\| + \|v_2\| \) for all \( v_1, v_2 \in V \).
\end{enumerate}
A \textbf{seminorm} is a function \( \|\cdot\| : V \rightarrow [0, \infty) \) which satisfies (2) and (3) but not necessarily (1), and a vector space equipped with a norm is called a \textbf{normed space}  (or normed vector space).
\end{definition}

\begin{proposition}[Metric induced by the norm]
Let \( (V, \|\cdot\| ) \) be a normed vector space. Then
\[
d(v, w) = \|v - w\|
\]
defines a metric on \( V \), which we call the \emph{metric induced by the norm}.
\end{proposition}

\begin{definition}
Let \( \{x_n\} \subset V \) be a sequence in a metric space \( (V, d) \). Then:
\[
\{x_n\} \text{ is \textbf{fundamental} if } d(x_n, x_m) \to 0 \text{ as } n, m \to \infty.
\]
\[
\{x_n\} \text{ \textbf{converges} to } x \in V \text{ if } d(x_n, x) \to 0 \text{ as } n \to \infty.
\]
\end{definition}

\begin{proposition}
If \( \{x_n\} \) converges in \( V \), then it is fundamental.
\end{proposition}
\begin{observation}
The converse is not always true. For example:
\begin{equation}
    x_n = \left(1 + \frac{1}{n}\right)^n  \in \mathbb{Q}, \forall n<\infty \underset{n\to \infty}{\longrightarrow}  e \notin \mathbb{Q}, \,
\end{equation}
i.e., $x_n \in \mathbb{Q}$ for all finite $n$, but the limit $\lim_{n\to \infty} x_n = e \notin \mathbb{Q}$.
\end{observation}
If the converse holds for all fundamental sequences in \( V \), then \( V \) is called \textbf{complete}.

The Euclidean norm on \( \mathbb{R}^n \) or \( \mathbb{C}^n \), given by
\[
\|x\|_2 = \left( \sum_{i=1}^{n} |x_i|^2 \right)^{1/2},
\]
is indeed a norm (the standard notion of “distance” that we're used to). But we can also define
\[
\|x\|_\infty = \max_{1 \le i \le n} |x_i|,
\]
which measures the largest magnitude among the components, and more generally (for \( 1 \le p < \infty \)),
\[
\|x\|_p = \left( \sum_{i=1}^{n} |x_i|^p \right)^{1/p},
\]
which is the $p-$norm. 

\begin{definition}[Banach space]
    A normed space is a \textbf{Banach} space if it is complete with respect to the metric induced by the norm.
\end{definition}
\noindent Why are Banach space relevant? because of the following
\begin{theorem}
For any metric space $X$, the space of bounded, continuous function on $X$, $C(X; \RR)$, is complete, and thus a Banach space with respect to the uniform norm (or sup norm)
\[
   \| F\|_\infty \defeq \sup_{x\in X} | F(x) | \,.
\]
\end{theorem}

Usually, we works with more restrictive spaces, where we have a concept of \textit{projectability}, or \textit{decomposition}, via a bilinear form (the scalar product).

\begin{definition}[bilinear forms]
Let \( X \) be a Banach space. A \textbf{form} is an application
\[
a : X \times X \rightarrow \mathbb{R}
\]
that assigns a real number to each pair of elements in \( X \). A form is called:
\begin{enumerate}
    \item \textbf{Bilinear} if it is linear in both arguments, i.e.,
    \[
        a(\lambda u + \mu w, v) = \lambda a(u, v) + \mu a(w, v), \quad
    a(u, \lambda w + \mu v) = \lambda a(u, v) + \mu a(u, w)
    \]
    for all \( \lambda, \mu \in \mathbb{R} \) and \( u, v, w \in X \);
    \item \textbf{Continuous} if there exists a constant \( M > 0 \) such that\note{As a consequence, $a(0, u) = a(u,0)=0 \; \forall u\in X$.}
    \[
    |a(u, v)| \leq M \|u\|_X \|v\|_X \quad \forall u, v \in X;  
    \]
    \item \textbf{Symmetric} if
    \[
    a(u, v) = a(v, u) \quad \forall u, v \in X.
    \]
    \item \textbf{Positive} if
    \[
    a(v,v) > 0, \forall v \in X \setminus \{0\} .
    \]
    \item \textbf{Coercive} if exists $\alpha\in \mathbb{R}^+$ (i.e. $\alpha > 0$) such that
    \[
    a(v,v) \ge \alpha \|v\|_X^2, \quad \forall v\in X .
    \]
    
\end{enumerate}
\end{definition}


\begin{definition}
Let \( X \) be a vector space over \( \mathbb{R} \).  
\( X \) is called a \textbf{pre-Hilbertian space} if there exists a bilinear form, called inner product \( \langle \cdot, \cdot \rangle : X \times X \rightarrow \mathbb{R} \), satisfying:
\begin{enumerate}
    \item \textbf{Positive definiteness:} \( \langle x, x \rangle \geq 0 \), and \( \langle x, x \rangle = 0 \Leftrightarrow x = 0 \),
    \item \textbf{Symmetry:} \( \langle x, y \rangle = \langle y, x \rangle \) for all \( x, y \in X \),
    \item \textbf{Bilinearity:} \( \langle z, a x + b y \rangle = a \langle z, x \rangle + b \langle z, y \rangle \) for all \( x, y, z \in X \) and \( a, b \in \mathbb{R} \).
\end{enumerate}
\end{definition}

\begin{remark}
    For complex vector spaces, the inner product is conjugate-linear in the first argument.
\end{remark}

\begin{definition}[Hilbert space]
If \( X \) is complete with respect to the metric induced by the norm \( \|x\| = \sqrt{\langle x, x \rangle} \), then \( X \) is called a \textbf{Hilbert space}.
\end{definition}

Why are Banach and (pre-)Hilbert spaces relevant for us? because we can study functions - and thus, solutions of our differential problems - in these spaces.
Let \( \Omega \subset \mathbb{R} \) be open, and let \( p \geq 1 \). Then \(  L^p(\Omega) \) is the set of functions \( f \) such that \( |f|^p \) is Lebesgue integrable, equipped with the norm
\[
\|f\|_{L^p(\Omega)} \equiv \left( \int_{\Omega} |f|^p \right)^{1/p}.
\]
This space is a Banach space (up to equivalence classes). When \( p = 2 \), (i.e., \(L^2(\Omega)\)) and we define the scalar product \( \langle \cdot, \cdot \rangle : L^2(\Omega) \times L^2(\Omega) \rightarrow \mathbb{R} \) by
\[
\langle f, g \rangle = \int_{\Omega} f g,
\]
then \( L^2(\Omega) \) becomes a Hilbert space.

Furthermore, Hilbert spaces are the ideal setting to solve problems in infinitely many dimensions.  They unify through the inner product and the induced norm, both an analytical and a geometric structure. We can use \textit{geometrical} structures of the space to separate the Hilbert space in two \textit{orthogonal} subsets (orthogonal w.r.t. the bilinear form), i.e. $H=V\oplus V^\perp$. This is possible thanks to the following

\begin{theorem}[Projection Theorem]
Let \( V \) be a closed subspace of a Hilbert space \( H \). Then, for every \( x \in H \), there exists a unique element \( P_V x \in V \) such that
\[
\|P_V x - x\| = \inf_{v \in V} \|v - x\|. 
\]
Moreover, the following properties hold:
\begin{enumerate}
    \item \( P_V x = x \) if and only if \( x \in V \),
    \item Let \( Q_V x = x - P_V x \). Then \( Q_V x \in V^\perp \defeq \{x \in H : \langle x, v \rangle = 0, \forall v \in V\}\) and
    \[
    \|x\|^2 = \|P_V x\|^2 + \|Q_V x\|^2.
    \]
\end{enumerate}
\end{theorem}

\begin{remark}
    Notice that this means that $H=V\oplus V^\perp$.
\end{remark}

This lays down the basis for a more restrictive, useful result, for certain Hilbert spaces (which, usually, are the one we have to work with):

\begin{definition}[Separable Hilbert Space]
A Hilbert space \( H \) is said to be \textbf{separable} if there exists a countable dense subset of \( H \).
\end{definition}

\noindent Those are relevant because they admits a countable orthonormal basis:
\begin{definition}
A countable orthonormal basis in a separable Hilbert space \( H \) is a countable set \( \{ \mathbf{v}_i\}_{i\in \mathbb{N}} \subset H \) whose linear span is dense in $H$ and 
\[
\langle \mathbf{v}_i, \mathbf{v}_j \rangle = \delta_{ij},
\]
where $\bra \cdot , \cdot \ket$ is the scalar product on $H$.
\end{definition}

\noindent With this concept at hand, we can represent any element of a separable Hilbert space (e.g., all functions in $L^2(\Omega)$) with a series over $\mathbb{N}$, the \textit{generalised Fourier series}.

\begin{definition}[Generalised Fourier Series]\label{def:1:3:genfourierseries}
If $H$ separable Hilbert space, any \( \mathbf{x} \in H \) can be written as
\[
\mathbf{x} = \sum_{i=1}^{\infty} \langle \mathbf{v}_i, \mathbf{x} \rangle \mathbf{v}_i.
\]
\end{definition}

This is already our \textbf{second} way to \textit{represent} a function we have seen so far; after the Taylor expansion, we have the (generalised) Fourier series. We will see a more relevant representation in a moment. 

\begin{figure}[t]
    \centering
    \includegraphics[width=0.75\linewidth]{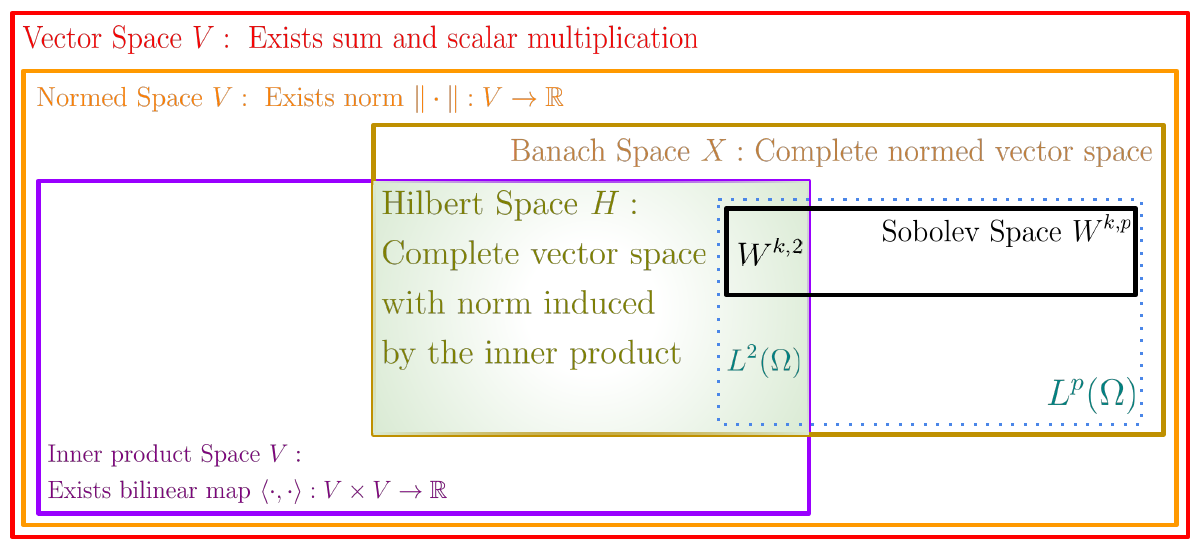}
    \caption{Pictorial representation of the spaces we have introduced.}
    \label{fig:1:3:spaces2}
\end{figure}

Before going on, we want to describe a more general function space, introducing also the derivatives of the elements of $L^p-$spaces.

\begin{definition}[Weak derivative]
    Let $\Omega \subset \RR^d$ open bounded, and let $f\in L^1(\Omega)$. We say that $f$ admits a \textit{weak derivative} $\tilde{\partial}_{x_j} f$ if there exists a function $g\in L^1(\Omega)$ such that
    \[
        \int_\Omega f \partial_{x_j} \varphi = - \int_\Omega g \varphi \,, \quad \forall \varphi \in C^1(\Omega), \left.\varphi\right|_{\partial \Omega} = 0 \,.
    \]
    We define 
    \[
        \tilde{\partial}_{x_j} f \defeq g .
    \]
\end{definition}

\begin{definition}[Sobolev space]
    Let \( \Omega \subset \mathbb{R}^d \) and let \( f \in L^p(\Omega) \). If $k\in \mathbb{N}$, we say that $f\in W^{k,p}(\Omega)$ if $\exists D^i f\in L^p(\Omega), \; \forall i=1, \dots,k$, such that
    \[
    \|f\|_{k,p} = \left(\sum_{|i| \le k} \|D^i f\|^p_{L_p}\right)^{1/p}, 
    \]
    where \( D^i f \) denotes the multi-index (weak) derivative of order \( i \), and \( \| \cdot \|_p \) is the \( L^p(\Omega) \) norm. The space \(  W^{k,p}(\Omega) \) equipped with this norm is called  Sobolev space, which is a Banach space.
\end{definition}

We have seen that $L^2(\Omega)$ is an Hilbert space. Similarly, in particular, \( W^{k,2}(\Omega) \) is a Hilbert space.

In \cref{fig:1:3:spaces2}, we see the relation between the spaces we have defined.

\subsection{From functions to function\textit{als}}

We know that \textit{functions} maps (vector) spaces to (vector) spaces. Yet, we are discussing (vector) spaces of \textit{functions}. So, how do we generalise these concept, i.e. a map from function spaces to function spaces?

\begin{definition}[operators \& functionals]
Let \( H_1, H_2 \) be Hilbert spaces. An \textbf{operator} \( L \in \mathcal{L}(H_1, H_2) \) is a map
\[
L : H_1 \rightarrow H_2,
\]
which is \textit{bounded}, i.e. \( \| L x\|_{H_2} \le c \|x\|_{H_1}, \, \forall x\in H_1 \).
If \( L(h_1 + a h_2) = L(h_1) + a L(h_2) \), then \( L \) is called a \emph{linear operator}.
If \( H_2 = \mathbb{R} \), i.e., \( L \in \mathcal{L}(H_1, \mathbb{R}) \), then \( L \) is called a (bounded) \emph{functional}.
\end{definition}

\begin{proposition}
    The space \( \mathcal{L}(H_1, H_2) \), endowed with the supremum norm, i.e.
    \[
    \| L\|_{\mathcal{L}(H_1, H_2)} = \sup_{\|x\|_{H_1} \le 1} \| L x \|_{H_2}
    \]
    is a Banach space. 
\end{proposition}

\begin{definition}[Dual space]
Let \( H \) be a Hilbert space. The collection of all bounded linear functionals on \( H \) is called the \textbf{dual space} of \( H \), and is denoted by \( H^* \) (instead of \( \mathcal{L}(H, \mathbb{R}) \)).
\end{definition}

\begin{remark}
    The dual of $L^2 (\Omega)$ is $L^2 (\Omega)$, i.e. $(L^2(\Omega))^* = L^2 (\Omega)$.
    
    More generally, for $1 < p < \infty$, the dual of $L^p(\Omega)$ is $L^q(\Omega)$ where $1/p + 1/q = 1$.
\end{remark}

To recap, we have that a \textbf{function} maps a space \( X \) to a space \( Y \), sending points in \( X \) to points in \( Y \).
An \textbf{operator} maps a function space \( H_1 \) to another function space \( H_2 \), sending functions to functions.
A \textbf{functional} maps a function space to a field (e.g., \( \mathbb{R} \)), associating scalar values to functions.

Like matrices represents the realisation of the action of a linear transformation on a vector space on a certain vector fields, linear operators are their generalisation on function spaces. 
Examples of linear operators on (certain) function spaces are derivatives, and thus differential operators.

\begin{example}
Let \( u, v \in C^{\infty}(\Omega) \), and \( a \in \mathbb{R} \). Then
\[
\Delta(u + a v) = \Delta u + a \Delta v.
\]
\end{example} 
\begin{example}
Let \( u, v \in C^{\infty}(\mathbb{R} \times (\Omega) \), \( a \in \mathbb{R} \), and define \( L = \partial_t - D^2 \Delta \). Then
\[
L(u + a v) = L(u) + a L(v).
\]
\end{example}

\begin{definition}
Let \( L \in \mathcal{L}(H_1, H_2) \) be a linear operator between Hilbert spaces. The set of points in \( H_1 \) mapped to the zero element of \( H_2 \) is called the \textbf{kernel} (or \emph{nucleus}) of \( L \),
\[
\ker(L) = \{ h \in H_1 : L(h) = 0_{H_2} \}.
\]
\end{definition}

\begin{observation}
Let \( u \in H(X) \) be a solution of the PDE defined by an operator \( L \), i.e., \( L(u) = f \). Then, for all \( v \in \ker(L) \), the function \( u + v \) is also a solution. This shows that solutions to linear PDEs are unique up to addition of elements from the kernel.
\end{observation} 

\noindent A relevant property of scalar products and Hilbert spaces is that exists a relevant representation theorem for bounded functionals (e.g., some bounded differential operators):

\begin{theorem}[Riesz Representation Theorem]\label{theorem:1:3:1:Riesz}
Let \( H \) be a Hilbert space. For each linear and bounded functional \( f \) on \( H \), there exists a unique element \( h_f \in H \) such that
\[
f(y) = \bra y, h_f \ket_H \quad \forall y \in H, \quad \text{and} \quad \|f\|_{H^*} = \|h_f\|_H. 
\]
Conversely, each element \( h \in H \) defines a linear and bounded functional \( f_h \) on \( H \) such that
\[
f_h(y) = \bra y, h\ket_H \quad \forall y \in H, \quad \text{and} \quad \|f_h\|_{H^*} = \|h\|_H. 
\]
\end{theorem}

In the case of $L^2(\Omega)$, another formulation of the Riesz representation theorem is that  any bounded linear functional $L \in \mathcal{L}(L^2(\Omega), \mathbb{R})$ can be represented as an integral 
\begin{equation}
    L[y] = \int_X y(x) h_f(x) \dd x \,,
\end{equation}
for some $h_f \in L^2(\Omega)$,  which can be rewritten, in Lebesgue integral formulation, as a \textit{measure} on the $X$ space:
\begin{equation}\label{eq:1:3:Riesz4Cyb}
    L[y] = \int_X y(x) \dd \mu(x) \,.
\end{equation}


Why all this functional analysis stuff? is it really useful? Well…. YES.

\begin{remark}
    The main idea behind the use of Hilbert (or, more generally, Banach) spaces is that the PDEs are difficult to solve using derivates in the classical sense (the one introduce before). While dealing with the \textit{weak formulation} of the PDE (which we will formally encounter later) is easier, which is, roughly speaking, the study of the functional associated to the PDE as the Energy is the functional associated to the Euler-Lagrange equations in Physics; and the study of these functionals requires spaces that admits a ``representation'', in the sense of \cref{theorem:1:3:1:Riesz}, to obtain existence and uniqueness of solutions; or, at least, spaces that are complete with respect to norms that descend from the functionals that we are considering (which are in general $L^p$ or $W^{k,p}$). 
    On the contrary, $C^m(\Omega)$ spaces are \textbf{not} complete w.r.t. those norms, so that we cannot apply functional results, which will also be used in numerical methods.  
\end{remark}

\noindent Let \( f \subset C^0(\Omega) \), with \( \Omega \subset \mathbb{R}^d \) open bounded, and let \( \Delta \equiv \nabla \cdot \nabla \) be the Laplace operator. Suppose we want to solve the PDE
\[
\Delta u = f.
\]
This is called the \textbf{strong formulation} of the PDE. Now, let \( v \in H \) be an arbitrary test function in a Hilbert space $H$ to be determined. Then
\[
\langle v, \Delta u \rangle = \langle v, f \rangle,
\]
where $\langle \cdot, \cdot \rangle$ is the inner product in $L^2(\Omega)$, i.e.,
\[
\int_{\Omega} v(x) \Delta u(x) \, \dd x = \int_{\Omega} v(x) f(x) \, \dd x.
\]
Applying the Gauss-Green-Ostrogradskiĭ-Stokes Theorem, 
\[
\int_\Omega v(x) \Delta u(x)  \, \mathrm{d}x = \int_{\partial\Omega} v(x)  \nabla u(x)  \cdot \dd \Sigma - \int_\Omega \nabla v(x)  \cdot \nabla u(x)  \, \mathrm{d}x , 
\]
yields
\begin{equation}
    \int_{\partial \Omega} v(x) \nabla u(x) \cdot \dd \Sigma - \int_{\Omega} \nabla v(x) \cdot \nabla u(x) \, \dd x = \int_{\Omega} v(x) f(x) \, \dd x,
\end{equation}
or, equivalently, 
\begin{equation}
      \int_{\Omega} \nabla v(x) \cdot \nabla u(x) \, \dd x = - \int_{\Omega} v(x) f(x) \, \dd x + \int_{\partial \Omega} v(x) \nabla u(x) \cdot \dd \Sigma,
\end{equation}
This is called the \textbf{weak formulation} of the PDE, if we impose that this relation must hold for all the \textit{test functions} $v$. Indeed, we can see that, while the first term represent the boundary terms, we have rewritten our PDE in terms of two bilinear form $a,b : H\times H \to \mathbb{R}$ on the Hilbert space $H$
\begin{equation}
    \begin{split}
        b(v,  u) &= a(v, f) ,
    \end{split}
\end{equation}
where $b : u,v \mapsto \int\nabla v \cdot \nabla u \, \dd x$.

\subsection{The Cybenko's Theorem}\label{subsec:1:3:2:UniversalApproxThm}

We now have almost all the ingredients to prove a very elegant theorem: the Cybenko's Universal Approximation Theorem \cite{Cybenko1989}. The last two ingredients we need are the Hahn-Banach separation theorem and the discriminatory measure definition

\begin{theorem}[Hahn-Banach Theorem]
Let \( V \) be a normed vector space, and let \( M \subset V \) be a subspace. Suppose \( u \in M^*\) is a linear map such that
\[
|u(t)| \leq C \|t\| \quad \forall t \in M,
\]
i.e., \( u \) is a bounded linear functional. Then there exists a continuous extension \( U \), with \( U \in   V^* \), such that
\[
U|_M = u \quad \text{and} \quad |U(t)| \leq C \|t\| \quad \forall t \in V,
\]
where the constant \( C \) is the same as above.
\end{theorem}
This mean that we can continuously extend a bounded linear map from a subspace of a functional space to the whole space, keeping its bounded property. 

Nevertheless, we need this result in a \textit{slightly} different setting, called the \textit{Hahn-Banach Separation Theorem}, which is actually a corollary of the former theorem

\begin{theorem}[Hahn-Banach Separation Theorem]\label{thm:1:3:HBST}
Let \(V\) be a normed vector space and let \(R \subset V\) be a proper closed subspace. 
Then there exists a bounded linear functional \(L \in V^*\), \(L \neq 0\), such that
\[
L(r) = 0 \qquad \forall r \in R.
\]
Equivalently, the annihilator \(\{L \in V^* : L(r)=0 \ \forall r \in R\}\) contains a non-zero functional.
\end{theorem}
\begin{proof}
We follow the steps of the original proof \cite{Cybenko1989}.

Let \(V\) be a normed vector space and \(R \subset V\) a proper closed subspace.
Since \(R\) is proper, pick \(f_{0} \in V \setminus R\)\note{1. Choose a point outside \(R\).}. Then the distance
\[
d = \inf_{r \in R} \|f_{0} - r\|
\]
is strictly positive.
Choose \(r_{0} \in R\) such that \note{2. Define a functional on a one-dimensional subspace.}
\[
\|f_{0} - r_{0}\| = d.
\]
Define the one-dimensional subspace
\[
M = \operatorname{span}\{f_{0} - r_{0}\},
\]
and define a linear functional \(u : M \to \mathbb{R}\) by
\[
u(\lambda (f_{0} - r_{0})) = \lambda d.
\]
This functional is bounded with \(\|u\| = 1\).
\note{3. Extend the functional using Hahn--Banach.}
By the Hahn--Banach extension theorem, \(u\) extends to a bounded linear
functional \(U : V \to \mathbb{R}\) with the same norm, i.e.
\[
U|_{M} = u, \qquad \|U\| = 1.
\]
For any \(r \in R\),\note{4. Show that the extension annihilates \(R\).}
\[
U(r) = U\big((r - r_{0}) + r_{0}\big)
      = U(r - r_{0}) + U(r_{0}).
\]
But \(r - r_{0} \in R\) and \(r_{0} \in R\), and by minimality of \(r_{0}\)
the extension satisfies
\[
U(r) = 0 \qquad \forall r \in R.
\]
Moreover,
\[
U(f_{0}) = U\big((f_{0} - r_{0}) + r_{0}\big)
         = U(f_{0} - r_{0}) = d \neq 0.
\]
Thus we have constructed a nonzero bounded linear functional \(U\) such that
\(U(r)=0\) for all \(r \in R\), proving the separation form of the
Hahn-Banach theorem.
\end{proof}

Finally, we introduce the concept of \textit{discriminatory} functions for a certain measure in a measurable space
\begin{definition}[Discriminatory function]
Let $X$ be a Banach space of functions over the real interval $I_n =[0,1]^n\subset\mathbb{R}^n$. A function \( \sigma \in X\) is said to be \emph{discriminatory} if the following implication holds:
Let \( \mu \in \mathcal{M}(I_n) \) denote the space of finite Borel measures on $I_n$. If\note{Notice that $y^T x = y\cdot x.$}
\[
\int_{I_n} \sigma(y^T x + \theta) \, \dd \mu(x) = 0
\]
for all \( y \in \mathbb{R}^n \) and \( \theta \in \mathbb{R} \), then \( \mu = 0 \). In other words, the vanishing of all such integrals implies that the measure \( \mu \) must be identically zero.
\end{definition}

The universal approximation property of neural networks is a key result that justifies their use as function approximators. We now sketch its proof, which relies on the functional analysis tools we just introduced.

\begin{tcolorbox}[colframe=ferrarired, colback=white, title=Universal Approximation Theorem]
Let $X$ be a Banach space of functions over the real interval $I_n =[0,1]^n\subset\mathbb{R}^n$. Let function \( \sigma : \mathbb{R} \to\mathbb{R}\) be any continuous, bounded,  discriminatory function. Then finite sums of the form
\begin{equation}\label{eq:1:3:ANNCyb}
    G(x) = \sum_{j=1}^{N} \alpha_j \sigma(w_j^T x + b_j) 
\end{equation}
are dense in \( C(I_n) \) (w.r.t. the uniform norm).
That is, for any $f \in C(I_n)$ and $\varepsilon > 0$, there exists a network \( G(x) \)  of the above form with sufficiently many neurons $N$ such that
\[
|G(x) - f(x)| < \varepsilon \quad \text{for all} \quad x \in I_n.
\]
\end{tcolorbox}

\begin{proof}
Let \( S \subset C(I_n) \) be the set of functions of the form \( G(x) \) as in \cref{eq:1:3:ANNCyb}. It is easy to see that \( S \) is a linear subspace of \( C(I_n) \). We will show  that the closure of \( S \) is all of \( C(I_n) \).

Assume  that the closure of \( S \) is not all of \( C(I_n) \)\note{1. We work by contraddiction. }. Then the closure of \( S \), called, let us say, \( R \), is a closed proper subspace of \( C(I_n) \). By the Hahn–Banach theorem\note{2. Use Hahn-Banach Separation Theorem \cref{thm:1:3:HBST}}, there is a bounded linear functional on \( C(I_n) \), call it \( L \), with the property that \( L \neq 0 \) but \( L(R) = L(S) = 0 \).

By the Riesz Representation Theorem\note{3. Use \cref{eq:1:3:Riesz4Cyb}.}, this bounded linear functional \( L \) is of the form
\[
L(h) = \int_{I_n} h(x) \, \dd\mu(x)
\]
for some \( \mu \in \mathcal{M}(I_n) \), for all \( h \in C(I_n) \). In particular, since $S$ is contained in its closure $R$, and \( \sigma(w^T x + b) \in S \) for all \( w,b \), it is also in $R$, and thus we must have
\[
\int_{I_n} \sigma(w^T x + b) \, \dd\mu(x) = 0
\]
for all \( w \in \mathbb{R}^n \), \( b \in \mathbb{R} \).

However, we assumed that \( \sigma \) is discriminatory\note{4. Impose discriminatory definition.}, so this condition implies \( \mu = 0 \), contradicting our assumption. Hence, the subspace \( S \) must be dense in \( C(I_n) \).
\end{proof}

 
\begin{remark}[Representations of functions]
    We have seen already three different way to represent \textit{any} function (under certain conditions). We have encountered \textit{Taylor expansion} (\cref{eq:1:taylorexp}), so that, for any $f\in C^\omega (\RR)$
    \[
        f(x) = \sum_{n=0}^\infty \frac{ f^{(n)} (x_0) }{n!} \,(x-x_0)^n =  \sum_{n=0}^\infty a[f](x_0) \,(x-x_0)^n \,,
    \]
    i.e., we can represent any function with a polynomial; furthermore, Taylor theorem furnishes two additional crucial ingredients:
    \begin{itemize}
        \item a way to define the polynomial coefficients $a[f](x_0)$, in terms of the function $f$ (and its derivatives) and of the point $x_0$ around which we are ``expanding'' it;
        \item a way to compute the \textit{residual} for a truncated expansion: 
        \[
            f(x) \approx \sum_{n=0}^N \frac{ f^{(n)} (x_0) }{n!} \,(x-x_0)^n + R_N(f; x_0) \,, \quad R_N (f; x_0)  \sim (x-x_0)^{N+1} \,,
        \]
        which allows us to estimate the numerical representation error efficiently and precisely. 
    \end{itemize}
    The Taylor representation naturally extends in higher dimensions, 
    \[
        f(\mathbf{x}) \approx f(\mathbf{x}_0) + \nabla f(\mathbf{x}_0) \cdot (\mathbf{x} - \mathbf{x}_0) + \frac{1}{2} \,  (\mathbf{x} - \mathbf{x}_0)^T \mathbf{H} (\mathbf{x}_0) (\mathbf{x} - \mathbf{x}_0) + R_2 \,,
    \]
    where $\mathbf{H}$ is the Hessian matrix, with \(H_{ij} = \partial_i \partial_j f\). 

    Unfortunately, the Taylor recipe is numerically expensive, and very limited to a certain region (within the ``radius of convergence'' of the series expansion). Another relevant expansion we have seen is the \textit{(generalised)-Fourier expansion} (\cref{def:1:3:genfourierseries}) of any element of an Hilbert space,
    \[
        \mathbf{x} = \sum_{i=1}^{\infty} \langle \mathbf{v}_i, \mathbf{x} \rangle \mathbf{v}_i ,
    \]    
    which, using the standard Fourier basis for an Hilbert space like $L^2([-\pi,+\pi]; \RR)$, for $f\in L^2([-\pi, +\pi]]; \RR)$ can be conveniently written as the more familiar formula:
    \[
        f(x) = \sum_{n=-\infty}^{+\infty} c_n \ee^{+\ii n x } \,, \qquad c_n = \frac{1}{2\pi}\int_{-\pi}^{+\pi} f(x) \, \ee^{- \ii n x }  \dd x \,,
    \]
    where the basis vector are \(\{v_n\} = \{\ee^{\ii n x }, n\in \mathbb{Z}\} \) and the product is 
    \[
        \langle g, f\rangle =  \frac{1}{2\pi}\int_{-\pi}^{+\pi} \overline{g(x)} f(x) \dd x.
    \]
    Also here we have the same two properties as above, with the \textit{finite approximation} being \cite{epstein2005well}
    \[
        f(x) \approx \sum_{n=-N}^{+N} c_n \ee^{+\ii n x } + R_N \,.
    \]
    The crucial point for Fourier representation, is that we have a relation between \textit{modes} (i.e., the $n$) and \textit{scale} of the \textit{features} in the function: lower $|n|$, long scale; higher $|n|$, short scale. 

    Nevertheless, we have a dependence on the number of basis elements used to the error rate, as expressed in the form of $R_N$. This is similar, yet even more convoluted, for the \textit{Universal Approximation Theorem} (\cref{eq:1:3:ANNCyb}); there, the $N$ we are referring to, are the \textit{Number of Neurons}, and the limit $N\to \infty$ is the \textit{infinite width limit} of the neural network. 
    
    Empirically, we observe that stacking multiple layers (deep networks) often achieves better approximation with fewer total neurons than widening a single layer, i.e. we can \textit{stack} multiple finite representation formulas (the \textit{layers} of the Multi-Layer Perceptron) and approximate the unknown function \textit{better} than widening the single-layer representation, in a more computationally efficient and effective manner. This `depth efficiency' is supported by theoretical results showing that certain functions require exponentially more neurons in shallow networks than in deep ones \cite{eldan2016power,telgarsky2016benefits}.
\end{remark}

\section{Recap of Monte Carlo Integration methods}\label{sec:1:4:MCI}

For an extended introduction to the subject, we refer to \cite{Landau2015, sanz2024first, Linge2020}, and references therein.

Sampling from a uniform distribution is straightforward using standard pseudorandom number generators (PRNGs), which produce sequences that approximate independent uniform random variables.
But what if we want to generate more general \textit{probability density functions} $\text{pdf}(x)$? Well, we can use the \textit{Monte Carlo Sampling Method} (one of its incarnations, at least). Indeed, we can draw $y\sim \mathcal{U}[0,1]$, the uniform distribution on the interval $[0,1]\subset \RR$. Its pdf is simply the constant function of value $1$ in the interval. The intuition here is that we can see the $[0,1]$ as the image of the cumulative distribution function (cdf) of \textit{any} pdf $f(x)$. Since the cdf is a right-continuous, monotone increasing, bounded function $F : \RR \to [0,1]$ for any pdf $f$\note{Equivalently, we can define the cdf as the linear, bounded functional over the Banach space $X \defeq L^1(\RR)$, $\Phi \in X^*= \mathcal{L}(X;\RR)$, with unit norm:
\begin{equation}
\begin{split}
      & \Phi : X \to \RR, \text{ s.t.} \\
      &\Phi : f \mapsto \int_{-\infty}^x f(s) \dd s .  
\end{split}
\end{equation}
}, we can see that any $x\sim f$ is mapped to $y=F(x)$, or, equivalently, that, drawing $y\in[0,1]$ can be mapped on $x = F^{-1}(y)$. But how are such $x$ distributed if $y\sim \mathcal{U}[0,1]$. Well... as $f(x)$! (i.e., they follow the distribution $f(x)$) Indeed, by a simple change of coordinates
\begin{equation}
    \begin{split}
        1 &= \int_0^1 \dd y = \int_{F^{-1}(0)}^{F^{-1}(1)} \frac{\dd F(x)}{\dd x} \, \dd x \quad \text{(where } y = F(x)\text{)} \\
        &= \int_{-\infty}^{+\infty} f(x) \, \dd x \qquad \qquad \qquad \quad\, \text{(} F'(x) = f(x)\text{).}
    \end{split}
\end{equation}
So that
\begin{equation}
    \begin{split}
        P(X\le x) &= P\left( F^{-1} (U) \right)  = P(U\le F(x)) = F(x) .\\
    \end{split}
\end{equation}

\noindent So the algorithm is: 
\begin{itemize}
    \item[0.] Choose the $f(x)$ to sample, (numerically) compute $F(x)$, and then its inverse, $F^{-1}(y)$.
    \item[1.] Draw $y \sim \mathcal{U}[0,1]$
    \item[2.] compute $x=F^{-1}(y)$, so that $x \sim f(x)$.
\end{itemize}

\begin{lstlisting}[caption={Python function to draw a normal distribution from the uniform distribution via Monte Carlo Sampling.}, label={lst:1:3:MCSampling}, language=Python]
import numpy as np
import scipy.special as scs

def inv_cumulative_func(y):
    return np.sqrt(2) * scs.erfinv(2*y-1) 

def draw_x(N: int = 1024):
    y = np.random.rand(N)      # random uniform extraction in [0,1]
    x = inv_cumulative_func(y) # retrieve x, x ~ Standard Normal 
    return x
\end{lstlisting}

\begin{figure}[t]
    \centering
    \includegraphics[width=0.75\linewidth]{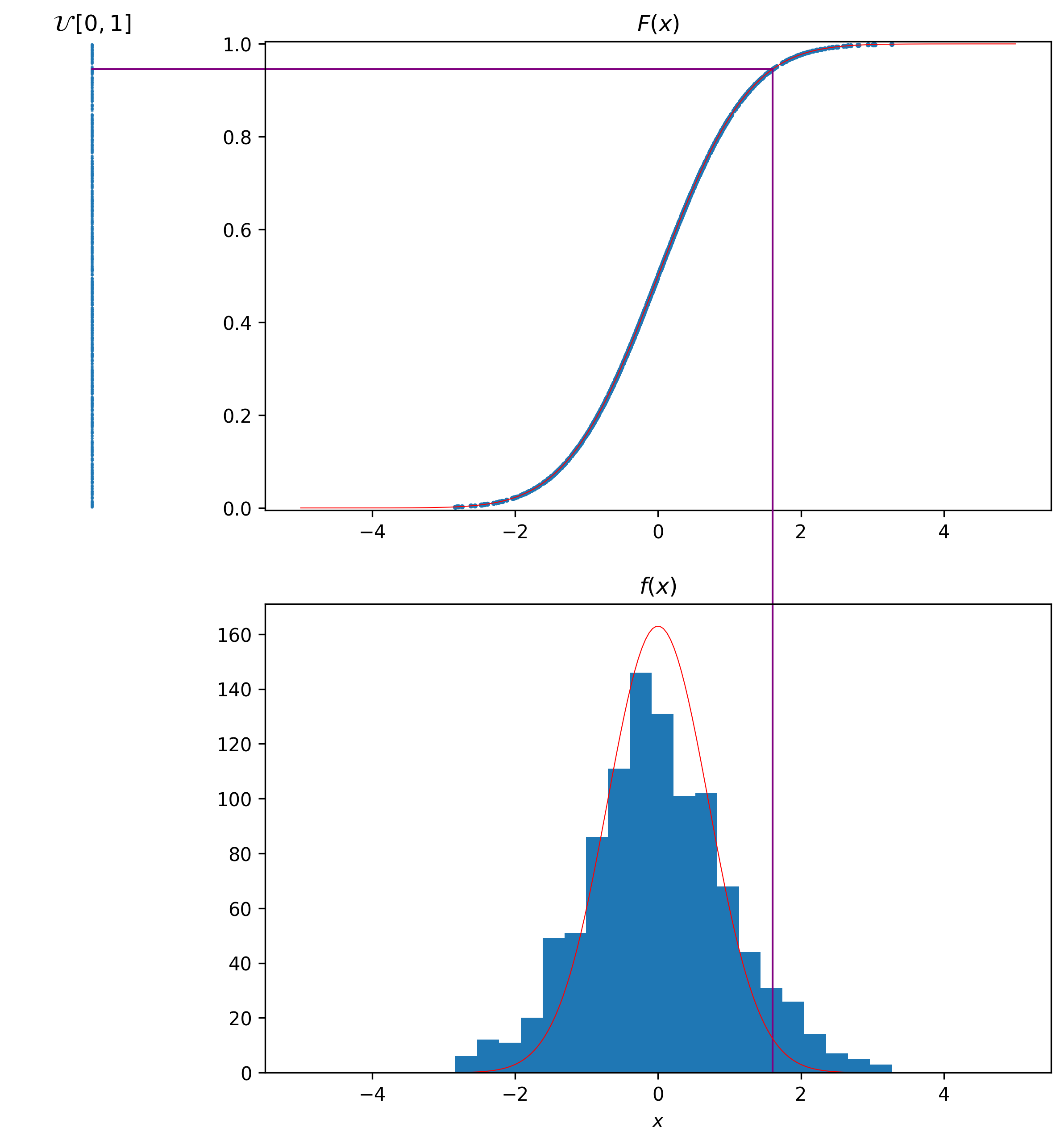}
    \caption{Example of the Monte Carlo Sampling of \cref{lst:1:3:MCSampling} with $N=1024$ draws.}
    \label{fig:1:3:MCSampling}
\end{figure}

In \cref{fig:1:3:MCSampling} we see an example of a normal distribution $\mathcal{N}(0,1)$ obtained using the Monte Carlo Sampling method. 

We can use a similar strategy to compute \textit{integrals}. We know that integrals can be thought of \textit{the area under the curve}; in higher dimensions, for a closed, bounded domain $\Omega$, we can think as the integral over $\Omega$ as the (hyper)volume contained within it. So we can use an accept/reject algorithm based on sampling the hypercube containing it. More formally, suppose we have $f : \Omega \subset \RR^d \to \RR^n$, and a (closed, bounded) domain $D\subseteq \Omega$. Suppose we want to compute
\[
I \defeq \int_D f(x) \dd \mu( x) \,.
\]
We can numerically evaluate $I$ via the so-called \textit{Monte Carlo integration}. Suppose we have a random sampler of the integration domain $D$, with a $\pdf : D \to \RR$, respecting
\[
\int_D \pdf (x) \dd \mu(x) = 1.
\]
We can then sample $\{X_i\}_{i=1, \ldots, N} $ i.i.d. from $D$, and compute
\begin{equation}
    I_N \defeq \frac{1}{N}\sum_{i=1}^N \frac{f(X_i)}{\pdf(X_i)} \,.
\end{equation}
We can see that $I_N$ approximate $I$, as it can be seen as its expectation value:
\begin{equation}
    \begin{split}
        \EE[I_N] &= \EE\left[ \frac{1}{N}\sum_{i=1}^N \frac{f(X_i)}{\pdf(X_i)} \right] \\
        &=  \frac{1}{N}\sum_{i=1}^N \EE\left[    \frac{f(X_i)}{\pdf(X_i)}\right]  \qquad\qquad\qquad \;(\text{linearity of } \EE)\\
        &= \frac{1}{N}\sum_{i=1}^N \int_D  \frac{f(x)}{\pdf(x)}  \pdf(x) \dd \mu(x) \qquad (\text{definition of } \EE) \\
        &= \int_D f(x) \dd \mu(x)  \qquad \qquad \qquad \qquad \quad \; \;  (\text{i.i.d.}) \\
        &= I.
    \end{split}
\end{equation}
So, we have seen that it is an \textit{unbiased} estimator of the integral. 
So we can \textit{numerically} integrate any function defined on a bounded region $D$, in any dimensions. Obviously, the rate of convergence depends on $N$. Indeed, let
\[
Y \coloneqq \frac{f(X)}{\pdf(X)}, \quad X \sim \pdf(x).
\]
Then $I_N = \frac{1}{N}\sum_{i=1}^N Y_i$ with i.i.d.\ copies $Y_i$ of $Y$.
We have $\mathbb{E}[Y] = I$ and
\begin{equation}
\begin{split}
    \mathrm{Var}(I_N)
    &= \mathrm{Var}\left(\frac{1}{N}\sum_{i=1}^N Y_i\right) \\
    &= \frac{1}{N^2}\sum_{i=1}^N \mathrm{Var}(Y_i) \\
    &= \frac{1}{N^2}\cdot N \cdot \mathrm{Var}(Y) \\
    &= \frac{\mathrm{Var}(Y)}{N}.
\end{split}
\end{equation}
Moreover,
\[
\mathrm{Var}(Y)
= \mathbb{E}\left[\left(\frac{f(X)}{\pdf(X)}\right)^2\right] - I^2
= \int_{\Omega} \frac{f(x)^2}{\pdf(x)}\,\dd \mu(x)- I^2.
\]
Therefore the mean squared error is
\[
\mathrm{Var}(I_N) = \mathbb{E}\big[(I_N - I)^2\big]
= \frac{1}{N}\left(\int_{\Omega} \frac{f(x)^2}{\pdf (x)}\,\dd \mu(x) - I^2\right),
\]
which gives a $O(N^{-1/2})$ convergence rate in  root‑mean‑square. 

\subsection{Relevance for Machine Learning}
Now one may ask: \textit{why should I care?} Well, we have seen that functions space may have the concept of \textit{distance}, like (pre-)Hilbert spaces, in particular $L^p(\Omega)$ spaces.

Let $(X,Y)$ be a pair of random variables with values in $\mathscr{X} \subseteq \mathbb{R}^m$ and $\mathscr{Y} \subseteq \mathbb{R}^n$, respectively, defined on some probability space, and let $P_X$ denote the distribution of $X$ on $\mathscr{X}$. We assume that there exists a (possibly unknown) target function
\[
f : \mathscr{X} \to \mathscr{Y}
\]
such that $f$ belongs to the Banach space $L^p(\mathscr{X}, P_X)$, equipped
with the norm
\[
\|F\|_{L^p(P_X)} \coloneqq
\left( \int_{\mathscr{X}} \|F(x)\|^p \, \dd P_X(x) \right)^{1/p},
\]
for $1 \le p < \infty$. We consider a parametric family of functions
\[
f_\theta : \mathscr{X} \to \mathscr{Y}, \quad \theta \in \Theta,
\]
for instance a deep neural network (DNN). The (population) $L^p$ distance between
$f_\theta$ and $f$ is given by
\[
\|f_\theta - f\|_{L^p(P_X)}^p
=
\int_{\mathscr{X}} \|f_\theta(x) - f(x)\|^p \, \dd P_X(x).
\]
In practice, the distribution $P_X$ and the function $f$ are unknown. 
Instead, we observe a dataset
\[
\{(X_i, Y_i)\}_{i=1, \ldots, N},
\]
where $(X_i, Y_i)$ are i.i.d.\ copies of $(X,Y)$. The integral above can be approximated by Monte Carlo integration using the samples $X_1,\dots,X_N$:
\[
    \int_{\mathscr{X}} \|f_\theta(x) - f(x)\|^p \, \dd P_X(x)
    \approx
    \frac{1}{N} \sum_{i=1}^N \|f_\theta(X_i) - f(X_i)\|^p.
\]

Since $f(X_i)$ is not observed, we replace it by the corresponding label $Y_i$.
This yields the empirical loss
\[
    R_N(f_\theta)
    \defeq
    \frac{1}{N} \sum_{i=1}^N \|f_\theta(X_i) - Y_i\|^p,
\]
which is precisely the \textit{empirical risk} associated with the loss $\ell(y',y) = \|y'-y\|^p$. This is the standard formulation in statistical learning theory, where the goal is to minimize the expected risk $\mathbb{E}[\ell(f_\theta(X), Y)]$ using its empirical approximation.

In the special case $p=2$, we obtain the mean squared error (MSE)
\[
R_N(f_\theta)
=
\frac{1}{N} \sum_{i=1}^N \|f_\theta(X_i) - Y_i\|^2,
\]
whose square root is the empirical $L^2$-norm (root mean squared error, RMSE)
of the difference between $f_\theta$ and the observed outputs. In a single formula, 
\[
\underbrace{\|f_\theta - f\|_{L^p(P_X)}^p}_{\text{Population Risk}} = \underbrace{\int_{\mathscr{X}} \|f_\theta(x) - f(x)\|^p \, \mathrm{d}P_X(x)}_{\text{Integral}} \approx \underbrace{\frac{1}{N}\sum_{i=1}^N \|f_\theta(X_i) - Y_i\|^p}_{\text{Empirical Risk (MCI)}} \,.
\]
Thus, we can see the entire \textit{learning process} as the iterative algorithm to minimise the distance between the approximator DNN and the real function, evaluating such distance using the Monte Carlo approach, which yields the minimisation of the \textit{empirical risk} \cite{scardapane2024alice}.

    \chapter{Lecture 2: An introduction to numerical resolution of differential equations}\label{chap:2}

For a broader introduction to the subject, see \cite{quarteroni2020numerical, Landau2015, iserles2009first, Suli2021, WierstrassInst2025}, and references therein. 

In this lecture, we will briefly introduce a few possible ways to \textit{numerically} solve a generic Partial Differential Cauchy problem, in the general form
\begin{equation}
    \begin{split}
        \mathcal{F}[u(x), x, \gamma] &= J(x) \,, \qquad x\in \Omega \subseteq \RR^d \,, \\
        \mathcal{B}[u(x), x, \gamma] &= g(x) \,, \qquad x\in \partial\Omega .
    \end{split}
\end{equation}

Overall, the main goal of numerical methods is to find a way to recast the \textit{differential} problem into an \textit{algebraic} problem, which can be easily solved with standard numerical methods. In essence, the different methods we will present, mainly differ in \textit{how} they recast the differential problem into the algebraic problem:
\begin{itemize}
    \item[1.] Finite Difference Methods (FDMs) recast it by approximating the function on the nodes of a regular grid, and replacing derivatives with incremental ratios of the function on neighbouring nodes;
    \item[2.] Finite Element Methods (FEMs) recast it by approximating the unknown function on the interior regions of an irregular mesh via a (multidimensional) linear approximation, and imposing stitching conditions on edges and nodes of the mesh, in order to solve the \textit{weak formulation} of the PDE;
    \item[3.] Finite Volume Methods (FVMs) recast it by approximating the unknown function on the interior regions of an irregular mesh via a (multidimensional) linear approximation, and imposing \textit{flux} conditions on edges, in order to satisfy continuity. 
\end{itemize}

We will close this Lecture with a completely different approach, which is a \textit{meshless} approach, called the \textbf{Kansa Method} \cite{KANSA1990147}, which instead uses a \textit{collocation} method to use an approximator to satisfy the differential equations, recasting the differential problem into an \textit{optimisation} problem.

\section{Finite Difference Methods (FDM)} \label{sec:2:1:FDM}

The core of FDM is to approximate the derivatives of unknown fields (multivariate functions) in terms of the value of the functions in a neighbourhood of the evaluation point. We have already encountered the Taylor expansion of a function
\[
\begin{split}
   u(x+\Delta x) &= u(x) + \sum_{n=1}^\infty \frac{(\Delta x)^n}{n!} \left. \frac{\partial^n u}{\partial  x^n} \right|_x \\
   &= u(x) + \Delta x \partial_x u(x) + \mathcal{O}(\Delta x) .
\end{split}
\]
This means that, up to an error $\mathcal{O}((\Delta x)^2)$, we can express the derivative as
\begin{equation}
    \partial_x u(x) = \frac{u(x+\Delta x) - u(x)}{\Delta x}  + \mathcal{O}(\Delta x)  \,, \qquad (\text{Forward Difference})
\end{equation}
which is the so-called \textit{Forward Difference Formula}. But we can also compute
\[
\begin{split}
   u(x - \Delta x) &= u(x) + \sum_{n=1}^\infty \frac{(-1)^n \Delta x^n}{n!} \left. \frac{\partial^n u}{\partial  x^n} \right|_x \\
   &= u(x) - \Delta x \partial_x u(x) + \mathcal{O}(\Delta x) .
\end{split}
\]
so that
\begin{equation}
    \partial_x u(x) = \frac{ u(x) - u(x-\Delta x)}{\Delta x}  + \mathcal{O}(\Delta x)  \,, \qquad (\text{Backward Difference})
\end{equation}
which is the so-called \textit{Backward Difference Formula}. Now, subtracting the two, we get a more precise approximation
\begin{equation}
    \begin{split}
    &-
    \begin{cases}
        u(x+\Delta x) = u(x) + \Delta x \partial_x u(x) + \mathcal{O}(\Delta x) \\
        u(x-\Delta x) = u(x) - \Delta x \partial_x u(x) + \mathcal{O}(\Delta x) 
    \end{cases} \\
    &\Rightarrow \partial_x u(x) = \frac{ u(x + \Delta x) - u(x-\Delta x)}{2 \Delta x}  + \mathcal{O}((\Delta x)^2)  \,, \qquad (\text{Central Difference})
\end{split}
\end{equation}
Notice that the \textit{Central Difference} has an higher precision on $ \mathcal{O}((\Delta x)^2)$; to notice this explicitly, we simply expand to a higher order
\begin{equation}
    \begin{split}
    &-
    \begin{cases}
        u(x+\Delta x) = u(x) + \Delta x \partial_x u(x) + \frac{\Delta x^2}{2} \, \partial_x^2 u(x)  + \mathcal{O}((\Delta x)^2) \\
        u(x-\Delta x) = u(x) - \Delta x \partial_x u(x)  + \frac{\Delta x^2}{2} \, \partial_x^2 u(x)  + \mathcal{O}((\Delta x)^2) 
    \end{cases} \\
    &\Rightarrow \partial_x u(x) = \frac{ u(x + \Delta x) - u(x-\Delta x)}{2\Delta x}  + \mathcal{O}((\Delta x)^2)  \,, \qquad (\text{Central Difference})
\end{split}
\end{equation}

Higher-order derivatives can be computed by systematically combining Taylor expansions at multiple points. \cite{Fornberg1988GenerationOF} provides an algorithm for computing the optimal weights for arbitrary stencil configurations, e.g.
\begin{align}
    \partial_x^2 u(x) &= \frac{u(x+2\Delta x) + u(x) -2u(x+\Delta x)}{\Delta x^2} + \mathcal{O}(\Delta x) , \qquad \text{(Forward Difference)} \\
    \partial_x^2 u(x) &= \frac{u(x-2\Delta x) + u(x) -2u(x-\Delta x)}{\Delta x^2} + \mathcal{O}(\Delta x) , \qquad \text{(Backward Difference)} \\
    \partial_x^2 u(x) &= \frac{u(x + \Delta x) -2 u(x) +u(x - \Delta x)}{\Delta x^2} + \mathcal{O}((\Delta x)^2) , \qquad \; \, \text{(Central Difference)} 
\end{align}

It is possible to compute higher-order derivatives, and also achieve higher precision at fixed derivative order, by enlarging the stencil of neighbouring points from $\{-1, 0, +1\}\cdot \Delta x$ to $\{-q, \ldots, 0, \ldots, +q\}\cdot \Delta x$.

\begin{figure}[t]
    \centering
    \includegraphics[width=0.75\linewidth]{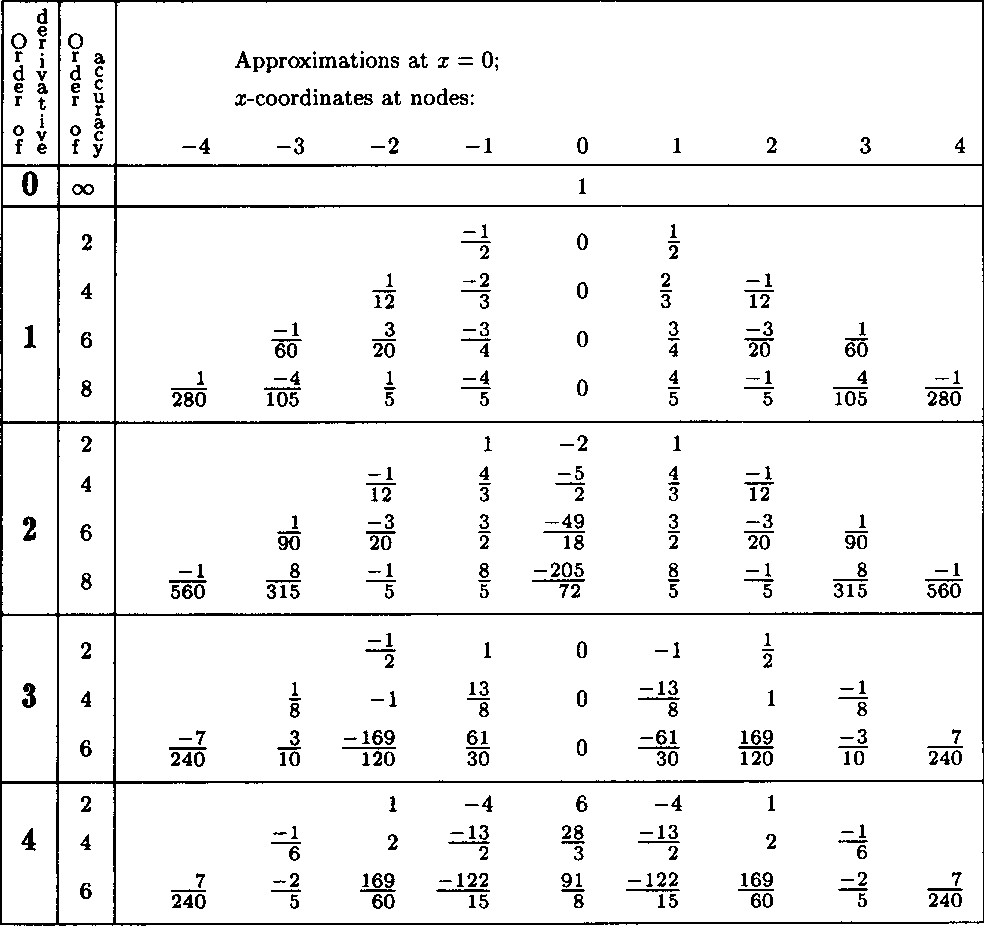}
    \caption{Higher order and Higher precision derivatives. From \cite{Fornberg1988GenerationOF}.}
    \label{fig:2:1:table_fdm_1}
\end{figure}

Now, let us pause for a moment. If we define $x_n \defeq n \Delta x, u_n \defeq u(x_n)$, we can see that we are expressing the derivative $\partial_x u_i \defeq \left. \partial_x u(x) \right|_{x=x_i}$ as a \textit{convolution}
\begin{equation}
    \pd_x u_i = k \star u \defeq \sum_{j=-q}^q k_j u_{i+j}  
\end{equation}
where $k$ is the \textit{discretised} differential operator, expressed as a \textit{linear kernel}. This is completely analogous to what a Convolutional Neural Network (CNN) is, except that CNNs have \textit{learnable} kernel parameters. Similarly to CNNs, enlarging the receptive field allows capturing longer-range dependencies, but increases computational cost and may lose fine-grained local features.

\subsection{A first application of finite differences: Sobel Edge detection algorithm}

In computer vision tasks, \textit{edge detection} has always been a relevant task. An algorithmic way to find the borders is the \textit{Sobel filter}.

\begin{algorithm}[H]
\caption{Sobel Edge Detection Algorithm}\label{alg:2:1:Sobel}
\small
\begin{algorithmic}[1]
\Procedure{Sobel}{image $I$, threshold $\theta$}
\State $G_x= \left[\begin{array}{c}
     1 \\ 2 \\ 1
\end{array}\right] \otimes [-1, 0, +1]$ 
\State $G_y= \left[\begin{array}{c}
     -1 \\ 0 \\ +1
\end{array}\right] \otimes [1, 2, 1]$ 
\For{$i\in [0, H]$} \Comment{Iterate over $x$ pixels}
    \For{$j\in [0, W]$} \Comment{Iterate over $y$ pixels}
        \State $X_{ij} = G_x \star I_{ij}$ \Comment{Apply $x-$derivative, $y-$smoothing}
        \State $Y_{ij} \,\, = G_y \star I_{ij}$  \Comment{Apply $x-$smoothing, $y-$derivative}
        \State $O_{ij} = \sqrt{X_{ij}^2 +Y_{ij}^2}$ \Comment{Compute edge magnitude}
        \If{$O_{ij}<\theta$} 
            \State $O_{ij} = 0$ \Comment{Threshold low values}
        \EndIf
    \EndFor
\EndFor
\State \textbf{return} $O$
\EndProcedure
\end{algorithmic}
\end{algorithm}
The $[-1, 0, +1]$ piece is the first order derivative in its finite difference approximation. The $[1,2,1]$ instead is a discrete gaussian smoothing kernel.

\begin{figure}[t]
    \centering
    \includegraphics[width=0.95\linewidth]{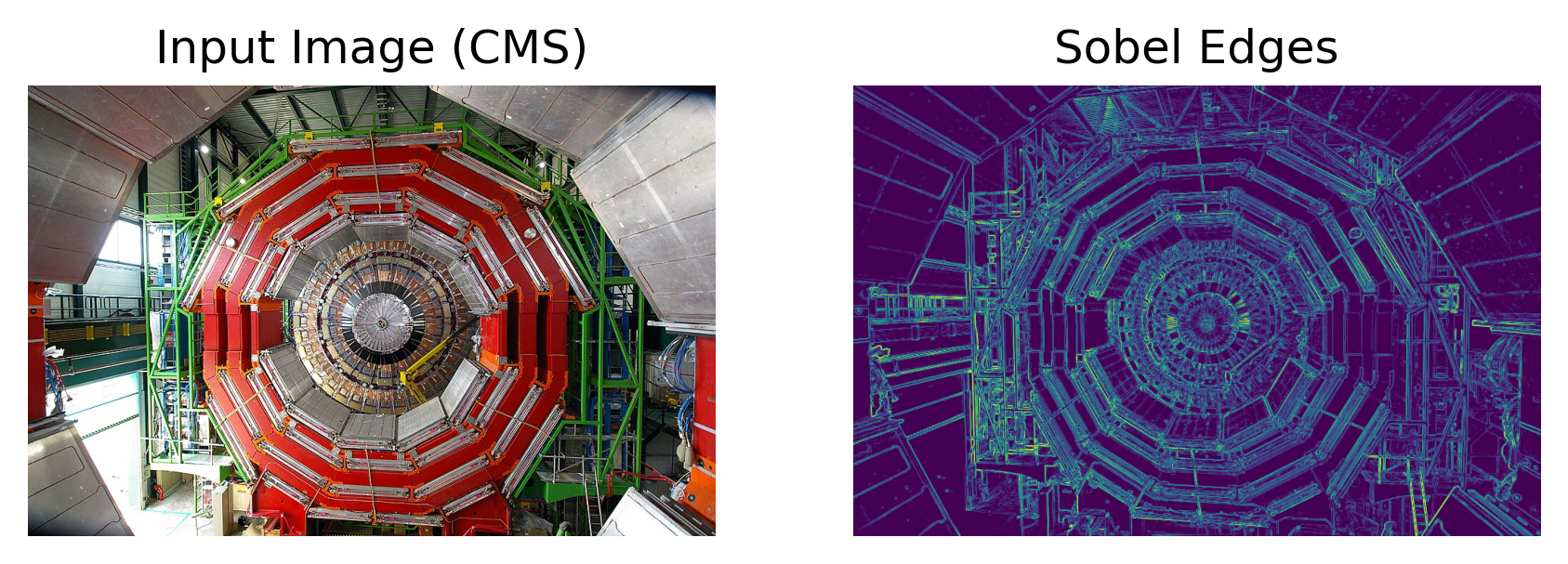}
    \caption{Example of the usage of Sobel Edge detection algorithm}
    \label{fig:2:1:sobel_edge_detection_img}
\end{figure}

\begin{lstlisting}[caption={Sobel Edge Detection Algorithm in Python}, label={lst:2:1:Sobel}, language=Python]
import numpy as np

def sobel_edges(gray: np.ndarray, theta: float, mode: str = "reflect", return_gradients: bool = False):
    """
    Compute Sobel edge magnitude for a 2D grayscale image using pure NumPy.

    Parameters
    ----------
    gray : np.ndarray
        2D array (H, W), grayscale image.
    mode : str
        Padding mode for np.pad (e.g. "reflect", "edge", "constant").
    return_gradients : bool
        If True, also return (Gx, Gy).

    Returns
    -------
    mag : np.ndarray
        Edge magnitude image, same shape as input.
    (optional) Gx, Gy : np.ndarray
        Horizontal and vertical gradients.
    """
    if gray.ndim != 2:
        raise ValueError("Input must be a 2D grayscale array")
    if gray.max() > 1:
        gray /= 255.
        theta/= 255.
    # Work in float for accuracy
    img = gray.astype(np.float32, copy=False)
    # Pad by 1 pixel on all sides
    p = np.pad(img, 1, mode=mode)
    # Horizontal gradient Gx
    Gx = (
        (p[1:-1, 2:] - p[1:-1, :-2]) +
        2 * (p[2:, 2:] - p[2:, :-2]) +
        (p[:-2, 2:] - p[:-2, :-2])
    )
    # Vertical gradient Gy
    Gy = (
        (p[2:, 1:-1] - p[:-2, 1:-1]) +
        2 * (p[2:, 2:] - p[:-2, 2:]) +
        (p[2:, :-2] - p[:-2, :-2])
    )
    # Edge magnitude
    mag = np.hypot(Gx, Gy)
    # threshold
    mag = mag * (mag > theta)
    if return_gradients:
        return mag, Gx, Gy
    return mag
\end{lstlisting}

\subsection{Using FDM to solve Poisson Equation: the Jacobi method}

Suppose we need to solve a Poisson equation in three dimension
\begin{equation}
    \begin{split}
        \Delta V(x,y,z) &= f(x,y,z) \,, \quad (x,y,z) \in  \Omega  \defeq [0,1]^3 \,, \\
        V(x,y,z) &= g(x,y,z)\,, \quad (x,y,z) \in \partial \Omega_D \,, \\
        \hat{n} \cdot \nabla V(x,y,z) &= h(x,y,z)\,, \quad (x,y,z) \in \partial \Omega_N \,, \\
    \end{split}
\end{equation}
with mixed Dirichlet/Neumann boundary conditions. As we have seen, the Finite Difference Method relies on the discretisation of the space, as $V(x_i, y_j, z_k) = V_{ijk}$, where
\begin{align}
 x_i = i  \Delta x, y_j = j  \Delta y, z_k = k  \Delta z,  \quad(i,j,k) = (1,1,1), \ldots,(N_x, N_y, N_z) \in \mathbb{N}^3 \,,
\end{align}
where $N_x \Delta x = N_y \Delta y = N_z \Delta z =  1$\note{The generalisation to less general interval is trivial, and left as an exercise to the reader.}.

Now, this implies that we can approximate the Laplace operator as\note{Again, the generalisation to different spatial dimension is trivial, as is trivial the generalisation to non-constant intervals $\Delta x\to \Delta x_i$.}
\begin{equation}\label{eq:2:1:lapl_fdm}
    \begin{split}
    \Delta V(\vec x) \to \Delta V_{ijk} \simeq \; & \,\frac{V_{(i+1),j,k} - 2V_{i,j,k} + V_{(i-1),j,k}}{(\Delta x)^2 } \\
    & + \frac{V_{i(j+1)k} - 2V_{ijk} + V_{i(j-1)k}}{(\Delta y)^2} \\
    & + \frac{V_{ij(k+1)} - 2V_{ijk} + V_{ij(k-1)}}{(\Delta z)^2} \,.
    \end{split}
\end{equation}
The three lines in \cref{eq:2:1:lapl_fdm} are, respectively, the three central difference of the operators $\partial_x^2, \pd_y^2, \pd_z^2$. 

This finite difference approximation of the Laplace operator allows us to recast the PDE as an algebraic relation
\begin{equation}
    \begin{split}
        f_{ijk} = \; & \,\frac{V_{(i+1),j,k} - 2V_{i,j,k} + V_{(i-1),j,k}}{(\Delta x)^2 } \\
                & + \frac{V_{i,(j+1),k} - 2V_{i,j,k} + V_{i,(j-1),k}}{(\Delta y)^2} \\
                & + \frac{V_{i,j,(k+1)} - 2V_{i,j,k} + V_{i,j,(k-1)}}{(\Delta z)^2} \,,
    \end{split}
\end{equation}
where $f(x_i, y_j, z_k) = f_{i,j,k}$. 

Now, we can solve this algebraic relation via a trick, that is known as the \textit{Jacobi Method}, due to its relation to the Jacobi Method to solve linear systems \cite{iserles2009first}. The intuition behind the Jacobi method (mutated from the approach in solving linear systems) is to update iteratively the values of the field by bringing on the LHS the $V_{ijk}$ terms, while the other terms are kept or put in the RHS, so that the field at the step $n+1$ is determined in terms of the field at steps $n$, as
\begin{equation}\label{eq:2:1:jacobi_poisson}
    \begin{split}
        V_{i,j,k}^{n+1} = \frac{\Delta^2}{2}  \Big(  \; & \frac{V_{i+1,j,k}^n + V_{i-1,j,k}^n}{ (\Delta x)^2 }  \\
        & + \frac{V_{i,j+1,k}^n + V_{i,j-1,k}^n}{ (\Delta y)^2 } \\
        & + \frac{V_{i,j,k+1}^n + V_{i,j,k-1}^n}{ (\Delta z)^2 } - f_{ijk} \Big) \defeq U^n_{ijk} ,
    \end{split}
\end{equation}
where we have defined the RHS as the update operator $ U^n_{ijk}$, and where we have defined
\begin{align}
    \frac{1}{\Delta^2} &\defeq \left( \frac{1}{(\Delta x)^2} + \frac{1}{(\Delta y)^2} +\frac{1}{(\Delta z)^2}  \right) \notag\\
    &= \frac{(\Delta x)^2 (\Delta y)^2 + (\Delta x)^2 (\Delta z)^2 + (\Delta y)^2 (\Delta z)^2}{(\Delta x)^2 (\Delta y)^2 (\Delta z)^2 } \,.
\end{align}

\noindent We thus have the following algorithm to solve the Poisson Cauchy problem:
\begin{algorithm}[H]
\caption{Jacobi Method for solving Poisson Problem}\label{alg:2:1:Jacobi}
\small
\begin{algorithmic}[1]
\Procedure{Jacobi}{grid $(N_z, N_y,N_z)$, discretisation $(\Delta x, \Delta y, \Delta z)$, threshold $\theta$}
\State init $V_{ijk}$ \Comment{Initialise $V_{ijk}$}
\State $\epsilon = \infty$ \Comment{Initialise update Error}
\While{$\epsilon > \theta$}
    \State compute update $U_{ijk}$
    \State $\epsilon = \| V_{ijk} - U_{ijk}\|_{\infty} $ \Comment{Update error}
    \State $V_{ijk} \leftarrow U_{ijk}$ \Comment{Update field}
    \State $\text{impose\_bc}(V_{ijk})$ \Comment{Impose BC}
\EndWhile
\State \textbf{return} $V$
\EndProcedure
\end{algorithmic}
\end{algorithm}

\subsubsection{Jacobi Method as a Relaxation of a Diffusion Equation}

Notice that, starting from the Jacobi method equation \cref{eq:2:1:jacobi_poisson}, we can add zero, written up as
\begin{equation}
    \begin{split}
        0 &= V^n_{ijk} - V^n_{ijk} \\
          &= V^n_{ijk} - \frac{\Delta^2}{2} \left( \frac{ 2 V_{ijk}^n}{ \Delta x^2 }  + \frac{ 2 V_{ijk}^n}{ \Delta y^2 }  + \frac{ 2 V_{ijk}^n}{ \Delta z^2 } \right),
    \end{split}
\end{equation}
and thus we get
\begin{equation}
    \begin{split}
        V_{ijk}^{n+1} - V^n_{ijk} = \frac{\Delta^2}{2}  \Big(  \; & \frac{V_{i+1,j,k}^n - 2 V_{ijk}^n + V_{i-1,j,k}^n}{ \Delta x^2 }  \\
        & + \frac{V_{i,j+1,k}^n - 2 V_{ijk}^n + V_{i,j-1,k}^n}{ \Delta y^2 } \\
        & + \frac{V_{i,j,k+1}^n - 2 V_{ijk}^n + V_{i,j,k-1}^n}{ \Delta z^2 } - f_{ijk} \Big) ,
    \end{split}
\end{equation}
Which is the Forward Euler approximation of the Diffusion equation with a constant source term
\begin{align}
\frac{\partial V(t, \vec x)}{\partial t} = \Delta V(t, \vec x) - f(\vec x) \,,
\end{align}
with 
$$\Delta t = \frac{\Delta^2}{2} \,$$
where the initialisation of the field is, here, interpreted as the initial condition.

Thus, whenever the Jacobi iteration is convergent, it can be interpreted as an explicit pseudo-time evolution approaching the steady state of the corresponding diffusion equation.

\subsection{An application in Computer Vision: image inpainting with Laplace Equation}

Image inpainting aims at reconstructing image values in regions where data are missing or have deliberately been removed. Since the lost information is not uniquely determined by the surviving pixels, the reconstruction is inherently ill-posed and necessarily relies on assumptions about how image information should propagate into the unknown region. PDE-based approaches provide a classical and mathematically well-studied family of inpainting methods \cite{Bertalmio200, Guillemot2014, hoeltgen2019theoretical}. A simple example is homogeneous diffusion: the missing values are reconstructed as a harmonic extension of the known image data. If $\Omega_K\subset\Omega$ denotes the set where the image is known, we can write
\begin{equation}\label{eq:2:1:inpainting_laplace}
    \begin{split}
        - \Delta u(x,y) &= 0 \,, \qquad \quad \, \, (x,y) \in \Omega\setminus\Omega_K \,, \\
        u(x,y) &= f(x,y)\,, \quad (x,y) \in \partial\Omega_K \,, \\
        \hat{n} \cdot \nabla u &= 0\,, \qquad \quad \, \,(x,y) \in \partial\Omega\setminus \pd\Omega_K \,,
    \end{split}
\end{equation}

\begin{figure}[t]
    \centering
    \includegraphics[width=0.55\linewidth]{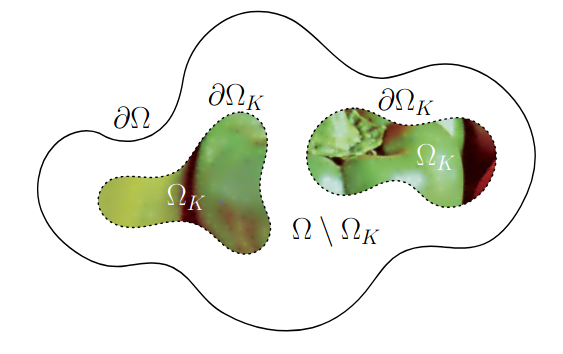}
    \caption{Schematic representation of the homogeneous-diffusion inpainting setting in \cref{eq:2:1:inpainting_laplace}: image data are known on $\Omega_K$, while the missing region $\Omega\setminus\Omega_K$ is reconstructed through harmonic extension. From \cite{hoeltgen2019theoretical}.}
    \label{fig:2:1:generic_inpainting_problem}
\end{figure}

\noindent  Here, $f$ represents known image data in a region $\Omega_K \subset \Omega$ (respectively on the boundary
$\pd \Omega_K$) of the whole image domain $\Omega$. A graphical sketch of the problem is given in \cref{fig:2:1:generic_inpainting_problem}. 

Another approach, equivalent to the former up to certain smoothness requirement, is to use the indicator function $c(x,y)$ of the optimal set $\Omega_K$ (i.e. $c(x,y) = 1$ if $(x,y) \in \Omega_K$, and $c(x,y)=0$ otherwise) and define the pure Laplace-Neumann problem on $\Omega$
\begin{align}
    c(x,y) [u(x,y) - f(x,y) ] - [1-c(x,y)] \nabla^2 u(x,y) &= 0, \quad \,(x,y) \in \Omega \notag \\
    \nabla_n u &= 0 \,,\quad (x,y) \in  \partial \Omega \setminus \Omega_K .
\end{align}

\begin{figure}[ht]
    \centering
    \begin{subfigure}[t]{0.32\textwidth}
        \centering
        \includegraphics[width=\textwidth]{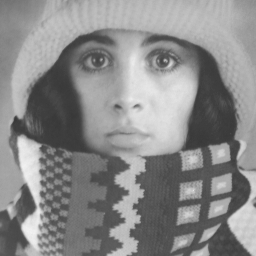}
        \caption{Original Image}
    \end{subfigure}
    \hfill
    \begin{subfigure}[t]{0.32\textwidth}
        \centering
        \includegraphics[width=\textwidth]{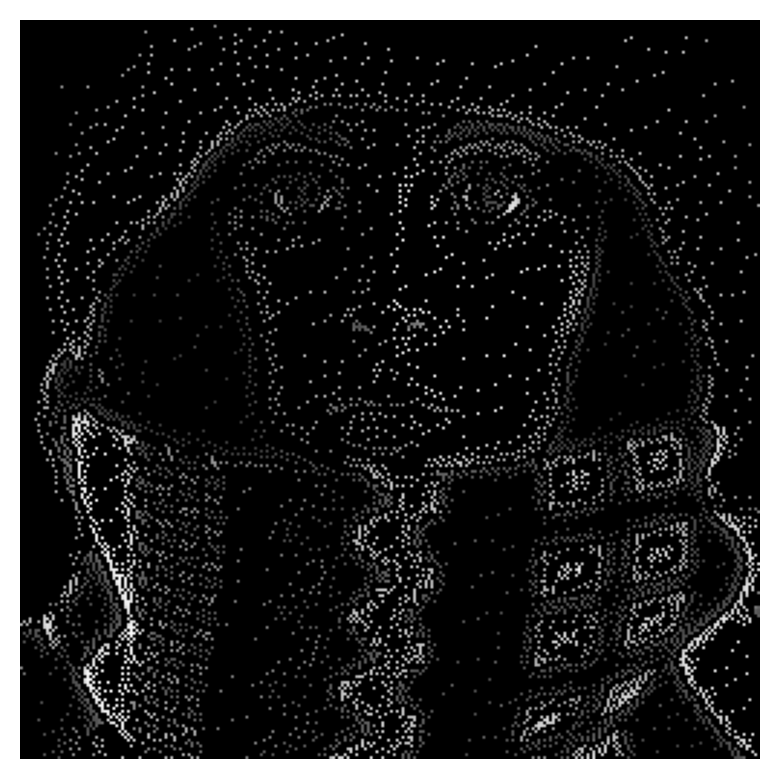}
        \caption{Optimal masking. It has $\sim 90\%$ of missing data.}
    \end{subfigure}
    \hfill
    \begin{subfigure}[t]{0.32\textwidth}
        \centering
        \includegraphics[width=\textwidth]{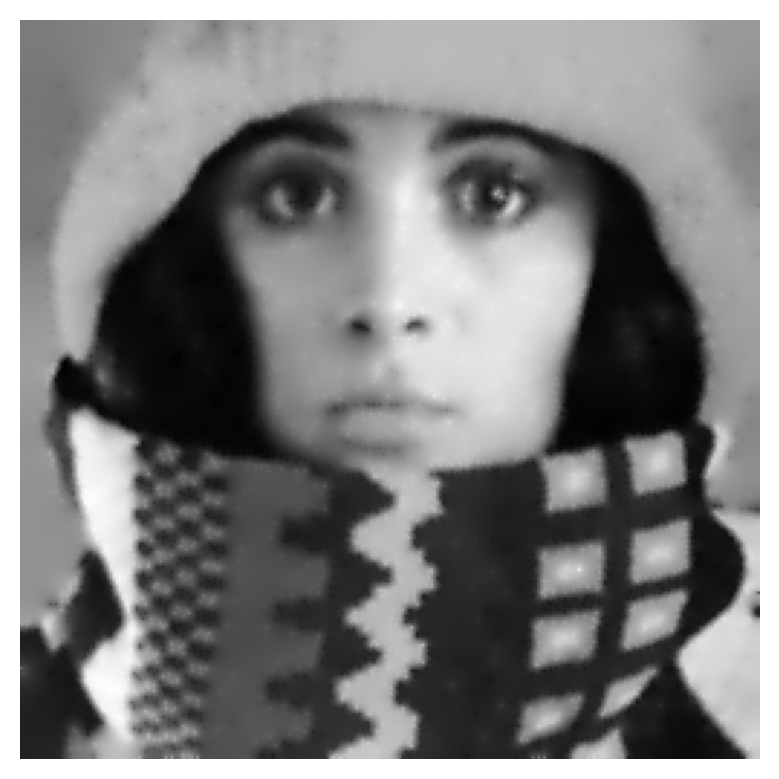}
        \caption{Inpainted solution using solving the Laplace problem \cref{eq:2:1:inpainting_laplace}.}
    \end{subfigure}
    \label{fig:2:1:inpainting_trui}
\end{figure}

A relevant result of this approach is that it is possible to have a high compression of the data. In \cref{fig:2:1:inpainting_trui}, an optimal compression, containing less than $10\%$ of pixels of the original image, is inpainted using the 2D Laplace pde. A python sketch of the algorithm is reported in \cref{lst:2:1:Laplace_Inpainting}. 

\begin{lstlisting}[caption={Laplace inpainting}, label={lst:2:1:Laplace_Inpainting}, language=Python]
import numpy as np
import tqdm 

def inpaint_with_laplace(real: np.array, bnd: np.array, N_iter: int):
    # mask: True = unknown pixel
    unknown = (bnd == 0)
    # initialize solution
    V = bnd.copy()    
    # Jacobi iteration
    for _ in tqdm.tqdm(range(N_iter)):
        V_new = V.copy()
        # Laplace update only on unknown pixels
        laplace =  (
            V[2:, 1:-1] + V[:-2, 1:-1] +
            V[1:-1, 2:] + V[1:-1, :-2]
        ) / 4.
        # update unknown pixels
        V_new[1:-1, 1:-1][unknown[1:-1, 1:-1]] = laplace[unknown[1:-1, 1:-1]]
        # enforce boundary conditions
        # neumann
        V_new[ 0, :] = V_new[ 1, :]
        V_new[-1, :] = V_new[-2, :]
        V_new[ :, 0] = V_new[ :, 1]
        V_new[ :,-1] = V_new[ :,-2]
        # data
        V_new[~unknown] = bnd[~unknown]
        # enforce V <= 1
        V_new[V_new>=1.]=1.
        # update V
        V = V_new
    return V
\end{lstlisting}

\subsection{Solving Burgers equation with FDM}
A simple example may be the tackling of the Burgers equation; a very nice set-up is the following Cauchy problem
\begin{equation}\label{eq:2:1:Burgers_FDM}
    \begin{split}
            \partial_t u(t,x) + u(t,x) \partial_x u(t,x) - \nu \partial_x^2 u(t,x) &= 0 \,,   \qquad \quad \quad (t,x) \in [0, 1]\times [-1, +1], \\
            u(t=0, x) &= - \sin (\pi x)\,,   \;  x \in [-1, +1], \\
            u(t, x=\pm1)&= 0\, ,   \qquad \qquad  t\in[0,1],
    \end{split}
\end{equation}
Where $\nu=0.01/\pi$. A very minimal, pure python (plus numpy) implementation is\note{As for any explicit time-stepping scheme, the choice of \(\Delta t\) is constrained by numerical stability conditions depending on the advective and diffusive terms; the parameters above are chosen to provide a stable illustrative computation.} 

\begin{lstlisting}[caption={FDM for Burgers problem \cref{eq:2:1:Burgers_FDM}}, label={lst:2:1:Burgers_FDM}, language=Python]
import numpy as np

# Parameters
nu = 0.01 / np.pi          # viscosity
Nx = 201                   # number of spatial points
Nt = 5000                  # number of time steps
xmin, xmax = -1.0, 1.0
tmax = 1.0
# Grid
x = np.linspace(xmin, xmax, Nx)
dx = x[1] - x[0]
dt = tmax / Nt

# Initial condition: u(0,x) = -sin(pi x)
u = np.zeros([Nt, Nx])
u[0, :] = - np.sin(np.pi * x)
# Enforce boundary conditions at t = 0
u[:,  0] = 0.0
u[:, -1] = 0.0
# Time stepping: forward Euler in time, central differences in space
for n in tqdm.tqdm( range(1, Nt) ):
    un = u[n-1,:].copy()
    # First derivative u_x (central difference)
    ux = (un[2:] - un[:-2]) / (2.0 * dx)
    # Second derivative u_xx (central difference)
    uxx = (un[2:] - 2.0 * un[1:-1] + un[:-2]) / dx**2
    # Interior update: u_t = -u u_x + nu u_xx
    u[n, 1:-1] = un[1:-1] + dt * (-un[1:-1] * ux + nu * uxx)
    # Dirichlet boundary conditions u(t, \\pm 1) = 0
    u[:, 0 ] = 0.0
    u[:, -1] = 0.0
\end{lstlisting}

\begin{figure}[t]
    \centering
    \includegraphics[width=0.75\linewidth]{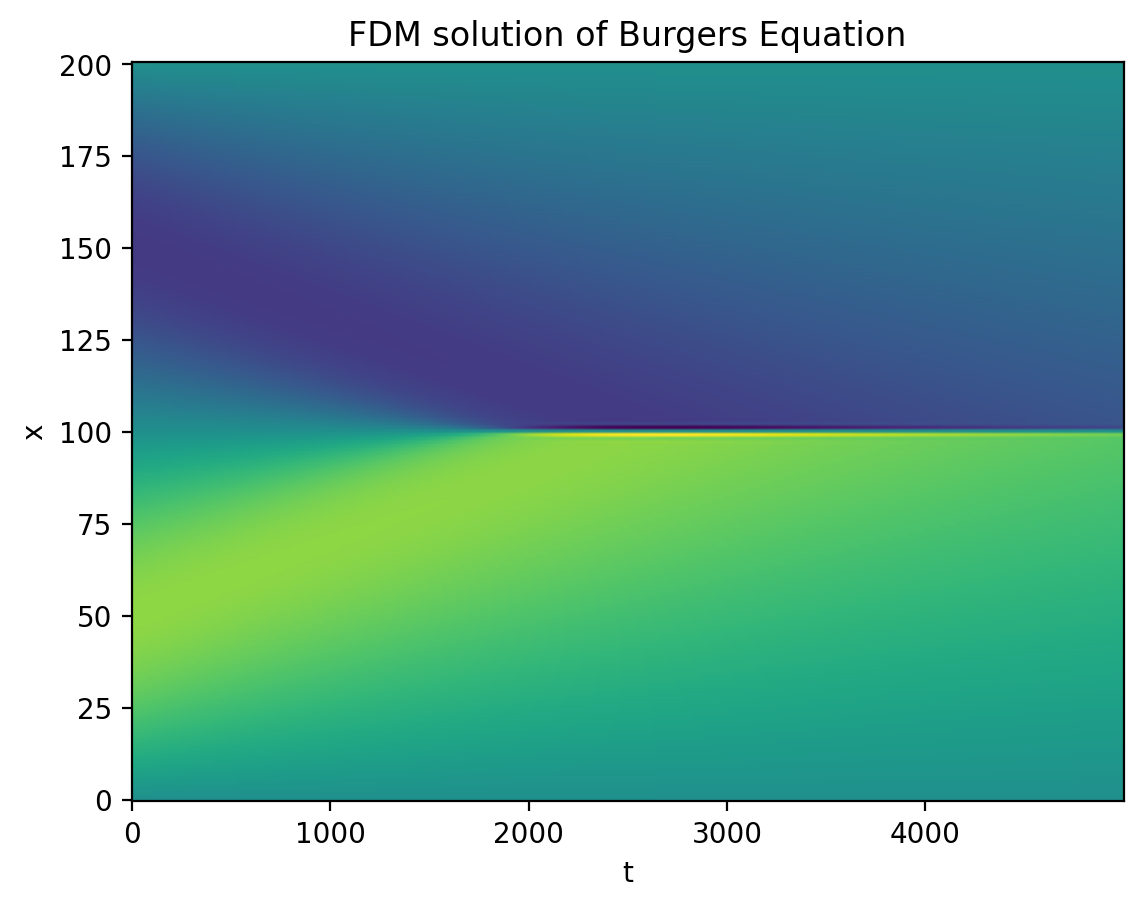}
    \caption{FDM results of the simple code \cref{lst:2:1:Burgers_FDM}.}
    \label{fig:2:1:FDM_burgers}
\end{figure}

\noindent The code reported in \cref{lst:2:1:Burgers_FDM} produces a solution which can be plotted in a 2D plot, as shown in \cref{fig:2:1:FDM_burgers}. 

We see why the Burgers equation is an interesting example: for this small viscosity, the nonlinear advection steepens the solution into a very sharp, shock-like internal layer. A genuine finite-time shock discontinuity occurs in the inviscid limit $\nu=0$; for $\nu>0$, diffusion smooths the solution.

\subsection{A heuristic connection to CNN and Graphs: Stencil representation \& Heat equation on Graphs}

We have seen in \cref{subsubsec:1:2:1:1:DiffEqDer} that we can derive the diffusion equation by balancing heat flux in spatially discretised cells, obtaining a relation between the rate change in time with the Laplacian in space (i.e., the rate of the rate of change in space).  We have seen here that we can stencil-approximate the Laplacian as
\[
    (\Delta u)_{i} = \frac{u_{i+1} - 2u_i + u_{i-1}}{h^2} \,,
\]
using central difference finite differences. We can go on, and generalise it to higher differences 
\[
    (\Delta u)_{ij} = \frac{u_{i+1;j} - 2u_{ij} + u_{i-1;j}}{h_x^2} + \frac{u_{i;j+1} - 2u_{ij} + u_{i;j-1}}{h_y^2} \,,
\]
and so on. We can represent \textit{graphically} this \textit{stencil} operators \cite{iserles2009first}, e.g.\note{Here we reported the case $h_x=h_y=h_z=h.$}
\begin{equation}
  (\Delta u)_{i} \; \defeq \; \frac{1}{h^2} \, \left[ 
  \tikz[baseline=(C.base), every node/.style={font=\small}]{
    \tikzset{stencil/.style={circle,draw,minimum size=7mm,inner sep=0pt,align=center}}
    \node[stencil] (C) at (0,0)   {\(\mathbf{-2}\)};
    \node[stencil] (W) at (-1.2,0){\(\mathbf{1}\)};
    \node[stencil] (E) at (1.2,0) {\(\mathbf{1}\)};
    \draw[thick] (W) -- (C);
    \draw[thick] (E) -- (C);
  }
  \right] \; \star\;  u_{i} \,,
\end{equation}
in 1D, or
\begin{equation}
  (\Delta u)_{ij} \; \defeq \; \frac{1}{h^2} \, \left[ 
  \tikz[baseline=(C.base), every node/.style={font=\small}]{
    \tikzset{stencil/.style={circle,draw,minimum size=7mm,inner sep=0pt,align=center}}
    \node[stencil] (N) at (0,1.2) {\(\mathbf{1}\)};
    \node[stencil] (C) at (0,0)   {\(\mathbf{-4}\)};
    \node[stencil] (S) at (0,-1.2){\(\mathbf{1}\)};
    \node[stencil] (W) at (-1.2,0){\(\mathbf{1}\)};
    \node[stencil] (E) at (1.2,0) {\(\mathbf{1}\)};
    \draw[thick] (N) -- (C);
    \draw[thick] (S) -- (C);
    \draw[thick] (W) -- (C);
    \draw[thick] (E) -- (C);
  }
  \right] \; \star\;  u_{ij} \,,
\end{equation}
in 2D, and 
\begin{equation}
  (\Delta u)_{ijk} \;\defeq\; \frac{1}{h^2} \,
  \left[
  \tikz[baseline=(C.base), every node/.style={font=\small}]{
    \tikzset{
      stencil/.style={circle,draw,minimum size=6mm,inner sep=0pt,align=center},
      depth/.style={stencil,draw=gray!60}
    }
    \node[stencil] (N) at (0,1.2)   {\(\mathbf{1}\)};
    \node[stencil] (C) at (0,0)     {\(\mathbf{-6}\)};
    \node[stencil] (S) at (0,-1.2)  {\(\mathbf{1}\)};
    \node[stencil] (W) at (-1.2,0)  {\(\mathbf{1}\)};
    \node[stencil] (E) at (1.2,0)   {\(\mathbf{1}\)};
    \node[depth]     (B) at (0.7,0.6)  {\(\mathbf{1}\)}; 
    \node[stencil]   (F) at (-0.7,-0.6){\(\mathbf{1}\)}; 
    \draw[thick] (N) -- (C);
    \draw[thick] (S) -- (C);
    \draw[thick] (W) -- (C);
    \draw[thick] (E) -- (C);
    \draw[thick] (F) -- (C);
    \draw[thick,dashed,gray!80] (B) -- (C);
  }
  \right] \;\star\; u_{ijk}\,.
\end{equation}
in 3D, and so on. From here, it is almost immediate to see the connection with convolutional neural networks, in the sense that we have a \textit{kernel} (the stencil operator) convolved with the input (the field), in 1D, 2D and 3D.

But we can draw an additional connection, to the field of Graph Theory, as well as Graph Machine Learning and Graph Neural Networks \cite{bacciu2020gentle, scardapane2024alice}. Indeed, let us suppose we have a graph $\mathcal{G} = \{ V, E \}$, where $V = \{v_i\}_{i=1, \ldots, N_{\text{nodes}}}$ is the set of vertices and $E = \{e_{ij} : (i,j) \text{ are connected}\}$ is the set of edges. 

Suppose we want to describe the evolution of temperature on a graph. It may be, for example, that we have a set of sensors in a building (the nodes), and few of those are ``connected'' by, e.g., corridors (the edges). We may define the ``temperature'' on the graph, i.e.~the temperature measured by the sensor $i$ at time $t$. Formally, we have a mapping $u : [0,T] \times\mathcal{G} \to \RR$, that we denote as $u_i(t)$. Now, as we have done in \cref{subsubsec:1:2:1:1:DiffEqDer}, we suppose that the rate of evolution of the temperature depends on the heat flow, which is proportional to the difference in temperature between a node $i$ and all its neighbouring nodes $j \in \mathcal{N}_i$, i.e.
\[
\partial_t u_i (t) \propto \sum_{j\in \mathcal{N}_i} [u_j(t) - u_i(t)] \,.
\]
Now, we recall the concept of \textit{Adjacency Matrix} $\mathrm{A}\in \mathcal{M}_{N\times N}(\RR)$\note{With $ \mathcal{M}_{N\times N}(\RR)$ we denote the space of the $N\times N$ matrices with entry values in $\RR$.}, with entries $A_{ij}$, which is the matrix defined as
\[
    A_{ij} = \begin{cases}
        1 \,, & (i,j) \text{ are connected} \,, \\
        0\,, & \text{otherwise} ,
    \end{cases}
\]
which is, by construction, symmetric, at least for undirected graphs, as in this case. 
Now, the above equation is, calling $\alpha$ the proportionality constant,
\[
    \begin{split}
        \partial_t u_i(t) &= \alpha \sum_{j=1}^N A_{ij} [u_j(t) - u_i(t)] \\
        &= \alpha \sum_{j=1}^N A_{ij} u_j(t) -   \alpha \left( \sum_{j=1}^N A_{ij}  \right)  u_i(t) \,, 
    \end{split}
\]
now, the last term is simply computing the \textit{connectivity} of the $i-$th node, i.e. the number of other nodes connected (sharing an edge) with the $i-th$ node, which is called \textit{degree}, $d_i$,
\[
    d_i \defeq \sum_{j=1}^N A_{ij}   \,,
\]
and the diagonal matrix defined as 
\[
    \mathrm{D} \defeq  \operatorname{diag}(d_1, \ldots, d_N) \,, \quad D_{ij} = \delta_{ij} \, d_i \,,
\]
is the \textit{Degree Matrix}. Now, we see that we can rewrite the \textit{diffusion equation in the graph} in a matricial form; called $\mathbf{u}(t) = (u_1(t), \ldots , u_N(t))$, we have
\begin{equation}\label{eq:2:1:5:GraphDiff}
    \partial_t\mathbf{u}(t) = -  \alpha [\mathrm{D} - \mathrm{A}] \cdot \mathbf{u}(t) .
\end{equation}
Thus, the matrix appearing there
\begin{equation} \label{eq:2:1:5:GraphLapl}
    \mathrm{L} \defeq \mathrm{D} - \mathrm{A} \,, \qquad L_{ij} = D_{ij}  - A_{ij}  \,,
\end{equation}
is, by analogy, called \textit{Graph Laplacian} matrix of the graph. 

\subsubsection{The spectral theory on the graph and Laplacian Eigenmaps as an unsupervised dimensional reduction method}

Disclaimer: for a more in-depth introduction to the subject, see e.g. \cite{chung1996lectures}. 

We have thus seen that the diffusion on graphs behave like the diffusion PDE \cref{eq:2:1:5:GraphDiff}, where the Laplacian is implemented as the above Graph Laplacian $\mathrm{L}$ \cref{eq:2:1:5:GraphLapl}. But, by definition, for undirected graphs, the Graph Laplacian is real and symmetric, $\mathrm{L}=\mathrm{L}^T$. Furthermore, for any function $g : \mathcal{G} \to \RR$, i.e. $g : v \mapsto g(v)$ (like the heat above at fixed time), we can see that
\begin{align*}
    \frac{\bra \mathrm{g} , \mathrm{L} \mathrm{g} \ket }{\bra \mathrm{g}, \mathrm{g}\ket} &= \frac{\sum_{i,j} L_{ij} g_i g_j}{\sum_{i = 1}^{N} g_i^2} \\
    &= \frac{\sum_{i,j} [D_{ij} - A_{ij}] g_i g_j}{\sum_{i = 1}^{N} g_i^2}  \\
    &= \half \frac{ \sum_{i,j}A_{ij} (g_i^2 + g_j^2) - 2 \sum_{i,j} A_{ij} g_i g_j}{\sum_{i = 1}^{N} g_i^2} \\
    &= \half\frac{\sum_{i,j} A_{ij} (g_i - g_j)  (g_i - g_j) }{\sum_{i = 1}^{N} g_i^2}  \\
    &= \half\frac{\sum_{i\sim j}  (g_i - g_j)^2 }{\sum_{i = 1}^{N} g_i^2} \ge 0 \,,
\end{align*}
where $i\sim j$ denotes that $(i,j)$ shares an edge, and where we have noted that, since $D$ is diagonal, $A$ here is symmetric with zero diagonal (no self-loops), and $d_i = \sum_jA_{ij}$, so
\begin{align*}
    \sum_{i,j} D_{ij} g_i g_j &= \sum_{i,j} D_{ij} g_i^2 = \sum_{i} g_i^2 \sum_j A_{ij} \\
    &= \half \sum_{i,j}A_{ij} (g_i^2 + g_j ^2) \,.
\end{align*}

Thus, we have proved that the Graph Laplacian is a (real symmetric) positive semi-definite matrix\note{The reader may have noticed the minus sign in the definition of the Graph Laplacian w.r.t. the Laplacian operator analogy in the diffusion equation. The sign is inserted to have the graph Laplacian positive semi-definite, and not negative semi-definite.}; thus, (i) by the Spectral Theorem, it is a diagonalisable matrix, i.e. $\exists \{\lambda_i,  \mathrm{v}_i\}_{i=1, \ldots , N}$ eigenvalues - (orthonormal) eigenvectors pairs, such that, 
\[
 \mathrm{L} = U \Lambda U^T \,, \quad \Lambda = \operatorname{diag} (\lambda_1, \ldots , \lambda_N) \,, \; U = ( \mathrm{v}_1 | \ldots, |  \mathrm{v}_N) \,,
\]
and that, since $\bra g, \mathrm{L} g \ket \ge 0, \; \forall g$, we can specialise to any $\mathrm{v}_i$ and see that
\[
0 \le \bra \mathrm{v}_i, \mathrm{L} \mathrm{v}_i \ket = \lambda_i \bra \mathrm{v}_i, \mathrm{v}_i \ket = \lambda_i \,, \qquad \forall i=1, \ldots, N \,,
\]
i.e. all eigenvalues of the graph Laplacian are greater or equal to zero. 

Notice that it is easy to see that there is an eigenvector with zero eigenvalue, which is $\mathrm{v}_1 = (1, \ldots, 1)/\sqrt{N}$, since $\mathrm{L}\mathrm{1} = 0$. Furthermore, if the graph is connected, this eigenvalue is simple. 

So the eigenvalues of the graph Laplacian can be ordered as\note{In general, the number of zero eigenvalues is the number of connected components present in the original graph.}
\[
    0 = \lambda_1 \le \lambda_2\le \cdots \le \lambda_N \,.
\]
This fact is quite relevant for the Laplacian Eigenmaps.

\paragraph{Laplacian Eigenmaps}
Laplacian Eigenmaps is an unsupervised learning method to non-linearly dimensionally reduce a dataset $X\in \RR^{N\times F}$, with $N$ data having $F$ features \cite{belkin2003laplacian}. In the abstract, they state

\quotespace
\begin{quote}
    \say{
    [...]\textit{We consider the problem of constructing a representation for data lying on a lowdimensional manifold embedded in a high-dimensional space. Drawing on the correspondence between the graph Laplacian, the Laplace Beltrami operator on the manifold, and the connections to the heat equation, we propose a geometrically motivated algorithm for representing the high dimensional data. The algorithm provides a computationally efficient approach to nonlinear dimensionality reduction that has locality-preserving properties and a natural connection to clustering.}
    }    
\end{quote}
\quotespace

\noindent Laplacian Eigenmaps utilises the fact that the eigenvalues of the graph Laplacian are ordered (and positive), and represents the scale of the rate of changes of the node features over the graph. Indeed, working in analogy with the Laplace (Beltrami) operator, $-\Delta$, and its eigenfunctions, the Fourier modes $\e^{\ii k \cdot x}$, the authors use the low-eingevalues eigenvectors of the weighted graph Laplacian, built using $W_{ij}$, i.e. the weighted-graph adjacency matrix, also called \textit{Heat Kernel}, defined as
\[
    W_{ij} = A_{ij} \, \exp\left[ - \beta \| \mathrm{x}_i - \mathrm{x}_j \|_2^2 \right] \,,
\]
(where $\beta$ is an hyper-parameter) to extract \textit{global} properties of the graph, built on top of the dataset $X$. The algorithm is
\begin{itemize}
    \item[1.] {Build the Graph}: Starting from the dataset $X$, an $\epsilon-$neighbourhood is built, i.e., for each data-point $\mathrm{x}_i$ (i.e., the node), we say that $\mathrm{x_j}$ is connected if $\| \mathrm{x}_i - \mathrm{x}_j \| \le \epsilon$, where the norm is computed in the feature space.\note{Another approach could be use kNN algorithm to selected $k$ neighbours for each data point.}
    \item[2.] The Heat Kernel is computed;\note{Sometimes, Laplacian Eigenmaps are performed on the standard graph Laplacian $L$, so using $W=A$.} Thus, from there, we defined the (weighted) Laplacian as
    \[
        \mathrm{L} = \mathrm{D} - \mathrm{W} \,, \qquad D_{ij} = \delta_{ij}
         \sum_k W_{ik} \,.
    \]
    \item[3.] The following (generalised) eigen-problem is solved:
    \[
        \mathrm{L} \mathrm{v} = \lambda \mathrm{D}  \mathrm{v} \,.
    \]
    \item[3.] The reduced dimensional components are thus selected as the first $k$ (non-trivial) eigenvectors ($\mathrm{v}_1$ is the trivial one), 
    \[
    \mathbf{x}_i \mapsto \left( \mathrm{v}_2 (i), \ldots,  \mathrm{v}_{k+1} (i) \right) \,.
    \]
\end{itemize}

The analogy $\mathrm{L} \longleftrightarrow-\Delta$ is 
\begin{align*}
    \mathrm{L} &\longleftrightarrow-\Delta \\
    (\lambda_i, \mathrm{v}_i ) &\longleftrightarrow (|k|^2, \e^{\ii k \cdot x})  \\
    g = \sum_i \hat{g}_i \;   \mathrm{v}_i \,, \quad \hat{g}_i \defeq \langle \mathrm{v}_i , g\rangle &\longleftrightarrow  f(x) = \int  f_k \e^{\ii k \cdot x}  \dd k \,, \quad f_k \defeq \langle \e^{\ii k \cdot x}, f \rangle 
\end{align*}
so that, low $k$ (or low $\lambda_k$) corresponds to slowly oscillating modes ($e^{ik\cdot x}$ or $\mathbf{v}_k$) thus extracting long-range behaviour of the function;  high $k$ corresponds to rapidly oscillating modes that capture fine-scale features.

The idea behind Laplacian Eigenmaps is thus to keep the lowest ones, because the highest ones may be contain noise effects in the data, effectively reducing the feature spaces in a non-linear way. 

Notice that the spectral problem solved here is a \textit{modified} one; or, better said (if $L$ is the standard Laplacian), we are solving the spectral problem for the \textit{random-walk} graph Laplacian, 
\[
    \mathrm{L}_{\text{random-walk}} \defeq  \mathrm{I} - \mathrm{D}^{-1} \mathrm{W} .
\]

About the Heat Kernel terminology: the authors of \cite{belkin2003laplacian} uses this terminology. In other works, it is sometime denoted as \textit{Similarity Matrix} or \textit{Affinity Matrix}, see e.g.~\cite{ng2001spectral}. What is interesting in using the Heat Kernel terminology is that, in Functional Analysis, the Heat Kernel  is the fundamental solution to the heat equation on a specified domain with appropriate boundary conditions. If we study the Heat PDE in $\mathbb{R}^d$, with zero Dirichlet boundary conditions at infinity, the Heat Kernel $K$ is
\begin{equation}
    K(t; x,y) = \frac{1}{(4\pi t)^{d/2}} \ \exp\left[ - \frac{\| x- y\|_2^2}{4t}  \right] ,
\end{equation}
which is defined $\forall x,y\in\RR^d, \forall t\in \RR, t>0$. Interestingly, in the $t\to0$ limit
\[
\lim_{t\to 0} K(t; x,y) = \delta(x-y) ,
\]
in the sense of the distributions, i.e., for all test function $f(x)$
\[
\lim_{t\to 0} \int K(t; x,y) f(y) \mathrm{d}^d y = f(x) .
\]
Now, notice that the Heat kernel can be written in terms of an infinite sum of the eigenfunctions of the Laplacian operator
\[
\Delta \phi_n + \lambda_n \phi_n = 0 \,, \quad \left.\phi_n \right|_{\partial \Omega} = 0 \,,
\]
and, as we have seen, since $-\Delta$ is positive semi-definite, $0\le \lambda_1 \le \lambda_2\le \cdots$, and 
\begin{equation}
    K(t; x,y) = \sum_{n=1}^\infty \ee^{-\lambda_n t} \, \phi_n(x) \phi_n(y) \,,
\end{equation}
which reminds the \textit{kernel trick} for the kernel PCA. Formally differentiating the series under the sign of the summation shows that this should satisfy the heat equation. However, convergence and regularity of the series are quite delicate.

The heat kernel is also sometimes identified with the associated integral transform $T$, defined for compactly supported smooth $f$ by
\[
T_t [f](x) = \int_\Omega K(t; x,y) f(y) \dd^d y \,,
\]
which can be represented in operatorial form via exponentiation of the Laplacian, 
\[
T_t = \ee^{t \Delta} = \ee^{(-t)  (-\Delta)} \,.
\]

\paragraph{Spectral Graph Convolutions}

In \cite{bruna2013spectral, defferrard2016convolutional} the authors wanted to extend the concept of \textit{convolution} from images to graphs. The idea was that, since in the graph context there is no simple meaningful way to define a \textit{translation} operator in the vertex domain, it is better to define the convolution operator on the graph in the Fourier space, i.e. the space of the eigenvectors of the graph Laplacian, such that
\[
    x \star_\mathcal{G} y = U \left[ (U^T x) \odot (U^T y)   \right],
\]
where $\odot$ is the element-wise Hadamard product. 
There, we can see that $x$ signal is filtered by (a possible learnable) $g_\theta$ via 
\[
y = g_\theta (L) x = g_\theta (U \Lambda U^T) x = U g_\theta(\Lambda) U^T x .
\]

So the idea of the graph spectral convolution is 
\begin{itemize}
    \item[(i)] Diagonalise the Graph Laplacian, and construct the Fourier space; 
    \item[(ii)] Transform the input into the Fourier space, $x\mapsto U^Tx$;
    \item[(iii)] Apply a learnable transformation $g_\theta$ in Fourier space
    \item[(iv)] Transform back to direct space, using $U^{-1} = U^T$, i.e., by multiplying by $U$. 
\end{itemize}

This spectral approach is computationally expensive, since it requires eigendecomposition, which motivated the development of \textit{approximate} spectral methods (e.g., ChebNet, GCN) that avoid explicit diagonalisation.

In \cref{chap:5:NeuralOperators}, 
we will encounter Fourier Neural Operators, which exploit a closely related spectral idea: transforming a field to Fourier space, applying a learnable transformation to its spectral modes, and transforming it back.

\section{Finite Element Methods (FEM)} \label{sec:2:2:FEM}

In this section, we briefly introduce the \textit{Finite Element Method} (FEM). We have no intention of furnishing an exhaustive introduction on the subject: this section is merely intendend as a first contact with the numerical resolution of PDEs in the weak sense. For a broader introduction to the subject, see \cite{iserles2009first, Gerbeau2009, Smith2022, quarteroni2020numerical}, and references therein. 

Following the useful summary of \cite{iserles2009first}, it is convenient to begin the section by outlining the five main ideas behind the finite element method:
\begin{itemize}
    \item[1.] Formulate the Cauchy problem\note{We will use the term \textit{Cauchy Problem} with a more general sense, as an encompassing term referring to the PDE + adequate number of Boundary and Initial Conditions.} as a \textit{variational} problem, e.g. by seeking a function that minimizes a nonlinear functional;
    \item[2.] If possible, reduce differentiability requirements in the above functional, using integration by parts;
    \item[3.] If integration by parts have been used in step 2., take appropriate care with boundary conditions, and distinguish between essential and natural conditions;
    \item[4.] Replace the underlying problem by an approximate one, restricted to a finite-dimensional space;
    \item[5.] Choose a finite-dimensional space which is spanned by functions with a small support (finite
element functions).
\end{itemize}

Ok, before delving into it further, let us recap the problem. We have to find $u : \Omega \to \mathbb{R}$, which satisfies the Cauchy problem
\begin{equation}
    \begin{split}
        \mathcal{F}[u(x)] &= J(x) \,, \quad x\in \Omega \quad\; \, (\text{PDE}) \\
        \mathcal{B}[u(x)] &= g(x) \,, \quad x\in \partial\Omega \quad (\text{BC}) \\
    \end{split}
\end{equation}
where, for the moment, we are focusing on \textit{pure  boundary} problems (i.e., time-free). We assume that the aforementioned problem can be written in its \textit{weak formulation} via a bilinear form $\mathfrak{g} : X(\Omega) \times X(\Omega) \to \RR$, where $X(\Omega)$ is the Banach space where $u$ lives, $u\in X(\Omega)$.

\begin{figure}[t]
    \centering
    \includegraphics[width=0.85\linewidth]{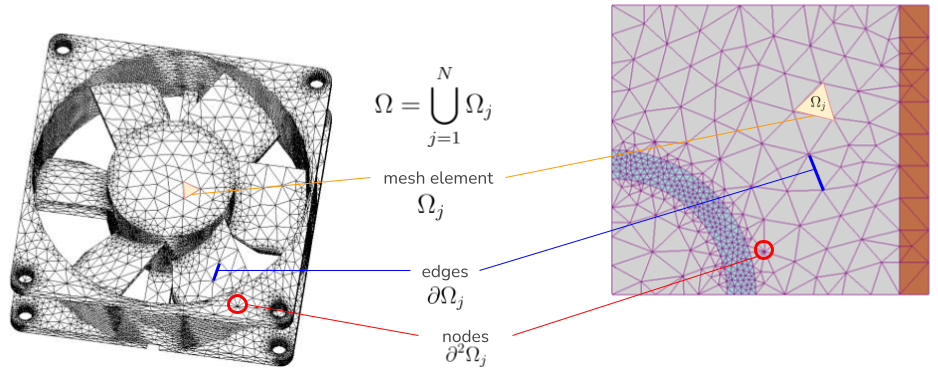}
    \caption{Visual representation of the \textit{meshing} of the domain $\Omega$ for the FEM analysis (2D surface in 3D space on the left, and a plain 2D region in the right). Adapted from \cite{BECHET20021}.}
    \label{fig:2:2:FEM1}
\end{figure}

We then define a \textit{mesh} of the $\Omega\subset \RR^d$, i.e. we represent $\Omega$ as an union of subdomains, the \textit{elements}, 
\[
\Omega = \bigcup_{j=1}^N \Omega_j \,,
\]
where \note{With $\overset{\circ}{\Omega}$ we denote the \textit{interior}, i.e., $\overset{\circ}{\Omega}=\Omega \setminus\partial\Omega$.} $\overset{\circ}{\Omega}_i \cap \overset{\circ}{\Omega}_j = \emptyset$.

\begin{figure}
    \centering
    \includegraphics[width=0.65\linewidth]{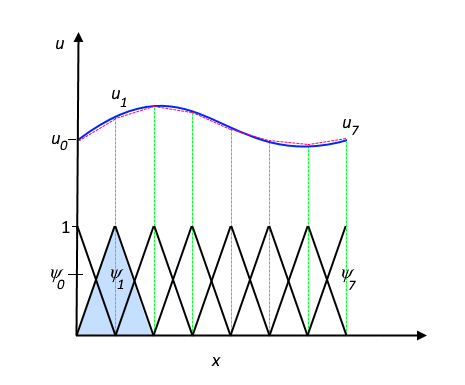}
        \caption{The function $u$ (solid blue line) is approximated with $u_h$ (dashed red line), which is a linear combination of linear basis functions $\{\phi_j\}$, represented by the solid black lines. From \cite{Comsol2025}.
    }
    \label{fig:2:2:FEM_repr}
\end{figure}

These elements have specific \textit{overlapping} regions, which are the \textit{edges}, $\partial \Omega_j$. A common approach in 2D is to use \textit{triangulation}, i.e. representing 2D domains as a union over triangles. In any case, we may have that our elements are manifold with \textit{corners}, i.e. we have a non-negligible $\partial^2\Omega_j$, which are \textit{corners}. 

Let us now build a basis family $\{\phi_j : \Omega \to \RR\}$, with appropriate behaviour under differentiation (up to a certain order, e.g. $\phi_j \in C^1(\Omega)$\note{Basis functions are defined on the entire domain $\Omega$, and have local support on $\Omega_j$}. We can represent the solution $u$ as 
\begin{equation}
    u(x) = \sum_{j=0}^N \alpha_j \phi_j(x) \,.
\end{equation}

\noindent Now, let us rewrite the weak formulation of the problem,
\begin{equation}
\begin{split}
    \mathfrak{g}(v, \mathcal{F}[u]) &= \mathfrak{g}(v, J) \,,   \qquad \forall v\in X(\Omega) ,\\
    \text{i.e. } \int_\Omega  v(x) \mathcal{F}[u(x)] \dd^d x &= \int_\Omega v(x) J(x) \dd^d x \,.
\end{split}
\end{equation}

Sometimes, we will use also $\mathscr{G}(u(x), v(x)) \defeq \langle u(x), v(x)\rangle_{\RR^d}$, so that we can write 
\[
 \int_\Omega  \mathscr{G} (\nabla v(x), \nabla u(x)) \dd^d x = \int_\Omega \nabla v(x)\cdot \nabla u(x) \dd^d x \,.
\]


\begin{figure}
    \centering
    \begin{subfigure}[t]{0.48\textwidth}
        \centering
        \includegraphics[width=\textwidth]{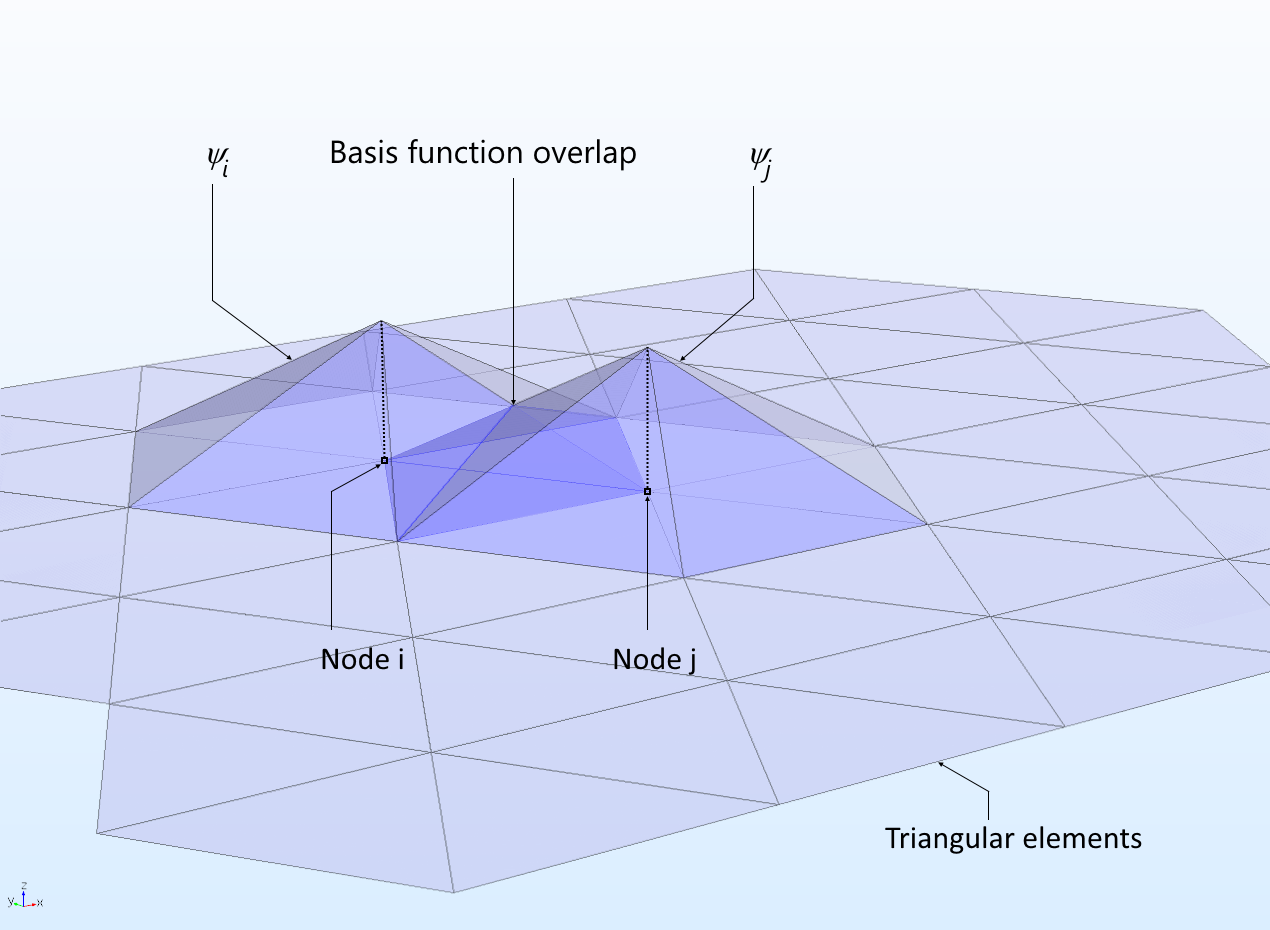}
        \caption{Overlap between basis functions. From \cite{Comsol2025}.}
    \end{subfigure}
    \hfill
    \begin{subfigure}[t]{0.48\textwidth}
        \centering
        \includegraphics[width=\textwidth]{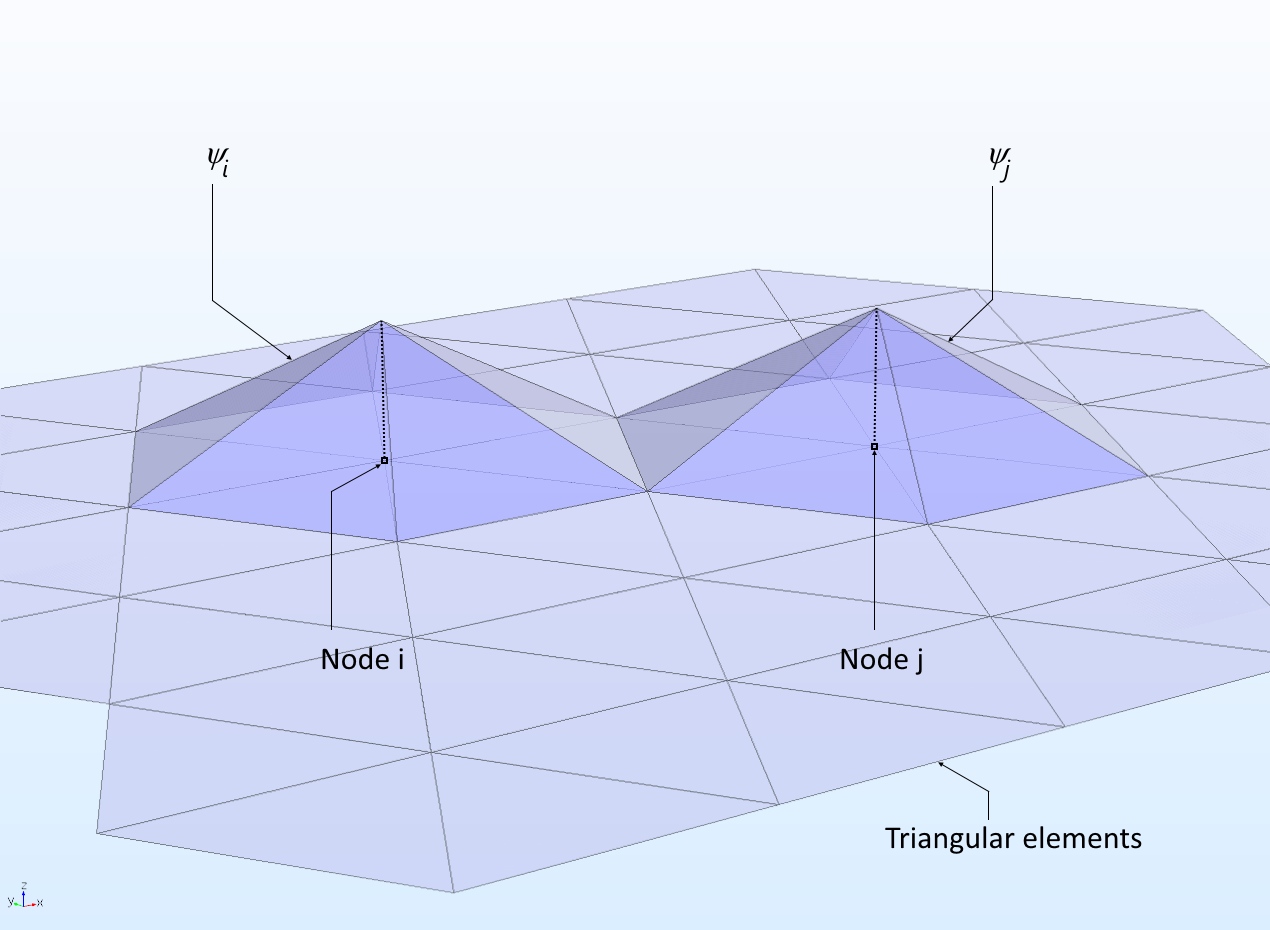}
        \caption{No overlap between basis functions. From \cite{Comsol2025}.}
    \end{subfigure}
    \label{fig:2:2:FEM2}
\end{figure}

Now, let us focus on a specific case: Laplace operator on $\RR^d$, i.e. we have the Poisson problem 
\[
    \nabla^2 u(x) = J(x), \quad x\in \Omega \subset \RR^d,
\]
so that
\[
    \int_\Omega v(x) \nabla^2u(x) \dd ^d x = \int_\Omega v(x) J(x) \dd ^d x \,.
\]
Now, integrating by parts via the Gauss-Green-Ostrogradskij-Stokes Theorem, we get
\begin{equation}
   \underbrace{ \int_{\partial \Omega} v(x) \nabla u \cdot \hat{n}_\Sigma \dd^{d-1}\Sigma }_{\text{Boundary Term}} - \int_\Omega \nabla v(x) \cdot \nabla u(x) \dd^d x = \int_\Omega v(x) J(x) \dd ^d x \,,
\end{equation}
obtaining the \textit{weak formulation} of the PDE. 

Why is it useful? Well, now, we can represent our functions on the $\{\phi_j\}$ basis\note{Notice that we are using the Riesz theorem under the hood, to see $J\in X(\Omega)$. Indeed, the Riesz representation theorem guarantees that $J \in X(\Omega)^*$ can be identified with an element of $X(\Omega)$ when $X(\Omega)$ is a Hilbert space.}
\begin{equation}
    u(x) = \sum_{j=0}^N \alpha_j \phi_j(x) \,, \qquad J(x) = \sum_{j=0}^N J_j \phi_j(x) \,.
\end{equation}
Using this, we recast our \textit{differential} problem into an \textit{algebraic} one, specialising the basis in a way that
\begin{align}
    \int_\Omega   \phi_j (x)\phi_k (x)  \dd^d x &\neq  0 \,, \qquad \text{if }\, \Omega_j \cap  \Omega_k \neq \emptyset  \, \\
    \int_\Omega \mathscr{G}\left(  \nabla \phi_j (x), \nabla \phi_k (x)  \right) \dd^d x &\neq  0 \,, \qquad \text{if } \, \Omega_j \cap  \Omega_k \neq \emptyset
\end{align}
so that we get\note{Here we assume that the boundary conditions are pure Dirichlet and that the construction of $u$ is such that the boundary term is zero. }
\begin{equation}
    - \sum_{j=0}^N \alpha_j \int_\Omega \mathscr{G}\left(  \nabla \phi_j (x), \nabla \phi_k (x)  \right) \dd^d x = \sum_{j=0}^N J_j  \int_\Omega     \phi_j (x) \phi_k (x)   \dd^d x \,,
\end{equation}
where we have picked a specific test function to simply the computations. Now, since the integrals can be explicitly computed once we have selected the basis, we can define the matrices $\mathrm{L}_{jk}$ (called \textit{stiffness} matrix) and $\mathrm{M}_{jk}$ (called \textit{mass matrix}) as 
\[
 \mathrm{L}_{jk} \defeq \int_\Omega \mathscr{G}\left(  \nabla \phi_j (x), \nabla \phi_k (x)  \right) \dd^d x  \,, \qquad  \mathrm{M}_{jk} \defeq  \int_\Omega   \phi_j (x)\phi_k (x)   \dd^d x  \,,
\]
and thus we get the \textit{algebraic} problem
\footnote{
    Notice that we don't need to expand $J$ on $\{\phi_j\}$; indeed, we can rewrite the problem 
    \[
    - \int_\Omega \nabla v \cdot \nabla u \dd x = \int_{\Omega} J v \dd x \,, \qquad \forall v \,,
    \]
    so that, expanding again $u$ and picking $v=\phi_k$, 
    \[
     - \sum_{j=0}^N \alpha_j \int_\Omega \mathscr{G}\left(  \nabla \phi_j (x), \nabla \phi_k (x)  \right) \dd^d x = \int_\Omega J \phi_k \dd x \,, 
    \]
    i.e., 
    \[
        \mathrm{L} \alpha = \mathrm{F} \,, \quad \text{where } \mathrm{F} \defeq  \int_\Omega J \phi_k \dd x \,.
    \]
    Expanding now $J$, we recover the mass matrix formulation
    \[
         \mathrm{F} = \mathrm{M} J \,.
    \]
}
\begin{equation}
    - \mathrm{L} \alpha = \mathrm{M} J \,,\qquad  \alpha = (\alpha_1, \ldots, \alpha_N), J = (J_1, \ldots, J_N),
\end{equation}
which can formally solved as 
\[
\alpha = - \mathrm{L}^{-1}  \mathrm{M} J  \,.
\]

\subsection{Ritz vs Galerkin method} \label{subsec:4:2:1:RitzVsGalerkin}

Let us now standardize a little bit the terminology, as well as introducing relevant formulation of the FEM approaches. 

\begin{definition}[Weak formulation]
Let \(\Omega\subset\mathbb{R}^d\) be a bounded domain and set
\(V\) to be a suitable Sobolev space (typically \(V=H_0^1(\Omega)\)).
Given a bilinear form \(a(\cdot,\cdot):V\times V\to\mathbb{R}\) and a bounded
linear functional \(f\in V'\), the \emph{weak} (or \emph{variational})
formulation of the boundary value problem is: find \(u\in V\) such that
\[
    a(u,v)=f(v)\qquad\forall v\in V\,.
\]
\end{definition}

Notice that, up to now, we  have said nothing on the bilinear forms. Let us relax the generalisation a bit, and let us keep in mind the Poisson problem of before. Using far away memories from physics courses, we may recall that, in some conservative systems, we may defined the solution trajectory of a point-particle of mass $m$ as the one which \textit{minimise} the energy \textit{functional}; indeed, if a particle moves in a \textit{potential} $U=U(q)$  with a trajectory $q : t\in[0, T] \mapsto q(t) \in \RR^d$, the energy functional $E : q\in X \mapsto \RR$ is
\[
    E[q] = \half \, m \dot{q}^2 + U(q) \,,
\]
which can be also seen as a function $E[t], t\in[0, T] \mapsto \RR$ once $q$ is selected; but energy is conserved, thus $\partial_t E=0$, i.e.
\begin{equation}
    \begin{split}
        0 &= \partial_t E  = \partial_t \left( \half \, m \dot{q}^2 + 
        U(q) \right) \\
        &= \dot{q} \left( m \ddot{q} + \nabla_q U  \right) \Longrightarrow m \ddot{q} = -  \nabla_q U, \quad\text{since } \dot{q}\neq0 ,
    \end{split}
\end{equation}
which is the standard Newton law for conservative forces, since the force is $\mathrm{F} = m \ddot{q}$. Thus, we can think to solve the differential problem by \textit{minimising} the Energy functional of the system (describing the \textit{status} of the system), instead of solving the differential equation (describing the \textit{dynamics} of the system)\note{The derivation above is heuristic: the correct variational route is via Hamilton's principle (or the Euler-Lagrange equations) rather than energy conservation alone, and a rigorous derivation requires specifying the admissible function space and boundary conditions. }.

These two approaches are at the core of two similar, yet different, schemes: the \textit{Ritz Method} and the \textit{Galerkin Method}. 

\begin{definition}[Ritz method (energy formulation)]
Assume the PDE admits an energy functional \(E:V\to\mathbb{R}\) whose
Euler--Lagrange equation is equivalent to the weak form above. The continuous
problem can then be written as the variational minimisation
\[
u=\arg\min_{v\in V} E[v].
\]
The \emph{Ritz} discretisation is obtained by restricting \(E\) to a finite
dimensional trial space \(V_h\subset V\) and solving
\[
u_h=\arg\min_{v\in V_h} E[v].
\]
Notice that, when \(E\) is quadratic, one obtains a symmetric coercive
bilinear form \(a(\cdot,\cdot)\) 
and the discrete equations follow from
\[
a   (u_h,v)=f(v), \qquad \forall v\in V_h \,.
\]
\end{definition}

Instead, if we don't want (or can't) use the energy functional minimisation trick, we are back to what we have seen before (maybe via the integration by parts of the Laplace operator): the Galerkin problem. 

\begin{definition}[Galerkin method]
Given the weak form \(a(u,v)=f(v)\), choose finite-dimensional trial and test
spaces \(V_h\subset V\) and \(W_h\subset V\). The \emph{Galerkin} (or
Petrov--Galerkin) method seeks \(u_h\in V_h\) such that
\[
a(u_h,w)=f(w)\qquad\forall w\in W_h.
\]
When \(W_h=V_h\) the method is the (Bubnov) Galerkin method; when
\(W_h\neq V_h\) it is a Petrov--Galerkin method. Galerkin projection requires
a well‑posed weak form but does not require the existence of an energy
functional.
\end{definition}

\begin{remark}
    The essential conceptual distinction is that \emph{Ritz} originates from an
    \emph{energy minimisation} (so it requires a variational/Euler-Lagrange
    structure and typically leads to symmetric positive definite systems), while
    \emph{Galerkin} is a \emph{projection} of the weak residual onto a test space
    and therefore applies more broadly (including non‑self‑adjoint operators).
    Integration by parts is a common step in deriving weak forms for both
    approaches and is not the defining difference.
\end{remark}

\begin{remark}
A relevant case where we encounter a Galerkin problem, but not a Ritz problem, is the general Elliptic PDE
\begin{equation}\label{eq:2:2:RAD_Eq}
     - \nabla\cdot [a(x) \nabla u(x)] + \mathrm{b}(x) \cdot \nabla u(x) + c(x) u(x) = f(x) ,
\end{equation}
where integration by parts cannot be made for \textit{all} the LHS\note{Better said, we can recast the problem in a \textit{weak form}; but the advection terms makes the bilinear form non-symmetric \cite{Gerbeau2009}.}. This PDE contains
\begin{itemize}
    \item in $a(x)$, the information about \textit{media anisotropy} for the diffusion of $u$; for modelling the diffusion in anistropic/porous media, we have the Fick Law relating the flux $q$ to the field $u$, $\mathbf{q} = - a(x) \nabla u(x)$, where $a$ is called \textit{diffusion matrix}. 
    \item in $\mathrm{b}(x)$, the information about the \textit{transport} or \textit{convection} of the field $u$; in such case, $\mathrm{b}$ is related to the velocity of the object the field $u$ is moving into (e.g., a substance concentration $u$ moving in a stream of liquid with velocity $\mathrm{b}$.
    \item in $c(x)$, the information about the \textit{reaction} of the media to the diffusion of $u$; for example, if we are modelling the concentration of a substance with $u$, $c$ may be the rate of \textit{decomposition} ($c<0$) or \textit{growth} ($c>0$).
\end{itemize}
This is why \cref{eq:2:2:RAD_Eq} is called \textit{Reaction-Advection-Diffusion} equation.
\end{remark}

\begin{example}[Poisson problem, reprise]
    For \(-\Delta u=f\) with \(u|_{\partial\Omega}=0\) the energy functional is
    \[
    E(v)=\frac{1}{2} \int_\Omega |\nabla v|^2\,\mathrm{d}x - \int_\Omega f v\,\mathrm{d}x,
    \]
    whose Euler-Lagrange equation yields the weak form
    \(\int_\Omega \nabla u\cdot\nabla v\,\mathrm{d}x=\int_\Omega f v\,\mathrm{d}x\)
    for all \(v\in H_0^1(\Omega)\). In this case the Ritz and (Bubnov) Galerkin
    formulations \textit{coincide}.
\end{example}

Why are these two methods relevant? Well, because of the following:

\begin{theorem}[Lax-Milgram Theorem]
Let \( a(\cdot, \cdot) : V \times V \rightarrow \mathbb{R} \) be a \textit{bounded} and \textit{coercive} \textit{bilinear} form on the Hilbert space \( V \). Then, for each bounded linear functional \( f \in V' \), there is exactly one \( u_M \in V \) with
\[
    a(u, v) = f(v) \quad \forall v \in V.
\]
\end{theorem}

Therefore, whenever the weak formulation satisfies boundedness and coercivity, the Lax–Milgram theorem guarantees existence and uniqueness of its weak solution.

\begin{tcolorbox}[colback=white,colframe=black,title=Ritz vs Galerkin - Quick Comparison]
\begin{tabularx}{\linewidth}{@{}lX X@{}}
\textbf{Aspect} & \textbf{Ritz} & \textbf{Galerkin} \\
\hline
Principle & Minimise an energy functional; solution is variational minimiser. & Enforce weak form residual orthogonality to test space. \\
Applicability & Requires PDE to admit a variational formulation; typically self-adjoint. & Applies to weak forms even without an energy functional. \\
Boundary handling & Natural BCs arise from variational derivative; essential BCs on trial space. & Essential BCs on trial space; natural BCs appear in weak form. \\
Advantages & Energetic interpretation; often symmetric positive definite systems. & Broad applicability; flexible test/trial choices (Petrov–Galerkin). \\
Drawbacks & Limited to variational problems. & May yield non-symmetric systems; stability depends on pairing.
\end{tabularx}
\end{tcolorbox}

\section{Finite Volume Methods (FVM)} \label{sec:2:3:FVM}

As with the Finite Element Method, this section does not aim to provide an exhaustive introduction to Finite Volume Methods; it is merely intended as a first contact with the numerical solution of PDEs through a conservation-based approach. For broader introductions to the subject, see \cite{quarteroni2020numerical, eymard02100732, leveque2002finite, toro2013riemann}, and references therein.

Finite Volume Methods are particularly natural for PDEs expressing conservation or balance laws, as commonly encountered in fluid mechanics, heat and mass transfer, and related transport problems. The domain is decomposed into a collection of \textit{control volumes}, and the governing equation is integrated over each cell. Divergence terms are converted into fluxes through the cell boundary, so that the discrete unknowns naturally represent cell-averaged quantities. Depending on the formulation, the control volume may coincide with a mesh cell (cell-centred FVM) or be constructed around mesh vertices or nodes.

A key property of FVM is \textit{local conservation}. At the discrete level, this requires assigning a single conservative numerical flux to every interior face shared by two neighbouring cells: the flux leaving one cell must enter the other with the opposite sign. When this condition is satisfied, internal face contributions cancel exactly when cell balances are summed, yielding global conservation up to physical boundary fluxes and source terms. This structure also makes finite-volume schemes well suited to unstructured meshes.

Suppose now that we have a scalar balance law in conservative form
\begin{equation}
    \partial_t q(t, \mathbf{x}) + \nabla \cdot \mathbf{F} (q(t, \mathbf{x}), t, \mathbf{x}) = f(t, \mathbf{x}) \, ,
\end{equation}
defined on a bounded domain $\mathcal{U} \subset \RR^d$. For $f=0$ this reduces to a pure conservation law. Let the domain be partitioned into control volumes $\{X_n\}_{n=1}^N$ such that
\[
    \overline{\mathcal U} = \bigcup_{n=1}^N \overline{X_n} \, ,
    \qquad
    \overset{\circ}{X_n}  \cap \overset{\circ}{X_k} = \emptyset \quad \text{for } n\neq k \, .
\]

\begin{figure}[t]
    \centering
    \includegraphics[width=0.85\linewidth]{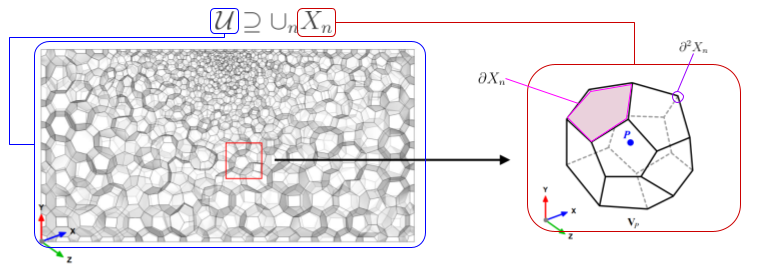}
    \caption{Example of finite volume tessellation.}
    \label{fig:2:3:FVM1}
\end{figure}

\noindent For the $n$-th control volume, we integrate the balance law over $X_n$:
\[
\int_{X_n} \partial_t q(t, \mathbf{x}) \, \dd \omega_n
+ \int_{X_n} \nabla \cdot \mathbf{F} (q(t, \mathbf{x}), t, \mathbf{x}) \, \dd \omega_n
= \int_{X_n} f(t, \mathbf{x}) \, \dd \omega_n  \,.
\]
Let us now define the cell averaged quantities
\[
    \bar{q}_n(t) \defeq \frac{1}{|X_n|} \int_{X_n} q(t, \mathbf{x}) \dd \omega_n \, ,
    \qquad
    \bar{f}_n(t) \defeq \frac{1}{|X_n|} \int_{X_n} f(t, \mathbf{x}) \dd \omega_n \, ,
\]
where $|X_n|$ is the volume of the cell. Assuming fixed control volumes and applying the divergence theorem, we obtain the exact cell balance
\begin{equation}
    |X_n| \, \partial_t \bar{q}_n(t)
    + \int_{\partial X_n} \mathbf{F}(q(t,\mathbf{x}),t,\mathbf{x}) \cdot \hat{\mathbf n}_n \, \dd \Sigma_n
    = |X_n| \, \bar{f}_n(t) \, .
\end{equation}
This equation already contains the basic idea of FVM: the evolution of the total amount stored in a cell is determined by the net flux through its boundary plus the source contribution inside the cell.

The exact flux on a cell face is generally not known because $q(t,\mathbf{x})$ on the face is not one of the discrete unknowns. We therefore introduce a \textit{numerical face flux}. Let $\mathcal{E}_n$ denote the set of faces of $X_n$, and let
\[
    \widehat{\Phi}_{i;n,\sigma}
    \approx
    \int_{\sigma}
    \mathbf{F}(q(t_i,\mathbf{x}),t_i,\mathbf{x})\cdot \hat{\mathbf n}_{n,\sigma}
    \, \dd \Sigma
\]
be the approximation of the outward physical flux through the face $\sigma\in\mathcal{E}_n$ at time $t_i$. On an interior face $\sigma=\partial X_n\cap\partial X_k$, a conservative discretisation requires
\begin{equation}
    \widehat{\Phi}_{i;n,\sigma} = -\widehat{\Phi}_{i;k,\sigma} \, .
\end{equation}
For boundary faces, the numerical flux is instead determined consistently with the prescribed boundary condition. Its explicit expression depends on the PDE and on the chosen finite-volume scheme; for hyperbolic conservation laws, for example, it is commonly built from the states on the two sides of a face, often through a Riemann solver.

Finally, discretising time with a Forward Euler step, $t_i=t_0+i\Delta t$, gives the explicit update
\begin{equation}
    |X_n|\frac{\bar q_{i+1;\,n}-\bar  q_{i;\,n}}{\Delta t}
    + \sum_{\sigma\in\mathcal E_n} \widehat{\Phi}_{i;n,\sigma}
    = |X_n|\bar f_{i;\,n} \, , \quad  q_{i;\,n} \defeq q(t_i, \mathbf{x}_n), 
\end{equation}
and where
\[
    \bar  f_{i;\,n}  \defeq \frac{1}{|X_n|}\int_{X_n} f(t_i,\mathbf{x})\dd\omega_n \, .
\]
If a cell-centre quadrature rule is used, one may further approximate $\bar  f_{i;\,n}  \simeq f(t_i,\mathbf{x}_n)$. Equivalently, after division by $|X_n|$,
\[
   \frac{\bar q_{i+1;\,n}-\bar  q_{i;\,n}}{\Delta t}
    + \frac{1}{|X_n|}\sum_{\sigma\in\mathcal E_n} \widehat{\Phi}_{i;n,\sigma}
    = \bar  f_{i;\,n}  \, .
\]

\section{Meshless methods; Kansa's approach} \label{sec:2:4:Kansa}

Again, the same warning as before. So, for a more detailed introduction to the subject, see \cite{DeMarchi2018, Schaback2007, Fasshauer2014}, as well as the original paper by Kansa \cite{KANSA1990147}.

Before delving into the Kansa method, let us (very briefly) recall the \textit{spline} approximation of functions, and the \textit{radial basis} approach. Sometimes, we have to face the problem of building an \textit{unknown} function $f$ from a set of data (which is usually \textit{small}). Suppose that these available data consist of two sets: the \textit{collocation points} (or \textit{data sites}) $X=\{x_1,\ldots, x_N\}$, and the \textit{data values} $f_j = f(x_j), \forall j=1,\ldots,N$.  Our goal is to find a \textit{reconstructing} function, $s$; we can either \textit{interpolate} the data, i.e. $s(x_j)=f_j$, or \textit{approximate} the data, $s(x_j) \approx f_j$. Usually, the latter is more relevant, especially in statistical applications, since the data may contain \textit{noise}. 

For the 1D case (i.e., the \textit{univariate} setting), we can approximate with a spline; indeed, it is a well-known fact that a large data set is better dealt with splines than by polynomials. 
Furthermore, unlike a single global high-degree polynomial, splines allow the approximation to be refined locally by reducing the knot spacing; their approximation accuracy depends on both the polynomial degree and the distribution/spacing of the knots.

We can represent any spline via the so-called \textit{B-splines} (Basis Spline) \cite{de1978practical, Embree2016}; B-splines are defined \textit{recursively} from the constant B-spline $B_{i,0}$
\[
B_{i,0} (x) = \begin{cases}
    1, & t_i \le x\le t_{i+1} \,, \\
    0, & \text{otherwise},
\end{cases}
\]
via the recurrence relation
\[
    B_{i,k} (x) \defeq \frac{x - t_i}{t_{i+k} - t_i} \, B_{i,k-1} (x) + \frac{t_{i+k+1} - x}{t_{i+k+1} - t_{i+1}} \, B_{i+1, k-1} (x) \, .
\]
So, to interpolate a dataset $\{(x_j, f_k)\}_{j=1, \ldots, N}$ via a spline $S_k(x)$ (using degree $k$ / order $k+1$ B-spline), such that $S_k(x_j) = f_j, \; \forall j=1, \ldots,N$, we have
\begin{equation}
    S_k(x) = \sum_{j=-k}^{N} c_{j,k} B_{j,k} (x) \,.
\end{equation}
Notice that, by construction, $S_k \in C^{k-1}([x_1, x_N])$. 

\begin{figure}[t]
    \centering
    \includegraphics[width=0.75\linewidth]{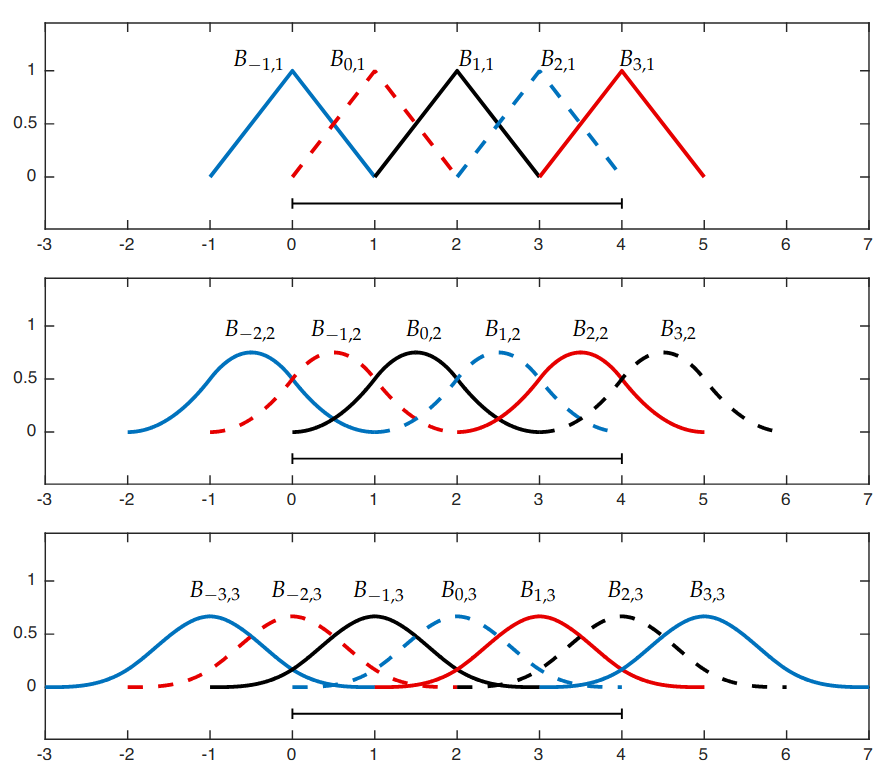}
    \caption{B-splines of degrees 1,2,3. (top, centre, bottom). From \cite{Embree2016}. }
    \label{fig:2:4:Bsplines}
\end{figure}

Notice that, to find the $\{c_{j,k}\}$ coefficients, we need to solve the algebraic problem
\[
    f_\ell = S_k(x_\ell) = \sum_{j=-k}^{N} c_{j,k} B_{j,k} (x_\ell) \,, \qquad \forall \ell =1, \ldots, N.
\]

Another relevant approximation type of approximation is the cubic spline $s : [a,b] \subset \RR \to \RR$ as\footnote{The notation $(\cdot)_+$ typically denotes the positive part: $(y)_+ = \max(0, y)$. However, for cubic splines, the truncated power function is usually written as $(x - x_j)_+^3 = \max(0, x - x_j)^3$.} 
\[
s(x) = \sum_{j=1}^N a_j (x -x_j)^3_+ + \sum_{k=0}^3 b_k x^k \,, \quad x\in[a,b].
\]

Nevertheless, the take away message is that \cite{DeMarchi2018}
\begin{proposition}
Every natural cubic spline \( s \) has the representation
\begin{equation} \label{eq:2:4:RBF1}
    s(x) = \sum_{j=1}^{N} a_j \phi(|x - x_j|) + p(x), \quad x \in \mathbb{R}
\end{equation}
where \( \phi(r) = r^3 \), \( r \geq 0 \), and \( p \in \mathcal{P}_1(\mathbb{R}) \), the space of polynomials of degree 1. The coefficients \( \{a_j\} \) have to satisfy the relations 
\begin{equation}\label{eq:2:4:RBFCond}
        \sum_{j=1}^N a_j = \sum_{j=1}^N a_j x_j = 0\,.
\end{equation}
On the other hand, for every set \( X = \{x_1, \ldots, x_N\} \subset \mathbb{R} \) of pairwise distinct points and for every \( f \in \mathbb{R}^N \), there exists a function \( s \) of the form \cref{eq:2:4:RBF1} satisfying \cref{eq:2:4:RBFCond}, that interpolates the data, i.e.\ \( s(x_j) = f(x_j), \quad 1 \leq j \leq N \).
\end{proposition}

The generalisation to functions of the form $\RR^d \to \RR$ is straightforward, and justifies where the name \textit{radial basis function} (RBF) becomes more evident \cite{buhmann2000radial}:
\begin{equation}
    s(x) = \sum_{j=1}^{N} a_j \phi(\|x - x_j\|_2) + p(x), \quad x \in \mathbb{R}^d \,,
\end{equation}
where $\phi : [0,\infty) \to \RR$ is a fixed univariate function and $p\in \mathcal{P}_{m-1}^d$ is a polynomial of degree at most $m-1$\note{Actually, depending on the chosen kernel, the polynomial term and corresponding side constraints may or may not be required \cite{buhmann2000radial}.}. 

The resulting interpolant is, up to a low-degree polynomial, a linear combination of shifts of a radial function.

Some relevant \textit{radial basis} are \cite{buhmann2000radial}
\begin{align}
    \phi (r) &= \exp(- \epsilon r^2) , \qquad \text{(gaussian)} \\
    \phi (r) &= \frac{1}{1 + \epsilon r^2} , \qquad \quad \text{(inverse quadratic)} \\
    \phi (r) &= \sqrt{1 + \epsilon r^2} , \qquad \text{(multiquadratic)}  \\ 
    \phi (r) &= r^2 \log r  , \qquad\quad \ \ \text{(thin-plate)}  \\ 
    \phi (r) &= \frac{1}{\sqrt{ 1 + \epsilon r^2} } . \qquad \ \text{(inverse multiquadratic)} 
\end{align}

\subsection{The Kansa approach: RBF for numerical PDE resolution}

We have now a way of interpolating, i.e. approximate, in a \textit{meshless} way, any function, via radial basis. 
We have seen that RBFs are obtained by composing a univariate radial profile with the distance from a centre. 

Let us now suppose we need to solve a Cauchy problem on a domain $\Omega\subset\RR^d$
\begin{align}
    L u(x) &= f(x), \qquad x\in \Omega \,, \\
    u(x) &= b(x), \qquad \,x\in \partial \Omega_D\,, \\
    \nabla_n u(x) &= v(x), \qquad  x\in \partial \Omega_N\,, \\
\end{align}
where $L$ is a linear operator acting on the function $u$, and where $\nabla_n u(x) \defeq  \hat{n} \cdot \nabla u(x)$ is the outward normal derivative at the boundary ($\hat{n}$ is the outward versor (unit vector) of $\partial \Omega$ in $x\in \partial \Omega$). 
Now, let us approximate the unknown function via a radial basis function on $N_c$ \textit{centres} $\{c_i\}_{i=1, \ldots, N_c}$:
\begin{equation}
    u(x) = \sum_{i=1}^{N_c} a_i \phi (r_i(x)) \,, \qquad r_i(x) \defeq \| x - c_i\|.
\end{equation}
We can now plug this in the PDE and compute \textit{analytically} the relation
\[
    L u(x) = L \sum_{i=1}^{N_c} a_i \phi (r_i(x)) = \sum_{i=1}^{N_c} a_i L \phi (r_i(x)) \defeq \sum_{i=1}^{N_c} a_i \ell (r_i(x)) ,
\]
where we have defined the \textit{explicitly, analytically} computable $ L \phi (r_i(x))$ as a convenient function $\ell$. Selecting now $N_I$ \textit{collocation points} $\{x_j\}_{j=1, \ldots , N_I}$, we can, once again, reduce the differential problem to an algebraic one
\begin{equation}
    L_x \Phi \cdot \mathbf{a} = \mathbf{f}\,, \qquad \text{ i.e. } \qquad  \sum_{i=1}^{N_c} (L_x\Phi)_{ji} a_i = f_j\,,\quad  \forall j=1, \ldots,N_I \,,
\end{equation}
where
\[
    (L_x\Phi)_{ji} = L_x \Phi_{ji} = L_x \phi(r_i(x_j)) \,, \qquad \Phi_{ji} \defeq \phi(r_i(x_j)) \,,
\]
and where we have defined 
\[ 
\mathbf{a} = (a_1, \ldots, a_{N_c})^T \,, \qquad \mathbf{f} = (f_1, \ldots, f_{N_I}) \,.
\]

We have additional constraints to satisfy, i.e. the boundary conditions; by selecting the \textit{boundary collocation points} $\{x_k\}_{k=1, \ldots, N_d} \subset \partial \Omega_D$, such that $\mathbf{b} = (b(x_1), \ldots, b(x_{N_d}))^T$, and similarly  $\{x_p\}_{p=1, \ldots, N_n}\in \partial \Omega_N$ such that $\mathbf{v}= (v(x_1), \ldots, v(x_{N_n}))^T$, we have two additional sets of constraints
\begin{align}
    \Phi \, \mathbf{a} &= \mathbf{b} \,, \qquad \text{ i.e. } \qquad  \sum_{i=1}^{N_c} \Phi_{ki} a_i = b_k\,,\qquad  \forall k=1, \ldots,N_d \,,  \\
    (\nabla_n \Phi) \, \mathbf{a} &= \mathbf{v} \,, \qquad \text{ i.e. } \qquad  \sum_{i=1}^{N_c} \nabla_n \Phi_{pi} a_i = v_p\,,\quad  \forall p=1, \ldots,N_n \,. 
\end{align}
Putting all together, we get a simple algebraic problem:
\begin{align}
    &\mathbf{M} \cdot \mathbf{a} = \mathbf{w} \quad  \Longrightarrow \sum_{i=1}^{N_c} M_{I i}\, a_i = w_I \, , \quad \forall I=1, \ldots, N_I + N_d + N_n\\
    &\mathbf{a} = a_i = (a_1, \ldots, a_{N_c})^T, \\
    & \mathbf{w} = w_{I = 1,\ldots, N_I + N_d + N_n} = (f_{j = 1,\ldots, N_I},\; b_{k = 1,\ldots, N_d},\; v_{p = 1,\ldots, N_n})^T, \\
    & \mathbf{M} = M_{I i} = \left[ (L_x \Phi_{j i})_{j = 1,\ldots, N_I},\; (\Phi_{k i})_{k = 1,\ldots, N_d},\; ( \nabla_n\Phi_{p i})_{p = 1,\ldots, N_n}
\right].
\end{align}

So...~is this the end of the story? Spoiler: NO. Finding $M^{-1}$ is \textit{hard}. Its well posedness is not a priori guaranteed\note{Actually, if it is \textit{not} a squared matrix, $M^{-1}$ does not even exists.}. 
Nevertheless, we don’t need to have an exact solvable system; we are interested in numerically viable solution. 
Something that minimises our error, or empirical risk, i.e.~we can recast the problem to an \textit{optimisation} problem 
\begin{equation}\label{eq:2:4:KansaLoss1}
    \mathbf{a}^* = \underset{\mathbf{a} \in \RR^{N_c}}{\text{argmin}} \; \mathcal{L}[a] \, , \qquad  \mathcal{L}[a]\defeq \text{MSE} [\mathbf{M} \cdot \mathbf{a} - \mathbf{w}] .
\end{equation}
We can also make the \textit{centres} learnable parameters in $\RR^d$, 
\begin{equation}
    \mathbf{a}^*, \mathbf{c}^* = \underset{\mathbf{a}, \mathbf{c}}{\text{argmin}} \; \mathcal{L}[a; c] \, , \qquad  \mathcal{L}[a; c]\defeq \text{MSE} [\mathbf{M}[\mathbf{c}] \cdot \mathbf{a} - \mathbf{w}] .
\end{equation}

Having set up our problem as an optimisation problem, we can employ our favourite optimisation scheme, e.g.~\textit{Stochastic Gradient Descent}, Newtonian methods, etc., and stop when a certain error threshold is reached. 

\begin{figure}[t]
    \centering
    \includegraphics[width=0.95\linewidth]{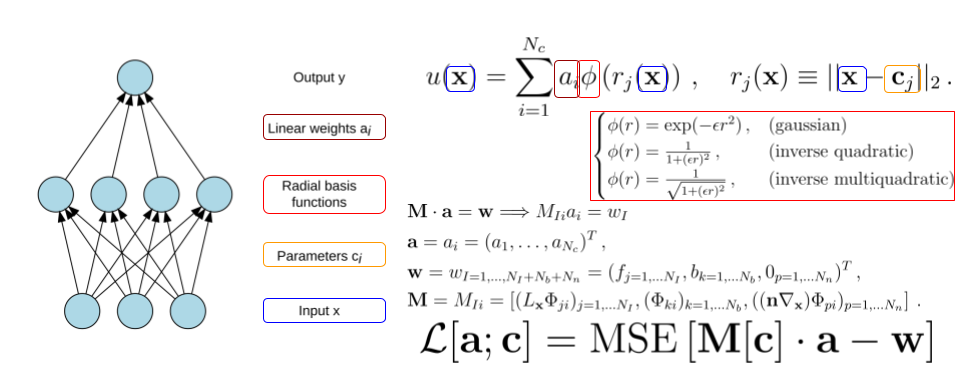}
    \caption{Visual representation of the Kansa method.}
    \label{fig:2:4:Kansa1}
\end{figure}

Let us stop here to think for a moment: what Kansa method, in the essence, is

\begin{recipe}{Kansa Method basic recipe}    \label{tbox:2:4:kansa_method}
\begin{itemize}
    \item[1.] Find a \textit{universal approximation} for any unknown function, here, the RBF. 
    \item[2.] Transform the Cauchy problem into an \textit{optimisation problem}, defining a certain error function(al) over an alternative representation of the differential problem (i.e., the algebraised representation). 
    \item[3.] Find the solution by solving the optimisation problem via our favourite optimisation algorithm, e.g., SGD.
\end{itemize}
\end{recipe}

\paragraph{An historical note:} In the classical square Kansa collocation method \cite{KANSA1990147} one chooses $N_{\mathrm{eq}}=N_c$ (often with centres located at the collocation sites) and solves the resulting linear system. One should not, in practice, form $\mathbf M^{-1}$ explicitly: the coefficients are obtained by solving $\mathbf M\mathbf a=\mathbf w$. The classical unsymmetric Kansa matrix is not guaranteed to be nonsingular for arbitrary choices of centres and collocation points, and its conditioning can also depend strongly on the kernel and its shape parameter \cite{Schaback2007, Fasshauer2014}. 

What we have described, i.e.,  \textit{overtest} the trial space, choosing $N_{\mathrm{eq}}>N_c$, and determine the coefficients through a linear least-squares problem, is a useful extension to the original Kansa method \cite{KANSA1990147}. We will still refer to it as the "Kansa method", with this historical understanding in place.

    \part{Physics Informed Neural Networks and Neural Operators}
    
    \chapter{Lecture 3: Introduction to Physics Informed Neural Networks} \label{chap:3}

Let us restart with our beloved differential problem
\begin{align}\label{eq:3:0:cauchy1}
    \mathcal{F}[u(x), x, \gamma] &= J(x) \,, \qquad x\in \Omega \,, \\
    \mathcal{B}[u(x), x] &= g(x) \,, \qquad x\in \partial \Omega \,, 
\end{align}
where we have explicit a few ingredients: 
\begin{itemize}
    \item \textbf{The Geometry}: the domain $\Omega$, its boundary $\partial \Omega$, and the behaviour of the field at the boundary;
    \item \textbf{The physics}: the source $J$, the dynamics $\mathcal{F}$, which may depend on a set of parameters $\gamma$.
\end{itemize}
Let us now introduce a bit of terminilogy here: 
\begin{definition}[Forward, inverse, parametric problems]
    We say we are solving a \textbf{forward problem} when the geometry and the physics are fixed, i.e., we have explicit $(\Omega, \mathcal{F}, \gamma)$, and we need to find the solution $u(x)$. We say we are solving an \textbf{inverse problem} when we have, as an additional set of informations, a set of \textit{observations} $\{(x_i, u_i^* \defeq u(x_i)\}_{i=1, \ldots, N}$, but we have no explicit knowledge on (some of) the physical parameters $\gamma$. We say we are solving a \textit{parametric problem} when we want to find a set of solutions $\{\gamma, u_\gamma\}$, or, more generally, a solution map $\gamma \to u_\gamma$, for the class of differential problems defined by $\mathcal{F}[\cdot,\cdot, \gamma]$.
\end{definition}

\section{Forward Problems: Vanilla PINNs}

We now suppose the reader knows what a Deep Neural Network (DNN) is, in particular, without loss of generality, what a Multi-Layer Perceptron (MLP) is\footnote{There is an incredible amount of fantastic books on the subject, e.g.~\cite{Goodfellow2016, prince2023understanding, scardapane2024alice, fleuret2024little, zhang2023dive, chollet2021deep, mehta2019high}.}. Nevertheless, to set up and standardise the notation, we say that a Deep Neural Network is a representative of the space of the functions satisfying the hypothesis of one of the \textit{Universal Approximation Theorems}. 

Now, suppose we need to approximate an unknown function $f : x \in \Omega \subseteq \RR^d \to \RR^n$ with one of those universally approximating functions $\DNN_\theta : x \mapsto \DNN_\theta (x)$, where $\theta\in \Theta$ are parameters, such that
\[
 \DNN_\theta (x) \approx f(x) .
\]
Usually this is done by solving an optimisation problem; we define a cost function (or loss function)
\[
\mathcal{L}[\theta] = \mathcal{L}[\DNN_\theta, f] \,,
\]
such that we define our optimal approximator as the one described by the optimal parameter set $\theta^*$
\[
\theta^* = \underset{\theta \in \Theta}{\text{argmin}} \; \mathcal{L}[\theta] \,.
\]
To achieve that, one commonly employs (a variation of) gradient descent, i.e.~an iterative process where we continuously update the parameter by a local exploration of the \textit{loss landscape}
\[
\theta \leftarrow\theta - \eta \nabla \mathcal{L}[\theta] \,.
\]
This is usually facilitated by the fact that the loss $\mathcal{L}[\theta]$ is a smooth function, with an explicit form for its derivatives.\note{We have seen in \cref{sec:1:4:MCI} that the loss function can be seen as a Monte Carlo approximation of the distance between the real function $f$, and the approximator $\DNN_\theta$. }
Also, $\DNN_\theta$ is itself a smooth function\note{Actually, that is not always true, since the smoothness of $\DNN_\theta$ may depends on the smoothness of the activation functions; nevertheless, the main comment holds: the derivate of $\DNN_\theta$ can be computed explicitly.}, arranged somehow in \textit{layers}, i.e. repeated composition of elemental functions, and, using repeatedly the formula for the derivative of a composite function, its derivative in each point can be \textit{analytically} \textit{explicitly} \textit{computed}.  

For example, the standard MLP is a composition of \textit{layers}, where each layer $\lambda_i : \RR^{m_i} \to \RR^{m_{i+1}}$ is the composition of a learnable \textit{affine transformation}  $x \mapsto W x + b$ and \textit{activation function} $\phi : \RR \to \RR$, whose derivative is explicitly, analytically known (in the sense of its analytical expression, $\phi'$):
\[
\lambda_i(\theta_i; x) = \phi(W_i x+b_i), \qquad \theta_i \defeq (W_i, b_i), \quad W_i \in \mathcal{M}(\RR^{m_i} \times \RR^{m_{i+1}} ), \; b_i \in \RR^{m_{i+1}} \,,
\]
so that a MLP is  \cite{scardapane2024alice, Goodfellow2016}
\begin{equation}
\begin{split}
    \DNN_\theta (x) &= \lambda_N(\theta_N) \circ \lambda_{N-1}(\theta_{N-1}) \circ \cdots \circ\lambda_1(\theta_1, x) \\
    &= \lambda_N (\theta_N , \lambda_{N-1}(\theta_{N-1}, \cdots\lambda_1(\theta_1, x)\cdots)) ,
\end{split}
\end{equation}
where $\theta = (\theta_1, \ldots, \theta_N)$. 
Since we can compute
\[
    \nabla_{\theta_i} \lambda_i(\theta_i; x) \defeq \begin{cases}
        \nabla_{W_i} \lambda_i(\theta_i; x) \,= \phi'(W_i x+b_i) \cdot x^T\\
        \nabla_{b_i} \lambda_i(\theta_i; x)  \;\; = \phi'(W_i x+b_i) , \\
    \end{cases}
\]
where we have used the rule of the derivative of a composed function, we can iterate over the MLP layers and easily compute the generic derivative $\nabla_{W_i, b_i} \DNN_\theta$, in a \textit{closed, analytical} form. 

But this means that we can compute also $\nabla_x \DNN_\theta(x)$ in  a \textit{closed, analytical} form, i.e., the derivative with respect to the input $x$. This is quite known and standard in the literature; for example \textbf{contractive autoencoders} \cite{Rifai2011} are denoising autoencoders with a penalisation term to the loss to reduce the variation under small variations of the input, which is the squared Frobenius norm (sum of squared elements) of the Jacobian matrix of partial derivatives associated with the encoder function $h(x)$, i.e. $\|\nabla_x h(x) \|_2$ \cite{Goodfellow2016}.

So this means that (almost) \textit{any} differential operator $\mathcal{F}$ acting on a DNN can be represented in a \textit{closed, analytical} form! so that, we can define a risk functional for our Cauchy problem in \cref{eq:3:0:cauchy1} where we are approximating the solution $u(x) \sim \DNN_\theta(x)$, simply by
\begin{equation}\label{eq:3:1:vanillapinnloss}
\begin{split}
    \mathcal{L}[\theta] &= 
    \| \mathcal{F}[\DNN_\theta (x), x, \gamma] - J(x)\|_{L^2(\Omega)}^2 + \lambda  \| \mathcal{B}[\DNN_\theta (x), x] - g(x)\|_{L^2(\partial\Omega)}^2 \\
    &= \int_\Omega \| \mathcal{F}[\DNN_\theta (x), x, \gamma] - J(x)\|^2_2 \dd \omega (x) \\
    & \qquad + \lambda \int_{\partial \Omega} \| \mathcal{B}[\DNN_\theta (x), x] - g(x) \|^2_2 \dd \Sigma(x) \\
    &\defeq \mathcal{L}_{PDE}[\theta] + \lambda \mathcal{L}_{BC}[\theta] 
\end{split}
\end{equation}
which, if we sample $N$ random points $\{x_i\}\in \Omega$, and $M$ random points in ${z_j} \in\partial \Omega$, can be computed via Monte Carlo approximation
\begin{equation}
    \mathcal{L}[\theta]  \approx \frac{|\Omega|}{N} \sum_{i=1}^N | \mathcal{F}[\DNN_\theta (x_i), x_i, \gamma] - J(x_i)|^2  +  \lambda\, \frac{ |\partial \Omega|}{M} \sum_{j=1}^M |\mathcal{B}[\DNN_\theta (z_j), z_j] - g(z_j)|^2 \,.
\end{equation}
This loss definition transforms our differential problem into an optimisation problem, i.e. the solution $u(x)$ of our differential problem \cref{eq:3:0:cauchy1} is approximated
\[
\begin{split}
    &u(x) \sim \DNN_{\theta^*} (x) \,, \\
    &\theta^* = \underset{\theta\in\Theta}{\text{argmin}} \, \Big[ \text{MSE}( \mathcal{F}[\DNN_\theta (x_i), x_i, \gamma] - J(x_i), 0) \\
    & \qquad  \qquad  \qquad  \qquad  + \lambda\,  \text{MSE}(\mathcal{B}[\DNN_\theta (z), z] - g(z), 0) \Big] .
\end{split}
\]
The optimisation problem can now solved using standard algorithms, like stochastic gradient descent; furthermore, (more or less) all the standard libraries used for training neural networks can be used, with very little fuzz, to solve this optimisation problem.

This framework for solving differential problem using deep neural networks is called Physics Informed Machine Learning; a DNN trained to solve a differential problem, is called (Vanilla) \textbf{Physics Informed Neural Network} (PINN).

We can summarise the basic recipe of a PINN to solve the differential problem \cref{eq:3:0:cauchy1}:

\begin{recipe}{Vanilla PINN basic recipe}    \label{tbox:3:1:vanilla_pinn}
\begin{itemize}
    \item[1.] Define a \textit{geometry sampler}, i.e. a method to draw $x_i \in \Omega$ (as well for boundary, initial surfaces, etc.).
    \item[2.] Define a Deep Neural Network to use as a function approximator (using your favourite framework, like PyTorch, TensorFlow, Jax, etc.), $\hat{u}_\theta(x) \leftarrow\DNN_\theta (x)$.
    \item[3.] Define the action of the differential operators on the DNN (e.g., using PyTorch autograd library), $\mathcal{F}[\hat{u}_\theta, x, \gamma], \mathcal{B}[\hat{u}_\theta (x), x]$.
    \item[4.] Choose an optimisation algorithm to solve the optimisation problem over the cost function defined in \cref{eq:3:1:vanillapinnloss}. 
\end{itemize}
\end{recipe}

\begin{figure}[t]
    \centering
    \includegraphics[width=0.95\linewidth]{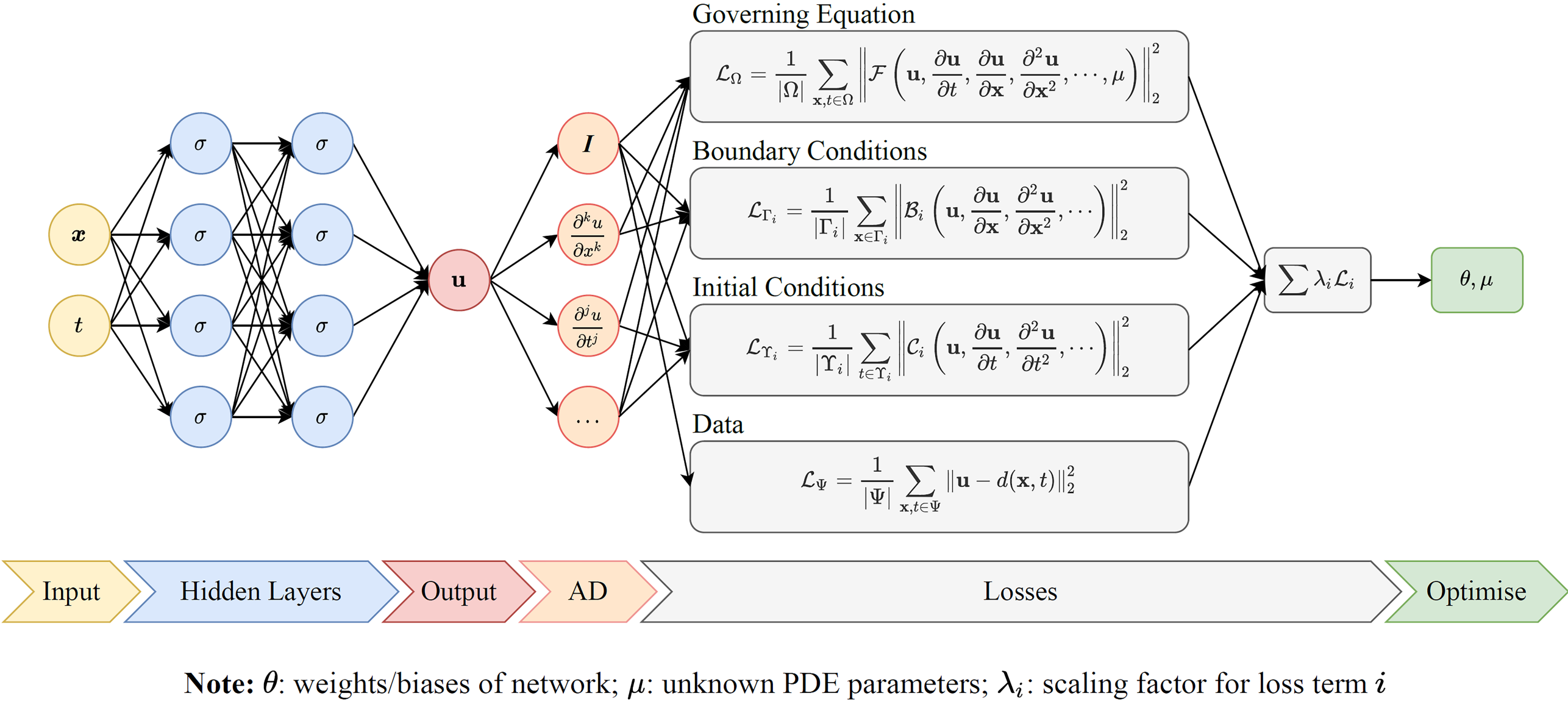}
    \caption{Visual representation of a PINN. From \href{https://kaufmann.ibk.ethz.ch/education/masterarbeiten/bischofr/_jcr_content/par/fullwidthimage/image.imageformat.1286.745131951.png}{here}.}
    \label{fig:3:1:PINN_visual_representation}
\end{figure}

A visual representation of a PINN is reported in \cref{fig:3:1:PINN_visual_representation}.

At this stage this should be understood as an algorithmic prescription: the trained network is a candidate approximation to the PDE solution.
The relation between discrete residual minimization and the actual solution error will be analysed in \cref{sec:3:3:Learning_Theory}.

\paragraph{A comment on losses - continuum-integral convention vs normalised-risk convention:} We have used here the continuum-integral convention, which brings, in its discretised form
\[
\mathcal{L}_\Omega = \int_\Omega \|r(x)\|_2^2 \dd \omega(x) \approx \frac{|\Omega|}{N} \sum_{i=1}^N |r(x_i)|^2 \,,
\]
where $r$ is the \textit{residual}; e.g., for the PDE term, $r(x) \defeq  \| \mathcal{F}[\DNN_\theta (x), x, \gamma] - J(x)\|^2_2$.
The usual optimisation problem, even for PINNs \cite{raissi2017physics}, uses the normalised-risk convention
\[
\mathcal{L}_\Omega = \mathbb{E}_{X\sim \mu_\Omega}\left[ r(X) \right] \approx \frac{1}{N} \sum_{i=1}^N |r(x_i)|^2 \,.
\]
Notice that the two formulations are equivalent, \textit{up to a relative factor} changing the relative weighting between the multi-objective losses:
\[
    \lambda_{CI} = \frac{|\Omega|}{|\partial\Omega|} \ \lambda_{NR} \,.
\]

\subsection{An historical note}

Before going on, let us make a brief historical note. One may wonder why \textit{such a simple} approach emerged only as recently as 2017. 

Indeed, the terminology \textit{Physics Informed Neural Network}  first appeared in a series of papers from 2017 by Maziar Raissi, Paris Perdikaris, George Em Karniadakis \cite{raissi2017physics,raissi2017physicsinformeddeeplearning,RAISSI2019686}, but the concept traces way back:
In 1990, Lee and Kang \cite{lee1990neural} presented a paper titled \textit{Neural algorithm for solving differential equations}, where they minimise FDM implicit relations by recasting the problem to an optimisation problem. 

In 1994, Dissanayake and Phan-Thien \cite{dissanayake1994neural} presented a paper with an abstract saying:

\quotespace
\begin{quote}
    \say{\textit{A numerical method, based on neural-network-based functions, for solving partial differential equations is reported in the paper. Using a ‘universal approximator’ based on a neural network and point collocation, the numerical problem of solving the partial differential equation is transformed to an unconstrained minimization problem.}}
\end{quote}
\quotespace

\noindent In 1998,  Lagaris, Likas, Fotiadis \cite{Lagaris98}, wrote in the abstract:

\quotespace
\begin{quote}
    \say{\textit{We present a method to solve initial and boundary value problems using artificial neural networks. [...] This part involves a feedforward neural network containing adjustable parameters (the weights)}.}
\end{quote}
\quotespace

\noindent Similarly, in 2001, Aarts \& Van Der Veer \cite{aarts2001neural} wrote in the abstract of paper:

\quotespace
\begin{quote}
    \say{\textit{A method is presented to solve partial differential equations (pde's) and its boundary and/or initial conditions by using neural networks. It uses the fact that multiple input, single output, single hidden layer feedforward networks with a linear output layer with no bias are capable of arbitrarily well approximating arbitrary functions and its derivatives.}}
\end{quote}
\quotespace

\noindent In 2011, a survey on such methods even appeared \cite{kumar2011multilayer}!

Furthermore, just few days \textit{prior} to the publication of the first PINNs paper, Berg and Nystr\"om \cite{berg2018unified} published a preprint on ArXiv titled ``A unified deep artificial neural network approach to partial differential equations in complex geometries'', where they used deep feed-forward networks, input derivatives and unconstrained gradient-based optimisation.

As usual, the success obtained by PINNs was that they were carefully branded to repack, reformulate, modernise, optimise, streamline, and simplify already known methods, uplifting them to be truly powerful. 

The novelty of PINNs was therefore not the invention of neural-network PDE solvers from scratch. Their impact came from synthesising earlier collocation ideas with modern deep architectures, automatic differentiation, scalable software frameworks, contemporary optimisation, and a unified treatment of forward, inverse and data-assisted problems.

This is somewhat similar to what happened with transformers and the attention mechanism\footnote{I would like to thank Prof.~Stefano Giagu for pointing this historical note about Attention Mechanism history out to me.} (for a nice historical review on Attention Mechanism, see \cite{soydaner2022attention}). The 2017 Transformer paper \cite{vaswani2017attention} triggered a revival because it reframed attention from an auxiliary component into the core computational engine of a model. Indeed, inceptions of similar ideas were as old as the early '90s, with the work on \textit{fast weights}, where a secondary mechanism dynamically modulates connections \cite{Schmidhuber92}. Earlier attention mechanisms (e.g., \cite{bahdanau2014neural}) were attached to recurrent or convolutional networks; the Transformer removed recurrence and convolution entirely, showing that self‑attention alone could model long‑range dependencies more efficiently and with superior performance. The paper’s clarity, empirical success, and architectural simplicity made attention not just a useful mechanism but the foundational primitive of contemporary deep learning. 

So, the early attempts to solve differential equations with neural networks, such as the works by \cite{Lagaris98, aarts2001neural}, introduced the essential idea that neural networks could approximate PDE solutions by embedding boundary conditions and minimizing physics‑related residuals. However, these methods remained isolated techniques, limited by the computational tools and modelling paradigms of their time. The 2017 introduction of Physics‑Informed Neural Networks (PINNs) by Raissi, Perdikaris, and Karniadakis played a role analogous to the paper ``Attention Is All You Need'' in the attention literature: it reframed a scattered set of earlier ideas into a unified and conceptually clean framework. PINNs demonstrated how automatic differentiation, modern optimization, and deep architectures could be systematically combined to enforce physical laws as soft constraints, transforming a niche methodology into a general paradigm for scientific machine learning. This synthesis, together with compelling demonstrations on forward, inverse, and data‑assimilation problems, catalysed the widespread adoption and rapid growth of PINNs in a way the earlier works could not.

\subsection{A brief comment on vanilla PINNs}

\begin{figure}[t]
    \centering
    \includegraphics[width=0.85\linewidth]{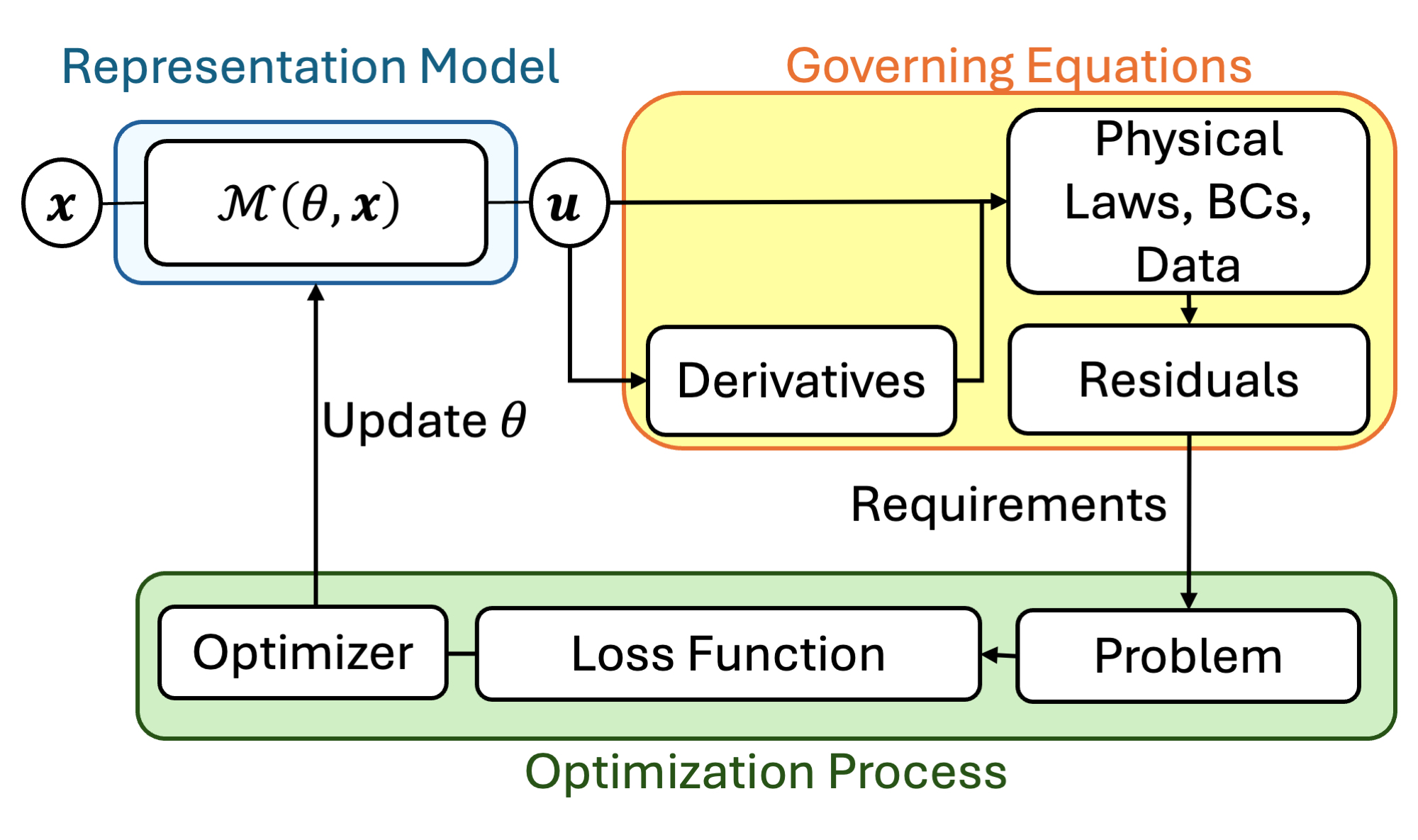}
    \caption{Visual representation of the PIML training components. A representation model provides an approximation $\hat u$ of the ODE/PDE solution $u$. The mismatch (residuals) between the approximated solution and the known components of the true solution, such as physical laws and boundary conditions, is calculated. These governing equations define a set of requirements that generate an optimization problem. This problem is reformulated as a loss function and then sent to an optimizer that updates the model parameters. Adapted from \cite{toscano2025pinns}.}
    \label{fig:3:1:PIML_overview}
\end{figure}

Let us stop here for a moment, again. Let us look again, closely, at the basic recipe of the Vanilla PINN \cref{tbox:3:1:vanilla_pinn}. The process we have described is essentially organised in three macro-blocks:
\begin{itemize}
    \item[1.] A Representation Model (e.g., the DNN)
    \item[2.] The Governing Equations (the Cauchy problem)
    \item[3.] An Optimisation process (e.g., the SGD)
\end{itemize}
At this moment, there is \textbf{no} data incorporation. Thus, strictly speaking, it is \textbf{not} machine learning. It is more similar to a \textit{meshless, numerical solver method}\note{With this strong wording we mean that, since there is no observational training data at this stage, a pure-forward PINN is algorithmically closer to a meshfree collocation solver than to conventional supervised data-driven learning. }. 

We have already encountered in \cref{sec:2:4:Kansa} a \textit{meshless, numerical solver method}, the Kansa Method \cref{tbox:2:4:kansa_method}. We have seen that we can use a radial basis function network  to solve a PDE problem, by minimising the loss using the Stochastic Gradient Descent (or any of its variant).

\begin{figure}[t]
    \centering
    \includegraphics[width=0.98\linewidth]{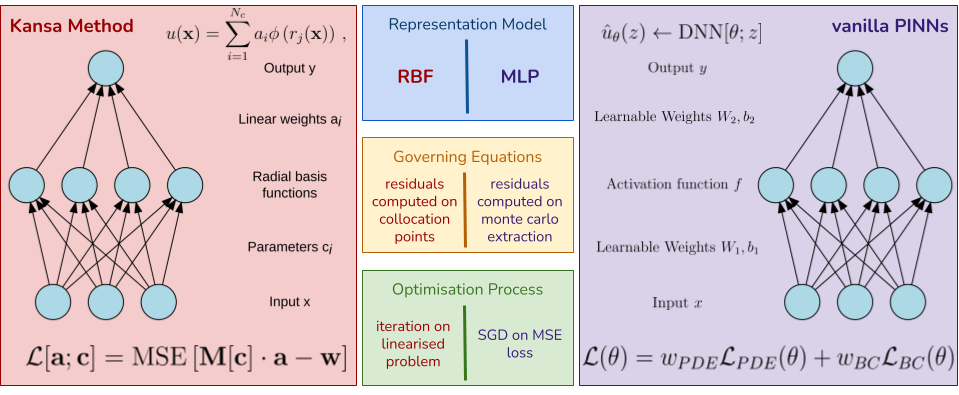}
    \caption{Visual comparison of the Kansa Method and the PINN method. The colour code of the centred box follows the one of \cref{fig:3:1:PIML_overview}.}
    \label{fig:3:1:KansaVsPINN}
\end{figure}

In \cref{fig:3:1:KansaVsPINN} we report a visual comparison of Kansa method \cref{tbox:2:4:kansa_method} and vanilla PINN method \cref{tbox:3:1:vanilla_pinn}. Let us observe them side-by-side, from the perspective of the three main ingredients above: 
\begin{itemize}
\item[1.]  \textbf{Representation Model}:
    \begin{itemize}
        \item[Kansa:] It uses the Radial Basis Function (RBF) as a representation of the solution, $\hat{u}_\theta (x) \sim \text{RBF}_\theta (x)$; \note{RBFs are local or semi‑global kernels with explicit smoothness.}
        \item[PINN:] It uses a Multi-Layer Perceptron (MLP) as a representation of the solution, $\hat{u}_\theta (x) \sim \DNN_\theta (x)$;\note{MLPs are global, compositional, adaptive basis functions. This helps explain why PINNs scale differently.}
    \end{itemize}
\item[2.]  \textbf{Governing Equations}:
    \begin{itemize}
        \item[Kansa:]  Classically, it leads to a dense linear system obtained by enforcing the PDE at collocation points. In modern reinterpretations, one may also minimise the residual via gradient‑based optimisation;
        \item[PINN:] It computes residuals on a Monte Carlo sampling of the domain;
    \end{itemize}
\item[3.]  \textbf{Optimisation Process}:
    \begin{itemize}
        \item[Kansa:]  At the vanilla level, it computes a linearised problem; more realistically, it computes SGD on the linear loss \cref{eq:2:4:KansaLoss1}.
        \item[PINN:] It uses SGD on the \textit{multi-objective} loss \cref{eq:3:1:vanillapinnloss}.
    \end{itemize}
\end{itemize}

To summarise, the structural similarity between the Kansa method and PINNs becomes evident when viewed through the lens of representation, physics enforcement, and optimisation. Both approaches construct a global trial function (RBF expansions in the Kansa method, and neural networks in PINNs) and enforce the governing equations at a set of collocation points. In the classical Kansa formulation, this yields a dense linear system derived from analytic derivatives of the RBFs, whereas PINNs rely on automatic differentiation and nonlinear optimisation to minimise a multi-objective residual loss. From this perspective, a vanilla PINN without data is not a machine-learning model but a modern meshless collocation method, differing from Kansa primarily in the choice of basis functions, the use of stochastic sampling, and the optimisation machinery (up to a certain point, as we have see in \cref{sec:2:4:Kansa}). This connection places vanilla PINNs within a long lineage of meshfree numerical solvers, rather than outside it.

\subsection{A \textit{key} comment on vanilla PINNs: strengths... and limits}

Before going on, let’s pause on Vanilla PINN for a moment. 

It seems that we have found the \textit{Holy Grail} of computational physics. We simply have to define a DNN, and compute the PDE (and BC/IC), and use it as Loss. \textit{That’s it}. 

Obviously, if life has a rule, it is:

\quotespace
\begin{quote}
    \say{\textit{There is no such thing as a free lunch} \cite{Heinlein66}\footnote{It was later popularised by Nobel-prize winner Milton Friedman \cite{friedman1975there}.}.}
\end{quote}
\quotespace

\noindent For example, in \cite{monaco2023training}, the authors have shown that 
\begin{itemize}
    \item (vanilla) PINNs are known to suffer from failures in some classes of problems.
    \item Literature provides many solutions to deal with these failures in specific contexts.
    \item The effectiveness of these strategies in different contexts \textit{is not always guaranteed}.
\end{itemize}

\noindent The main takeaway lessons are (in the author's opinion):
\begin{itemize}
    \item[1.] (vanilla) PINNs are an \textit{additional} tool to tackle the issue of solving differential problems. 
    \item [2.] (vanilla) PINNs have difficulties in performing the task (as the normal solver do, but in a different, novel way).
    \begin{itemize}
        \item This means that we need to address such difficulties in a \textit{different way} w.r.t. standard numerical solver;
        \item But we can tackle (most of) them using insights from \textit{Deep learning approaches}. 
    \end{itemize}
    \item[3.]  PINNS works better when they are additionally \textit{data-informed}. 
\end{itemize}

\begin{figure}[t]
    \centering
    \includegraphics[width=0.95\linewidth]{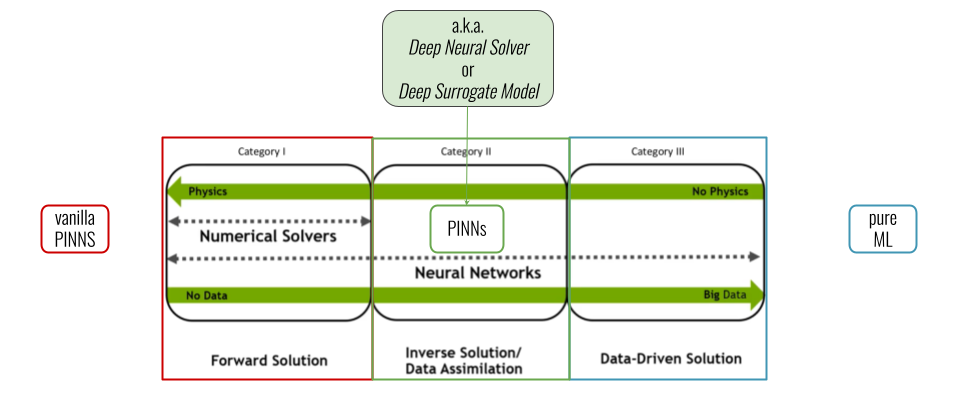}
    \caption{Visual representation of the Physics Informed Machine Learning landscape. Adapted from the NVIDIA Physics-Informed NeMo SYM documentation, which categorizes scientific machine learning approaches by their reliance on physics and data \cite{NvidiaPhysicsNemoDocumentation}.}
    \label{fig:3:1:PINN_landscape}
\end{figure}


In \cref{fig:3:1:PINN_landscape}, we present a schematic overview of the landscape of machine learning applications to physics-based systems, particularly differential problems, whether well-posed or not. Vanilla PINNs occupy the same corner as traditional numerical solvers: they rely solely on the governing equations, without incorporating observational data. In this regime, PINNs function as meshless solvers, leveraging neural networks as expressive trial spaces\note{The term \textit{expressive power} refers to the ability of neural networks to approximate complex, high-dimensional functions with intricate structure. This capacity stems from their compositional architecture and nonlinearity, which allow them to represent solution manifolds that would be intractable with classical basis functions.}. 

What distinguishes PINNs, however, is their capacity to transcend this regime. Neural networks offer remarkable expressive power and an inherent ability to model complex, high-dimensional data. This makes them particularly well-suited for tasks involving \textit{data assimilation} or \textit{inverse problems}, where observational data must be integrated into the solution process. In these contexts, PINNs (no longer vanilla) may offer practical advantages by integrating sparse observations and governing equations within a single differentiable framework. Their ability to extract latent structure from data while respecting physical laws positions them as a versatile and powerful tool within the broader PIML landscape.

\section{Inverse \& Parametric Problems}

\subsection{PINNs for inverse problems}

Let us return to \cref{eq:3:0:cauchy1}, 
\[
\begin{split}
    \mathcal{F}[u(x), x, \gamma] &= J(x) \,, \qquad x\in \Omega \,, \\
    \mathcal{B}[u(x), x] &= g(x) \,, \qquad x\in \partial \Omega \,, \\
\end{split}
\]
and suppose for now that the parameter(s) $\gamma$ are \textit{unknown}, but we have access to a set of \textit{observations}, i.e., a dataset $\{(x_i, u_i^*) \}_{i=1,\ldots,N_{\text{data}}}$. This setting defines a prototypical \textbf{inverse problem}, where the goal is to infer hidden parameters, coefficients, or source terms from partial observations of the solution field. More formally \cite{tarantola2005inverse, kaipio2005statistical}, 

\begin{definition}[Inverse Problem]
    Given a physical system described by a differential operator $\mathcal{F}[u(x), x, \gamma] = J(x)$, an inverse problem seeks to infer unknown parameters $\gamma$, source terms, or boundary conditions from partial observations of the solution $u(x)$, typically under constraints imposed by the governing equations. Unlike forward problems, which compute $u(x)$ from known $\gamma$, inverse problems are often ill-posed and require regularization or prior knowledge to ensure stability and uniqueness.
\end{definition}

Inverse problems are notoriously challenging because they often violate the classical notion of \emph{well-posedness}, introduced by Jacques Hadamard in the early 20th century \cite{Hadamard1902}: 
\begin{definition}[Well-posedness]
    A problem is said to be \emph{well-posed} if it satisfies three conditions:
    \begin{enumerate}
        \item \textbf{Existence}: a solution $u$ to the problem exists;
        \item \textbf{Uniqueness}: the solution is unique;
        \item \textbf{Stability}: the solution depends continuously on the data (small perturbations in the input produce small perturbations in the output).
    \end{enumerate}
\end{definition}

Inverse problems frequently violate one or more of these criteria. For instance, multiple parameter configurations may explain the same observations (non-uniqueness), or small measurement noise may lead to large deviations in the inferred parameters (instability). This intrinsic ill-posedness motivates the use of regularization strategies, prior knowledge, or physics-based constraints, such as those embedded in PINNs, to obtain meaningful and stable solutions.

Physics-Informed Neural Networks (PINNs) offer a natural framework for tackling such problems. By embedding the governing equations into the loss function and simultaneously fitting the observed data, PINNs enable the recovery of unknown physical quantities while respecting the underlying physics. This dual enforcement of data fidelity and physical consistency makes PINNs particularly attractive for inverse problems, especially in regimes where traditional solvers struggle due to ill-posedness, sparsity, or noise.
\note{Nevertheless, physics constraints can reduce the admissible solution space and provide useful regularisation, but they do not in general remove the intrinsic non-identifiability or instability of an inverse problem. Additional priors, regularisation, uncertainty modelling, or suitable observation design may still be required, especially when facing noisy data \cite{yang2021b}.}

Inverse problems addressed with PINNs span a wide range of applications; just to name a few examples:
\begin{itemize}
    \item \textbf{Petroleum engineering}: monitoring multiphase flow in reservoirs \cite{Calad2020,Calad2023}.
    
    \item \textbf{Quantum mechanics}: solving eigenvalue problems with unknown potentials; reconstruction of unknown potentials and eigenvalues \cite{Jin2022}.

    \item \textbf{Groundwater hydrology}: estimation of hydraulic conductivity fields in unsaturated flow models \cite{Depina2022}.
    
    \item \textbf{Structural mechanics}: identification of stiffness parameters in beam and rail systems \cite{Kapoor2023}.
    
    \item \textbf{Nano-optics and metamaterials}: recovery of effective permittivity and material parameters from scattering measurements \cite{Chen2020}.    
    
    \item \textbf{Fluid dynamics}: parameter estimation in Burgers, Euler, and Navier--Stokes equations \cite{raissi2017physicsinformeddeeplearning,Jagtap2022}.
    
    \item \textbf{Epidemiology}: inference of transmission rates and latent variables in compartmental models \cite{Patacchio2020}.

    \item \textbf{Battery modelling}: estimation of reaction kinetics and diffusion coefficients in electrochemical PDEs \cite{wang2024physics}.

\end{itemize}

A living repository of PINN-based inverse problem examples is maintained at \cite{ZhangGithubReview}, showcasing implementations across domains such as harmonic oscillators, reaction-diffusion systems, and parameter estimation in PDEs.

These diverse examples illustrate the flexibility of PINNs in inverse modeling: the neural network acts as a surrogate solution model, while the unknown parameters are treated as trainable variables. The optimisation process seeks to minimize both the PDE residuals and the mismatch with observed data, yielding physically consistent reconstructions even in challenging settings.

\begin{recipe}{Inverse Problem PINN basic recipe}    \label{tbox:3:1:inverse_pinn}
\begin{itemize}
    \item[1.] Define a \textit{geometry sampler}, i.e. a method to draw $x_i \in \Omega$.
    \item[2.] Define a Deep Neural Network to use as a function approximator \textbf{\& parameter estimator} (using your favourite framework, like PyTorch, TensorFlow, Jax, etc.), $\hat{u}_\theta(x), \textcolor{red}{\gamma} \leftarrow\DNN_\theta (x)$.
    \item[3.] Define the action of the differential operators on the DNN (e.g., using PyTorch autograd library), $\mathcal{F}[\hat{u}_\theta, x, \gamma], \mathcal{B}[\hat{u}_\theta (x), x]$.
    \item[4.] Choose an optimisation algorithm to solve the optimisation problem over the cost function defined in \cref{eq:3:1:vanillapinnloss}, plus an additional \textit{data loss} term:
    \begin{equation}
        \mathcal{L}_{data} [\theta] = \frac{1}{N_{data}} \sum_{i=1}^{N_{data}} \left| \hat{u}_\theta (x_i) - u_i^* \right|^2 .
    \end{equation}
\end{itemize}
\end{recipe}

A very simple implementation of a PINN for an inverse problem is simply to add a learnable parameter to the deep neural network modules (e.g., in PyTorch, using \textsc{nn.Parameter}). 

Notice that joint trainability does not imply identifiability: successful parameter recovery requires the observations and the governing equations to contain sufficient information to distinguish the unknown parameters.

\paragraph{A very simple example: PINNs for Burgers equation}

A simple example is the Burgers equation; a very nice set-up is the following Cauchy problem
\begin{align}
    \partial_t u(t,x) + \lambda u(t,x) \partial_x u(t,x) - \nu \partial_x^2 u(t,x) &= 0 \,, & (t,x) \in [0, 1]\times [-1, +1], \\
    u(t=0, x) &= - \sin (\pi x)\,, & x \in [-1, +1], \\
    u(t, x=\pm1)&= 0\, , &t\in[0,1].
\end{align}
which is the generalisation of the one reported in \cref{eq:2:1:Burgers_FDM}. We have already seen that the case where $\lambda=1, \nu=0.01/\pi$ is a nice set up (see \cref{fig:2:1:FDM_burgers}); furthermore, from there, we can have a \textit{dataset}, obtained from the FDM method, which we can employ to mimic an inverse problem: 

\noindent \textbf{Problem:} Given the data $\{(t_n, x_i), u^*_{ni}\defeq u(t_n, x_i)\}_{n=1, \ldots, N_t, \; i=1, \ldots N_x}$ (obtained from the FDM method), define a PINN to solve the inverse problem. 

To solve it, we simply need a basic DNN, a learnable parameter, and a method to compute the PDE operator on the DNN via automatic differentiation methods, see \cref{lst:3:2:Burgers_PINN_inverse}.

\begin{lstlisting}[caption={PINN sketch for Burgers inverse problem}, label={lst:3:2:Burgers_PINN_inverse}, language=Python]
import torch
from torch import nn
from .utils import MLP # standard MLP implementation

class BurgersInverse(nn.Module):
    def __init__(self):
        super().__init__()
        # Network
        self.dnn = MLP(
            n_inputs  = 2, # (t,x)
            n_outputs = 1, # (u)
            hidden_dims=list([32, 32, 16]),
            activation_func=nn.Tanh,
        )
        # learnable parameter
        self.lam = nn.Parameter(torch.tensor([0.8]), requires_grad=True) # requires_grad=True is Default
        self.nu  = nn.Parameter(torch.tensor([0.005]), requires_grad=True)

    def forward(self, coords: torch.Tensor):
        # coords of shape [N_points, 2], 
        # with t=coords[:, 0], x=coords[:, 1]
        u_pred = self.dnn(coords)
        return u_pred, self.lam, self.nu

# compute PDE 
def compute_burgers_pde(
    coords: torch.Tensor, 
    u: torch.Tensor,
    lam: torch.Tensor,
    nu: torch.Tensor
) -> torch.Tensor:
    """
    coords: (N, 2) tensor with columns [t, x]
    u:      (N, 1) predicted solution u(t,x)
    """
    # First derivatives wrt (t, x)
    grads = torch.autograd.grad(
        u, coords,
        grad_outputs=torch.ones_like(u),
        create_graph=True
    )[0]

    u_t = grads[:, 0:1]   # shape (N,1)
    u_x = grads[:, 1:2]

    # Second derivative wrt x
    grads2 = torch.autograd.grad(
        u_x, coords,
        grad_outputs=torch.ones_like(u_x),
        create_graph=True
    )[0]

    u_xx = grads2[:, 1:2]

    # 
    u2_x = torch.autograd.grad(
        u.pow(2), coords,
        grad_outputs=torch.ones_like(u),
        create_graph=True
    )[0][:, 1:2] / 2

    # PDE residual
    return u_t + lam.unsqueeze(0) * u2_x - (nu.unsqueeze(0)) * u_xx
\end{lstlisting}

\noindent Notice that in \cref{lst:3:2:Burgers_PINN_inverse} we have used the \textit{conservative} form of the Burgers equation, where we have simply used that
\[
 u\partial_xu = \frac{1}{2}\, \partial_x (u^2) \,.
\]

\subsection{PINNs for parametric problems}

\begin{figure}[t]
    \centering
    \includegraphics[width=0.85\linewidth]{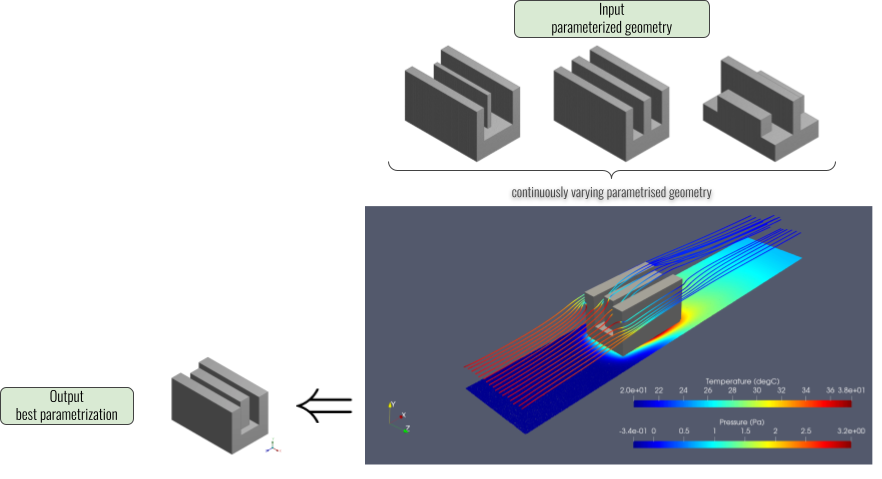}
    \caption{Visual example of parametrised geometries, and its connected optimisation task. }
    \label{fig:3:2:ParametricPINN_geom}
\end{figure}


In many practical scenarios, solving a single instance of a PDE is not sufficient. What we often require is a \emph{family} of solutions, smoothly indexed by one or more parameters~$\gamma$. This situation arises naturally in design, control, uncertainty quantification, and inverse modelling. Similarly to the inverse setting, and in close analogy with the vanilla PINN formulation, we can extend the basic PINN recipe so that the neural network does not approximate a single function $u(t,x)$, but rather a \emph{solution generator}
\[
(t,x,\gamma) \;\mapsto\; u_\gamma(t,x),
\]
capable of producing the solution for any admissible value of the parameter~$\gamma$.

This parametric viewpoint is especially valuable when the dependence on $\gamma$ cannot be conveniently embedded into the loss function or optimised through gradient descent---for example, when $\gamma$ is discrete, non-differentiable, or when the downstream objective is itself non-smooth. By learning a surrogate model that spans the entire parameter space, we gain the ability to explore, optimise, or control the system without repeatedly solving the PDE from scratch.

Although in \cref{eq:3:0:cauchy1} the parameter $\gamma$ appears explicitly only inside the differential operator $\mathcal{F}$, in practice parametric problems arise in several distinct forms. It is therefore useful to classify them according to where the parameter enters the Cauchy problem:
\begin{itemize}
    \item[A.] \textbf{Parametric geometries}: The parameter modifies the \emph{domain} itself (e.g., shape optimisation, moving boundaries, geometry-dependent coefficients).
    \item[B.] \textbf{Parametric PDEs}: The parameter indexes a \emph{family of operators} or physical regimes (e.g., varying diffusivity, Reynolds number, material properties).
    \item[C.] \textbf{Parametric BC/IC}: The parameter controls the \emph{boundary or initial conditions}, effectively generating a family of admissible solutions.
\end{itemize}

Each of these cases requires slightly different architectural choices and sampling strategies, but the overarching idea remains the same: we embed $\gamma$ directly into the neural network input, and train a single model that captures the entire solution manifold. Parametric PINNs thus provide a unified framework for learning high-dimensional, continuous families of PDE solutions in a single training phase, enabling fast evaluation and flexible downstream optimisation.

\paragraph{An example of parametric geometry}

In \cref{fig:3:2:ParametricPINN_geom} we show a typical \emph{design optimisation problem}, taken from \cite{NvidiaPhysicsNemoDocumentation}, namely the \emph{Parametrised 3D Heat Sink} example\footnote{See \url{https://docs.nvidia.com/physicsnemo/latest/physicsnemo-sym/user_guide/advanced/parametrized_simulations.html}. Accessed \today.}. The key idea is that, once the training is complete, the network can be evaluated for many different values of the geometric parameter $\gamma$ as a post-processing step, without solving the forward problem again. The parametrisation increases the computational cost only marginally, while covering the entire design space in a single training phase. 
In this particular benchmark, the authors report that a single parameterised neural solver substantially reduces the total cost of exploring the chosen design space relative to the corresponding collection of OpenFOAM runs.

In this example the goal is to solve a conjugate heat-transfer problem coupling the fluid flow and thermal fields for a family of heat sink geometries. The parameters describe the fin dimensions (thickness, length, and height), so that the design space is generated by varying the height $h$, length $\ell$, and thickness $t$ of the central fin and the two side fins (see again \cref{fig:3:2:ParametricPINN_geom}). A parametric PINN learns the mapping
\[
(\mathbf{x}, \gamma) \longmapsto T_{\gamma}(\mathbf{x}) \, ,
\]
where $\gamma$ encodes the geometric configuration.

Once the model is trained, it becomes possible to perform design optimisation directly at inference time. A typical optimisation problem specifies an objective function that is minimised or maximised under physical or engineering constraints. For heat sink design, a common constraint is the maximum allowable temperature at the source chip, which is dictated by the chip’s operating limits. Another constraint is the maximum pressure drop that the cooling system can sustain while driving the flow around the heat sink. Parametric PINNs allow the entire optimisation loop to be carried out efficiently, since evaluating new geometries does not require solving the PDE again.

\paragraph{An example of a parametric PDE: stationary wave propagation and the Helmholtz equation}

A simple yet representative example of a parametric PDE is the Helmholtz equation, which describes stationary wave propagation. In this setting the wave-number $k$ plays the role of a free parameter:
\[
\nabla^{2} f(\mathbf{x}) + k^{2} f(\mathbf{x}) = 0 \, .
\]

For each admissible value of $k$ we obtain a different PDE, and the goal of a parametric PINN is to learn the mapping
\[
(\mathbf{x}, k) \longmapsto f_{k}(\mathbf{x}) \, ,
\]
that is, a generator of solutions for the entire family of Helmholtz problems.

The Helmholtz equation arises naturally from the standard wave equation (for a field component)
\[
\square E(t,\mathbf{x}) \defeq \partial_{t}^{2} E(t,\mathbf{x}) - c^{2} \nabla^{2} E(t,\mathbf{x}) = 0 \, ,
\]
which models, for example, the propagation of the electromagnetic field in a waveguide. In the stationary regime we look for solutions of the form
\[
E(t,\mathbf{x}) = f(\mathbf{x}) \, \mathrm{e}^{\mathrm{i}\omega t} \, ,
\qquad
k^{2} \defeq \frac{\omega^{2}}{c^{2}} \, ,
\]
and by substituting this ansatz into the wave equation we obtain the Helmholtz equation for the spatial profile $f(\mathbf{x})$. Different values of the frequency $\omega$ correspond to different values of the parameter $k$, which makes this a natural example of a parametric PDE.

Notice that the two cases are \textit{hugely} different; indeed, the latter is an operator map, since we need to approximate with our beloved neural network not a family of functions, but instead an entire \textit{operator} mapping $f \to G[f]$.

\paragraph{An example of parametric BC/IC: the vibrating string with a loose end}

A natural example of a parametric boundary condition arises from the vibrating string model discussed in \cref{subsec:1:2:1:Loose_String}. In that setting the string satisfies the 1+1-dimensional wave equation, and the behaviour at the right endpoint is controlled by a Robin condition of the form
\[
u(t,1) + \alpha\, \partial_{x} u(t,1) = 0 \, ,
\]
where the parameter $\alpha$ describes how tightly the string is held by the peg. The case $\alpha = 0$ corresponds to a perfectly fixed end, while increasing $\alpha$ models a progressively looser constraint. For each value of $\alpha$ the PDE has a different family of eigenvalues $\{\lambda_{n}(\alpha)\}$, and therefore a different vibration spectrum.

This can be recast as a typical \emph{parametric BC} problem: the differential operator is fixed\note{We could actually treat also the wave velocity as a parameter.}, but the admissible solutions depend on a parameter that enters through the boundary condition. A parametric PINN can be trained to learn the mapping
\[
(t,x,\alpha) \longmapsto u_{\alpha}(t,x) \, ,
\]
so that, once training is complete, the model can predict the vibration pattern of the string for any value of the looseness parameter $\alpha$ without solving the PDE again.

A similar situation occurs for \emph{parametric initial conditions}. In the same example the pluck profile $f(x)$ determines the coefficients of the modal expansion. If we regard $f$ as a parameter, the task becomes learning the operator
\[
f \longmapsto u_{f}(t,x) \, ,
\]
that is, the solution corresponding to a given initial displacement. Parametric PINNs provide a unified framework for both types of dependence, since the parameters (either $\alpha$ or a representation of $f$) can be included directly in the network input.

\paragraph{Connection with Neural Operators.}

Parametric PINNs naturally point toward a broader and more rigorous framework, namely the theory of Neural Operators. While a parametric PINN learns a map $(x,t,\gamma)\mapsto u_\gamma(t,x)$, a Neural Operator is designed to learn a map between infinite-dimensional function spaces, for example from a coefficient field or forcing term to the corresponding solution of a PDE. We will discuss Neural Operators in \cref{chap:5:NeuralOperators}. 
In this sense\note{More precisely, Parametric PINNs and Neural Operators address related multi-query PDE settings, but they represent different modelling paradigms. A parametric PINN usually approximates a finite-dimensional parameter-to-solution map through a coordinate network, whereas a Neural Operator is explicitly designed to approximate maps between function spaces.}, Neural Operators provide a mathematically principled generalisation of the parametric PINN idea, since they treat the parameter not as a finite-dimensional vector but as an entire input function. This viewpoint clarifies the role of parametric dependence and offers a systematic way to learn solution operators for families of PDEs. Recent work explicitly analyses the relationship between PINNs and Neural Operators in the context of parametric PDEs, showing how both approaches can be interpreted as operator-learning strategies with different inductive biases and training regimes\note{For a detailed discussion of this connection, see for example the analysis in \cite{lu2021learning, wang2021learning, kovachki2023neural, li2024physics, zhang2025pinnsNOs}.}.

\section{A brief introduction to the Learning Theory of PINNs} \label{sec:3:3:Learning_Theory}

Having introduced the basic formulations of PINNs, we now examine their theoretical foundations, focusing on the sources of error and the dynamics of training, and their theoretical bounds.

The next section follows the theoretical framework developed in \cite{cuomo2022scientific, toscano2025pinns, de2021approximation, de2022error, kutyniok2022mathematics, de2024numerical, khanra2026physics}.

Physics-Informed Neural Networks (PINNs) are trained by minimizing a loss function that enforces the differential equation, the boundary conditions, and any additional physical constraints. From a learning-theoretic perspective, this raises a natural question: under which conditions does a PINN converge to the true solution of the Cauchy problem? In other words, when does the trained predictor $u_{\theta}$ approximate the exact solution $u$ of the PDE?

This question is subtle, because PINNs do not follow the classical supervised learning paradigm. In standard machine learning the loss is defined on a finite dataset, and the goal is to generalize from the training samples to unseen data. In PINNs the loss is defined by the physics itself, through the residual of the PDE and the boundary or initial conditions. The training points are not given by a dataset, but are chosen by the practitioner, and the loss approximates a continuum functional through numerical quadrature, making the estimation error fundamentally a quadrature error rather than a statistical generalization error. 

\begin{figure}[t]
    \centering
    \includegraphics[width=0.95\linewidth]{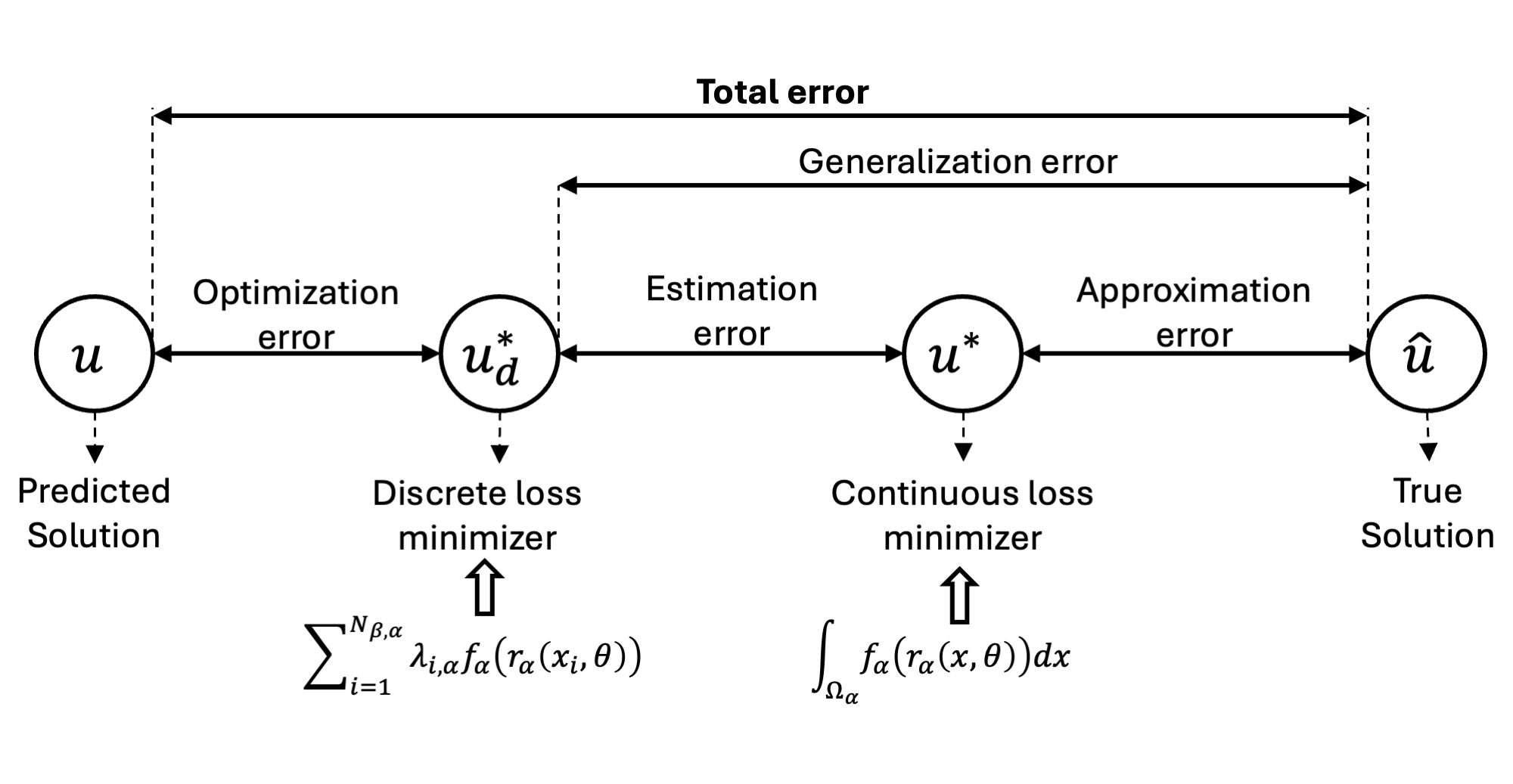}
    \caption{Decomposition of total error. The total error in a PIML model can be decomposed into three parts: (1) an optimization error resulting from the optimizer’s inability to fully minimize the loss function, (2) an estimation or quadrature error due to discretizing the loss function over a finite number of points, and (3) an approximation error, which reflects the expressivity or capacity of the representation model to capture the true solution. Adapted from \cite{toscano2025pinns}}
    \label{fig:3:3Errors_PIML}
\end{figure}

For this reason, the learning theory of PINNs must account for three distinct sources of error (for a visual representation, see \cref{fig:3:3Errors_PIML}):
\begin{itemize}
    \item the \emph{approximation error}, which measures the expressive power of the neural network class;
    \item the \emph{estimation error}\note{We use the term 'estimation error' (also called 'quadrature error' in the PINN literature).}, which arises from discretizing the continuous loss functional over a finite set of collocation points;
    \item the \emph{optimization error}, which reflects the fact that the PINN loss is non-convex and the optimizer may not reach a global minimizer.
\end{itemize}

These three contributions combine to determine the total error between the trained model $u_{\hat{\theta}}$ and the true solution $u$. Understanding their role is essential for explaining both the successes and the limitations of PINNs, and provides a principled framework for improving architectures, sampling strategies, and optimization schemes.

\subsection{The PINN Learning Problem as Risk Minimization}

The training objective of a Physics-Informed Neural Network can be interpreted as the minimization of a suitable risk functional. Given a PDE of the form
\[
\mathcal{F}[u(x), x] = J(x) \qquad x \in \Omega ,
\]
together with boundary or initial conditions, the PINN loss is constructed by evaluating the residual of the PDE and the constraints at a set of collocation points. This leads to a discrete loss of the form
\[
\hat{\mathcal{R}}[u_\theta] = \sum_{\alpha} \sum_{i=1}^{N_\alpha} \lambda_{i,\alpha}\, f_\alpha\bigl(\tau_\alpha(x_i,\theta)\bigr),
\]
where the index $\alpha$ runs over the different components of the loss (PDE residual, boundary conditions, initial conditions, and so on), the points $x_i$ are the collocation nodes, where $\tau_\alpha$ denotes the residual of constraint $\alpha$,  the functions $f_\alpha$ encode the contribution of each constraint, and $\lambda_{i,\alpha}$ are quadrature weights.

From a continuum perspective, the ideal objective is the minimization of the continuous risk
\[
\mathcal{R}[u_\theta] = \sum_{\alpha} \int_{\Omega_\alpha} f_\alpha\bigl(\tau_\alpha(x,\theta)\bigr)\, \dd x ,
\]
which measures how well the neural network satisfies the PDE and the associated constraints over the entire domain. The discrete loss $\hat{\mathcal{R}}$ is therefore a numerical quadrature approximation of the continuous functional $\mathcal{R}$.

This viewpoint highlights an important difference between PINNs and classical supervised learning. In standard machine learning the empirical risk is defined on a fixed dataset, and generalization refers to performance on \textit{unseen data} drawn from the same distribution. In PINNs the training points are chosen by the practitioner, and the empirical loss approximates a continuum integral. The discrepancy between $\mathcal{R}$ and $\hat{\mathcal{R}}$ is therefore more related to a quadrature error rather than a statistical generalization error. This discrepancy is fundamentally a \textit{quadrature error} (due to numerical integration) rather than a \textit{statistical generalization error} (due to finite data sampling), which distinguishes PINNs from standard supervised learning.

The minimization of $\hat{\mathcal{R}}$ is performed by gradient-based optimization, which produces a trained parameter vector $\hat{\theta}$. The quality of the resulting predictor $u_{\hat{\theta}}$ depends on three factors: the expressivity of the neural network class, the accuracy of the quadrature approximation, and the ability of the optimizer to reach a good minimum of the discrete loss. These three aspects form the basis of the error decomposition discussed in the following subsections.

\subsection{Approximation Error}

The first source of error in a PINN is the \emph{approximation error}, which measures the expressive power of the neural network class used to represent the solution. Given the continuous risk functional $\mathcal{R}$, the approximation error is defined as\note{In the generic learning decomposition \cite{kutyniok2022mathematics}, the approximation error does not assume that the exact solutions satisfies all residuals:\[
     \mathcal{E}_A = \inf_{\theta \in \Theta} \mathcal{R}[u_\theta] - \mathcal{R} ,
     \] where \[
     \mathcal{R}^* \defeq \inf_{v\in \mathcal{X}} \mathcal{R} [v] \,.
\]}
\begin{equation}
    \mathcal{E}_A = \inf_{\theta \in \Theta} \mathcal{R}[u_\theta] ,
\end{equation}
that is, the smallest possible value of the loss over all neural networks in the chosen architecture. This quantity reflects how well the network class can represent functions that satisfy the PDE and the associated constraints.

Approximation theory for PINNs relies on classical results for neural networks in Sobolev spaces. A key result \cite{de2021approximation, de2022error}, shows that shallow networks with $\tanh$ activation can approximate functions and their derivatives with controlled accuracy. More precisely, for a target function $u \in W^{s,\infty}([0,1]^d)$, a two-hidden layers network with  width $O(N^d)$ satisfies the bound
\[
\| u_{\theta_N} - u \|_{W^{k,\infty}} \le c_1 \frac{\log(c_2 N)^k}{N^{s-k}},
\qquad k = 0, \ldots, s-1 ,
\]
for suitable constants $c_1$ and $c_2$, assuming the target function is $W^{s, \infty}$; this means that if we have a PDE of order $m$, we need that our DNN is at least $W^{m,\infty}$. This estimate shows that neural networks can approximate smooth functions with algebraic convergence rates, and that the approximation improves as the width increases.

In the context of PINNs, the approximation error quantifies the ability of the network to represent a function that satisfies the PDE residual and the boundary or initial conditions. Since the loss involves derivatives of $u_\theta$, the relevant function space is typically a Sobolev space of order one or higher. The approximation error therefore depends on the smoothness of the true solution, the choice of activation function, and the architecture of the network.

Several influential studies indicate that, for many standard PINN benchmarks, optimisation rather than representational capacity can be the dominant bottleneck \cite{wang2021understanding, krishnapriyan2021characterizing}.
Modern neural networks are highly expressive, and even moderately sized architectures can approximate smooth PDE solutions with good accuracy. However, the approximation error becomes relevant when the solution has sharp gradients, discontinuities, or multi-scale features, or when the PDE involves high-order derivatives. In such cases, architectural choices such as depth, width, activation functions, and positional encodings can significantly influence the expressive power of the model. We will see the impact of this fact in the following sections.

\subsection{Estimation (Quadrature) Error}

The second contribution to the total error in a PINN arises from the discrepancy between the continuous risk $\mathcal{R}$ and its discrete approximation $\hat{\mathcal{R}}$. This is the \emph{estimation error}, defined as
\begin{equation}
    \mathcal{E}_G = \sup_{\theta \in \Theta} \left| \mathcal{R}[u_\theta] - \hat{\mathcal{R}}[u_\theta] \right|.
\end{equation}

In classical supervised learning this quantity is interpreted as a generalization error, since the empirical risk is computed on a finite dataset and the continuous risk corresponds to the expected loss over the data distribution. In PINNs the situation is different. The training points are not drawn from a distribution, but are chosen by the practitioner, and the discrete loss is a numerical quadrature approximation of a continuum integral. The estimation error is therefore a quadrature error rather than a statistical generalization error.

The magnitude of $\mathcal{E}_G$ depends on the sampling strategy used to select the collocation points. Uniform sampling, Latin hypercube sampling, adaptive sampling, and residual-based refinement all lead to different quadrature accuracies. In high-dimensional problems the estimation error can dominate the total error, since accurate quadrature becomes challenging. However, for many PDEs the structure of the residual and the regularity of the solution mitigate the curse of dimensionality.

A rigorous analysis of the estimation error has been obtained for certain classes of PDEs. In particular, De Ryck and Mishra \cite{de2022error} proved that for Kolmogorov-type equations, such as the heat equation or the Black–Scholes equation, the following bound holds:
\[
\mathcal{R}[u_\theta] \le \left( C\, \mathcal{R}[u] + \mathcal{O}(N^{-1/2}) \right)^{1/2},
\]
where $N$ is the number of collocation points, where $C$ depends on the PDE coefficients and domain geometry, and the $\mathcal{O}(N^{-1/2})$ term reflects Monte Carlo quadrature error.. This result shows that small training error implies small continuous risk, and that the estimation error decays at a rate comparable to Monte Carlo quadrature. Remarkably, the bound does not depend on the dimension of the problem (at least for Kolmogorov-type equations), which suggests that PINNs can generalize well for certain PDEs even in high-dimensional settings.

In practice, the estimation error is often a critical factor in the performance of PINNs. If the collocation points do not adequately resolve the regions where the PDE residual is large, the discrete loss may underestimate the true error. This explains why adaptive sampling and residual-based refinement can significantly improve the accuracy of PINNs. The estimation error therefore reflects the interplay between numerical quadrature, PDE structure, and the sampling strategy used during training.

\subsection{Optimization Error}

The third contribution to the total error in a PINN is the \emph{optimization error}, which reflects the fact that the discrete loss $\hat{\mathcal{R}}$ is minimized by a gradient-based algorithm that may not reach a global minimizer. The optimization error is defined as
\begin{equation}
    \mathcal{E}_O = \hat{\mathcal{R}}[u_{\hat{\theta}}] - \inf_{\theta \in \Theta} \hat{\mathcal{R}}[u_\theta],
\end{equation}
where $\hat{\theta}$ denotes the parameter vector obtained after training. Since the PINN loss is highly non convex, the optimizer may converge to a local minimum, a saddle point, or a flat region of the loss landscape.

The optimization landscape of PINNs is significantly \textit{more complex} than that of standard supervised learning models. The loss involves derivatives of the neural network, and the PDE residual may contain stiff terms, multi-scale features, or high-order operators. These factors create sharp valleys, flat plateaus, and regions where gradients vanish or explode. As a consequence, \textit{the optimization error can dominate the total error}, especially in problems with strong nonlinearities or multi-scale behaviour \cite{wang2021understanding, krishnapriyan2021characterizing}.

Empirical evidence suggests that the choice of optimizer plays a crucial role \cite{toscano2025pinns}. First-order methods such as Adam are effective in the early stages of training, where they rapidly reduce the residual. Second-order methods such as L-BFGS often perform better in the later stages, where they refine the solution and reduce the stiffness of the residual \cite{jin2021nsfnets, urban2025unveiling, wang2022and}. Hybrid strategies that combine both methods are widely used in practice.

Despite these empirical successes, the theoretical understanding of optimization in PINNs remains limited. The non-convexity of finite-width PINN training prevents general, architecture-independent global convergence guarantees, and the interaction between the PDE structure and the optimization dynamics is not fully understood. Recent studies indicate that the optimization trajectory may exhibit distinct phases, with an initial period of rapid error reduction followed by a slower regime where the network adjusts to satisfy the PDE constraints more precisely \cite{toscano2025pinns}. These observations motivate the study of optimization dynamics and information-theoretic perspectives, which will be discussed later on.

\subsection{Total Error Decomposition}

The three sources of error described above combine to determine the total discrepancy between the trained predictor $u_{\hat{\theta}}$ and the true solution $u$ of the PDE. A central result in the learning theory of PINNs is the following inequality, which bounds the continuous risk of the trained model \cite{kutyniok2022mathematics}:
\begin{equation}
    \mathcal{R}[u_{\hat{\theta}}] \le \mathcal{E}_O + 2\,\mathcal{E}_G + \mathcal{E}_A .
\end{equation}
The factor of 2 arises from applying the triangle inequality to bound the difference between continuous and discrete risks.

This decomposition, introduced in \cite{kutyniok2022mathematics}, separates the contributions of optimization, estimation, and approximation. Each term reflects a different aspect of the learning process:
\begin{itemize}
    \item $\mathcal{E}_A$ measures the expressive power of the neural network architecture. It depends on depth, width, activation functions, and the regularity of the true solution \cite{de2021approximation}.
    \item $\mathcal{E}_G$ quantifies the quadrature error introduced by discretizing the continuous loss. It depends on the sampling strategy, the number of collocation points, and the structure of the PDE \cite{de2022error}.
    \item $\mathcal{E}_O$ captures the limitations of gradient-based optimization. It depends on the optimizer, initialization, and the geometry of the loss landscape \cite{wang2021understanding,krishnapriyan2021characterizing}.
\end{itemize}

This decomposition provides a conceptual framework for understanding both the strengths and the weaknesses of PINNs. For example, if the sampling is insufficient, the estimation error dominates and the network may satisfy the discrete residual while violating the PDE in regions not covered by collocation points. If the optimizer becomes trapped in a poor local minimum, the optimization error dominates. If the architecture is not expressive enough, the approximation error becomes the limiting factor.

In practice, the three errors interact in non-trivial ways. Improving the sampling strategy may reduce the estimation error but increase the stiffness of the optimization landscape. Increasing the network size may reduce the approximation error but make optimization more difficult.

\subsection{Training Dynamics of PINNs: connection to Optimization Dynamics and Information Flow} 

The error decomposition provides a static view of the learning process, but PINN training is inherently dynamic. The evolution of the parameters $\theta(t)$ under gradient-based optimization plays a central role in determining which error component dominates at different stages of training. Recent empirical studies show that PINNs often exhibit distinct learning phases \cite{anagnostopoulos2025learning,toscano2025pinns}, similar to those observed in deep supervised networks \cite{tishby2015deep,shwartz2017opening,westphal2025generalized}  (for a nice non-technical introduction, see the \textit{Quanta Magazine} blog post \cite{Wolchover2017}, while for a well rounded critic, see \cite{saxe2019information}).

\begin{figure}[t]
    \centering
    \includegraphics[width=0.99\linewidth]{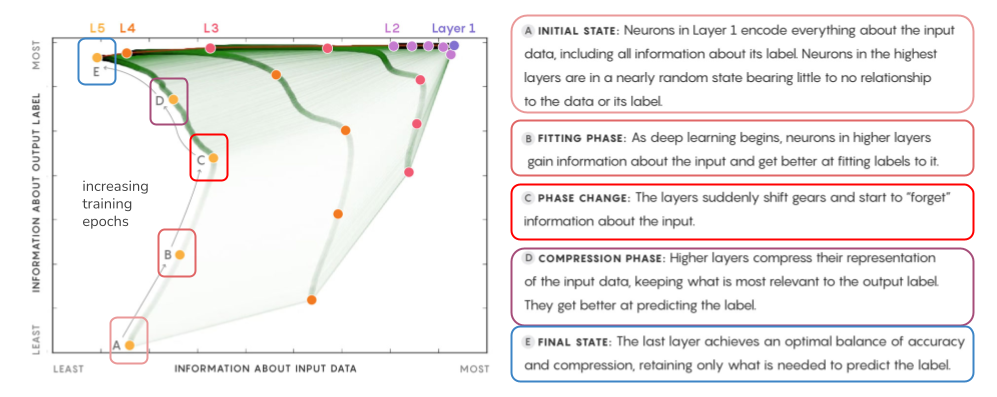}
    \caption{Qualitative evolution of the internal representations of a deep neural network during training, visualized in the \emph{information plane}. 
    The horizontal axis reports $I(X;T)$, the mutual information between the input $X$ and the layer representation $T$, which measures how much information about the input is retained. 
    The vertical axis reports $I(T;Y)$, the mutual information between the representation and the output label $Y$, quantifying how predictive the representation is. 
    Shwartz-Ziv and Tishby reported, in their experimental setup and with their mutual-information estimator, layers tend to follow a two-phase dynamics: an initial \emph{fitting phase}, where both $I(X;T)$ and $I(T;Y)$ increase, followed by a \emph{compression phase}, where $I(X;T)$ decreases while $I(T;Y)$ stabilizes. 
    The descriptive points A to E indicate the evolution across training epochs: the model is initialized at A and, during training, moves through B, C, and D, eventually (and hopefully) reaching E. 
    The example shown corresponds to a 5-layer MLP; the layer indices L1-L5 (from last to first) are reported above the trajectories. 
    This picture is inspired by \cite{tishby2015deep, shwartz2017opening} and adapted from \cite{Wolchover2017}.}
    \label{fig:3:3:DNN_IB}
\end{figure}

\paragraph{Hints on DNN training dynamics from Information Bottleneck Theory}

The Information Bottleneck (IB) theory \cite{tishby2000information, slonim2002information, hu2024survey} provides an information-theoretic perspective on the training and performance of neural networks.

It presents a framework for forming a condensed representation of layer activations with respect to an input variable $x\in\ \mathscr{X}$, retaining as much information as possible about an output variable, $y\in\mathscr{Y}$. A key concept in this theory is the mutual information which suggests that optimal model representations preserve all relevant information about the output while discarding irrelevant input information, thus creating an “information bottleneck”.  The formal definition is
\begin{definition}[Mutual Information]
    Let $X,Y$ be two random variables with a joint distribution $p(x,y)$. Their \textit{mutual information} is defined as
    \begin{equation}
        \begin{split}
            I(X;Y) &\defeq  D_{KL} \left[ p(x,y) \| p(x)p(y)  \right] ,
        \end{split}
    \end{equation}
    where $D_{KL}$ is the Kullback-Leibler divergence. 
\end{definition}
\noindent We can show that
\[
\begin{split}
     I(X;Y) &=D_{KL} \left[ p(x,y) \| p(x)p(y)  \right] \\
     &=  \mathbb{E}_{p(x,y)} \left[ \log \frac{p(x,y)}{p(x)p(y)} \right] \\
     &= \sum_{x\in \mathscr{X}, y\in \mathscr{Y}}  p(x,y) \log \left( \frac{p(x,y)}{p(x)p(y)} \right) \\
     &= \sum_{x\in \mathscr{X}, y\in \mathscr{Y}}  p(x,y) \log \left( \frac{p(x,y)}{p(x)} \right)  - 
     \sum_{x\in \mathscr{X}, y\in \mathscr{Y}}  p(x,y) \log p(y) \\
     &=   \sum_{x\in \mathscr{X}, y\in \mathscr{Y}}  p(x) p(y|x) \log p(y|x) - 
     \sum_{x\in \mathscr{X}, y\in \mathscr{Y}}  p(x,y) \log p(y) 
     \\
     &= \sum_{x\in \mathscr{X}}  p(x)  \left[  \sum_{y\in \mathscr{Y}} p(y|x) \log p(y|x)  \right] - 
     \sum_{y\in \mathscr{Y}} \left[  \sum_{x\in \mathscr{X}} p(x,y) \right] \log p(y) \\
     &= - H(Y|X) + H(Y) = H(Y) - H(Y|X) ,\\
\end{split}
\]
where $H(Y)$ is the marginal entropy, while $H(Y|X)$ is the conditional Entropy, and where we have used $p(x,y) = p(y|x) p(x)$. Notice that, thus, the Mutual Information is symmetric under $X\longleftrightarrow Y$, so 
\[
I(X;Y) =  H(Y) - H(Y|X)  =  H(X) - H(X|Y)  = I(Y;X)
\]
Furthermore, we have \cite{shwartz2017opening}\footnote{Actually, the following properties of mutual information are standard \cite{cover2006elements}; in \cite{shwartz2017opening} these definition are used for their application to the study of the training behaviour of DNNs.}
\begin{remark}
    The Mutual Information has the following properties:
    \begin{itemize}
        \item[1.] \textbf{Nonnegativity}: $I(X;Y)\ge 0$, by Jensen's Inequality. 
        \item[2.] \textbf{Symmetry}: $I(X;Y) = I(Y;X)$.
        \item[3.] \textbf{Relation to Entropies}: We have
        \[
        \begin{split}
            I(X;Y) &= H(Y) - H(Y|X)  =  H(X) - H(X|Y) \\
            &= H(X) + H(Y)  - H(X,Y) \\
            &= H(X,Y) - H(X|Y) - H(Y|X) \,,
        \end{split}
        \]
        where $H(X,Y)$ is the joint entropy of $X, Y$. 
        \item[4.] \textbf{Invariance under invertible transformations}:
        \[
            I(X;Y) = I(\psi(X), \phi(Y)) \,,
        \]
        for any measurable invertible functions $\psi, \phi$\footnote{For continuous variables, we also need the transformations to be \textit{diffeomorphisms} (smooth with smooth inverses) to preserve the differential entropy structure. }.
        \item[5.] \textbf{Data Processing Inequality} (DPI): For any 3 variables that form a Markov chain $X\to Y \to Z$,  
        \[
            I(X;Y) \ge I(X; Z) .
        \]
    \end{itemize}
\end{remark}

The last two properties are very important in the context of DNNs. Indeed, in supervised learning we are interested in good representations, $T (X)$, of the input patterns $x\in \mathscr{X}$, to enable good predictions of the target/label (for regression/classification tasks) $y \in \mathscr{Y}$ . Moreover, we want to efficiently learn such representations from an empirical sample of the (unknown) joint distribution $p(X, Y )$, in a way  that provides good generalisation. 
Furthermore, or a sequential feed-forward network, once the parameters are fixed, the hidden representations satisfy a computational chain\note{This layer-wise Markov description assumes a sequential computational graph. In architectures with skip connections, individual layers do not necessarily form a Markov chain; however, an appropriate coarser-grained representation (e.g., residual blocks) often does \cite{achille2018emergence}.}
\[
Y\to X\to T_1\to\cdots\to T_L\to\hat Y.
\]
%

The insight of \cite{shwartz2017opening} is to use each whole layer, $T$, as a single random variable, which can be characterised by the distribution $P(T|X)$ (the ``encoding'' distribution) and $P(Y|T)$ (the ``decoding'' distribution). As we are only interested in the information that flows through the network, invertible transformations of the representations, that preserve information, generate equivalent representations even if the individual layers encode entirely different features of the input. For this reason, and for the property 4 of the mutual information, it is possible to quantify the representations by the two mutual informations, $I(T;X)$, the mutual information of $T$ with the input $X$, and $I(Y;T)$, the mutual information of the output $Y$ with $T$, and they may play the role of ``order parameters''; furthermore, by property 4., they are invariant to any invertible re-parametrisation of T. 

Another crucial concept, which can be exploit thanks to property 5.~of the mutual information, is the \textit{information  plane}. Any representation variable, $T$ , defined as a (possibly stochastic) map of the input $X$, is characterised by its joint distributions with $X$ and $Y$, $P (T |X)$ and $P (Y |T )$. Given $P (X; Y )$, $T$ is uniquely mapped to a point in the \textit{information plane} $(I(X; T ), I(T ; Y ))$. 

When applied to the deep neural networks with $k$ layers, seen as $k-$step Markov Chain, labelling with $i=1, \ldots, k$ the layer index, the layers are mapped to $k$ monotonic connected points in the plane, called the \textit{information path}. By property 5.~of mutual information, it satisfies the DPI chains:
\begin{align}
    I(X;Y) \ge I(T_1; Y) \ge I(T_2; Y)  \ge \cdots \ge  I(T_k; Y)  \ge  I(\hat{Y}; Y)  \,, \\
    H(X)   \ge I(X; T_1) \ge I(X; T_2)  \ge \cdots \ge  I(X; T_k)   \ge  I(X; \hat{Y})  \,.
\end{align}
If $X$ is discrete, one may additionally write $H(X)\ge I(X;T_1)$. The pair
\[
    \bigl(I(X;T),I(T;Y)\bigr)
\]
defines the usual \textit{information plane}. \cite{shwartz2017opening}  proposed to follow the hidden layers through this plane during optimization and to interpret their trajectories as coarse order parameters for representation learning. 

This plane is where the action happens, during training. It was empirically observed that there is a non-trivial, interesting dynamics showing in the information plane during training (treating the epoch as a discretisation of a training time). 

There is, however, an important technical caveat. In an ordinary deterministic network $T=f_\theta(X)$, $P(T|X)$ ma degenerate. For continuous inputs, $I(X;T)$ can be infinite under common conditions; for finite/discrete empirical inputs it can instead become constant or otherwise uninformative for broad classes of deterministic representations. Consequently, the finite values plotted in an information plane generally require an additional modelling or estimation choice, such as quantisation/binning, injected noise, kernel estimators, or variational estimators \cite{goldfeld2018estimating,amjad2019learning}. The observed trajectory can therefore depend materially on that choice.

This caveat is central to the historical debate around the IB interpretation of deep learning. \cite{shwartz2017opening} reported, for the networks and estimator considered in their study, an initial \emph{fitting} or \emph{drift} regime followed by a slower \emph{compression/diffusion} regime \cite{shwartz2017opening}. In \cref{fig:3:3:DNN_IB} we report an illustrative example. \cite{saxe2019information} subsequently showed that this behaviour is not universal: compression can depend strongly on the activation function and estimator, networks can generalize without displaying the same compression pattern, and compression-like behaviour can also occur under full-batch gradient descent. \cite{goldfeld2018estimating} further clarified that, in deterministic networks, binning-based mutual-information estimates can effectively track geometric clustering rather than a unique intrinsic information loss.

Following \cite{shwartz2017opening}, we can detect five distinct points:
\begin{itemize}

    \item[A.] \textbf{Initial State:} 
    Neurons in the first layer encode essentially almost all information about the input data, including information relevant to the label. 
    Neurons in deeper layers are nearly random and bear little to no relationship to either the input or the output.

    \item[B.] \textbf{Fitting Phase:} 
    As training begins, neurons in higher layers progressively acquire information about the input and become increasingly effective at fitting the labels.

    \item[C.] \textbf{Phase Change:} 
    During training, due to the stochasticity of SGD, the network undergoes a transition in which layers begin to ``forget'' irrelevant information about the input.

    \item[D.] \textbf{Compression Phase:} 
    Higher layers compress their internal representation of the input, retaining primarily the information that is most relevant for predicting the output label. 
    Predictive performance improves as irrelevant variability is discarded.

    \item[E.] \textbf{Final State:} 
    The final layer reaches a balance between accuracy and compression, keeping only the information necessary to predict the label.

\end{itemize}

\begin{figure}[t]
    \centering
    \includegraphics[width=0.95\linewidth]{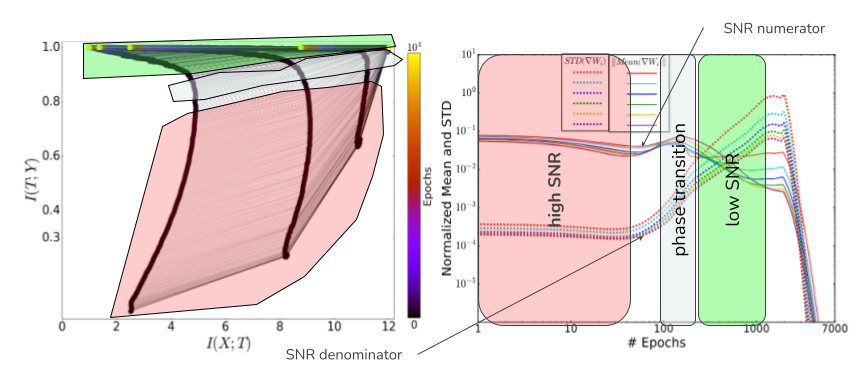}
    \caption{
    On the left, the layers information plane paths during training. On the right, the stochastic gradients means and standard deviations, during training (epochs are on $x-$axis). We have highlighted the three phases of the $\SNR$. 
    Adapted from \cite{shwartz2017opening}.
    }
    \label{fig:3:3:DNN_IB_SNR}
\end{figure}

\paragraph{From Information Bottleneck ideas to gradient diagnostics in PINNs}

A physics-only forward PINN does not possess supervised labels $Y$ in the same sense as an ordinary classification problem: the training signal is provided by differential-equation and constraint residuals. Therefore, the classical supervised IB picture should be viewed here as an analogy for analysing representation and optimization dynamics, not as a direct identification of the PINN objective with the IB functional.

A more direct diagnostic to monitor the PINN training dynamics is the batch-wise \textit{Signal-to-Noise} ratio (SNR), defined as \cite{shukla2024comprehensive, anagnostopoulos2025learning}
\begin{equation}
    \SNR \defeq \frac{\|\mu\|_2}{\|\sigma\|_2} \defeq \frac{\|\mathbb{E}\left[ \nabla_\theta \mathcal{L} \right]\|_2}{\|\std \left[ \nabla_\theta \mathcal{L} \right]\|_2} \,.
\end{equation}
Here the expectation and standard deviation are taken over the chosen batches/partitions while keeping $\theta$ fixed. A high SNR indicates substantial agreement among batch gradients, whereas a low SNR indicates strong heterogeneity and cancellation. Importantly, this diagnostic does \emph{not} require the optimizer itself to perform stochastic mini-batch updates: it can also be evaluated during full-batch training by partitioning the collocation set only for the purpose of measuring gradient agreement \cite{anagnostopoulos2025learning}. Thus, the ``noise'' appearing in this SNR should not automatically be identified with stochastic SGD noise.

As shown in \cref{fig:3:3:DNN_IB_SNR}, the phases of training can be seen as the behaviour of $\SNR$. Furthermore, \cite{shukla2024comprehensive, anagnostopoulos2025learning} have shown that PINNs exhibit \textit{three} distinct training phases:

\begin{figure}[t]
    \centering
    \includegraphics[width=0.95\linewidth]{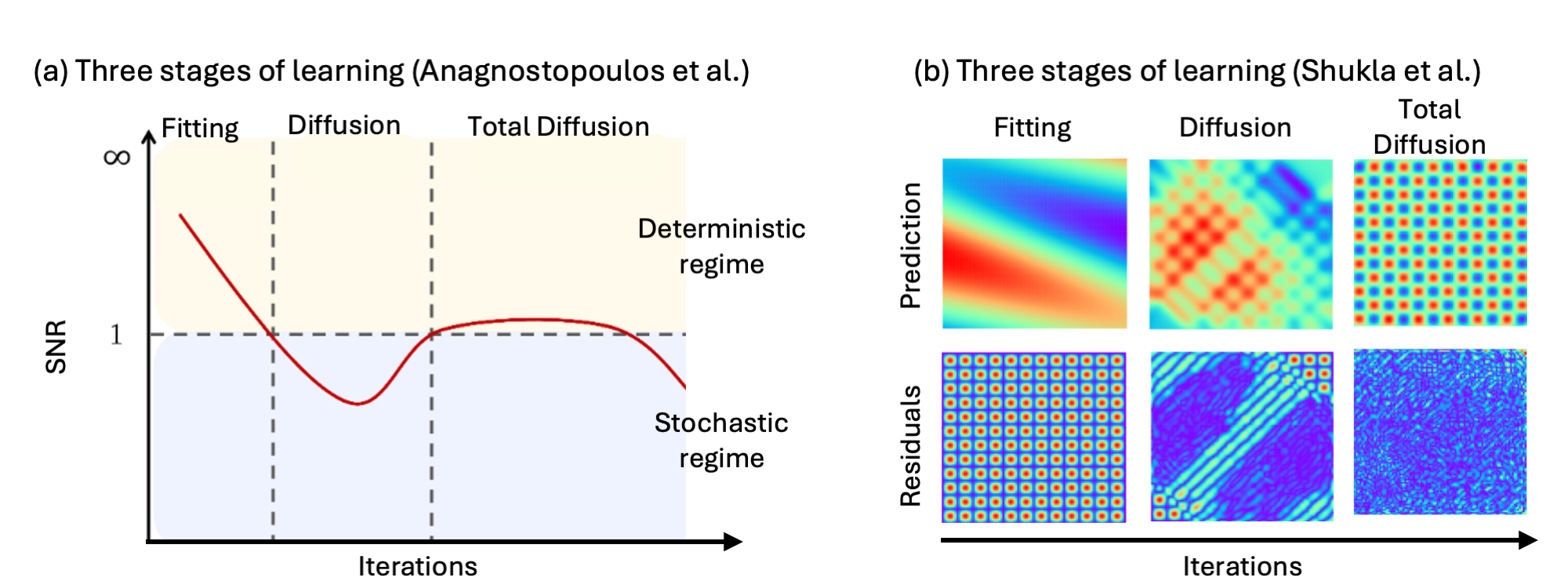}
    \caption{
    Illustration of the three stages of learning in PINNs. 
    (a) The evolution of the batch-wise signal-to-noise ratio (SNR) reveals distinct training regimes: 
    a high-SNR \emph{fitting} phase corresponding to deterministic descent; 
    a low-SNR \emph{diffusion} phase characterized by stochastic exploration; 
    and a final \emph{total diffusion} phase in which the SNR rises sharply and stabilizes, indicating the emergence of a coherent descent direction and accelerated reduction of the generalization error. 
    (b) Prediction fields and residual distributions for the Helmholtz equation at the three stages. 
    Residuals are highly ordered during fitting, become progressively disordered during diffusion, and remain noisy in total diffusion, where the model simplifies its internal representation and closely matches the analytical solution. 
    From \cite{toscano2025pinns}.
    }
    \label{fig:3:3:PINN_Stages_learning}
\end{figure}

\begin{itemize}
    \item[1.] \textbf{Fitting Phase}:
    At the beginning of training, both the loss and its gradients are large across all subdomains. 
    This produces a high batch-wise SNR, since the gradient signal dominates the noise. 
    During this stage, the optimizer follows a clear descent direction and the residuals exhibit an ordered structure. 
    As training progresses and the loss decreases, disagreement between subdomains increases, causing the SNR to drop. 
    This phase is therefore predominantly \emph{deterministic}, transitioning from high to low SNR \cite{shukla2024comprehensive}.

    \item[2.] \textbf{Diffusion Phase}:
    Once the model has adequately fitted the data, it enters a stochastic exploration regime. 
    Here, the gradients across subdomains fluctuate, the SNR remains low, and the network weights undergo a diffusion-like evolution. This diffusion arises from the stochasticity of SGD, where gradient noise dominates the signal.
    This diffusion disrupts the initial ordering of parameters and is associated with improved generalization, as the model searches for a direction that consistently reduces the global training error. 
    Residuals become disordered and the SNR oscillates \cite{shukla2024comprehensive, anagnostopoulos2025learning}.

    \item[3.] \textbf{Total Diffusion Phase}: [New phase in PINNs!]
    After the model identifies a coherent descent direction that reduces the generalization error across all subdomains, the SNR rises sharply and stabilizes. 
    In this phase, the network simplifies its internal representation by retaining only the most relevant features while discarding irrelevant ones, effectively reducing model complexity. 
    Residuals remain highly disordered, but the generalization error (e.g., relative $L^2$) decreases rapidly, indicating convergence. 
    Empirically, the best-performing PINN models are those that enter the total diffusion phase earliest \cite{anagnostopoulos2025learning, shukla2024comprehensive}. This phase appears to be a signature of successful PINN training.

\end{itemize}

The three phases described above are visually summarized in \cref{fig:3:3:PINN_Stages_learning} (Adapted in \cite{toscano2025pinns} from \cite{anagnostopoulos2025learning, shukla2024comprehensive}). 
Panel (a) shows how the batch-wise SNR evolves during training, providing a clear diagnostic of the underlying learning regime: 
the fitting phase corresponds to a high-SNR, largely deterministic descent; 
the diffusion phase emerges when the SNR drops and oscillates, reflecting stochastic exploration; 
and the total diffusion phase begins once the model identifies a coherent descent direction, causing the SNR to rise and stabilize. 
Panel (b) complements this view by displaying predictions and residuals for a Helmholtz problem at each stage. 
Residuals transition from ordered (fitting) to disordered (diffusion), and remain noisy in total diffusion, where the model achieves its best generalization and most accurate reconstruction of the analytical solution. 
Together, these visualizations reinforce the interpretation of PINN training as a progression from deterministic fitting to stochastic exploration and finally to a stabilized, highly generalizable regime.

\begin{figure}[t]
    \centering
    \includegraphics[width=0.95\linewidth]{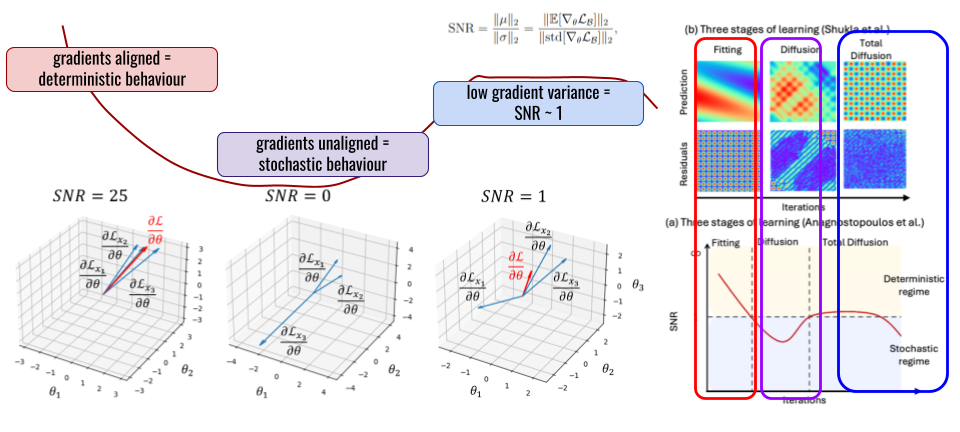}
    \caption{Visual representation of the alignment of the mini-batch gradients during the three PINN training phases. Adapted from \cite{anagnostopoulos2025learning}.}
    \label{fig:3:3:PINN_IB_align}
\end{figure}


These three training phases are fundamentally governed by the geometry of the batch-wise gradients (see \cref{fig:3:3:PINN_IB_align}). As demonstrated in \cite{shukla2024comprehensive, anagnostopoulos2025learning}, the Signal-to-Noise Ratio (SNR) provides a quantitative measure of this geometry: during the \emph{fitting phase}, gradients computed at different collocation points are largely aligned, producing a high SNR and a clear deterministic descent direction; in the subsequent \emph{diffusion phase}, these gradients become increasingly misaligned and partially cancel, yielding a low SNR and a stochastic exploration regime; finally, in the \emph{total diffusion phase}, a coherent global descent direction re-emerges, the SNR rises sharply, and the model rapidly improves its generalization. 

Importantly, we have seen that PINNs are trained through a \textit{multi-objective loss} that typically includes PDE residuals, boundary conditions, and possibly data terms. This decomposition can introduce additional sources of gradient conflict, further contributing to misalignment and potentially creating multiple local transitions between deterministic and stochastic regimes. 

    \chapter{Lecture 4: Advanced PINNs methods - \textit{Learning strategies, Architectures, Losses, and other approaches}} \label{chap:4}

The aim of this lecture is to introduce and examine various techniques to improve PINN training, grouped into five categories: (i) adaptive weighting of loss terms, (ii) advanced sampling and geometry handling, (iii) architectural innovations, (iv) optimisation strategies, and (v) a synthesis of best practices. This section covers a vast amount of research put into optimising PINNs for various tasks. Furthermore, it aims to show the interconnection of PINNs with other machine learning topics, highlighting how hints from one topic may be beneficial in the other.

\section{Dynamic Hyperparameter Optimisation}\label{sec:4:1:DynHypOp}

In \cref{chap:3} we have seen that PINNs, whether in their vanilla, inverse, or parametric formulation, are inherently \textit{multi-objective} optimisation problems. The loss functional typically combines several heterogeneous components,
\begin{equation}
    \mathcal{L}[\theta] 
    = w_{\mathrm{pde}}\,\mathcal{L}_{\mathrm{pde}}[\theta] 
    + w_{\mathrm{bc}}\,\mathcal{L}_{\mathrm{bc}}[\theta] 
    + w_{\mathrm{data}}\,\mathcal{L}_{\mathrm{data}}[\theta] 
    = \sum_{i} w_i\,\mathcal{L}_i[\theta] \,,
\end{equation}
where the coefficients \(w_i\) determine the relative importance of the PDE residual, the boundary or initial conditions, and any available data. It is customary to fix \(w_{\mathrm{pde}} = 1\), but the remaining weights strongly influence the optimisation landscape. For instance, if \(w_{\mathrm{bc}} \ll w_{\mathrm{pde}}\), the network may satisfy the PDE well but violate boundary conditions.

As discussed in \cref{sec:3:3:Learning_Theory}, the training dynamics of PINNs are already intricate due to the evolving alignment of per-sample gradients and the associated transitions between deterministic and stochastic learning regimes. The multi-objective structure of the loss further amplifies this complexity. Different loss terms may operate at different scales, exhibit different curvature properties, or generate gradients that point in conflicting directions. This imbalance can lead to optimisation pathologies such as stiffness, slow convergence, or premature stagnation, phenomena that have been widely documented in the PINN literature \cite{wang2021understanding, wang2022and, bischof2025multi}.

These challenges motivate the development of \textit{dynamic hyperparameter optimisation} strategies, where the weights \(w_i\) are adapted during training rather than fixed a priori. Several approaches have been proposed, including \textit{GradNorm} \cite{chen2018gradnorm}, \textit{Learning Rate Annealing }\cite{wang2021understanding}, \textit{SoftAdapt} \cite{heydari2019softadapt}, \textit{Relative Loss Balancing with Random Lookback} (ReLoBRaLo) \cite{bischof2025multi}, conflict-aware weighting schemes such as \textit{Conflict-Free Inverse Gradients} (ConFIG) \cite{liu2024config}, and NTK-based adaptive weighting inspired by neural tangent kernel theory \cite{jacot2018neural, Weng2022, wang2022and}. These methods aim to rebalance the contributions of the different loss components in real time, mitigate gradient conflicts, and guide the optimiser toward a regime where all objectives are satisfied coherently and efficiently.

In the following paragraphs we review these approaches, analyse their underlying principles, and discuss their practical impact on the training of PINNs.

\subsection{GradNorm}

Multi-objective learning is not unique to PINNs; on the contrary, it arises quite naturally in many Computer Vision applications, often denoted as deep multi-task learning \cite{crawshaw2020multi}. Indeed, often, even single computer vision problems can even be framed as multi-task problems, such as in Mask R-CNN for instance segmentation \cite{he2017mask} or the YOLo family for object detection \cite{redmon2016you, redmon2017yolo9000}. \note{For an explicit example of multi-objective loss, dubbed as ``multi-partite'' loss function, see Equation (3) in \cite{redmon2016you}.}

\begin{figure}[t]
    \centering
    \includegraphics[width=0.95\linewidth]{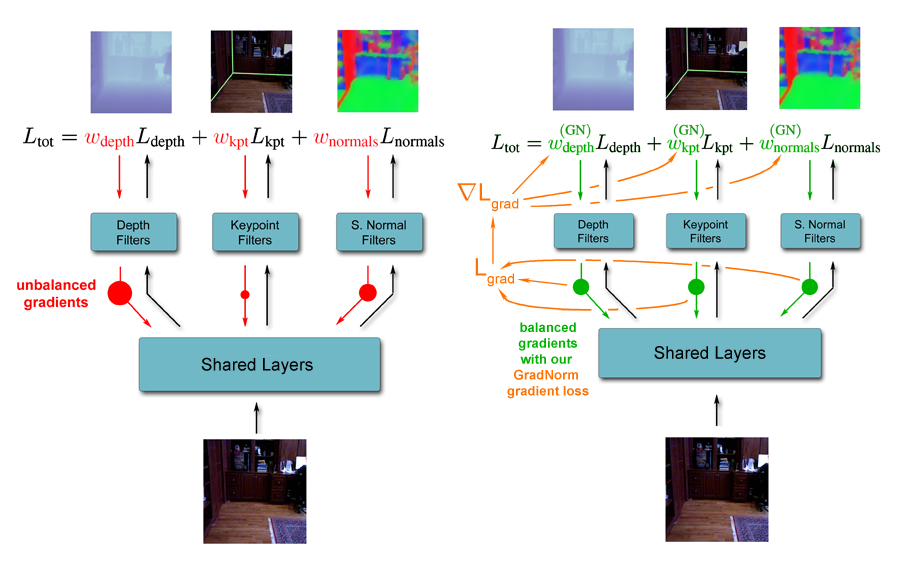}
    \caption{Visual representation of the GradNorm algorithm. Imbalanced gradient norms across tasks (left) result in suboptimal training within a multitask network. GradNorm algorithm is implemented through computing a novel gradient loss $L_{grad}$ (right) which tunes the loss weights $w_i$ to fix such imbalances in gradient norms. From \cite{chen2018gradnorm}.
    }
    \label{fig:4:1:GradNorm1}
\end{figure}

In \cite{chen2018gradnorm}, the authors defined a dynamic hyperparameter re-weighting for multi-objective losses in the form
\[
    \LL (\theta(t) ) = \sum_{i} w_i (t)\,\LL_i[\theta(t) ] \,,
\]
where $t$ is the \textit{training time}, by defining $w_i(t)$ with the goals:
\begin{itemize}
    \item[(1)] gradient norms across tasks are scaled to a common baseline;
    \item[(2)] tasks with slower training rates are assigned higher gradient magnitudes to accelerate their convergence.
\end{itemize}

The new loss, $\LL_{grad}$, is defined in terms of a certain sub-network with parameters $W\subset \mathcal{W}$; the weighted norm of its gradient at a certain epoch $t$
\[
    G_W^{(i)} (t) \defeq \left\| \nabla_W \left( w_i(t) \LL_i (\theta(t))  \right)\right\|_2 \,;
\]
its average across all tasks at step $t$, 
\[
    \overline{G}_W (t) \defeq \mathbb{E}_{task} \left[ G_W^{(i)} (t)  \right] \,,
\]
and the relative inverse training rate of task $i$, 
\[
    r_i (t) \defeq \frac{\tilde{\LL}_i (\theta(t))}{\mathbb{E}_{task} \left[ \tilde{\LL}_i (\theta(t))  \right] } \,, \qquad \tilde{\LL}_i (\theta(t)) \defeq \frac{\LL_i (\theta(t))}{\LL_i (\theta(0))} \,.
\]
The goal of GradNorm is to establish a common scale for gradient magnitudes, and also to balance training rates of different tasks. The common scale for gradients is the average gradient norm, $\overline{G}_W(t)$, which establishes a baseline at each training step $t$ by which we can determine relative gradient sizes. The relative inverse training rate of task $i$, $r_i(t)$, can be used to rate balance our gradients: the higher the value of $r_i(t)$, the higher the gradient magnitudes should be for task $i$ in order to encourage the task to train more quickly. Therefore, our desired gradient norm for each task $i$ is:
\[
G_W^{(i)}(t)  \overset{\text{target}}{\longrightarrow} \overline{G}_W(t)\cdot [r_i(t)]^{\alpha},
\]
where $\alpha$ is an additional hyperparameter, which sets the strength of the restoring force that pulls tasks back to a common training rate.

\begin{algorithm}[tb]
   \caption{Training with GradNorm (from \cite{chen2018gradnorm})}
   \label{alg:4:1:gradnorm}
\begin{algorithmic}
   \State initialise $w_i(0)=1$ $\forall i$
   \State initialise network weights $\mathcal{W}$
   \State Pick value for $\alpha> 0$ and pick the weights $W$ (usually the final layer of weights which are shared between tasks)
   \For{$t=0$ {\bfseries to} $N_{epochs}$}
       \State {\bfseries Input} batch $x_i$ to compute $L_i(t)$ $\forall i$ and   $L(t) = \sum_i w_i(t)L_i(t)$ [standard forward pass]
       \State Compute $G_W^{(i)}(t)$ and $r_i(t)$ $\forall i$
       \State Compute $\overline{G}_W(t)$ by averaging the $G_W^{(i)}(t)$
       \State Compute $L_{\text{grad}}= \sum_i\rvert G_W^{(i)}(t) - \overline{G}_W(t)\cdot [r_i(t)]^{\alpha}\rvert_1$
       \State Compute GradNorm gradients $\nabla_{w_i} L_{\text{grad}}$, keeping  targets $\overline{G}_W(t)\cdot [r_i(t)]^{\alpha}$ constant
       \State Compute standard gradients $\nabla_{\mathcal{W}} L(t)$
       \State Update $w_i(t) \mapsto w_i(t+1)$ using $\nabla_{w_i} L_{\text{grad}}$
       \State Update $\mathcal{W}(t) \mapsto \mathcal{W}(t+1)$ using $\nabla_{\mathcal{W}} L(t)$ [standard  backward pass]
       \State Renormalize $w_i(t+1)$ so that $\sum_iw_i(t+1) = T.$
   \EndFor
\end{algorithmic}
\end{algorithm}

The core idea is that now, we can treat $\{w_i\}$ as \textit{learnable parameters}, adapted at each step, together with the network parameters $W$, by an optimisation algorithm to minimise a single loss. The above equation gives a target for each task $i$'s gradient norms, and thus a natural choice for the loss is
\begin{equation}
    \LL_{grad}(\theta(t); w_i(t) ) \defeq \sum_{i} \left\|  G_W^{(i)} (t) -  \bar{G}_W (t) \cdot \left[ r_i (t) \right]^\alpha \right\|_1 \, ,
\end{equation}
where $\| \cdot \|_1$ is the standard 1-norm, and $\cdot$ is a normal multiplication operator, added for clarity. We report a schematic representation of GradNorm in \cref{alg:4:1:gradnorm}.

\subsection{Homoscedastic Task Uncertainty}

\begin{figure}[t]
    \centering
    \includegraphics[width=0.95\linewidth]{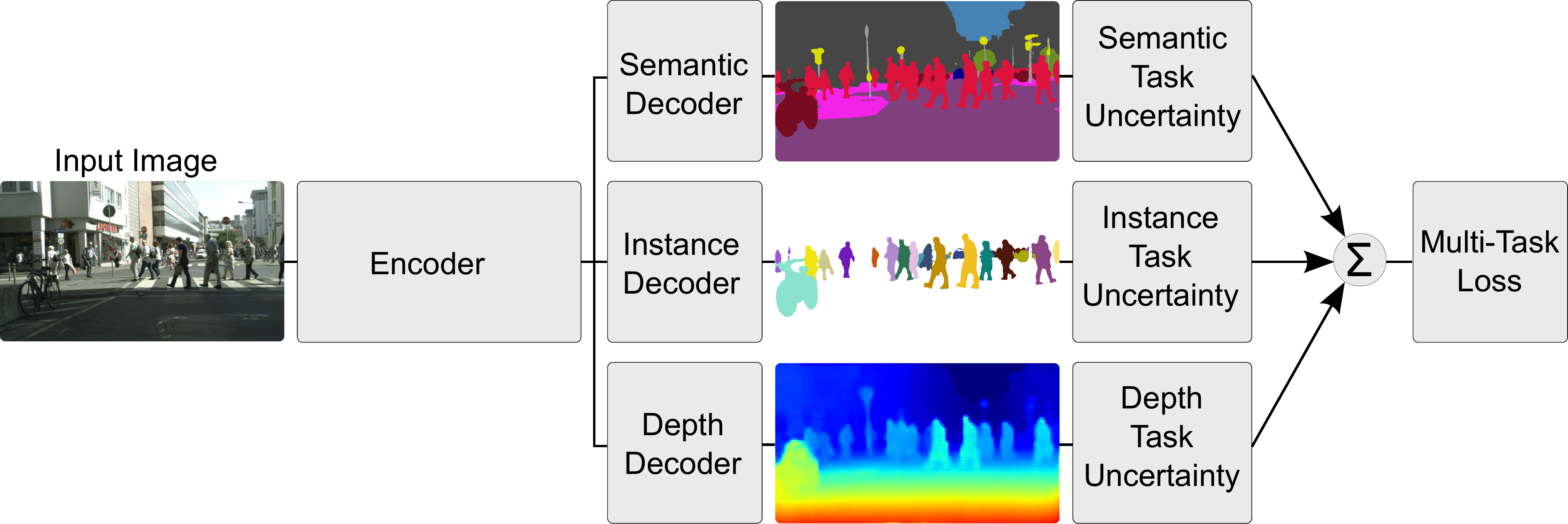}
    \caption{Visual example of a Multi-task deep learning model.  Homoscedastic Task Uncertainty is introduced as a principled way of combining multiple regression and classification loss functions for multi-task learning. 
    As an example, the architecture takes a single monocular RGB image as input and produces a pixel-wise classification, an instance  semantic segmentation and an estimate of per pixel depth. 
    Notice that Multi-task learning can improve accuracy over separately trained models because cues from one task, such as depth, are used to regularize and improve the generalization of another domain, such as segmentation. From \cite{kendall2018multi}.
    }
    \label{fig:4:1:HomoscedasticModel}
\end{figure}

Another approach coming from the field of Computer Vision is \textit{Homoscedastic Task Uncertainty} \cite{kendall2018multi}, where the goal is to balance semantic task, instance task, and depth task uncertainty of a single model (see \cref{fig:4:1:HomoscedasticModel}).

The idea was to use a probabilistic approach to dynamic loss weighting, interpreting each task-specific loss as arising from a likelihood model with an associated observation noise. 
In this framework, the relative weight of each loss term is not treated as a free hyperparameter but is instead derived from the \emph{homoscedastic uncertainty}\note{In statistics, a sequence of random variables is \textit{homoscedastic} if all its random variables have the same finite variance; this is also known as homogeneity of variance. } of the corresponding task. 
This uncertainty is independent of the input data and reflects the intrinsic noise level of the task itself.

In Bayesian modelling, there are two main types of uncertainty one can model \cite{kendall2017uncertainties}: the \textit{Epistemic uncertainty}, describing the uncertainty in the model, which captures what the model does not know due to lack of training data; and the \textit{Aleatoric uncertainty},  capturing the uncertainty with respect to information which the data cannot explain. Epistemic uncertainty can be explained away with the ability to observe all explanatory variables with increasing precision. 
Furthermore, Aleatoric uncertainty  can be further categorized;  If this noise varies with the input (e.g., higher uncertainty in certain regions), it is dubbed \textit{Data-dependent} or \textit{Heteroscedastic} uncertainty. If it remains constant across the input space but differs between tasks, it is \textit{Task-dependent} or \textit{Homoscedastic} uncertainty.

In a multi-task setting, the authors \cite{kendall2018multi} show that the task uncertainty captures the relative confidence between tasks, reflecting the uncertainty inherent to the regression or classification task. They thus propose a multi-task loss function based on maximising the Gaussian likelihood with homoscedastic uncertainty. 

Additionally, they show that, for a regression task with Gaussian likelihood, the negative log-likelihood takes the form (up to an additive constant)
\[
\mathcal{L}_j(\theta, \sigma_j)
=
\frac{1}{2\sigma_j^2}\,\|y_j - f_j(\theta)\|^2
+
\log \sigma_j,
\]
where \(\sigma_j\) is the task-dependent noise parameter. 
The key observation is that the coefficient \(1/\sigma_j^2\) acts as an \emph{automatic loss weight}, while the additional \(\log \sigma_j\) term prevents the trivial solution \(\sigma_j \to \infty\). This regularization term prevents the network from arbitrarily increasing \(\sigma_j\) to down-weight the loss, ensuring a meaningful trade-off.
When multiple tasks are present, the total loss becomes
\[
\mathcal{L}(\theta, \{\sigma_j\})
=
\sum_{j=1}^{n_{\mathrm{loss}}}
\left[
\frac{1}{2\sigma_j^2}\,\mathcal{L}_j(\theta)
+
\log \sigma_j
\right],
\]
and the uncertainty parameters \(\sigma_j\) are learned jointly with the network parameters \(\theta\). 
Tasks with higher intrinsic noise (larger \(\sigma_j\)) receive smaller weights, while tasks with lower noise are emphasised.

In the context of PINNs, this idea can be applied by treating the PDE residual, boundary conditions, and data terms as separate tasks, each with its own homoscedastic uncertainty \cite{NvidiaPhysicsNemoDocumentation}. 
The resulting PINN loss takes the form
\[
\begin{split}
 \mathcal{L}(\theta; \{\sigma_{i}\})
&=
\frac{1}{2\sigma_{\mathrm{pde}}^2}\,\mathcal{L}_{\mathrm{pde}}
+
\frac{1}{2\sigma_{\mathrm{bc}}^2}\,\mathcal{L}_{\mathrm{bc}}
+
\frac{1}{2\sigma_{\mathrm{data}}^2}\,\mathcal{L}_{\mathrm{data}} + 
\log (\sigma_{\mathrm{pde}} \sigma_{\mathrm{bc}} \sigma_{\mathrm{data}} ).
\end{split}
\]

This formulation allows the network to learn the appropriate balance between the different components of the PINN loss directly from the data and the PDE structure, without requiring manual tuning of the weights. 

An alternative and numerically safer way to define the loss, is to redefine the parameters 
\[
2\sigma^2_j \to \exp[  \gamma_j] 
\]
so that
\begin{equation}
     \mathcal{L}(\theta; \{\gamma_{i}\}) = 
     \ee^{-\gamma_{\mathrm{pde}}}\,\mathcal{L}_{\mathrm{pde}} +
     \ee^{-\gamma_{\mathrm{bc}}}\,\mathcal{L}_{\mathrm{bc}} +
     \ee^{-\gamma_{\mathrm{data}}}\,\mathcal{L}_{\mathrm{data}} + 
     \frac{1}{2}\left(\gamma_{\mathrm{pde}} + \gamma_{\mathrm{bc}} + \gamma_{\mathrm{data}}  \right).
\end{equation}

\subsection{Learning Rate Annealing}

One of the earliest and most widely used dynamic weighting strategies for PINNs is \emph{Learning Rate Annealing}, introduced in \cite{wang2021understanding}. 
The key idea is to adapt the relative weight between the PDE residual loss and the boundary (or initial) condition loss by monitoring the magnitude of their gradients during training. 
Since different loss components may operate on different numerical scales, their gradients can differ by several orders of magnitude, which leads to unbalanced optimisation dynamics and poor convergence. 
Learning Rate Annealing addresses this issue by enforcing that the gradients of the different loss terms contribute comparably to the parameter update.

Given a loss of the form
\[
\mathcal{L}(\theta) = \mathcal{L}_{\mathrm{pde}}(\theta) 
+ \lambda^{(i)} \mathcal{L}_{\mathrm{BC}}(\theta),
\]
the adaptive weight \(\lambda^{(i)}\) at iteration \(i\) is computed from the ratio of gradient magnitudes,
\[
\bar{\lambda}^{(i)} 
= 
\frac{
\max\!\left( \left| \nabla_\theta \mathcal{L}_{\mathrm{pde}}(\theta^{(i)}) \right| \right)
}{
\mathrm{mean}\!\left( \left| \nabla_\theta \mathcal{L}_{\mathrm{BC}}(\theta^{(i)}) \right| \right)
},
\]
and smoothed through an \textit{exponential moving average}\note{The role of exponential moving average, or exponential smoothing, is to act as low-pass filters to remove high-frequency noise.},
\[
\lambda^{(i)} 
= 
\alpha\, \bar{\lambda}^{(i)} 
+ (1 - \alpha)\, \lambda^{(i-1)}.
\]

This procedure ensures that the optimiser receives balanced gradient contributions from the PDE and boundary losses, preventing one term from dominating the training dynamics. 
In practice, Learning Rate Annealing aims to mitigate stiffness in the optimisation landscape and improves convergence, especially in problems where the PDE residual and boundary conditions evolve at different rates during training.

\begin{remark}
Although the gradient ratio used in Learning Rate Annealing is not directly related to the batch-wise SNR discussed in \cref{sec:3:3:Learning_Theory}, both quantities share a similar qualitative purpose. 
The SNR measures the relative strength of aligned versus misaligned gradient components across training samples, while the annealing ratio compares the magnitudes of gradients arising from different loss terms. 
In both cases, the goal is to diagnose and correct imbalances in the optimisation dynamics, ensuring that no single component dominates the parameter updates. Both diagnostics can thus be used together: SNR monitors gradient alignment within a loss term, while the annealing ratio monitors balance between loss terms.
\end{remark}

\subsection{SoftAdapt}

\textit{SoftAdapt} is another dynamic loss-balancing strategy that has been successfully applied to PINNs, although it was originally developed in the context of computer vision and multi-part loss functions \cite{heydari2019softadapt}. 
In its original formulation, SoftAdapt was designed to stabilise training in tasks such as image reconstruction with Sparse Autoencoders and synthetic data generation with Introspective Variational Autoencoders, where different loss components (for example, pixel-wise reconstruction, perceptual loss, and adversarial loss) evolve at different rates. 
This makes SoftAdapt particularly interesting for readers with a background in computer vision, since it demonstrates how techniques developed for deep learning in perception tasks can transfer effectively to physics-informed settings, \textit{and vice-versa}.

The core idea of SoftAdapt is to monitor the \emph{relative progress} of each loss term during training. 
Instead of comparing gradient magnitudes, as in Learning Rate Annealing, SoftAdapt compares the ratio of each loss value at the current iteration to its value at the previous iteration. 
Loss terms that decrease slowly (or even increase) are assigned larger weights, while those that decrease rapidly are down-weighted. 
The weights are computed through a softmax transformation, ensuring positivity and smoothness\note{The original SoftAdapt paper uses $s_j(i) = L_j(i) - L_j (i-1)$. The formula reported here is a subsequent variation.}:
\[
\lambda_j^{(i)} 
= 
\frac{
\exp\!\left( \frac{L_j(i)}{L_j(i-1)} \right)
}{
\sum_{k=1}^{n_{\mathrm{loss}}} 
\exp\!\left( \frac{L_k(i)}{L_k(i-1)} \right)
},
\]
where  \(L_j(i-1)\) is the loss at the previous $(j-1)$-th iteration.

This adaptive mechanism encourages the optimiser to focus on the most challenging components of the loss, preventing situations where one term is minimised quickly while others stagnate. 
In the context of PINNs, SoftAdapt has been shown to mitigate imbalance between PDE residuals, boundary conditions, and data terms, often leading to faster and more stable convergence. 
Although not originally designed for scientific machine learning, its transfer to PINNs highlights the broader relevance of multi-objective optimisation techniques across different deep learning domains.

\subsection{ReLoBRaLo}

The \emph{Relative Loss Balancing with Random Lookback} (ReLoBRaLo) method \cite{bischof2025multi} is a refinement of SoftAdapt that addresses two of its main limitations: (a) sensitivity to noise in the loss ratios, and (b) instability caused by relying exclusively on the most recent loss values. 
While SoftAdapt reacts immediately to short-term fluctuations in the loss landscape, ReLoBRaLo introduces two stabilising mechanisms: (i) a moving average over past weights, and (ii) a random lookback window that prevents the optimiser from overfitting to transient loss behaviour.

Mathematically speaking, given a multi-objective loss 
\[
\mathcal{L}(\theta) = \sum_{j=1}^{n_{\mathrm{loss}}} w_j\,\mathcal{L}_j(\theta),
\]
ReLoBRaLo computes the adaptive weights through a combination of exponential smoothing and a stochastic lookback:
\[
w_j(i)
=
\alpha\!\left( \beta\, w_j(i-1) + (1-\beta)\,\hat{w}_j^{(i;0)} \right)
+ (1-\alpha)\,\hat{w}_j^{(i;i')},
\]
where \(\beta\) is a Bernoulli random variable with expectation close to one, and \(i'\) is a randomly selected past iteration. 
The quantities \(\hat{w}_j^{(i;i')}\) are SoftAdapt-style weights computed using a temperature parameter \(\tau\),
\[
\hat{w}_j^{(i;i')}
=
\frac{
\exp\!\left( \frac{L_j(i)}{\tau\,L_j(i')} \right)
}{
\sum_{k=1}^{n_{\mathrm{loss}}}
\exp\!\left( \frac{L_k(i)}{\tau\,L_k(i')} \right)
}.
\]

The temperature \(\tau\) controls the sharpness of the weighting: large values yield nearly uniform weights, while \(\tau \to 0\) recovers an \(\arg\max\)-like behaviour. 
The random lookback mechanism prevents the method from reacting too aggressively to short-term oscillations, while the moving average smooths the evolution of the weights across iterations.

In practice, ReLoBRaLo has been shown to outperform SoftAdapt in some cases in PINN training, 
particularly in stiff problems where different loss components evolve on widely separated scales \cite{bischof2025multi}. 
By combining adaptivity with temporal smoothing and stochasticity, ReLoBRaLo provides a more robust and stable approach to dynamic loss balancing.

\subsection{ConFIG}

The \emph{Conflict-Free Gradient} (ConFIG) method \cite{liu2024config} takes a fundamentally different approach to dynamic hyperparameter optimisation. 
Instead of adapting scalar weights for each loss term, ConFIG directly modifies the \emph{gradient direction} used for the parameter update. 
The motivation stems from the observation that, in multi-objective PINNs, the gradients associated with different loss components often point in conflicting directions. 
Such conflicts can severely hinder optimisation, especially in stiff PDEs where the PDE residual, boundary conditions, and data terms may pull the parameters toward incompatible updates.

To resolve these conflicts, ConFIG constructs a single, unified descent direction. It does so by geometrically aligning the individual gradient vectors, effectively finding a consensus direction that minimizes destructive interference between the competing objectives.
Given the individual gradients \(g_1, g_2, \ldots, g_m\) associated with the \(m\) loss components, ConFIG first normalises each gradient to extract its direction, then computes a consensus direction \(g_u\) by summing the orthogonal components:
\[
g_u 
= 
\mathcal{U}\!\left( 
\left[ 
\mathcal{U}(g_1),\, \mathcal{U}(g_2),\, \ldots,\, \mathcal{U}(g_m) 
\right]^{-T} 
\mathbf{1}_m 
\right),
\]
where \(\mathcal{U}(\cdot)\) denotes normalisation to unit length. 
The final ConFIG gradient is obtained by projecting each loss-specific gradient onto this consensus direction and summing the contributions,
\[
g_{\mathrm{ConFIG}}
=
\mathcal{G}(g_1, g_2, \ldots, g_m)
=
\left( \sum_{i=1}^{m} g_i^{T} g_u \right) g_u.
\]

This construction ensures that the update direction is free of destructive interference between loss terms, while still respecting the relative magnitudes of their projections. 
Unlike weighting-based methods such as SoftAdapt or ReLoBRaLo, ConFIG does not attempt to rebalance the loss values themselves; instead, it enforces geometric coherence at the level of the gradients. 
This makes ConFIG particularly effective in scenarios where gradient conflicts, rather than scale imbalances, dominate the training dynamics.

In practice, ConFIG has been shown to stabilise optimisation and accelerate convergence in challenging PINN problems, especially those involving multiple competing objectives or highly anisotropic residual landscapes. 
Its gradient-level formulation also makes it compatible with standard optimisers such as Adam, as demonstrated in the M-ConFIG algorithm proposed in \cite{liu2024config}.

The ConFIG method can be interpreted as a geometric modification of the standard Adam optimiser \cref{alg:4:1:adam} \cite{kingma2017adammethodstochasticoptimization}. 
Both algorithms maintain first and second moment estimates of the gradients, and both update the parameters through a normalised, momentum-driven descent direction. 
The key difference lies in how the gradient used by the optimiser is constructed. 
Adam relies directly on the raw gradient of the full loss, while M-ConFIG \cref{alg:4:1:ConFIG} replaces this gradient with a conflict-free direction obtained by projecting the individual loss-specific gradients onto a consensus direction. 
This substitution leaves the Adam machinery intact but fundamentally alters the descent geometry, enabling the optimiser to avoid destructive interference between competing objectives.

\begin{algorithm}[H]
\caption{Adam Optimiser}
\label{alg:4:1:adam}
\begin{algorithmic}[1]
\Require $\theta_0$ (initial parameters), $\eta$ (learning rate), $\beta_1, \beta_2$ (momentum coefficients), $\epsilon$
\State $m_0 \gets 0$ \Comment{Init first moment}
\State $v_0 \gets 0$ \Comment{Init second moment}
\For{$t = 1,2,\ldots$}
    \State $g_t \gets \nabla_\theta \mathcal{L}(\theta_{t-1})$ \Comment{compute gradient of full loss}
    \State $m_t \gets \beta_1 m_{t-1} + (1 - \beta_1) g_t$  \Comment{Update biased $1^\circ$ momentum}
    \State $v_t \gets \beta_2 v_{t-1} + (1 - \beta_2) g_t^2$ \Comment{Update biased  $2^\circ$ raw momentum}
    \State $\hat{m}_t \gets m_t / (1 - \beta_1^t)$ \Comment{Compute bias-corrected  $1^\circ$ momentum}
    \State $\hat{v}_t \gets v_t / (1 - \beta_2^t)$ \Comment{Compute bias-corrected $2^\circ$ raw momentum}
    \State $\theta_t \gets \theta_{t-1} - \eta\, \hat{m}_t / (\sqrt{\hat{v}_t} + \epsilon)$ \Comment{Update parameters}
\EndFor
\end{algorithmic}
\end{algorithm}

\begin{algorithm}[H]
\caption{M-ConFIG Optimiser}
\label{alg:4:1:ConFIG}
\begin{algorithmic}[1]
\Require $\theta_0$, $\eta$, $\beta_1, \beta_2$, $\epsilon$, $m_0$ (pseudo first momentum)
\State $\{m_{g_1,0}, \ldots, m_{g_m,0}\} \gets 0$ \Comment{Init  $1^\circ$  momentums for each loss term}
\State $v_0 \gets 0$ \Comment{Init  $2^\circ$  momentum}
\State $\{t_{g_1}, \ldots, t_{g_m}\} \gets 0$
\For{$t = 1,2,\ldots$}
    \State $i \gets t \bmod m + 1$
    \State $t_{g_i} \gets t_{g_i} + 1$
    
    \State $m_{g_i,t_{g_i}} \gets \beta_1 m_{g_i,t_{g_i}} + (1 - \beta_1)\nabla_\theta \mathcal{L}_i(\theta_{t-1})$  \Comment{Update $1^\circ$ momentum of $g_i$ }
    
    \State $\textcolor{red}{\hat{m}_{g_k} \gets m_{g_k,t_{g_k}} / (1 - \beta_1^{t_{g_k}}), \; \forall k}$ \Comment{Bias corrections for $1^\circ$ momentum } 
    
    \State \textcolor{red}{$\hat{m}_g \gets \mathcal{G}(\hat{m}_{g_1}, \ldots, \hat{m}_{g_m})$} \Comment{ConFIG consensus gradient}
    
    \State  $g_c \gets [\hat{m}_g (1 - \beta_1^t) - \beta_1 m_{t-1}]/(1-\beta_1)$ \Comment{Obtain the estimated gradient}
    
    \State $m_t \gets \beta_1 m_{t-1} + (1 - \beta_1) g_c$ \Comment{Update pseudo $1^\circ$ momentum}
    
    \State $v_t \gets \beta_2 v_{t-1} + (1 - \beta_2) g_c^2$ \Comment{Update $2^\circ$ momentum}
    
    \State $\hat{v}_t \gets v_t / (1 - \beta_2^t)$ \Comment{ Bias correction}
    
    \State $\theta_t \gets \theta_{t-1} - \eta\, \hat{m}_t / (\sqrt{\hat{v}_t} + \epsilon)$ \Comment{Update weights}
\EndFor
\end{algorithmic}
\end{algorithm}

The M-ConFIG algorithm preserves the update structure of Adam but replaces the raw gradient with a conflict-free direction obtained through the ConFIG operator. 
The red-highlighted lines indicate the only modifications relative to Adam: the computation of the consensus direction, the projection of the loss-specific gradients onto this direction, and the construction of the conflict-free gradient $g_c$. 
All remaining steps, including moment estimation and bias correction, follow the standard Adam formulation. 
This design allows ConFIG to inherit the robustness and adaptivity of Adam while resolving gradient conflicts that arise in multi-objective PINN training.

\noindent\textit{Computational overhead:}
Although M-ConFIG introduces additional operations compared to standard Adam, its computational overhead remains modest. 
The extra cost arises primarily from computing the individual loss-specific gradients and constructing the consensus direction through the ConFIG operator. 
As noted in \cite{liu2024config}, these steps scale linearly with the number of loss terms and involve only vector normalisation and inner products, which are negligible compared to the cost of backpropagation. 
In practice, the overhead is typically less than a few percent of the total training time, while the improvement in optimisation stability and convergence often outweighs this additional cost.

\subsection{Neural Tangent Kernel-based Adaptive Weighting} \label{subsec:4:1:7:NTK}

The Neural Tangent Kernel (NTK) provides a rigorous framework for understanding how neural networks evolve during gradient descent. In particular, in the abstract of \cite{jacot2018neural}, the authors write

\begin{center}
    \begin{quote}
    \say{\textit{We prove that the evolution of an ANN during training can also be described by a kernel: during gradient descent on the parameters of an ANN, the network function} $f(\theta ; \cdot) \defeq f_\theta$ \textit{(which maps input vectors to output vectors) follows the kernel gradient of the functional cost (which is convex, in contrast to the parameter cost) w.r.t. a new kernel: the \emph{Neural Tangent Kernel} (NTK).}}
\end{quote}
\end{center}

\noindent The starting point is the parameter-space dynamics\note{In the function-space view, the loss $\mathcal{L}$ can be seen as a functional of the network output $f_\theta$.},
\[
\frac{\dd\theta}{\dd t} = -\eta\,\nabla_\theta \mathcal{L}[f_\theta],
\]
which expresses how the network parameters change under continuous-time gradient descent. 
Applying the chain rule to the network output \(f_\theta(\mathbf{x})\) yields
\[
    \frac{\dd f(\mathbf{x};\theta)}{\dd t}
    =
    \nabla_\theta f(\mathbf{x};\theta)\cdot \frac{\dd\theta}{\dd t}
    =
    -\eta\,\nabla_\theta f(\mathbf{x};\theta)\cdot \nabla_\theta \mathcal{L}.
\]
For a loss of the form 
\(\mathcal{L} = \frac{1}{N}\sum_{i=1}^N \ell(f(\mathbf{x}^{(i)};\theta), y^{(i)})\),
this becomes\note{Notice that \(\delta \ell/\delta f\) denotes the \textit{functional derivative} of the loss with respect to the network output. }
\[
\frac{\dd f(\mathbf{x};\theta)}{\dd t}
=
-\frac{\eta}{N}\sum_{i=1}^N 
\nabla_\theta f(\mathbf{x};\theta)\cdot 
\nabla_\theta f(\mathbf{x}^{(i)};\theta)\,
\frac{\delta \ell(f(\mathbf{x}^{(i)};\theta), y^{(i)})}{\delta f}.
\]
The inner product
\[
K(\mathbf{x},\mathbf{x}';\theta)
\equiv 
\nabla_\theta f(\mathbf{x};\theta)\cdot \nabla_\theta f(\mathbf{x}';\theta)
\]
is the \emph{Neural Tangent Kernel}. 
It encodes how changes in the parameters affect the network outputs at different input locations. 
Substituting this definition yields the NTK-driven evolution equation,
\[
\frac{\dd f(\mathbf{x};\theta)}{\dd t}
=
-\frac{\eta}{N}\sum_{i=1}^N 
K(\mathbf{x},\mathbf{x}^{(i)};\theta)\,
\frac{\delta \ell(f(\mathbf{x}^{(i)};\theta), y^{(i)})}{\delta f}.
\]
Furthermore, still in the abstract, the authors say
\begin{center}
    \begin{quote}
        \say{\textit{We then focus on the setting of least-squares regression and show that in the infinite-width limit, the network function} $f_\theta$ \textit{follows a linear differential equation during training. The convergence is fastest along the largest kernel principal components of the input data with respect to the NTK, hence suggesting a theoretical motivation for early stopping.}}
    \end{quote}
\end{center}

Indeed, at least in the infinite-width limit, the NTK becomes constant during training \cite{jacot2018neural}, and the network output follows a linear differential equation whose convergence is governed by the eigenvalues of the kernel. 
This provides a spectral interpretation of optimisation and explains why certain components of the solution converge faster than others. 
The NTK perspective has been applied to PINNs in \cite{wang2022and}, where it reveals that imbalance in the NTK spectrum between PDE residuals and boundary conditions can lead to stiffness and slow convergence. 
For an accessible introduction to NTK intuition and its implications for deep learning, see also \cite{Weng2022}.

The NTK perspective provides not only a theoretical description of training dynamics but also a principled mechanism for rebalancing the loss terms in PINNs. 
As shown in \cite{wang2022and}, the convergence rate of each loss component is governed by the eigenvalues of the corresponding NTK blocks. 
In particular, the PDE residual and boundary-condition losses often exhibit NTK spectra that differ by several orders of magnitude, which leads to stiffness and slow convergence when the loss terms are weighted uniformly. 
This motivates an adaptive weighting strategy in which each loss term is scaled according to the magnitude of its associated NTK eigenvalues.

Let us move on to how to use NTK to dynamically adapt the loss hyperparameters during training. 
Let \(K(t)\) denote the NTK matrix at iteration \(t\), partitioned according to the PDE residual and boundary-condition samples,
\[
    K(t)
    =
    \begin{bmatrix}
    K_{uu}(t) & K_{ur}(t) \\
    K_{ru}(t) & K_{rr}(t)
    \end{bmatrix}
    =
    J(t)J(t)^{T},
\]
where \(J(t)\) is the Jacobian of the residuals (and not the network outputs) with respect to the parameters. 
The eigenvalues of \(K_{uu}(t)\) and \(K_{rr}(t)\) determine the convergence rates of the boundary and PDE losses, respectively. 
To ensure uniform convergence across all objectives, \cite{wang2022and} proposes to scale the loss terms by the ratio of the total NTK trace to the trace of each block:
\[
\lambda_b(t)
=
\frac{\mathrm{Tr}(K(t))}{\mathrm{Tr}(K_{uu}(t))},
\qquad
\lambda_r(t)
=
\frac{\mathrm{Tr}(K(t))}{\mathrm{Tr}(K_{rr}(t))}.
\]
These weights are updated periodically during training and used in the composite loss
\[
\mathcal{L}(\theta)
=
\lambda_b(t)\,\mathcal{L}_b(\theta)
+
\lambda_r(t)\,\mathcal{L}_r(\theta).
\]

This NTK-based reweighting ensures that each loss term evolves at a comparable rate, preventing the optimiser from overfitting to the components associated with larger NTK eigenvalues \cref{alg:4:1:NTK}. 
From a spectral viewpoint, the method equalises the effective conditioning of the optimisation problem by compensating for anisotropy in the NTK. 
In practice, NTK-based weighting has been shown to significantly improve convergence in stiff PINN problems, particularly those where the PDE residual and boundary conditions exhibit highly unbalanced sensitivities.

\begin{algorithm}[H]
\caption{Adaptive Weights for Physics-Informed Neural Networks \cite{wang2022and}}
\label{alg:4:1:NTK}
\begin{algorithmic}[1]
\Require $\theta_0$ (initial parameters), $\eta_\theta$ (learning rate), $S$ (number of iterations)
\State Initialise $\lambda_b \gets 1$, $\lambda_r \gets 1$

\For{$n = 1$ to $S$}
    \State 1. Compute the NTK matrix $K(n)$ and its blocks $K_{uu}(n)$ and $K_{rr}(n)$
    
    \State 2. Compute eigenvalues $\{\lambda_i(n)\}$ of $K(n)$
    
    \State 3. Compute eigenvalues $\{\lambda_i^{uu}(n)\}$ of $K_{uu}(n)$
    
    \State 4. Compute eigenvalues $\{\lambda_i^{rr}(n)\}$ of $K_{rr}(n)$
    
    \State 5. Update adaptive weights; Compute $\lambda_b$ and $\lambda_r$ by
     \begin{align}
          & \lambda_b =  \frac{\sum_{i=1}^{N_r + N_b}\lambda_i(n)}{\sum_{i=1}^{N_b}\lambda_i^{uu}(n)} = \frac{\Tr( {K}(n))}{\Tr( {K}_{uu}(n))} \\
          &\lambda_r =  \frac{\sum_{i=1}^{N_r + N_b}\lambda_i(n)}{\sum_{i=1}^{N_r}\lambda_i^{rr}(n)} = \frac{\Tr( {K}(n))}{\Tr( {K}_{rr}(n))}
      \end{align}
      \State where $\lambda_i(n), \lambda_i^{uu}$ and $\lambda_i^{rr}(n)$ are eigenvalues of ${K}(n), {K}_{uu}(n),  {K}_{rr}(n)$ at $n$-th iteration.
    
    \State 6. Form the weighted loss:  
    \[
    \mathcal{L}(\theta_n)
    =
    \lambda_b\,\mathcal{L}_b(\theta_n)
    +
    \lambda_r\,\mathcal{L}_r(\theta_n)
    \]

    \State 7. Update parameters via gradient descent:
    \[
    \theta_{n+1}
    =
    \theta_n
    -
    \eta \,\nabla_\theta \mathcal{L}(\theta_n)
    \]

\EndFor
\end{algorithmic}
\end{algorithm}

\section{Advanced Schemes}

In this section, we examine a collection of advanced schemes that address specific weaknesses of the vanilla PINN formulation. 
Rather than a single unified theory, these methods can be grouped into three categories: 
(i) Loss \& Gradient Enhancement, which modifies the objective to include derivative information or exact boundary enforcement; 
(ii) Sampling \& Quadrature Optimization, which improves estimation error by choosing collocation points more intelligently; and 
(iii) Training Schedule \& Initialization, which guides the optimization trajectory to avoid poor local minima. 
While diverse, all these schemes share a common goal: to reduce the optimization and estimation errors identified in \cref{chap:3}.

\subsection{Sobolev training}

Sobolev training, also known as gradient-enhanced training, is an advanced technique that improves the learning of PDE solutions by incorporating derivative information directly into the loss function. The idea is to measure the discrepancy between the predicted solution $\hat{u}_\theta$ and the true solution $u$ not only in an $L^p$ sense, but in a Sobolev norm  $W^{p,q}$ (\cref{sec:1:3:func_analys}), which also controls derivatives up to order $q$. 

While Sobolev norms are standard in numerical analysis, their application to neural networks predates PINNs. In \cite{czarnecki2017sobolev}  Sobolev Training was introduced to stabilise training by enforcing agreement on both function values and derivatives.  This idea proved particularly effective in tasks where the target function exhibits sharp variations or where derivative information is available or can be estimated. 

The transition of these ideas to physics-informed learning was natural. Since PDEs explicitly involve derivatives, enforcing residuals in Sobolev norms provides a principled way to incorporate the structure of the differential operator into the training process. This led to the development of Sobolev or gradient-enhanced PINNs \cite{son2021sobolev}, where the PDE residual and its derivatives are minimized simultaneously.

In standard PINN training, the PDE loss enforces the residual in an $L^p$ norm. Sobolev training generalizes this by defining
\[
\mathcal{L}_{PDE} = \left\| \mathcal{F}[\hat{u}_\theta] - f \right\|_{W^{k,p}(\Omega)} = \| r_\theta \|_{W^{p,q}}\; ,
\]
where
\[
\left\| u \right\|_{W^{k,p}(\Omega)} =
\begin{cases}
\left( \sum_{|\alpha| \leq k} \left\| D^\alpha u \right\|_{L^p(\Omega)}^p \right)^{\frac{1}{p}}, & 1 \leq k < \infty, \\
\max_{|\alpha| \leq k} \left\| D^\alpha u \right\|_{L^p(\Omega)}, & p = \infty,
\end{cases}
\]
and $D^\alpha u$ denotes the partial derivative of multi-index order $\alpha$.

In practice, this means that the PINN loss includes not only the PDE residual but also its spatial derivatives:
\[
\mathcal{L}_{Sobolev} = \mathcal{L}_{PDE} + \sum_{|\alpha|=1}^{k} \lambda_\alpha \left\| D^\alpha r_\theta] \right\|_{L^2(\Omega)}^2 .
\]
In \cite{son2021sobolev}, authors found that Sobolev training may lead to faster convergence, and furthermore it may be beneficial for solving high-dimensional PDEs. 

However, care must be taken when choosing the relative weights of the additional loss terms with respect to the standard PDE residual and boundary condition loss terms;  indeed, Sobolev training may even adversely affect the training convergence and accuracy. Furthermore, Sobolev or gradient-enhanced training increases the training time as the differentiation order will be increased and thus extra backpropagation will be required. 

\subsection{Quasi-random sampling}

The choice of collocation points plays a central role in the performance of PINNs. Since the PDE loss is a quadrature approximation of a continuous functional, the sampling strategy directly affects the estimation error $\mathcal{E}_G$. A common baseline is uniform random sampling, which corresponds to Monte Carlo quadrature. While simple and dimension-independent, random sampling suffers from clustering and large gaps, which lead to poor coverage of the domain and slow convergence.

\begin{figure}[t]
    \centering
    \includegraphics[width=0.95\linewidth]{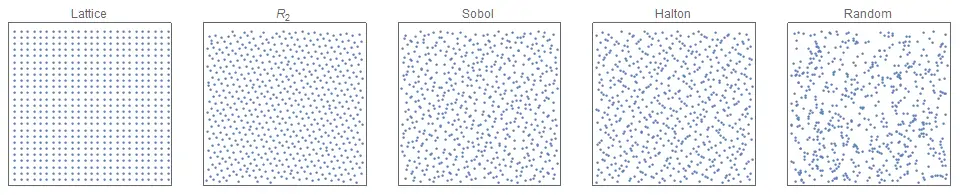}
    \caption{Comparison of a regular lattice (left) with 3 different quasirandom functions (middle), and a simple random distribution (right). Notice that the quasirandom distributions appear less regular than a lattice but do not have as many ‘clumps’ or ‘gaps’ as the random distribution. From \cite{viswanathan2019unreasonable}}
    \label{fig:4:2:Sampling_strategies}
\end{figure}

Quasi-random sampling methods originate from the theory of Quasi-Monte Carlo (QMC) integration, where the goal is to approximate high-dimensional integrals using deterministic point sets with low discrepancy. Classical Monte Carlo sampling achieves an error rate of order $\mathcal{O}(N^{-1/2})$, independent of dimension, but suffers from clustering and irregular coverage. In contrast, low-discrepancy sequences such as Halton and Sobol achieve convergence rates of order $\mathcal{O}(N^{-1} (\log N)^d)$ under mild regularity assumptions \cite{niederreiter1992random,lemieux2009monte}. These sequences fill the domain more uniformly, leading to more accurate quadrature and reduced variance. Their use in PINNs is therefore natural, since the PINN loss is itself a quadrature approximation of a continuous PDE residual. A visual comparison of different sampling strategies is shown in \cref{fig:4:2:Sampling_strategies}.

\paragraph{Halton sequences:}
One of the classical low-discrepancy sequences is the Halton sequence \cite{halton1960efficiency}. It is based on the one-dimensional van der Corput sequence, defined using the radical inverse function. Let us now denote the $N$ prime numbers as $m_1$, $m_2$, $\ldots$, $m_N$. The Halton sequence for the $[0,1]^d$ dimensional space is
\begin{equation}
    x_n = \Big(\Phi_{m_1} (n), \ldots, \Phi_{m_i} (n), \ldots, \Phi_{m_N} (n)\Big)
\end{equation}
where $\Phi_{m_i} (n)$ is the $i$-th \textit{radical inverse function} 
\begin{equation}
    \Phi_{m_i} (n) \defeq \sum_{j=0}^{l(i)} a_j (i,n) m_i^{-(j+1)}
\end{equation}
In the above equation, $a_j$ is an integer coefficient, and it is given as $a_j (i,n) \in [0, m_i -1]$. Also, the integer terms $n$ and $l$ are given by
\begin{equation}
    n = \sum_{j=0}^{l(i)} a_j (i,n) m_i^{j}, \hspace{9mm} l(i) = \ceil{\log _{m_i} n} \,.
\end{equation}
The samples generated using the Halton sequence are only bounded in between $[0,1]^d$ which are spread more evenly within the design space compared to random sampling.




\paragraph{Sobol sequences:}
Sobol sequences \cite{sobol1967distribution, joe2008constructing} are another widely used family of low-discrepancy sequences. Sobol sequences use a binary expansion based on direction numbers derived from primitive polynomials over $\mathbb{F}_2$. This construction guarantees excellent equidistribution across all projections, making them particularly effective for high-dimensional PDEs where uniformity is critical.

The samples are drawn from a special binary fraction ($v_i^{(j)}$) of length $w$ where $i$ = 1, 2, $\ldots$, $w$ and $j$ = 1, 2, $\ldots$, $N$ and where $N$ is the space dimension. The numbers $v_i^{(j)}$ are known as \textit{direction numbers}. To generate direction numbers of dimension $j$, a primitive polynomial over the field $\mathbb{F}_2$ with elements \{0,1\} is considered, which is expressed as
\begin{equation}
    p_j (x) = x^q + b_1 x^{q-1} + \cdots + b_{q-1} x + 1
\end{equation}
Thus, the direction numbers, $v_i^{(j)}$ are generated using the following recurrence relation
\begin{equation}
\begin{split}
    v_i^{(j)} = b_1 v_{i-1}^{(j)} \oplus b_2 & v_{i-2}^{(j)} \oplus \cdots \oplus b_{q-1} v_{i-(q-1)}^{(j)} \oplus \\
    & v_{i-q}^{(j)} \oplus \big(v_{i-q}^{(j)} / 2^q\big) ;\hspace{3mm} i > q \,,
\end{split}
\end{equation}
where $\oplus$ represents the bitwise XOR. Finally, in the dimension $j$, the Sobol sequence is written as
\begin{equation}
    x_n^{(j)} = a_1 v_1^{(j)} \oplus a_2 v_2^{(j)} \oplus \cdots \oplus a_w v_w^{(j)}
\end{equation}
where $n = \sum_{i=0}^{w} a_i 2^i$. The coefficient $a_i$ are deterministic binary digits of the integer $n$.


\paragraph{Hammersley point set:}
The Hammersley point set  \cite{hammersley1960monte} is a finite low-discrepancy design closely related to the Halton sequence. For a given number of points $N$ and dimension $d$, the Hammersley set is defined as
\[
x_n = \Big( \frac{n}{N}, \Phi_{m_1} (n), \ldots, \Phi_{m_i} (n), \ldots, \Phi_{m_{d-1}} (n)\Big)
\]
where $\Phi_{m_1} (n)$ are the radical inverse functions in pairwise coprime bases defined above.   The Hammersley sequence differs from the Halton sequence in that the Hammersley sequence employs ($N$-1)-$\Phi$ sequence; furthermore, while the Halton sequence can generate an infinite number of samples (i.e., infinite $N$), the Hammersley sequence requires an upper bound on the number of samples.
The Hammersley point set is a finite low-discrepancy design that differs from Halton by using a perfectly stratified first coordinate (uniform grid) combined with van der Corput sequences for the remaining coordinates. This hybrid structure yields low discrepancy for finite $N$, making it ideal for design of experiments  \cite{hammersley1960monte}.

\paragraph{Faure sequence:}
The Faure sequence \cite{faure1982faure} is another classical low-discrepancy sequence based on permutations of digits in a fixed prime base.  

Let $m$ is the first prime number such that $m \geq n$, where $n$ is the total number of samples, and let
\[
c_{ij} = \binom{i}{j} \; \operatorname{mod} m \,, \qquad 0 \leq j \leq i \leq p
\]  
This implies $\big\{c_{ij} - \binom{i}{j}\big\}$ is a multiple of $m$. Thus, the base $m$ representation of $n$ is expressed as
\begin{equation}
    n = \sum_{i=0}^{p-1} a_i (n) m^i \,,
\end{equation}
where the coefficients $a_i \in [0, m)$ takes integer values. The samples produced by the Faure sequence are denoted by
\begin{equation}
    x_n = \sum_{j=0}^{p-1} a_j' (n) m^{-(j+1)} \,
\end{equation}
and where 
\[
    a_j' = \sum_{l=j}^{p-1} c_{lj} a_l(n) \; \operatorname{mod} p \,, \qquad j \in \{0, 1, \ldots, p-1\} .
\]

The upper bound of the sample size for the Faure sequence is $m^p$. This is the main difference between the Faure and the Halton sequences.  
Faure sequences use permutations of digits in a fixed prime base to achieve excellent equidistribution. A key advantage over Halton sequences is their ability to fill gaps more rapidly in high-dimensional problems, while avoiding the correlation artifacts that can arise in Halton sequences.

\paragraph{The $R_d$ sequence:}
The $R_d$ sequence, introduced in \cite{viswanathan2019unreasonable}, is a modern low-discrepancy sequence in $d$ dimensional spaces designed to improve uniformity and reduce correlation artifacts present in classical sequences. It is based on irrational rotations on the unit square and exhibits excellent performance in practice. The sequence is defined by
\[
\mathbf{t}_n =  \left(\{ n \alpha_1 \}, \{ n \alpha_2 \}, \cdots, \{ n \alpha_d\},\right)
\]
where $\{\cdot\}$ denotes the fractional part, and where
\[
\alpha_k = \varphi_d^{-k} \,, \qquad k =1,\ldots, d,
\]
while $\varphi_d$ is the solution of the $d-$generalised golden number, i.e. the positive root of
\[
x^{d+1} = x + 1.
\]

Empirical studies show that the $R_d$ sequence often outperforms Halton and Sobol sequences in terms of discrepancy and visual uniformity.

\paragraph{Latin Hypercube Sampling (LHS):}
Latin Hypercube Sampling (LHS)  \cite{mckay1979lhs} is a stratified technique that ensures every coordinate interval is sampled exactly once per dimension. While not strictly low-discrepancy, LHS significantly reduces variance compared to random sampling and is widely used in experimental design.

In $d$ dimensions, the domain is divided into $N$ intervals along each axis, and points are sampled such that each interval is used exactly once per dimension. LHS is not strictly low-discrepancy, but it significantly reduces variance compared to random sampling and is widely used in design of experiments.
For each dimension $j$, one draws a random permutation $\pi_j$ of $\{1,\ldots,N\}$ and defines the sample points as
\[
x_n^{(j)} = \frac{\pi_j(n) - U_{n,j}}{N},
\qquad n = 1,\ldots,N,
\]
where $U_{n,j} \sim U(0,1)$ are independent uniform random variables.  
This guarantees that each interval is used exactly once per dimension, reducing variance compared to pure Monte Carlo sampling \cite{mckay1979lhs}.

\begin{algorithm}[ht]
 \caption{Pseudo-code for generation of basic Latin Hypercube Sampling (from \cite{das2022doe})} 
 \label{alg:4:2:LatinHypercube}
     \begin{algorithmic}[1]
         \State Aim: Generate $n$ samples over the $N$-dimensional space.
         \State Let $x_{ij}$ is the $j$-th coordinate of the $i$-th sample, $x_i$.
         \State \textbf{Begin} 
         \For {$j$ = 1 to $N$}
            \State Generate $P_j$, a random permutation of the set \{1, $\ldots$, n\}.
         \EndFor
         \For {$j = 1$ to $N$}
             \For {$i = 1$ to $n$}
                \State Generate $x_{ij} = \frac{P_j (i) - U_j (i)}{n}$,
            \EndFor
         \EndFor
     \end{algorithmic}
 \end{algorithm}

\begin{figure}[t]
    \centering
    \includegraphics[width=0.95\linewidth]{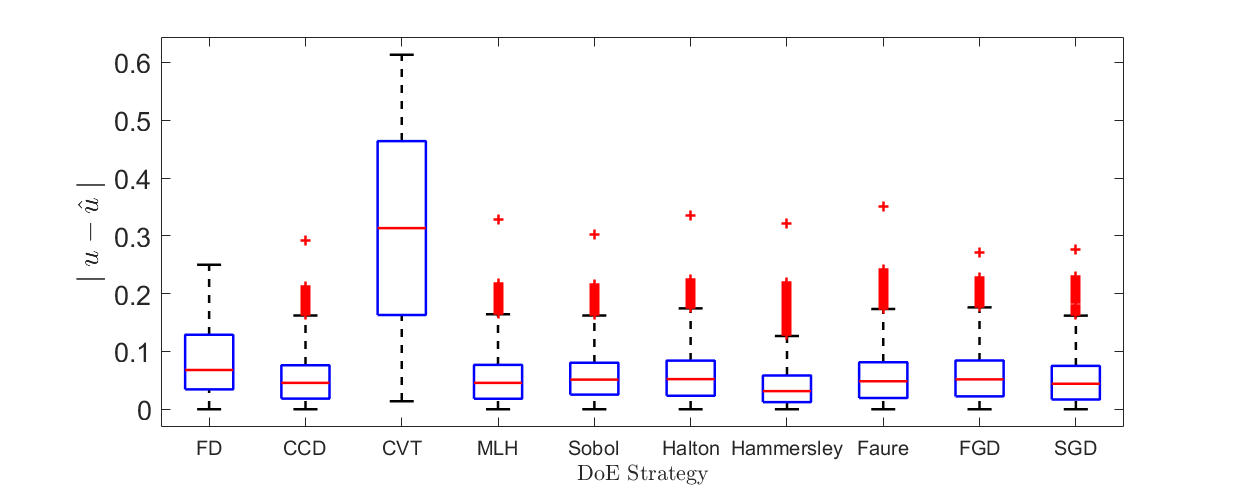}
    \caption{Absolute error between original and predicted responses of Heat equation considering different DoE strategies. From \cite{das2022doe}.}
    \label{fig:4:2:Sampling_res}
\end{figure}

\paragraph{Impact on PINN performance:}
The effect of sampling strategies on PINN accuracy has been systematically studied in \cite{das2022doe, wu2023comprehensive}. Their results, summarized in \cref{fig:4:2:Sampling_res}, show that quasi-random sequences such as Sobol, Halton, Hammersley, and Faure consistently outperform purely random or deterministic designs in terms of prediction error. The improved uniformity of quasi-random sampling leads to more accurate quadrature of the PDE residual, reducing the estimation error $\mathcal{E}_G$ and improving stability during training. 

In particular, the study highlights that quasi-random sequences yield lower median error and fewer outliers, and it emphasises the importance of choosing an appropriate sampling strategy in order to obtain accurate predicted responses using a physics-informed neural network.

\subsection{Importance sampling} \label{subsec:4:2:3:ImportanceSampling}

As we have seen, the standard formulation of PINNs, collocation points are drawn from a uniform distribution over the computational domain. This corresponds to a Monte Carlo approximation of the continuous loss functional and directly influences the estimation error $\mathcal{E}_G$. Uniform sampling is simple and dimension independent, but it is often inefficient when the PDE solution contains sharp gradients, boundary layers, or localized features. In such situations, many points fall in regions where the residual is already small, while too few points are allocated where the solution is difficult to approximate. This imbalance leads to a poor approximation of the continuous risk and slows down convergence.

Importance sampling provides a principled alternative. Instead of sampling from a uniform density $f(\mathbf{x})$, one draws points from a different distribution $q(\mathbf{x})$ that concentrates samples in regions where the PDE residual is expected to be large.\note{Please notice that in \cref{sec:1:4:MCI} we have discussed Monte Carlo integration with generic $\pdf(x)$.} The loss is then re-weighted to preserve an unbiased estimator of the continuous objective, following the classical importance sampling identity. In the context of PINNs, this leads to the modified empirical risk \cite{nabian2021importance})
\[
\theta^* \approx \arg\min_{\theta} \frac{1}{N} \sum_{i=1}^{N} \frac{f(\mathbf{x}_i)}{q(\mathbf{x}_i)} \, \ell(\theta; \mathbf{x}_i),
\qquad \mathbf{x}_i \sim q(\mathbf{x}),
\]
where $f$ is the uniform pdf, and $q$ is the recomputed pdf. 
The key idea is that the sampling distribution should reflect the local difficulty of the PDE, so that the quadrature approximation of the loss becomes more accurate where it matters most\note{Actually, the most correct viewpoint is that we \textit{sample high residual regions more frequently}, and afterwards we reweight them. }.

The authors suggest, as a means of computing the \textit{importance sampling} $\pdf$, either: 
\begin{itemize}
    \item[(a)] based on results on deep learning \cite{katharopoulos2018not, alain2015variance}, where authors shown that, in principle, the rate of convergence is maximised if the training samples are drawn with a $\pdf$ proportional to the relative magnitude of the Loss gradients, 
    \[
        q_j^{(n)} \approx \frac{ \| \nabla_{\theta^{(n)}}\ell[\theta^{(n)}; x_j ]\|_2}{\sum_{j=1}^N \| \nabla_{\theta^{(n)}}\ell[\theta^{(n)}; x_j ]\|_2} \, ,
    \]
    where $\ell$ is the non-negative per sample loss, $n$ is the epoch, and $j$ is the sample index. However, this formula requires computing the 2-norm of this gradient for all of the collocation point at each iteration, and thus requires extra
    backpropagations through the computational graph, which can be computationally expensive be slow and computationally inefficient.

    \item[(b)] A simplified version, relying on the results of \cite{katharopoulos2017biased},  where it was shown theoretically and numerically that a linear transformation of the loss value at a training example is always greater than the 2-norm of loss gradient at that. For this, the $\pdf$ is related directly to the (normalised) residuals:
    \[
        q_j^{(n)} \approx \frac{ \ell[\theta^{(n)}; x_j ]}{\sum_{j=1}^N \ell[\theta^{(n)}; x_j ] } \, .
    \]

\end{itemize}

\begin{figure}[t]
    \centering
    \begin{subfigure}[t]{0.55\textwidth}
        \centering
        \includegraphics[width=\textwidth]{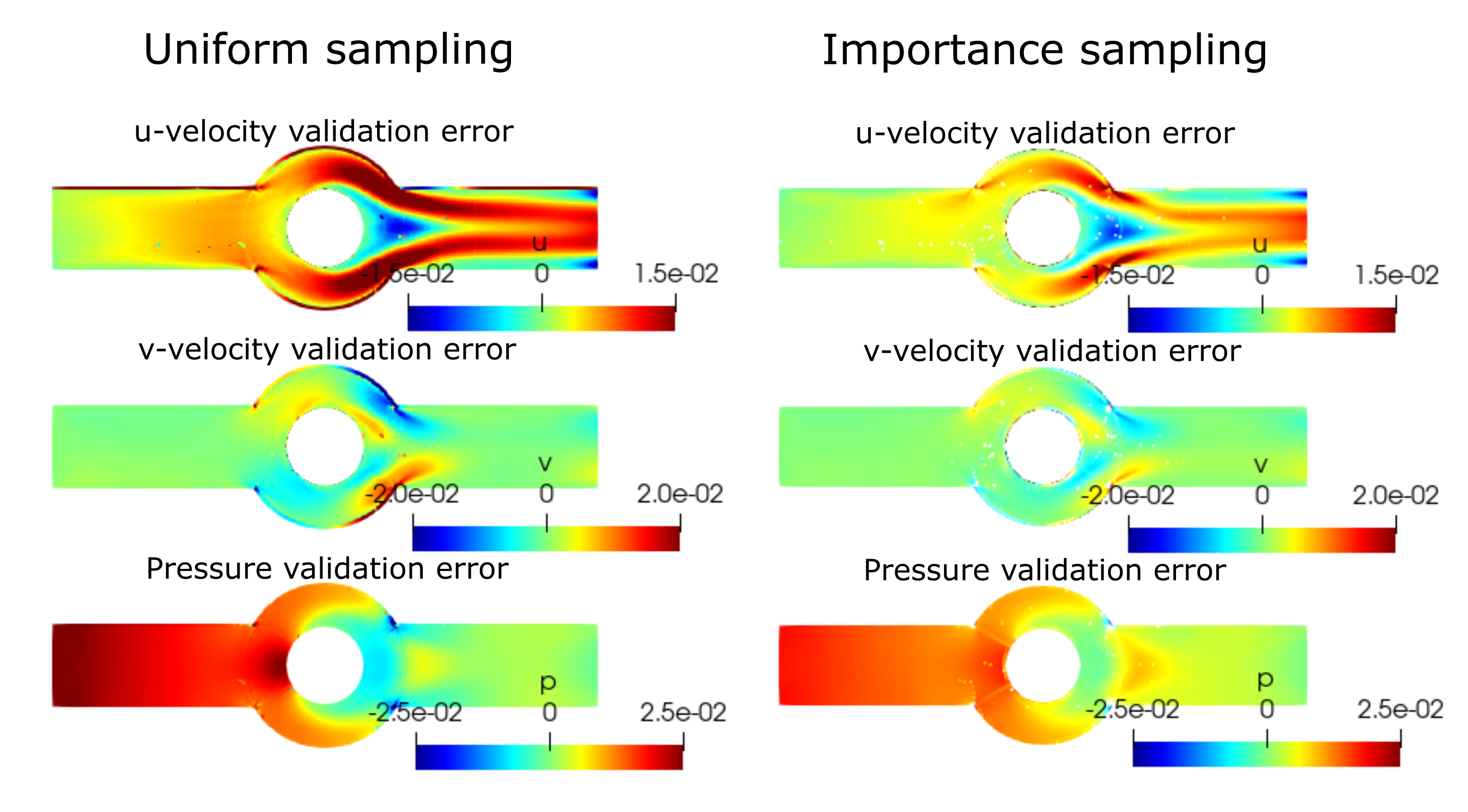}
        \caption{A comparison between the uniform and importance sampling validation error results for stationary navier stokes in an annular ring geometry. From \cite{NvidiaPhysicsNemoDocumentation}.}
    \end{subfigure}
    \hfill
    \begin{subfigure}[t]{0.40\textwidth}
        \centering
        \includegraphics[width=\textwidth]{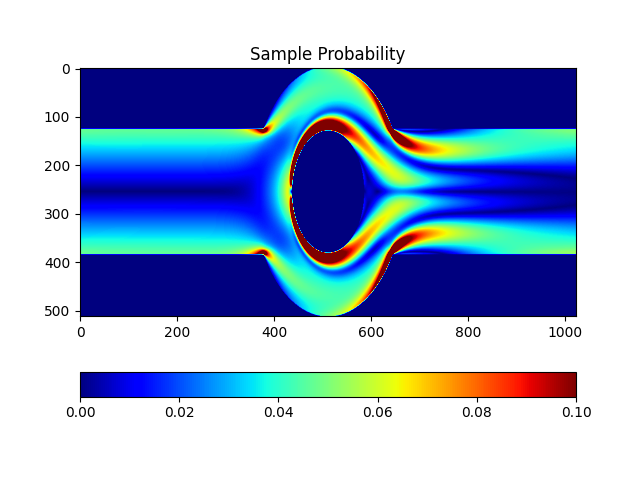}
        \caption{A visualization of the training point sampling probability at iteration 100K for the annular ring example. From \cite{NvidiaPhysicsNemoDocumentation}.}
    \end{subfigure}
    \label{fig:4:2:ImportanceSamplingAR}
\end{figure}

Furthermore, \cite{nabian2021importance}) introduces an adaptive strategy for constructing the sampling distribution \cref{alg:4:2:Importance_sampling}. After a short training phase, the PDE residual is evaluated on a candidate set of points, and the sampling density is updated to be proportional to a power of the residual magnitude. This procedure is repeated periodically during training. The resulting distribution allocates more points to regions where the residual is large, improving the approximation of the continuous loss and accelerating convergence. The authors show that this adaptive refinement leads to a more accurate and stable training process, particularly for problems with localized features or multiscale behaviour.

\begin{algorithm}[ht]
	\caption{Efficient training of PINNs via importance sampling. From \cite{nabian2021importance}}\label{alg:4:2:Importance_sampling}
	\begin{algorithmic}[1]
		\State Generate $N$ collocation points $\{ t_j,\mathbf{x}_j \}_{j=1}^{N}$ sampled from $[0,T] \times \mathcal{D}$, $n$ boundary points $\{\underbar{$\mathbf{x}$}_j\}_{j=1}^{N}$ sampled from ${\partial \mathcal{D}}$, and $S$ seeds $\{ t_s,\mathbf{x}_s \}_{s=1}^{S}$ sampled from $[0,T] \times \mathcal{D}$.
		\State For each collocation point, find the nearest seed.
		\State Set the model architecture (number of layers, dimensionality of each layer, and nonlinearities). Also specify optimizer hyper-parameters, $\lambda_1, \lambda_2$, batch size $m$, and error tolerance $\epsilon$.
		\State initialise model parameters $\mathbf{\theta}^{(0)}$.
		\While {$J( \mathbf{\theta} )>\epsilon$ }
		\State Compute  $\{J(\mathbf{\theta}^{(i)}; \mathbf x_s)\}_{s=1}^S$.
		\State Compute $\tilde{q}_j^{(i)} = {  J(\mathbf{\theta}^{(i)}; \mathbf x_{\rho(j)}) }/{\sum_{j=1}^{N} J(\mathbf{\theta}^{(i)}; \mathbf x_{\rho(j)}) }\; \forall j \in \{1, \cdots, N\}$.
		\State Select a batch of collocation points according to  $\mathbf p^{(i)}=\{\tilde{q}_1^{(i)}, \cdots, \tilde{q}_N^{(i)}\}$.
		\State $\mathbf{\theta}^{(i+1)} = \mathbf{\theta}^{(i)} - \frac{\eta^{(i)}}{mN} \sum_{j \in M^{(i)}} \frac{1}{\tilde{q}_j^{(i)}} \nabla_{\mathbf{\theta}}J(\mathbf{\theta}^{(i)};\mathbf x_j)$.
		\EndWhile
	\end{algorithmic}
\end{algorithm}

Although importance sampling introduces additional computational overhead due to the need to estimate and update the sampling distribution, it provides a systematic way to reduce the estimation error $\mathcal{E}_G$ and to improve the robustness of PINN training. It is compatible with any PINN architecture or optimizer and complements other advanced sampling strategies such as quasi-random sequences or residual-based adaptive sampling. For PDEs with strong spatial variability, it seems that importance sampling may offer a significant improvement over uniform sampling and may represent one of the most effective ways to focus computational effort where the PDE is hardest to satisfy.

In \cref{sec:4:6:Optim_schemes} we will see how similar approaches involving some kind of ``loss re-weighting'' are introduced and generalised. 

\subsection{Approximate distance functions and R‑functions for exact boundary conditions}

We have seen that, in the PINN framework, the boundary/initial conditions are imposed, as for the PDE, as a soft enforcement via penalty terms in the loss, which is a common source of error and instability: the trained network need not satisfy \(\mathcal{B}[u]=g\) on \(\partial\Omega\) exactly, and near-boundary errors may pollute the interior solution. A constructive remedy is to build a geometry‑aware trial space so that Dirichlet data are satisfied by design \cite{Sukumar2022}. Let \(g:\Omega\to\mathbb{R}\) denote the extended function of the prescribed boundary data and \(\widehat u_\theta\) a neural network. If \(\phi:\overline{\Omega}\to\mathbb{R}\) is a scalar field with \(\phi|_{\partial\Omega}=0\) and appropriate normal derivative behaviour on \(\partial\Omega\), then the ansatz
\[
u(\mathbf{x}) = g(\mathbf{x}) + \phi(\mathbf{x})\,\widehat u_\theta(\mathbf{x})
\]
enforces \(u|_{\partial\Omega}=g\) exactly for any choice of \(\widehat u_\theta\). The function \(\phi\) is commonly called an \emph{approximate distance function} (ADF): it vanishes on the boundary, is sign‑consistent with the interior/exterior of \(\Omega\), and is sufficiently smooth so that automatic differentiation yields stable PDE residuals in the interior\note{We will re-encounter ADFs in \cref{sec:4:6:Optim_schemes}.}. The construction of \(\phi\) for complex geometries is most conveniently performed using the algebra of \textbf{R‑functions}, which provides smooth\note{At least away from joining corners.}, differentiable analogues of Boolean set operations on implicit functions and preserves zero level sets under composition \cite{Shapiro1991,rvachev1975method, shapiro2007semi}.

R‑functions realize logical operations on level‑set representations. Given two implicit scalar functions \(r_A(\mathbf{x})\) and \(r_B(\mathbf{x})\) whose zero level sets define primitive subdomains \(A\) and \(B\), one may form smooth approximations of intersection and union that retain the zero level set structure. Two frequently used binary R‑function formulas are
\[
\mathcal{R}_{\cap}(r_A,r_B) \;=\; r_A + r_B - \sqrt{r_A^2 + r_B^2},
\qquad
\mathcal{R}_{\cup}(r_A,r_B) \;=\; r_A + r_B + \sqrt{r_A^2 + r_B^2}.
\]
These expressions are smooth (away from joining corners), sign‑preserving, and satisfy
\[
\mathcal{R}_{\cap}(r_A,r_B)=0 \quad\Longleftrightarrow\quad r_A=0 \text{ or } r_B=0 \quad\text{on the boundary of } A\cap B,
\]
so that the zero level set of the composed function represents the boundary of the composed domain. By iteratively composing primitives with R‑function operators one obtains a single implicit function \(r_\Omega\) whose sign encodes membership in \(\Omega\) and whose zero level set approximates \(\partial\Omega\) with controlled smoothness \cite{Shapiro1991,rvachev1975method}.

To convert an implicit representation \(r_\Omega\) into an ADF \(\phi\) that vanishes on \(\partial\Omega\) and has a prescribed normal derivative, one may apply a smooth monotone transform and a local regularization. A practical choice is
\[
\phi(\mathbf{x}) = \frac{r_\Omega(\mathbf{x})}{\sqrt{r_\Omega(\mathbf{x})^2 + \delta^2}} \,,
\]
with \(\delta>0\) small; this yields \(\phi\to\operatorname{sign}(r_\Omega)\) away from the boundary while ensuring differentiability near \(\partial\Omega\). If a unit normal derivative is desired on the boundary, one can rescale \(\phi\) locally or apply a further normalization so that \(\nabla_n\phi|_{\partial\Omega}=1\) in a weak or pointwise sense. Signed‑distance functions and level‑set primitives (for circles, rectangles, half‑spaces, etc.) are natural building blocks for \(r_\Omega\); standard references on level‑set methods provide practical formulas and numerical techniques for these primitives \cite{osher2004level}.

The PINN workflow with an ADF then proceeds by:
\begin{itemize}
    \item[(i)] constructing primitive implicit functions \(r_i\) for simple geometric pieces;
    \item[(ii)] composing them with R‑function operators to obtain \(r_\Omega\);
    \item[(iii)] regularizing and transforming \(r_\Omega\) into \(\phi\);
    \item[(iv)] training \(\widehat u_\theta\) using the interior PDE residual only, since Dirichlet conditions are satisfied exactly by construction.
\end{itemize}
In practice one implements the composition and regularization symbolically or numerically so that \(\phi\) is available as a differentiable function compatible with automatic differentiation frameworks. The use of an ADF reduces the effective search space of the PINN, improves conditioning near \(\partial\Omega\), and eliminates the need for Dirichlet penalty terms in the loss; these advantages are demonstrated empirically in \cite{Sukumar2022}, who provide algorithmic recipes and numerical examples for a variety of two‑dimensional geometries.

For clarity, a compact algorithmic description of the ADF construction and PINN integration is given below in pseudocode form (the algorithm is intended as a high‑level recipe rather than a line‑by‑line implementation):

\begin{algorithm}[H]
\caption{Construct ADF and train PINN with exact Dirichlet conditions}
\begin{algorithmic}[1]
\State Define primitive implicit functions \(r_i(\mathbf{x})\) for geometric primitives (circles, half‑spaces, rectangles, \dots).
\State Compose primitives using R‑function operators to obtain \(r_\Omega(\mathbf{x})\) representing \(\Omega\).
\State Regularize and transform: \(\phi(\mathbf{x}) \leftarrow r_\Omega(\mathbf{x})/\sqrt{r_\Omega(\mathbf{x})^2+\delta^2}\).
\State Define trial solution \(u(\mathbf{x}) = g(\mathbf{x}) + \phi(\mathbf{x})\,\widehat u_\theta(\mathbf{x})\).
\State Minimize the interior PDE residual \(\|\mathcal{F}[u]\|_{L^2(\Omega)}\) with respect to \(\theta\) using automatic differentiation; omit Dirichlet penalty terms.
\end{algorithmic}
\end{algorithm}

An implementation note: when primitives or R‑function formulas involve non-differentiable operations (e.g., \(\max\), \(\min\)), usually they are replaced with smooth approximations or use analytic signed‑distance expressions to preserve differentiability. The regularisation parameter \(\delta\) must be chosen small enough to keep \(\phi\) close to a signed‑distance near the boundary but large enough to avoid numerical underflow in gradients; corners and cusps require special care and may benefit from local smoothing or mesh‑based refinement.

\subsection{Curriculum learning} \label{subsec:4:2:4:currlearn}

Curriculum learning adapts the training schedule so that the model is first exposed to simpler instances of the task and only later to the full, hard problem. The idea, originally formalized in the machine‑learning literature \cite{bengio2009curriculum}, is to shape the optimization trajectory: early training on easy examples guides the parameters into basins of attraction from which the final, more difficult task can be learned more reliably (for nice recent surveys on the topic, see \cite{wang2021survey, soviany2022curriculum}). In the context of PINNs this paradigm is natural because many PDEs admit families of problems parametrised by physical coefficients, forcing terms, and/or boundary/initial data; by starting from a benign parameter regime and progressively moving toward the target regime, one can reduce both the estimation error and the optimization error.

A canonical example where, despite the simplicity of the problem, vanilla PINNs may experience training difficulties, is the linear continuity (advection) equation on \([0,1]\times[-1,1]\) (see \cite{monaco2023training}),
\[
\begin{split}
    \partial_t u(t,x) + \beta\,\partial_x u(t,x) &= 0,\\
    u(0,x) &= -\sin(\pi x),\\
    u(t,+1) &= u(t,-1),
\end{split}
\]
whose exact solution, easily obtainable by the method of characteristics, is 
\[
u(t,x) = -\sin(\pi(x-\beta t)) .
\]
As \(\beta\) increases the solution becomes increasingly oscillatory in time, and standard PINN training may fail to converge for large \(\beta\) because the residual landscape becomes highly nonconvex and gradients become ill‑conditioned. Curriculum learning addresses this by initialising training at a small velocity \(\beta_0\) and gradually increasing \(\beta\) to the target value, or by using a sequence of intermediate tasks that interpolate between smooth and oscillatory regimes. Empirical studies show that such schedules reduce training time and improve final accuracy for advection‑dominated problems \cite{krishnapriyan2021characterizing,monaco2023training}.

Curriculum strategies for PINNs can be implemented in several ways: (i) parameter continuation, where a physical parameter is ramped according to a predefined schedule or an adaptive rule; (ii) data curriculum, where collocation points or boundary/initial data are introduced progressively (for instance, coarse-to-fine in time or space); and (iii) loss curriculum, where loss terms are weighted dynamically so that easier constraints dominate early training. These variants are complementary and can be combined: for example, one may first train on a low‑\(\beta\) problem with coarse sampling, then increase \(\beta\) while refining the collocation set and rebalancing loss weights.

From an optimization perspective, curriculum learning acts as a form of informed initialization or regularization: by steering parameters toward favourable regions of the loss landscape, it reduces the probability of becoming trapped in poor local minima and mitigates gradient pathologies. The approach is particularly effective when the family of easier problems is chosen so that their solutions are continuously connected to the target solution; in practice, simple linear schedules or a small number of intermediate steps are often sufficient to obtain substantial improvements \cite{bengio2009curriculum,krishnapriyan2021characterizing}.

\section{Architectures}

\begin{table}[hbt!]
\centering
\resizebox{\textwidth}{!}{%
\begin{tabular}{@{} c  l  l  @{}}
\toprule
\textbf{NN family} &
  \textbf{NN type} &
  \textbf{Papers} \\ \midrule
\multirow{6}{*}{FF-NN} &
  1 layer / EML &
  \begin{tabular}[c]{@{}l@{}}\cite{Dwi2020_PhysicsInformedExtreme_SriDS},\\ \cite{Sch2021_ExtremeTheoryFunctional_FurSFL}\end{tabular} \\ \cline{2-3} 
 &
  2-4 layers &
  \begin{tabular}[c]{@{}l@{}}32 neurons/layer \cite{He2020_PhysicsInformedNeural_BarHBTT} \\ 50 neurons/layer \cite{Tar2020_PhysicsInformedDeep_MarTMP}\end{tabular} \\ \cline{2-3} 
 &
  5-8 layers &
  250 neurons/layer \cite{Zhu2021_MachineLearningMetal_LiuZLY} \\ \cline{2-3} 
 &
  9+ layers &
  \begin{tabular}[c]{@{}l@{}}\cite{Che2021_DeepLearningMethod_ZhaCZ}\\ \cite{Wah2021_PinneikEikonalSolution_HagWHA}\end{tabular} \\ \cline{2-3} 
 &
  Sparse &
  \cite{Ram2021_SpinnSparsePhysics_RamRR} \\ \cline{2-3} 
 &
  multi FC-DNN &
  \begin{tabular}[c]{@{}l@{}}\cite{Ami2021_PhysicsInformedNeural_HagANHC} \\ \cite{Isl2021_ExtractionMaterialProperties_ThaITMH} \end{tabular} \\ \midrule
\multirow{2}{*}{CNN} &
  plain CNN &
  \begin{tabular}[c]{@{}l@{}}\cite{Gao2021_PhygeonetPhysicsInformed_SunGSW}\\ \cite{Fan2021_HighEfficientHybrid_Fan} 
  \end{tabular} \\ \cline{2-3} 
 &
  AE CNN &
  \begin{tabular}[c]{@{}l@{}}\cite{Zhu2019_PhysicsConstrainedDeep_ZabZZKP}, \\ \cite{Gen2020_ModelingDynamicsPde_ZabGZ} \\ \cite{Wan2021_TheoryGuidedAuto_ChaWCZ}\end{tabular} \\ \midrule
\multirow{2}{*}{RNN} &
  RNN &
  \cite{Via2021_EstimatingModelInadequacy_NasVNDY} \\ \cline{2-3} 
 &
  LSTM &
  \begin{tabular}[c]{@{}l@{}}\cite{Zha2020_PhysicsInformedMulti_LiuZLS}\\ \cite{Yuc2021_HybridPhysicsInformed_ViaYV}\end{tabular} \\ \midrule
\multirow{2}{*}{Other} &
  BNN &
  \cite{Yan2021_BPinnsBayesian_MenYMK} \\ \cline{2-3} 
 &
  GAN &
  \cite{Yan2020_PhysicsInformedGenerative_ZhaYZK} \\ \bottomrule
\end{tabular}%
}
\caption{A list of standard neural network architecture, such as Feedforward neural networks (FFNNs), convolutional neural networks (CNNs), and recurrent neural networks (RNN), utilised in PINN implementations.
A publication is reported for each type that either used this type of network first or best describes its implementation.
From \cite{cuomo2022scientific}
}
\label{tab:4:3:NN4PINNv1}
\end{table}

We have introduced PINN as a general training recipe, without any specification on the model architectures; apparently, everything that is approximating a function, and whose parameter can be learned by solving the multi-task problem, can do the job. This is in principle true (provided that the architecture and activation functions have sufficient regularity for the derivatives entering the chosen PDE residual); nevertheless, as we will see, in the PINN practice some intervention on the model architecture may be required, if not even \textit{necessary}. In \cite{cuomo2022scientific}, authors list a set of standard architectures used in PINNs: fully-connected models, convolutional models, Long-Short Term Memory (LSTM), as well as Bayesian Neural Networks and Generative Adversarial Neural Networks (see \cref{tab:4:3:NN4PINNv1}), are all have been used to tackle PINNs. 

Nevertheless, empirical evidence has suggested that PINNs may need special care when discussing neural network architectures; for example, their \textit{sizes}: as already noted in \cite{cuomo2022scientific}, and more broadly discussed in \cite{Rowan2026small}, most of the time, when designing model for PINNs, it is more convenient to use \textit{small} networks.

Furthermore, when dealing with Physics-Informed problem we easily encounter a less known issue with deep learning models: the \textit{spectral bias}, which we will discuss in depth in \cref{subsec:4:4:3:SpectralBias}. This issue is relevant also in other fields \cite{liu2024finer}, most notably in the context of \textbf{Implicit Neural Representations} (INR) \cite{sitzmann2020implicit}, which serve as the foundational building blocks for neural scene representations, where INR are designed to learn continuous functions using a multilayer perception (MLP) that maps coordinates to visual signals, such as images, videos, and 3D scene.

\subsection{Adaptive Activations and Weight Factorisation}

\subsubsection{Adaptive Activations} \label{subsubsec:4:3:1:1:AdaptiveAct}
While designing the neural networks to be used to tackle the differential problem, \cite{jagtap2020adaptive,jagtap2020locally}, in addition to leaning the linear transformations, suggest to also learn the non-linear transformations to potentially improve the convergence as well as the accuracy of the model, by using the global adaptive activation functions.

Global adaptive activations consist of a single trainable parameter that is multiplied by the input to the activations in order to modify the slope of activations:
\begin{equation}
    \mathcal{N}^{(\ell)} \left(H^{(\ell-1)}; \theta^{(\ell)}, a \right) = \sigma\left(a \, {L}_{\theta^{(\ell)}}^{(\ell)} \left(H^{(\ell-1)}\right) \right),
\end{equation}
where $\mathcal{N}^{(\ell)}$ the non-linear transformation at layer $\ell$, $H^{(\ell-1)}$ is the output of the $(\ell-1)-$hidden layer, $\theta^{(\ell)}$ is the set of weights and biases of layer $\ell$, $a$ is the global adaptive activation parameter, which is one \textit{per-network}, $\sigma$ is the activation function, and $ {L}^{(\ell)} : x \mapsto W^{(\ell)} x + b^{(\ell)}$ is the linear transformation at layer $\ell$. 

Similar to the network weights and biases, the global adaptive activation parameter $a$  is also a trainable parameter which needs to be optimised during training.

The intuition of introducing a scalable hyper-parameter in the activation function, which has to be optimised, is that it changes dynamically the topology of the loss function involved in the optimization process. Empirically,  \cite{jagtap2020adaptive} shows that the adaptive activation function has better learning capabilities with respect to the fixed activation as it improves the convergence rate, especially at early training, as well as the solution accuracy.

\begin{remark}
    At first sight, multiplying each pre-activation by a learnable scalar may look like a trivial rescaling of the weights. For a single layer one can indeed write
    \[
        \sigma\!\left(a(Wx+b)\right)
        =
        \sigma\!\left((aW)x+ab\right).
    \]
    The important point, however, is not that $a$ simply rescales the parameter norm or the learning rate. Once the slope is shared across several nonlinear layers and optimized jointly with the remaining parameters, it changes the parameterisation of the nonlinear feature map and therefore modifies the network Jacobian and the curvature of the loss in a state-dependent manner.
    
    For example, already for a scalar activation $y=\sigma(az)$,
    \[
        \frac{\partial y}{\partial W}
        =
        a\,\sigma'(az)\frac{\partial z}{\partial W},
    \]
    which is not, in general, $a$ times the derivative of the corresponding unscaled network because $\sigma'(az)\neq\sigma'(z)$. Consequently, there is no generic transformation law $K\mapsto a^2K$ for the Neural Tangent Kernel, nor $H\mapsto a^2H$ for the Hessian of the loss. If $a$ is itself trainable, the full NTK also contains derivatives with respect to $a$.
    
    The more precise interpretation developed in \cite{jagtap2020locally} is that adaptive activations modify the gradient dynamics through implicit, parameter-dependent conditioning matrices. Thus, their effect is better viewed as a learned reparameterisation of the optimization geometry than as a scalar adaptive learning-rate rescaling.
\end{remark}

\subsubsection{Random Weight Factorisation} \label{4:2:1:2:WeightFactor}

Another similar approach was outlined in \cite{wang2022random}, where authors introduced the concept of \textit{Random Weight Factorisation}. 

Let $\xx \in \R^d $ be the input, ${g}^{(0)}(\xx) = \xx$ and $d_0 = d$. They define a standard multi-layer Perceptron (MLP) $f_{{\theta}}(\xx)$ recursively defined by
\begin{align}
    {f}_\theta^{(l)}(\xx) = {W}^{(l)} \cdot {g}^{(l-1)}(\xx) + {b}^{(l)}, \quad {g}^{(l)}(\xx) = \sigma({f}_\theta^{(l)}(\xx)), \quad l = 1,2, \dots, L,
\end{align}
with a final layer 
\begin{align}
    f_{{\theta}}(\xx) &= {W}^{(L+1)} \cdot {g}^{(L)}(\xx) + {b}^{(L+1)},
\end{align}
where ${W}^{(l)} \in \R^{d_l \times d_{l-1}}$ is the weight matrix in $l$-th layer and $\sigma$ is an element-wise activation function. Here, the ensemble of trainable parameters are ${\theta}=\left({W}^{(1)}, {b}^{(1)}, \ldots, {W}^{(L+1)},  {b}^{(L+1)}\right)$ . 

MLPs are commonly trained by minimizing an appropriate  loss function $\mathcal{L}({\theta})$ via gradient descent. To improve convergence, we propose to factorise the weight parameters associated with each neuron in the network as follows
\begin{align}
    \label{eq:4:3:1:2:neuron_fact}
    {w}^{(k, l)} =  s^{(k, l)}  \cdot {v}^{(k, l)}, \quad k=1, 2, \dots, d_l, \quad l = 1,2, \dots, L+1,
\end{align}
where ${w}^{(k,l)} \in \R^{d_{l-1}}$ is a weight vector representing the $k$-th row of the weight matrix ${W}^{(l)}$, $s^{(k,l)} \in \R$ is a trainable scale factor assigned to each individual neuron, and ${v}^{(k,l)} \in \R^{d_{l-1}}$. Thus, the proposed weight factorisation can be rewritten as
\begin{align}
    {W}^{(l)} = \mathrm{diag}({s}^{(l)}) \cdot  {V}^{(l)}, \quad l = 1,2, \dots, L+1.
\end{align}
with ${s} \in \R^{d_l}$.

A way to interpret the relevance of this approach is to look its associated gradient updates. We recall that a standard gradient descent update with a learning rate $\eta$ takes the form 
\begin{align}  
    \mathbf{w}^{(k,l)}_{n+1} &= \mathbf{w}^{(k,l)}_{n} - \eta \frac{\partial \mathcal{L}}{\partial \mathbf{w}^{(k,l)}_{n}}.
\end{align}
In \cite{wang2022random}, they prove the following
\begin{theorem} Under the weight factorisation of \eqref{eq:4:3:1:2:neuron_fact}, the gradient descent update is given by
\begin{align}
    \mathbf{w}^{(k, l)}_{n+1} = \mathbf{w}^{(k, l)}_{n} - \eta \left( [s^{(k,l)}_n]^2 + \| \mathbf{v}^{(k,l)}_n\|^2_2 \right)  \frac{\partial \mathcal{L}}{\partial \mathbf{w}^{(k,l)}_{n}}  + \mathcal{O}(\eta^2),
\end{align}
for $l=1, 2,\dots, L+1$ and $k=1, 2, \dots, d_l$.
\end{theorem}

We notice that the weight factorisation $\mathbf{w} = s \cdot \mathbf{v}$ re-scales the learning rate of $\mathbf{w}$ by a factor of $(s^2 + \|\mathbf{v}\|_2^2)$. Since $\mathbf{s}, \mathbf{v}$ are both trainable parameters, this suggests that weight factorisation effectively assigns a self-adaptive learning rate to each neuron in the network.


It is worth pointing out that RWF is related to adaptive activations in that both introduce trainable scale degrees of freedom, but it acts through a weight-space reparameterisation and is more directly related to weight normalisation \cref{subsubsec:4:3:1:1:AdaptiveAct}.

A major difference is the initialisation strategy. The random weight factorisation is applied as follows; we first initialise the parameters of an MLP network via the Glorot scheme \cite{glorot2010understanding}. Then, for every weight matrix $\mathbf{W}$, we proceed by initialising a scale vector $\exp(\mathbf{s})$ where $\mathbf{s}$ is sampled from a multivariate normal distribution $\mathcal{N}(\mathbf{\mu}, \sigma \mathrm{I})$. Finally, every weight matrix is factorised by the associated scale factor as  $ \mathbf{W} = \mathrm{diag}(\exp(\mathbf{s})) \cdot \mathbf{V}$ at initialization.  We train this network by gradient descent on the new parameters $\mathbf{s}, \mathbf{V}$ directly. See \cref{alg:4:3:1:2:RandomWeightAlg}.

\begin{algorithm}[H]
\caption{Random weight factorisation (RWF). From \cite{wang2022random}.}\label{alg:4:3:1:2:RandomWeightAlg}
\begin{algorithmic}
\State {1. initialise a neural network $f_{\mathbf{\theta}}$ with $\mathbf{\theta} = \{\mathbf{W}^{(l)}, \mathbf{b}^{(l)}\}_{l=1}^{L+1}$ (e.g., using the Glorot scheme).}
      \For{$l = 1,2, \dots, L$}
        \State {(a) initialise each scale factor as $\mathbf{s}^{(l)}\sim\mathcal{N}(\mu, \sigma I)$.} 
        \State {(b)  Construct the factorised weight matrices as $\mathbf{W}^{(l)} = \text{diag}( \exp( \mathbf{s}^{(l)})) \cdot \mathbf{V}^{(l)}$.  }
      \EndFor
      \State {2. Train the network by gradient descent on the factorised parameters $\{\mathbf{s}^{(l)}, \mathbf{V}^{(l)}, \mathbf{b}^{(l)}\}_{l=1}^{L+1}$.}

\State {The recommended hyper-parameters are $\mu=1.0, \sigma=0.1$.} 
\end{algorithmic}
\end{algorithm}

\begin{lstlisting}[language=Python, caption=Torch implementation of Random Weight Factorisation]
import torch
import torch.nn as nn
import math

def factorised_glorot_normal(mean=1.0, stddev=0.1):
    """
    Returns a function that initialises (s, v) such that:
        w = glorot_normal(shape)
        s ~ LogNormal(mean, stddev)  # shape = (out_features,)
        v = w / s
    """
    def init(shape):
        in_features, out_features = shape
        # Glorot normal for w
        fan_in, fan_out = in_features, out_features
        glorot_std = math.sqrt(2.0 / (fan_in + fan_out))
        w = torch.randn(shape) * glorot_std
        # Log-normal scaling vector s (per output unit)
        s = torch.exp(torch.normal(mean=mean, std=stddev, size=(out_features,)))
        # Broadcast divide: v[i,j] = w[i,j] / s[j]
        v = w / s
        return s, v
    return init

class factorisedLinear(nn.Module):
    def __init__(self, in_features, out_features, mean=1.0, stddev=0.1):
        super().__init__()
        self.in_features  = in_features
        self.out_features = out_features
        # initialise s and v using the factorised initialiser
        s, v = factorised_glorot_normal(mean, stddev)((in_features, out_features))
        # Register as parameters
        self.s = nn.Parameter(s)              # shape (out_features,)
        self.v = nn.Parameter(v)              # shape (in_features, out_features)
        self.bias = nn.Parameter(torch.zeros(out_features))

    def forward(self, x):
        # Recompose kernel
        W = self.v * self.s      # broadcasting: (in, out) * (out,)
        return x @ W + self.bias

# example Model
pinn_model = nn.Sequential(
    factorisedLinear(3, 16),
    nn.ReLU(),
    factorisedLinear(16, 16),
    nn.ReLU(),
    factorisedLinear(16, 1)
)
\end{lstlisting}

\subsection{Deep Galerkin Method, Deep Ritz Method, and Variational PINNs}

We have seen in \cref{subsec:4:2:1:RitzVsGalerkin} that, in finite‑element practice, two closely related but conceptually distinct families of weak‑form discretisations are commonly used: the \emph{Ritz} (energy) method and the \emph{Galerkin} (projection) method. 
The Ritz approach begins from an energy or action functional \(E[u]\) and seeks the solution as the minimiser of \(E\); discretisation is obtained by restricting \(E\) to a finite trial space and minimising there. 
By contrast, the Galerkin method starts from the weak form of the PDE (often obtained by integration by parts) and enforces that the residual be orthogonal to a chosen test space. 
Consequently, Ritz methods require that the PDE admit a variational (Euler-Lagrange) formulation (typically self‑adjoint and coercive), whereas Galerkin projection is more general and applies even when no energy functional exists. 
Natural boundary conditions arise automatically from the variational derivative in the Ritz setting, while in Galerkin formulations essential boundary conditions are imposed on the trial space and natural conditions appear in the weak form.

Schematically, the methods discussed below can be separated as follows:
\[
    \begin{array}{c|c|c}
    \text{method} & \text{neural trial representation} & \text{physics enforcement} \\
    \hline
    \text{PINN} & u_\theta & \text{pointwise strong residual} \\
    \text{DGM} & u_\theta\ \text{(originally gated)} & \text{pointwise strong residual with random batches} \\
    \text{Deep Ritz} & u_\theta & \text{energy/variational minimisation} \\
    \text{vPINN} & u_\theta & \text{finite Petrov--Galerkin test space}
    \end{array}
\]

\subsubsection{The Deep Ritz Method}

In \cite{yu2018deep}, authors aim at solving the Ritz problem
\[
    \min_{u\in H} I(u), \qquad I(u) \defeq \int_{\Omega} \left( \half |\nabla u(x)|^2 - f(x) u(x)  \right) \dd x \,,
\]
where $H$ is the set of admissible functions, or trial functions. To solve it, they build on the following ideas:
\begin{itemize}
    \item[1.] Approximate the trial function with multi-layer perceptron;
    \item[2.] Use a numerical quadrature rule for the functional;
    \item[3.] Use an algorithm for solving the final optimization problem.
\end{itemize}

If you think that these ideas are similar to the PINN concept, well... you are right. This work originate in the same year of PINNs, roughly two months \textit{earlier}. 

In particular, they used a 5-layer MLP with two skip-connections 
\[
    x \mapsto y_i=\lambda_1(x) \mapsto y_2=\lambda_2(y_1) \mapsto y_3 = \lambda_3(y_2 + r_1(x)) \mapsto y_4 = \lambda_4(y_3) \mapsto u = \lambda_5(y_4 + r_2(y_2)))
\]
where $r_1, r_2$ are the two residual connections. In this case, they get a $\hat{u}(\theta; x)$ approximator, and thus defining
\[
    g(\theta, x) \defeq \half |\nabla \hat{u}(\theta; x)|^2 - f(x) \hat{u}(\theta; x) \,,
\]
we get the optimisation problem
\[
    \min_{\theta \in \Theta} \LL (\theta) \,, \quad \LL(\theta) = \int_\Omega  g(\theta, x) \dd x \,.
\]

The numerical quadrature rule they use to solve the problem is just Stochastic Gradient Descent over $N$ domain points sampled i.i.d., so that
\[
    \theta_{n+1} = \theta_n - \eta \nabla_\theta \left[ \frac{1}{N} \sum_{j=1}^N g(\theta_n, x_{j;n})  \right] ,
\]
where they \textit{re-draw} the $N$ points at each epoch $n$ (as denoted by $x_{j;k}$).

\begin{figure}[t]
    \centering
    \begin{subfigure}[t]{0.48\textwidth}
        \centering
        \includegraphics[width=\textwidth]{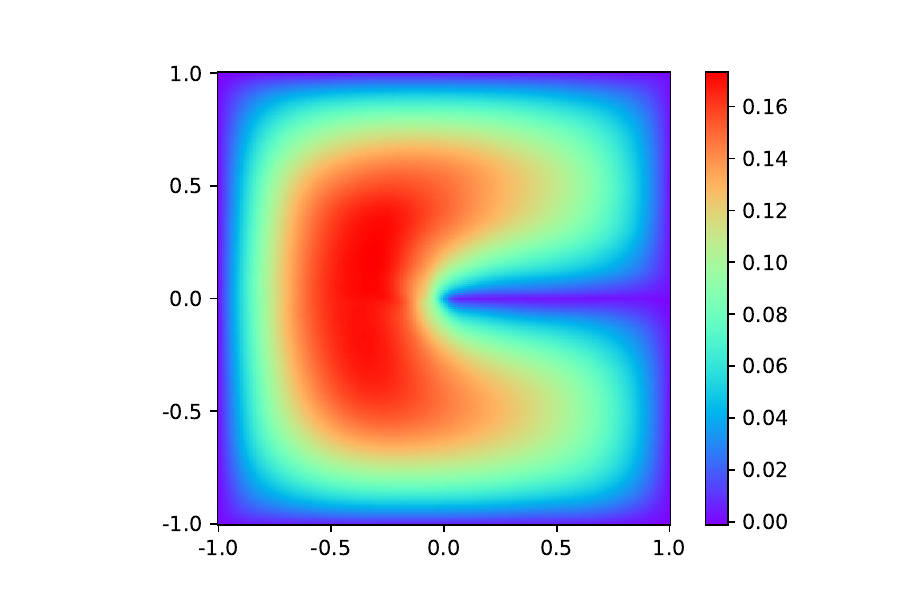}
        \caption{Solution of Deep Ritz method, 811 parameters. From \cite{yu2018deep}.}
    \end{subfigure}
    \hfill
    \begin{subfigure}[t]{0.48\textwidth}
        \centering
        \includegraphics[width=\textwidth]{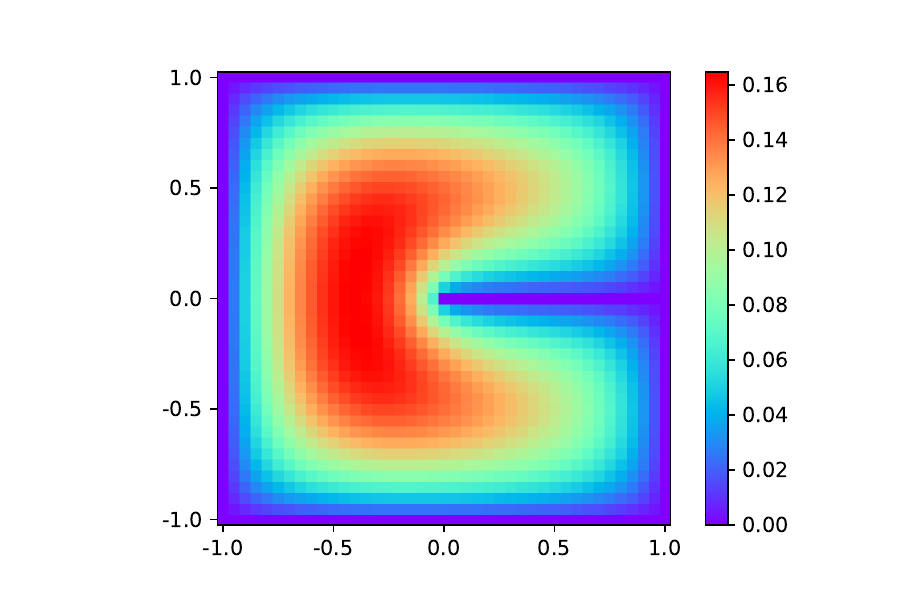}
        \caption{Solution with FDM, 1,681 parameters. From \cite{yu2018deep}.}
    \end{subfigure}
    \label{fig:4:3:DeepRitz}
\end{figure}

\noindent In practice, they want to solve the Cauchy problem
\begin{align}
    - \Delta u(x)  &= 0 \,, \quad x \in  \Omega \,, \\
    u(x) &= r^{\half} \sin \frac{\theta}{2} = u_D(x) \,, \quad x \in \partial \Omega \,;
\end{align}
this problem is a well known problem, since its solution suffers from a corner-singularity, caused by the nature of the domain \cite{strang1973analysis}.  Additionally, this problem has an exact solution
\[
    u^* (x) = r^{\half} \sin \frac{\theta}{2} \,.
\]
This allows for error comparison of standard Ritz method and DRM. To handle the B.C., they minimised 
\[
    I(u) =\half  \int_\Omega |\nabla \hat{u}(\theta; x)|^2 \dd x + \beta \int_{\partial \Omega} [u(x) - u_D(x)]^2 \dd x \,,
\]
with $\beta$ hyper parameter selected to be $\beta = 500$. 
With a simple MLP with 811 trainable parameters, they got the solution with $L_2$ error of $0.0072$; additional results are shown in \cref{tab:4:3:3:DRMvsRM}; for the visual comparison, see \cref{fig:4:3:DeepRitz}. 

\begin{table}[hb]
\caption{Error of Deep Ritz method (DRM) and finite difference. From \cite{yu2018deep}.
method (FDM)}\label{tab:4:3:3:DRMvsRM}
\begin{center}
\begin{tabular}{c|ccccc}
\hline
Method  & Blocks Num & Parameters & relative $L_2$ error \\ \hline
DRM & 3 & 591 & 0.0079\\ \hline
& 4 & 811 & 0.0072\\ \hline
& 5 & 1031 & 0.00647\\ \hline
& 6 & 1251 & 0.0057\\ \hline
FDM & & 625 & 0.0125\\ \hline
& & 2401 & 0.0063\\ \hline
\end{tabular}
\end{center}
\end{table}

\subsubsection{The Deep Galerkin Method}\label{subsec:4:3:3:DGM}

In \cite{sirignano2018dgm}, instead, the authors introduce the Deep Galerkin Method (DGM); starting from the Cauchy problem
\begin{align}
    \partial_t u(t,x) +L[u(t,x)] &= 0 \,, \quad x\in [0,T]\times\Omega \,,\\
    u(t=0, x) &= u_0 (x) \,, \\
    u(t,x) &= g(t, x) \,, \qquad x\in \partial \Omega \,,
\end{align}
where $\Omega \subset \RR^d$. The DGM algorithm approximates $u(t,x)$ with a deep neural network $f(t,x; \theta)$ where $\theta \in \mathbb{R}^K$ are the neural network's parameters. The DGM algorithm is 
\begin{itemize}
    \item[1.] Generate random points $(t_n, x_n)$ from $[0,T] \times \Omega$ and $(\tau_n, z_n)$ from $[0,T] \times \partial \Omega$ according to respective probability densities $\nu_1$ and $\nu_2$.  Also, draw the random point $w_n$ from $\Omega$ with probability density $\nu_3$.
    \item[2.] Calculate the squared error $G(\theta_n ,s_n)$ at the randomly sampled points $s_n = \{ (t_n, x_n), ( \tau_n, z_n), w_n \}$ where:
    \begin{align}\label{eq:4:3:2:2:dgmloss}
        G(\theta_n , s_n ) = &   \bigg{(} \frac{\partial f}{\partial t}(t_n ,x_n; \theta_n) +  \mathcal{L} f(t_n,x_n; \theta_n) \bigg{)}^2 \\
        & \quad +   \bigg{(} f(\tau_n ,z_n ; \theta_n ) - g(\tau_n,z_n) \bigg{)}^2 + \bigg{(}  f(0, w_n; \theta_n) - u_0(w_n) \bigg{)}^2. 
    \end{align}
    \item[3.] Take a descent step at the random point $s_n$:
    \begin{eqnarray*}
    \theta_{n+1} = \theta_n - \alpha_n \nabla_{\theta} G(\theta_n, s_n)  \notag
    \end{eqnarray*}
    \item[4.] Repeat until convergence criterion is satisfied.
\end{itemize}

\noindent Do you have a sensation of \textit{d\'ej\`a vu}? well, you are (almost) right. This is the same basic recipe we gave for PINNs \cref{tbox:3:1:vanilla_pinn}. Indeed, as outlined in \cite{jentzen2023mathematical}, the only major difference between the (original paper version of) PINN and DGM is the \textit{sampling strategy}:
\begin{itemize}
    \item PINNs draws a large set of points \textit{before} the training process, and, per each epoch, they draw a subsample from there;
    \item DGM re-draws each sample \textit{per-epoch}. 
\end{itemize}

Beyond the sampling strategy, the original DGM paper introduced a specialized architecture inspired by Long-Short Term Memory (LSTM) recurrent neural network units \cite{hochreiter1997long, staudemeyer2019understanding, scardapane2024alice}. This architecture uses gating mechanisms to control information flow across layers, which proved effective for high-dimensional PDEs.  Explicitly, the architecture  is \cite{sirignano2018dgm}
\begin{align}
S^{1} &=  \sigma( W^1 \mathbf{x} + b^1 ), \notag \\
Z^{\ell} &=  \sigma ( U^{z,\ell} \mathbf{x}  + W^{z,\ell} S^{\ell} + b^{z,\ell} ), \phantom{....} \ell = 1, \ldots, L, \notag \\
G^{\ell} &=  \sigma ( U^{g,\ell} \mathbf{x}  + W^{g,\ell} S^1 + b^{g,\ell} ), \phantom{....} \ell = 1, \ldots, L, \notag \\
R^{\ell} &=  \sigma ( U^{r,\ell} \mathbf{x}  + W^{r,\ell} S^{\ell} + b^{r, \ell} ), \phantom{....} \ell = 1, \ldots, L, \notag \\
H^{\ell} &=  \sigma ( U^{h,\ell} \mathbf{x}  + W^{h,\ell}  ( S^{\ell} \odot R^{\ell} ) + b^{h, \ell} ), \phantom{....} \ell = 1, \ldots, L, \notag \\
S^{\ell+1} &=  (1 - G^{\ell} ) \odot H^{\ell} + Z^{\ell} \odot S^{\ell}, \phantom{....} \ell = 1, \ldots, L, \notag \\
f(t,x; \theta) &= W S^{L+1} + b,
\label{eq:4:3:2:2:DGMArch}
\end{align}
where $\mathbf{x}  = (t,x)$, \textcolor{black}{the number of hidden layers is $L+1$}, and $\odot$ denotes element-wise multiplication (i.e., $z \odot v = \big{(} z_0 v_0, \ldots, z_N v_N \big{)}$). The parameters are
\begin{eqnarray}
    \theta = \bigg{\{} W^1, b^1, \bigg{(} U^{z, \ell}, W^{z, \ell}, b^{z, \ell} \bigg{)}_{\ell=1}^L,  \bigg{(} U^{g, \ell}, W^{g, \ell}, b^{g, \ell} \bigg{)}_{\ell=1}^L,  \notag \\
    \bigg{(} U^{r, \ell}, W^{r, \ell}, b^{r, \ell} \bigg{)}_{\ell=1}^L, 
    \bigg{(} U^{h, \ell}, W^{h, \ell}, b^{h, \ell} \bigg{)}_{\ell=1}^L, W, b \bigg{\}}. 
\end{eqnarray}
In later literature, DGM may refer either to the overall Sirignano–Spiliopoulos solver or, more narrowly, to its gated neural architecture. 

\begin{remark}
    We have seen that the DGM objective \cref{eq:4:3:2:2:dgmloss} is  a sum of pointwise squared strong residuals, so, mathematically speaking, the DGM method is more similar to a strong-form residual least squares or a meshfree collocation method, rather than a Galerkin method; the authors themselves say \cite{sirignano2018dgm}:
    \quotespace
    \begin{quote}
        \say{\textit{We call the algorithm a “Deep Galerkin Method (DGM)” since it is similar in spirit to Galerkin methods, with the solution approximated by a neural network instead of a linear combination of basis functions.}}
    \end{quote}
    \quotespace

    Thus, we are \textit{not} facing a true weak-formulation based method, which we will encounter instead in the next subsection.
\end{remark}

\subsubsection{Variational PINNs}

The variational PINN (vPINN) approach \cite{kharazmi2019variational} instead, is fully inspired by the weak form of the PDE; so, to follow the notation of \cite{kharazmi2019variational}, if PINN aim to solve a problem in the form strong form
\begin{align}
\mathcal{L}^\textbf{q} u(\textbf{x}) & = f(\textbf{x}), \quad \textbf{x}\in \Omega 
\\
u(\textbf{x}) & = h(\textbf{x}), \quad \textbf{x}\in \partial \Omega
\end{align}
using a DNN $u_{NN}(\theta; x)$ to approximate the solution, and optimising the parameters $\theta$ by minimising the \textit{strong-form residual} 
\begin{align}
%
r(\textbf{x}) &
= \mathcal{L}^\textbf{q} u_{NN}(\textbf{x}) - f(\textbf{x}), \quad \forall \textbf{x} \in \textbf{x}_r,
\\ \nonumber
r^b(\textbf{x}) 
&= u_{NN}(\textbf{x}) - h(\textbf{x}) , \,\,\quad\quad \forall \textbf{x} \in \textbf{x}_u.
\end{align}
This give rise to \emph{strong-form loss function} as 
\begin{align}
L^{\mathfrak{s}} 
& = L_{r}^{\mathfrak{s}} + L_{u} ,
\\ \nonumber
L_{r}^{\mathfrak{s}}
& 
= \frac{1}{N_r} \sum_{i = 1}^{N_r} \Big|r(\textbf{x}_{r_i})\Big|^2 ,
\quad
L_{u}
= \tau \, \frac{1}{N_u} \sum_{i = 1}^{N_u} \Big|r^b(\textbf{x}_{u_i}) \Big|^2,
\end{align}
in which $\tau $ denotes a penalty parameter. Here, the superscript $\mathfrak{s}$ refers to the loss function associated with the strong form of residual, to distinguish between this from and other considered cases. In this setting, the problem is solved by 
\begin{align}
\text{find } \tilde u(\textbf{x}) = u_{NN}(\textbf{x}; \textbf{w}^*, \textbf{b}^*) \text{ such that } \{\textbf{w}^*, \textbf{b}^*\} = \text{argmin}(L^{\mathfrak{s}}(\textbf{w}, \textbf{b})).
\end{align}

vPINNs instead aim to solve the \textit{weak-form} formulation of the differential problem
\note{In this sense, vPINNs are actually true \textit{Deep Galerkin Method}, i.e. the Deep learning approach to (Petrov-)Galerkin Methods, unlike the DGM method of \cref{subsec:4:3:3:DGM}. But it is what it is.}. 

The goal is thus to find an approximation form $ \tilde u (\textbf{x}) = u_{NN}(\textbf{x}; \textbf{w} , \textbf{b})$ with weights and biases $\{\textbf{w} , \textbf{b} \}$, respectively. Let $v(\textbf{x})$ be a properly chosen test function. The differential problem can be recast in the variational form
\begin{align}
&
\left( \mathcal{L}^\textbf{q} u_{NN}(\textbf{x}), v(\textbf{x}) \right)_{\Omega} = \left( f(\textbf{x}) , v(\textbf{x}) \right)_{\Omega} 
\\
& u(\textbf{x}) = h(\textbf{x}), \quad \textbf{x}\in \partial \Omega
\end{align}
where $(\cdot , \cdot)$ denotes the inner product. This variational form leads to the \emph{variational residual}, defined as  
\begin{align}
%
& \mathcal{R} 
= \left( \mathcal{L}^\textbf{q} u_{NN} , v \right)_{\Omega},
\quad
F = \left( f, v \right)_{\Omega},
\end{align}
which is enforced for any admissible test function $v_k , \,\, k = 1,2,\cdots$ and $r^b$ has the same form as in the strong form. 

Hence, the authors of \cite{kharazmi2019variational} construct a discrete finite dimensional space $V_K $ by choosing a finite set of admissible test functions and let $$V_K = \text{span} \lbrace v_k , \, k=1, 2, \cdots, K \rbrace. $$ Subsequently, they define the \emph{variational loss function} as
\begin{align}
&
L^{\mathfrak{v}} = L_{R}^{\mathfrak{v}} + L_{u} ,
\\ \nonumber
&
L_{R}^{\mathfrak{v}} = \frac{1}{K} \sum_{k = 1}^{K} \Big| \mathcal{R}_k - F_k \Big|^2 ,
\quad
L_{u} = \tau \frac{1}{N_u} \sum_{i = 1}^{N_u} \Big|r^b(\textbf{x}_{u_i}) \Big|^2,
\end{align}
in which $\tau $ denotes the penalty parameter. Here, the superscript $\mathfrak{v}$ refers to the loss function associated with the variational form of residual. In this setting, the problem is this 
\begin{align}
\text{find } \tilde u(\textbf{x}) = u_{NN}(\textbf{x}; \textbf{w}^*, \textbf{b}^*) \text{ such that } \{\textbf{w}^*, \textbf{b}^*\} = \text{argmin}(L^{\mathfrak{v}}(\textbf{w}, \textbf{b})). 
\end{align}

\begin{remark}
    In \cref{sec:2:2:FEM}, we have seen that an important advantage of the variational formulation is that integration by parts can lower the differential order that must be evaluated through automatic differentiation. For example, for
    \[
        -\Delta u=f,
    \]
    testing against $v$ gives
    \[
        \int_\Omega(-\Delta u_\theta)v\,\dd x
        =
        \int_\Omega f v\,\dd x.
    \]
    After integration by parts,
    \[
        \int_\Omega \nabla u_\theta\cdot\nabla v\,\dd x
        -
        \int_{\partial\Omega}\partial_nu_\theta\,v\,\dd s
        =
        \int_\Omega fv\,\dd x.
    \]
    If $v|_{\partial\Omega}=0$, the boundary term vanishes, so only first derivatives of $u_\theta$ are required rather than the second derivatives appearing in the strong Poisson residual. This reduction in differential order is one of the main motivations for the variational construction \cite{kharazmi2019variational}.
    
    The inner products defining $R_k$ must themselves be evaluated numerically, so quadrature accuracy becomes part of the vPINN discretisation. Thus vPINNs trade pointwise strong-form collocation for a finite collection of numerically integrated residual moments: the method is neural in its trial space, variational in its testing procedure, and still numerical in its quadrature.

\end{remark}

\subsection{Spectral Bias and Fourier Embeddings}\label{subsec:4:4:3:SpectralBias}

An outstanding work \cite{rahaman2019spectral} presented\footnote{Additional studies \cite{xu2019training, xu2019frequency} help shed light in what they called the \textit{frequency principle} (F-Principle).}, using Fourier Analysis, a relevant property of (deep ReLU) neural networks: 
during gradient-based fitting, low-frequency components of a target function tend to be learned earlier than high-frequency components. They refer to this frequency-dependent learning preference as \textit{spectral bias}\note{It is important to distinguish this statement from a limitation of expressive power. ReLU networks can represent functions with substantial high-frequency content; \textit{spectral bias} concerns instead which components are preferentially reached, and how rapidly they are fitted, during training.}.

\begin{figure}[t]
    \centering
    \begin{subfigure}[t]{0.48\textwidth}
        \centering
        \includegraphics[width=\textwidth]{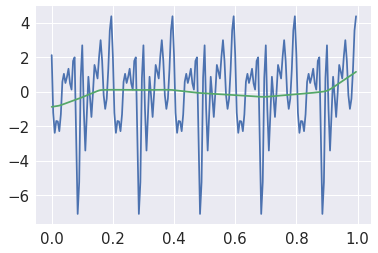}
        \caption{Iteration 100. From \cite{rahaman2019spectral}.}
    \end{subfigure}
    \hfill
    \begin{subfigure}[t]{0.48\textwidth}
        \centering
        \includegraphics[width=\textwidth]{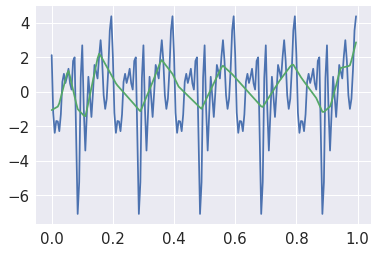}
        \caption{Iteration 1000. From \cite{rahaman2019spectral}.}
    \end{subfigure}

    \vspace{0.8em} 

    \begin{subfigure}[t]{0.48\textwidth}
        \centering
        \includegraphics[width=\textwidth]{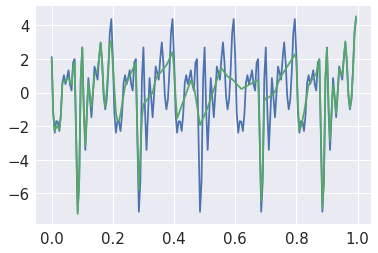}
        \caption{Iteration 10000. From \cite{rahaman2019spectral}.}
    \end{subfigure}
    \hfill
    \begin{subfigure}[t]{0.48\textwidth}
        \centering
        \includegraphics[width=\textwidth]{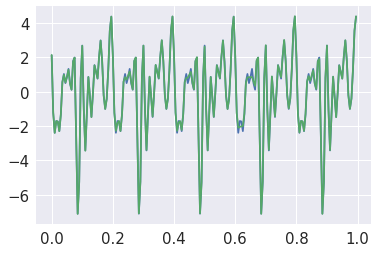}
        \caption{Iteration 80000. From \cite{rahaman2019spectral}.}
    \end{subfigure}

    \caption{The learnt function (green) overlayed on the target function (blue) as the training progresses. The target function is a superposition of sinusoids of frequencies $\kappa = (5, 10, \dots, 45, 50)$, equal amplitudes and randomly sampled phases. From \cite{rahaman2019spectral}.}
    \label{fig:4:4:SpectralBiasEx1}
\end{figure}

A simple, yet illustrative example they propose is the following:
Given frequencies $\kappa = (k_1, k_2, ...)$ with corresponding amplitudes $\alpha = (A_1, A_2, ...)$, and phases $\phi = (\varphi_1, \varphi_2, ...)$, define the mapping $\lambda: [0, 1] \to \mathbb{R}$ given by
\[
\lambda(z) = \sum_i A_i \sin(2\pi k_i z + \varphi_i).
\]
A 6-layer deep 256-unit wide ReLU network $f_\theta$ is trained to regress $\lambda$ with $\kappa = (5, 10, ..., 45, 50)$ and $N=200$ input samples spaced equally over $[0, 1]$; 
its spectrum $\tilde{f}_\theta(k)$ in expectation over $\varphi_i \sim U(0, 2 \pi)$ is monitored as training progresses. The amplitude are a linear ramp from $A_1 = 0.1$ to $A_{10} = 1$. \cref{fig:4:4:SpectralBiasEx1} shows  the learned function at intermediate training iterations in an example setting. The result is that  lower frequencies (i.e. smaller $k_i$'s) are regressed first, regardless of their amplitudes. 

But \textit{why} this happens? the point is that it happens because parameters that contribute towards expressing high-frequency components occupy \textit{a small volume} in the parameter space, \textit{regardless} of their amplitude.   

Authors also proved this fact. Let us sketch their main line of reasoning. The key insight is that ReLU networks have a Fourier spectrum that decays as $\mathcal{O}(k^{-n-1})$, meaning that high-frequency components occupy a vanishingly small volume in the parameter space. We sketch the main steps: first, decompose the network's Fourier transform into a sum of rational functions; second, show that the volume of parameters capable of representing high frequencies scales as $\mathcal{O}(k^{-\Delta-1})$.

Let us follow \cite{rahaman2019spectral} and define \textit{ReLU network} a scalar function $f: \R^d  \mapsto \R$ defined by a neural network with $L$ hidden layers of widths  $d_1, \cdots d_L$ and a single output neuron: 
\[
f(\xx) = \left(T^{(L+1)} \circ \sigma \circ T^{(L)} \circ \cdots \circ \sigma \circ T^{(1)}\right)(\xx)
\]
where each ${T}^{(k)}: \RR^{d_{k-1}} \to \RR^{d_k}$ is an affine function ($d_0 = d$ and $d_{L+1} = 1$) and $\sigma(\mathrm{u})_i = \max(0, u_i)$ denotes the ReLU activation function acting elementwise on a vector $\mathrm{u} = (u_1, \cdots u_n)$. In the standard basis, ${T}^{(k)}(\xx) = W^{(k)} \x + \mathbf{b}^{(k)} $ for some weight matrix $W^{(k)}$  and bias vector $\mathbf{b}^{(k)}$. 

The goal is now to study the structure of ReLU networks in terms of their Fourier representation\note{Strictly speaking, a general ReLU network on $\RR^d$ need not be integrable (or even square-integrable), since it may grow linearly at infinity. Accordingly, the Fourier transform appearing here should in general be understood in the sense of tempered distributions.}, 
\[
    f(\xx) \defeq \frac{1}{(2\pi)^d} \int \tilde{f}(\mathbf{k}) \, \e^{\ii\mathbf{k}\cdot \xx} \dd k,
\]
where $\tilde{f}(\mathbf{k}) \defeq \int f(\xx)\, \e^{- \ii \mathbf{k}\cdot \xx} \dd x$ is the Fourier transform.
We can now write down the following\note{For the proofs, see the Appendix C and D of \cite{rahaman2019spectral}.}

\begin{lemma}
Let
\[
    f_\theta(\mathbf{x})
    =
    \sum_{\epsilon}
    1_{P_\epsilon}(\mathbf{x})
    \left(
        W_\epsilon\mathbf{x}+b_\epsilon
    \right)
\]
be the piecewise-affine decomposition of a ReLU network over its linear
regions $P_\epsilon$. Then,
\[
    \tilde f_\theta(\mathbf{k})
    =
    -\ii
    \sum_{\epsilon}
    \frac{W_\epsilon\mathbf{k}}{\|\mathbf{k}\|_2^2}
    \,
    \tilde 1_{P_\epsilon}(\mathbf{k}),
\]
where
\[
    \tilde 1_{P}(\mathbf{k})
    =
    \int_P
    \e^{-\ii\mathbf{k}\cdot\mathbf{x}}\,
    \dd\mathbf{x},
\]
is the Fourier transform of the indicator function of $P$. 
\end{lemma}

\begin{lemma}
Let $P$ be a full-dimensional polytope in $\RR^d$. Its Fourier transform
can be written in the form
\[
    \tilde 1_P(\mathbf{k})
    =
    \sum_{n=0}^{d}
    \frac{
        D_n(\mathbf{k})\,1_{G_n}(\mathbf{k})
    }{
        \|\mathbf{k}\|_2^{\,n}
    },
\]
where $G_n$ is a finite union of $n$-dimensional subspaces orthogonal to
$n$-codimensional faces of $P$, and
\[
    D_n(\mathbf{k})=\Theta(1),
    \qquad
    \|\mathbf{k}\|_2\to\infty
\]
along the corresponding directions,  and $1_{G_n}$ the indicator over $G_n$. 
\end{lemma}

\noindent These two lemmas give the following

\begin{theorem}\label{theorem:4:3:4:SpectralThm1}
For a ReLU network $f_\theta$, the Fourier transform has the structured form
\[
    \tilde f_\theta(\mathbf{k})
    =
    \sum_{n=0}^{d}
    \frac{
        C_n(\theta,\mathbf{k})\,
        1_{H_n^\theta}(\mathbf{k})
    }{
        \|\mathbf{k}\|_2^{\,n+1}
    },
\]
where $H_n^\theta$ is a finite union of $n$-dimensional subspaces orthogonal
to $n$-codimensional faces of the linear regions of the network, and the
coefficients $C_n(\theta,\mathbf{k})$ remain bounded in magnitude along the
corresponding asymptotic directions.
\end{theorem}

The important consequence is that the spectral decay is strongly
\textit{direction-dependent}. For a generic direction $\hat{\mathbf{k}}$, the
polytope recursion reaches vertices and one obtains asymptotically
\[
    \left|
    \tilde f_\theta(k\hat{\mathbf{k}})
    \right|
    =
    \mathcal O(k^{-d-1}).
\]
Along exceptional directions orthogonal to faces of the linear regions,
the recursion may terminate earlier and the decay may be slower; in the
slowest non-trivial case discussed in \cite{rahaman2019spectral}, it can be
of order $\mathcal O(k^{-2})$.

Therefore, the result should not be interpreted as a single isotropic law
$\tilde f(k)=\mathcal O(k^{-n-1})$ with one fixed exponent $n$.

With these tools at hand, let us define what we mean by parameter space volume occupied by a certain frequence:
\begin{definition}
Given a ReLU network $f_{\theta}$ of fixed depth, width and weight clip $K$ with parameter vector $\theta$, an $\epsilon > 0$ and $\Theta = B^{\infty}_{K}(0)$ a $L^{\infty}$ ball around $0$, we define:
$$
\Xi_{\epsilon}(k) \defeq \{\theta \in \Theta :  \exists k' > k, |\tilde{f}_{\theta}(k')| > \epsilon \}
$$
as the set of all parameters vectors $\theta \in \Xi_{\epsilon}(k)$ that contribute more than an $\epsilon$ in expressing one or more frequencies $k'$ above a \textit{cut-off frequency} $k$.

The \textit{relative volume} of the parameter sub-space is thus defined as the integral over the set indicator function:
\[
\operatorname{vol}(\Xi_{\epsilon}(k) ) \defeq \int_{\theta \in \Theta } 1_{k}^{\epsilon}(\theta) \dd \theta . 
\]
\end{definition}

\begin{lemma}\label{lemma:4:3:4:indicatorbound}
Let $1_{k}^{\epsilon}(\theta)$ be the indicator function on $\Xi_{\epsilon}(k)$. Then:
$$
\exists \, \kappa > 0:  \forall k \ge \kappa, 1_{k}^{\epsilon}(\theta) = 0.
$$
Furthermore, this implies that we have $1_{k}^{\epsilon}(\theta) \le |\tilde{f}_{\theta}(k)|$ for large enough $k$ (i.e. for $k \ge \kappa$), since $|\tilde{f}_{\theta}(k)| \ge 0$.
\end{lemma}
\begin{proof}
From theorem \ref{theorem:4:3:4:SpectralThm1}, we know that
$|\tilde{f}_{\theta}(k)| = \mathcal{O}(k^{-\Delta - 1})$ for an integer $1 \le \Delta \le d$. In the worse case where $\Delta = 1$, we have that $\exists M < \infty: |\tilde{f}_{\theta}(k)| < \frac{M}{k^2}$. Now, simply select a $\kappa > \sqrt{\frac{M}{\epsilon}}$ such that $\frac{M}{\kappa^2} < \epsilon$. This yields that $|\tilde{f}_{\theta}(\kappa)| < \frac{M}{\kappa^2} < \epsilon$, and given that $\frac{M}{\kappa^2} \le \frac{M}{k^2} \,\forall \, k \ge \kappa$, we find $|\tilde{f}_{\theta}(k)| < \epsilon \,\forall\, k \ge \kappa$. Now by definition, $\theta \not\in \Xi_{\epsilon}(\kappa)$, and since $\Xi_{\epsilon}(k) \subseteq \Xi_{\epsilon}(\kappa)$, we have $\theta \not\in \Xi_{\epsilon}(k)$, implying $1_{k}^{\epsilon}(\theta) = 0 \,\forall\, k \ge \kappa$.
\end{proof}

\noindent We thus have the main

\begin{tcolorbox}[colframe=darkblue, colback=white, title=Spectral Volume Theorem]
The relative volume of $\Xi_{\epsilon}(k)$ w.r.t. $\Theta$ is $\mathcal{O}(k^{-\Delta - 1})$ where $1 \le \Delta \le d$. 
\end{tcolorbox}

\begin{proof}
For a large enough $k$, we have from remark \cref{lemma:4:3:4:indicatorbound}, the monotonicity of the Lebesgue integral and theorem \cref{theorem:4:3:4:SpectralThm1} that: 

\begin{align*}
\text{vol}(\Xi_{\epsilon}(k)) &= \int_{\theta \in \Theta} 1_{k}^{\epsilon}(\theta) d\theta \\
&\le \int_{\theta \in \Theta} |\tilde{f}_{\theta}(k)| d\theta = \mathcal{O}(k^{-\Delta - 1}) \text{vol}(\Theta) \\
&\implies \frac{\text{vol}(\Xi_{\epsilon}(k))}{\text{vol}(\Theta)} = \mathcal{O}(k^{-\Delta - 1}) .    
\end{align*}
\end{proof}

\begin{lstlisting}[caption={Spectral Bias}, label={lst:4:3:3:SpectralBias}, language=Python]
import os
import math
import numpy as np
from typing import List, Tuple

import torch
from torch import nn
from torch.utils.data import TensorDataset, DataLoader
import lightning as L

# Utility: target generator
def make_target(k_list: List[int], A_list: List[float], seed: int = 0) -> Tuple[callable, np.ndarray]:
    rng = np.random.RandomState(seed)
    phases = rng.uniform(0, 2 * np.pi, size=len(k_list))
    def target(z: np.ndarray) -> np.ndarray:
        z = np.asarray(z).reshape(-1)
        out = np.zeros_like(z, dtype=float)
        for k, A, phi in zip(k_list, A_list, phases):
            out += A * np.sin(2 * np.pi * k * z + phi)
        return out
    return target, phases

# MLP model (6 layers, 256)
class MLP(nn.Module):
    def __init__(self, in_dim=1, hidden=256, depth=6, out_dim=1):
        super().__init__()
        layers = []
        layers.append(nn.Linear(in_dim, hidden))
        layers.append(nn.ReLU())
        for _ in range(depth - 2):
            layers.append(nn.Linear(hidden, hidden))
            layers.append(nn.ReLU())
        layers.append(nn.Linear(hidden, out_dim))
        self.net = nn.Sequential(*layers)
        self._init_weights()

    def _init_weights(self):
        for m in self.net:
            if isinstance(m, nn.Linear):
                nn.init.xavier_uniform_(m.weight)
                if m.bias is not None:
                    nn.init.zeros_(m.bias)

    def forward(self, x):
        return self.net(x)

# Single LightningModule encapsulating the experiment
class SpectralBiasPL(L.LightningModule):
    def __init__(
        self,
        k_list: List[int],
        A_list: List[float],
        N_train: int = 200,
        x_eval_n: int = 4096,
        hidden: int = 256,
        depth: int = 6,
        lr: float = 1e-3,
        seed: int = 0,
    ):
        super().__init__()
        self.save_hyperparameters(ignore=["k_list", "A_list"])
        self.k_list = np.array(k_list)
        self.A_list = np.array(A_list)
        self.N_train = N_train
        self.lr = lr
        # target function and phases
        self.target_fn, self.phases = make_target(k_list, A_list, seed=seed)
        # training samples (equally spaced)
        self.x_train = np.linspace(0.0, 1.0, N_train, endpoint=False)
        self.y_train = self.target_fn(self.x_train)
        # dense eval grid for spectrum and plotting
        self.x_eval = np.linspace(0.0, 1.0, x_eval_n, endpoint=False)
        # model
        self.model = MLP(in_dim=1, hidden=hidden, depth=depth, out_dim=1)
        # loss
        self.criterion = nn.MSELoss()
        # for deterministic behavior
        L.seed_everything(seed)

    def forward(self, x: torch.Tensor) -> torch.Tensor:
        return self.model(x)

    def train_dataloader(self):
        X = torch.tensor(self.x_train.reshape(-1, 1), dtype=torch.float32)
        Y = torch.tensor(self.y_train.reshape(-1, 1), dtype=torch.float32)
        ds = TensorDataset(X, Y)
        # full-batch gradient descent: batch_size = N_train
        return DataLoader(ds, batch_size=self.N_train, shuffle=False)

    def configure_optimizers(self):
        # full-batch Adam as in the paper
        opt = torch.optim.Adam(self.parameters(), lr=self.lr)
        return opt
    
    def training_step(self, batch, batch_idx):
        x, y = batch
        y_hat = self.model(x)
        loss = self.criterion(y_hat, y)
        # log loss for monitoring
        self.log("train/loss", loss, on_step=True, on_epoch=False, prog_bar=True)
        # periodic logging of spectrum and function
        step = int(self.global_step)
        # if wanted to reproduce plot figure, do:
        if (step % self.log_every_steps == 0) or (step == 0):
            self._save_function_and_spectrum(step) # to implement
        return loss
\end{lstlisting}

\subsubsection{Fourier Feature Embedding and Fourier Network}

This spectral property of MLPs is somehow a beneficial property for standard deep learning tasks, since it means that its training may be more stable under noise oscillation in data.  When the target signal is dominated by smooth, low-frequency components, a preference for low frequencies may act as an implicit smoothness bias and can reduce sensitivity to high-frequency perturbations. On the other hand, when fine-scale or oscillatory components are essential to the target function, the same bias can severely slow down their learning.


This issue is particularly important for \emph{coordinate-based MLPs}, i.e.\ networks that directly map low-dimensional coordinates $\mathbf{x}=(x,y,z,\ldots)$ to a signal value. In such settings, high-frequency components of the target may converge much more slowly than low-frequency ones, with important consequences in computer vision, implicit neural representations, and, of course, PINNs \cite{tancik2020fourier,toscano2025pinns}.

The solution is to use a \textit{positional encoding} of the input coordinate data, originally introduced in \cite{rahimi2007random}, and later on popularised  by the revolutionary \textit{Neural Radiance Field} (NeRF)  paper \cite{mildenhall2021nerf}:
\begin{equation}
    \begin{split}
        \xx \mapsto\gamma(\xx) = \; & \left[
         a_1 \cos (2 \pi \mathbf{b}_1 \cdot \mathbf{x}), \dots , a_m  \cos (2 \pi \mathbf{b}_m \cdot \mathbf{x}) , \right. \\
        & \quad \left.
         a_1 \sin (2 \pi \mathbf{b}_1 \cdot \mathbf{x}), \dots , a_m  \sin (2 \pi \mathbf{b}_m \cdot \mathbf{x}) 
        \right] \,,
    \end{split}
\end{equation}
where $(a_i, \mathbf{b}_i)_{i=1, \ldots, m}$ are parameters, which can be \textit{fixed} (as in \cite{mildenhall2021nerf}) or \textit{learnable}.

\begin{figure}[t]
\centering
\resizebox{0.99\textwidth}{!}{%
\begin{tabular}{@{}c@{\,\,}c@{\,\,}c@{}}
    \makecell{
        \includegraphics[width=0.175\textwidth]{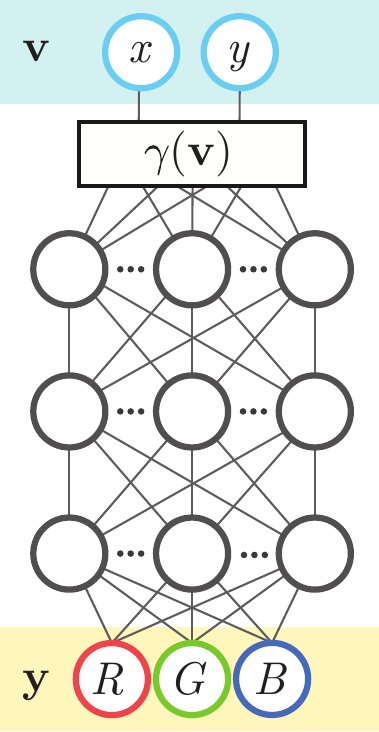} \\
        \scriptsize{(a) Coordinate-based MLP}
    }
    &
    \begin{tabular}{@{}c@{\,\,}c@{\,\,}c@{\,\,}c@{\,\,}c@{\,\,}c@{\,\,}c@{}}
        \scriptsize\rotatebox{90}{\quad No Fourier features }  &
        \scriptsize\rotatebox{90}{\quad\quad\, $\gamma(\xx) = \xx$ }  &
        \includegraphics[width=0.177\textwidth]{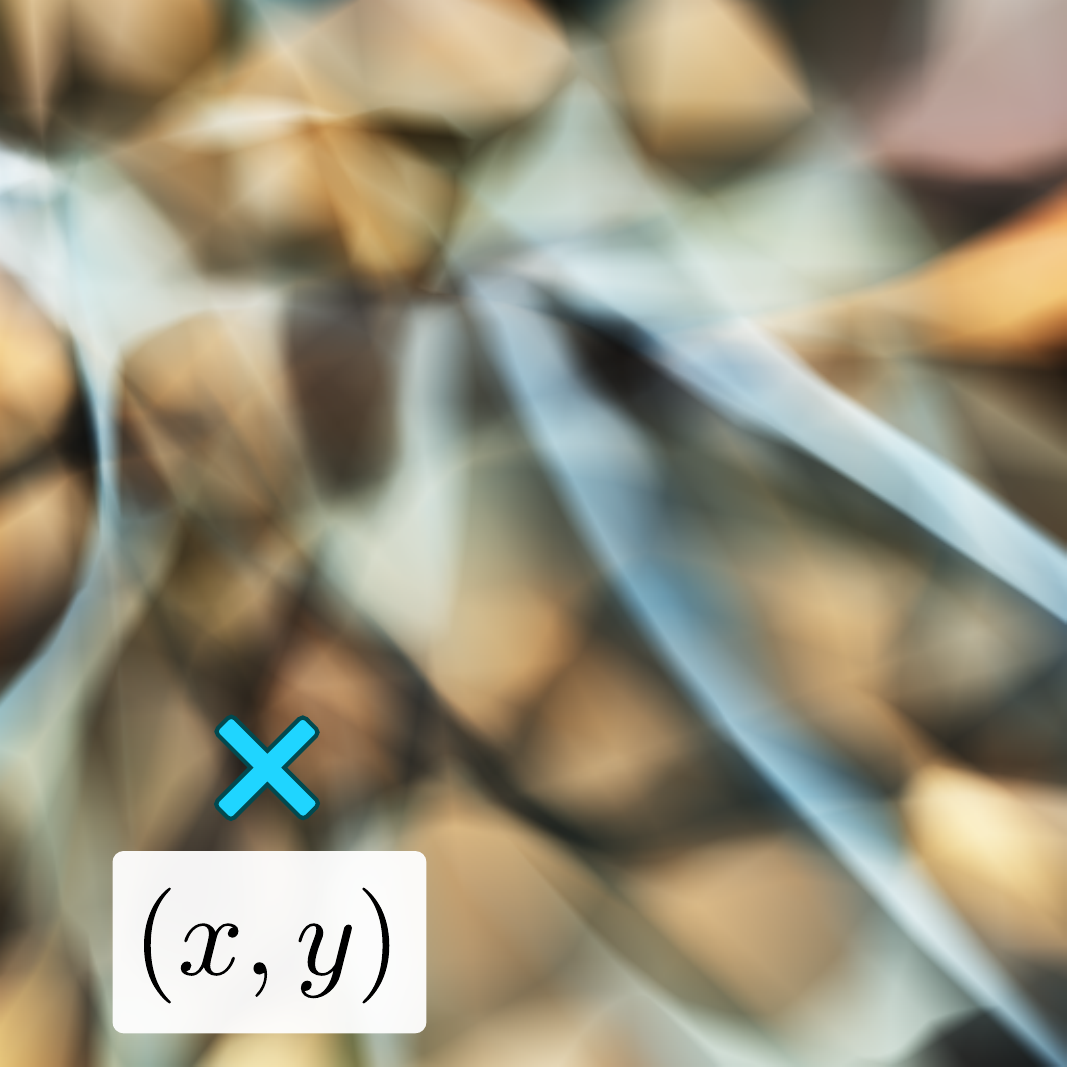} &
        \includegraphics[width=0.177\textwidth]{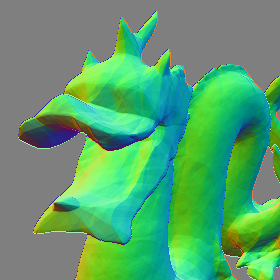} &
        \includegraphics[width=0.177\textwidth]{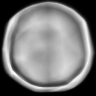} &
        \includegraphics[width=0.177\textwidth,trim= 100 100 50 50,clip]{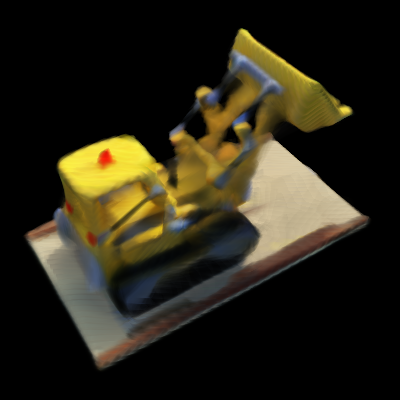} \\
        \scriptsize\rotatebox{90}{\,\, With Fourier features }  &
        \small\rotatebox{90}{\,\,\, $\gamma(\xx) = \operatorname{FF}(\xx)$ }  & \includegraphics[width=0.177\textwidth]{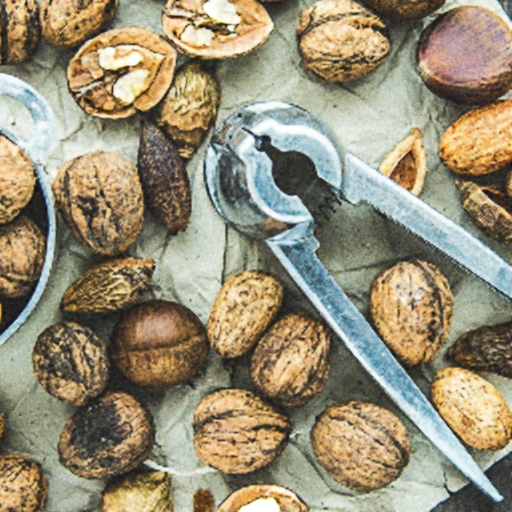} &
        \includegraphics[width=0.177\textwidth]{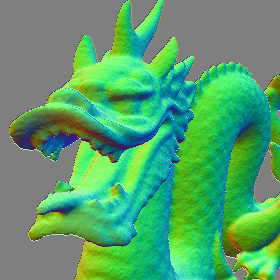} & 
        \includegraphics[width=0.177\textwidth]{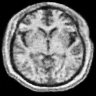} & 
        \includegraphics[width=0.177\textwidth,trim= 100 100 50 50,clip]{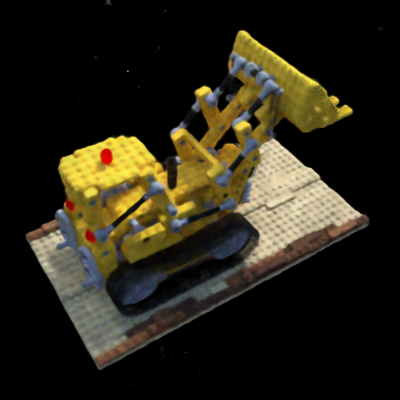} \\
        && \scriptsize{(b) Image regression}
        & \scriptsize{(c) 3D shape regression}
        & \scriptsize{(d) MRI reconstruction}
        & \scriptsize{(e) Inverse rendering} \\
        && \scriptsize{$\!(x,\!y)\!\rightarrow \textrm{RGB}$}
        & \scriptsize{$\!(x,\!y,\!z)\!\rightarrow \textrm{occupancy}$}
        & \scriptsize{$\!(x,\!y,\!z)\!\rightarrow \textrm{density}$}
        & \scriptsize{$\!(x,\!y,\!z)\!\rightarrow\!\textrm{RGB, density}\!$}
    \end{tabular}
\end{tabular}
} 
\caption{Fourier features improve the results of coordinate-based MLPs for a variety of high-frequency low-dimensional regression tasks, both with direct (b, c) and indirect (d, e) supervision. We visualize an example MLP (a) for an image regression task (b), where the input to the network is a pixel coordinate and the output is that pixel's color. Passing coordinates directly into the network (top) produces blurry images, whereas preprocessing the input with a Fourier feature mapping (bottom) enables the MLP to represent higher frequency details. From \cite{tancik2020fourier}.
}
\label{fig:4:3:3:1:FourierNetteaser}
\end{figure}

What we achieve is that, with this positional encoding, we uplift the $\RR^d \ni \xx$ data space to a $\RR^{2m} \ni \gamma(\xx)$, where we have already either decompose the relevant Fourier modes, i.e., the model starts learning from a set of $m$ Fourier modes, or we \textit{learn} these relevant Fourier modes. The full networks is thus \cite{tancik2020fourier, wang2021eigenvector}
\begin{eqnarray}\label{eq:4:3:3:1:FourierNet}
    u_{NN} (\theta; \xx) = \mathrm{W}_n \{\lambda_{n-1} \circ \cdots \circ \lambda_1 \circ\gamma \} (\xx) + \mathbf{b}_n \,, \quad \lambda_i (\xx_i) \defeq \phi_i (\mathrm{W}_i \xx_i + \mathbf{b}_i).
\end{eqnarray}

\begin{remark}
    The role of the Fourier embedding is more subtle than simply ``providing the correct Fourier modes'' to the network. In the NTK analysis of
    \cite{tancik2020fourier}, the sinusoidal feature map induces a stationary
    kernel
    \[
        k_\gamma(\mathbf{x},\mathbf{x}')
        =
        \gamma(\mathbf{x})^T\gamma(\mathbf{x}')
        =
        h_\gamma(\mathbf{x}-\mathbf{x}').
    \]
    Consequently, in the corresponding infinite-width NTK regime, the MLP acting
    on the embedded coordinates is governed by the composed stationary kernel
    \[
        k_{\mathrm{eff}}(\mathbf{x},\mathbf{x}')
        =
        h_{\mathrm{NTK}}
        \!\left(
            h_\gamma(\mathbf{x}-\mathbf{x}')
        \right).
    \]
    
    Changing the frequency distribution of the feature map therefore changes the
    spectral bandwidth of the effective kernel. A wider spectral support generally
    accelerates the convergence of higher-frequency components, while an
    excessively wide kernel may also increase high-frequency overfitting or
    aliasing. Thus, Fourier features modify both the optimization dynamics and the
    implicit spectral bias of the coordinate MLP.
\end{remark}

\subsubsection{Modified Fourier Network family}

The Fourier Network architecture of \cref{eq:4:3:3:1:FourierNet} is just the first step in the design of Fourier-modes-aware neural networks, especially in the PINN domain. 

\paragraph{Modified Fourier networks: } in Fourier network, a standard fully-connected neural network is used as the nonlinear mapping between the Fourier features and the model output. 

In \textit{modified Fourier networks} \cite{wang2021understanding}, two transformation layers are introduced to project the Fourier features to another learned feature space, and are then used to update the hidden layers through element-wise multiplications:
\begin{eqnarray}
    \phi_{i}(\mathbf{x}_i) = \left[1 - \sigma \left( \mathbf{W}_i \mathbf{x}_i + \mathbf{b}_i \right) \right] \odot \sigma \left( \mathbf{W}_{T_1} \phi_E + \mathbf{b}_{T_1} \right) \notag \\
    + \sigma \left( \mathbf{W}_i \mathbf{x}_i + \mathbf{b}_i \right) \odot \sigma \left( \mathbf{W}_{T_2} \phi_E + \mathbf{b}_{T_2} \right),     
\end{eqnarray}
where from now on $\phi_E$ is the \textit{learnable} Fourier encoding
\begin{eqnarray}
    \phi_E(\xx) = \big[ \sin \left( 2\pi \mathbf{B} \cdot \mathbf{x} \right); \cos \left( 2\pi  \mathbf{B} \cdot 
\mathbf{x} \right) \big]^T,
\end{eqnarray}
where $\mathbf{B} \in\RR^{n_f \times d}$ is a learnable embedding matrix. Furthermore, $\mathbf{W}_{T_{1,2}}$ and $\mathbf{b}_{T_{1,2}}$ are also learnable parameters. 

\paragraph{Highway Fourier Network: } Inspired by \cite{srivastava2015training, srivastava2015highway}, in \cite{NvidiaPhysicsNemoDocumentation} it was introduced the \textit{Highway Fourier network}. 
Highway Fourier networks incorporate adaptive gating units, inspired by highway networks \cite{srivastava2015training}, to control the flow of information. The gating mechanism allows the network to learn how much of the Fourier-encoded information to pass through to subsequent layers, which can be particularly useful for very deep architectures. The hidden layers in this network are
\begin{align}
    \phi_{i}(\mathbf{x}_i) =\; &\sigma \left( \mathbf{W}_i \mathbf{x}_i + \mathbf{b}_i \right) \odot \sigma_s \left( \mathbf{W}_{T} \phi_E + \mathbf{b}_{T} \right) \notag \\
    &+ \left( \mathbf{W}_P \mathbf{x} + \mathbf{b}_P \right) \odot \left (1 - \sigma_s \left( \mathbf{W}_{T} \phi_E + \mathbf{b}_{T} \right) \right).
\end{align}

\paragraph{Multi-scale Fourier Feature Network: } In \cite{wang2021eigenvector}, authors proposed a \textit{multi-scale Fourier feature network} architecture that aim to tackle partial differential equations exhibiting multi-scale behaviours. 

Multi-scale Fourier feature networks address multi-scale PDEs by applying multiple Fourier embeddings with different frequency ranges to the input. These embeddings are processed through shared network layers, and the outputs are concatenated before the final linear layer. This allows the network to simultaneously capture both low- and high-frequency features.

\begin{figure}[t]
    \centering
    \begin{subfigure}[t]{0.48\textwidth}
        \centering
        \includegraphics[width=\textwidth]{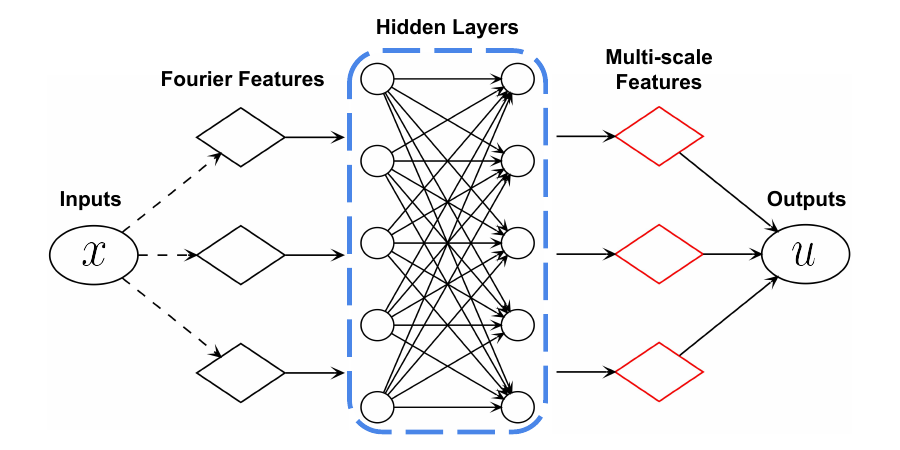}
        \caption{
        \textit{Multi-scale Fourier feature architecture}: multiple Fourier feature embeddings (initialised with different $\sigma$) are applied to input coordinates and then passed through the same fully-connected neural network, before the outputs are finally concatenated with a linear layer. From \cite{wang2021eigenvector}.}
    \end{subfigure}
    \hfill
    \begin{subfigure}[t]{0.48\textwidth}
        \centering
        \includegraphics[width=\textwidth]{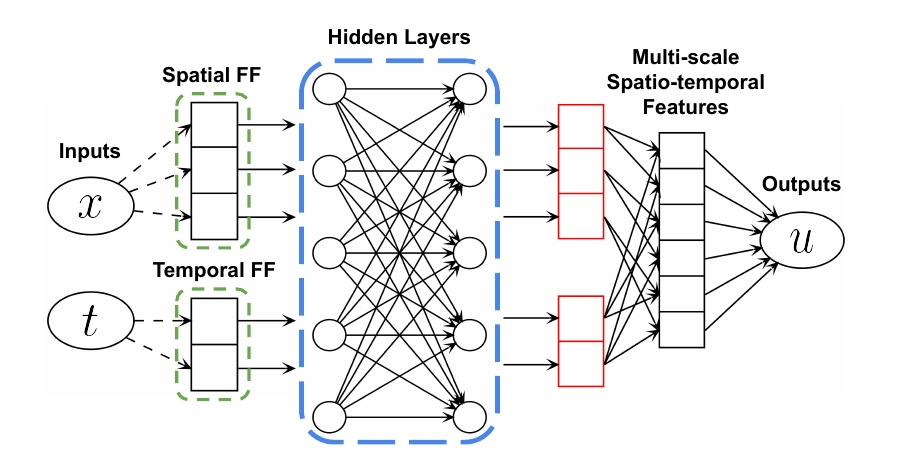}
        \caption{
        Spatio-temporal multi-scale Fourier feature architecture: Multi-scale Fourier feature architecture: multiple Fourier feature embeddings (initialised with different $\sigma$) are separately applied to spatial and temporal input coordinates and then passed through the same fully-connected neural network. Merging of the spatial and temporal outputs is performed using a point-wise multiplication layer, before obtaining the final outputs through a linear layer.
        From \cite{wang2021eigenvector}.}
    \end{subfigure}
    \caption{Visual representations of modified fourier networks.}
    \label{fig:4:3:3:1:ModFourNet}
\end{figure}

The forward pass of the multi-scale Fourier feature networks is given by
\begin{align}
     &\phi_{E}^{(i)}(\mathbf{x})=[\sin (2 \pi \mathbf{B}^{(i)} \cdot \mathbf{x}) ; \cos (2 \pi \mathbf{B}^{(i)} \cdot \mathbf{x})]^{T},  \quad \text{ for } i=1, 2, \dots, M  \notag\\
    &\mathbf{H}^{(i)}_1 = \sigma(\mathbf{W}_1 \cdot\phi_{E}^{(i)}(\mathbf{x})  + \mathbf{b}_1),  \hspace{24.5mm} \text{ for } i=1, 2, \dots, M \notag\\
    & \mathbf{H}^{(i)}_\ell = \sigma(\mathbf{W}_\ell \cdot \mathbf{H}^{(i)}_{\ell - 1}  + \mathbf{b}_\ell),  \hspace{27.5mm} \text{ for } \ell=2,   \dots, L;  i=1, 2, \dots, M \notag\\
    & \mathbf{u}_{NN}(\mathbf{x}, {\mathbf{\theta}}) = \mathbf{W}_{L+1} \cdot \left[  \mathbf{H}^{(1)}_L,  \mathbf{H}^{(2)}_L, \dots,   \mathbf{H}^{(M)}_L  \right] + \mathbf{b}_{L+1}.
\end{align}
For a visual representation of this architecture, see \cref{fig:4:3:3:1:ModFourNet} (a).

\paragraph{Spatio-temporal Multi-scale Fourier Feature Network:}
For time-dependent problems, distinct frequency scales may occur in the spatial and temporal coordinates. In \cite{wang2021eigenvector} authors therefore introduce separate families of random Fourier embeddings for space and time,
\begin{align}
    \gamma_x^{(i)}(\mathbf{x})
    &=
    \left[
        \cos(2\pi B_x^{(i)}\mathbf{x});
        \sin(2\pi B_x^{(i)}\mathbf{x})
    \right],
    &&
    i=1,\ldots,M_x,
    \notag\\
    \gamma_t^{(j)}(t)
    &=
    \left[
        \cos(2\pi B_t^{(j)}t);
        \sin(2\pi B_t^{(j)}t)
    \right],
    &&
    j=1,\ldots,M_t.
\end{align}
The entries of $B_x^{(i)}$ and $B_t^{(j)}$ are independently sampled from Gaussian distributions with characteristic scales $\sigma_{x,i}$ and $\sigma_{t,j}$, respectively, and are held fixed during training.

The spatial and temporal embeddings are passed through the \emph{same} fully-connected network,
\begin{align}
    H_{x,1}^{(i)}
    &=
    \sigma\!\left(
        W_1\gamma_x^{(i)}(\mathbf{x})+b_1
    \right),
    &&
    i=1,\ldots,M_x,
    \notag\\
    H_{t,1}^{(j)}
    &=
    \sigma\!\left(
        W_1\gamma_t^{(j)}(t)+b_1
    \right),
    &&
    j=1,\ldots,M_t,
    \notag\\
    H_{x,\ell}^{(i)}
    &=
    \sigma\!\left(
        W_\ell H_{x,\ell-1}^{(i)}+b_\ell
    \right),
    &&
    \ell=2,\ldots,L,
    \notag\\
    H_{t,\ell}^{(j)}
    &=
    \sigma\!\left(
        W_\ell H_{t,\ell-1}^{(j)}+b_\ell
    \right),
    &&
    \ell=2,\ldots,L.
\end{align}

At the last hidden layer, every spatial branch is combined with every temporal
branch through a Hadamard product,
\begin{equation}
    H_L^{(i,j)}
    =
    H_{x,L}^{(i)}
    \odot
    H_{t,L}^{(j)},
    \qquad
    i=1,\ldots,M_x,
    \quad
    j=1,\ldots,M_t.
\end{equation}
The resulting $M_xM_t$ feature vectors are concatenated before the final
linear layer,
\begin{equation}
    u_{NN}(\mathbf{x},t;\theta)
    =
    W_{L+1}
    \left[
        H_L^{(1,1)},
        \ldots,
        H_L^{(M_x,M_t)}
    \right]
    +b_{L+1}.
\end{equation}

\subsection{Sinusoidal Representation Networks} \label{subsec:4:4:4:Siren}

In \cite{sitzmann2020implicit} (the webpage, presenting the results, the data and the code, is available at \cite{sitzmann2020implicitlink}.), a simple neural network architecture for implicit neural representations that uses the sine as a periodic activation function called \textsc{SiRen} was proposed. It is defined as
\begin{equation}
\Phi \left( \mathbf{x} \right) = \mathbf{W}_n \left( \phi_{n-1} \circ \phi_{n-2} \circ \ldots \circ \phi_0 \right) \left( \mathbf{x} \right) + \mathbf{b}_n, \quad \mathbf{x}_i \mapsto \phi_i \left( \mathbf{x}_i \right) = \sin \left( \mathbf{W}_i \mathbf{x}_i + \mathbf{b}_i \right).
\end{equation}
Here, $\phi_i: \mathbb{R}^{M_i} \mapsto \mathbb{R}^{N_i}$ is the $i^{th}$ layer of the network. It consists of the affine transform defined by the weight matrix $\mathbf{W}_i \in \mathbb{R}^{N_i \times M_i}$ and the biases $\mathbf{b}_i\in  \mathbb{R}^{N_i}$ applied on the input $\xx \in\mathbb{R}^{M_i}$, followed by the sine nonlinearity applied to each component of the resulting vector.

An interesting property of this architecture is that any derivative of a \textsc{SiRen} \emph{is itself a composition of} \textsc{SiRen}, as the derivative of the sine is a cosine, which can be interpreted as a phase-shifted sine. 
Therefore, the derivatives of a \textsc{SiRen} inherit the properties of \textsc{SiRen}s, enabling to supervise any derivative of \textsc{SiRen} with  ``complicated'' signals.

This network has similarities to the Fourier networks described in \cref{subsec:4:4:3:SpectralBias}, because using a Sin activation function has the same effect as having a learnable Fourier input encoding for the first layer of the network. The main difference is that in \textsc{SiRen} \textit{all} layers have a sinusoidal activation function. Something similar will be seen in \cref{4:4:5:Pinnsformer}, where we will introduce the \textsc{Wavelet} activation function. 

A critical component of \textsc{SiRen} is their initialization scheme. Weights are initialized from a uniform distribution scaled by the fan-in
\[
W \sim \mathcal{U} \left( - \sqrt{\frac{6}{\text{fan\_in}}} ,  + \sqrt{\frac{6}{\text{fan\_in}}} \right)\,,
\]
which ensures that the inputs to the sine activation follow a roughly Gaussian normal distribution. Crucially, the output of the sine function follows an arcsine distribution, which preserves the statistical properties of the activations across layers. This allows for the training of very deep networks without the vanishing or exploding gradient problems common with other activations.

The first layer of the network is scaled by a factor $\omega$ to span multiple periods of the $\sin$ function. This was empirically shown to give good performance and is in line with the benefits of the input encoding in the Fourier network. The authors suggest $\omega=30$ to perform well under many circumstances. 

\begin{figure}[t]
	\includegraphics[width=\textwidth]{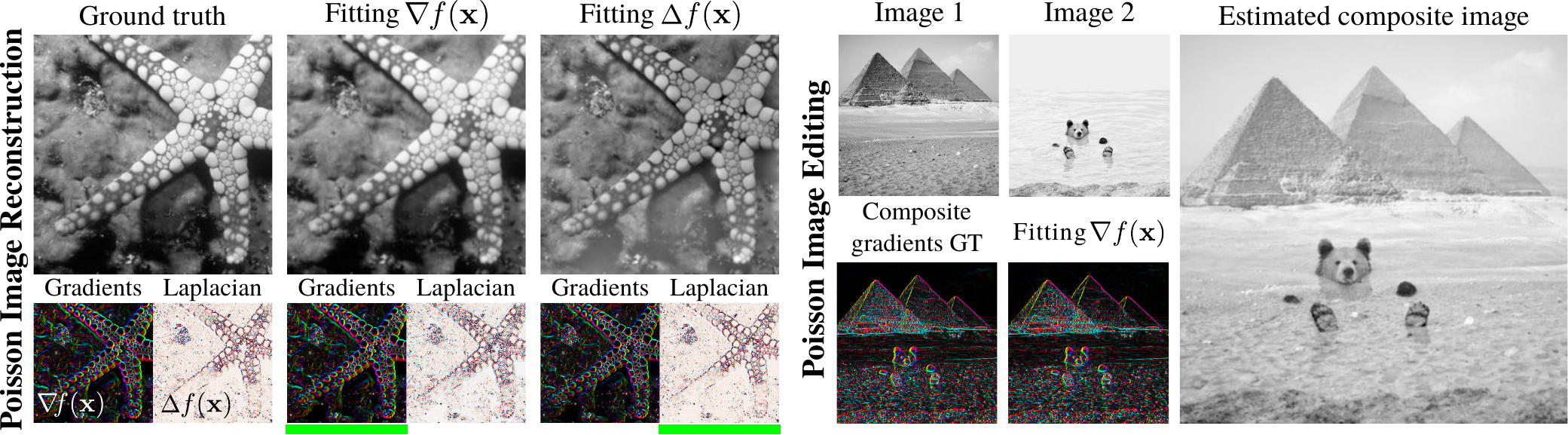}
	\caption{\textbf{Poisson image reconstruction:} An image (left) is reconstructed by fitting a \text{SiRen}, supervised either by its gradients or Laplacians (underlined in green). The results, shown in the centre and right, respectively, match both the image and its derivatives well. \textbf{Poisson image editing:} The gradients of two images (top) are fused (bottom left). \text{SiRen} allows for the composite (right) to be reconstructed using supervision on the gradients (bottom right).
    From \cite{sitzmann2020implicit}.
    }
	\label{fig:4:4:4:Siren_poisson_results}
\end{figure}

\begin{figure}[t]
	\includegraphics[width=\textwidth]{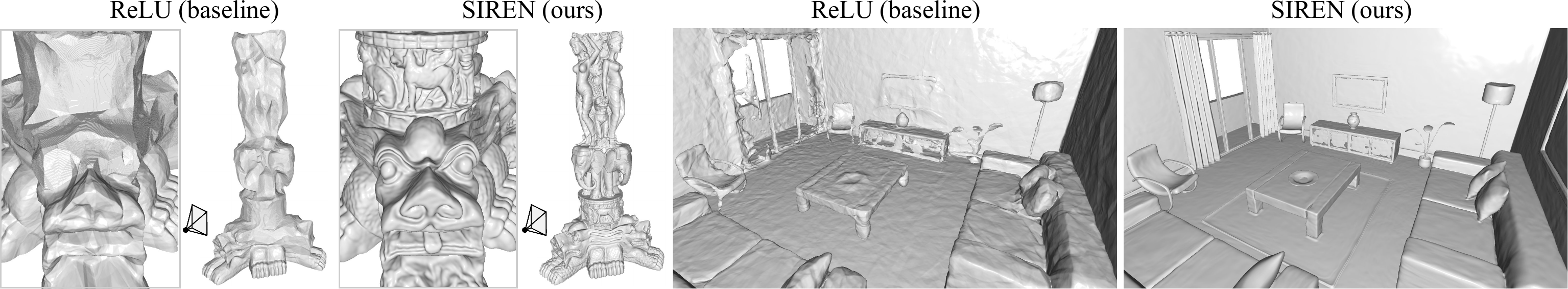}
	\caption{Shape representation. The authors of \cite{sitzmann2020implicit} fit signed distance functions parametrised by implicit neural representations directly on point clouds. Compared to ReLU implicit representations, the sinusoidal activations improve detail of objects (left) and complexity of entire scenes (right).
    From \cite{sitzmann2020implicit}.
    }
	\label{fig:4:4:4:sdf_results}
\end{figure}

A very interesting fact about \textsc{SiRen} is that, mainly, it was introduced in the context of Computer Vision, to obtain a network capable of learning representation of images (2D data), videos (2+1D data), and 3D representations (see \cref{fig:4:4:4:Siren_poisson_results} and \cref{fig:4:4:4:sdf_results}). 

The idea is that shape representation can be modelled by with differentiable \textit{signed distance functions} (SDFs) \cite{park2019deepsdf}; thus, it is possible to fit SDFs directly on oriented point clouds using implicit neural representations.
This amounts to solving a particular \textit{Eikonal boundary value problem }
\begin{align} 
    \Phi (\xx) &= 0\,, \qquad \xx \in \Omega_0\,, \notag\\
    |\nabla_\xx \Phi| &= 1 \,, \qquad \xx \in \Omega \,.
\end{align}

They fit the \textsc{SiRen} with a loss
\begin{align}
	\mathcal{L}_{\mathrm{sdf}} \! &= \! \int_{\Omega}
	\big\| \left|\nabla_\mathbf{x} \Phi(\mathbf{x}) \right| - 1 \big\|
	\dd^d x \notag \\
	& \quad+ \int_{\Omega_0}
	\!\!
		  \left[ \left\| \Phi ( \mathbf{x} ) \right\|
	 	+ \big(1 - \langle \nabla_\mathbf{x} \Phi(\mathbf{x}), \mathbf{n}(\mathbf{x}) \rangle  \big)
	\right] \dd \Sigma \notag \\
	& \qquad +	\int_{\Omega\setminus\Omega_0} \!\!\!\!
	\psi \big(\Phi(\mathbf{x})\big)
    \dd^d {x},
\end{align}
Here, $\psi(\mathbf{x})=\exp(-\alpha\cdot|\Phi(\mathbf{x})|),\alpha\gg 1$ penalizes off-surface points for creating SDF values close to 0. $\Omega$ is the whole domain and  $\Omega_0$ is the zero-level set of the SDF. The model $\Phi(x)$ is supervised using oriented points sampled on a mesh, where we require the \textsc{SiRen} to respect $\Phi(\mathbf{x})=0$ and its normals $\mathbf{n}(\mathbf{x})=\nabla f(\mathbf{x})$. During training, each mini-batch contains an equal number of points on and off the mesh, each one randomly sampled over $\Omega$.

As seen in Fig.~\ref{fig:4:4:4:sdf_results}, \textsc{SiRen} significantly increase the details of objects and the complexity of scenes that can be represented by these neural SDFs, parametrising a full room with only a single five-layer fully connected neural network.

\subsubsection{A simple example of emergence of Spectral Bias effects in PINN training: continuity equation}

\begin{figure}[t]
    \centering
    \includegraphics[width=0.95\linewidth]{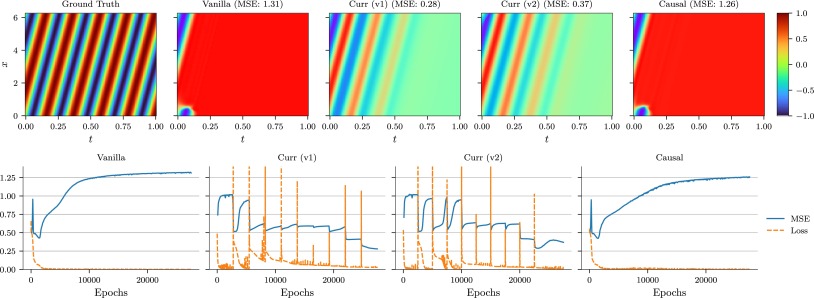}
    \caption{Worst case prediction for each method. The upper row shows the output of the trained network, the lower row reports the total loss and the mean squared error over the training epochs. Adapted from \cite{monaco2023training}.}
    \label{fig:4:3:4:1:continuity_instability}
\end{figure}

In \cite{monaco2023training}, authors noticed that a simple 1+1D dimensional problem may pose a significant difficulty if attacked with PINNs; the problem is the simple \textit{continuity equation} 
\begin{align}
    \partial_t u(t,x) + \beta \partial_x u(t,x) &= 0 \,, \qquad (t,x) \in [0, T]\times [-1,+1]\,, \\
    u(t=0, x) &= -\sin \pi x \,,\\
    u(t,x=+1) &= u(t, x=-1) .
\end{align}
Notice that, w.r.t. the conventions used in the paper, we adimensionalised the problem via the change of variable $x\to (x - \pi)/\pi$, and the redefinition $\beta \to \beta /\pi$. 
Authors in \cite{monaco2023training} have shown that, regardless of different training techniques, such as Fourier feature embeddings, curriculum learning (\cref{subsec:4:2:4:currlearn}), and causal loss (which we will see in \cref{subsubsec:4:4:2:4:CausalLoss}), the neural network shows difficulties in properly learning the solution for the problem \textit{in the high $\beta$ regime} ($\beta > 30$). In particular, the performance depends strongly on the number of collocation samples $N_f$: Vanilla and Causal training converge in some regimes, while the two Curriculum variants show comparatively high convergence rates across the tested training sizes.
In \cref{fig:4:3:4:1:continuity_instability} we present a visual example of this behaviour. 

The key point that is that this PDE is easily solvable \textit{analytically}, using the method of characteristics; simply selecting a solution in the form of $u(t,x)=f(x - \beta t)$, which trivially solves the PDE, gives us the solution simply requiring to satisfy both IC and BC;
$$
(\mathrm{IC}) \Rightarrow f(x) = - \sin \pi x  \Rightarrow u(t,x) = \sin \pi (\beta t  - x ) \,,
$$
which, indeed, on the BC
$$
 u(t, x=\pm 1) =  \sin  (\pi\beta t  \mp \pi) = - \sin \pi \beta t \Rightarrow u(t, x=+1) = u(t, x=-1) \quad \cvd
$$

Finally, notice the expression for the solution can be recast, using that $\sin(a+b) = \sin a \cos b + \cos a \sin b$, as
$$
u(t,x) = \sin (\pi \beta t ) \cos (\pi x) - \cos (\pi \beta t) \sin (\pi x) \,.
$$
This form immediately highlights \textit{why} high $\beta$ cases may be difficult for PINNs: high $\beta$ means \textit{high frequency modes} in the solution has to be regressed.   
This is an evident \textit{spectral bias} problem.

This is, however, only one possible contribution to the observed PINN failure. For this family of convection problems, \cite{krishnapriyan2021characterizing} provides evidence that increasing the PDE coefficient also makes the optimization problem induced by the physics residual progressively more difficult, producing a more complex loss landscape. Other effects, including differential-operator scaling, loss imbalance, sampling, and temporal propagation, may therefore interact with the network's spectral learning dynamics.


Let us assume now a different approach: we have the freedom to \textit{change the coordinates}, and we can pick the transform $t\mapsto \tau = \beta t$. In this case, the problem becomes
\[
\begin{split}
    \partial_\tau u(\tau,x) +  \partial_x u(\tau,x) &= 0 \,, \qquad (\tau,x) \in [0, \beta T]\times [-1,+1]\,, \\
    u(\tau=0, x) &= -\sin \pi x \,,\\
    u(\tau,x=+1) &= u(\tau, x=-1) ,   
\end{split}
\]
whose solution is
\[
    u(\tau, x) = \sin \pi (\tau   - x ).
\]

From here, it is immediately evident that the high frequency solution \textit{is gone}. So, no more spectral bias? No, of course. The point is that we traded \textit{high frequency solution on a small time domain} to a \textit{low frequency solution on a large time domain}!\note{This is due to the fact that $\tau\in[0, \beta T]$ while $t\in[0,T].$} 
So this means that the number of oscillations in the domain is \textit{the same} as before. 

This latter problem presents issues too, since we need to enforce \textit{causality} in the solution on a \textit{longer time scale}, which itself is a well known problem, even in standard numerical methods, since errors may accumulate over repeated time-integration steps. 

Furthermore, notice that the curriculum learning approach over $\beta$ (i.e., gradually increasing $\beta$ during training), proposed in \cite{monaco2023training}, can be seen as a \textit{moving time window} approach (which we will describe in more details in \cref{subsubsec:4:4:2:4:CausalLoss}), which justifies ex-post its effectiveness, at least for the initial part of the time domain (see \cref{fig:4:3:4:1:continuity_instability}).

\subsection{Transformers in PINNs} \label{4:4:5:Pinnsformer}

\begin{figure}[t]
    \centering
    \begin{subfigure}[t]{0.65\textwidth}
        \centering
        \includegraphics[width=\textwidth]{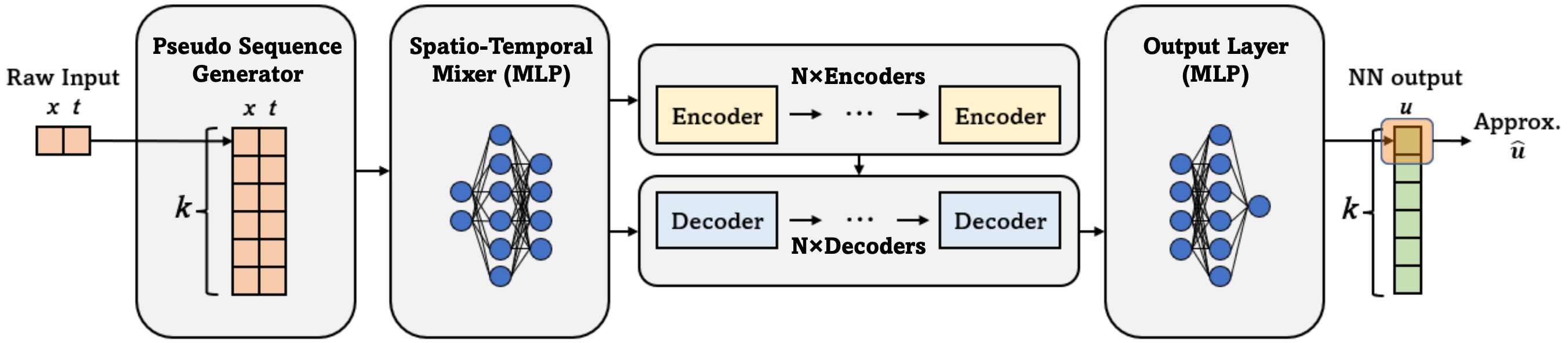}
        \caption{
        Architecture of the PINNsFormer. PINNsFormer generates a pseudo sequence based on pointwise input features. It outputs the corresponding sequential approximated solution. The first approximation of the sequence is the desired solution $\hat{u}_{NN}$.
        From \cite{zhao2023pinnsformer}.
        }
        \end{subfigure}
    \hfill
    \begin{subfigure}[t]{0.3\textwidth}
        \centering
        \includegraphics[width=\textwidth]{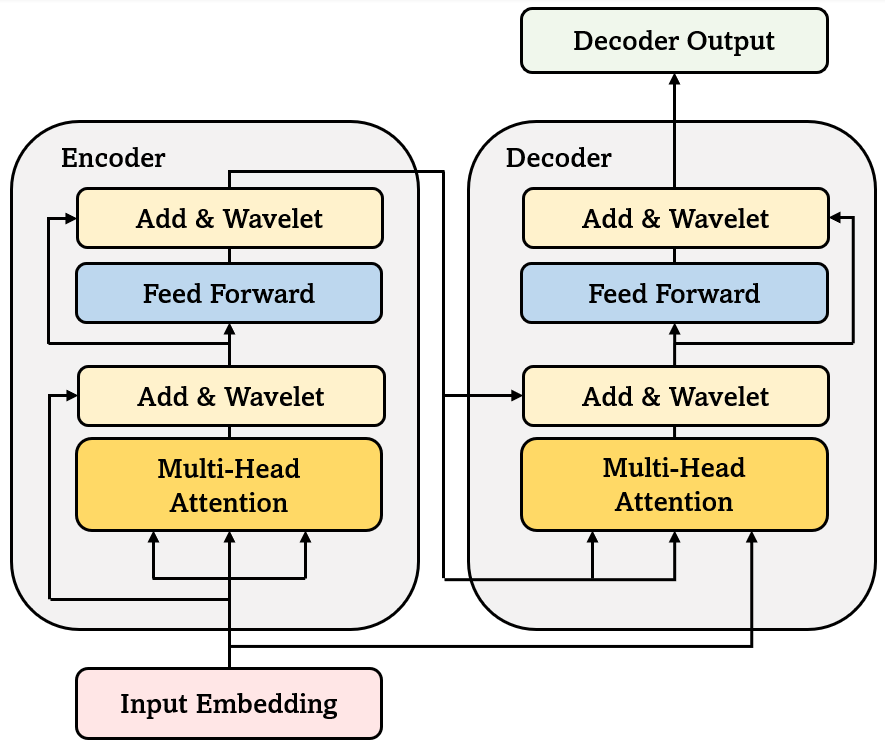}
        \caption{
        The architecture of PINNsFormer’s Encoder-Decoder Layers. The decoder is \textbf{not} equipped with self-attentions. From \cite{zhao2023pinnsformer}.
        }
    \end{subfigure}
    \label{fig:4:3:5:PINNsformer}
\end{figure}

In \cite{zhao2023pinnsformer}, authors introduce a methodology to use Transformers \cite{vaswani2017attention, scardapane2024alice} in the context of PINNs, in particular to attack time-evolution problems. It may seems a complex task, since transformers usually works on (discrete) sequences and/or structured patches (ViT for example \cite{dosovitskiy2020image}). Additionally, they are \textit{permutation equivariant}, and great care must be used to break this property via an appropriate, so-called \textit{Positional Embedding}. 

Thus, they introduce four main steps to allow for the usage of a Transformer architecture:
\begin{itemize}
    \item[1.] A novel activation function, Wavelet, to achieve Fourier-transforms with forward passing:
        \begin{equation}
            \texttt{Wavelet} (\xx) = \omega_1 \sin \xx + \omega_2 \cos \xx\,,
        \end{equation}
        where $\omega_{1,2}$ are learnable parameter. 
    \item[2.] a pseudo-sequence generator, to create \textit{phrases} as a discrete set of time-steps, i.e. a map $\RR^{d+1} \to \RR^{m\times (d+1)}$:
    \begin{equation}
        [\xx,t] \mapsto \{ [\xx, t], [\xx, t+\Delta t], \dots \, [\xx, t+m\Delta t] \} .
    \end{equation}
    \item[3.] A learnable embedding, the so-called\textit{spatio-temporal mixer} (see \cref{fig:4:3:5:PINNsformer}).
\end{itemize}

\paragraph{The \texttt{Wavelet} activation: } The intuition behind \texttt{Wavelet} activation simply follows Real Fourier Transform. It can be shown that \texttt{Wavelet} can approximate arbitrary functions giving sufficient approximation power, as presented in the following proposition:
\begin{proposition}
    Let $\mathcal{N}$ be a two-hidden-layer neural network with infinite width, equipped with \texttt{Wavelet} activation function, then $\mathcal{N}$ is a universal approximator for any real-valued target $f$.
\end{proposition}
\textit{Proof sketch:} The proof follows the Real Fourier Transform. For any given input $x$ and its corresponding real-valued target $f(x)$, we can represent it by its Fourier Integral:
\begin{gather*}
    f(x) = \int_{-\infty}^{\infty} F_c(\omega) \cos(\omega x) \, \dd\omega + \int_{-\infty}^{\infty} F_s(\omega) \sin(\omega x) \, \dd\omega
\end{gather*}
where $F_c$ and $F_s$ are the coefficients of Sines and Cosines respectively. Second, by Riemann sum approximation, the integral can be approximated by the infinite sum such that:
\begin{gather*}
    f(x) \approx \sum_{n=1}^{N} \left[ F_c(\omega_n) \cos(\omega_n x) + F_s(\omega_n) \sin(\omega_n x) \right] \equiv W_2(\texttt{Wavelet}(W_1 x))
\end{gather*}
where $W_1$ and $W_2$ are the weights of $\mathcal{N}$'s first and second hidden layer. As $W_1$ and $W_2$ are infinite-width, we can divide the piecewise summation into infinitely small intervals, making the approximation arbitrarily close to the true integral. Hence, $\mathcal{N}$ is a universal approximator for any given $f$. 

Notice that $\texttt{Wavelet}$ is used in the PINNsformer work just for its expressive power, and, as \textsc{SiRen} before, and Fourier Embedding even before, may have applications in other different contexts.

\paragraph{The Spatio-Temporal Mixer: } Most PDEs contain low-dimensional spatial or temporal information. Directly feeding low-dimensional data to encoders may fail to capture the complex relationships between each feature dimension. Hence, in an analogy with Fourier Embedding, the authors of \cite{zhao2023pinnsformer} suggest to embed original sequential data in higher-dimensional spaces such that more information is encoded into each vector, using an MLP on the \textit{phrases}.

Instead of embedding raw data in a high-dimensional space where the distance between vectors reflects the semantic similarity \cite{vaswani2017attention}, PINNsFormer constructs a linear projection that maps spatio-temporal inputs onto a higher-dimensional space using a fully-connected MLP. The embedded data enriches the capability of information by mixing all raw spatio-temporal features together, that they call the linear projection Spatio-Temporal Mixer.

\paragraph{Sequential physics-informed loss:}
A second important difference with respect to a conventional PINN is that the
physics-informed loss is evaluated over the generated pseudo-sequences.

For residual samples, PINNsFormer uses
\begin{equation}
    \mathcal L_{\mathrm{res}}
    =
    \frac{1}{kN_{\mathrm{res}}}
    \sum_{i=1}^{N_{\mathrm{res}}}
    \sum_{j=0}^{k-1}
    \left\|
        \mathcal D
        \left[
            \hat u(\mathbf{x}_i,t_i+j\Delta t)
        \right]
        -
        f(\mathbf{x}_i,t_i+j\Delta t)
    \right\|_2^2 .
\end{equation}
Analogously,
\begin{equation}
    \mathcal L_{\mathrm{bc}}
    =
    \frac{1}{kN_{\mathrm{bc}}}
    \sum_{i=1}^{N_{\mathrm{bc}}}
    \sum_{j=0}^{k-1}
    \left\|
        \mathcal B
        \left[
            \hat u(\mathbf{x}_i,t_i+j\Delta t)
        \right]
        -
        g(\mathbf{x}_i,t_i+j\Delta t)
    \right\|_2^2 .
\end{equation}

The initial condition is different: only the first element of a sequence
generated from an initial point remains at $t=0$. Hence
\begin{equation}
    \mathcal L_{\mathrm{ic}}
    =
    \frac{1}{N_{\mathrm{ic}}}
    \sum_{i=1}^{N_{\mathrm{ic}}}
    \left\|
        \mathcal I[\hat u(\mathbf{x}_i,0)]
        -
        h(\mathbf{x}_i,0)
    \right\|_2^2 .
\end{equation}

The complete objective is then
\begin{equation}
    \mathcal L_{\mathrm{PINNsFormer}}
    =
    \lambda_{\mathrm{res}}\mathcal L_{\mathrm{res}}
    +
    \lambda_{\mathrm{bc}}\mathcal L_{\mathrm{bc}}
    +
    \lambda_{\mathrm{ic}}\mathcal L_{\mathrm{ic}}.
\end{equation}

\subsection{Mixture-of-Experts}

\begin{figure}[t]
    \centering
    \includegraphics[width=\textwidth]{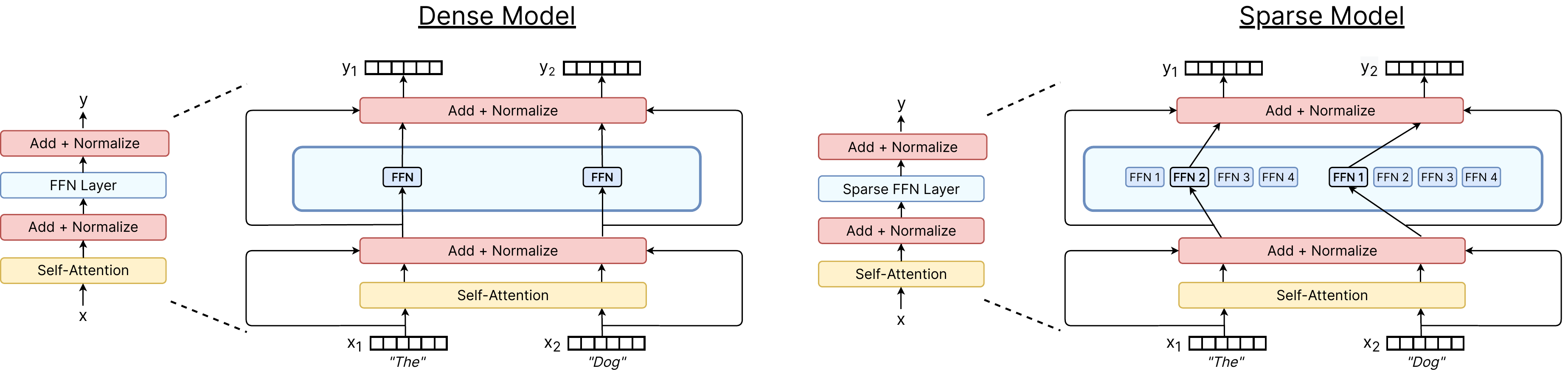}
    \caption{
    \textbf{Comparing a dense and sparse expert Transformer.} A dense model \textbf{(left)} sends both input tokens to the same feed-forward network parameters (FFN). A sparse expert model \textbf{(right)} routes each input token independently among its four experts (FFN1 $\cdots$ FFN4). In this diagram, each model uses a similar amount of computation, but the sparse model has more unique parameters.
    Note while this figure showcases a specific and common approach of sparse feed-forward network layers in a Transformer \cite{vaswani2017attention}, the technique is more general. From \cite{fedus2022reviewsparseexpertmodels}
    }.
    \label{fig:4:3:6:MoE_sparsity_diagram}
\end{figure}

One powerful concept emerged already at the beginning of the '90s (see, e.g., \cite{jacobs1991adaptive}) was the possibility of gather a set of \textit{experts} model to perform a certain task, and then aggregate their prediction in a gated manner, for example by training a specialised network to learn how to do that. This approach is increasingly becoming relevant, especially in the context of LLMs (see, e.g., \cite{fedus2022reviewsparseexpertmodels, zhang2025mixture} and references therein), where it is possible to reduce the computational costs and the latency by using, instead of an extremely large model, a set of \textit{expert} smaller models, of whose only a few are activated to perform a reply, token-by-token, without impacting the performances, as shown, for example, in \cite{jiang2024mixtral}. In particular, in \cite{jiang2024mixtral}, they used eight 7B (pre-trained, and then fine-tuned) transformer model, where the fully-connected layers are replaced by the sparse Mixture-of-Experts, and defined a \textit{router} model to select the $\text{TopK}$ models (see \cref{fig:4:3:6:MoE_sparsity_diagram}).

\begin{figure}[t]
    \centering
    \includegraphics[width=0.75\linewidth]{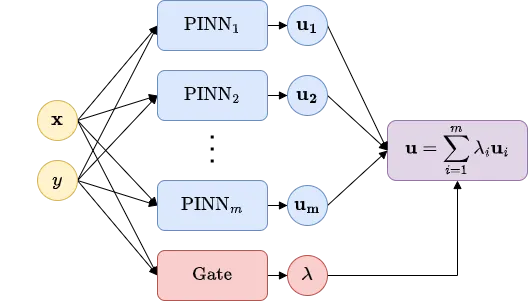}
    \caption{Mixture of Experts of PINNs. An arbitrary number $m$ of PINNs, possibly with varying architectures and properties, is initialised together with a gating network. All models receive the same input and the gate produces weights that are used for aggregating the results. From \cite{BischofMixtureofExpertsEnsembleMF}.}
    \label{fig:4:3:6:MoE_pinn1}
\end{figure}

The idea was extended in the PINN domain in \cite{BischofMixtureofExpertsEnsembleMF}, with a few simplification to streamline its interpretability. In particular, MoE-PINN approach uses multiple networks (i.e., the experts) to infer the output multiple time, and then uses an additional network (the gate), which will weight the experts output. The $m$ model may all have a different hyperparameter set (number of layers, activations, number of nodes, different embeddings, etc). 
Furthermore, by leveraging multiple networks and a gate model to divide the domain,  each expert can specialise in solving a distinct part of the problem, allowing for improved accuracy and reduced bias-variance trade-off. Dividing the problem into smaller sub-problems has many advantages:
\begin{itemize}
    \item By using several learners on distinct subdomains, the complexity of the problem is reduced.
    \item The gate in MoE-PINNs is a continuous function, resulting in smooth transitions between the domains. Therefore, more complex regions of the domain can be equally divided amongst several learners, whereas simpler regions can be attributed to a single expert.
    \item  The gate can be any type of neural network, from a linear layer to a deep neural network, which allows it to adapt to different types of domains and divide them in an arbitrary way.
    \item MoE can be easily parallelised, since only the weights lambda need to be shared amongst learners on different devices. In theory, each learner could be placed on a distinct GPU.
    \item By initialising a large number of PINNs with different architectures, the need for labor-intensive hyperparameter tuning is reduced.
\end{itemize}

The set up is the following: we have $m$ networks $u_i(\theta_i) : (t,\xx) \mapsto u_i(\theta_i; t, \xx)$, each of wich, alone, can approximate the solution. Furthermore, we have a gating network $g(\theta_g) : (t,\xx) \mapsto g_i(t, \xx)$, with $i=1,\dots,m$. Then, using a softmax, we associate to the gating network prediction a probability
\[
    \lambda_i (t,\xx) =  \frac{\exp [g_i (t,\xx)]}{\sum_{j=1}^m \exp [g_j (t,\xx)]} \, ,
\]
which is the \textit{importance} of  the expert $i$. Then, the general approximation of the MoE-PINN is
\[
    u_{MoE}(\theta; t, \xx) = \sum_{i=1}^m \lambda_i (t,\xx)  u_{i}(\theta_i; t, \xx) \,.
\]

Furthermore, MoE-PINNs have a nice \textit{interpretable} feature, if sparsified, i.e. if there is added a sparsifying term to the loss to incentivise the router model $\lambda : (\xx, t) \mapsto(\lambda_1(\xx, t) , \dots, \lambda_m(\xx, t))$ the focus of each expert in a region, 
\[
    \LL_{sp} = \sum_{i=1}^m \left|  \frac{1}{|B|} \sum_{(x,t)\in B} \lambda_i(\xx, t)\right |^p \,, \quad p\in (0,1).
\]

\begin{figure}
    \centering
    \includegraphics[width=0.95\linewidth]{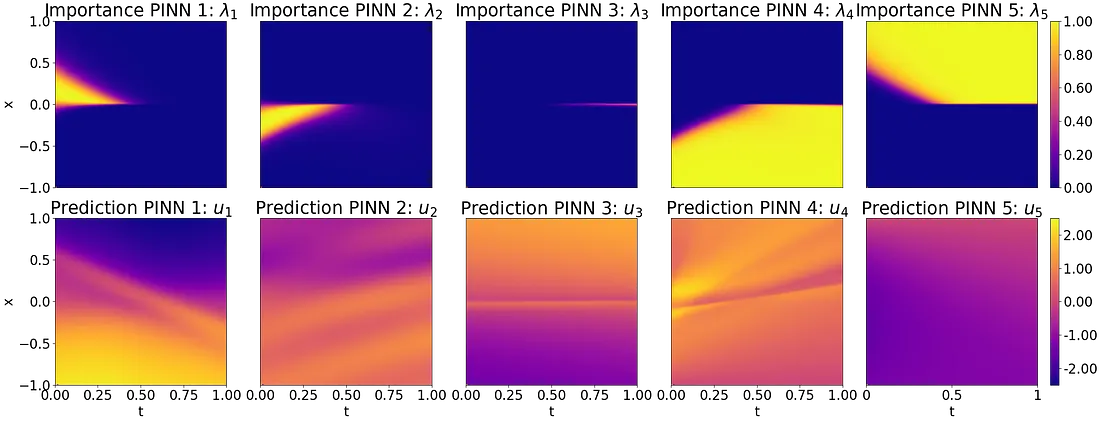}
    \caption{Weights lambda produced by the gating network (top row) for each of the expert (columns) as well as the predictions from each expert (bottom row) for Burgers’ equation. From \cite{BischofMixtureofExpertsEnsembleMF},
    }
    \label{fig:4:3:6:MoE_burgers}
\end{figure}

In \cref{fig:4:3:6:MoE_burgers} is reported an example of MoE-PINNs application to solve the Burgers equation:
\begin{align*}
    \partial_t u(t,x) + u(t,x) \partial_x u(t,x)  &= \nu \partial_x^2 u(t,x)  \,, \quad (t,x) \in [0,T]\times[-1,+1], \\
    u(0, x) &= - \sin (\pi x) \,, \\
    u(t, -1) = u(t, +1) &= 0 \,. 
\end{align*}

\noindent They designed the experiment with $m=5$ experts network with
\begin{itemize}
    \item Expert 1: 2 layers with 64 nodes each and tanh activation;
    \item Expert 2: 2 layers with 64 nodes each and sine activation;
    \item Expert 3: 2 layers with 128 nodes each and tanh activation;
    \item Expert 4: 3 layers with 128 nodes each and tanh activation;
    \item Expert 5: 2 layers with 256 nodes each and swish activation;
\end{itemize}

Notice how the gating network in MoE-PINNs was able to effectively allocate tasks to each expert based on their respective capacity for modelling different parts of the domain. The experts with fewer layers and nodes were assigned to the smoother regions, which are relatively easier to model, i.e. close to the initial conditions. Meanwhile, the more complex experts, i.e. those with deeper and wider architectures, were utilised in the regions with discontinuities, where a more complex model was necessary for accurate representation. This is particularly evident in the case of expert 3, which was solely dedicated to capturing the discontinuity.

\section{Optimisation Schemes} \label{sec:4:6:Optim_schemes}

\subsection{Optimisers: second order over first order}

Let us now stop for a moment to look at our optimisation problem. As standard in machine and deep learning, what we are actually dealing with is some form of an optimisation problem, where we need to solve 
\[
    \min_{\theta \in \RR^n} \LL(\theta) \,,
\]
or, just to simplify the notation, 
\[
    \min_{\xx\in \RR^n} f(\xx) .
\]
The most famous families of algorithms to solve this problem are first order methods, such as the Gradient Descent and its derivatives algorithms (SGD, RMSprop, Adam, etc). Symbolically, an iteration of
\[
    \xx_{k+1} = \xx_k - \eta \nabla_\xx f(\xx_k) \,,
\]
where $\eta$ is a positive real number called \textit{learning rate}, which controls the step size taken at each iteration. This is clearly a \textit{first-order} optimisation scheme, such it utilise only the first-derivative (the gradient) of our loss. Furthermore remains the issue of picking $\eta$; if it is too large, the model diverges out of control; if it is too small, the model becomes highly inefficient, taking unnecessarily long to converge to the minimum. There is no a priori way of choosing a good learning rate, and in practice the optimal learning rate is typically found empirically, by looping through various values and seeing how they perform. 

\begin{figure}[t]
    \centering
    \includegraphics[width=0.95\linewidth]{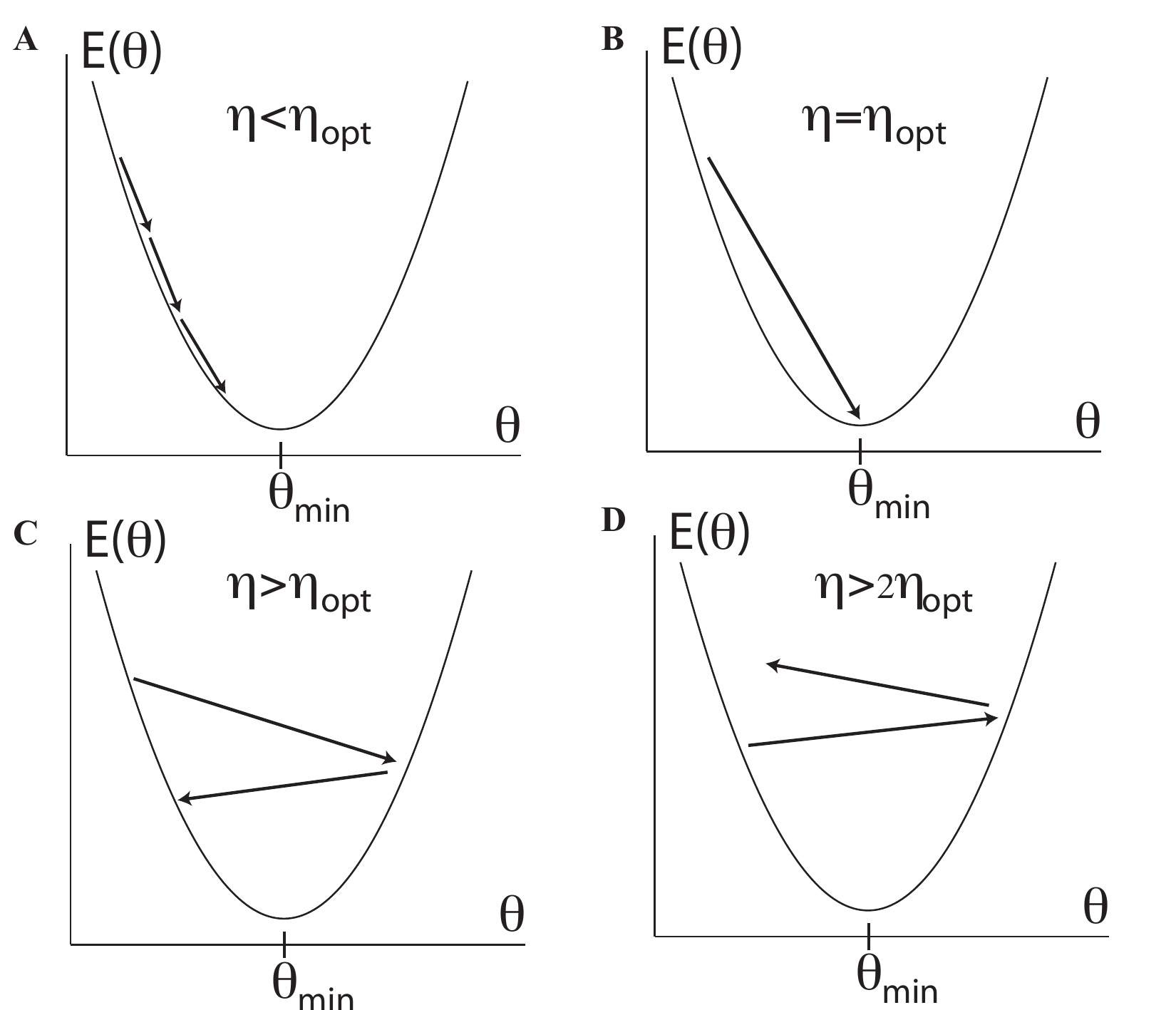}
    \caption{    \textbf{Effect of learning rate on convergence}. For a one dimensional quadratic potential, one can show that there exists four different qualitative behaviours for gradient descent (GD) as a function of the learning rate $\eta$ depending on the relationship between $\eta$ and $\eta_{\mathrm{opt}}= [\partial_\theta^2 E(\theta)]^{-1}$. (a) For $\eta <\eta_{\mathrm{opt}}$, GD converges to the minimum. (b) For $\eta =\eta_{\mathrm{opt}}$, GD converges in a single step. (c) For $\eta_{\mathrm{opt}} <\eta < 2\eta_{\mathrm{opt}}$, GD oscillates around the minima and eventually converges. (d) For $\eta  > 2\eta_{\mathrm{opt}}$, GD moves away from the minima.  From \cite{mehta2019high}, adapted from \cite{lecun2002efficient}.}
    \label{fig:4:5:GD3regimes}
\end{figure}

\paragraph{A physicist's point-of-view: } 

Before going on, let us look again at our optimisation problem, and put up our physicist glasses (adapted from \cite{mehta2019high}). We need to minimise something, which is estimated via some sort of Monte Carlo integration scheme. Usually, in physical system, we minimise the \textit{energy}
\[
    E(\boldsymbol{\theta}) = \sum_{i=1}^n e_i(\theta; \mathbf{x}_i),
\]
and this is somehow the total energy of a system of $n$ (non-interacting) particles of energy $e_i$; additionally, if we use the mean squared error (MSE) as a cost function, we have
\[
e_i(\theta; \mathbf{x}_i)  = (\yy_i - f_\theta (\xx_i))^2 \,,  
\]
which is the elastic potential energy of a spring (of elastic constant $k=1$) extending from $f_\theta(\xx_i)$ to $\yy_i$. We can thus see the plain gradient descent algorithm
\begin{align*}
    \mathbf{v}_t&= \eta_t \nabla_\theta E({\theta}_t), \nonumber \\
    \boldsymbol{\theta}_{t+1}&= \boldsymbol{\theta}_t -\mathbf{v}_t \,,
\end{align*}
where we used the bold symbol to emphasize the vectorial nature of $\boldsymbol{\theta}$.  It is clear that for sufficiently small choice of the learning rate $\eta_t$  this methods will converge to a \emph{local minimum} (in all directions) of the cost function. However, choosing a small $\eta_t$ comes at a huge computational cost. The smaller $\eta_t$, the more steps we have to take to reach the local minimum. In contrast, if $\eta_t$ is too large, we can overshoot the minimum and the algorithm becomes unstable; it either oscillates or even moves away from the minimum (see \cref{fig:4:5:GD3regimes}).

To better understand this behaviour, and highlight some of the shortcomings of plain gradient descent, it is useful to contrast first-order Gradient descent with a second-order method, the \textit{Newton's Method}. In Newton's method, we choose the step $\mathbf{v}$ for the parameters in such a way as to minimize a second-order Taylor expansion to the energy function
\[
E(\boldsymbol{\theta}+\mathbf{v}) \approx E(\boldsymbol{\theta} )+ \nabla_\theta E(\boldsymbol{\theta})\mathbf{v} + \frac{1}{2}\, \mathbf{v}^T H(\mathbf{\theta})\mathbf{v},
\]
where $H(\boldsymbol{\theta})$ is the Hessian matrix of second derivatives.  Differentiating this equation respect to $\mathbf{v}$ and noting that for the optimal value $\mathbf{v}_\mathrm{opt}$ we expect the minimum, i.e. a stationary point for its derivate $\nabla_\theta E(\boldsymbol{\theta} +\mathbf{v}_{\mathrm{opt}})=0$, yields  
\[
\nabla_\theta E(\boldsymbol{\theta})+ H(\mathbf{\theta})\mathbf{v}_\mathrm{opt} = 0,
\]
which gives this algorithm
\begin{align*}
    \mathbf{v}_t &= H^{-1}(\mathbf{\theta}_t)\nabla_\theta E(\boldsymbol{\theta}_t) \\
    \boldsymbol{\theta}_{t+1} &= \boldsymbol{\theta}_t -\mathbf{v}_t.
\end{align*}
Since there is no guarantee that the Hessian is well conditioned, in almost all applications of Newton's method, the inverse of the Hessian $ H^{-1}(\mathbf{\theta}_t)$ is usually replaced by some suitably regularised pseudo-inverse such as $[H(\mathbf{\theta}_t)+\epsilon I]^{-1}$ with $\epsilon$ a small parameter \cite{battiti1992first}.

We can now use a simple case with Newton's method to develop intuition about the role of the learning rate in first-order methods. Let us first consider the special case of using GD to find the minimum of a quadratic energy function of a single parameter $\theta$ \cite{lecun2002efficient}. Given the current value of our parameter $\theta$, we can ask what is the optimal choice of the learning rate $\eta_{\mathrm{opt}}$, where $\eta_{\mathrm{opt}}$ is defined as the value of $\eta$ that allows us to reach the minimum of the quadratic energy function in a single step. To find $\eta_{\mathrm{opt}}$, we expand the energy function to second order around the current value
\[
E(\theta +v) = E(\theta_c )+ \partial_\theta E(\theta)v+ \frac{1}{2}\, {\partial_\theta^2 E(\theta)}v^2.
\]
Now, differentiating with respect to $v$, and setting $\theta_{\mathrm{min}}=\theta-v$ we get
\[
\theta_{\mathrm{min}}=\theta- [\partial_\theta^2 E(\theta)]^{-1} \partial_\theta E(\theta),
\]
so that, by looking at these expression, we see that the optimal learning rate is
\[
\eta_{\mathrm{opt}} = [\partial_\theta^2 E(\theta)]^{-1}.
\]

\subsubsection{BFGS and its limited memory version}

We have seen that Newton's method does not require the learning rate hyperparameter. It seems too good to be true!
It turns out that the benefits from Newton’s method comes at a cost. There are two main issues with Newton’s method \cite{nocedal2006numerical, Johnson2019bfgs} (for a nice blog post, see \cite{Lam2020BFGS}):
\begin{itemize}
    \item Newton's method is sensitive to initial conditions. This is especially apparent if our objective function is non-convex. Unlike gradient descent, which ensures that we’re always going downhill by always going in the direction opposite the gradient, Newton’s method fits a paraboloid to the local curvature and proceeds to move to the stationary point of that paraboloid. Depending on the local behaviour of our initial point, Newton’s method could take us to a maximum or a saddle point instead of a minimum. Thus Newton’s method is really only appropriate for minimizing convex objective functions.
    \item Newton’s method is very computationally expensive. While the computation of the gradient scales as $O(n)$, the computation of the inverse Hessian scales as $O(n^3)$\note{computing the Hessian scales as $O(n^2)$, inverting it scales as $O(n^3)$.}.  Almost immediately, the benefit of an increased convergence rate will be far outweighed by the large cost of the additional computation time.
\end{itemize}

Quasi-Newtonian methods aim to handle these issues, still inserting Hessian hints of the landscape while keeping the computational cost low. The most famous method is the Broyden-Fletcher-Goldfarb-Shanno (BFGS) algorithm
\cite{broyden1970convergence, fletcher1970new, goldfarb1970family, shanno1970conditioning}, especially in its Limited-memory/Low storage version (L-BFGS)\note{The PyTorch version of L-BFGS is inspired to the Matlab version presented in \cite{Schmidt2005}.}.

\begin{algorithm}[t]
\caption{BFGS quasi-Newton method}
\label{alg:4:5:1:1:BFGS}
\begin{algorithmic}[1]
\Require initial point \(x_0\), tolerance \(\varepsilon>0\), maximum iterations \(K_{\max}\), initial inverse Hessian approximation \(B_0\) (symmetric positive definite, e.g. \(B_0=I\))
\State \(k \leftarrow 0\)
\While{\(\|\nabla f(x_k)\| > \varepsilon\) and \(k < K_{\max}\)}
    \State \(\;p_k \leftarrow - B_k \nabla f(x_k)\) \Comment{ Compute search direction}
    \State \(\alpha_k \leftarrow \text{StrongWolfeLineSearch}(f, x_k, p_k)\) \Comment{Pick $\alpha_k$}
    \State \(x_{k+1} \leftarrow x_k + \alpha_k p_k\) \Comment{Compute update}
    \State \(s_k \leftarrow x_{k+1} - x_k\)
    \State \(y_k \leftarrow \nabla f(x_{k+1}) - \nabla f(x_k)\)
    \If{\(y_k^{T} s_k > 0\)} 
        \State \(\rho_k \leftarrow \dfrac{1}{y_k^{T} s_k}\)
        \State \(B_{k+1} \leftarrow \left(  I - \rho_k s_k y_k^{T}\right) B_k \left(  I - \rho_k s_k y_k^{T}\right)^{T} + \rho_k s_k s_k^{T}\) \Comment{Update $B$}
    \Else
        \State Option: reset \(B_{k+1} \leftarrow B_0\) or skip update
    \EndIf
    \State \(k \leftarrow k+1\)
\EndWhile
\State return \(x_k\)
\end{algorithmic}
\end{algorithm}

The key idea of quasi-Newton methods is that, instead of computing the actual Hessian, we just approximate it with a positive-definite matrix $B$, which is updated from iteration to iteration using information computed from previous steps, thus effectively reducing the computational cost of inverting $H$ at each iteration. 

To describe the BFGS algorithm (following \cite{nocedal2006numerical}), we start by identifying the objective function at the iteration $k$
\[
m_k(p) = f_k + \nabla f_k ^T p + \half p^T H_k p \,, 
\]
whose minimiser is
\[
    p_k = - H_k^{-1} \nabla_k f \,,
\]
which is used as the search direction, and the new iterate is
\[
    x_{k+1} = x_k + \alpha_k p_k, 
\]
where $\alpha_k$ can be found performing a one-dimensional optimization (line search) to find an acceptable step-size $\alpha_k$ such that
\[
    \alpha_k = \arg \min f(x_k + \alpha_k p_k) .
\]
Practically, an inexact line search usually suffices, with $\alpha_k$ is chosen to satisfy the \note{The (not strong) Wolfe Condition is the same, but without the absolute values in the second condition.}\textit{Strong Wolfe Condition}
\begin{align}
    f(x_k + \alpha_k p_k) &\le f(x_k) + c_1 \alpha_k \nabla f_k^T p_k \,, \notag \\
    |\nabla f(x_k + \alpha_k p_k)^T p_k| &\le c_2 |\nabla f_k^T p_k| \,,  \notag \\
    \text{with } 0 < c_1 &< c_2 < 1.
\end{align}
The following \cref{lst:4:5:1:1:StrongWolfe} reports a sketch of a plain python implementation of the Strong Wolfe line search.
\begin{lstlisting}[caption={Line Search with Strong Wolfe Condition}, label={lst:4:5:1:1:StrongWolfe}, language=Python]
def strong_wolf_line_search(f,x,p,nabla):
    a  = 1.0
    c1 = 1e-4 # standard choice
    c2 = 0.9  # standard choices are in [0.1, 0.9] 
    fx = f(x)
    x_new = x + a * p 
    nabla_new = grad(f,x_new)
    while ( (f(x_new) >= fx + (c1*a*nabla.T @ p) ) 
            or (abs(nabla_new.T @ p) >= c2*abs(nabla.T @ p) ) ) : 
        a *= 0.5
        x_new = x + a * p 
        nabla_new = grad(f,x_new)
    return a
\end{lstlisting}

\paragraph{L-BFGS}

We notice that applying the BFGS update directly requires $O(n^2)$ storage for $H_k^{-1}$ and $O(n^2)$ work on each step to update $H_{k+1}$. This is fine for up to a few thousand parameters, but for truly large-scale optimization problems it is prohibitive. 
%

Fortunately, there is a solution: L-BFGS avoids storing $H_k$ explicitly. Instead, it retains only the most recent $m$ curvature pairs
\[
    (s_i,y_i),
    \qquad
    s_i=x_{i+1}-x_i,
    \qquad
    y_i=g_{i+1}-g_i,
\]
together with
\[
    \rho_i=\frac{1}{y_i^Ts_i}.
\]
Explicitly, one does not explicitly represents $B_k$, but computes its contraction $B_k g_k,\; g_k\defeq \nabla f(x_k)$ needed for the algorithm, and defines a two-step algorithm \cref{alg:4:5:1:1:L-BFGS2S}, which replaces the inverse hessian approximation $B_k$ update  \cref{alg:4:5:1:1:L-BFGS}.
Consequently, storage and the cost of the recursion scale as $O(mn)$ rather than $O(n^2)$, with $m\ll n$   \cite{liu1989limited}.

\begin{algorithm}
    \caption{L-BFGS two-step algorithm}
    \label{alg:4:5:1:1:L-BFGS2S}
    \begin{algorithmic}[1]
    \Require stored pairs $\{(s_i,y_i)\}_{i=k-m+1}^k$, scalars $\rho_i=1/(y_i^T s_i)$, initial $B_0$ (often $\gamma I$), vector $q=g$
    \For{$i=k$ \textbf{downto} $k-m+1$}
      \State $\alpha_i \leftarrow \rho_i\, s_i^{T} q$
      \State $q \leftarrow q - \alpha_i\, y_i$
    \EndFor
    \State $r \leftarrow B_0\, q$ \Comment{apply initial matrix}
    \For{$i=k-m+1$ \textbf{to} $k$}
      \State $\beta_i \leftarrow \rho_i\, y_i^{T} r$
      \State $r \leftarrow r + s_i\,(\alpha_i - \beta_i)$
    \EndFor
    \State \Return $r$ \Comment{this is $B_k g$}
    \end{algorithmic}
\end{algorithm}

\begin{algorithm}
\caption{Limited‑memory BFGS (L‑BFGS)}
\label{alg:4:5:1:1:L-BFGS}
\begin{algorithmic}[1]
\Require initial point \(x_0\), tolerance \(\varepsilon>0\), maximum iterations \(K_{\max}\),
\Statex \hspace{\algorithmicindent} memory parameter \(m\) (number of stored pairs), initial scalar \(\gamma_0>0\) (for \(B_0=\gamma_0 I\))
\State \(k \leftarrow 0\)
\State initialise empty lists \(\mathcal{S}\leftarrow[],\ \mathcal{Y}\leftarrow[],\ \mathcal{R}\leftarrow[]\) \Comment{store recent \(s_i,y_i,\rho_i\)}
\While{\(\|\nabla f(x_k)\| > \varepsilon\) and \(k < K_{\max}\)}
    \State \(g_k \leftarrow \nabla f(x_k)\)
    \State \textcolor{ferrarired}{$r \leftarrow \text{LBFGS2Steps}(g_k, \mathcal{S}, \mathcal{Y}, \mathcal{R}, B_0)$} \Comment{Two step algorithm}
    \State $p_k \leftarrow -r$ \Comment{search direction}
    \State \(\alpha_k \leftarrow \text{StrongWolfeLineSearch}(f, x_k, p_k)\)
    \State \(x_{k+1} \leftarrow x_k + \alpha_k p_k\)
    \State \(s_k \leftarrow x_{k+1} - x_k\)
    \State \(y_k \leftarrow \nabla f(x_{k+1}) - g_k\)
    \If{\(y_k^{T}s_k > 0\)}
        \State \(\rho_k \leftarrow 1/(y_k^{T}s_k)\)
        \State Append \(\mathcal{S}.\text{append}(s_k),\ \mathcal{Y}.\text{append}(y_k),\ \mathcal{R}.\text{append}(\rho_k)\)
        \If{\(|\mathcal{S}| > m\)} \Comment{drop oldest pair to keep memory \(m\)}
            \State  \(\mathcal{S}.\text{pop}(0),\ \mathcal{Y}.\text{pop}(0),\ \mathcal{R}.\text{pop}(0)\) \Comment{remove oldest elements}
        \EndIf
        \State Optionally update \(\gamma_0 \leftarrow \dfrac{s_k^{T}y_k}{y_k^{T}y_k}\) \Comment{scaling for next iteration}
    \Else
        \State Option: skip storing the pair or reset memory
    \EndIf
    \State \(k \leftarrow k+1\)
\EndWhile
\State \Return \(x_k\)
\end{algorithmic}
\end{algorithm}

\paragraph{Relevance for PINN training: }

Empirically, it was seen that, in some applications, quasi-Newton Methods, mostly the limited memory version of the BFGS algorithm (L-BFGS) \cite{liu1989limited, byrd1995limited}, can be used to achieve better performance with fewer iterations with respect to first order methods, though they are more prone to getting trapped at saddle point \cite{urban2025unveiling}. To mitigate this, some studies recommend using a first-order optimisation algorithm, mostly Adam, during the initial stages of training, followed by L-BFGS for fine-tuning \cite{raissi2017physics, tartakovsky2018learning, urban2025unveiling}.

\subsubsection{Self-Scaled Broyden (SSBroyden)}

In \cite{urban2025unveiling} investigated the use of self-scaled quasi-Newton algorithm, which has an huge impact in optimising PINNs \cite{toscano2025pinns, khodakarami2026spectral}. 

Self‑scaled Broyden (SSBroyden) \cite{al1998numerical, al2005wide, al2014broyden} is a quasi‑Newton update that generalises the classical Broyden/BFGS family by introducing \emph{self‑scaling} parameters that adapt the inverse‑Hessian approximation to improve robustness and global convergence in large‑scale, ill‑conditioned problems. The method maintains an approximation \(H_k\) of the inverse Hessian and updates it using the secant information
\begin{align*}
    \mathbf{s}_k \;&=\; \Theta_{k+1}-\Theta_k,\\
    \mathbf{y}_k \;&=\; \nabla\mathcal{J}(\Theta_{k+1})-\nabla\mathcal{J}(\Theta_k).
\end{align*}
Introducing the auxiliary vector
\[
\mathbf{v}_k \;=\; \sqrt{\mathbf{y}_k^\top H_k \mathbf{y}_k}\,\left(\frac{\mathbf{s}_k}{\mathbf{y}_k^\top\mathbf{s}_k}
- \frac{H_k\mathbf{y}_k}{\mathbf{y}_k^\top H_k\mathbf{y}_k}\right), 
\]
the SSBroyden inverse‑Hessian update reads
\[
\begin{aligned}
H_{k+1}
&= \frac{1}{\tau_k}\Bigg[\,H_k
- \frac{H_k\mathbf{y}_k\,\mathbf{y}_k^\top H_k}{\mathbf{y}_k^\top H_k\mathbf{y}_k}
+ \phi_k\,\mathbf{v}_k\mathbf{v}_k^\top \Bigg]
+ \frac{\mathbf{s}_k\mathbf{s}_k^\top}{\mathbf{y}_k^\top\mathbf{s}_k},
\end{aligned}
\]
where \(\tau_k>0\) and \(\phi_k\) are scalar scaling/update parameters chosen to enforce stability and (when possible) superlinear convergence. For \(\tau_k=\phi_k=1\) the formula reduces to the standard BFGS update.

\paragraph{Practical parameter choices (SSBroyden variant):}
A robust choice used in practice sets \(\phi_k\) and \(\tau_k\) from local curvature diagnostics. Defining
\begin{align*}
    b_k &= \frac{\mathbf{s}_k^\top H_k^{-1}\mathbf{s}_k}{\mathbf{y}_k^\top\mathbf{s}_k}, \\
    h_k &= \frac{\mathbf{y}_k^\top H_k \mathbf{y}_k}{\mathbf{y}_k^\top\mathbf{s}_k}, \\
    a_k &= h_k b_k - 1,\\
    c_k &= \sqrt{\frac{a_k}{a_k+1}} \\
    \rho_k^- &= \min\big(1,\, h_k(1-c_k)\big),\\
    \theta_k^- &= \frac{\rho_k^- - 1}{a_k},\\
    \theta_k^+ &= \frac{1}{\rho_k^-}, \\
    \theta_k &= \max\!\Big(\theta_k^-,\; \min\!\big(\theta_k^+,\; \tfrac{1-b_k}{b_k}\big)\Big),\\
    \sigma_k &= 1 + a_k\theta_k.
\end{align*}
A practical self‑scaled choice for \(\tau_k\) is
\[
\tau_k =
\begin{cases}
\tau_k^{(1)}\min\!\big(\sigma_k^{-1/(n-1)},\, 1/\theta_k\big) & \text{if }\theta_k>0,\\
\min\!\big(\tau_k^{(1)}\sigma_k^{-1/(n-1)},\, \sigma_k\big) & \text{if }\theta_k\le 0,
\end{cases}
\]
where \(\tau_k^{(1)}=\min\{1,\;(\mathbf{y}_k^\top\mathbf{s}_k)/(\mathbf{s}_k^\top H_k^{-1}\mathbf{s}_k)\}\) and \(n=\dim(\Theta)\). A recommended choice for \(\phi_k\) is
\[
\phi_k = \frac{1-\theta_k}{1+a_k\theta_k}.
\]

\begin{algorithm}[H]
\caption{SSBroyden quasi‑Newton step}
\begin{algorithmic}[1]
    \State \textbf{Input:} current iterate \(\Theta_k\), inverse‑Hessian approx. \(H_k\), objective \(\mathcal J\), step \(\alpha_k\) (line search)
    \State Compute gradient \(g_k=\nabla\mathcal J(\Theta_k)\)
    \State Compute search direction \(p_k = -H_k g_k\) and line search to get \(\alpha_k\)
    \State Set \(\Theta_{k+1}=\Theta_k+\alpha_k p_k\); compute \(g_{k+1}=\nabla\mathcal J(\Theta_{k+1})\)
    \State \(\mathbf{s}_k=\Theta_{k+1}-\Theta_k,\quad \mathbf{y}_k=g_{k+1}-g_k\)
    \State Compute scalars \( \mathbf{y}_k^\top\mathbf{s}_k,\; \mathbf{y}_k^\top H_k\mathbf{y}_k\)
    \State Form \(\mathbf{v}_k = \sqrt{\mathbf{y}_k^\top H_k\mathbf{y}_k}\big(\mathbf{s}_k/(\mathbf{y}_k^\top\mathbf{s}_k) - H_k\mathbf{y}_k/(\mathbf{y}_k^\top H_k\mathbf{y}_k)\big)\)
    \State Compute \(b_k,h_k,a_k,c_k,\rho_k^-,\theta_k,\sigma_k\) and choose \(\tau_k,\phi_k\) as above
    \State Update
    \[
    H_{k+1} \leftarrow \frac{1}{\tau_k}\Big[H_k - \frac{H_k\mathbf{y}_k\mathbf{y}_k^\top H_k}{\mathbf{y}_k^\top H_k\mathbf{y}_k} + \phi_k\,\mathbf{v}_k\mathbf{v}_k^\top\Big] + \frac{\mathbf{s}_k\mathbf{s}_k^\top}{\mathbf{y}_k^\top\mathbf{s}_k}
    \]
    \State \textbf{Return} \(\Theta_{k+1}, H_{k+1}\)
\end{algorithmic}
\end{algorithm}

\subsection{Loss functions, residuals, and geometry}
\subsubsection{Lambda Weighting}

We have already seen schemes to dynamically adapt the Losses combination coefficients during training in \cref{sec:4:1:DynHypOp}. In a sense, we have seen how to change them during \textit{training time} $\tau$
\[
\LL(\theta(\tau)) = \sum_{i=1}^{n_{loss}} w_i(\tau) \LL_i (\theta (\tau)) \,.
\]

\noindent We have also seen that we can do something similar in spirit (changing how we compute the optimisation risk of out multi-objective task), but different in practice: the importance measure \cref{alg:4:2:Importance_sampling}. Indeed, each individual loss is a (Monte Carlo) integral over some distance
\[
\LL_i (\theta) = \int_D \ell(\theta; \xx) \dd^d x \,
\]
Such terms should be \textit{strongly zero}, i.e.~zero for each point in the domain; the formulation may give rise to ill-behaved solutions in negligible-size region (i.e., their contribution to the integral is small w.r.t. the other regions), but giving rise to pathologies due to the non-linear nature of the network, thus affecting the training process towards global convergence. 

That fact was behind the intuition of importance sampling; a more general formulation is to change (during training) how we \textit{locally weight} the loss terms, 
\[
\LL_i (\theta) = \int_D \lambda(\xx) \ell(\theta; \xx) \dd^d x \,.
\]
This is what is called, borrowing the terminology form \cite{NvidiaPhysicsNemoDocumentation}, \note{Obviously, is it possible to have the lambda weighting factor dependent on the training time $\tau$ (as well as the PDE time $t$), as $\lambda(\tau; t, \xx)$.}\textit{lambda weighting}, or, following \cite{toscano2025pinns}, \textit{local weights}. The importance sampling algorithm of \cref{alg:4:2:Importance_sampling} can thus be seen as a way to specify a certain lambda weighting. 

Before moving on to show a few relevant examples of lambda weighting, we first ask: what is the ground of this idea? We recall that the plain loss can be seen as the (Monte Carlo integral of the) distance in the Banach/Hilbert space (e.g., the 2-norm)
\[
    \LL_i(\theta) = d(f_\theta, f) = ||f_\theta (\xx) - f(\xx) ||_2 = \int_\Omega (f_\theta (\xx) - f(\xx) )^2 \dd^d x \,,
\]
where we have, e.g., $f_\theta(\xx) = \mathcal{F}[\hat{u}_\theta (\xx)]$ and $f(\xx) = J(\xx)$. Notice that the $\dd^d x$ term represents, on a flat open subset $\Omega \subset\RR^d$, simply the standard metric $\dd^d x = \dd x_1 \dd x_2 \cdots \dd x_d$. This means that we can interpret the lambda weighting as a \textit{change of measure}, 
\[
\dd^d x \mapsto \lambda(\xx) \dd^d x = \dd \omega(\xx) \,, 
\]
which, again, can be heuristically interpreted as adding an \textit{intrinsic curvature} to the domain $\Omega$. 

\subsubsection{Signed Distance Function as Lambda Weighting}

Sometimes, numerical problems have complex boundary conditions defined on complex geometries, which may have sharp corners; think about a simple squared region, $[-1, +1]\times[-1, +1] \subset\RR^2$, with a scalar field having to satisfy mixed boundary conditions on the the $x=-1$ and $y=+1$ boundaries. The corner point $(x,y)=(-1,+1)$ is a \textit{numerically} problematic point. One approach is thus weighting the equation residuals by the \textit{signed distance function} (SDF) of the geometries. If the geometry has sharp corners this often results in sharp gradients in the solution of the differential equation. Weighting by the SDF tends to weight these sharp gradients lower and often results in a convergence speed increase, and sometimes also improved accuracy \cite{NvidiaPhysicsNemoDocumentation}. 

We have already introduced Signed Distance Functions in \cref{subsec:4:4:4:Siren}, and the fact that they can be obtained as the solution of the \textit{Eikonal Equation}; let us formalise them a little better. 

The signed distance function (sometimes denoted as signed distance field) is the orthogonal distance of a given point $x\in \RR^d$ to the boundary of a set $\Omega\subset \RR^d$ in a metric space. The sign is determined by whether or not $x\in \overset{\circ}{\Omega}$, i.e.~it is in the interior of the set. Usually, the convention is that the function has positive values for $x\in \overset{\circ}{\Omega}$, it decreases in value as $x \to \partial \Omega$, the signed distance function is zero $\forall x \in \partial\Omega$, and it takes negative values $\forall x\notin\overline{\Omega}$. So that
\begin{definition}[Signed Distance Function]
    Let $\Omega$ be a subset of a metric space $\mathscr{X}$, and let $d : \mathscr{X} \times \mathscr{X} \to \RR$ be a distance function on $\mathscr{X}$. The \textit{unsigned distance function} of $x\in\mathscr{X}$ from $\Omega$ is the infimum of all the distances between $x$ and any $y\in\partial\Omega$, i.e.
    \[
        d_{\partial\Omega}(x) \defeq   \inf_{y \in {\partial} \Omega} d(x, y) ,
    \]
    The \textit{signed distance function} $\sdf$ is
    \begin{equation}
        \sdf_{\partial\Omega}(x) \defeq \begin{cases}
            + d_{\partial\Omega}(x) \,, &x \in \overset{\circ}{\Omega} \,, \\
            - d_{\partial\Omega}(x) \,, &x \notin \overline{\Omega} \,, \\
            0 \,, & x\in \partial \Omega \,.
        \end{cases} 
    \end{equation}
\end{definition}

An interesting fact is that signed distance functions satisfy a simple Eikonal equation almost everywhere 
\[
|\nabla \sdf (x) | = 1 \,,
\]
and, on $\partial \Omega$, we have
\[
\left. \nabla\sdf(x) \right|_{\partial \Omega} = n(x) \,, 
\]
where $n$ is the orthonormal vector to $\partial \Omega$. 

While various algorithm to solve the Eikonal equation to compute the signed distance function exists, sometimes it is beneficial to resort to different approximations of the Eikonal equation, due to the nature of the domain; a relevant example is the \textit{Screened Poisson Equation} $\Omega$ \cite{whytock2014dynamic}:
\begin{align*}
    v(x) - t \Delta v(x) &= 0 \,, \qquad x \in \Omega, \\
    v(x) &= 1 \,, \qquad x\in\partial \Omega ,
\end{align*}
where $t$ is a small, positive, real arbitrary parameter. In \cite{varadhan1967behavior} it was shown that
\[
    \lim_{t\to 0} - \sqrt{t} \, \log [v(x)] = d_{\partial\Omega}(x) \,.
\]
Indeed, we can heuristically retrieve that result; if we call $u(x)\defeq - \sqrt{t} \, \log |v(x)| $, and placing it inside the Screened Poisson Equation\note{Plugging it in \(v(x)=\exp [ - u(x)/\sqrt{t}]\) as a change of variable.}, we express it as a \textit{regularised Eikonal equation}:
\begin{align*}
    (1 - |\nabla u|^2)  + \sqrt{t} \Delta u(x) &=0 \,, \qquad x\in\Omega, \\
    u(x) &= 0, \qquad x\in \partial \Omega,
\end{align*}
which is, in the $t\to 0$ limit, the Eikonal Equation for the SDF. This ``Screened Poisson Equation'' version is the one implemented in NVIDIA PhysicsNemo \cite{NvidiaPhysicsNemoDocumentation}.

\begin{figure}[t]
    \centering
    \includegraphics[width=0.75\linewidth]{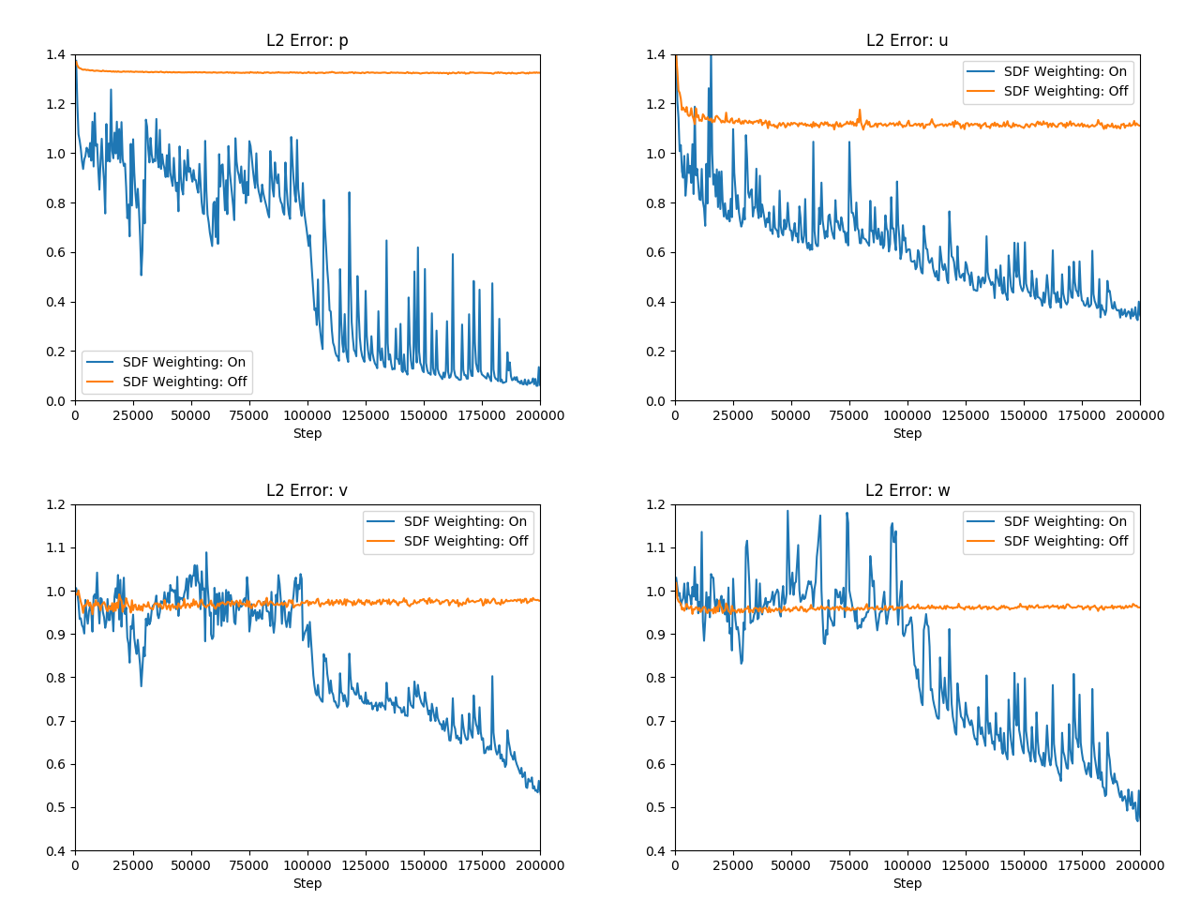}
    \caption{$L_2$ errors for one example of laminar flow (Reynolds number 50) over a 17 fin heat sink in the initial 100,000 iterations. The multiple closely spaced thin fins lead to several sharp gradients in flow equation residuals in the vicinity of the heat sink. Weighting them spatially, the dominance of these sharp gradients during the iterations are essentially minimised, thus achieving a faster rate of convergence. From \cite{NvidiaPhysicsNemoDocumentation}.}
    \label{fig:4:5:2:1:SDF_nvidia}
\end{figure}

In \cref{fig:4:5:2:1:SDF_nvidia} is reported the comparison of a training with and without SDF lambda weighting in a complex geometry, a 17 fin heat sink. It is evident a convergence speed up, for all four fields (pressure $p$, fluid velocities $u,v,w$ along $x,y,z$ axis respectively).

\subsubsection{Residual Based Attention as Lambda Weighting}

A relevant, yet numerically simple choice for the lambda weighting function was introduced in \cite{anagnostopoulos2023residual}; the authors introduced \textit{residual-based attention} (RBA) weights, where $\lambda_i(\tau; t, \xx_i)$ is computed using the exponentially weighted moving average of the residuals $r_i(t, \xx)$. Since residuals contain information about regions with high error, this method proved to be highly effective, outperforming previous approaches with minimal computational cost. 

As we have already noticed, one of the inherent challenges of PINN training is that the residuals of key collocation points can get overlooked by the mean calculation of the objective function. Consequently, despite a decrease in total loss during training, certain spatial (or temporal) characteristics of the searched solution might not be fully captured. 

The goal of \cite{anagnostopoulos2023residual} was the development of a simple, gradient-less weighting scheme based on the rolling history of the cumulative residuals to extend the ``attention'' span\note{Not in the sense of Multi-Head Attention.} of the optimizer. The proposed scheme aims to increase attention to the challenging regions of information propagation in both space and time dimensions defining the problem.
  
The update rule for RBA for any collocation point $i$ at the iteration $k$ is given by:

\begin{equation}
  \lambda_i^{(k+1)} \leftarrow \gamma\lambda_i^{(k)}+\eta^*\frac{|r_i|}{\max_{i}({|r_i|})},\qquad i \in \{0, 1, ..., N_r\},
\end{equation}
  
\noindent   where $N_r$ is the number of collocation points, $\gamma$ is the decay parameter, $\eta^*$ is the learning rate,  and $r_i$ is the PDE residual for point $i$
\[
    r_i \defeq (\mathcal{F}[\hat{u}_\theta(\xx_i)]-J(\xx_i))^2 \,.
\]
Note that the learning rate of the optimizer $\eta$ and the learning rate of the weighting scheme $\eta^*$ are two different hyperparameters.
  
\begin{algorithm}
\caption{Residual‑Based Attention Gradient Descent (from \cite{anagnostopoulos2023residual})}
\label{alg:4:4:2:3:residual-attention}
\begin{algorithmic}[1]
\Require training set, learning rates $\eta,\ \eta^*$, decay $\gamma\in[0,1]$, lower bound $c\ge0$, batch size, stopping criterion
\State initialise $w_i$ with Xavier for all $i\in\{0,\dots,N_r\}$
\State initialise $\lambda_i \leftarrow 0$ for all $i$
\While{stopping condition not met}
  \For{each batch of training examples}
    \State $\lambda_i \leftarrow \gamma\,\lambda_i + \eta^*\,\dfrac{|r_i|}{\max_j |r_j|}$ \Comment{RBA weight update}
    \State $\lambda_i \leftarrow \lambda_i + c$ \Comment{enforce lower bound shift}
    \State $\mathcal{L}_r^* \leftarrow \dfrac{1}{N_r}\sum_{i=1}^{N_r} \big(\lambda_i\, r_i\big)^2$ \Comment{weighted residual loss}
    \State $\mathcal{L} \leftarrow \lambda_{ic}\,\mathcal{L}_{ic} + \lambda_{bc}\,\mathcal{L}_{bc} + \mathcal{L}_r^*$ \Comment{total loss}
    \State \(\;w_i \leftarrow w_i - \eta\, \nabla_w \mathcal{L}\) \Comment{Update parameters}
  \EndFor
\EndWhile
\State \Return optimized parameters $\{w_i\}$
\end{algorithmic}
\end{algorithm}

\noindent The main advantages of the RBA weighting scheme are:
\begin{enumerate}
  \item Deterministic operation scheme, bounded by the $\gamma$ and $\eta^*$ parameters, which ensure the absence of exploding multipliers. 
  \item No training or gradient calculation is involved for computing $\lambda$, leading to negligible additional computational cost.
  \item Scaling with the cumulative residuals guarantees increased attention on the solution fronts where the PDEs are unsatisfied in both spatial and temporal dimensions.
\end{enumerate}

The modified training process for an indicative standard Gradient Descent optimization method is outlined in \cref{alg:4:4:2:3:residual-attention}. 
A constant $c$ may be added to the updated weights to raise the lower bound and adjust the ratio $\max(\lambda_i)/\min(\lambda_i)$.

\subsubsection{Temporal Loss Weighting: causal loss} \label{subsubsec:4:4:2:4:CausalLoss}
We have seen methods to change the local weighting of the loss w.r.t. the space dependence. But usually, when dealing with time-dependent problems, a major issue is to ensure a smooth, well-behaved time evolution of the solution. This is especially crucial in auto-regressive models, which treat time evolution as an auto-regressive predictive step (in \cref{chap:5:NeuralOperators}, we will encounter this kind of behaviour when dealing with Neural Operators). The most important task is to ensure \textit{causality} in the solution, which is quite relevant for non-linear equations which may present chaotic behaviour. 

Assume we need to solve a time-dependent evolution problem with a PDE of the form
\[
\partial_t u (t, \xx) + \mathcal{N}[u(t,\xx)] = 0 \,, \quad (t,\xx) \in [0,T] \times \Omega \,.
\]

A very simple approach is to blend Temporal loss weighting and a time-marching schedule \cite{krishnapriyan2021characterizing}, a specific case of curriculum learning strategy; a simple example of temporal loss weighting is, e.g., linear decay
\[
    \lambda(t) = 1 + C_T \left( 1 - \frac{t}{T} \right) \,, \quad t\in [0, T].
\]
In this way, the learning is penalised more highly when for errors at earlier time. Additionally, a sampling strategy called time-marching schedule can be used to increase the time span of the simulation during training time, for example having a linearly moving time interval
\[
t \in \left[ 0, T(n) \right] \,, \quad T(n) \defeq T \min \left( 1, \frac{n}{N}\right) \,,
\]
where $n$ is the training epoch and $N$ is the ending epoch of the time marching schedule (for example, half the whole training epochs total).

\begin{figure}[t]
    \centering
    \includegraphics[width=0.95\linewidth]{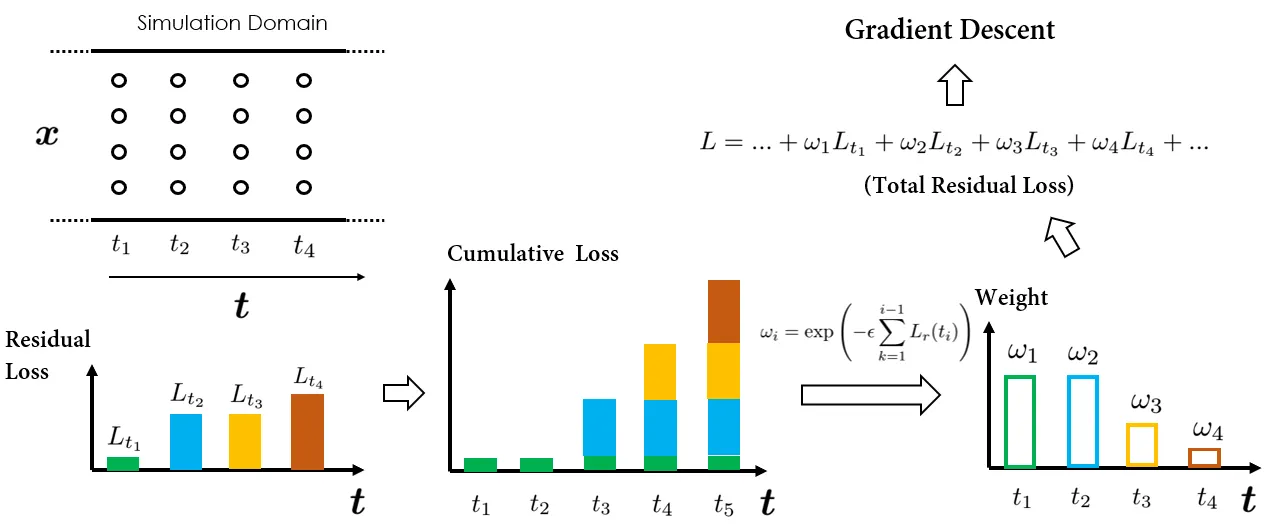}
    \caption{Illustration of the causal training algorithm. From \cite{Guo2023causal}.}
    \label{fig:4:4:2:4:CausaLoss}
\end{figure}

A most common approach, called \textit{causal loss}, was proposed in \cite{wang2022respecting}. Indeed, they show that continuous-time PINNs models can violate temporal causality, and hence are susceptible to converge towards erroneous solutions for transient problems. Thus, causal loss was introduced to address this issue and a key source of error by reformulating the PDE residual loss to account explicitly for physical causality during model training.
The idea is to split the time interval into chunks $\{ [t_i, t_{i+1}] \}_{i=0}^{N-1}$ and then compute the loss for each chunk
\[
\mathcal{L}_i(\mathbf{\theta}) = \sum_j \left| \frac{\partial u_{\mathbf{\theta}}}{\partial t}(t_j, \xx_j)
                    + \mathcal{N}[u(t_j, \xx_j)] \right|^2 \,, \quad \{t_j, x_j\} \subset [t_{i-1}, t_i] \times \Omega \,.
\]

\begin{algorithm}[ht]
\caption{Causal Training for Physics-Informed Neural Networks. From \cite{wang2022respecting}.}
\label{alg4:4:2:4:causalloss}
\begin{algorithmic}[1]
\Require training set $\{t_i\}_{i=0}^{N_t}$, learning rate $\eta$, 
         hyperparameters $\{\epsilon_j\}_{j=1}^k$, decay threshold $\delta$,
         loss weights $\lambda_{ic},\lambda_{bc}$
\State initialise network parameters $\boldsymbol{\theta}$ (e.g., Xavier)
\State initialise temporal weights $w_i \leftarrow 1$ for all $i=0,\dots,N_t$
\For{each $\epsilon$ in $\{\epsilon_1,\dots,\epsilon_k\}$}   \Comment{outer causal loop}
    \For{$n = 1,\dots,S$}   \Comment{inner gradient descent steps}
        \State Compute losses $\mathcal{L}(t_i,\boldsymbol{\theta})$ for all $i$
        \Statex \hspace{\algorithmicindent} where 
        \Statex \hspace{\algorithmicindent} $\mathcal{L}(t_0,\boldsymbol{\theta}) = \lambda_{ic}\,\mathcal{L}_{ic}(\boldsymbol{\theta})$
        \Statex \hspace{\algorithmicindent} $\mathcal{L}(t_i,\boldsymbol{\theta})$ for $i\ge 1$ is the PDE residual loss
        \State \textbf{Update temporal weights:}
        \For{$i = 2,\dots,N_t$}
            \State $w_i \leftarrow \exp\!\left(-\epsilon \sum_{k=1}^{i-1} \mathcal{L}(t_k,\boldsymbol{\theta})\right)$
        \EndFor
        \State \textbf{Compute weighted total loss:}
        \State $\displaystyle 
            \mathcal{L}(\boldsymbol{\theta}) 
            = \frac{1}{N_t}\sum_{i=0}^{N_t} w_i\,\mathcal{L}(t_i,\boldsymbol{\theta})
        $
        \State \textbf{Gradient descent update:}
        \State $\boldsymbol{\theta} \leftarrow 
                \boldsymbol{\theta} - \eta\,\nabla_{\boldsymbol{\theta}}\mathcal{L}(\boldsymbol{\theta})$
        \If{$\min_i w_i > \delta$}
            \State \textbf{break}   \Comment{stop inner loop early}
        \EndIf
    \EndFor
\EndFor
\State \Return $\boldsymbol{\theta}$
\end{algorithmic}
\end{algorithm}

The full PDE loss is then obtained by summing over chunks \cite{wang2022respecting, penwarden2023unified}, 
\[
\mathcal{L}_r(\mathbf{\theta}) = \sum_{i=1}^N w_i \mathcal{L}_i(\mathbf{\theta}) \,,
\]
each of which is weighted as 
\[
w_i = \exp\left[-\epsilon \sum_{k=1}^{i-1} \mathcal{L}_k(\mathbf{\theta}) \right], \quad \forall i=2,3, \dots, N.
\]

\noindent In full form, we have
\begin{equation}\label{eq:4:4:2:4:CausaLoss}
    \mathcal{L}_{pde} (\theta)  = \frac{1}{N}\sum_{i=1}^N \e^{-\epsilon \sum_{k=1}^{i-1} \mathcal{L}_k(\mathbf{\theta}) } \, \mathcal{L}_i(\mathbf{\theta}) \,.
\end{equation}

Note that the causal weight $w_i$ is inversely exponentially proportional to the magnitude of the cumulative residual loss from the previous chunks. As a consequence, the loss at step (chunk) $i$, $\mathcal{L}_i(\mathbf{\theta})$ will almost not be affected during training unless all previous residuals decrease to some small value such that its weight $w_i$ is large enough.
Notice that the $w_i$ must be computed as a parameter, i.e. they have to remain \textit{outside} of the computational graph (e.g., by using \textsc{detach} in pytorch.).

Now, notice that \cref{eq:4:4:2:4:CausaLoss} can be seen as the discretised version of 
\[
    \mathcal{L}_{pde} (\theta)  = \int_0^T \e^{-\epsilon\int_0^t \mathcal{L}(\mathbf{\theta}; s) \dd s } \, \mathcal{L}(\mathbf{\theta}; t)  \, \dd t\,,
\]
with the appropriate redefinition of $\epsilon \mapsto \epsilon \Delta t_k$. This form is strikingly similar to the generic solution of the first-order ordinary differential problem
\begin{align*}
    \dot{y}(t) + \epsilon \mathcal{L}(t) y(t) &= \mathcal{L}(t) \,,  \qquad y(0) = 0\,, \\
    \Rightarrow\dot{y}(t) + \epsilon  \mathcal{L}(t) \left( y(t) - \frac{1}{\epsilon}\right) &= 0 \,,  \qquad \quad y(0) = 0\,, \\
    \Rightarrow\dot{z}(t) + \epsilon  \mathcal{L}(t) z(t) &= 0 \,, \qquad \quad z(0) =  -\frac{1}{\epsilon} \,, \text{ where }z(t) \defeq y(t) -  \frac{1}{\epsilon} \,.
\end{align*}
Heuristically, we can see our causal loss objective $y(t) = \int_0^T \e^{-\epsilon\int_0^t \mathcal{L}(\mathbf{\theta}; s) \dd s } \, \mathcal{L}(\mathbf{\theta}; t)  \, \dd t$ as a exponentially damped state, relaxed at a rate proportional to $\epsilon  \mathcal{L}(t)$.

\section{Recap-ish: \textit{An Expert's Guide to Training Physics-informed Neural Networks}}

In this lecture, we have seen multiple peculiarities, issue, and difficulties experienced when training deep neural networks to solve differential problems of the form
\begin{align}
    \partial_tu(t,\xx) + \mathcal{N}[u(t, \xx)] &= 0 \,, \qquad \qquad (t, \xx) \in [0, T] \times \Omega  \qquad (\text{PDE}) , \label{eq:4:5:PDE}\\
    u(0, \xx) &= g(\xx) \,, \qquad \quad \xx \in \Omega \qquad\qquad\qquad\; (\text{IC}), \label{eq:4:5:IC}\\
    \mathcal{B}[u(t, \xx)] &= 0 \,, \qquad \qquad (t, \xx) \in [0, T] \times \partial\Omega \quad (\text{BC}). \label{eq:4:5:BC}
\end{align}
Let us recap the basic steps for training a PINN. 

We can define a Monte Carlo sampling of the domain $[0,T]\times \Omega$ as $\{t_{r}^i, \mathbf{x}_{r}^i\}_{i=1}^{N_{r}}$ to impose locally the PDE constraint, and, similarly, $\{\mathbf{x}_{ic}^i\}_{i=1}^{N_{ic}}$ and $\{t_{bc}^i, \mathbf{x}_{bc}^i\}_{i=1}^{N_{bc}}$ for the IC/BC constraint imposition. From here, by defining a deep neural network as an universal approximator of the solution, $u_\theta(t, \xx) \sim u(t, \xx)$, where $\theta$ are the learnable parameters describing the neural network, we can compute the PDE residuals as
\begin{align}
    \label{eq:4:5:pde_residual}
    \mathcal{R}_{\mathbf{\theta}}(t, \mathbf{x}) = \frac{\partial \mathbf{u}_{\mathbf{\theta}}}{\partial t}(t_r, \mathbf{x}_r) + \mathcal{N}[\mathbf{u}_{\mathbf{\theta}}(t_r, \mathbf{x}_r)] \,.
\end{align}
From here, as usual, we have the full multi-objective loss
\begin{align}
    \label{eq:4:5:PINN_loss}
    \mathcal{L}(\mathbf{\theta}) =  \mathcal{L}_{ic}(\mathbf{\theta}) +  \mathcal{L}_{bc}(\mathbf{\theta}) +   \mathcal{L}_r(\mathbf{\theta}), 
\end{align}
where each component is 
\begin{align}
    \label{eq:4:5:loss_ic}
     \mathcal{L}_{ic}(\mathbf{\theta}) &= \frac{1}{N_{ic}} \sum_{i=1}^{N_{ic}} \left| \mathbf{u}_{\mathbf{\theta}}(0, \mathbf{x}_{ic}^i) - \mathbf{g}(\mathbf{x}_{ic}^i) \right|^2, \\
     \label{eq:4:5:loss_bc}
     \mathcal{L}_{bc}(\mathbf{\theta}) &= \frac{1}{N_{bc}} \sum_{i=1}^{N_{bc}} \left| \mathcal{B}[\mathbf{u}_{\mathbf{\theta}} ( t_{bc}^i, \mathbf{x}_{bc}^i)] \right|^2, \\
    \label{eq:4:5:loss_r}
    \mathcal{L}_r(\mathbf{\theta}) &= \frac{1}{N_r} \sum_{i=1}^{N_r} \left| \mathcal{R}_{\mathbf{\theta}}(t^i_r, \mathbf{x}^i_r) \right|^2. 
\end{align}

In \cite{wang2023expert}, authors propose a baseline training pipeline for PINNs, blending Fourier embeddings \cref{subsec:4:4:3:SpectralBias}, Random Weight Factorisation \cref{4:2:1:2:WeightFactor}, causal loss training \cref{subsubsec:4:4:2:4:CausalLoss}, and a hyperparameter update (NTK-based or gradient-norm) \cref{subsec:4:1:7:NTK}. The main ingredients of the suggested recipe is what follows (see \cref{alg:4:5:PINN_training_pipline}):
\begin{itemize}
    \item[0.] Adimensionalise the governing equations and variables;
    \item[1.] Use a MLP with Fourier embeddings and Random Weight Initialisation;
    \item[2.] For time-dependent problems, use causal loss for PDE residuals
    \item[3.] Every tot epochs, rebalance the multi-objective loss balancing parameters, using an exponential moving average. 
\end{itemize}

In \cite{wang2023expert}, the recommended default values for hyper-parameters are as follows: the NTK update every $f=1,000$ epochs, and its exponential moving average hyperparameter $\alpha=0.9$;  the causal loss hyperparameter $\epsilon=1.0$. As already noticed in \cref{subsubsec:4:4:2:4:CausalLoss}, it is crucial that the the back-propagation of the weights $w_i$'s and $\lambda_i$'s with respect to network parameters $\mathbf{\theta}$ is frozen.

\begin{algorithm}[ht]
\caption{Training pipeline of physics-informed neural networks. From \cite{wang2023expert}.}
\label{alg:4:5:PINN_training_pipline}
\begin{algorithmic}
{\small 
\State {1. Non-dimensionalise the PDE system \eqref{eq:4:5:PDE}.}

\State{2. Represent the PDE solution by a multi-layer Perceptron network (MLP) $\mathbf{u}_{\mathbf{\theta}}$  with Fourier feature embeddings and random weight factorisation. In general, we recommend using $\tanh$ activation and initialised using the \textit{Glorot} scheme.}

\State{3. Formulate the  weighted loss function according to the PDE system:
\begin{align}
    \mathcal{L}(\mathbf{\theta}) =  \lambda_{ic} \mathcal{L}_{ic}(\mathbf{\theta}) + \lambda_{bc} \mathcal{L}_{bc}(\mathbf{\theta}) +  \lambda_r \mathcal{L}_r(\mathbf{\theta}), 
\end{align}
where $\mathcal{L}_{ic}(\mathbf{\theta})$ and $\mathcal{L}_{bc}(\mathbf{\theta})$ are defined in \eqref{eq:4:5:loss_ic}, \eqref{eq:4:5:loss_bc} respectively, and 
\begin{align}
    \label{eq:4:5:splited_res_loss}
    \mathcal{L}_r(\mathbf{\theta}) = \frac{1}{M} \sum_{i=1}^M w_i \mathcal L_r^i(\mathbf{\theta}).
\end{align}
The temporal domain is partitioned into $M$ equal sequential segments and $\mathcal L_r^i$ denotes the PDE residual loss within the $i$-th segment of the temporal domain. 
}

\State{4. Set all global weights $\lambda_{ic}, \lambda_{bc}, \lambda_{r}$ and temporal weights $\{w_i\}_{i=1}^M$ to 1.}

\State{5. Use $S$ steps of a gradient descent algorithm to update the parameters $\theta$ as:}

\For{$n = 1, \dots, S$} 

 \State{(a) Randomly sample $\{\mathbf{x}_{ic}^i\}_{i=1}^{N_{ic}}$, $\{t_{bc}^i, \mathbf{x}_{bc}^i\}_{i=1}^{N_{bc}}$ and $\{t_{r}^i, \mathbf{x}_{r}^i\}_{i=1}^{N_{r}}$ in the computational domain and evaluated each loss terms $ \mathcal{L}_{ic}, \mathcal{L}_{bc}$ and $\{\mathcal{L}_r^i\}_{i=1}^M$.}

  \State{(b) Compute and update the temporal weights by
  \begin{align}
  \label{eq:4:5:temporal_update}
    w_i = \exp\left(- \epsilon \sum_{k=1}^{i-1}  \mathcal{L}_r^k( \mathbf{\theta})\right)    , \text{ for } i = 2, 3, \dots, M.
  \end{align}
  Here $\epsilon > 0$ is a user-defined hyper-parameter that determines the ``slope'' of temporal weights.   
 }
    
 \If{$n \mod f = 0$ }

 \State{(c) Compute the global weights by
  \begin{align}
  \label{eq:4:5:lambda_ic_update}
     \hat{\lambda}_{ic} &=  \frac{  \|\nabla_{\theta}  \mathcal{L}_{ic}(\theta)\| +  \|\nabla_{\theta}  \mathcal{L}_{bc}(\theta)\| +  \|\nabla_{\theta}  \mathcal{L}_{r}(\theta)\|   } {\|\nabla_{\theta}  \mathcal{L}_{ic}(\theta)\|}, \\
     \label{eq:4:5:lambda_bc_update}
      \hat{\lambda}_{bc} &= \frac{  \|\nabla_{\theta}  \mathcal{L}_{ic}(\theta)\| +  \|\nabla_{\theta}  \mathcal{L}_{bc}(\theta)\| +  \|\nabla_{\theta}  \mathcal{L}_{r}(\theta)\|   } {\|\nabla_{\theta}  \mathcal{L}_{bc}(\theta)\|},  \\
      \label{eq:4:5:lambda_r_update}
       \hat{\lambda}_{r} &=  \frac{  \|\nabla_{\theta}  \mathcal{L}_{ic}(\theta)\| +  \|\nabla_{\theta}  \mathcal{L}_{bc}(\theta)\| +  \|\nabla_{\theta}  \mathcal{L}_{r}(\theta)\|   } {\|\nabla_{\theta}  \mathcal{L}_{r}(\theta)\|}, 
  \end{align}
where $\|\cdot\|$ denotes the $L^2$ norm.
   } 
   
\State{(d) Update the global weights $\mathbf{\lambda} = (\lambda_{ic}, \lambda_{bc}, \lambda_{r})$ using a moving average of the form
  \begin{align}
  \label{eq:4:5:lambda_update}
      \mathbf{\lambda}_{\text{new}} =  \alpha  \mathbf{\lambda}_{\text{old}}  + (1 - \alpha) \hat{\mathbf{\lambda}}_{\text{new}} .
  \end{align}}
where the parameter $\alpha$ determines the balance between the old and new values

 \EndIf
  
 \State{(e) Update the parameters $\theta$ via gradient descent
  \begin{align}
  \label{eq:4:5:theta_update}
      \theta_{n+1} &= \theta_{n} - \eta \nabla_{\mathbf{\theta}}\mathcal{L}(\mathbf{\theta}_n) 
  \end{align}}

 \EndFor

} 
\end{algorithmic}
\end{algorithm}

\newpage
\section{[EXTRA]: Neural ODEs}

Another interesting field bridging differential equations, numerical methods and neural network is the topic of \textit{Neural ODEs}, introduced in \cite{chen2018neural} (part of this section is based on \cite{murari2025nnasdyn, munari2025phd, okamoto2025learning, kidger2022neural}).

The motivation of Neural ODEs comes from residual connections \cite{he2016deep}, where we add the identity to the layer forward pass:
\[
  y_{t+1} = F_\theta (y_t;t) \;\longrightarrow\; y_{t+1} = y_t + \epsilon  F_\theta (y_t; t) \,.
\]
From here, we can notice, as done in \cite{lu2018beyond}, that residual connections can be interpreted as a discretised derivative (with forward Euler):
\[
    \partial_t y(t) \simeq \frac{y_{t+1} - y_{t}}{\epsilon } = F_\theta (y(t); t).
\]

More precisely, we have a dynamical model with full Cauchy formulation
\begin{equation}
\begin{split}
    \dot{y}(t) &= F_\theta (y(t); t) \,, \qquad F_\theta : \RR \times \RR^h \to \RR^h , \\ 
    y(0) &= A x_0 + b \in \RR^h \,, 
\end{split}  
\end{equation}

The idea of Neural ODEs is, given a task to map $x_0 \to y$ to model $F_\theta$ with a neural network, and then \textit{solve} the ODE system to get the prediction $x_0 \to y(T)$, where $T$ is  a prescribed terminal integration horizon (which should be around the convergence time of the system). This means that \textit{during inference}, supposing to have access to a method $\mathrm{ODESolver}$ which solves the ODE system, $\mathrm{ODESolver} : (t, F_\theta, x_0) \to y(t)$, 
\begin{enumerate}
    \item The input $x_0 \in \RR^p$ becomes the initial condition for the ODE, via the map $x_0 \mapsto Ax_0 + b = y(0) \in \RR^h$;
    \item The neural net $F_\theta$ defines the dynamics;
    \item The ODE solver computes the final hidden state, i.e., the output:
    \begin{equation}
        y(t) = \mathrm{ODESolver}(t, F_\theta, x_0) .
    \end{equation}
\end{enumerate}

This means that the solver adaptively chooses how many steps to take and how fine-grained they need to be, depending on how complex the function $F_\theta$ is for a particular input.

If the $\mathrm{ODESolver}$ is \textit{differentiable}, we can train the Neural ODE with the standard backpropagation method, with the crucial difference that we do not compare the output of the neural network $F_\theta$ \textit{directly} with the output, but it is used to produce the output via the $\mathrm{ODESolver}$; i.e., if I have a dataset $\mathcal{D} = \{x_i, y_j\}_{i=1,\ldots, N}$, we have
\begin{equation}
    \mathcal{L}[\theta] = \frac{1}{N}\sum_{i=1}^N \left( y_j - \mathrm{ODESolver}(t, F_\theta, x_0)\right)^2
\end{equation}

If the $\mathrm{ODESolver}$ is not differentiable, or if we see that the standard backprop here is kind of inefficient, we can use a different optimisation approach: the \textit{adjoint method} \cite{pontryagin2018mathematical}.

The adjoint method is based on a different approach to compute the updating gradient, $g_\theta = \nabla_\theta \mathcal{L}[\theta]$; 
Let us restart from $\mathcal{L}[\theta]$. We see that, by chain rule,
\[
\nabla_\theta \mathcal{L}[\theta] = \left(  \frac{\partial y(t)}{\partial\theta}\right)^T \frac{\partial \mathcal{L}}{\partial y(t)} \,.
\]
Now let us introduce the so-called \textit{adjoint variable} $a(t)$, 
\begin{equation}\label{eq:4:6:adjointvariable}
    a(t) \defeq \frac{\partial \mathcal{L}}{\partial y(t)} \,.
\end{equation}
Now, notice that, selected a small $\epsilon$,
\[
    y(t+\epsilon) = y(t) + \int_t^{t+\epsilon} F_\theta (y(t); t ) \ \mathrm{d}t  \approx y(t) + \epsilon\,  F_\theta (y(t); t ) \,,
\]
and thus
\[
    a(t) = \frac{\partial \mathcal{L}}{\partial y(t)} = \frac{\partial \mathcal{L}}{\partial y(t+\epsilon)} \frac{\partial y(t+\epsilon)}{\partial y(t)} = a(t+\epsilon) \frac{\partial y(t+\epsilon)}{\partial y(t)} 
\]
Now, let us compute, for $\epsilon\to0$,
\begin{equation}
    \begin{split}
        \dot{a}(t) &= \lim_{\epsilon\to 0} \frac{a(t+\epsilon) - a(t)}{\epsilon} \\
        &=  \lim_{\epsilon\to 0} \frac{ a(t+\epsilon) - a(t+\epsilon) \frac{\partial y(t+\epsilon)}{\partial y(t)} }{\epsilon} \\
        &= \lim_{\epsilon\to 0}  \frac{ a(t+\epsilon) - a(t+\epsilon) \frac{\partial}{\partial y(t)}\left(  y(t) + \epsilon \, F_\theta(y(t); t) \right) + \mathcal{O}(\epsilon^2) }{\epsilon} \\
        &=  -  \lim_{\epsilon\to 0}  a(t+\epsilon) \frac{\partial  F_\theta (y(t); t)}{\partial y(t)} + \mathcal{O}(\epsilon)     \\
        &=  - a(t) \frac{\partial  F_\theta (y(t); t)}{\partial y(t)}   \,.
    \end{split}
\end{equation}
We can now define
\[
    J_\theta (t) \defeq   \frac{\partial y(t)}{\partial\theta} \,,
\]
so that
\[
\nabla_\theta \mathcal{L}[\theta] = \left(J_\theta(t)\right)^T \frac{\partial \mathcal{L}}{\partial y(t)} = \left(J_\theta(t)\right)^T a(t) \,.
\]
Now, notice that $J_\theta$ satisfies
\[
\begin{split}
   \frac{\dd J_\theta (t)}{\dd t} &= \frac{\dd }{\dd t}\frac{\dd y(t)}{\dd\theta} \\ 
   &= \frac{\dd}{\dd \theta} \frac{\dd }{\dd t} y(t)  \\
   &= \frac{\dd}{\dd \theta} \, F_\theta (y(t; \theta), t)\\
   &= \frac{\partial F_\theta}{\partial \theta} + \frac{\partial F_\theta}{\partial y(t)}  \frac{\partial y(t)}{\partial \theta}  \\
   &= \frac{\partial F_\theta}{\partial \theta} + \frac{\partial F_\theta}{\partial y(t)} J_\theta(t) \,.
\end{split}
\]
Formally, we can write
\[
 \left(J_\theta(t)\right)^T a(t) = \left(J_\theta(0)\right)^T a(0) + \int_0^t  \left[ \frac{\dd }{\dd s} \, \left(J_\theta(s)\right)^T a(s) \right] \mathrm{d}s \,,
\]
where $J_\theta(0) = \frac{\partial y_0}{\partial \theta} = 0$, since the input variable is model-independent; this is helpful since we can write
\begin{equation}
    \begin{split}
        \frac{\dd }{\dd t} \, \left(J_\theta(t)\right)^T a(t) &= \frac{\dd\left(J_\theta(t)\right)^T }{\dd t} \,  a(t)  + \left(J_\theta(t)\right)^T \dot{a}(t) \\
        &=\left[ \frac{\partial F_\theta}{\partial \theta} + \frac{\partial F_\theta}{\partial y(t)} J_\theta(t) \right]^T a(t) - \left(J_\theta(t)\right)^T  a(t) \frac{\partial  F_\theta (y(t); t)}{\partial y(t)} \\
        &= a(t) \left( \frac{\partial F_\theta}{\partial \theta} \right)^T
    \end{split}
\end{equation}

This allows to compute directly the loss gradient:
\begin{equation}
    \begin{split}
        \nabla_\theta \mathcal{L} &= \left(J_\theta(T)\right)^T a(T) \\
        &= \int_0^T  \left[ \frac{\dd }{\dd t} \, \left(J_\theta(t)\right)^T a(t) \right] \mathrm{d}t \\
        &= \int_0^T  \left[  \left( \frac{\partial F_\theta}{\partial \theta} \right)^T  a(t) \right] \mathrm{d}t \\
        &= - \int_T^0  \left[  \left( \frac{\partial F_\theta}{\partial \theta} \right)^T  a(t) \right] \mathrm{d}t\,,
    \end{split} 
\end{equation}
which can be used to update the weights of our Neural ODE.  

The adjoint method provides a memory-efficient alternative to standard backpropagation by solving an additional ODE backward in time, avoiding the need to store the entire computational graph of the forward pass. 

Notice that we have found three (interrelated) ODEs, describing the system:
\begin{align}
    \dot{y} &= F_\theta (y(t; \theta), t) \, , \qquad \quad \; (\text{Neural ODE def})\\
    \dot{a} &=  - a(t) \frac{\partial  F_\theta (y(t); t)}{\partial y(t)}  \, , \quad (\text{Adjoint ODE})\\
    \dot{g}_\theta &= \left( \frac{\partial F_\theta}{\partial \theta} \right)^T  a(t) \, , \qquad\quad (\text{Gradient ODE})
\end{align}
Now, while in the \textit{forward pass} we integrate the Neural ODE relation to compute the model output, in the \textit{backward pass}, we simply start from the end-point and integrate \textit{backward} in time, i.e. we send $t\to -t$.
To summarise it, the passage are:
\begin{enumerate}
    \item \textbf{Forward:} Solve $\dot{y} = F_\theta(y,t)$, $y(0) = Ax_0 + b$ up to time $T$.
    \item \textbf{initialise:} $a(T) = \partial_{y(t)} \mathcal{L}$, and the gradient $g_\theta = 0$.
    \item \textbf{Backward:} Solve the augmented system \textit{backwards} from $T$ to $0$:
    \[
    \frac{\dd}{\dd t} \begin{bmatrix} y \\ a \\ g_\theta \end{bmatrix} = 
    \begin{bmatrix} -F_\theta(y,t) \\ (\partial_{y(t)} F_\theta)^\top a \\ -(\partial_\theta F_\theta)^\top a \end{bmatrix}.
    \]
    \item \textbf{Output:} $\nabla_\theta \mathcal{L} = g_\theta(0)$.
\end{enumerate}

This allows us to compute gradients \textit{without backpropagating through through the operations of the solver}. Furthermore, this approach allows us to not store any intermediate quantities of the forward pass, which allows us to train Neural ODEs with constant memory cost as a function of depth, which, instead, is a major bottleneck of training standard deep models. 

\subsection{Neural ODEs Applications and Connections}

Neural ODEs have found successful application across a variety of domains that require continuous-time or irregularly-sampled modelling. 
A prominent use case is the analysis of \textit{irregularly-sampled time series}, where the continuous state evolution of a Neural ODE naturally accommodates non-uniform observation intervals~\cite{rubanova2019latent}. 
The model has also been applied in \textit{physical sciences and system identification}, where the differential formulation can recover underlying dynamical laws from observational data~\cite{chen2018neural}. 
Other areas of application include \textit{time-series forecasting}~\cite{zhang2025survey}, \textit{tabular data generation}~\cite{zhang2025survey}, \textit{image outpanting and image reconstruction}, and \textit{graph representation learning}~\cite{zhang2025survey}, showcasing the versatility of continuous-depth models.

We have started this section with a conceptual link between Neural ODEs and residual networks, in particular from the observation that a ResNet layer resembles a forward Euler discretization of the ODE with step size \(\epsilon\)~\cite{chen2018neural}. 
While this intuition is pedagogically useful, it is important to clarify that the rigorous correspondence holds primarily under the assumption of \textit{shared weights across layers}. 
Indeed, \cite{avelin2019neural} proved that only in the deep limit of a ResNet where each layer shares the same weight matrix does the stochastic gradient descent dynamics converge to that of a Neural ODE. 
For general ResNets with layer-varying parameters, the continuous limit is described by a more complex setting, which needs to keep into account the variation of the parameters $\theta$ with the ``depth/time'', such as the Neural Generalised ODEs \cite{yu2022neural}. For example, in ANODEV2 \cite{zhang2019anodev2}, authors suggest a \textit{coupled} Neural ODE formulation
\begin{align}
    \dot{y}(t) &= f_\theta(t; y(t), \theta(t)) \,, \\
    \dot{\theta}(t) &= h_\theta(t; \theta(t))  \,.
\end{align}

Neural ODEs also exhibit a strong relationship with recurrent neural networks (RNNs) \cite{Goodfellow2016, scardapane2024alice}, as both architectures process informationdomand sequentially through a dynamic state evolution. 
In \cite{rubanova2019latent} it was introduced the concept of ODE-RNNs, a hybrid architecture where a Neural ODE governs the continuous-time evolution of the hidden state \textit{between} observation times, while a standard RNN updates the state \textit{at} each observation. 
This design enables the model to naturally handle time series with arbitrary and irregular sampling intervals, overcoming a key limitation of conventional RNNs. 
A complementary perspective is provided in \cite{niu2019recurrent}, who defined a general family of RNNs, termed ODERNNs, by relating the composition rules of recurrent architectures to integration methods of ODEs. 
There, it was shown that popular architectures, such as LSTM and URNN, correspond to specific orders of these ODE-inspired RNNs, establishing a formal correspondence between recurrent computation and numerical ODE integration.

\subsection{Comparison with Physics-Informed Neural Networks}

While both Physics-Informed Neural Networks (PINNs) and Neural ODEs leverage differential equations in their formulations, they differ fundamentally in their objectives, the role of physics, and the training methodology.

The two approaches address substantially different tasks.
We have seen that a Neural ODE learns an \emph{unknown} dynamics $\dot{y} = f_\theta(y(t),t)$ directly from observed state trajectories, treating the vector field as a black-box function to be discovered~\cite{chen2018neural}.
In contrast, a PINN solves a \emph{known} differential equation by representing its solution with a neural network $\hat{y}(t;\theta)$ and enforcing the governing physics as a soft constraint during training.
In this sense, Neural ODEs \textbf{learn the equation} from data, whereas PINNs \textbf{learn the solution} of a given equation.

The distinction is most visible in the loss function.
A PINN minimizes a composite loss that includes the residual of the known ODE/PDE at collocation points, together with any data-fitting terms (e.g., initial conditions, boundary conditions, or sparse measurements)~\cite{matzakos2026comparing}:
\begin{equation}
\mathcal{L}_\text{pinn} = \frac{1}{N_c}\sum_{i=1}^{N_c} \bigl\|\dot{\hat{y}}(t_i) - f\bigl(t_i,\hat{y}(t_i)\bigr)\bigr\|^2 \;+\; \text{data-fitting terms}.
\end{equation}
Here, the known function~$f$ encodes the physical law and is not learnable.
A Neural ODE, by contrast, learns the vector field $f_\theta$ itself by comparing the numerically integrated state against observations:
\begin{equation}
\mathcal{L}_\text{node} = \frac{1}{N}\sum_{i=1}^{N} \bigl\| y_i - \text{ODESolver}\bigl(f_\theta, y(t_0), t_i\bigr) \bigr\|^2.
\end{equation}
Physics is not embedded as a constraint; it is the object of discovery.

    \chapter{Neural Operators} \label{chap:5:NeuralOperators}

\section{Learning Operators, Part I: Deep Operator Networks}

\subsection{Why learning \textit{operators}?}

We have seen in \cref{subsec:1:3:2:UniversalApproxThm} that the conceptual basis underpinning function approximation with neural networks is the Universal Approximation Theorem, which ensures the representability of any unknown function. This theorem exists in several forms, including Cybenko's version \cite{Cybenko1989} and the Hornik, Stinchcombe, and White version \cite{hornik1989multilayer}.

From this, we learn that it is possible to devise an universal approximator function, capable of representing any wanted unknown function. In particular, when dealing with differential problems, we have seen that we can use a neural network $u_\theta(\xx)$ to approximate a function $u\sim u_\theta$ by recasting the differential problem into an optimisation problem. 

In this way, we can solve a single differential problem, using the map from the coordinate space $\RR^d \ni \xx$ to the target space of the function, $u(\xx) = \yy \in \RR^n$ (for scalar functions, $\RR$). A very simple case, as a representative, is the Poisson equation:
\[
    \Delta u(\xx) = f(\xx) \,, \quad \xx\in D \subset \RR^d \,.
\]
Having the source term $f$ fixed, is it possible to numerically find a solution with an approximating function $u_\theta$. 

Now, let us now notice that we can \textit{formally solve}, in an exact manner, this PDE, via the so-called Green Functional method \cite{salsa2016partial}; we recast the problem into a distributional problem
\[
    \Delta_\xx \mathcal{G}(\xx, \yy) = \delta(\xx - \yy) \,,
\]
where $\mathcal{G} : D \times D \to \RR$ is the \textit{Green's Function}, while $\delta(\xx, \yy)$ is the Dirac Delta, defined as a distribution as
\[
    \int_D \delta(\xx - \yy) g(\yy) \dd^d y = g(\xx) .
\]
Heuristically, it can be represented as a continuous generalisation of the Kronecker delta, 
\[
    \delta(\xx - \yy) = \begin{cases}
        0, & \xx \neq \yy \,, \\
        \infty, & \xx = \yy,
    \end{cases}
\]
such that
\[
    \int_D \delta(\yy) \dd^d y = 1 .
\]

\begin{figure}[t]
    \centering
    \includegraphics[width=0.80\linewidth]{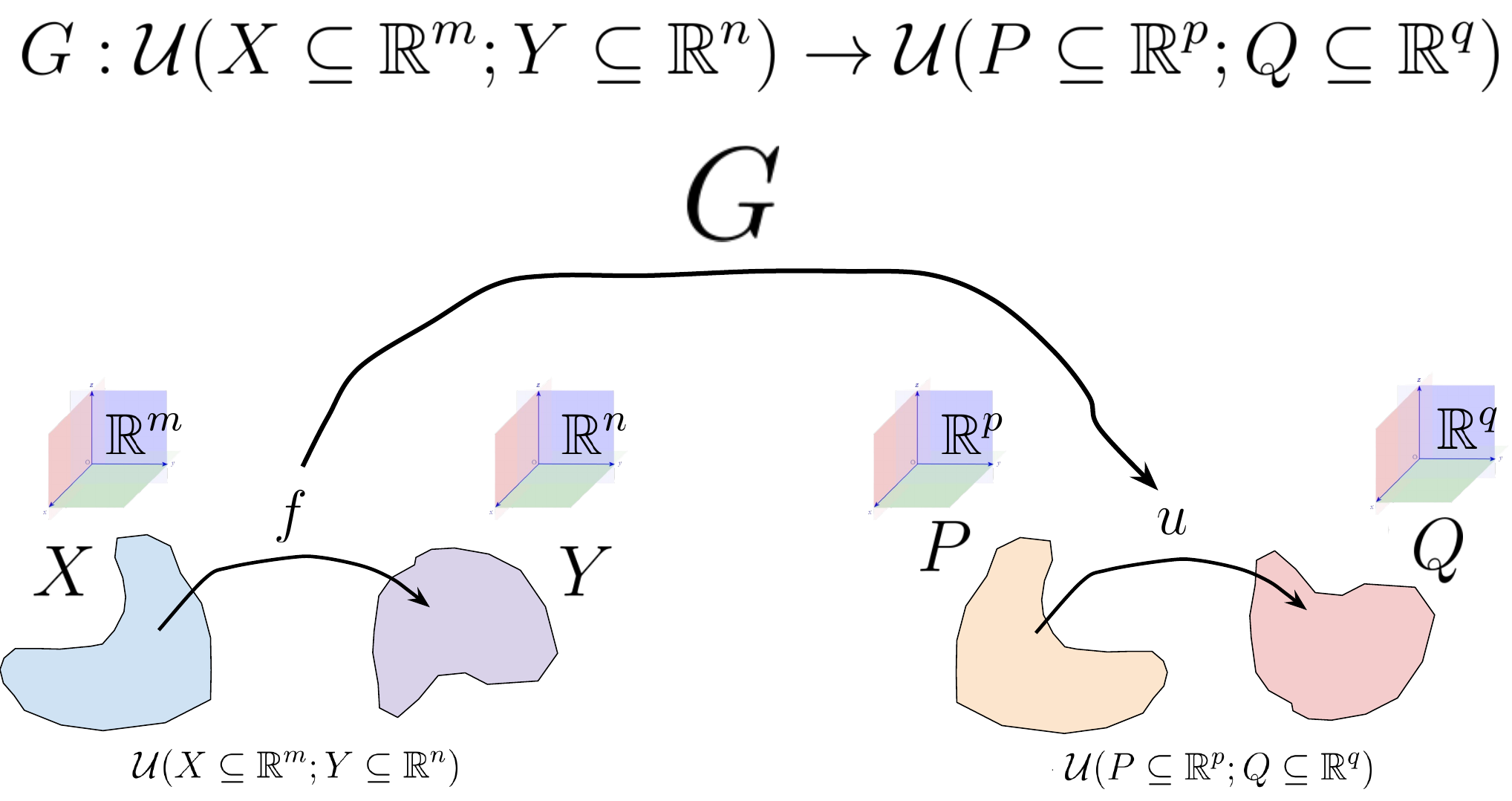}
    \caption{Visual representation of an operator.}
    \label{fig:5:1:OperViz}
\end{figure}

Let us, for a moment, suppose that the Green's function can be analytically computed. In that case, we can see that we can express the solution of the Poisson PDE $u$, as 
\[
u(\xx) = \int_D \mathcal{G}(\xx, \yy) f(\yy) \dd^d y .
\]
Indeed, applying $\Delta_\xx$, and (with the appropriate conditions satisfied), moving the derivation inside the integral sign, 
\begin{align*}
    \Delta_\xx u(\xx) &= \Delta_\xx \int_D \mathcal{G}(\xx, \yy) f(\yy) \dd^d y \\
    &= \int_D (\Delta_\xx \mathcal{G}(\xx, \yy)) f(\yy) \dd^d y \\
    &= \int_D \delta (\xx - \yy) f(\yy) \dd^d y \\ 
    &= f(\xx). \qquad \square
\end{align*}

Notice now that, denoting the (Banach) function space where $u$ lives as $\mathcal{U}\ni u$, and with $\mathcal{U}^* \ni f$ the space where the source function $f$ lives\note{In the following we will show that, indeed, $f$ lives in the dual of the space where $u$ lives, since we will tackle this kind of differential problems in a \textit{weak sense} (\cref{sec:1:3:func_analys}).}, we can write the Green's function as an operator mapping the two infinite-dimensional, functional spaces, i.e.
\begin{align*}
    G : \mathcal{U}^* \to \mathcal{U} \,, \qquad G : f \mapsto G[f] \defeq \int_D \mathcal{G}(\cdot, \yy) f(\yy) \dd^d y \,.
\end{align*}

What does it mean? It means that, through the Green's function, is possible to solve \textit{any} differential problem of the Poisson form, not only one specific realisation, simply changing $f$. 

The question now is: is it possible to \textit{learn} such operator? differently stated: is it possible to represent with an \textit{operator approximator}, dependent on a set of parameters $\theta$, which allows us to translate the differential problem \textit{family} to an \textit{optimisation} problem?

The answer is \textbf{yes}. Via \textit{Neural Operators}.

\subsection{The Universal Approximation Theorem for function\textit{als} (and operators)}

We have thus seen that underpinning the multi-layer perceptron for function approximation we have the Universal Approximation theorem for functions. The question now is: do we have something similar for function\textit{als} and for operators? the answer is yes \cite{chen1993approximations, chen1995universal}. In \cite{chen1993approximations, chen1995universal}, it was proven that there exists an universal approximation theorem for both functionals and operators. 
Using the original notation, let $X$ be a Banach space with norm $\| \cdot \|$, $K\subset X $ a compact subset, $C(K)$ the Banach space of all continuous functions on $K$, with the sup-norm. Furthermore, let us have

\begin{definition}[Tauber-Wiener functions]
    Let \(g:\RR\to\RR\) be a function. We call \(g\) a \textit{Tauber-Wiener} function if, for every compact interval \([a,b]\subset\RR\), the set of all finite linear combinations
    \[
        \left\{
        \sum_{i=1}^{N} c_i\,g(w_i x+b_i)
        \;:\;
        N\in\N,\;
        c_i,w_i,b_i\in\RR
        \right\}
    \]
    is dense in \(C([a,b])\) with respect to the uniform norm.
    The set of Tauber-Wiener functions is denoted by \((\TW)\).
\end{definition}


\begin{remark}
The Tauber-Wiener (TW) requirement on a scalar activation \(g:\mathbb{R}\to\mathbb{R}\) is a \emph{sufficient} non-degeneracy condition that guarantees single‑hidden‑layer networks of the form \(\sum_i c_i\,g(w_i\cdot x+b_i)\) are dense in spaces of continuous functions; in other words, TW ensures the classical universal approximation property. It is not a necessary condition (other, weaker hypotheses may also suffice), but it is convenient because it is simple to state and verify. Many commonly used activations (sigmoidal functions and a broad class of nonpolynomial activations; under mild technical hypotheses certain ReLU‑type and smooth activations) satisfy the TW/non-degeneracy requirements used in the operator‑approximation results of Chen and Chen. Thus, when we assume \(g\in(\mathrm{TW})\) we are adopting a standard sufficient hypothesis that covers most practical activation choices while keeping the proofs and statements compact. \cite{chen1993approximations,chen1995universal,hornik1989multilayer}
\end{remark}

\noindent Indeed, the authors of \cite{chen1993approximations, chen1995universal} thus proved 

\begin{theorem}[Universal Approximation Theorem for Functions]\label{thm:5:1:2:UnivThmFunctions}
    Let $K\subset \RR^m$ compact, $U\subset C(K)$ compact, $g\in \TW$, and let $f\in U$. Then,  $\forall \epsilon>0$, there exists a positive integer $N$, $\{b_i\}_{i=1,\dots,N} \in \RR, \{w_i\}_{i=1,\dots,N} \in \RR^m$, independent of $f$, and $\{c_i(f)\}_{i=1,\dots,N}\in\R$, depending on $f$, such that 
    \[
        \left| f(x) - \sum_{i=1}^N  c_i(f) g(w_i x + b_i)  \right| \le \epsilon    \, ,\qquad \forall x\in K. 
    \]
    Moreover, each $c_i(f)$ is a linear continuous functional defined on $U$.
\end{theorem}

\cref{thm:5:1:2:UnivThmFunctions} shows that  for a function (continuous or discontinuous) to be qualified as an activation function, a sufficient condition is that it belongs to Tauber-Wiener class.\note{Actually, while it gives the graso of the concept, this is a little misleading; Actually, \cref{thm:5:1:2:UnivThmFunctions} shows that if $g \in \TW$, then a single-hidden-layer network with activation $g$ can approximate any continuous function on a compact set. }
Furthermore, the equiuniform convergence property in \cref{thm:5:1:2:UnivThmFunctions} plays a crucial role in approximation to non-linear operators by neural networks (see the paper for more details).

But the two crucial results are the following

\begin{theorem}[Universal Approximation Theorem for Functionals]\label{thm:5:1:2:UnivThmFunctionals}
    Suppose that \(g\in(\mathrm{TW})\), \(X\) is a Banach space, \(K\subseteq X\) is a compact set, \(V\subset C(K)\) is a compact set, and \(f:V\to\mathbb{R}\) is a continuous \emph{functional}. Then for any \(\epsilon>0\) there exist a positive integer \(N\), an integer \(m\), points \(x_1,\dots,x_m\in K\), and real constants \(c_i,b_i,\xi_{ij}\) for \(i=1,\dots,N\) and \(j=1,\dots,m\), such that
    \[
    \left| f(u) - \sum_{i=1}^{N} c_i\, g\!\left( \sum_{j=1}^{m} \xi_{ij}\, u(x_j) + b_i \right) \right| < \epsilon 
    \]
    holds for all \(u\in V\).
\end{theorem}

\begin{theorem}[Universal Approximation Theorem for Operators]\label{thm:5:1:2:UnivThmOps}
    Suppose that \(g\in(\mathrm{TW})\), \(X\) is a Banach space, \(K_1\subseteq X\) and \(K_2\subseteq\mathbb{R}^n\) are compact sets, \(V\subset C(K_1)\) is a compact set, and \(G:V\subset C(K_1)\to C(K_2)\) is a nonlinear continuous \emph{operator}. Then for any \(\epsilon>0\) there exist positive integers \(M,N,m\), real constants \(c_i^k,\zeta_k,\xi_{ij}^k\), points \(\omega_k\in\mathbb{R}^n\) and \(x_j\in K_1\) for \(i=1,\dots,M\), \(k=1,\dots,N\), \(j=1,\dots,m\), such that
    \begin{equation}\label{eq:5:1:2:UnivThmOpsEq}
        \left| G(u)(y) - \sum_{k=1}^{N}\sum_{i=1}^{M} c_i^k\, g\!\left( \sum_{j=1}^{m} \xi_{ij}^k\, u(x_j) + b_i^k \right)\, g(\omega_k\cdot y + \zeta_k) \right| < \epsilon 
    \end{equation}
    holds for all \(u\in V\) and all \(y\in K_2\).
\end{theorem}

\begin{figure}[t]
    \centering
    \includegraphics[width=0.65\linewidth]{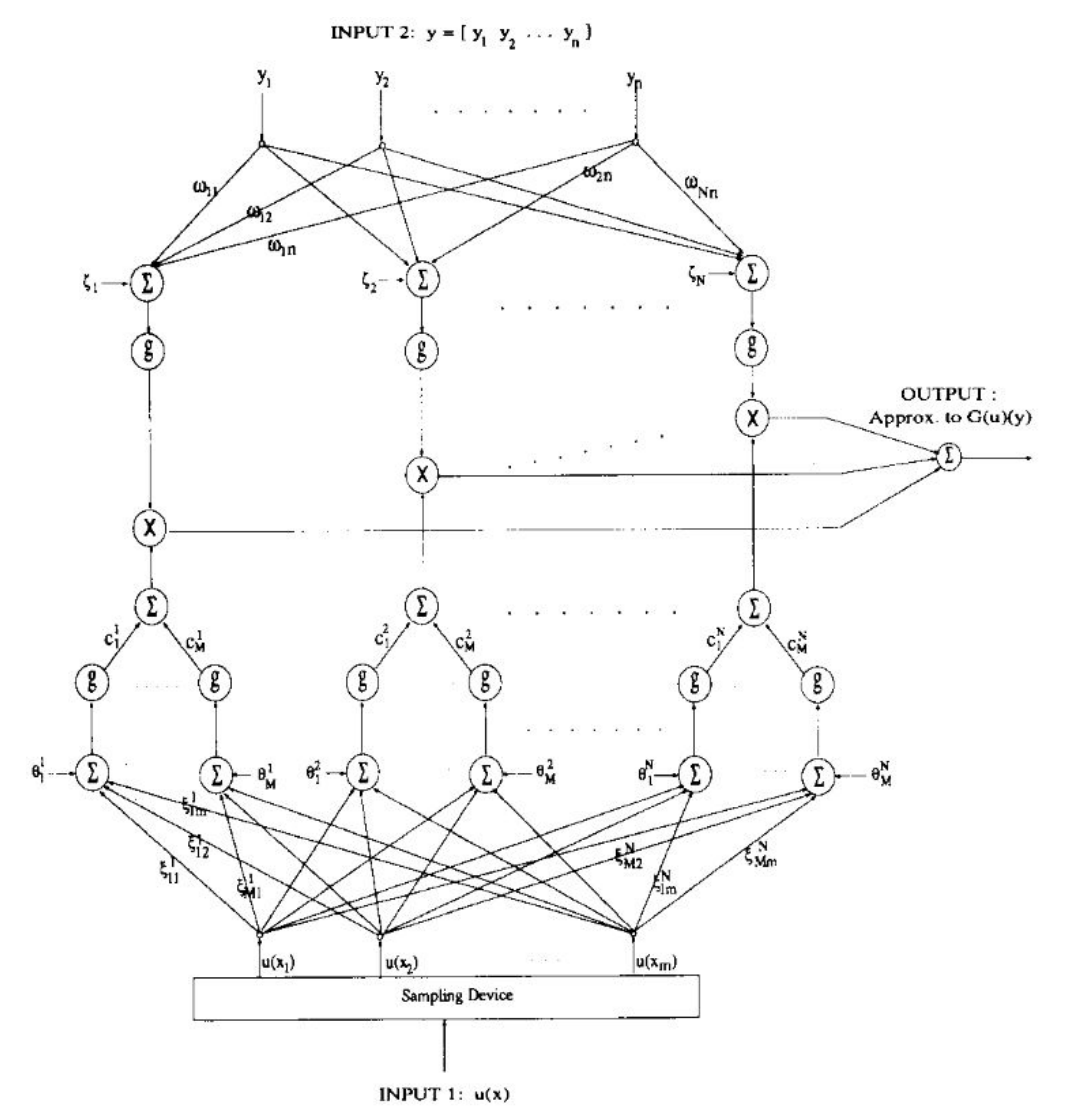}
    \caption{A neural network approximation to nonlinear operator $G(u)(y)$ based on theorem 5 of \cite{chen1995universal}. Original Figure 1 of the same paper.}
    \label{fig:5:1:2:LearningOpOrigImg}
\end{figure}

\begin{remark}
    \cref{thm:5:1:2:UnivThmOps} is existential: it guarantees approximation power but not a constructive procedure for selecting the number of terms or sensor locations; 
\end{remark}

\begin{remark}
    \cref{thm:5:1:2:UnivThmOps} guarantees uniform approximation on compact sets (equicontinuity/equiuniform convergence). It’s worth noticing  that the approximation is uniform in both the input function (over the compact set $V\subset C(K_1)$ and the output function variable $y\in K_2$).
\end{remark}

Let us now look more deeply at \cref{eq:5:1:2:UnivThmOpsEq}; we can rewrite the approximator
\begin{equation}\label{eq:5:1:2:DeepONet1}
    \begin{split}
        G_\theta [u](y)&\defeq \sum_{k=1}^{N} 
    \underbrace{\textcolor{darkgreen}{
        \sum_{i=1}^{M} c_i^k\, g\!\left( \sum_{j=1}^{m} \xi_{ij}^k\, u(x_j) + b_i^k \right)
    } }_{\text{branch}}
    \ \  
    \underbrace{\textcolor{blue}{
        g(\omega_k\cdot y + \zeta_k)
    } }_{\text{trunk}} \\ 
    &\defeq \sum_{k=1}^N \textcolor{darkgreen}{
         \sum_{i=1}^{M} c_i^k\, g\!\left( \sum_{j=1}^{m} \text{branchnet}_{ij}^k(u(x_j)) \right)
        } \cdot \textcolor{blue}{
            \text{trunknet} (\omega_k, \zeta_k  ; y)\,, 
    }
    \end{split}
\end{equation}
in terms of two blocks: a so-called \textit{branch}, representing the action of the operator on the functional, and the \textit{trunk},  representing the embedding of the output function's input. The original pictorial representation of this \textit{Operator Network} is reported in \cref{fig:5:1:2:LearningOpOrigImg}. 

\begin{figure}[t]
    \centering
    \includegraphics[width=0.95\linewidth]{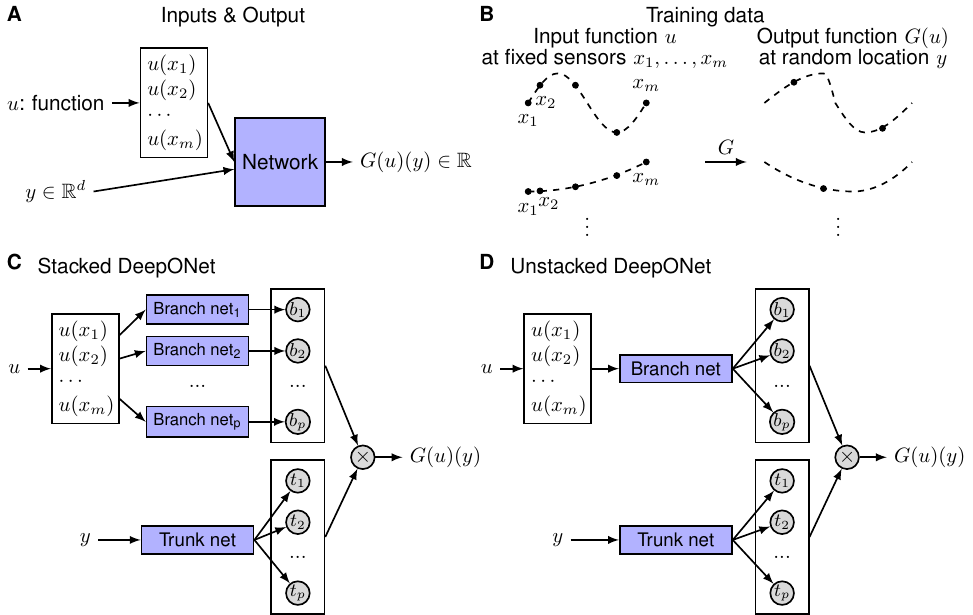}
    \caption{
    \textbf{Illustrations of the Operator Learning setup and architectures of DeepONets.} (\textbf{A}) The network to learn an operator $G: u \mapsto G(u)$ takes two inputs $[u(x_1), u(x_2), \dots, u(x_m)]$ and $y$. (\textbf{B}) Illustration of the training data. For each input function $u$, we require that we have the same number of evaluations at the same scattered sensors $x_1, x_2, \dots, x_m$. However, we do not enforce any constraints on the number or locations for the evaluation of output functions. (\textbf{C}) The stacked DeepONet in \cref{thm:5:1:2:UnivThmOps} has one trunk network and $p$ stacked branch networks. (\textbf{D}) The unstacked DeepONet has one trunk network and one branch network. From \cite{lu2021learning}.
    }
    \label{fig:5:1:2:deeponet}
\end{figure}


To clarify further, we borrow the \textit{physical intuition} from
\cite{lu2021learning, wang2021learning}. Suppose that we want to learn a continuous operator
\[
    G:V\to C(K_2),
\]
where \(V\subset C(K_1)\) is a compact family of input functions and
\(K_1\subset\RR^{d_x}\), \(K_2\subset\RR^{d_y}\) are compact domains.

Suppose that we observe each input function \(u\in V\) through a fixed set of
``sensors''
\[
    x_1,\dots,x_m\in K_1,
\]
which provide the evaluations
\[
    u(x_1),\dots,u(x_m).
\]
For example, \(u\) may represent a temperature field and the \(x_i\)'s the
locations of thermometers, while \(G[u](y)\) represents the corresponding
output field evaluated at a coordinate \(y\in K_2\); see
\cref{fig:5:1:2:deeponet}(B)\note{Notice that the assumption \(V\subset C(K_1)\) makes these pointwise sensor evaluations well-defined. More general input spaces, such as \(L^p(K_1)\), require an appropriate definition of the observation functionals rather than unqualified point evaluations.}.

\subsection{Deep Operator Networks}

We have now at our disposal the needed tools: an universal approximation theorem, furnishing solid grounds, and a way to formulate  the approximator model in terms of \textit{learnable parameters}. Indeed, starting from here, \cite{lu2021learning} introduced \textit{Deep Operator Networks} (DeepONets), an approach to learning operators. To design and train DeepONets, in this general setting, the network inputs consist of two separate components:  $[u(x_1), u(x_2), \dots, u(x_m)]^T$ and $y$ (see \cref{fig:5:1:2:deeponet}). 
Now, we can use a multi-layer fully connected networks (as an MLP) to model both the ``trunk'' network, which takes $y$ as the input and outputs $[t_1, t_2, \dots, t_p]^T \in \mathbb{R}^p$, and the ``branch'' network. Formally, we should have $N$ independent branch-nets (see \cref{eq:5:1:2:UnivThmOpsEq} and especially \cref{eq:5:1:2:DeepONet1}). This is clearly difficult to implement in practice (we should train $m+1$ networks, where $m$ is the number of ``sensors''); to overcome this fact, in practice one merges the branch-nets in what is called ``Unstacked DeepONet''
\begin{equation}\label{eq:5:1:2:DeepONet2Unstacked}
\begin{split}
    G_\theta [u](y) &= \sum_{k=1}^N \textcolor{darkgreen}{
        \left[\text{branchnet}(u) \right]_k
        } \;  \textcolor{blue}{
            \text{trunknet} (\omega_k, \zeta_k  ; y)
        } + b_0 \\ 
        &= \langle \textcolor{darkgreen}{ 
            \text{branchnet}(u(x_1), \ldots , u(x_N))
        } \, , \, \textcolor{blue}{
            \text{trunknet}(y) 
        } \rangle_{\RR^N} + b_0 \,, 
\end{split} 
\end{equation}
The scalar bias \(b_0\) is not required by the operator approximation theorem; it is an optional architectural addition introduced in DeepONet and found empirically to improve performance in the experiments of \cite{lu2021learning}. 

The DeepONet version most faithful to \cref{thm:5:1:2:UnivThmOps} and to \cref{eq:5:1:2:DeepONet1}, is called ``Stacked DeepONet''. In \cref{fig:5:1:2:LearningOpOrigImg} it is reported a pictorial representation of both Stacked (\cref{fig:5:1:2:deeponet}(C)) and Unstacked DeepONets (\cref{fig:5:1:2:deeponet}(D)).

Notice that, in the context of Unstacked DeepONets, the $\textcolor{darkgreen}{\text{branchnet}(u) } $ is a single fully connected network  $\text{branchnet} : \RR^m \to \RR^N$, mapping $  [u(x_1), \ldots , u(x_m)] \mapsto \text{branchnet}(u) \sim (N)$\note{Here we use the notation of \cite{scardapane2024alice}, where  $X \sim (n,k,l)$ means that, as an array, it has a shape of $[n,k,l]$.}.  

To train DeepONets, we simply start from a training dataset $\mathcal{D}= \{X; Y\} = \{  \mathbf{u}_n, \mathbf{y}_n; G( \mathbf{u}_n) (\mathbf{y}_n) \rdefeq \mathbf{o}_n \}_{n=1}^{N_{data}}$, where $n$ represents the data index, $\mathbf{u}_n,  \in \RR^m, \mathbf{y}_n \in K_2 \subseteq \RR^{d_y}$, $\mathbf{u}_n \defeq [u_n(x_1), \ldots u_n(x_m)]$\note{Notice that we \textit{don't} need to know $[x_1, \ldots, x_m]$ explicitly.}, and $G( \mathbf{u}_n) (\mathbf{y}_n) \rdefeq \mathbf{o}_n $ is the target value of the unknown function $G( \mathbf{u}_n)$ (mapped by $G$ from $\mathbf{u}_n$) evaluated on $\mathbf{y}_n$. 
Notice that this dataset consists of triples $(\mathbf{u}_n, \mathbf{y}_n, \mathbf{o}_n)$, where $\mathbf{u}_n$ are sensor evaluations of the input function $u_n$, and $\mathbf{o}_n$ are target evaluations of the output function $G(u_n)$. Indeed, $G$ maps functions to functions (being an operator) $u \mapsto G(u)$, but, for each simulation $n$, we have access to $u$ only as a set of its evaluation on certain points, $\mathbf{u}_n \defeq [u_n(x_1), \ldots u_n(x_m)]$, and we have access to $G(u)$ again on some of its evaluation points, $G( \mathbf{u}_n) (\mathbf{y}_n)$. Nevertheless, once trained, the DeepONets furnish a \textit{continuous} (approximated) operator mapping for each (evaluated) $\mathbf{u}_p \notin \mathcal{D}$, in the form of 
\[
   G(\mathbf{u}_p)(\cdot) : \RR^m \to C(K_2).
\]

A crucial question remains open: \textit{how many sensor points do we need to achieve a certain accuracy}? The authors replied also to this question, at least for ODE systems of the form 
\begin{align*}
    \frac{\dd \mathbf{s}(x)}{\dd x} &=\mathbf{g}(\mathbf{s}(x),u(x),x) \\
    \mathbf{s}(a) &=\mathbf{s_0} \,.
\end{align*}
For systems of this kind, we can formally integrate the ODEs as
\[
    (Gu)(x)=\mathbf{s_0}+\int_a^x\mathbf{g}((Gu)(t),u(t),t) \dd t.
\]
where $u\in V$ (a compact subset of $C[a,b]$) is the input signal, and $\mathbf{s}:[a,b]\rightarrow\mathbb{R}^{K}$ is the solution of system, serving as the output signal.

Now, we choose uniformly $m+1$ points $x_{j}=a+j(b-a)/m,j=0,1,\cdots,m$ from $[a,b]$, and define the function $u_{m}(x)$ as follows:
\begin{equation*}
    u_m(x)=u(x_j)+\frac{u(x_{j+1})-u(x_j)}{x_{j+1}-x_j} (x-x_j),\ x_j\leq x\leq x_{j+1},\ j=0,1,\cdots,m-1.
\end{equation*}

Denote the operator mapping $u$ to $u_{m}$ by $\mathcal{L}_{m}$, and let $U_{m}=\{\mathcal{L}_{m}(u)|u\in V\}$, which is obviously a compact subset of $C[a,b]$ since $V$ is compact and continuous operator $\mathcal{L}_{m}$  keeps the compactness. Naturally, $W_{m}\coloneqq V\cup U_{m}$ as the union of two compact sets is also compact. 
To ensure that the infinite union $W\coloneqq\bigcup_{m=1}^{\infty}W_{m}$ remains compact, we rely on the specific structure of the interpolation operators $\mathcal{L}_{m}$. In the context of piecewise linear interpolation on uniformly refined grids, the sequence $\{W_m\}$ satisfies a \textit{Cauchy condition} in the Hausdorff metric: the approximation error $\kappa(m,V) = \sup_{u\in V} \|u - \mathcal{L}_m u\|_\infty$ decays to zero as $m \to \infty$. Consequently, the closure $\overline{W}$ is compact, and $W$ is pre-compact (totally bounded) in $C[a,b]$\note{Strictly speaking, the infinite union of compact sets is not always compact. However, in this specific setting, the uniform convergence of the interpolation operators $\mathcal{L}_m \to \text{id}$ on the compact set $V$ ensures that $W$ is totally bounded. By the Ascoli-Arzelà theorem, its closure is compact. See \cite{lu2021learning} Appendix A for a detailed proof.}.
Since $G$ is a continuous operator,  $G(W)$ is compact in $C([a,b];\mathbb{R}^{K})$.
The subsequent discussions are mainly within $W$ and $G(W)$. For convenience of analysis, we assume that $\mathbf{g}(\mathbf{s},u,x)$ satisfies the Lipschitz condition with respect to $\mathbf{s}$ and $u$ on $G(W)\times W$, i.e., there is a constant $c>0$ such that
\begin{equation*}
    \begin{array}{l}
    \|\mathbf{g}(\mathbf{s_1},u,x)-\mathbf{g}(\mathbf{s_2},u,x)\|_{2}\leq c\|\mathbf{s_1}-\mathbf{s_2}\|_{2}\\
    \|\mathbf{g}(\mathbf{s},u_1,x)-\mathbf{g}(\mathbf{s},u_2,x)\|_{2}\leq c|u_1-u_2|
    \end{array}.
\end{equation*}
Note that this condition is easy to achieve, for instance, as long as $\mathbf{g}$ is differentiable with respect to $\mathbf{s}$ and $u$ on $G(W)\times W$.

For $u\in V,u_{m}\in U_{m}$, there exists a constant $\kappa(m,V)$ depending on $m$ and compact space $V$, such that
\begin{equation}
    \max_{x\in[a,b]}|u(x)-u_m(x)|\leq\kappa(m,V),\quad \kappa(m,V)\rightarrow 0\ \ as\ \ m\rightarrow\infty.
\end{equation}

\begin{remark}
    \textit{Discretization versus approximation error:} The total error of a learned DeepONet decomposes into the interpolation (discretization) error \(\|u-u_m\|_\infty\) (due to finite sensors), and the neural approximation error of the branch/trunk parametrisation; both terms must be balanced when choosing \(m\) and the network sizes.
\end{remark}

Based on these concepts, the authors of \cite{lu2021learning} proved the following

\begin{theorem}
Suppose that $m$ is a positive integer making $c(b-a)\kappa(m,V)e^{c(b-a)} < \varepsilon$, then for any $d\in [a,b]$, there exist $\mathcal{W}_{1}\in \mathbb{R}^{n\times (m+1)},b_{1}\in \mathbb{R}^{n},\mathcal{W}_{2}\in \mathbb{R}^{K\times n},b_{2}\in \mathbb{R}^{K}$, such that
\begin{equation*}
    \|(Gu)(d)-(\mathcal{W}_{2}\cdot\sigma(\mathcal{W}_{1}\cdot [u(x_{0})\quad\cdots\quad u(x_{m})]^{T}+b_{1})+b_{2})\|_{2}<\varepsilon
\end{equation*} 
holds for all $u\in V$.
\end{theorem}

\subsubsection{A prototypical application: learning $\partial^{-1}$}\label{subsubsec:5.1.3.1:antider}

Let us start (following also the original paper \cite{lu2021learning}) by trying to learn the anti-derivative operator; i.e., suppose we have a (huge) dataset for 
\begin{align}
    \frac{\dd s(x)}{\dd x} &= u(x) \,, \qquad x \in [0,1] \,, \\
    s(0) &= 0 \,.
\end{align}

\noindent The formal solution of this problem is the anti-derivative operator 
\[
    G : u(x) \mapsto s(x) = \int_0^x u(t) \dd t . 
\]

To generate the dataset, it is simple to select random analytical symbolic functions $s(x)$, and computes their derivative $s' = u$, and evaluate it on certain domain point $x\in[0,1]$.

Actually, in their experiments, \cite{lu2021learning}, they did the opposite: picked $u$ and integrated back to get $s$. In particular, for the sensors, they used uniformly distributed points 
\[
x_j = a + j \, \frac{b-a}{m} \,, \quad j=0,\ldots, m, 
\]
where $a=0, b=1$\note{The above equation is for generating uniformly distributed points in $[a,b]$.}; 
for the functions, they mainly consider two function spaces: Gaussian random field (GRF) and orthogonal (Chebyshev) polynomials. For  the mean-zero GRF
\[
u \sim \mathcal{G}(0, k_l(x_1, x_2)),
\]
where the covariance kernel $k_l(x_1, x_2) = \exp(-\| x_1 -x_2 \|^2 / 2l^2)$ is the radial-basis function (RBF) kernel with a length-scale parameter $l > 0$. The length-scale $l$ determines the smoothness of the sampled function, and larger $l$ leads to smoother $u$. For the Chebyshev polynomials (of the first kind), they define the orthogonal polynomials of degree $N$ as: 
\[
V_{\text{poly}}=\left\{ \sum_{i=0}^{N-1} a_i T_i(x): |a_i| \leq M \right\}.
\]
They generate the dataset from $V_{\text{poly}}$ by randomly sampling $a_i$ from $[-M,M]$ to get a sample of $u$.

Thus, after sampling $u$ from the chosen function spaces, they numerically solve the ODE systems by  Runge-Kutta (4, 5) and PDEs by a second-order finite difference method to obtain the reference solutions. 

\begin{remark}
    Note that one data point is a triplet $(u, y, G(u)(y))$, and thus one specific input $u$ may appear in multiple data points with different values of $y$. For example, a dataset of size $10\,000$ may only be generated from 100 $u$ trajectories, and each evaluates $G(u)(y)$ for 100 $y$ locations.
\end{remark}

\subsection{Physics-Informed DeepONets}

\begin{figure}[t]
    \centering
    \includegraphics[width=0.85\linewidth]{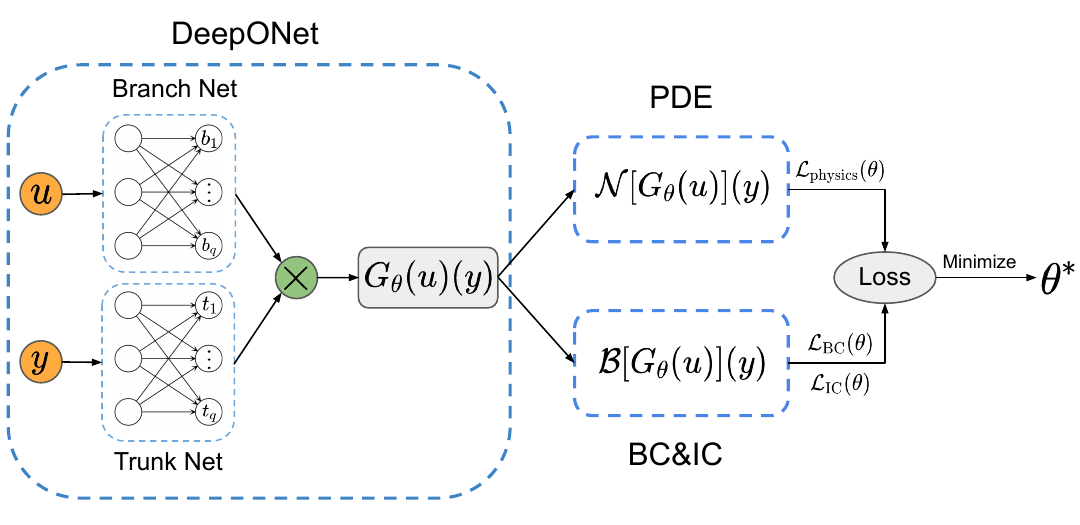}
    \caption{
    {\em Making DeepONets physics-informed:}  We have seen that the DeepONet architecture \cite{lu2021learning} consists of two sub-networks: the \textit{branch}, to extract latent  representations of input functions; and the \textit{trunk}, to extract latent representations of input coordinates at which the output functions are evaluated. A continuous and differentiable representation of the output functions is then obtained by merging the latent representations extracted by each sub-network via a dot product. To make it physics-informed, automatic differentiation can then be employed to formulate appropriate regularization mechanisms for biasing the DeepONet outputs to satisfy a given system of PDEs. There, the differential operator defining the PDE, $\mathcal{F}$, is denoted as $\mathcal{N}$. 
    From \cite{wang2021learning}.}
    \label{fig:5:1:4:physics_informed_DeepONet}
\end{figure}

The plain, vanilla DeepONet seen in the previous section is purely \textit{data informed}. The next step is to make it (also) \textit{physics informed}. This step was done by \cite{wang2021learning}. This is done, as usual, by leveraging automatic differentiation to impose the underlying physical laws via soft penalty constraints during model training. 

To move towards this direction, let us reformulate the problem, in a way which will be useful also in the remainder of the chapter. Let $(\mathcal{U}, \mathcal{V}, \mathcal{S})$ be a triplet of Banach spaces and $\mathcal{F}: \mathcal{U} \times \mathcal{S} \rightarrow \mathcal{V}, \; \mathcal{B} : \mathcal{S} \rightarrow \mathcal{V}$ be  (linear or non-linear) differential operators. We consider a differential problem
\begin{align}
    \mathcal{F}(u, {s}) &= 0,\\
    \mathcal{B}(s)&=0
\end{align}
where ${u} \in \mathcal{U}$ denotes the  input functions, and ${s} \in \mathcal{S}$ is the corresponding unknown solution of the PDE system. In particular, we assume that, for any ${u} \in \mathcal{U}$, there exists an unique solution ${s} = {s}[{u}] \in \mathcal{U}$ (subject to appropriate initial and boundary conditions). Then, we can define the solution operator $G: \mathcal{U} \rightarrow \mathcal{S}$ as
\begin{align}
    G({u}) = {s}({u}).
\end{align}
We then approximate that $G$ with $G_\theta : \mathcal{U}\times \Theta \rightarrow \mathcal{S}$, where $\theta \in \Theta$ are the trainable parameters. Usually, $G_\theta$ is an unstacked DeepONet, 
\begin{align}
    G_\theta [u](y) &= \sum_{k=1}^N \textcolor{darkgreen}{
        b_k ( u(x_1), \ldots , u(x_m)  )
        } \;  \textcolor{blue}{
            t_k ( y)
        } + b_0 .
\end{align}

As already noticed, the outputs of a DeepONet model are continuously differentiable with respect to their input coordinates. Therefore, one may use automatic differentiation to formulate an appropriate regularization mechanism for biasing the target output functions to satisfy any given differential constraints. Thus, we can build \textit{physics-informed DeepONets} by formulating the following loss function
\begin{align}
    \mathcal{L}( {\theta}) = \mathcal{L}_{\text{operator}}( {\theta}) +  \mathcal{L}_{\text{physics}}( {\theta}),
\end{align}
where 
\begin{align}
    \mathcal{L}_{\text{operator}} ( {\theta}) &= \frac{1}{NP} \sum_{i=1}^N \sum_{j=1}^P  \left|G_{ {\theta}} \left[ {u}^{(i)}\right]  \left(y^{(i)}_{uj}  \right)- s^{(i)} \left(y^{(i)}_{uj} \right)  \right|^2, \\
    \mathcal{L}_{\text{physics}}( {\theta}) &= \frac{1}{NQm} \sum_{i=1}^N \sum_{j=1}^Q \sum_{k=1}^m \left|\mathcal{F}\left(  u^{(i)}\left( {x}_k\right), G_{ {\theta}}\left[ {u}^{(i)} \right]  \left( {y}^{(i)}_{rj}  \right) \right) \right|^2.
\end{align}
Here, $\{u^{(i}\}_{i=1,\dots,N}$ denotes $N$ distinct input functions, sampled from $\mathcal{U}$. For each $u\in\mathcal{U}$, $\{{y}^{(i)}_{uj}\}_{j=1,\dots,P}$ are $P$ locations where the true field is evaluated. Furthermore, $\{{y}^{(i)}_{rj}\}_{j=1,\dots,Q}$   are $Q$ random collocation points, sampled in the domain of $G[u^{(i)}]$, and may (or may not) be different from $\{{y}^{(i)}_{uj}\}_{j=1,\dots,P}$. Furthermore, they \textit{may vary} for different $u^{(i)}$ (thus the subscript $u$). 

In this way, $\mathcal{L}_{\text{operator}} ( {\theta}) $ drives, during training, the approximator towards the available solutions measurements, while $\mathcal{L}_{\text{physics}} ( {\theta}) $ enforces the underlying PDE constraints. We may want also to add the $\mathcal{L}_{\text{boundary}} (\theta)$ loss, as
\begin{equation}
    \mathcal{L}_{\text{boundary}} (\theta) = \frac{1}{N Q} \sum_{i=1}^{N} \sum_{j=1}^B \left|\mathcal{B}\left(    G_{ {\theta}}\left[ {u}^{(i)} \right]  \left( {y}^{(i)}_{bj}  \right) \right) \right|^2 ,
\end{equation}
where $\{{y}^{(i)}_{bj}\}_{j=1,\dots,B}$ are $B$ random collocation points, sampled in the boundary of the domain of $G[u^{(i)}]$.

\subsubsection{learning $\partial^{-1}$, the \textit{physics-informed} way}
Let us go back to the problem discussed in \cref{subsubsec:5.1.3.1:antider}, i.e., let us discuss how to learn the anti-derivative operator. We restart from  
\begin{align*}
    \frac{\dd s(x)}{\dd x} &= u(x) \,, \qquad x \in [0,1] \,, \\
    s(0) &= 0 \,.
\end{align*}
Now, the aim is to learn the solution operator $G_\theta : u \mapsto s$ \textbf{without any paired input-output} data. 
What we use, instead, to have a \textit{pure-forward} physics informed DeepONet formulation, is that we \textit{sample} the Banach space where $u$ lives (something like $x\in \Omega \subseteq \RR^d,  u\in X(\Omega)$), obtaining discrete evaluations $\{(x_i, u_i \defeq u(x_i))\}_{i=1, \ldots, N} \in (\Omega, X(\Omega))$. 
Thus, after modelling the solution operator $G_\theta$ with a fully connected neural network (for both branch and trunk net, in an unstacked manner), we recast the general differential problem to an optimisation one with loss
\begin{align}
    \mathcal{L}( {\theta}) &= \mathcal{L}_{\text{boundary}}( {\theta}) +  \mathcal{L}_{\text{physics}}( {\theta}) \\
    &= \frac{1}{N} \sum_{i=1}^N  \left|G_{ {\theta}} [ {u}^{(i)} ](0) \right|^2  + \frac{1}{NQ} \sum_{i=1}^N \sum_{j=1}^Q \left| \frac{\dd G_{ {\theta}}[ {u}^{(i)}] (y)}{\dd y}\Big|_{y = x_j} - u^{(i)}(x_j)   \right|^2. 
\end{align}

In the specific experiment of \cite{wang2021learning}, the input functions are sampled from a Gaussian random-field prior and the residual collocation points are chosen to coincide with the branch sensors, so that \(Q=m\) and \(y_j=x_j\).

\section{Using a different representation theorem for functions: Kolmogorov-Arnold Network}

We have seen that our approach to solve various approximating problems is grounded into \textit{universal approximation theorems}. One major result, due to Kolmogorov and Arnold (\cite{kolmogorov1957representations, arnold2009collected, arnol1960representation}; see \cite{braun2009constructive, schmidt2021kolmogorov} for a modern perspective), directly related to Hilbert 13th problem \cite{morris2021hilbert}. The Hilbert 13th problem is to prove the following conjecture 

\quotespace
\begin{center}
    \begin{quote}
    \say{\textit{A solution of the general equation of degree 7 cannot be represented as a superposition of continuous functions of two variables.}}
\end{quote}
\end{center}
\quotespace

\noindent Working on that, Kolmogorov and Arnold proved that \textit{any} multivariate function (i.e., any function whose domain $\Omega \subseteq\RR^m, m>1$), continuous on $[0,1]^n$, can be written as a \textit{finite} composition of \textit{univariate} functions (i.e., function whose domain is $I\subseteq\RR)$. The most common version of the Kolmogorov-Arnold Representation Theorem (sometimes abbreviated as KAT, or KART) is the following \cite{ismailov2025addressing} 

\begin{theorem}[(original) Kolmogorov-Arnold Representation Theorem]
    Let $I^n = [0,1]^n \subset \RR^n, n\ge2$ be the unit cube. Then, there exists $n(2n+1)$ universal, continuous, one-variable functions $\phi_{q,p} : [0,1] \to \RR$, with $q=1,\dots,2n+1, p=1,\dots,n$, such that any continuous function $f\in C(I^n)$ admits the precise representation
    \begin{equation}
        f(x_1,\dots,x_n) = \sum_{q=1}^{2n+1} \Phi_q \left( \sum_{p=1}^n \phi_{q,p}  (x_p)  \right) \,, \qquad \forall (x_1, \dots , x_n) \in I^n ,
    \end{equation}    
    where $\Phi_q : \RR \to \RR$ are $2n+1$ continuous one-variable functions depending on $f$.
\end{theorem}

Notice that the second activation $\Phi_q$ is $f-$dependent, as it happened in \cref{thm:5:1:2:UnivThmFunctions} for $c_i(f)$. 

Notice also that the original formulation above is about \textit{continuous} function on the unit cube in $\RR^n$. Fortunately, it is possible to extend the theorem to approximate  (i) a non-continuous function; and (ii) a function with generic bounded domain \cite{ismailov2023three, ismayilova2024kolmogorov, ismailov2025addressing}. Furthermore, more recent work have shown that the outer functions $\Phi_q$ can be replaced by a single $\Phi$. In particular \cite{ismailov2025addressing}

\begin{theorem}[(extended) Kolmogorov-Arnold Representation Theorem]
    Let $D \subset \RR^n, n\ge2$ be a bounded domain. Then, there exists $n(2n+1)$ universal, continuous, one-variable functions $\phi_{q,p} : \RR \to \RR$, with $q=1,\dots,2n+1, p=1,\dots,n$, such that any  function $f$ (not necessarily continuous) admits the precise representation
    \begin{equation}\label{eq:5:2:KARTeq}
        f(x_1,\dots,x_n) = \sum_{q=1}^{2n+1} \Phi_q \left( \sum_{p=1}^n \phi_{q,p}  (x_p)  \right) \,, \qquad \forall (x_1, \dots , x_n) \in D ,
    \end{equation}    
    where $\Phi : \RR \to \RR$ is a one-variable function depending on $f$. If $f$ is continuous, then $\Phi$ can be chosen to be continuous; if $f$ is discontinuous and bounded, $\Phi$ is also discontinuous and bounded; if $f$ is unbounded, $\Phi$ is also unbounded. 
\end{theorem}

Furthermore, it has been shown that $\phi_{q,p}$ can be replaced by a family of functions of simpler structure, e.g. \cite{sprecher1965structure}
\[
    \phi_{q,p} (x_p)  \to \lambda_p \phi(x_p +aq) , \quad \lambda_p, a \in \RR.
\]

There have been multiple attempts to use this representation theorems for deep learning purposes, some of which dates back to early 90's and 2000's \cite{lin1993realization, koppen2002training}, up to modern times \cite{ismayilova2024kolmogorov}. There have been also criticism on the usefulness of KART for this task \cite{girosi1989representation, poggio2020theoretical}. Nevertheless, \cite{liu2024kan} popularised this approach by blending Kolmogorov-Arnold Representation Theorem, symbolic regression, and explainability, giving rise to KAN: Kolmogorov-Arnold Network.

\subsection{Kolmogorov-Arnold Network}

\begin{figure}[t]
    \centering
    \includegraphics[width=0.95\linewidth]{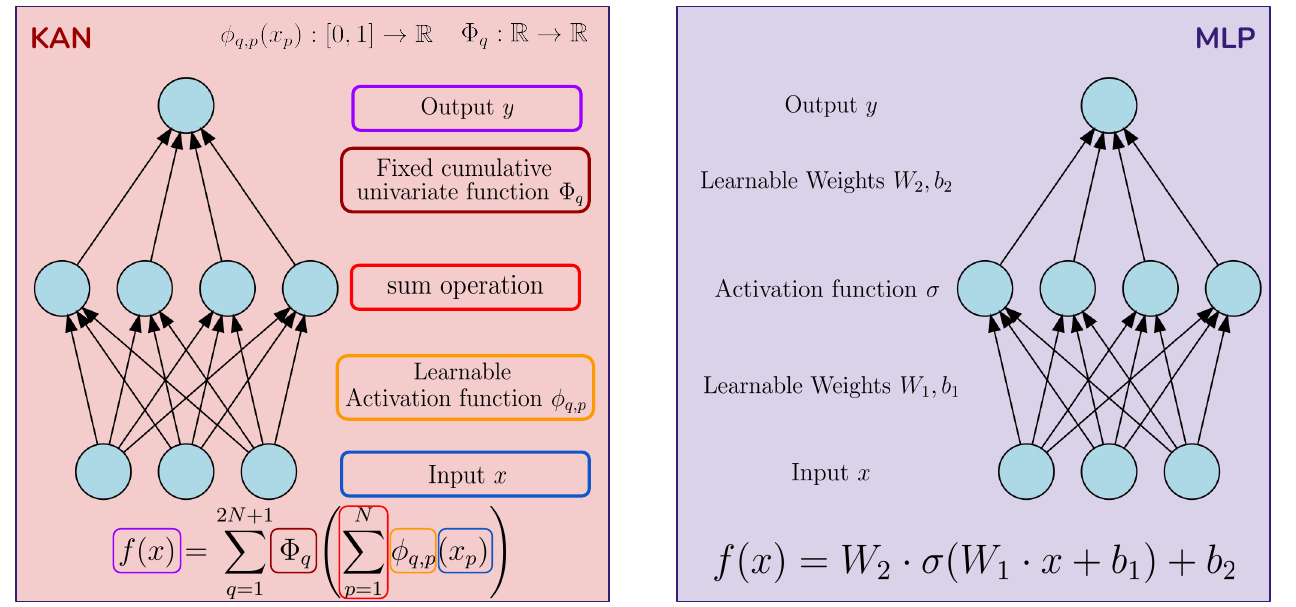}
    \caption{Visual comparison between single-hidden layer Kolmogorov-Arnold Networks and Multi-Layer Perceptrons. }
    \label{fig:5:2:1:KANvsMLP}
\end{figure}

In  \cite{liu2024kan}, the authors introduced the Kolmogorov-Arnold Network (KAN) \cref{fig:5:2:1:KANvsMLP}. Its shallow version, analogous to the shallow ANN stemming from the Cybenko/Hornik Universal Approximation Theorem (UAT), can be easily done by  looking at \cref{eq:5:2:KARTeq}:
\begin{itemize}
    \item \textbf{Input Layer: } This layer has $n$ neurons, mapping the input $(x_1,\dots, x_n) \in \RR$. 
    \item \textbf{First Hidden Layer:} This layer has two ingredients:
        \begin{itemize}
            \item[a.] a set of $n(2n+1)$ \textit{learnable}, \textit{univariate} activation functions $\phi_{q,p}$, mapping each \textit{individual} input components in $\RR$, $x_p \mapsto \phi_{q,p}(x_p)$;
            \item[b.] A sum operation on each $q$ node, $(\phi_{q,1}(x_1), \dots , \phi_{q,n}(x_n)) \to \sum_{p=1}^n \phi_{q,p}(x_p) \rdefeq z_q$. 
        \end{itemize}
    \item \textbf{Second hidden layer:} This layer takes the $2n+1$ nodes (neurons), and applies the cumulative activation $\Phi_q : z_q \to \Phi_q(z_q)$;
    \item \textbf{Output Layer: } This layer simply maps the activated nodes of the second hidden layer into the output, $\Phi_q(z_q) \to \tilde{f}$, via the sum $(\Phi_1(z_1), \dots ,\Phi_{2n+1}(z_{2n+1})) \mapsto\sum_{q=1}^{2n+1} \Phi_q(z_q)$.
\end{itemize}

\begin{figure}[t]
    \centering
    \includegraphics[width=0.95\linewidth]{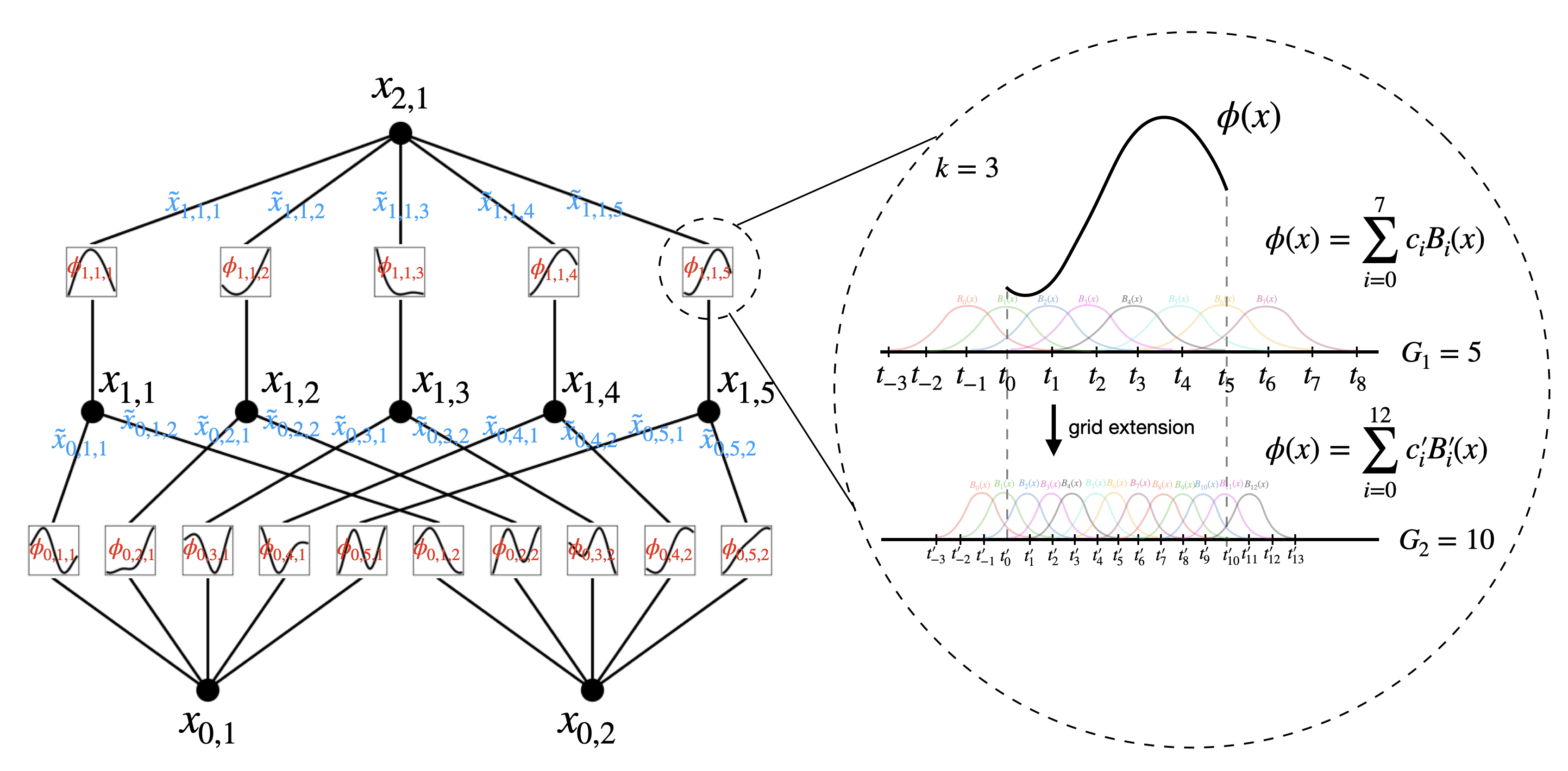}
    \caption{\textit{Left}: Notations of activations that flow through the network. \textit{Right}: an activation function is parametrised as a B-spline, which allows switching between coarse-grained and fine-grained grids. From \cite{liu2024kan}.}
    \label{fig:KAN_spline_notation}
\end{figure}

\begin{table}[t]
    \centering
    \resizebox{0.98\textwidth}{!}{%
    \begin{tabular}{|c|c|c|}
        \hline
        \rowcolor[HTML]{DAE8FC}
        \textbf{model} & \textbf{KAN} & \textbf{MLP} \\ \hline
        \cellcolor[HTML]{FFFFC7}\textbf{shallow} &
        \(f(x_1,\dots,x_n) = \displaystyle\sum_{q=1}^{2n+1} \Phi_q\!\Big( \sum_{p=1}^n \phi_{q,p}(x_p) \Big)\) &
        \(f(x_1,\dots,x_n) = \displaystyle\sum_{i=1}^N c_i\, g(w_i\!\cdot\! x + b_i)\) \\ \hline
        \cellcolor[HTML]{FFFFC7}\textbf{deep} &
        \(\mathrm{KAN}(\mathbf{x}) = (\Phi_L \circ \cdots \circ \Phi_1)(\mathbf{x})\) &
        \(\mathrm{MLP}(\mathbf{x}) = W_L\;\sigma\big(W_{L-1}\;\sigma(\cdots\sigma(W_1\mathbf{x}+b_1)\cdots)+b_{n-1}\big)+b_n\) \\ \hline
    \end{tabular}%
    }
    \caption{Comparison between Kolmogorov–Arnold networks (KAN) and standard multilayer perceptrons (MLP). The \textit{shallow} row shows canonical representation; the \textit{deep} row shows compositional forms.}
    \label{tab:5:2:1:kan_vs_mlp}
\end{table}

This construction is quite analogous to the standard shallow ANN; furthermore, \cite{liu2024kan} furnished a way to \textit{go deep} with KANs, by stacking multiple layers of the form 
\[
    \Phi_\ell = (\Phi_1^{(n)}, \dots, \Phi_{2n+1}^{(\ell)} ), \quad   \Phi_q^{(\ell)} \defeq \Phi_q\!\Big( \sum_{p=1}^n \phi_{q,p}(x_p) \Big) .
\]

More crucially, they suggest a way to make $\phi_{q,p}$ \textit{learnable}, by means of B-splines (see \cref{sec:2:4:Kansa}, and \cref{fig:2:4:Bsplines}). The key tricks to define the KAN layer are:
\begin{enumerate}
    \item \textbf{Residual Activation Functions:} They include a basis function $b(x)$ (analogous to residual connection, so to speak) such that the generic activation function is the sum of the basis function $b(x)$ and the spline:
    \begin{equation}\label{eq:5:2:1:KANlayer}
        \phi(x) = w_b b(x) + w_s \, \text{spline} (x) \,,
    \end{equation}
    where
    \[
        b(x) = \text{silu}(x) \defeq \frac{x}{1+\ee^{-x}} = x \sigma(x) \,,
    \]
    and, crucially for \cite{liu2024kan},
    \begin{equation}\label{eq:5:2:1:BsplineKANeq}
        \text{spline} (x) = \sum_{i} c_i B_i(x) \,,
    \end{equation}
    where $c_i$ are \textit{trainable} parameters, and also $w_b$ and $w_s$ are trainable parameters. 
    \item \textbf{Initialisation scales:}  Each activation function is initialized to have $w_s =1$ and $\text{spline}(x) \approx 0$, while $w_b$ is initialised using Xavier initialisation. 
    \item \textbf{Update of spline grids: } They  update each grid on the fly according to its input activations, to address the issue that splines are defined on bounded regions but activation values can evolve out of the fixed region during training.
\end{enumerate}

\begin{figure}[t]
    \centering
    \includegraphics[width=0.95\linewidth]{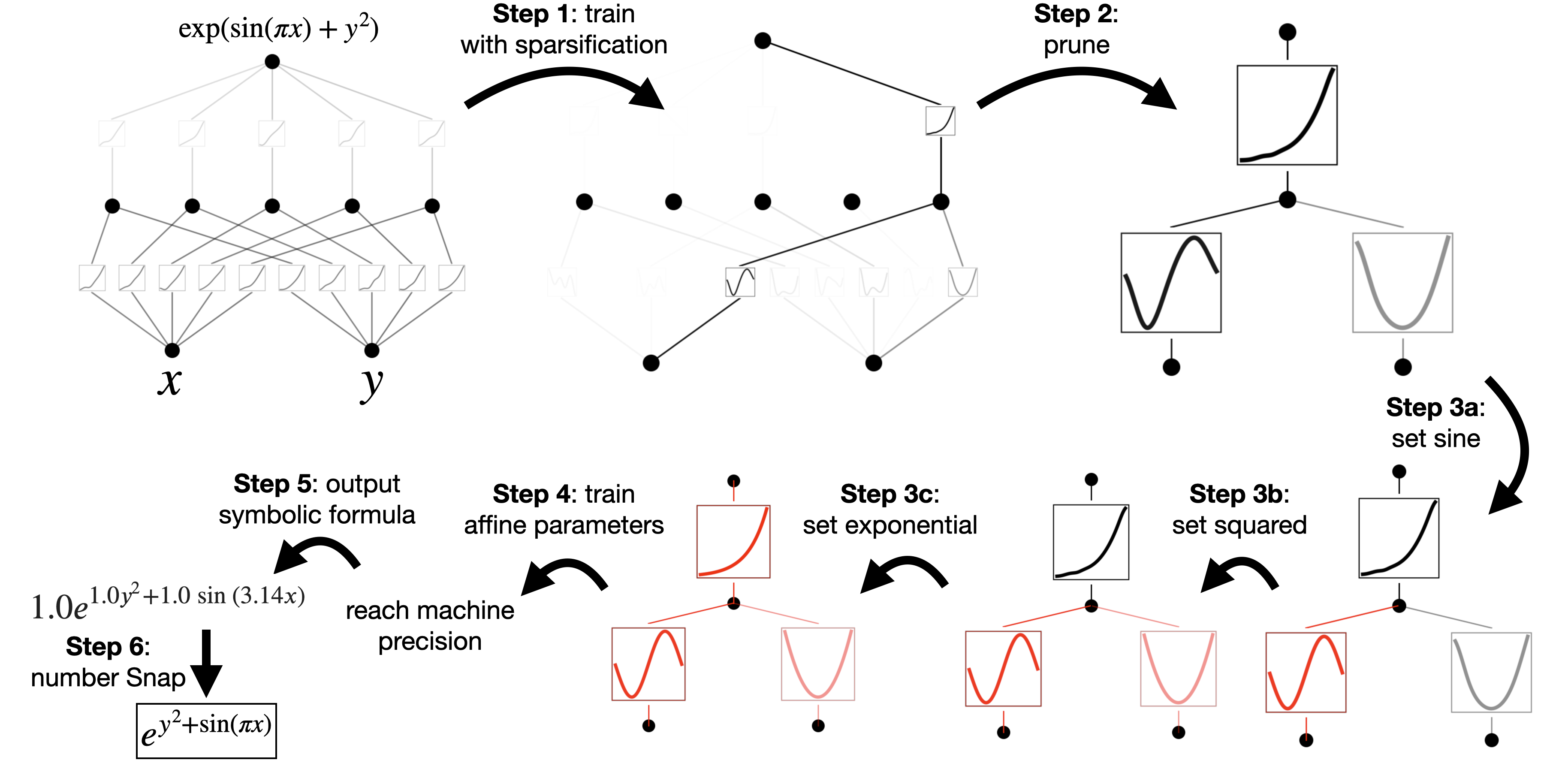}
    \caption{ An example of how to do symbolic regression with KAN. From \cite{liu2024kan}.}
    \label{fig:5:2:1:KAN_symbolic_regression}
\end{figure}

One of the main, crucial, aspects of KANs, is their potential for interpretability: by sparsifying the training, is it possible to use KANs for \textit{symbolic regression}, i.e., to get a compact, symbol representation of the trained function (see \cref{fig:5:2:1:KAN_symbolic_regression}). 

Authors suggest to adapt the standard L1 regularisation for MLP, which favours sparsity. The crucial difference is that (a) in KANs, the norms is defined on the learned activation functions; and (b) L1 is (found empirically to be) insufficient for sparsification of KANs, so entropy regularisation is added. 

The total sparsification loss term is
\begin{equation}
    \ell_{\text{spars}} = \lambda \left( \mu_1 \sum_{l=1}^{L} |\Phi_l|_1  + \mu_2 \sum_{l=1}^{L} S(\Phi_l)\right)  \,,
\end{equation}
where
\begin{align}
    |\Phi_l|_1 & \defeq \sum_{i=1}^{n_{in}} \sum_{j=1}^{n_{out}} \left| \phi_{i,j} \right|_1 \,, \qquad |\phi|_1 \defeq \frac{1}{N_p} \sum_{s=1}^{N_p} \left| \phi(x^{(s)}) \right|  , \\
    S(\Phi) & \defeq - \sum_{i=1}^{n_{in}} \sum_{j=1}^{n_{out}} \frac{\left| \phi_{i,j} \right|_1}{\left| \Phi \right|_1} \log \left[  \frac{\left| \phi_{i,j} \right|_1}{\left| \Phi \right|_1}  \right] \,,
\end{align}
i.e., the L1 norm of an activation function $\phi$ is its average magnitude over its $N_p$ inputs, while for the entire KAN layer $\Phi$ with $n_{in}$ inputs and $n_{out}$ outputs, its norm is simply the sum of the L1 norms of all its activation functions. 

The crucial steps, after sparsified training, are \textit{pruning} and \textit{symbolification}; 

\noindent \textbf{Pruning:}  Sparsify KANs on the node level (rather than on the edge level). For each node (say the $i^{th}$ neuron in the $l^{th}$ layer), let us define its \textit{incoming} and \textit{outgoing} score as
\begin{align}
    I_{l, i} &\defeq \max_k \left( \left| \phi_{l-1, i, k} \right|_1 \right) \,, \\
    O_{l, i} &\defeq \max_k \left( \left| \phi_{l+1, i, k} \right|_1  \right) \,, 
\end{align}
and consider a node to be important if both incoming and outgoing scores are greater than a threshold hyperparameter (usually, $\theta = 10^{-2}$). Then, all unimportant neurons are pruned.

\noindent \textbf{Symbolification:}  Symbolification means replacing the (unpruned) activation functions with symbolic expression via per-function fitting. Getting preactivations $x$ and postactivations $y$ from samples, we simply fit the affine parameter sets $(a,b,c,d)$ such that, called $f$ the symbolic expression (e.g., $\sin, \exp, \tanh,$ etc.), we have $y\approx cf(ax+b)+d$. The fitting is done by iterative grid search of $(a,b)$ and linear regression.

\subsubsection{Critiques to KAN}


Following the introduction of Kolmogorov-Arnold Networks, it is essential to critically examine their practical limitations and theoretical foundations. As noted before, early critiques were pointed out on the applicability of the Kolmogorov-Arnold representation theorem in the context of neural networks. These classical theoretical objections, combined with recent systematic empirical evaluations, provide a nuanced perspective on the promise and current limitations of KANs.

\paragraph{Historical Theoretical Critiques:}
The relevance of the Kolmogorov-Arnold representation theorem as a direct foundation for practical learning architectures was already questioned in the late 1980s, notably by \cite{girosi1989representation}. These critiques should
not be interpreted as objections to the mathematical correctness of KART, but rather as objections to the inference that an exact representation theorem automatically yields a useful learning architecture.

First, there is a genuine \emph{regularity obstruction}. Results originating with Vitushkin show that the universal inner functions required by Kolmogorov-type exact representations cannot, in general, be required to be continuously differentiable. Later constructions show that they may be chosen with Lipschitz regularity, but \(C^1\) smoothness is incompatible with the universal exact superposition property \cite{vitushkin1954hilbert,vitushkin2004hilbert,girosi1989representation}.
This limits the direct usefulness of the classical universal inner functions as smooth trainable building blocks.

Second, the classical representation is not a learning prescription. Importantly, the inner functions \(\phi_{q,p}\) \emph{are universal} and can be chosen independently of the target function. It is the outer functions \(\Phi_q\) that depend on the particular \(f\). KART therefore guarantees existence of an exact representation, but does not by itself provide a
finite-dimensional parametrisation, an optimisation procedure, or statistical learning guarantees for recovering the required target-dependent functions from data.

Third, exact representability should not be confused with efficient approximation or learnability. Universal approximation results can still suffer from approximation complexity that grows rapidly with input dimension; special architectural advantages against the curse of dimensionality require additional structural assumptions on the target class \cite{poggio2020theoretical}. Thus KART alone does not establish that a KART-inspired neural architecture avoids the curse of dimensionality in practical learning problems.

\paragraph{Modern Empirical Findings:}
Regardless of these theoretical concerns, recent systematic and fair comparisons have revealed significant practical challenges regarding computational efficiency, training stability, and scalability of KANs.

A primary critique, as rigorously investigated by \cite{yu2024kan}, concerns the fairness of comparisons often made in favour of KANs. 
Their main finding is that MLPs generally outperform KANs on the tested tasks, with a notable exception in symbolic-formula representation. Their ablation study indicates that much of this advantage can be attributed to the B-spline activation itself: equipping an MLP with B-spline activations substantially improves its symbolic-regression performance and can match or exceed that of KAN. They also report stronger catastrophic forgetting for KAN in their class-incremental continual-learning experiment.

Further deepening this critique, \cite{pal2024understanding} focus specifically on the convergence dynamics and computational overhead of B-spline-based KANs. Their analysis reveals that the training dynamics of these KANs are fundamentally different from those of MLPs, often leading to slower convergence and difficulties in optimization. The study highlights that the expressive power gained from learnable spline functions comes at a steep computational cost, as the piecewise polynomial evaluations and the associated parameter updates are significantly more expensive than the simple linear transformations in MLPs. This computational inefficiency, as argued by \cite{pal2024understanding}, severely limits the scalability of KANs to larger datasets and deeper architectures.

In the specific context of scientific machine learning, \cite{shukla2024comprehensive} provide a comprehensive comparison between MLPs and KANs for solving differential equations and within operator networks. Their findings corroborate and extend the previous critiques. While acknowledging that KANs can achieve high accuracy with a very small number of parameters for certain problems, \cite{shukla2024comprehensive} demonstrate that this efficiency does not translate to faster training. On the contrary, they report that KANs are consistently slower to train than MLPs by a significant margin. Moreover, their study reveals that KANs can be unstable and highly sensitive to the choice of hyperparameters like grid size and spline order, whereas modern MLPs with appropriate activations and architectures often provide a more robust and computationally efficient baseline for solving partial differential equations and learning operators \cite{shukla2024comprehensive}.

For additional critiques, studies, applications of KANs, see also \cite{hou2024kolmogorov, noorizadegan2025practitioner, dutta2025first, GuideToKAN_GitHub}.

\subsection{Alternative KAN implementations}
Following the revival of Kolmogorov-Arnold Networks, the immediate goal was to speed up their training and inference. Furthermore, multiple architectural modifications were introduced, replacing B-splines with faster, easier, more common univariate function interpolators.  A nice living review repo on KANs, their implementations, and applications (with link to GitHub code bases and papers) is \cite{AwesomeKan2024Git}.

\subsubsection{EfficientKAN}

The first relevant work to speed up the original KAN implementation was Efficient-KAN (eKAN) \cite{EfficientKAN2024Git}; as its author stated, 

\quotespace
\begin{quote}
    \say{\textit{the performance issue of the original implementation is mostly because it needs to expand all intermediate variables to perform the different activation functions. For a layer with} \texttt{in\_features} \textit{input and} \texttt{out\_features}\textit{ output, the original implementation needs to expand the input to a tensor with shape }(\texttt{batch\_size, out\_features, in\_features})\textit{ to perform the activation functions. However, all activation functions are linear combination of a fixed set of basis functions which are B-splines; given that, we can reformulate the computation as activate the input with different basis functions and then combine them linearly. This reformulation can significantly reduce the memory cost and make the computation a straightforward matrix multiplication, and works with both forward and backward pass naturally.}}
\end{quote}
\quotespace

Another fact hampering the training speed was the L1 regularisation to foster sparsification. As we have seen, the original L1 regularization is defined on the input samples, which requires non-linear operations on the  (\texttt{batch\_size, out\_features, in\_features}) tensor, and is thus not compatible with the Efficient-KAN reformulation. So, in Efficient-KAN, the L1 regularization of the activation function is replaced with a L1 regularization on the weights, which is more common in neural networks and is compatible with the reformulation. 

This kind of implementation was immediately used in the PINN community, as one of the earlier work using KAN for PINNs, \cite{wang2025kolmogorov}, one of the first to introduce the PIKAN nomenclature (together with \cite{shukla2024comprehensive}), used Efficient-KAN for tackling differential problem. The workflow is as easy as simply replacing the approximator function $\hat{u}_\theta(\xx) \leftarrow \mathrm{DNN}_\theta(\xx)$ with $\hat{u}_\theta(\xx) \leftarrow \mathrm{eKAN}_\theta(\xx)$.

\subsubsection{Using Radial Basis: FastKAN}

In \cite{li2024kolmogorov}, the author pointed out that

\quotespace
\begin{quote}
    \say{\textit{the 3-order B-splines used in Kolmogorov-Arnold Networks (KANs) can be well approximated by Gaussian radial basis functions. Doing so leads to FastKAN, a much faster implementation of KAN which is also a radial basis function (RBF) network.}}
\end{quote}
\quotespace

The point is quite easy; in \cref{eq:5:2:1:KANlayer}, instead of using the B-splines of \cref{eq:5:2:1:BsplineKANeq}, FastKAN uses gaussian radial basis function
\begin{equation}
    \phi(x) = w_b b(x) + w_s  \left(\sum_{i} k_i \, \text{rbf}(\left|\text{Lin}(x) -c_i\right|) \right)  \,, \quad k_i, c_i \in \RR,
\end{equation}
where $\text{Lin}(x) = Ax + b$ is a linear layer. Notice that this expression can be easily parallelised!
\begin{equation}
    \phi(\xx) =  w_b b(\xx) +   w_s    \left(  \mathbf{k}  \odot \, \text{rbf}(\left|\text{Lin}(\xx) - \mathbf{c}\right|) \right)\cdot \mathrm{1}_d ,
\end{equation}
where $ \mathbf{k} ,\mathbf{c}\sim (\texttt{in\_dim},\, \texttt{grid\_size})$, $\odot$ represents the per-element product, and $\cdot \mathrm{1}_d  $ denotes the sum over the \texttt{grid\_size} dimension. 

This fact will return in the next approaches.

\subsubsection{Chebyshev-KAN}

As already noticed with FastKAN, the main innovation of KAN is to rely in Kolmogorov-Arnold Theorem, and find a way to learn the univariate activation functions. KAN/eKAN used B-splines, FastKAN uses RBF. But a plethora of common basis functions exists, such as the Chebyshev Polynomials \cite{vcebyvsev1853theorie}. Using that, authors of \cite{ss2024chebyshev} introduced the Chebyshev-KAN (cKAN). For $x\in [-1,+1]$, Chebyshev polynomials of the first kind are special functions defined iteratively as 
\begin{align*}
    T_0(x) &= 1, \\
    T_1(x) &= x \,, \\
    T_n(x) &= 2x T_{n-1} (x)-T_{n-2} (x), \quad \forall n\ge 2 .
\end{align*}
It can be proven that they admit the compact form
\[
T_n(x) = \cos \left( n \arccos(x) \right),
\]
which may or may not be more useful from the numerical standpoint. 
Chebyshev polynomials of the second kind, $U_n(x)$, satisfy a similar recurrence relation to the first kind, but with $U_1(x) = 2x$ instead of $U_1(x) = x$. Both families form orthogonal bases on $[-1, 1]$ with respect to different weight functions; thus
\begin{align*}
    U_0(x) &= 1, \\
    U_1(x) &= 2x \,, \\
    U_n(x) &= 2x U_{n-1} (x)-U_{n-2} (x), \quad \forall n\ge 2 .
\end{align*}
They can be expressed also as
\[
U_n(x) = \frac{\sin ( (n+1) \arccos(x) )}{\sqrt{1-x^2}} \,.
\]
They are \textit{orthogonal polynomials}, i.e.
\begin{align}
    \int_{-1}^{+1} \frac{T_m(x) T_n(x)}{\sqrt{1-x^2}} \, \dd x = \begin{cases}
        0, & n\neq m \\
        \pi, & n=m=0, \\
        \frac{\pi}{2} , &n=m\neq 0.
    \end{cases} \\
    \int_{-1}^{+1}   U_m(x) U_n(x)  \sqrt{1-x^2} \, \dd x = \begin{cases}
        0, & n\neq m \\
        \frac{\pi}{2}, & n=m, \\
    \end{cases}
\end{align}

The cKAN layer output is computed by performing an Einstein summation operation (einsum) over the Chebyshev polynomials tensor $\mathrm{T}\sim(B, \texttt{input\_dim}, \texttt{degree}+1)$ and the Chebyshev coefficients tensor $\mathrm{C}\sim(B, \texttt{output\_dim}, \texttt{degree}+1)$. This operation effectively combines the polynomial bases with the learned coefficients to produce the final output tensor 
\[
    \mathbf{y}_o = \sum_{i=1}^{\texttt{input\_dim}} \sum_{j=0}^{\texttt{degree}} \mathrm{T}_{bij} \cdot \mathrm{C}_{boj} (\xx) \,.
\]

The $\mathrm{C}_{boj} (\xx)$ tensor is built easily following three steps:
\begin{itemize}
    \item[1.] Nornmalise the input $x \mapsto\tilde{x} = \tanh(x)$
    \item[2.] The Chebyshev polynomials - up to a degree $n$ (hyperparameter) are generated, $\tilde{x} \mapsto [T_0(\tilde{x}), T_1 (\tilde{x}), \dots , T_n(\tilde{x})]$
    \item[3.] They are combined using a set of learnable parameters $\Theta\sim (d_{in}, d_{out}, n+1)$.
\end{itemize}

\begin{lstlisting}[language=Python, caption=Torch implementation of ChebyshevKAN (cKAN).  From \cite{JacobiKan2024Git}.]
import torch
import torch.nn as nn

class ChebyKANLayer(nn.Module):
    def __init__(self, input_dim, output_dim, degree):
        super(ChebyKANLayer, self).__init__()
        self.inputdim = input_dim
        self.outdim = output_dim
        self.degree = degree

        self.cheby_coeffs = nn.Parameter(torch.empty(input_dim, output_dim, degree + 1))
        nn.init.normal_(self.cheby_coeffs, mean=0.0, std=1/(input_dim * (degree + 1)))

    def forward(self, x):
        x = torch.reshape(x, (-1, self.inputdim))  # shape = (batch_size, inputdim)
        # Since Chebyshev polynomial is defined in [-1, 1]
        # We need to normalize x to [-1, 1] using tanh
        x = torch.tanh(x)
        # Initialize Chebyshev polynomial tensors
        cheby = torch.ones(x.shape[0], self.inputdim, self.degree + 1, device=x.device)
        if self.degree > 0:
            cheby[:, :, 1] = x
        for i in range(2, self.degree + 1):
            cheby[:, :, i] = 2 * x * cheby[:, :, i - 1].clone() - cheby[:, :, i - 2].clone()
        # Compute the Chebyshev interpolation
        y = torch.einsum('bid,iod->bo', cheby, self.cheby_coeffs)  # shape = (batch_size, outdim)
        y = y.view(-1, self.outdim)
        return y
\end{lstlisting}

\subsubsection{Jacobi-KAN}
A more general basis function for the $[-1, +1]$ interval are the Jacobi Polynomials, which, for certain values of their defining parameters, reduces to Chebyshev polynomials. 

For parameters $\alpha,\beta>-1$, the \emph{Jacobi polynomials} $\{P_n^{(\alpha,\beta)}(x)\}_{n\ge 0}$ form a family of orthogonal polynomials on $[-1,1]$ with respect to the weight
\[
    w_{\alpha,\beta}(x) = (1-x)^{\alpha}(1+x)^{\beta}.
\]
They satisfy the orthogonality relation
\[
    \int_{-1}^{1} P_m^{(\alpha,\beta)}(x)\,P_n^{(\alpha,\beta)}(x)\,
    (1-x)^{\alpha}(1+x)^{\beta}\,\mathrm{d}x
    = \delta_{mn} \mathcal{N}_{m,n}^{\alpha, \beta} \,, 
\]
where the normalsiation factor is
\[
    \mathcal{N}_{m,n}^{\alpha, \beta}  = \frac{2^{\alpha+\beta +1}}{2n+\alpha + \beta +1} \, \frac{\Gamma(n+\alpha+1)\Gamma(n+\beta+1)}{n! \Gamma(n+\alpha+\beta+1)} \,.
\]
Jacobi polynomials obey the recurrence
\[
\begin{aligned}
    P_0^{(\alpha,\beta)}(x) &= 1, \\
    P_1^{(\alpha,\beta)}(x) &= \tfrac12\big[(2+\alpha+\beta)x + (\alpha-\beta)\big],
\end{aligned}
\]
and for $n\ge 1$,
\[
\begin{aligned}
    &2(n+1)(n+\alpha+\beta+1)(2n+\alpha +\beta)\,P_{n+1}^{(\alpha,\beta)}(x) = \\
    &\quad (2n+\alpha+\beta+1)\Big[(2n+\alpha+\beta)(2n+\alpha+\beta+2)x + \alpha^2-\beta^2\Big]P_n^{(\alpha,\beta)}(x) \\
    &\quad -\,2(n+\alpha)(n+\beta)(2n+\alpha+\beta+2)\,P_{n-1}^{(\alpha,\beta)}(x).
\end{aligned}
\]

Jacobi polynomials include several classical families:
\begin{align*}
    P_n^{(0,0)}(x) &= P_n(x) \,, \hspace{3cm} \text{(Legendre polynomials)} \\
    P_n^{(-\tfrac12,-\tfrac12)}(x) &= \frac{\Gamma(n+\tfrac12)}{\sqrt{\pi}\,\Gamma(n+1)}\,T_n(x) \,, \qquad  \text{(Chebyshev of the first kind)} \\
    P_n^{(\tfrac12,\tfrac12)}(x) &= \frac{2\Gamma(n+\tfrac32)}{\sqrt{\pi}\,\Gamma(n+1)}\,U_n(x) \,, \qquad  \text{(Chebyshev of the second kind)}.
\end{align*}

Thus Chebyshev and Legendre polynomials arise as particular choices of $(\alpha,\beta)$, making Jacobi polynomials an unifying basis family.

From here, it is immediate to replace Chebyshev polynomials with Jacobi Polynomials, and thus define the JacobiKAN (jKAN).

\begin{lstlisting}[language=Python, caption=Torch implementation of JacobiKAN (jKAN). From \cite{JacobiKan2024Git}.]
import torch
import torch.nn as nn

class JacobiKANLayer(nn.Module):
    def __init__(self, input_dim, output_dim, degree, a=1.0, b=1.0):
        super(JacobiKANLayer, self).__init__()
        self.inputdim = input_dim
        self.outdim   = output_dim
        self.a        = a
        self.b        = b
        self.degree   = degree

        self.jacobi_coeffs = nn.Parameter(torch.empty(input_dim, output_dim, degree + 1))
        nn.init.normal_(self.jacobi_coeffs, mean=0.0, std=1/(input_dim * (degree + 1)))

    def forward(self, x):
        x = torch.reshape(x, (-1, self.inputdim))  # shape = (batch_size, inputdim)
        # Since Jacobi polynomial is defined in [-1, 1]
        # We need to normalize x to [-1, 1] using tanh
        x = torch.tanh(x)
        # Initialize Jacobian polynomial tensors
        jacobi = torch.ones(x.shape[0], self.inputdim, self.degree + 1, device=x.device)
        if self.degree > 0: 
            # degree = 0: jacobi[:, :, 0] = 1 (already initialized) ; 
            # degree = 1: jacobi[:, :, 1] = x ; 
            jacobi[:, :, 1] = ((self.a-self.b) + (self.a+self.b+2) * x) / 2
        for i in range(2, self.degree + 1):
            theta_k  = (2*i+self.a+self.b)*(2*i+self.a+self.b-1) / (2*i*(i+self.a+self.b))
            theta_k1 = (2*i+self.a+self.b-1)*(self.a*self.a-self.b*self.b) / (2*i*(i+self.a+self.b)*(2*i+self.a+self.b-2))
            theta_k2 = (i+self.a-1)*(i+self.b-1)*(2*i+self.a+self.b) / (i*(i+self.a+self.b)*(2*i+self.a+self.b-2))
            jacobi[:, :, i] = (theta_k * x + theta_k1) * jacobi[:, :, i - 1].clone() - theta_k2 * jacobi[:, :, i - 2].clone()  # 2 * x * jacobi[:, :, i - 1].clone() - jacobi[:, :, i - 2].clone()
        # Compute the Jacobian interpolation
        y = torch.einsum('bid,iod->bo', jacobi, self.jacobi_coeffs)  # shape = (batch_size, outdim)
        y = y.view(-1, self.outdim)
        return y
\end{lstlisting}

\subsection{KAN for PDEs: PIKAN and DeepOKan}

\begin{figure}[t]
    \centering
    \begin{subfigure}[t]{0.48\textwidth}
        \centering
        \includegraphics[width=\textwidth]{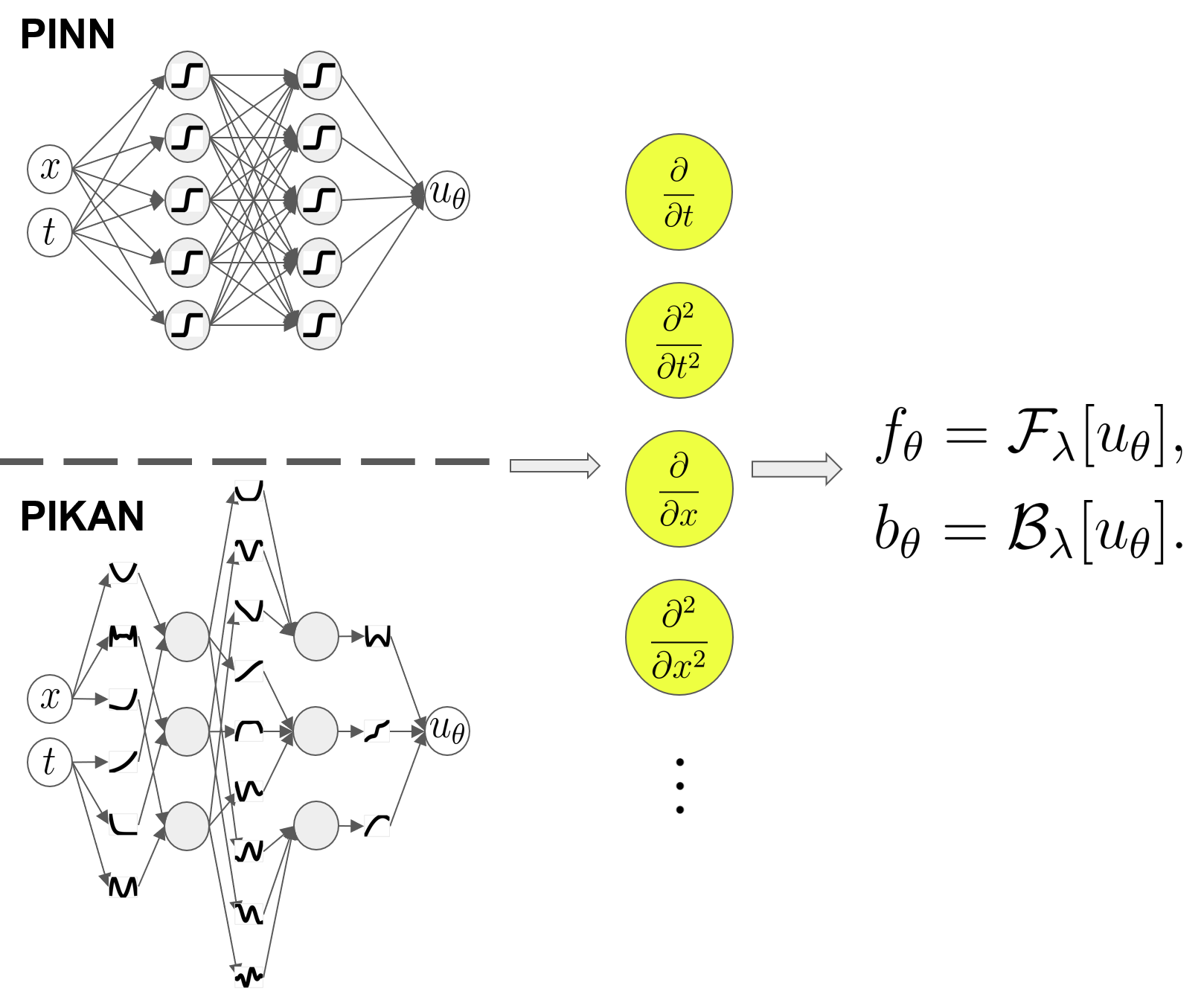}
        \caption{Physics-Informed Networks. Top, PINN; bottom, PIKAN.}
    \end{subfigure}
    \hfill
    \begin{subfigure}[t]{0.48\textwidth}
        \centering
        \includegraphics[width=\textwidth]{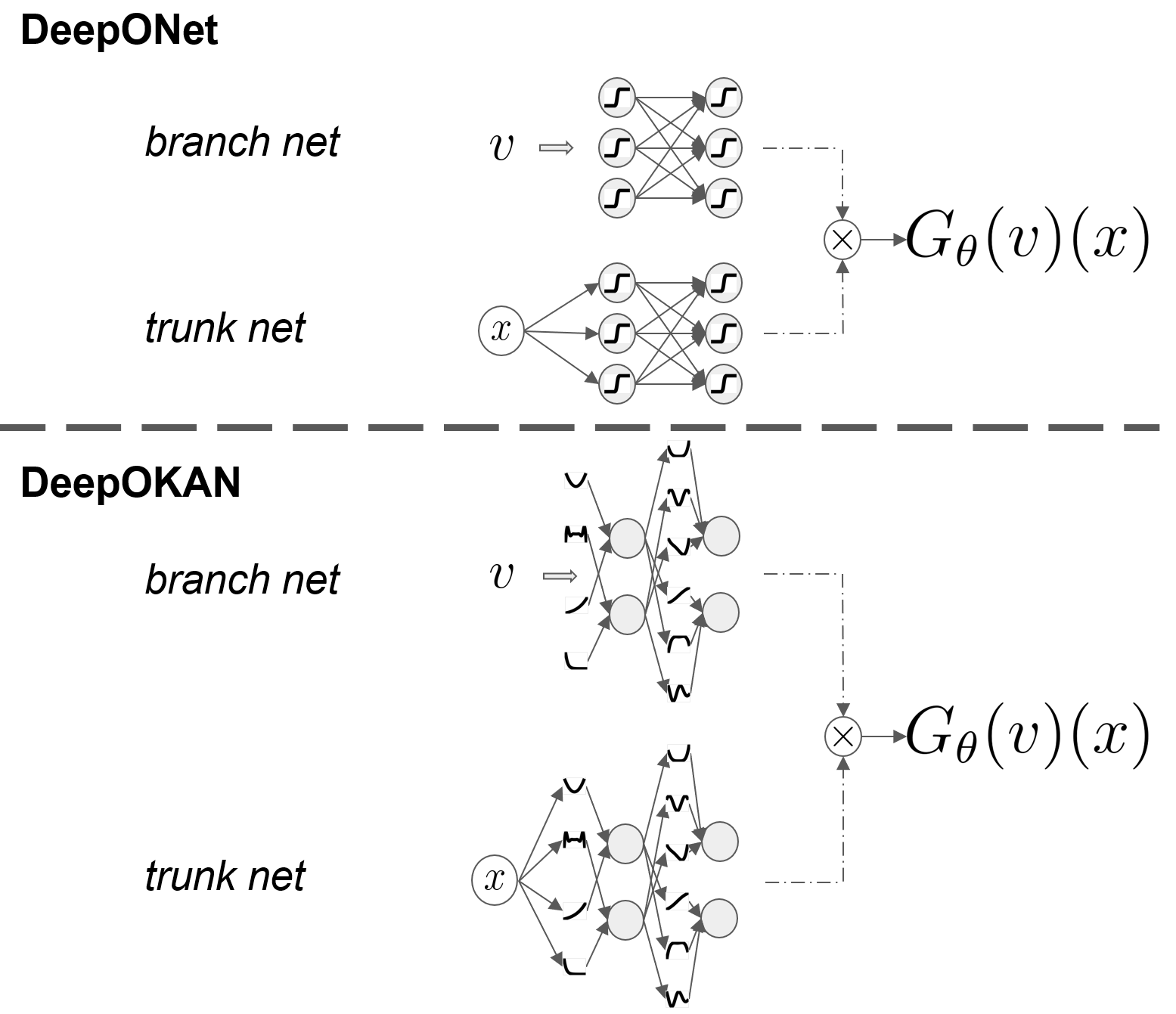}
        \caption{Operator Networks. Top, DeepONet; bottom, DeepOKAN.}
    \end{subfigure}
    \caption{An illustration of MLP and KAN for \textbf{(a)} differential equations and \textbf{(b)} operator networks (DeepONet \cite{lu2021learning} is used as the representation model for operator learning). From \cite{shukla2024comprehensive}.}
    \label{fig:5:2:3:PIKANvsPINN}
\end{figure}

As discussed earlier, KANs can be readily adapted to Physics-Informed Machine Learning (PIML) tasks, including forward problems, inverse problems, and parametric PDE solving \cite{shukla2024comprehensive}. Furthermore, it is possible to replace MLP blocks in Deep Operator Networks to build up a Deep Operator Kolmogorov Arnold Networks (DeepOKAN).  In \cite{shukla2024comprehensive}, authors compared performances of PINN, PIKAN on certain benchmark tasks, and DeepONets vs DeepOKANs on certain others. 

They found that both MLPs and KANs achieve comparable accuracies, but KANs generally exhibit slower training convergence and lower data efficiency. Interestingly, cPIKAN (Chebyshev PIKAN) can achieve the same accuracy with far less network parameters, but it takes more GPU hours to train. Nevertheless, the DeepOKAN has shown competitive performance in operator learning compared to the DeepONet, indicating DeepOKAN as a promising alternative representation model. Furthermore, DeepOKAN demonstrates significantly greater robustness to noisy input functions at test time when trained on clean data. 

\section{Learning Operators, Part II: Neural Operators  - formal theory} \label{sec:5:3:NOsTheory}

Let us now return to Operator Learning. Following \cite{kovachki2023neural, bhattacharya2021model, duruisseaux2025fourier}, we want now to formally introduce the concept of Neural Operators, its formal definition, and its underlying representation properties. 

\begin{table}[h!]
\centering
\resizebox{\textwidth}{!}{%
\begin{tabular}{l|c|c|c|c}
\diagbox{Property\hspace{.5cm}}{\hspace{2em}\raisebox{-.3cm}{Model}}&
    NNs & DeepONets & Interpolation & Neural Operators \\
    \hline
    Discretization Invariance & \xmark & \xmark & \cmark & \cmark \\
    \hline
    Is the output a function? & \xmark & \cmark & \cmark & \cmark \\
    \hline
    Can query the output at any point? & \xmark & \cmark & \cmark & \cmark \\
    \hline
    Can take the input at any point? & \xmark & \xmark & \cmark & \cmark \\
    \hline
    Universal Approximation of operators & \xmark & \cmark & \xmark & \cmark \\
    \hline
    \end{tabular}%
}
\caption{Comparison of deep learning models. The first row indicates whether the model is discretization invariant. The second and third rows indicate whether the output and input are functions. The fourth row indicates whether the model class is a universal approximator of operators. Neural Operators are discretization invariant deep learning methods that output functions and can approximate any operator. From \cite{kovachki2023neural}.}
\label{tab:5:4:deeplearning_comparison}
\end{table}

The main properties we want our \textit{Neural Operators} $G_\theta$ to obey are, loosely speaking
\begin{itemize}
    \item[1.] Maps infinite-dimensional (Banach/Hilbert) spaces to infinite-dimensional (Banach/Hilbert) spaces;
    \item[2.] Admits an explicit realisation for certain input-function values, so that we can use the Neural Operator to approximate a target Banach space function at any point in its domain\note{Statements concerning evaluation ``at any point'' implicitly assume a function-space setting in which pointwise evaluation is meaningful, for example spaces continuously embedded in \(C(\overline D)\). More general function spaces may instead require other finite-dimensional representations, such as basis coefficients or bounded linear functionals.}; 
    \item[3.] Satisfy an Universal Approximation Theorem of some sort. 
\end{itemize}

We have already sees an Operator Network formalism, the Deep Operator Network (DeepONet), relying on the universal approximation theorem for operators. These are a class of operators that consist of a branch net and a trunk net. The trunk-net allows queries at any point, but the branch-net constrains the input to fixed locations. This fact breaks explicitly the discretisation invariance property (\cref{tab:5:4:deeplearning_comparison}). However, it is possible to modify the branch-net to make the methodology discretization invariant, for example by using
the PCA-based approach as used in \cite{de2022cost}.

Let us follow \cite{kovachki2023neural} and require a discretization-invariant model with a fixed number of parameters to satisfy the following:
\begin{enumerate}[leftmargin=*]
    \item  acts on any discretization of the input function, i.e. accepts any set of points in the input domain,
    \item can be evaluated at any point of the output domain,
    \item  converges to a continuum operator as the discretization is refined. 
\end{enumerate} 

\noindent The first two requirements of accepting any input and output points in the domain is a natural requirement for discretization invariance, while the last one ensures consistency in the limit as the discretization is refined. 

Before delving into the formalism of Neural Operators, let us stop for a moment and re-focus on a specific task: a pure-boundary problem of the form
\begin{align}
    \mathrm{L}_a [u](x) &= f(x) \,, \quad x\in D\subset \RR^d , \\
    u(x) &= 0 \,, \qquad \; x\in \partial D .
\end{align}

\noindent This may represent, for example, the stationary Advection-Reaction-Diffusion PDE, stationary Navier-Stokes, etc. Relying on intuition from the weak formulation of such a boundary problem, we can see that $u \in \mathcal{U}(D; \RR)$ where $u: D \to \RR$, while the source term $f$ lives in the dual space $\mathcal{U}^*$, acting on $u$ via the duality pairing
\[
 u \mapsto \langle f, u \rangle = \int_D f u .
\]
The subscript $a$ of the differential operator $\mathrm{L}_a$ refers to a function $a\in\mathcal{A}(D'\subset\RR^{d_a};\RR)$ (for example, the permeability field in the stationary Darcy Flow PDE). In this sense, $\mathrm{L}: \mathcal{A}\times \mathcal{U} \to \mathcal{U}^*$, or, chosen $a\in\mathcal{A}$, $\mathrm{L}_a : \mathcal{U} \to \mathcal{U}^*$. Thus, if such operator can be inverted, we can build
\begin{equation}
    G^\dagger[a] \defeq \mathrm{L}_a^{-1} f : \mathcal{A} \to \mathcal{U}, \text{ s.t. }  G^\dagger : a \mapsto u ,
\end{equation}
i.e., $G^\dagger$ maps the \textit{parameter function} $a$ to the solution of the PDE, 
\begin{equation}
    u_{\text{sol}} (x) = G^\dagger [a] (x), \Rightarrow \mathrm{L}_a [G^\dagger [a]](x) = \underbrace{ \mathrm{L}_a  \mathrm{L}_a^{-1} }_{1} \;  f = f. 
\end{equation}

Our goal is to learn this mapping, using a finite collection of observations (input/output). I.e. we aim to build
\begin{equation}
    G_\theta : \mathcal{A}\times \Theta \to \mathcal{U},
\end{equation}
where we want to pick up the optimal $\theta^\dagger \in \Theta$ such that $G_{\theta^\dagger}\approx G^\dagger$.

We sample the parameter functions as i.i.d. samples from a certain probability measure, while the solutions are push-forwards
\[
\left\{ a^{(i)} , u^{(i)} =G^\dagger[a^{(i)}] \right\}_{i=1, \dots, N} 
\]
where $a^{(i)}\sim \mu$ are i.i.d. samples drawn from the probability measure $\mu$ with support on $\mathcal{A}$. But these functions are defined on a certain domain, $D'\subset\RR^{d_a}$ and $D\subset\RR^{d}$. Then
\begin{definition}
We call a \textit{discrete refinement} of the domain \(D \subset \RR^d\) any sequence of nested sets \(D_1 \subset D_2 \subset \dots \subset D\) with \(|D_L| = L\) for any \(L \in \N\) such that, for any \(\epsilon > 0\), there exists a number \(L = L(\epsilon) \in \N\) such that
\[D \subseteq \bigcup_{x \in D_L} \{y \in \RR^d : \|y - x\|_2 < \epsilon\}.\]
Here, $|D_L| = L$ denotes the cardinality (number of points) in the discrete set $D_L$.
\end{definition}

\begin{figure}[t]
    \centering
    \begin{subfigure}[t]{0.32\textwidth}
        \centering
        \includegraphics[width=\textwidth]{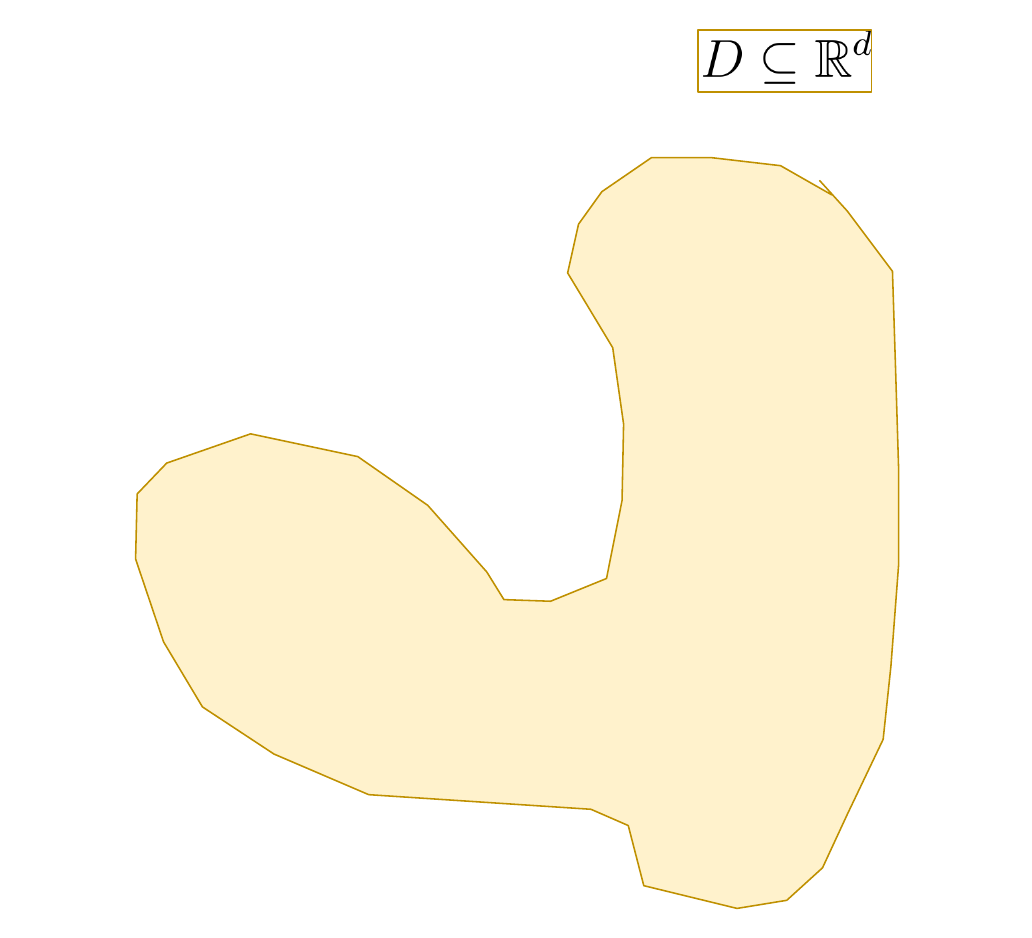}
    \end{subfigure}
    \hfill
    \begin{subfigure}[t]{0.32\textwidth}
        \centering
        \includegraphics[width=\textwidth]{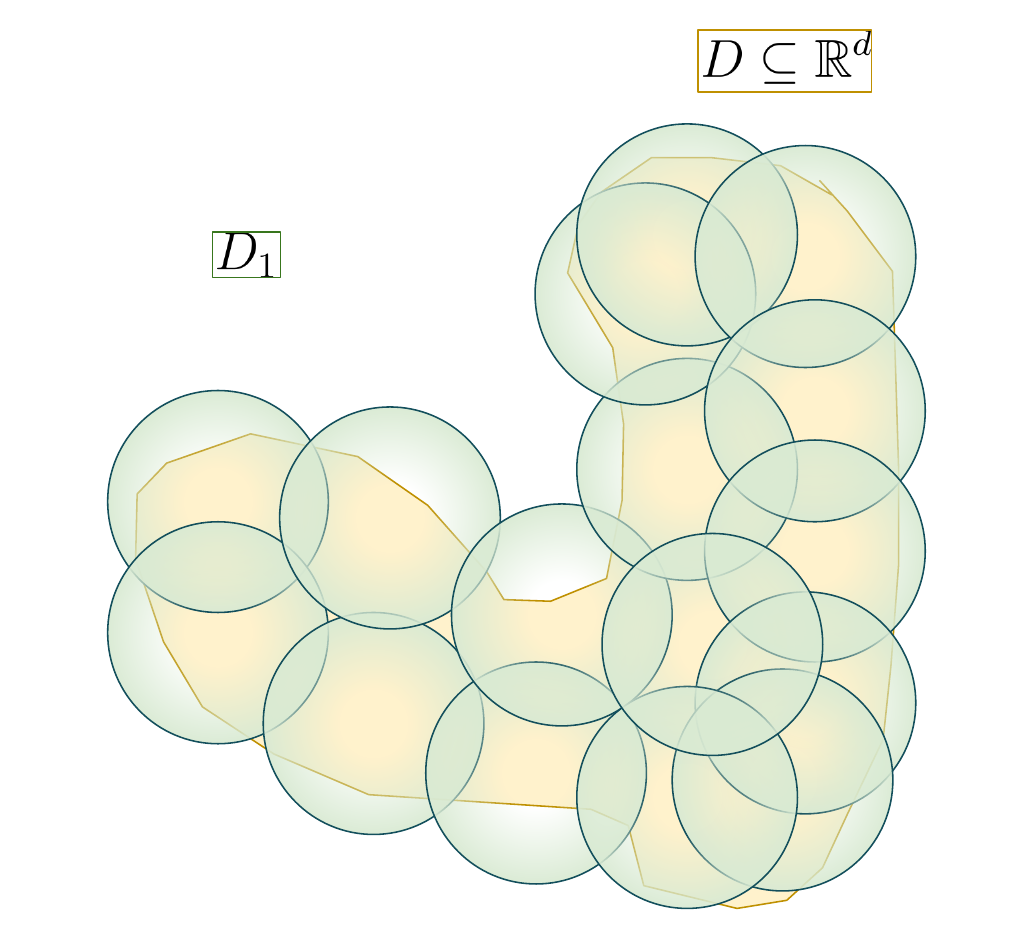}
    \end{subfigure}
    \hfill
    \begin{subfigure}[t]{0.32\textwidth}
        \centering
        \includegraphics[width=\textwidth]{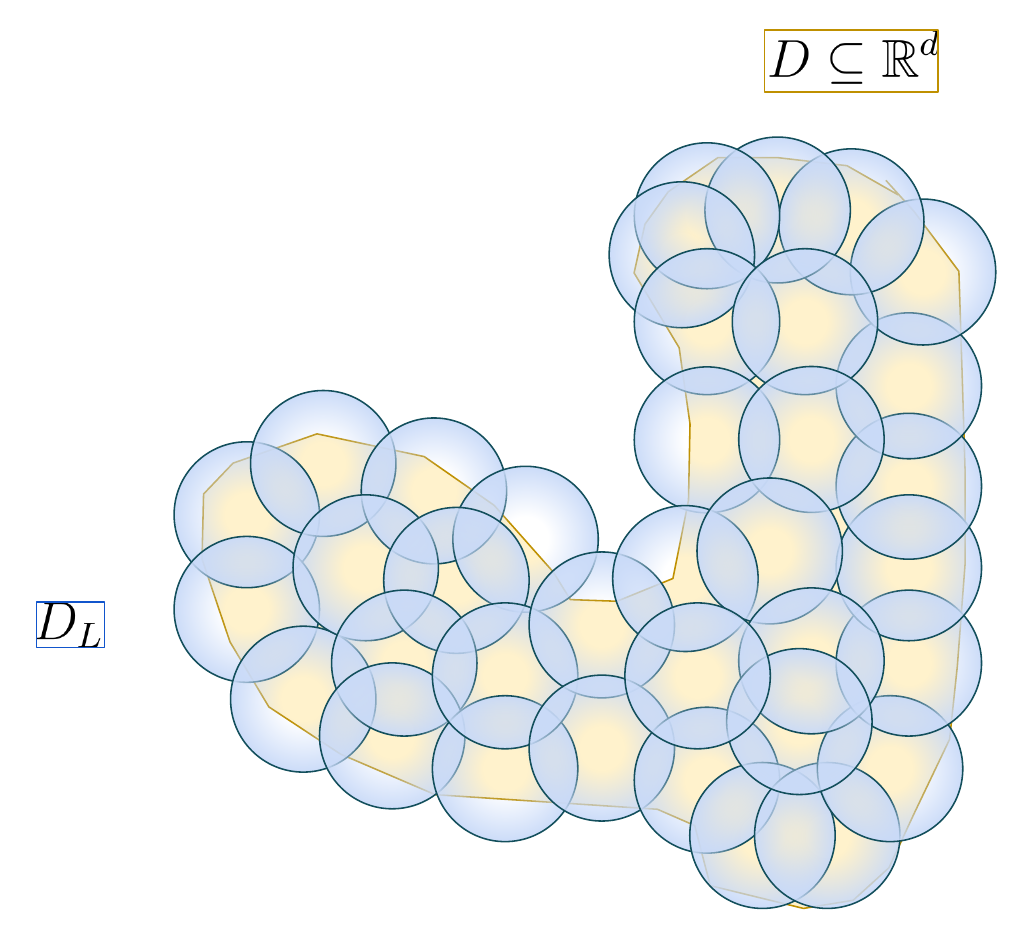}
    \end{subfigure}
    \caption{Visual representation of the Domain refinement. }
    \label{fig:5:3:Domain_refinement}
\end{figure}

\begin{definition}
Given a discrete refinement \((D_L)_{L=1}^\infty\) of the domain \(D \subset \RR^d\), any member \(D_L\) is called a \textit{discretization} of \(D\).
\end{definition}

To build our approximation $G_\theta$, we need to cast the problem to an optimisation problem, where we can control the error of the approximation (on average) with respect to the probability measure\note{$\| G^\dagger - G_\theta \|_{L^2_\mu (\mathcal{A}; \mathcal{U})}$ is the $L^2$ norm over the probability space $\mathcal{A}$ with values in the Banach space $\mathcal{U}$.}
\begin{equation}
    \| G^\dagger - G_\theta \|_{L^2_\mu (\mathcal{A}; \mathcal{U})}^2 = \mathbb{E}_{a\sim \mu} \left[   \| G^\dagger - G_\theta \|^2_{\mathcal{U}} \right] = \int_{\mathcal{A}} \| G^\dagger[a] - G_\theta[a] \|^2_{\mathcal{U}} \, \dd \mu(a) ,
\end{equation}
which, using the Monte Carlo integration approach, give rise to the following minimisation problem
\begin{equation}
    \theta^\dagger =   \underset{\theta \in \Theta}{ \operatorname{argmin} } \, \mathbb{E}_{a\sim \mu} \left[   \| G^\dagger - G_\theta \|^2_{\mathcal{U}} \right] \stackrel{  \text{\tiny{M.C.I.}}}{\approx}  \underset{\theta \in \Theta}{ \operatorname{argmin} } \frac{1}{N} \sum_{i=1}^N \left\|  u^{(i)} - G_\theta [a^{(i)}]\right\|^2_{\mathcal{U}} \,.
\end{equation}

Now, since we are treating \textit{functions}, and since we have defined a way to \textit{discrete refine} the domains where these functions are defined on, we can use the \textit{pointwise evaluation} of these functions at a set of $L$ points, giving rise to the (discrete) dataset
\[
    \{ \left(x_\ell, a(x_\ell)  \right) \}_{\ell=1,\dots, L},
\]
which may be viewed as a vector in $(\RR^{d} \times \RR^{m})^L = \RR^{d\cdot L} \times \RR^{m \cdot L}$. The increase of $L$ to obtain a denser domain is often called a \textit{mesh refinement}. 

\noindent We have all the needed ingredients to define the optimisation problem:

\begin{definition}[Discretised Uniform Risk]
    Suppose \(\AA\) is a Banach space of \(\RR^m\)-valued functions on the domain \(D \subset \RR^d\). Let \(\GG : \AAA \to \UU\) be an operator, \(D_L\) be an \(L\)-point discretization of \(D\), and \(\hat{\GG} : \RR^{Ld} \times \RR^{Lm} \to \UU\) some map. For any \(K \subset \AAA\) compact, we define the \textit{discretized uniform risk} as
    \[
    R_K (\GG,\hat{\GG},D_L) = \sup_{a \in K} \left\|\hat{\GG}(D_L,a|_{D_L}) - \GG  (a) \right\|_{\UU}.
    \]
\end{definition}

\subsection{Neural Operators}
With these concept in mind, we can define the Neural Operator as composed by the following operations:
\begin{itemize}
    \item[1.] \textbf{Lifting:}  Using a pointwise function \(\RR^{d_a} \to \RR^{d_{v_0}}\), map the input $\{a: D \to \RR^{d_a}\} \mapsto \{v_0: D \to \RR^{d_{v_0}}\}$ to its first hidden representation. Usually, we choose \(d_{v_0} > d_a\) and hence this is a lifting operation performed by \textit{a fully local operator}.
    \item[2.] \textbf{Iterative Kernel Integration}: For \(t=0,\dots,T-1\), map each hidden representation to the next\note{$D_t$ represents the domain at layer $t$, and in practice $D_t = D$ for all $t$ in most architectures (except for multi-scale NOs).} $\{v_t: D_t \to \RR^{d_{v_t}}\} \mapsto \{v_{t+1}: D_{t+1} \to \RR^{d_{v_{t+1}}}\}$ via the action of the sum of a local linear operator, a non-local integral kernel operator, and a bias function, composing the  sum with a fixed, pointwise nonlinearity. Here we set \(D_0 = D\) and \(D_T = D'\) and impose that \(D_t \subset \RR^{d_t}\) is a bounded domain.
    \item[3.] \textbf{Projection}: Using a pointwise function \(\RR^{d_{v_T}} \to \RR^{d_u}\), map the last hidden representation $\{v_T: D' \to \RR^{d_{v_T}}\} \mapsto \{u: D' \to \RR^{d_u}\}$ to the output function. Analogously to the first step, we usually pick \(d_{v_T} > d_u\) and hence this is a projection step performed by \textit{a fully local operator}.
\end{itemize}

\begin{figure}[t]
    \centering
    \includegraphics[width=0.95\linewidth]{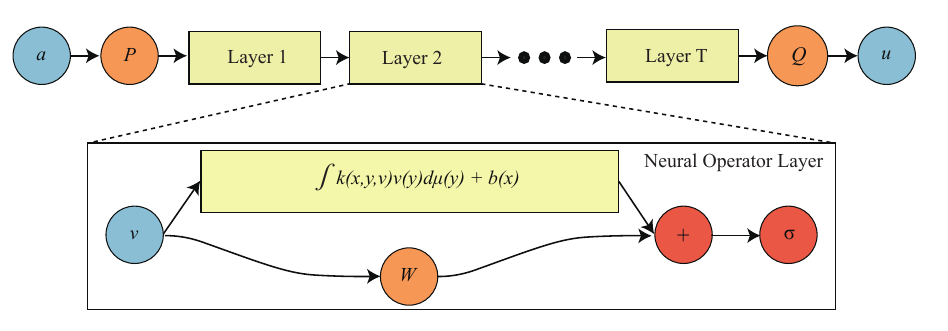}
    \caption{Visual representation of the architecture of a generic Neural Operator. Adapted from \cite{kovachki2023neural}.}
    \label{fig:5:3:1:NeurOpSchema}
\end{figure}

The outlined structure mimics that of a finite dimensional neural network where hidden representations are successively mapped to produce the final output. In particular, we have 
\begin{equation}
    \cG_{\theta} \coloneqq \cQ \circ \sigma_T(W_{T-1} + \cK_{T-1} + b_{T-1}) \circ \cdots \circ \sigma_1(W_0 + \cK_0 + b_0) \circ \cP
\end{equation}
where
\begin{itemize}
    \item \(\cP: \RR^{d_a} \to \RR^{d_{v_0}}\), \(\cQ: \RR^{d_{v_T}} \to \RR^{d_u}\) are the local lifting and projection mappings respectively;
    \item \(W_t \in \RR^{d_{v_{t+1}} \times d_{v_t}}\) are local linear operators (i.e., matrices), \(b_t : D_{t+1} \to \RR^{d_{v_{t+1}}}\) are bias functions;
    \item \(\sigma_t\) are fixed activation functions acting locally as maps \(\RR^{v_{t+1}} \to \RR^{v_{t+1}}\) in each layer;
    \item crucially, \(\cK_t: \{v_t: D_t \to \RR^{d_{v_t}}\} \to \{v_{t+1}: D_{t+1} \to \RR^{d_{v_{t+1}}}\}\) are the \textit{integral kernel operators};
\end{itemize}

The output dimensions \(d_{v_0},\dots,d_{v_T}\) as well as the input dimensions \(d_1,\dots,d_{T-1}\) and domains of definition \(D_1,\dots,D_{T-1}\) are \textit{hyperparameters} of the architecture. 

Notice that, by \textit{local maps}, here we mean that the action is pointwise. In particular, for the \textit{lifting} and \textit{projection} maps, we have \((\cP(a))(x) = \cP(a(x))\) for any \(x \in D\) and  \((\cQ(v_T))(x) = \cQ(v_T(x))\) for any \(x \in D'\) and similarly, for the activation,
\((\sigma(v_{t+1}))(x) = \sigma(v_{t+1}(x))\) for any \(x \in D_{t+1}\). See \cref{fig:5:3:1:NeurOpSchema} for a visual representation of a Neural Operator. 

The crucial difference between this Neural Operator architecture and a standard feed-forward neural network is that all operations are directly defined in function space, and therefore do \textit{not} depend on any discretization of the data.

Intuitively, the lifting step locally maps the data to a space where the non-local part of \(\GG^\dagger\) is easier to capture. Each integral kernel operator is the function space analogue of the weight matrix in a standard feed-forward network, since they are infinite-dimensional linear operators mapping one function space to another. Furthermore the biases, which are normally vectors, are uplifted to functions and, using intuition from the ResNet architecture, there is also a local linear operator acting on the output of the previous layer before applying the nonlinearity. The final projection step simply gets us back to the space of output function.  

But what are those \textit{integral kernel operators}?

In literature we may identify three main versions of the integral kernel operator $\cK_t$. 

\paragraph{First integral operator family (basic Neural Operator): } let \(\kappa^{(t)} \in C(D_{t+1} \times D_t; \RR^{d_{v_{t+1}} \times d_{v_t}})\) and let \(\nu_t\) be a Borel measure on \(D_t\). Then we define \(\cK_t\) by
\begin{equation}
    \label{eq:5:3:1:IntOp1}
    (\text{F1}) \qquad (\cK_t(v_t))(x) = \int_{D_t} \kappa^{(t)} (x,y) v_t(y) \: \text{d}\nu_t(y) \,,
    \qquad \forall x \in D_{t+1}.
\end{equation}
This is the easiest family which can be created, with the kernel function $\kappa$ be dependent only on the layer-input coordinates. By relaxing this, and allowing for multiple dependences, we obtain the other two families.

\paragraph{Second integral operator family: } let \(\kappa^{(t)} \in C(D_{t+1} \times D_t \times \RR^{d_a} \times \RR^{d_a}; \RR^{d_{v_{t+1}} \times d_{v_t}})\). Then we define \(\cK_t\) by
\begin{equation}
    \label{eq:5:3:1:IntOp2}
    (\text{F2}) \quad (\cK_t(v_t))(x) = \int_{D_t} \kappa^{(t)} \left(x,y; a\left(\Pi_{t+1}^D(x)\right),a\left(\Pi_t^D(y)\right)  \right) v_t(y) \: \text{d}\nu_t(y) \,,
    \quad \forall x \in D_{t+1},
\end{equation}
where \(\Pi_t^{D} : D_t \to D\) are fixed mappings from the layer-input domain to the $\AAA$ functions domain. Notice that, now, $\kappa$ depends on the layer-input coordinates and, thanks to the custom, fixed map $\Pi_t^D$, also on the Neural Operator input, $a$. 

Usually, this family is numerically more performant with respect to the first family. A way to look at it is that, if we think to \cref{eq:5:3:1:IntOp2} as a discrete time dynamical system, then the input \(a \in \A\) only enters through the "initial condition"; hence its influence to the Neural Operator forward process diminishes layer by layer. By directly putting in an \(a\)-dependence into the kernel, its influence in the entire architecture is ensured.

\paragraph{Third integral operator family - Transformers: }
let \(\kappa^{(t)} \in C(D_{t+1} \times D_t \times \RR^{d_{v_t}} \times \RR^{d_{v_t}}; \RR^{d_{v_{t+1}} \times d_{v_t}})\). Then define \(\cK_t\) by
\begin{equation}
    \label{eq:5:3:1:IntOp3}
     (\text{F3}) \quad (\cK_t(v_t))(x) = \int_{D_t} \kappa^{(t)} \left(x,y; v_t\left(\Pi_t(x)\right)),v_t(y)\right) \, v_t(y) \, \text{d}\nu_t(y)\,,
    \quad \forall x \in D_{t+1}.
\end{equation}
where \(\Pi_t : D_{t+1} \to D_t\) are, again, fixed mappings. 

\begin{remark}
    Family F1 uses a fixed kernel depending only on spatial coordinates. Family F2 incorporates the input function $a$ into the kernel, ensuring its influence persists across layers. Family F3 introduces nonlinearity by making the kernel depend on the hidden state $v_t$, which, we will see, generalises the self-attention mechanism.
\end{remark}

Notice that, in contrast to previous families, the integral operator here is \textit{non-linear}, since the kernel $\kappa$ depends on the input function \(v_t\). With this definition, and a particular choice of kernel \(\kappa_t\) and measure \(\nu_t\), we it is possible to show that neural operators are a continuous input/output space generalization of the  transformer architecture \cite{vaswani2017attention}. We adapt the following proposition from \cite{kovachki2023neural, calvello2025continuum}.

\begin{proposition}[Transformers are Neural Operators]
    The Attention Mechanism in transformer models is a special case of Neural Operator layer.
\end{proposition}

\begin{proof}
    Let us start by writing down the single-layer representation of a Neural Operator with \cref{eq:5:3:1:IntOp3} integral operator, with a simplified setting, i.e., we pick the kernel to \textit{not} explicitly depend on the spatial variables, but only on the input pair :
    \[
    u(x) = \sigma \left( 
        v(x) + \int_D \kappa_v ( v(x), v(y)) \, v(y) \, \dd y,
    \right) \quad \forall x\in D.
    \]
    Here, \(v: D \to \RR^n \) is the input function to the layer, and \(u: D \to \RR^n\) the output function. \(\kappa_v\) indicates that the kernel depends on the entirety of the function \(v\) as well as on its pointwise values $v(x)$ and $v(y)$.  
    
    We can now make a choice, and pick a specific form for kernel. In particular, we assume \(\kappa_v : \RR^n \times \RR^n \to \RR^{n \times n}\) does not explicitly depend on the spatial variables \((x,y)\) but only on the input pair \((v(x),v(y))\), and 
    \[
    \kappa_v(v(x),v(y)) = g_v(v(x),v(y)) \; R , 
    \]
    where \(R \in \RR^{n \times n}\) is a matrix of learnable parameters, and \(g_v: \RR^n \times \RR^n \to \RR\) is defined as
    \[
        g_v(v(x),v(y)) = \left ( \int_D \text{exp}\left( \frac{\langle A v(s), Bv(y) \rangle}{\sqrt{m}} \right ) \: \text{d}s \right )^{-1} \text{exp} \left ( \frac{\langle A v(x), Bv(y) \rangle}{\sqrt{m}} \right ),
    \]
    where $s$ is a dummy integration variable. 
    
    Here \(A,B \in \RR^{m \times n}\) are again matrices of learnable parameters, \(m \in \N\) is a hyperparameter, and \(\langle \cdot, \cdot \rangle\) is the Euclidean inner-product on \(\RR^m\).

    From here, it is easy to see that
    \begin{equation}
        u(x) = \sigma \left ( v(x) + \int_D \frac{\text{exp} \left ( \frac{\langle A v(x), Bv(y) \rangle}{\sqrt{m}} \right )}{\int_D \text{exp}\left( \frac{\langle A v(s), Bv(y) \rangle}{\sqrt{m}} \right ) \: \text{d}s} \; R \, v(y) \, \text{d}y \right ), \quad \forall x \in D.
    \end{equation}
    This equation can be seen as a \textit{continuum limit} of the single-head self-attention (plus the skip connection). Indeed, Monte Carlo integration of $\int_D$ gives
    \[
    \int_D \text{exp}\left( \frac{\langle A v(s), Bv(y) \rangle}{\sqrt{m}} \right ) \: \text{d}s \approx \frac{|D|}{k} \sum_{l=1}^k \text{exp} \left ( \frac{\langle A v_l, B v(y) \rangle}{\sqrt{m}} \right ),
    \]
    where \(\{x_1,\dots,x_k\} \subset D\) is a uniformly-sampled, \(k\)-point discretization of \(D\) and  \(v_j = v(x_j) \in \RR^n\) and \(u_j = u(x_j) \in \RR^n\) for \(j=1,\dots,k\), the consequent discretisations. Plugging it in, and re-doing the Monte Carlo integrations, gives
    \begin{equation}
    u_j = \sigma \left ( v_j + \sum_{q=1}^k \frac{\text{exp} \left ( \frac{\langle A v_j, Bv_q \rangle}{\sqrt{m}} \right )}{\sum_{l=1}^k \text{exp} \left ( \frac{\langle A v_l, B v_q \rangle}{\sqrt{m}} \right )} \, R \, v_q \right ), \quad j=1,\dots,k.
    \end{equation}
    Stated differently, and using
    \[
    \text{softmax}(x)_i = \frac{\ee^{x_i}}{\sum_i \ee^{x_i}} \,,
    \]
    we have
    \begin{equation}
        u_j = \sigma \left ( v_j + \sum_{q=1}^k 
        \text{softmax}\left(
        \frac{
            \textcolor{blue}{ v_q^T B^T}  \; 
            \textcolor{red}{ A v_j }
        }{\sqrt{m}}
        \right)
        \textcolor{green}{R \, v_q} \right ), \quad j=1,\dots,k,
    \end{equation}
    i.e., 
    if we pass to the vectorised form,
    \[
    X : X_j = v_j,\quad  \textcolor{red}{K}\defeq X B \quad \textcolor{blue}{Q}\defeq X A \quad \textcolor{green}{V}\defeq  X R,
    \]
    we have the standard Single-Head Attention (SHA) \cite{vaswani2017attention, scardapane2024alice}
    \begin{equation}
         U = \sigma \Bigg(
         X +
         \underbrace{
              \text{softmax}\left( \frac{
              \textcolor{blue}{Q} \, \textcolor{red}{K}^T
              }{\sqrt{m}} \right) \, \textcolor{green}{V}
         }_{\text{SHA}(X)}
         \Bigg) .
    \end{equation}
\end{proof}

\subsection{Universal Approximation Theorems for Neural Operators}

In \cite{kovachki2023neural}, authors standardised and proved a set of universal approximation theorems for Neural Operators. Here, we briefly reports those results, and we refer the interested reader to the original work for the proofs. 

In order to report the universal approximation theorems for neural operators, we first need to fix the notation. First, we can write the hidden single-layer update as 
\[
 v_{t+1} (x) = \sigma \left( W_t v_t \left(\Pi_t(x)\right) + \int_{D_t} \kappa^{(t)} (x,y) \, v_t (y) \, \dd \nu_t (y)   + b_t(x)   \right), 
\]
valid $\forall x \in D_{t+1}$, and where $\Pi_t: D_{t+1} \to D_t$ are fixed mapping which, since we often consider functions on the same domain, can usually be taken to be the identity. So, to mimic the Cybenko's theorem, we can consider the following two-layer neural operator architecture:
\begin{equation}
    \label{eq:5:3:2:singlehiddenlayer}
    \cG_\theta[a](x) = \cQ \left (\int_D \kappa^{(1)}(x,y) \sigma \left ( W_0 \cP(a(y)) + \int_D \kappa^{(0)}(y,z) \cP (a(z)) \: \text{d}z + b_0(y) \right )  \text{d}y \right) ,
\end{equation}
for any \(x \in D'\). We can now standardise and define the main ingredients of Neural Operators:
\begin{itemize}
    \item[(a)] Neural Networks;
    \item[(b)] Activation functions;
    \item[(c)] Kernel Integral Operators;
\end{itemize}

\paragraph{Neural Networks ($\NN_n$):} for any \(n \in \N\) and \(\sigma :\RR \to \RR\), we define the set of real-valued \(n\)-layer neural networks
on \(\RR^d\) by
\begin{align*}
\NN_n (\sigma;\RR^d) \coloneqq \{ f : \RR^d \to \RR : \: &f(x) = W_n \sigma ( \dots W_1 \sigma (W_0 x + b_0) + b_1 \dots ) + b_n, \\
&W_0 \in \RR^{d_0 \times d}, W_1 \in \RR^{d_1 \times d_0}, \dots, W_n \in \RR^{1 \times d_{n-1}}, \\
&b_0 \in \RR^{d_0}, b_1 \in \RR^{d_1}, \dots, b_n \in \RR, \: d_0,d_1,\dots,d_{n-1} \in \N\}.
\end{align*}
We define the set of \(\RR^{d'}\)-valued neural networks simply by stacking real-valued networks
\[
\NN_n(\sigma;\RR^d, \RR^{d'}) \coloneqq \{f : \RR^d \to \RR^{d'} : f(x) = \bigl ( f_1(x), \dots, f_{d'}(x) \bigl), \: f_1,\dots,f_{d'} \in \NN_n(\sigma;\RR^d)\}.
\]

\paragraph{Activation functions ($\mathsf{A}_m $): } For any \(m \in \N_0\), we define the set of allowable activation functions as the continuous \(\RR \to \RR\) maps which make
neural networks dense in \(C^m(\RR^d)\) on compact set at any fixed depth,
\[
\mathsf{A}_m \coloneqq \{\sigma \in C(\RR) : \exists n \in \N \text{ s.t. } \NN_n(\sigma;\RR^d) \text{ is dense in } C^m(K) \;\; \forall K \subset \RR^d \text{ compact}\}.
\]
Furthermore, we define the set of linearly bounded activations as
\[\mathsf{A}^{\text{L}}_m \coloneqq \left \{ \sigma \in \mathsf{A}_m : \sigma \text{ is Borel measurable }, \:\: \sup_{x \in \RR} \frac{|\sigma(x)|}{1 + |x|} < \infty \right \},\]
noting that any globally Lipschitz, non-polynomial, \(C^m\)-function is contained in \(\mathsf{A}^{\text{L}}_m\). This definition is quite relevant since most activation functions used in practice fall within this class, for example, \(\text{ReLU} \in \mathsf{A}^{\text{L}}_0\), \(\text{ELU} \in \mathsf{A}^{\text{L}}_1\)
while \(\tanh, \text{sigmoid} \in \mathsf{A}^{\text{L}}_m\) for any \(m \in \mathbb{N}_0\). 

Finally, saying that  the diameter of any set \(S \subseteq \RR^d\) is defined as, for
$|\cdot|_2$ the Euclidean norm on $\RR^d$,
\[
    \text{diam}_2(S) \coloneqq \sup_{x,y \in S} |x - y|_2, 
\]

\begin{definition}[Bounded Activation]
    We denote by \(\mathsf{BA}\) the set of activations
    \(\sigma\in\mathsf A_0\) with the following property\note{The corresponding Definition 7 in \cite{kovachki2023neural}
    appears to contain a typographical inconsistency: it uses
    \(C\ge\operatorname{diam}(K)\), which is insufficient for compact sets
    translated far from the origin, and writes an output dimension \(d'\) although
    the identity approximation requires input and output dimensions to coincide.
    The formulation above follows the clipping property used in the cited
    construction.}. 
    
    For every compact set \(K\subset\RR^d\), every \(\epsilon>0\), and every
    constant
    \[
        C>\sup_{x\in K}|x|_2,
    \]
    there exist \(n\in\N\) and a neural network
    \[
        f\in\NN_n(\sigma;\RR^d,\RR^d)
    \]
    such that
    \begin{align*}
        |f(x)-x|_2 &\le\epsilon,
        &&\forall x\in K,\\
        |f(x)|_2 &\le C,
        &&\forall x\in\RR^d.
    \end{align*}
\end{definition}

\paragraph{Kernel Integral Operators ($\mathsf{IO}$):} Let \(D \subset \RR^d\) be a domain.
For any \(\sigma \in \mathsf{A}_0\), we define the set of affine kernel integral operators by 
\begin{align*}
\mathsf{IO}(\sigma;D,\RR^{d_1},\RR^{d_2}) = \{f \mapsto \int_D \kappa(\cdot, y) f(y) \dd y + b : \: &\kappa \in \NN_{n_1}(\sigma;\RR^{d} \times \RR^{d},\RR^{d_2 \times d_1}), \\
&b \in \NN_{n_2}(\sigma;\RR^{d},\RR^{d_2}), \: n_1, n_2 \in \mathbb{N}\},
\end{align*}
for any \(d_1,d_2 \in \mathbb{N}\).

Finally, we have
\begin{definition}[$n$-layer neural operator]
    For any \(n \in \mathbb{N}_{\geq 2}\), \(d_a,d_u \in \mathbb{N}\), \(D \subset \RR^d\), \(D' \subset \RR^{d'}\) domains, and \(\sigma_1 \in \mathsf{A}^{\text{L}}_0\), \(\sigma_2, \sigma_3 \in \mathsf{A}_0\), 
    we define the set of \(n\)-layer neural operators by
    \begin{align*}
    \mathsf{NO}_n(\sigma_1,\sigma_2,\sigma_3; & D,D',\RR^{d_a},\RR^{d_u}) = \{ \\
    &f \mapsto \int_{D} \kappa_n (\cdot, y) \bigl ( S_{n-1} \sigma_1 ( \dots S_2 \sigma_1(S_1(S_0 f)) \dots) \bigl )(y) \dd y : \\
    &S_0 \in \mathsf{IO}(\sigma_2,D;\RR^{d_a},\RR^{d_1}), \dots S_{n-1} \in \mathsf{IO}(\sigma_2,D;\RR^{d_{n-1}},\RR^{d_n}), \\
    &\kappa_n \in \NN_l (\sigma_3; \RR^{d'} \times \RR^{d}, \RR^{d_u \times d_n}), \: d_1,\dots,d_n, l \in \mathbb{N}\}.
    \end{align*}
\end{definition}

With this  set of definitions, it is possible to state the two main results of the approximation theory for neural operators: \textit{discretisation invariance} and \textit{universal approximation theorem}.

\subsubsection{Discretisation Invariance}
First, let us state the following definition:
\begin{definition}[Discretisation invariance]
    Let \(\Theta \subseteq \RR^p\) be a finite dimensional parameter space and \(\GG : \AAA \times \Theta \to \UU\) a map representing a parametric class of operators with parameters \(\theta \in \Theta\). Given a discrete refinement \((D_n)_{n=1}^\infty\) of the domain \(D \subset \RR^d\), we say \(\GG\) is \textit{discretization-invariant} if there exists a sequence of maps \(\hat{\GG}_1, \hat{\GG}_2, \dots\) where \(\hat{\GG}_L : \RR^{Ld} \times \RR^{Lm} \times \Theta \to \UU\) such that, for any \(\theta \in \Theta\) and any compact set \(K \subset \AAA\),
    \[
    \lim_{L \to \infty} R_K\left(
        \GG(\cdot,\theta), \hat{\GG}_L(\cdot,\cdot,\theta),D_L
    \right) = 0.
    \]
\end{definition}
\noindent From here, it is possible to state the theorem
\begin{theorem}[Discretisation invariance property of Neural Operators]
    Let \(D \subset \RR^{d}\) and \(D' \subset \RR^{d'}\) be two domains for some \(d, d' \in \mathbb{N}\). Let
    \(\cA\) and \(\cU\) be real-valued Banach function spaces on \(D\) and \(D'\) respectively. Suppose that
    \(\cA\) and \(\cU\) can be continuously embedded in \(C(\bar{D})\) and \(C(\bar{D'})\) respectively and that
    \(\sigma_1, \sigma_2, \sigma_3 \in C(\RR)\). Then,
    for any \(n \in \mathbb{N}\), the set of neural operators \(\mathsf{NO}_n(\sigma_1,\sigma_2,\sigma_3;D,D')\) whose elements are viewed 
    as maps \(\cA \to \cU\) is discretization-invariant. 
\end{theorem}
The proof builds a sequence of finite dimensional maps which approximate the neural operator by Riemann sums and shows uniform convergence of the error over compact sets of \(\cA\).  See Appendix E of \cite{kovachki2023neural}.

\paragraph{Why do we care about Discretisation invariance? Zero-shot super-resolution}

\begin{figure}[t]
    \centering
    \includegraphics[width=0.95\linewidth]{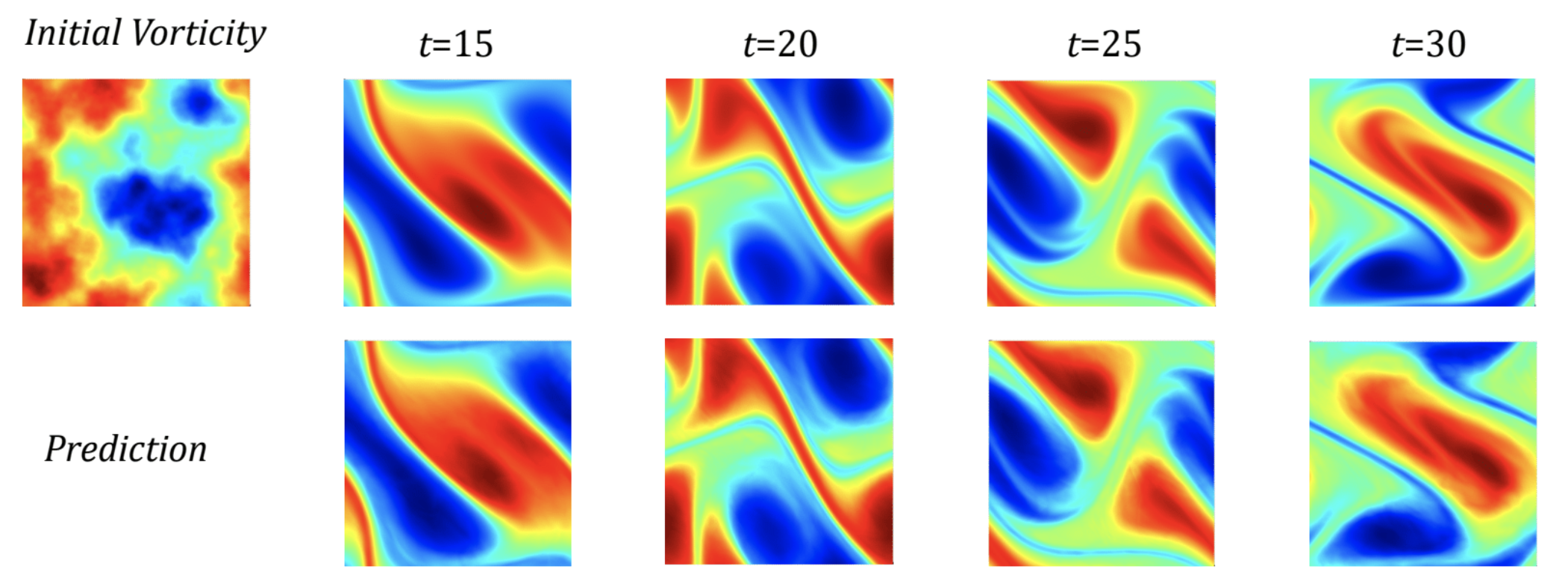}
    \caption{
     Vorticity field of the solution to the two-dimensional Navier-Stokes equation with viscosity $\nu =10^{-4}$. 
    (Re$\approx200$); Ground truth on top and prediction on bottom. The model is trained on data that is discretized on a uniform $64\times64$ spatial grid and on a $20$-point uniform temporal grid. The model is evaluated with a different initial condition that is discretized on a uniform $256\times256$ spatial grid and a $80$-point uniform temporal grid. 
     From \cite{kovachki2023neural}.}
    \label{fig:5:3:2:1:SuperRes}
\end{figure}

\phantom{.} \\
The crucial results from here is that, any Neural Operator, being discretisation invariant, can be trained on a set of numerical results computed on lower-resolution grid, and evaluated at higher resolution, without seeing any higher resolution data! This is the \textit{zero-shot super-resolution}. An example is presented in \cref{fig:5:3:2:1:SuperRes}, where the Neural Operator is trained on 64x64 data, and evaluated on 256x256. On the top row are presented the real data, and on the bottom row the prediction, ad different time steps. The single image on the left is the initial vorticity. The system resolved is the two-dimensional Navier-Stokes equation for a viscous, incompressible fluid:
\begin{align}
\begin{split}
\partial_t u(x,t) + u(x,t) \cdot \nabla u(x,t) + \nabla p(x,t) &= \nu \Delta u(x,t) + f(x), \quad x \in \mathbb{T}^2, t \in (0,\infty)  \\
\nabla \cdot u(x,t) &= 0, \quad \qquad \qquad \qquad \:\:\: x \in \mathbb{T}^2, t \in [0,\infty)  \\
u(x,0) &= u_0(x), \quad \qquad \qquad \:\:\: x \in \mathbb{T}^2 
\end{split}
\end{align}
where \(\mathbb{T}^2\) is the unit torus i.e. \([0,1]^2\) equipped with periodic boundary conditions, and \(\nu \in \RR_+\) is a fixed viscosity.
Here $u : \mathbb{T}^2 \times \RR_+ \to \RR^2$ is the velocity field,  
$p : \mathbb{T}^2 \times \RR_+ \to \RR$ is the pressure field, and
 \(f: \mathbb{T}^2 \to \RR^2\) is a fixed forcing function.

\subsubsection{Universal Approximation Theorem}

Let \(\cA\) and \(\cU\) be Banach function spaces on the domains \(D \subset \RR^d\) and \(D' \subset \RR^{d'}\), respectively. We are interested in the approximation of non-linear operators \(\cG^\dagger : \cA \to \cU\)
by neural operators. We will make the following assumptions on the spaces \(\cA\) and \(\cU\):
\begin{assumption}
    \label{assump:5:3:2:2:input}
    Let \(D \subset \RR^d\) be a Lipschitz domain for some \(d \in \mathbb{N}\).
    One of the following holds
    \begin{enumerate}
    	\item \(\cA = L^{p_1} (D)\) for some \(1 \leq p_1 < \infty\).
    	\item \(\cA = W^{m_1,p_1}(D)\) for some \(1 \leq p_1 < \infty\) and \(m_1 \in \mathbb{N}\),
    	\item \(\cA = C (\bar{D})\).
    \end{enumerate}
\end{assumption}

\begin{assumption}
\label{assump:5:3:2:2:output}
Let \(D' \subset \RR^{d'}\) be a Lipschitz domain for some \(d' \in \mathbb{N}\).
One of the following holds
\begin{enumerate}
	\item \(\cU = L^{p_2} (D')\) for some \(1 \leq p_2 < \infty\),
	and \(m_2 = 0\),
	\item \(\cU = W^{m_2,p_2}(D')\) for some \(1 \leq p_2 < \infty\) and \(m_2 \in \N\),
	\item \(\cU = C^{m_2}(\bar{D}')\) and \(m_2 \in \N_0\).  
\end{enumerate}
\end{assumption}

From here, we have

\begin{theorem}[Universal Approximation Theorem for Neural Operators]
Let \cref{assump:5:3:2:2:input} and \cref{assump:5:3:2:2:output} hold and suppose \(\cG^\dagger : \cA \to \cU\) is continuous.
Let \(\sigma_1 \in \mathsf{A}^{\emph{\text{L}}}_0\), \(\sigma_2 \in \mathsf{A}_0\), and \(\sigma_3 \in \mathsf{A}_{m_2}\).
Then for any compact set \(K \subset \cA\) and \(0 < \epsilon \leq 1\), there exists a number \(N \in \mathbb{N}\) and a neural operator 
\(\cG \in \mathsf{NO}_{N}(\sigma_1,\sigma_2,\sigma_3;D,D')\) such that
\[
    \sup_{a \in K} \|\cG^\dagger (a) - \cG (a) \|_{\cU} \leq \epsilon.
\]
Furthermore, if \(\cU\) is a Hilbert space and \(\sigma_1 \in \mathsf{BA}\) and, for some \(M > 0\), we have that \(\|\cG^\dagger (a)\|_{\cU} \leq M\) for all 
\(a \in \cA\) then \(\cG\) can be chosen so that
\[
    \|\cG(a)\|_{\cU} \leq 4M, \qquad \forall a \in \cA.
\]
\end{theorem}

\section{Learning Operators, Part III: Neural Operators - Architectures}

\subsection{Fourier Neural Operator} \label{subsec:5:4:1:FNO}

The first example of a Neural Operator architecture is the Fourier Neural Operator (FNO) \cite{li2020fourier}.

\begin{figure}
    \centering
    \includegraphics[width=0.95\linewidth]{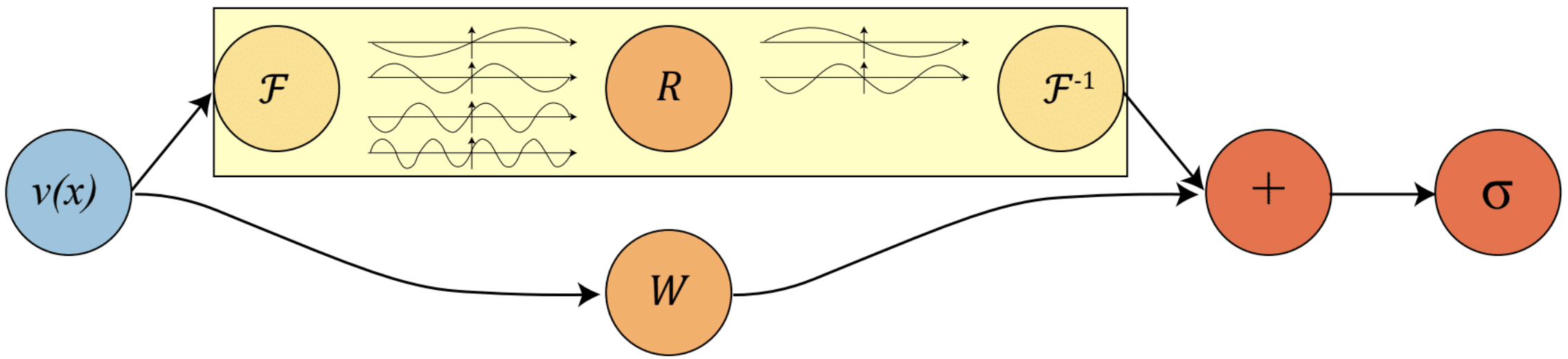}
    \caption{
    \textbf{(a) The full architecture of neural operator}: start from input $a$. 1. Lift to a higher dimension channel space by a neural network $\cP$. 2. Apply four layers of integral operators and activation functions. 3. Project back to the target dimension by a neural network $\cQ$. Output $u$.
    \textbf{(b) Fourier layers}: Start from input $v$. On top: apply the Fourier transform $\cG$; a linear transform $R$ on the lower Fourier modes and filters out the higher modes; then apply the inverse Fourier transform $\cG^{-1}$. On the bottom: apply a local linear transform $W$. 
    From \cite{li2020fourier}.
    }
    \label{fig:5:4:1:fourier_layer_fno}
\end{figure}

The main ingredient of the Fourier Neural Operator is the Fourier representation of the kernel integral operator. Indeed, we have seen in the previous subsection that the Neural Operator layer is written as
\[
 v_{t+1} (x) = \sigma \left( W_t v_t \left(\Pi_t(x)\right) + \int_{D_t} \kappa^{(t)} (x,y) \, v_t (y) \, \dd \nu_t (y)   + b_t(x)   \right), 
\]
so, now, let us assume that the integral kernel is translation equivariant and can thus be written as
\[
    \kappa^{(t)} (x,y) = \kappa^{(t)} (x-y).
\]
We can now \textit{Fourier Transform it}, and parametrise its Fourier components with $\theta_t$. The main idea of FNO is to exploit this fact by parameterising $\kappa$ directly in Fourier space and using the Fast Fourier Transform (FFT) to efficiently compute the integral. 
Indeed, let \(\FFT\) denote the Fourier transform of a function $f: D \to \RR^{d_v}$ and $\FFT^{-1}$ its inverse; then
\begin{align*}
    (\FFT f)_j(k) = \int_{D} f_j(x) \ee^{- 2i \pi \langle x, k \rangle} \mathrm{d}x, \qquad
    (\FFT^{-1} f)_j(x) = \int_{D} f_j(k) \ee^{2i \pi \langle x, k \rangle} \mathrm{d}k ,
\end{align*}
for $j=1,\dots,d_v$ where \(i = \sqrt{-1}\) is the imaginary unit. By assuming \( \kappa^{(t)} (x,y) = \kappa^{(t)} (x-y)\), and by using the Convolution Theorem,  we find that
\[
\bigl(\cK_\theta v_t\bigr)(x) = \FFT^{-1} \bigl( \FFT(\kappa) \cdot \FFT(v_t) \bigr )(x), \qquad \forall x \in D.
\]

Now, the intuition is to directly  parametrise $\kappa$ in Fourier space;
\begin{definition}[Fourier integral operator $\cK$] 
Define the Fourier integral operator
\begin{equation}
    \bigl(\cK(\theta)v_t\bigr)(x)=   
    \FFT^{-1}\Bigl(R_\theta \cdot (\FFT [v_t]) \Bigr)(x) \quad \forall x \in D 
\end{equation}
where $R_\theta\in \mathbb C^{d_{v}\times d_{v}}$ is the Fourier transform of a periodic function $\kappa: \bar{D} \to \RR^{d_v \times d_v}$, parametrised by \(\theta \in \Theta_\cK\).
An illustration is given in Figure \cref{fig:5:4:1:fourier_layer_fno} (b).
\end{definition}

\paragraph{Invariance to discretization:}
The Fourier layers are discretization-invariant because they can learn from and evaluate functions which are discretized in an arbitrary way. Since parameters are learned directly in Fourier space, resolving the functions in physical space simply amounts to projecting on the basis $\ee^{2\pi i \langle x, k \rangle}$ which are well-defined everywhere on $\RR^d$. This allows us to achieve zero-shot super-resolution, as explained before. 

\paragraph{Quasi-linear complexity:}
The weight tensor $R$ contains $k_{\text{max}} < n$ modes, so the inner multiplication has complexity $O(k_{\text{max}})$. Therefore, the majority of the computational cost lies in computing the Fourier transform $\FFT(v_t)$ and its inverse. 

General Fourier transforms have complexity $O(n^2)$; however, since the series is truncated, the complexity is in fact $O(n k_{\text{max}})$, while the FFT has complexity $O(n \log n)$.

\subsection{Adaptive Fourier Neural Operator} \label{subsec:5:4:2:AFNO}
One crucial aspect of Self-Attention Mechanism \cite{vaswani2017attention}, also in vision tasks \cite{dosovitskiy2020image}, is that it imposes \textit{graph structures}, and uses the similarity among the tokens to capture the long-range dependencies. 

This makes self-attention parameter efficient and adaptive, but it suffers from a quadratic complexity in the sequence size. To achieve efficient mixing with linear complexity, several approximations have been introduced for self-attention. More recently, alternatives have been introduced for self-attention that relax the graph assumption for efficient mixing. Instead, they leverage the \textit{geometric structures} using Fourier transform.

\begin{figure}
    \centering
    \includegraphics[width=0.95\linewidth]{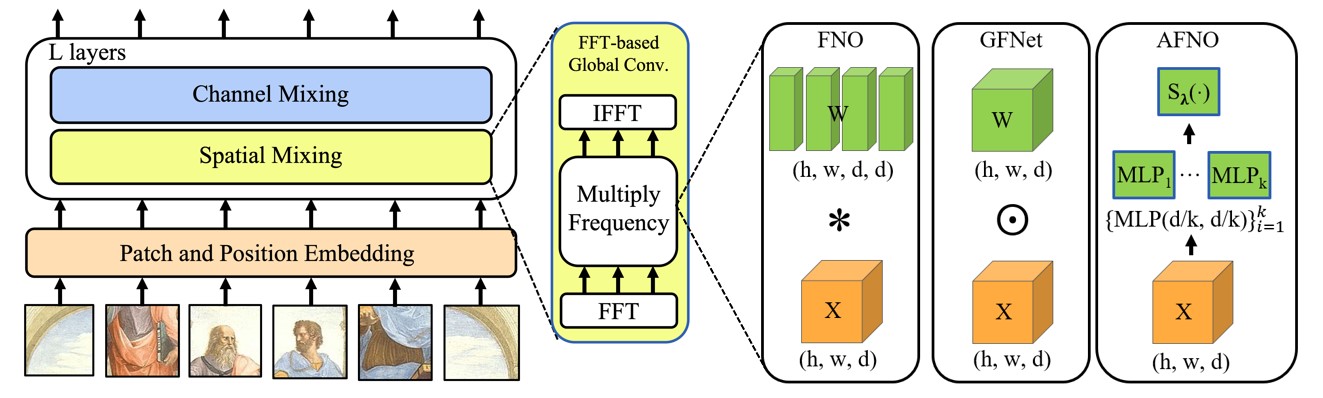}
    \caption{ The multi-layer transformer network with FNO, GFN, and AFNO mixers. GFNet performs element-wise matrix multiplication with separate weights across channels ($k$). FNO performs full matrix multiplication that mixes all the channels. AFNO performs block-wise channel mixing using MLP along with soft-thresholding. The symbols $h$, $w$, $d$, and $k$ refer to the height, width, channel size, and block count, respectively.
    From \cite{guibas2021adaptive}.
    }
    \label{fig:5:4:2:AFNO}
\end{figure}

The idea of~\cite{guibas2021adaptive, guibas2021efficient}  was to blend intuition from  Fourier Neural Operators, by  framing token mixing as operator learning that learns mappings between continuous functions in infinite dimensional spaces. Tokens are treated as
continuous elements in the function space, and  token mixing is modelled as continuous global convolution, which captures global relationships in the geometric space. But, to ease the quadratic complexity in the channel size for channel mixing of standard FNOs,  authors impose a block-diagonal structure on the channel mixing weights. Furthermore, to enhance
generalisation, inspired by sparse regression, they sparsify the frequencies via soft-thresholding. Finally, for parameter efficiency, the MLP layer shares weights across tokens. These are the ingredients of Adaptive Fourier Neural Operator (see \cref{fig:5:4:2:AFNO}).

The first step in the architecture involves dividing the input image into a regular grid with $h \times w$ equal sized patches of size $p\times p$. The parameter $p$ is referred to as the \textit{patch size}, as in Vision Transformers \cite{dosovitskiy2020image}. For simplicity, we consider a single channel image. Each patch is embedded into a token of size   $d$ , the embedding dimension. The patch embedding operation results in a token tensor   $X\sim(h,w,d)$.       
 
The patch size and embedding dimension are user selected parameters. A smaller patch size allows the model to capture fine scale details better while increasing the computational cost of training the model; on the other hand, a higher embedding dimension also increases the parameter count of the model. 

The token tensor is then processed by multiple layers of the transformer architecture performing spatial and channel mixing. The AFNO architecture implements the following operations in each layer.

\begin{table}[t]
\centering
\resizebox{\textwidth}{!}{%
    \renewcommand{\arraystretch}{1.3}
    \begin{tabular}{>{\bfseries}l >{\raggedright\arraybackslash}p{0.4\textwidth} >{\raggedright\arraybackslash}p{0.4\textwidth}}
    \toprule
    \textbf{Step} & \textbf{FNO} & \textbf{AFNO} \\
    \midrule
    (i) Token mixing & 
    \( X_{m,n:d} \mapsto z_{m,n} = \text{DFT}[X_{m,n:d}] ,  \) & 
    \( X_{m,n:d} \mapsto z_{m,n} = \text{DFT}[X_{m,n:d}] , \) \\
    \midrule
    (ii) Block mixing & 
    \( z_{m,n} \mapsto \tilde{z}_{m,n} = W_{m,n|m',n'}z_{m',n'} ,  \) & 
    \( z_{m,n} \mapsto \tilde{z}_{m,n} = \sum_{\ell=1,\dots,k} S_\lambda\left(W_{m,n|m',n'}^{(\ell)}z_{m',n'}^{(\ell)}\right) , \) \\
    \midrule
    (iii) Token unmixing & 
    \( X_{m,n:d} \mapsto y_{m,n} = \text{IDFT}[\tilde{z}_{m,n}] ,  \) & 
    \( X_{m,n:d} \mapsto y_{m,n} = \text{IDFT}[\tilde{z}_{m,n}] ,  \) \\
    \bottomrule
    \end{tabular}
}
\caption{Comparison of the operational steps between FNO and AFNO. The notation \(d\) represents the channel dimension, \(k\) the number of blocks, and \(S_\lambda(\cdot)\) the soft-thresholding operator. In AFNO, the block mixing step incorporates a summation over blocks and the nonlinear shrinkage function.}
\label{tab:5:4:2:AFNOvsFNOsteps}
\end{table}

A crucial element in the AFNO's block mixing step is the application of soft-thresholding and shrinkage. Images are inherently sparse in the Fourier domain, with most of the energy concentrated around low frequency modes. Thus, one can adaptively mask the tokens according to their importance towards the end task. This allows the model to focus its expressivity on representing the important tokens. To sparsify the tokens, instead of linear combination, we use the nonlinear LASSO channel mixing as follows:
\[
\min \|\tilde{z}_{m,n} - W_{m,n}z_{m,n}\|^2 + \lambda \|\tilde{z}_{m,n}\|_1
\]
This can be solved via the soft-thresholding and shrinkage operation
\[
\tilde{z}_{m,n} = S_\lambda(W_{m,n}z_{m,n}),
\]
that is defined as \( S_\lambda(x) = \text{sign}(x)\max\{|x| - \lambda, 0\} \), where \( \lambda \) is a tuning parameter that controls the sparsity. It is also worth noting that the promoted sparsity can also regularize the network and improve the robustness.

\begin{lstlisting}[language=Python, caption={Pseudocode for AFNO with adaptive weight sharing and adaptive masking. Adapted from \cite{guibas2021adaptive}.}, label=lst:5:4:2:afno]
def AFNO(x):
    bias = x
    x = RFFT2(x)                          # (b, h, w, d) -> (b, h, w//2+1, d)
    x = x.reshape(b, h, w//2+1, k, d//k)  # Split into k blocks
    x = BlockMLP(x)                       # Shared MLP per block
    x = x.reshape(b, h, w//2+1, d)        # Merge blocks
    x = SoftShrink(x)                     # Soft-thresholding
    x = IRFFT2(x)                         # (b, h, w//2+1, d) -> (b, h, w, d)
    return x + bias                       # Residual connection

# Tensor shapes
# x : Tensor[b, h, w, d]                    # Input: batch, height, width, channels
# W_1, W_2 : ComplexTensor[k, d//k, d//k]   # Weights per block
# b_1, b_2 : ComplexTensor[k, d//k]         # Biases per block

def BlockMLP(x):
    """Two-layer MLP with ReLU activation, shared across all tokens"""
    x = MatMul(x, W_1) + b_1                 # Linear projection
    x = ReLU(x)                              # Non-linearity
    return MatMul(x, W_2) + b_2              # Second projection
\end{lstlisting}

\begin{lstlisting}[language=Python, caption={Full pytorch code for AFNO layer. Adapted from \cite{guibas2021adaptive}.}, label=lst:5:4:2:afnofull]
import math
import torch
import torch.fft
import torch.nn as nn
import torch.nn.functional as F

class AFNO2D(nn.Module):
    """
    hidden_size: channel dimension size
    num_blocks: how many blocks to use in the block diagonal weight matrices (higher => less complexity but less parameters)
    sparsity_threshold: lambda for softshrink
    hard_thresholding_fraction: how many frequencies you want to completely mask out (lower => hard_thresholding_fraction^2 less FLOPs)
    """
    def __init__(
        self, 
        hidden_size: int, 
        num_blocks : int = 8,
        sparsity_threshold: float = 0.01, 
        hard_thresholding_fraction: float = 1, 
        hidden_size_factor: float = 1
    ):
        super().__init__()
        assert hidden_size % num_blocks == 0, f"hidden_size {hidden_size} should be divisble by num_blocks {num_blocks}"

        self.hidden_size = hidden_size
        self.sparsity_threshold = sparsity_threshold
        self.num_blocks = num_blocks
        self.block_size = self.hidden_size // self.num_blocks
        self.hard_thresholding_fraction = hard_thresholding_fraction
        self.hidden_size_factor = hidden_size_factor
        self.scale = 0.02

        self.w1 = nn.Parameter(self.scale * torch.randn(2, self.num_blocks, self.block_size, self.block_size * self.hidden_size_factor))
        self.b1 = nn.Parameter(self.scale * torch.randn(2, self.num_blocks, self.block_size * self.hidden_size_factor))
        self.w2 = nn.Parameter(self.scale * torch.randn(2, self.num_blocks, self.block_size * self.hidden_size_factor, self.block_size))
        self.b2 = nn.Parameter(self.scale * torch.randn(2, self.num_blocks, self.block_size))

    def forward(self, x):
        bias = x

        dtype = x.dtype
        x = x.float()
        B, C, H, W = x.shape
        N = H*W
        x = x.moveaxis(1, -1) # (B, C, H, W) -> (B, H, W, C) # x = x.reshape(B, H, W, C)
        x = torch.fft.rfft2(x, dim=(1, 2), norm="ortho")
        x = x.reshape(B, x.shape[1], x.shape[2], self.num_blocks, self.block_size)

        o1_real = torch.zeros([B, x.shape[1], x.shape[2], self.num_blocks, self.block_size * self.hidden_size_factor], device=x.device)
        o1_imag = torch.zeros([B, x.shape[1], x.shape[2], self.num_blocks, self.block_size * self.hidden_size_factor], device=x.device)
        o2_real = torch.zeros(x.shape, device=x.device)
        o2_imag = torch.zeros(x.shape, device=x.device)

        total_modes = N // 2 + 1
        kept_modes = int(total_modes * self.hard_thresholding_fraction)

        o1_real[:, :, :kept_modes] = F.relu(
            torch.einsum('...bi,bio->...bo', x[:, :, :kept_modes].real, self.w1[0]) - \
            torch.einsum('...bi,bio->...bo', x[:, :, :kept_modes].imag, self.w1[1]) + \
            self.b1[0]
        )

        o1_imag[:, :, :kept_modes] = F.relu(
            torch.einsum('...bi,bio->...bo', x[:, :, :kept_modes].imag, self.w1[0]) + \
            torch.einsum('...bi,bio->...bo', x[:, :, :kept_modes].real, self.w1[1]) + \
            self.b1[1]
        )

        o2_real[:, :, :kept_modes] = (
            torch.einsum('...bi,bio->...bo', o1_real[:, :, :kept_modes], self.w2[0]) - \
            torch.einsum('...bi,bio->...bo', o1_imag[:, :, :kept_modes], self.w2[1]) + \
            self.b2[0]
        )

        o2_imag[:, :, :kept_modes] = (
            torch.einsum('...bi,bio->...bo', o1_imag[:, :, :kept_modes], self.w2[0]) + \
            torch.einsum('...bi,bio->...bo', o1_real[:, :, :kept_modes], self.w2[1]) + \
            self.b2[1]
        )

        x = torch.stack([o2_real, o2_imag], dim=-1)
        x = F.softshrink(x, lambd=self.sparsity_threshold)
        x = torch.view_as_complex(x)
        x = x.reshape(B, x.shape[1], x.shape[2], C)
        x = torch.fft.irfft2(x, s=(H, W), dim=(1, 2), norm="ortho")
        x = x.moveaxis(-1, 1) # (B, H, W, C) -> (B, C, H, W)
        x = x.type(dtype)
        return x + bias
\end{lstlisting}

\subsection{Physics Informed Neural Operator}  \label{subsec:5:4:3:PINO}

Up to now we have seen Neural Operators for surrogate modelling. The \textit{Physics Informed Neural Operators} (PINO) \cite{li2024physics} approach effectively combines the data-informed supervised learning framework of the Fourier Neural Operator with the physics-informed learning framework. PINO incorporates a PDE loss to the Fourier Neural Operator training. This reduces the amount of data required to train a surrogate model, since the PDE loss constrains the solution space. The PDE loss also enforces physical constraints on the solution computed by a surrogate ML model, making it an attractive option as a verifiable, accurate and interpretable ML surrogate modelling tool.

Let us set up the notation, to align with \cite{li2024physics}. Suppose we have to face the pure-boundary differential problem like 
\begin{equation} 
    \begin{split}
    \mathcal{P}(u,a) &= 0, \qquad \text{in } D \subset \RR^d \\
    u &= g, \qquad \text{in } \partial D
    \end{split}
\end{equation}
where \(D\) is a bounded domain, \(a \in \mathcal{A} \subseteq \mathcal{V}\) is a PDE coefficient/parameter, \(u \in \mathcal{U}\) is the unknown, and \(\mathcal{P}: \mathcal{U} \times \mathcal{A} \to \mathcal{F}\) is a possibly non-linear partial differential operator with \((\mathcal{U}, \mathcal{V}, \mathcal{F})\) a triplet of Banach spaces. Usually, the function $g$ is a fixed boundary condition but can also potentially enter as a parameter.
This formulation gives rise to the solution operator \(\Gtrue: \mathcal{A} \to \mathcal{U}\) defined by \(a \mapsto u\). A prototypical example is the second-order elliptic equation $ \mathcal{P}(u,a) = -\nabla \cdot (a \nabla u) + f$. Another class of problem we may want to tackle is the dynamical problem
\begin{align} 
\begin{split}
    \frac{du}{dt} &= \mathcal{R}(u), \qquad  \text{in } D \times (0,\infty) \\
    u &= g, \qquad \quad \:\:\: \text{in } \partial D \times (0,\infty) \\
    u &= a \qquad \qquad \text{in } \bar{D} \times \{0\}
\end{split}
\end{align}
where \(a = u(0) \in \mathcal{A} \subseteq \mathcal{V}\) is the initial condition, \(u(t) \in \mathcal{U}\) for \(t > 0\) is the unknown, and \(\mathcal{R}\) is a possibly non-linear partial differential operator with \(\mathcal{U}\), and \(\mathcal{V}\) Banach spaces. As before, we take \(g\) to be a known boundary condition. We assume that \(u\) exists and is bounded for all time and for every $u_0 \in \mathcal{U}$.

This formulation gives rise to the solution operator \(\Gtrue : \mathcal{A} \to C \big ( (0,T]; \mathcal{U} \big )\), mapping \(a \mapsto u\).

We have seen that the standard surrogate modelling with neural operators is as follows: given a differential system, and the corresponding solution operator \(\Gtrue\), one can use a neural operator ${\cG_{\theta}}$ with parameters ${\theta}$ as a surrogate model to approximate \(\Gtrue\). 
Usually we assume a dataset $\{a_j, u_j\}_{j=1}^N$ is available, where $\Gtrue(a_j) = u_j$ and $a_j \sim \mu$ are i.i.d.~samples from some distribution \(\mu\) supported on \(\mathcal{A}\). In this case, one can optimize the solution operator by minimizing the empirical data loss on a given data pair
\begin{align}
\mathcal{L}_{\text{data}}(u, \cG_\theta(a)) = \|u - \cG_\theta(a)\|_{\mathcal{U}}^2= \int_D |u(x) - \cG_\theta(a)(x) |^2 \mathrm{d}x
\end{align}

The operator data loss is defined as the average error across all possible inputs
\begin{align}
\begin{split}
\mathcal{J}_{\text{data}}( \cG_\theta) &= \|\Gtrue - \cG_\theta\|^2_{L^2_\mu(\mathcal{A};\mathcal{U})} \\
&= \EE_{a \sim \mu}[ \mathcal{L}_{\text{data}}(a,\theta)] \approx \frac{1}{N} \sum_{j=1}^N \int_D |u_j (x) - {\cG_\theta}(a_j)(x) |^2 \mathrm{d}x.
\end{split}
\end{align}

\noindent We can now define physics-informed loss, e.g. 
\begin{align}
\begin{split}
    \mathcal{L}_{\text{pde}}(u_\theta, a) &= \Big\|\mathcal{P}(a,u_{\theta})\Big\|^2_{L^2(D)} + \alpha \Big\|u_{\theta}|_{\partial D} - g \Big\|^2_{L^2(\partial D)} \\ 
    &= \int_D |\mathcal{P}(u_{\theta},a)(x) |^2\mathrm{d}x  +\alpha\int_{\partial D} |u_{\theta}(x) - g(x)|^2 \mathrm{d}x
\end{split}
\end{align}
for the stationary case, or 
\begin{align}
\begin{split}
\mathcal{L}_{\text{pde}}(a, u_\theta) 
&= \Big\|\frac{du_{\theta}}{dt} - \mathcal{R}(u_{\theta})\Big\|^2_{L^2(T;D)} + \alpha \Big\|u_{\theta}|_{\partial D} - g \Big\|^2_{L^2(T; \partial D)} + \beta \Big\|u_{\theta}|_{t=0} - a \Big\|^2_{L^2(D)} \\
&= \int_{0}^T \int_D |  \frac{du_{\theta}}{dt}(t,x) -\mathcal{R}(u_{\theta})(t,x)|^2 \mathrm{d}x \mathrm{d}t \\
&\:\:\:+ \alpha \int_{0}^T \int_{\partial D} | u_{\theta}(t,x) - g(t,x)|^2 \mathrm{d}x \mathrm{d}t \\
&\:\:\:+ \beta \int_{D} |u_{\theta}(0,x) - a(x)|^2 \mathrm{d}x
\end{split}
\end{align}
for the dynamic case.
\noindent These, for the operator case, can be compactly written as
\begin{align}
\begin{split}
\mathcal{J}_{\text{pde}}( \cG_\theta) = \EE_{a \sim \mu}[ \mathcal{L}_{\text{pde}}(a,  \cG_\theta(a)) ].
\end{split}
\end{align}

A distinctive feature of PINO is that the two losses need not be evaluated at the same resolution. The supervised data may be available only on a relatively coarse discretization, while the PDE residual can be evaluated on a finer grid using the same underlying neural operator.

Schematically,
\[
    \mathcal J(\theta)
    =
    \lambda_{\mathrm{data}}
    \mathcal J_{\mathrm{data}}^{(L_{\mathrm{data}})}(\theta)
    +
    \lambda_{\mathrm{pde}}
    \mathcal J_{\mathrm{pde}}^{(L_{\mathrm{pde}})}(\theta),
\]
with potentially
\[
    L_{\mathrm{pde}}>L_{\mathrm{data}}.
\]

The framework also admits a physics-only regime in which labelled input/output solution pairs are unavailable.

The point is that, in general, it is non-trivial to compute the derivatives $d\cG_\theta (a)/dx$ and $d\cG_\theta (a)/dt$ for model $\cG_\theta$. In the following, we discuss the options the authors \cite{li2024physics}  suggested how to compute these derivatives for Fourier neural operator.

In the FNO framework, the surrogate ML model is given by a the solution operator \( \mathcal{G}_\theta^\dagger \), which maps any given coefficient in the coefficient space \( a \) to the solution \( u \). The FNO is trained in a supervised fashion using training data in the form of input/output pairs \( \{a_j, u_j\}_{j=1}^N \). The training loss for the FNO is given by summing the data loss, \( \mathcal{L}_{\text{data}}(\mathcal{G}_\theta) = \|u - \mathcal{G}_\theta(a)\|^2 \) over all training pairs \( \{a_i, u_i\}_{i=1}^N \).

Indeed, in the PINO framework, the solution operator is optimized also with PDE loss given by \( \mathcal{L}_{\text{pde}}(a, \mathcal{G}_\theta(a)) \) computed over i.i.d. samples \( a_j \) from an appropriate supported distribution in parameter/coefficient space.

In general, the PDE loss involves computing the PDE operator which in turn involves computing the partial derivatives of the Fourier Neural Operator ansatz. In general this is non-trivial. The key set of innovations in the PINO are the various ways to compute the partial derivatives of the operator ansatz. 
%
PINO implements differentiation via three main strategies: (1) Numerical differentiation (FDM/Spectral), which is fast but approximate; (2) Pointwise autograd, which is exact but memory-intensive; (3) Function-wise differentiation in Fourier space, which exploits the spectral structure for exact, efficient gradients.

Let us discuss and compare these three methods; 

\paragraph{Numerical differentiation:} A simple but efficient approach is to use conventional numerical derivatives such as finite difference and Fourier differentiation. These numerical differentiation methods are fast and memory-efficient: given a $n$-points grid, finite difference requires $O(n)$, and the Fourier method requires $O(n\log n)$.

However, the numerical differentiation methods face the same challenges as the corresponding numerical solvers: finite difference methods require a fine-resolution uniform grid; spectral methods require smoothness and uniform grids. Furthermore, these numerical errors on the derivatives will be amplified on the output solution.

\paragraph{Pointwise differentiation:} in analogy with PINN, the most general way to compute the exact derivatives is to use the auto-differentiation library of neural networks (autograd). However, it is not straightforward to write out the solution function in the neural operator which directly outputs the numerical solution $u = \cG_\theta(a)$ on a grid, especially for FNO which uses FFT. To apply autograd, we design a query function $u$ that input $x$ and output $u(x)$.  Recall $\cG_{\theta} \coloneqq \cQ \circ  (W_{L} + \cK_{L}) \circ \cdots \circ \sigma(W_1 + \cK_1) \circ \cP$ and $u = \cG_{\theta}a = \cQ v_L = \cQ (W_{L} + \cK_{L}) v_{L-1} \ldots$. Since $\cQ$ is pointwise, 
\begin{equation}
\label{eq:autograd}
u(x) = \cQ(v_L)(x) = \cQ(v_L(x)) = \cQ \bigl( (W_{L} v_{L-1})(x) + \cK_{L}v_{L-1}(x) \bigr)
\end{equation}
For both the kernel operator and Fourier operator, we either remove the pointwise residual term of the last layer $(W_{L} v_{L-1})(x)$ or define the query function as an interpolation function on $W_{L}$. 

For kernel integral operator, the kernel function can directly take the query points as input. So we can apply auto-differentiation to compute the derivatives
\begin{equation}
    u'(x) = \cQ' \bigl(v_L(x)\bigr) \cdot \sum_{B(x)} {\kappa^{(l)}}' (x,y, v_{L-1}(y))
\end{equation}

Similarly, for the Fourier convolution operator, we need to evaluate the Fourier convolution $\cK_{L}v_{L-1}(x)$ on the query points $x$. It can be done by writing out the output function as  Fourier series composing with $Q$ :
\[
\begin{split}
   u(x) &= \cQ \circ 
\FFT^{-1}\Bigl(R \cdot (\FFT [v_{L-1}])) \Bigr)(x) \\
&= 
\cQ \Bigl(
\frac{1}{k_{max}}\sum_{k=0}^{k_{max}} \big(R_k  (\FFT [v_{L-1}])_k\big) \exp\frac{i 2\pi k}{D}  (x) 
\Bigr) 
\end{split}
\]
Where $ \FFT$ is the discrete Fourier transform. The inverse discrete Fourier transform is the sum of $k_{max}$ Fourier series with the coefficients $\big(R_k  (\FFT [v_{L-1}])_k\big)$ coming from the previous layer. 
\begin{equation}
    u'(x) = \cQ'\bigl(v_L(x)\bigr) \cdot \frac{1}{k_{max}}\sum_{k=0}^{k_{max}} \big(R_k  (\FFT [v_{L-1}])_k\big) \exp'\frac{i 2\pi k}{D}  (x)
\end{equation}
Notice $\exp'\frac{i 2\pi k}{D}  (x) = \frac{i 2\pi k}{D} \exp\frac{i 2\pi k}{D}  (x)$, just as the numerical Fourier method. If the query points $x$ are a uniform grid, the derivative can be efficiently computed with the Fast Fourier transform.

The autograd method is general and exact, however, it is less efficient.
Since the computational graph for autograd scales with the number of parameters, which is often larger than the grid size, numerical differentiation methods are significantly faster.
Empirically, the autograd method is usually slower and memory-consuming. 

\paragraph{Function-wise differentiation:} while it is expensive to apply the auto differentiation per query point, the derivative can be batched and computed in a function-wise manner.
\cite{li2024physics} developed an efficient and exact derivatives method based on the architecture of the neural operator that can compute the full gradient field. The idea is similar to the autograd, but the derivatives on the Fourier space are explicitly computed.
Given the explicit form, $u'$ can be directly computed on the Fourier space. 
\begin{align}
    u' 
    =   \cQ'\bigl(v_L \bigr) \cdot 
    \FFT^{-1}\left[\frac{i 2\pi }{D}K \cdot (\FFT [v_{L}]) \right]
\end{align}
Therefore, to exactly compute the derivative of the Fourier neural operator, one just needs to run the numerical Fourier differentiation. Especially the derivative can be efficiently computed with the Fast Fourier transform when the query is uniform. Similarly, if the kernel function $\kappa^{(l)}$ has a structured form, it is also possible to write out its gradient field explicitly.


\section{Learning Operators, Part IV: Neural Operators - Applications}

\subsection{Neural operators for Weather forecasting: FourCastNet}

One of the most relevant applications of Neural Operator lies in the field of weather forecasting, where Neural Operator zero-shot superresolution property is quite appealing. A set of relevant papers on this matter are the ones introducing \textit{FourCastNet}, a FNO-based neural operator for weather forecasting \cite{pathak2022fourcastnet, kurth2023fourcastnet, bonev2025fourcastnet}. 

\textit{FourCastNet} (Fourier Forecasting Neural Network) is a global, data-driven weather forecasting model that leverages the AFNO architecture to deliver accurate short- to medium-range predictions at a high spatial resolution of 0.25\textdegree, corresponding to roughly 30 km $\times$ 30 km at the equator. By framing token mixing as a continuous global convolution implemented efficiently in the Fourier domain, FourCastNet achieves a quasi-linear computational complexity of $O(N \log N)$ with respect to the sequence length, enabling it to process the $720 \times 1440$ global grid with unprecedented efficiency.

The model accurately forecasts high-resolution, fast-timescale variables, including surface wind speed, precipitation, and atmospheric water vapor, quantities that exhibit complex fine-scale structure and have been challenging for previous data-driven approaches. This capability has important implications for planning wind energy resources at onshore and offshore farms, as well as for predicting extreme weather events such as tropical cyclones, extratropical cyclones, and atmospheric rivers.

In terms of forecast skill, FourCastNet matches the accuracy of the ECMWF Integrated Forecasting System (IFS), a state-of-the-art Numerical Weather Prediction (NWP) model, at short lead times for large-scale variables like geopotential height, while outperforming IFS for variables with complex fine-scale structure, including precipitation and near-surface wind speeds. Perhaps most remarkably, FourCastNet generates a week-long forecast in less than 2 seconds, orders of magnitude faster than traditional NWP models, enabling the rapid and inexpensive creation of large-ensemble forecasts with thousands of members. This dramatic speedup, quantified as approximately $45\,000$ times faster on a node-hour basis than IFS, revolutionizes probabilistic forecasting by allowing well-calibrated uncertainty quantification for extreme events that smaller ensembles cannot reliably capture. Furthermore, FourCastNet's training time has been scaled to as little as 67 minutes on $3\,072$ NVIDIA A100 GPUs, achieving 140.8 petaFLOPS in mixed precision and demonstrating the viability of data-driven deep learning models as a valuable addition to the meteorology toolkit to aid and augment traditional NWP models.

\begin{figure}[t]
    \centering
    \includegraphics[width=0.65\textwidth]{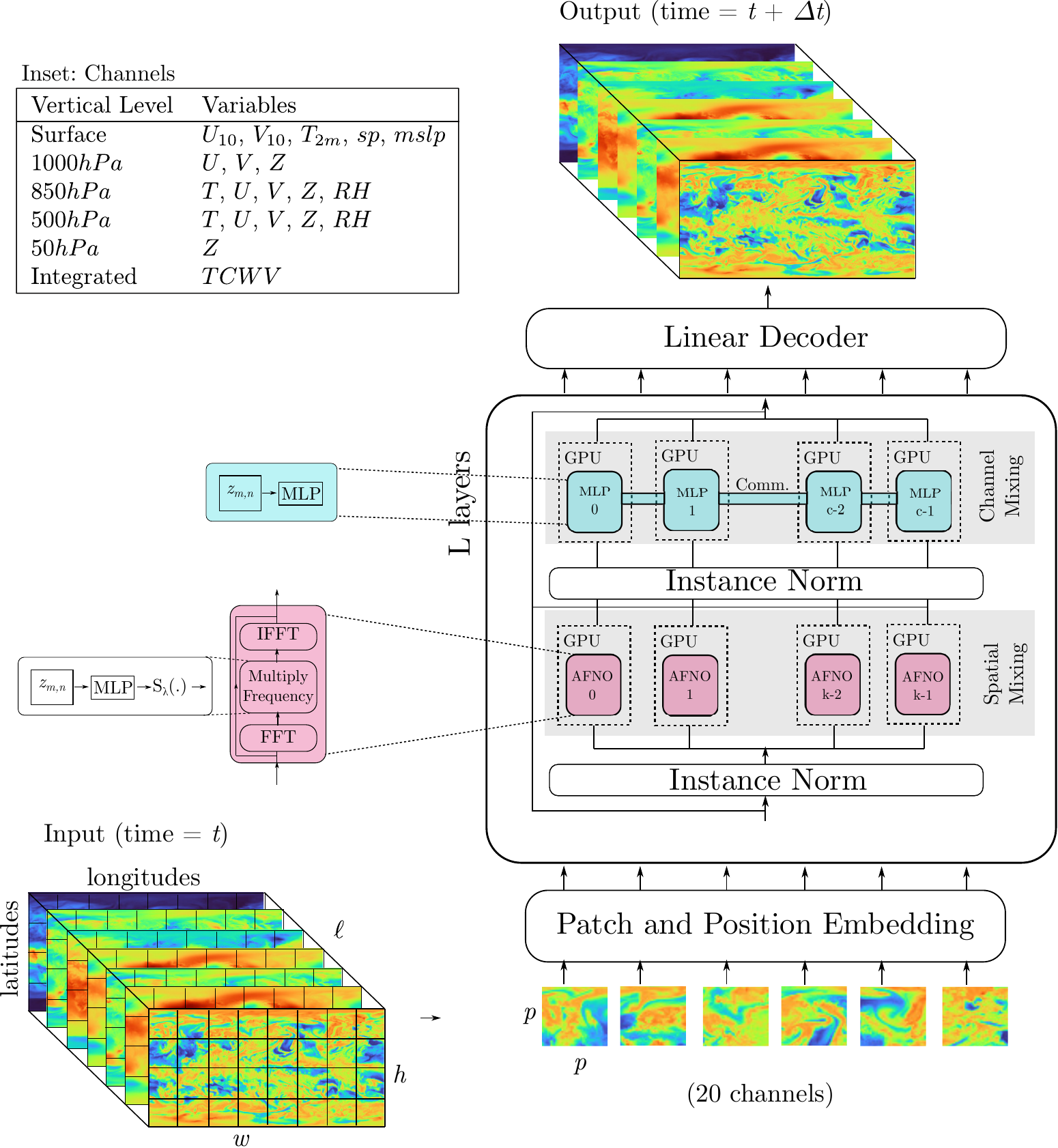}
    \caption{The AFNO architecture showing the key operations performed on the the input tensor with dimensions ($20 \times 720 \times 1440$) to produce a 6 hour single time step forecast with the same dimensions. Model parallelism is implemented by splitting the channels (feature maps) across GPUs. Channel mixing MLP operations require communication across the model parallel ranks, while the FFT based spatial-mixing operates on disjoint blocks that are embarrassingly parallel. From \cite{kurth2023fourcastnet}.} 
    \label{fig:5:5:1:afnocastnet}
\end{figure}
As said, the original implementation of FourCastNet employs an AFNO-based transformer model. 

As we can see from \cref{fig:5:5:1:afnocastnet}, The atmospheric state is represented initially by a tensor with
\(20\times720\times1440\) values (channels, latitude, longitude). A patch-embedding layer with \(8\times8\) patches maps this field to a \(90\times180\) token grid with embedding dimension \(768\). 
The resulting hidden tensor is then processed by a stack of Transformer-like blocks in which AFNO replaces dense multi-head
self-attention as the spatial-mixing mechanism.

\begin{figure}
    \centering
    \includegraphics[width=0.9\textwidth]{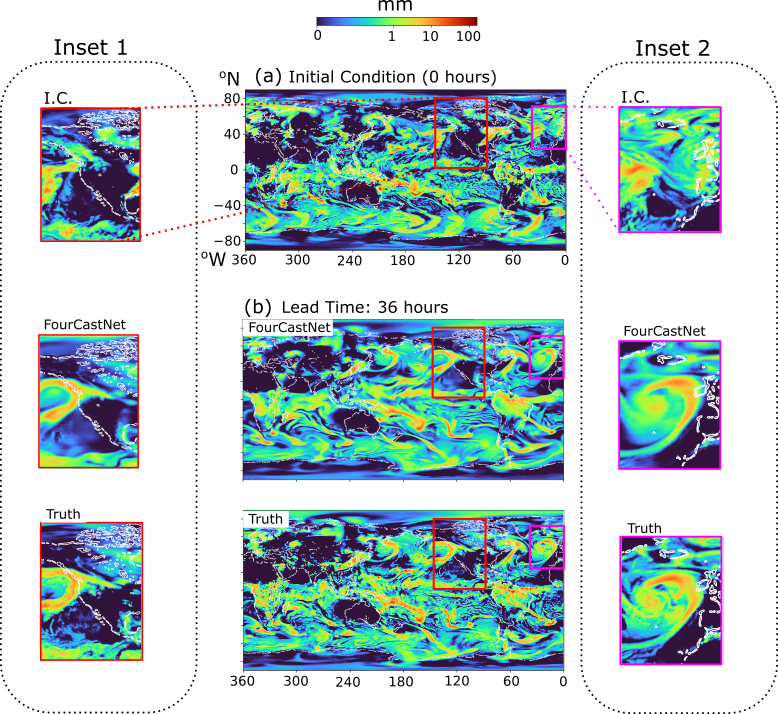}
    \caption{Illustration of a global Total Precipitation (TP) forecast using the FourCastNet model. Land-sea borders are shown using a thin white trace. For ease of visualization, the precipitation field is plotted as a log-transformed field in all panels. Panel (a) shows the TP fields at the time of forecast initialization. Panel (b) shows the TP forecast generated by the FourCastNet model (upper panel) over the entire globe at $0.25^{\circ}$-lat-long resolution with the corresponding truth (lower panel). Inset 1 shows the I.C., forecast and true precipitation fields at a lead time of 36 hours over a local region along the western coast of the United States. This highlights the ability of the FourCastNet model to resolve and predict localized regions of high precipitation, in this case due to an atmospheric river. Inset 2 shows the I.C., forecast, and true precipitation fields near the coast of the U.K. and highlights an extreme precipitation event due to an extra-tropical cyclone that is predicted very well by the FourCastNet model. The calendar time-stamp of the initial condition used to generate this forecast was 00:00 UTC on April 4, 2018. The high-resolution FourCastNet model demonstrates excellent skill in capturing small scale features that are key to precipitation forecasting. From \cite{pathak2022fourcastnet}.}
    \label{fig:5:5:1:precip}
\end{figure}

As an example of application, we report their result on a global Total Precipitation (TP) forecast using the FourCastNet model, \cref{fig:5:5:1:precip}. The total precipitation ($TP$) in the ERA5 re-analysis dataset is a variable that represents the accumulated liquid and frozen water that falls to the Earth's surface through rainfall and snow. It is defined in units of length as the depth of water that would accumulate if spread evenly over a unit grid box of the model.

\subsubsection{FourCastNet3: Spherical Neural Operator Blocks}

\begin{figure}
    \centering
    \includegraphics[width=0.95\linewidth]{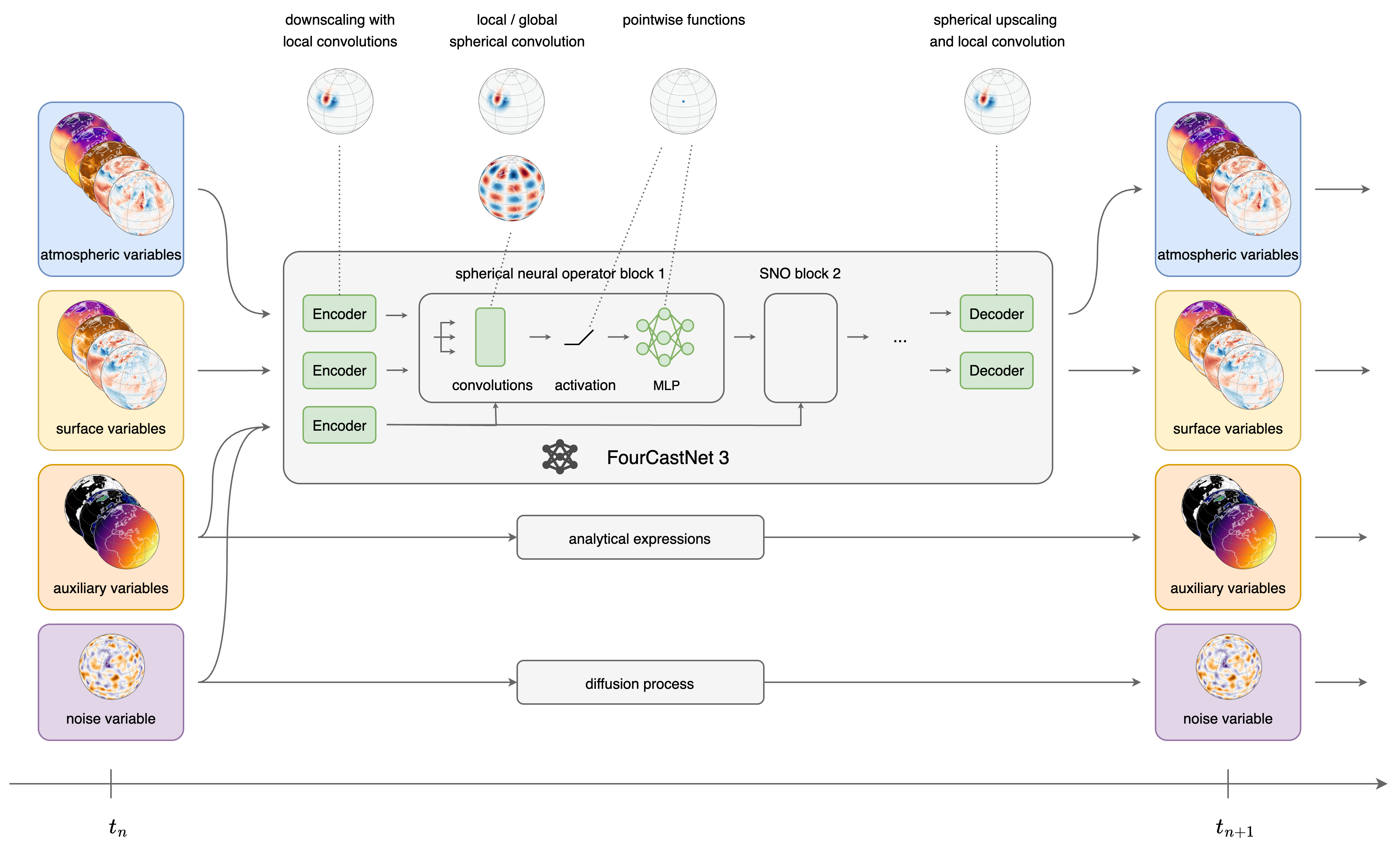}
    \caption{Schematic of the FourCastNet 3 model. The model predicts the state of the atmosphere at the next timestep, given the state at the previous timestep. Auxiliary variables such as the cosine zenith angle are computed from analytical expressions for each timestep and appended to the input. A hidden Markov model is obtained by conditioning FourCastNet 3 on a stochastic latent variable whose temporal dynamics are governed by a diffusion process on the sphere. The model itself is formed by an encoder, a decoder and 8 neural operator blocks. Each of these operations can be grouped into local, global and pointwise operations and therefore be formulated on arbitrary grids and resolutions, making FourCastNet 3 discretization independent. Green boxes illustrate learnable operations. From \cite{bonev2025fourcastnet}.}
    \label{fig:5:5:1:1:FCN3}
\end{figure}

Building upon the success of the original FourCastNet, the recently introduced FourCastNet 3 (FCN3) advances global weather modeling by implementing a scalable, geometric machine learning approach specifically designed for probabilistic ensemble forecasting \cite{bonev2025fourcastnet}. Unlike its predecessor, which, as we have seen, relied on the Adaptive Fourier Neural Operator (AFNO) within a vision transformer backbone, FCN3 adopts a purely convolutional architecture built around Spherical Neural Operator (SNO) blocks that are tailored to the geometry of the Earth. These blocks leverage two complementary types of spherical convolutions: global convolution filters parameterised in the spectral domain via the spherical harmonic transform (SHT), which resemble classical pseudo-spectral methods such as IFS, and local spherical group convolutions implemented using discrete-continuous (DISCO) convolutions, which enable anisotropic, locally supported filters that are better suited to capturing phenomena like adiabatic flow and orographic effects. The architecture organizes these blocks into an encoder-processor-decoder structure inspired by ConvNeXt \cite{liu2022convnet}, with a carefully calibrated ratio of four local blocks to one global block per stage, ensuring effective multiscale processing while respecting the spherical geometry and its inherent rotational symmetries. This geometric foundation enables FCN3 to better model the spatially correlated probabilistic nature of atmospheric dynamics, resulting in stable spectra and realistic dynamics across multiple scales.

\begin{figure}
    \centering
    \includegraphics[width=0.95\linewidth]{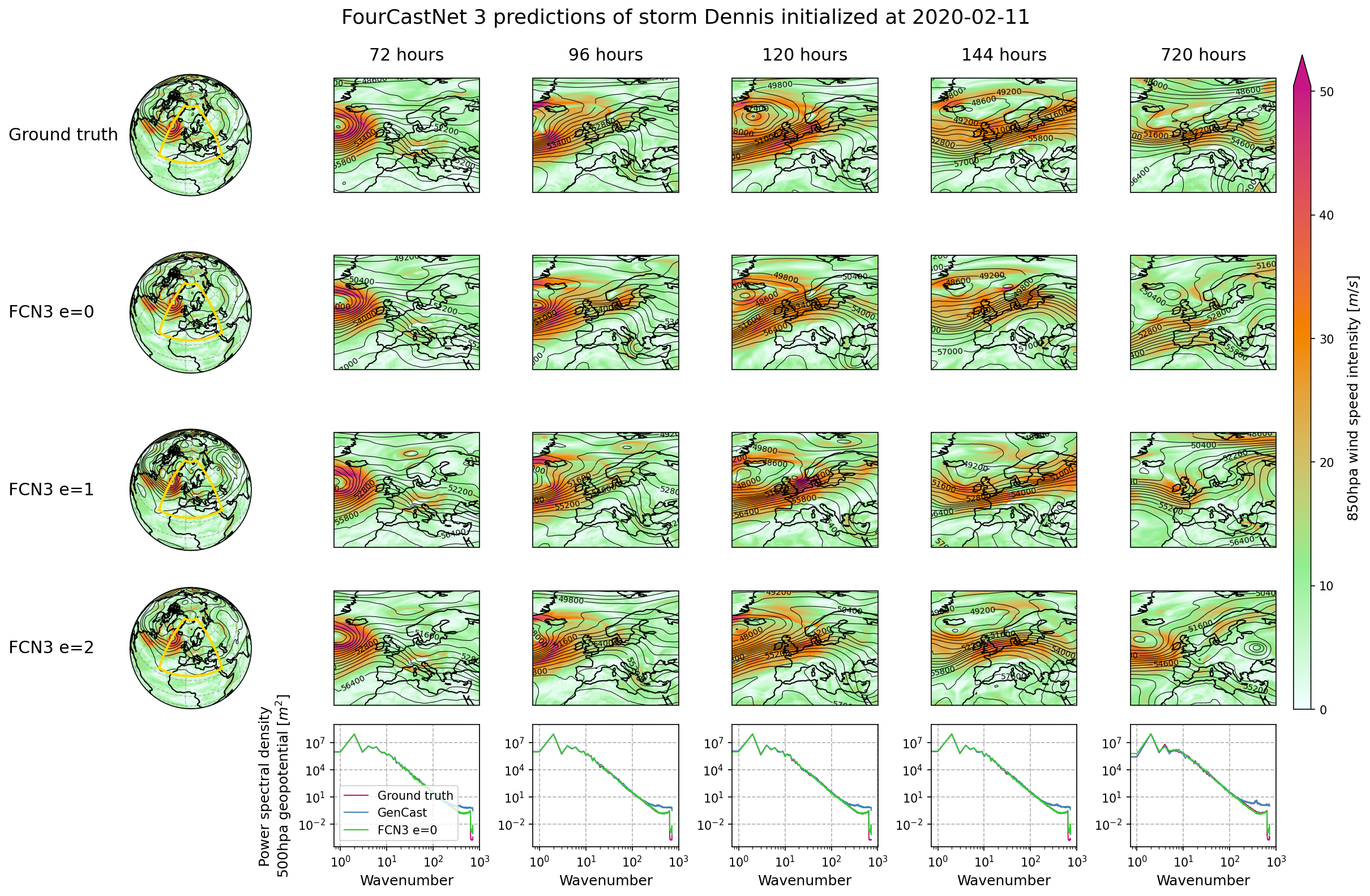}
    \caption{FourCastNet 3 prediction of storm Dennis initialized on 2020-02-11 at 00:00:00 UTC. The plot depicts wind-speeds at a pressure level of 850hPa and isohypses (height contours) of the 500hPa geopotential height. FCN3 accurately predicts the storm and its landfall 5 days in advance, with different ensemble members depicting different scenarios. FCN3 predicts global weather phenomena at a spatial resolution of 0.25\textdegree{} and a temporal resolution of 6 hours. From \cite{bonev2025fourcastnet}.}
    \label{fig:5:5:1:1:storm_dennis}
\end{figure}

Another key innovation of FCN3 is its end-to-end probabilistic formulation as a hidden Markov model, trained with a composite loss function that combines the Continuous Ranked Probability Score (CRPS) in both spatial and spectral domains. This approach effectively addresses the common issues of blurring in deterministic models and spurious high-frequency noise in diffusion-based probabilistic models, achieving well-calibrated ensembles with spread-skill ratios approaching unity. In terms of forecast skill, FCN3 surpasses the gold-standard IFS ensemble system and nearly matches the state-of-the-art diffusion-based model GenCast, while being 8 to 60 times faster in inference. A single 15-day global forecast at 0.25\textdegree{} resolution with 6-hourly outputs can be generated in under 60 seconds on a single NVIDIA H100 GPU.

\subsection{Neural Operators for Reservoir Simulation}

\subsubsection{Fourier Neural Operators for ECHELON software}

The application of neural operators has extended beyond weather forecasting into the domain of subsurface energy exploration, where they address critical challenges in reservoir simulation. Reservoir simulation involves modelling multiphase flow in porous media to predict oil, gas, and water movement over time. Traditional full-physics numerical simulators, while accurate, are computationally expensive, particularly when performing uncertainty quantification, history matching, or field optimization tasks that require thousands of forward simulations. These tasks often involve complex geological models with hundreds of thousands to millions of active grid cells, making them computationally prohibitive even with high-performance computing resources \cite{srt2024physicsnemo}.

\begin{figure}
    \centering
    \includegraphics[width=0.95\linewidth]{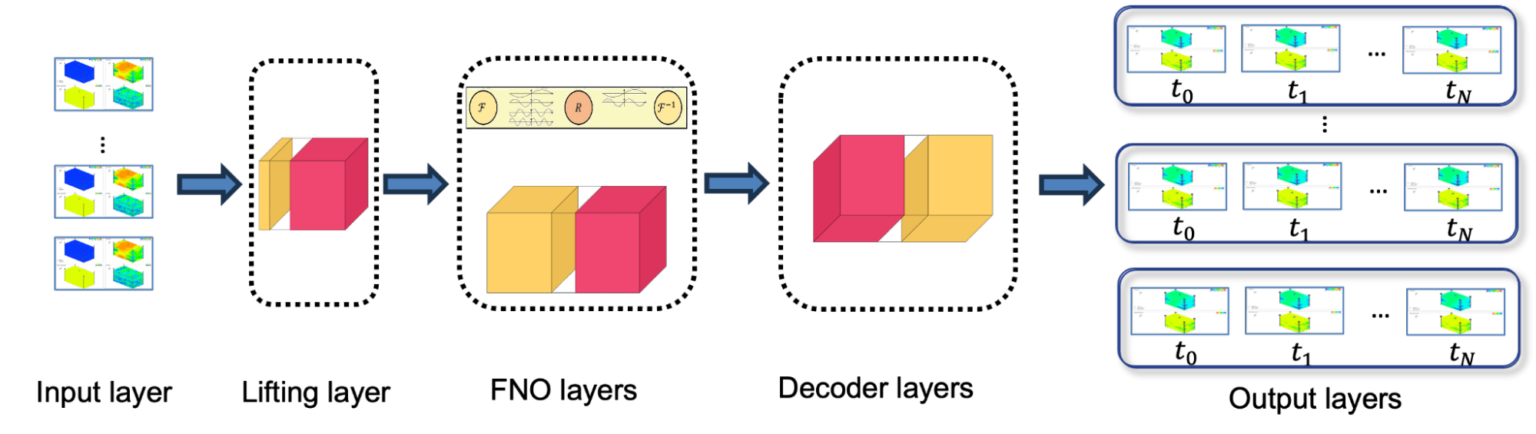}
    \caption{ Fourier neural operator architecture for generating spatio - temporal reservoir proxies with NVIDIA PhysicsNeMo on AWS. From \cite{srt2024physicsnemo}.}
    \label{fig:5:5:2:srt_arch}
\end{figure}

A promising solution to this computational bottleneck lies in the use of full-field proxy models that approximate the solution of primary state variables (such as pressure and saturation) directly in space and time, analogous to a full-physics simulator. Stone Ridge Technology (SRT) has developed a scalable framework that integrates its ECHELON reservoir simulator \cite{srt_echelon_software} with NVIDIA PhysicsNeMo \cite{NvidiaPhysicsNemoDocumentation} on AWS to generate such full-field proxy models using Fourier neural operators (FNOs) \cite{srt2024physicsnemo} (see \cref{fig:5:5:2:srt_arch}). The workflow begins with ECHELON generating large volumes of training, validation, and test data, which are stored on Amazon S3 for efficient retrieval. The neural operator model is then trained within the PhysicsNeMo framework to learn the spatio-temporal mapping from reservoir parameters (such as porosity, permeability, and well locations) to the resulting pressure and saturation fields. Once trained, these proxy models can perform inference orders of magnitude faster than forward simulations, achieving speedups of 10x to 100x while maintaining reasonably accurate agreement with the full-physics numerical solutions.

\begin{figure}[t]
\centering
\includegraphics[width=0.9\textwidth]{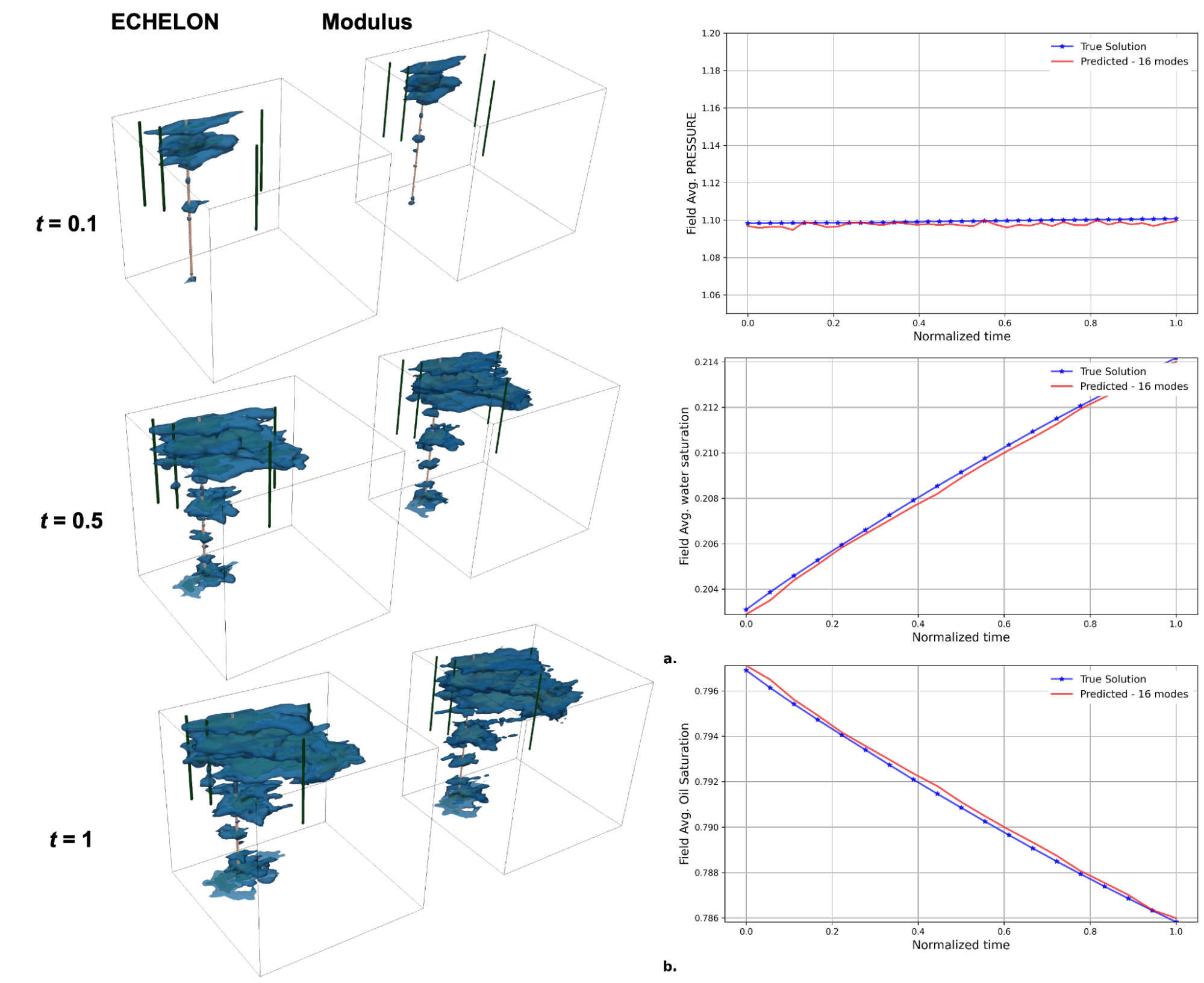}
\caption{Comparison of Fourier neural operator predictions against ground-truth ECHELON simulations for water front propagation in a heterogeneous permeability field under different well placement scenarios. The complex topological details of the waterfront are well captured by the neural operator proxy. Adapted from \cite{srt2024physicsnemo}.}
\label{fig:5:5:2:srt_waterfront}
\end{figure}

The effectiveness of this approach has been demonstrated on two representative use cases. The first involves well placement optimization in a reservoir with heterogeneous permeability fields. By varying the positions of injectors and producers across 500 training samples, the FNO-based proxy captured the complex topological details of waterfront propagation, showing strong agreement with the ground-truth ECHELON simulations (\cref{fig:5:5:2:srt_waterfront}). The second case addresses geological uncertainty in a real-field model: the Norne field in the Norwegian Sea, a complex reservoir with faults, heterogeneous anisotropic permeability, and 35 wells. Using 500 realizations with variations in porosity, permeability, and fault transmissibility multipliers, the FNO proxy demonstrated strong agreement with ECHELON for both pressure and water saturation fields across time.

This integration of a high-performance full-physics simulator with neural operator-based surrogate modeling provides a generalized framework for addressing subsurface challenges, including uncertainty quantification and field development optimization, with enhanced performance and high accuracy. The use of modern GPUs (A100, H100, H200) on AWS enables the computational scale required for training these models, while the resulting proxies open new possibilities for engineering workflows not previously feasible.

\begin{figure}[t]
    \centering
    \begin{subfigure}[b]{0.48\textwidth}
        \centering
        \includegraphics[width=\textwidth]{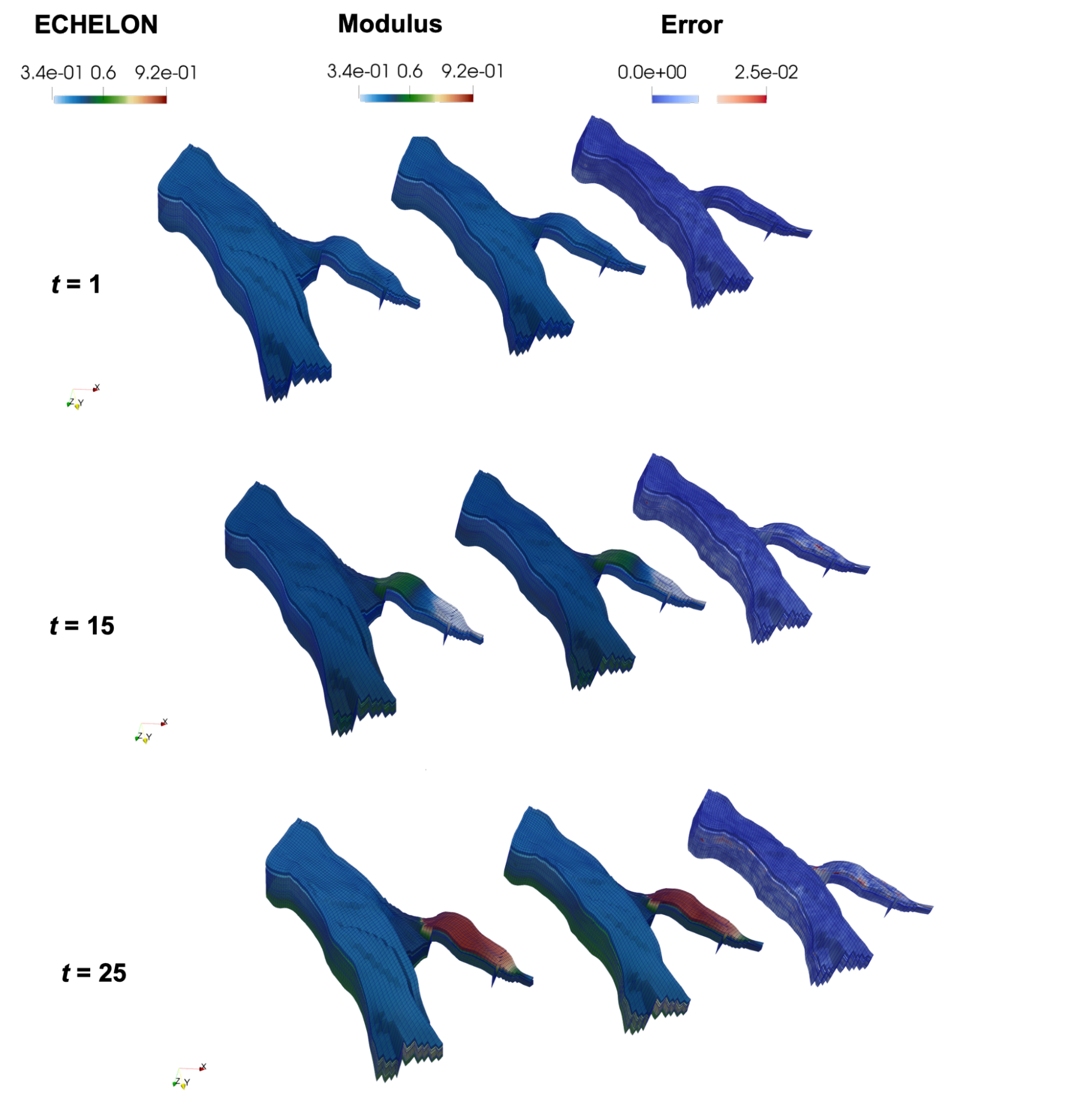}
        \caption{Pressure field comparison.}
    \end{subfigure}
    \hfill
    \begin{subfigure}[b]{0.48\textwidth}
        \centering
        \includegraphics[width=\textwidth]{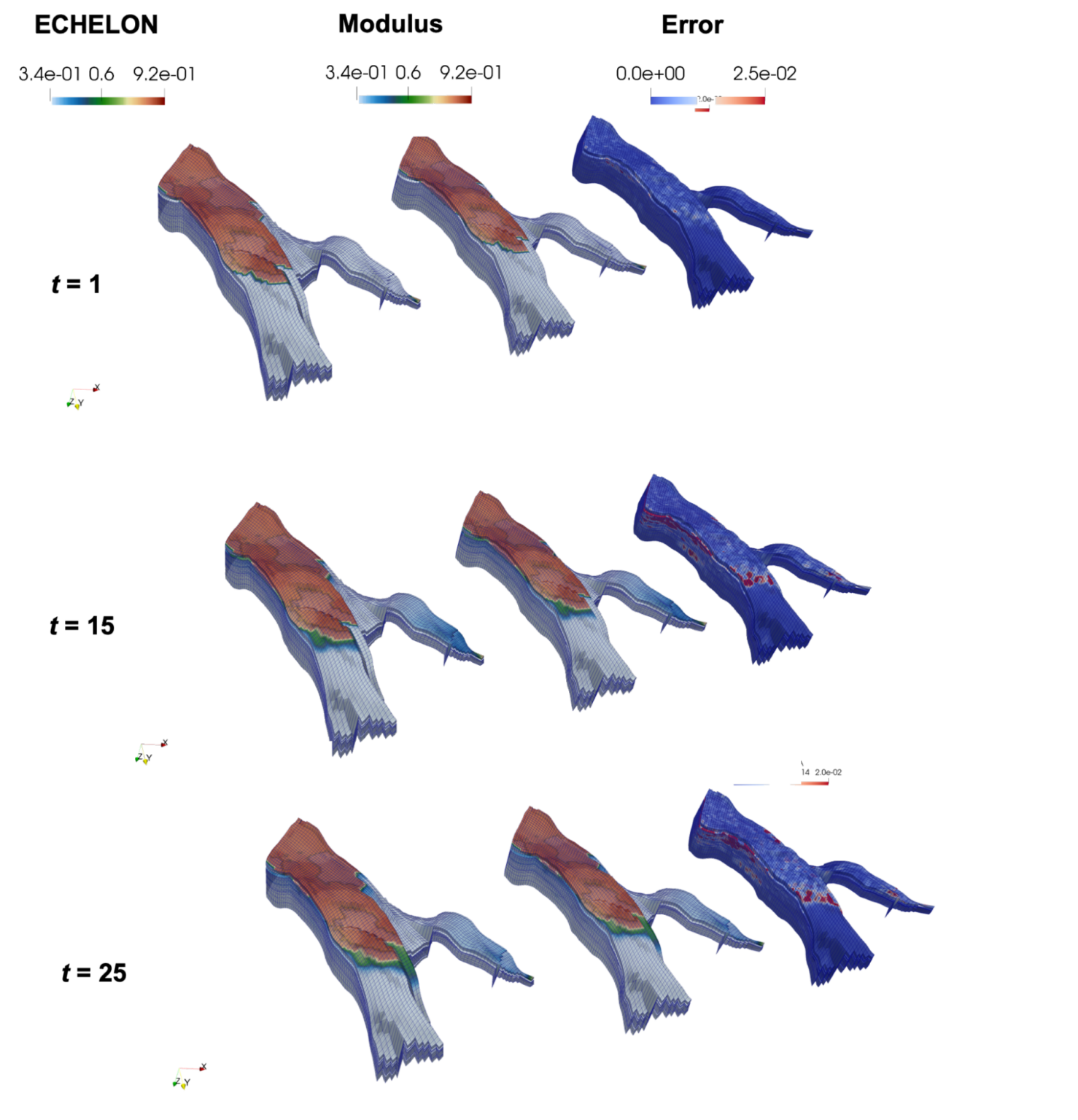}
        \caption{Water saturation comparison.}
    \end{subfigure}
    \caption{Validation of Fourier neural operator (FNO) proxy against full-physics ECHELON simulations for the Norne field. The FNO-based proxy, trained using NVIDIA PhysicsNeMo, demonstrates strong agreement with the ground-truth ECHELON solutions for both (a) pressure fields and (b) water saturation fields across the 3,298-day simulation period. Only small errors are observed, validating the use of neural operators as accurate surrogates for full-physics reservoir simulation. Adapted from \cite{srt2024physicsnemo}.}
    \label{fig:fno_echelon_validation}
\end{figure}

\subsubsection{3D Reservoir Simulation with Physics-Informed Neural Operators}

As a complement to the data-driven proxy modeling approach discussed in the previous section, NVIDIA PhysicsNeMo Sym provides a framework for physics-informed machine learning that embeds the governing partial differential equations directly into the training process. A representative example is the 3D reservoir simulation problem, which models two-phase (oil-water) flow in porous media using a physics-informed neural operator (PINO) architecture \cite{etienam_reservoir_readme}.

\paragraph{Governing Equations:} the two-phase black-oil model for reservoir simulation is governed by a system of coupled PDEs describing pressure and saturation evolution. The conservation equations for water and oil phases are given by \cite{etienam_reservoir_readme}:
\begin{align}
\phi(\mathbf{x}) \frac{\partial S_w}{\partial t} - \nabla \cdot \left[ T_w \left( \nabla p_w + \rho_w g \nabla z \right) \right] &= Q_w,   \\
\phi(\mathbf{x}) \frac{\partial S_o}{\partial t} - \nabla \cdot \left[ T_o \left( \nabla p_o + \rho_o g \nabla z \right) \right] &= Q_o, 
\end{align}
where $\phi(\mathbf{x})$ is the porosity, $S_w$ and $S_o$ are the water and oil saturations, $p_w$ and $p_o$ are the phase pressures, $\rho_w$ and $\rho_o$ are the phase densities, $g$ is gravitational acceleration, and $Q_w$, $Q_o$ are source/sink terms representing wells. The phase transmissibilities are defined as:
\begin{equation}
T_w = \frac{K(\mathbf{x}) K_{rw}(S_w)}{\mu_w}, \qquad 
T_o = \frac{K(\mathbf{x}) K_{ro}(S_o)}{\mu_o}, \qquad 
T = T_w + T_o,
\end{equation}
where $K(\mathbf{x})$ is the absolute permeability tensor, $K_{rw}(S_w)$ and $K_{ro}(S_o)$ are the relative permeabilities, and $\mu_w$, $\mu_o$ are the phase viscosities. The system is closed by the saturation constraint and capillary pressure relation:
\begin{equation}
S_w + S_o = 1, \qquad p_c(S_w) = p_o - p_w.
\end{equation}

The pressure equation is derived by combining the two phase equations, yielding:
\begin{equation}
-\nabla \cdot \left[ T(\mathbf{x}, S_w) \, \nabla p \right] = Q, \qquad Q = Q_o + Q_w, \label{eq:pressure}
\end{equation}
while the water saturation equation can be expressed as:
\begin{equation}
\phi(\mathbf{x}) \frac{\partial S_w}{\partial t} - \nabla \cdot \left[ T_w(\mathbf{x}, S_w) \, \nabla p_w \right] = Q_w. 
\end{equation}

\begin{figure}[t]
    \centering
    \includegraphics[width=0.95\textwidth]{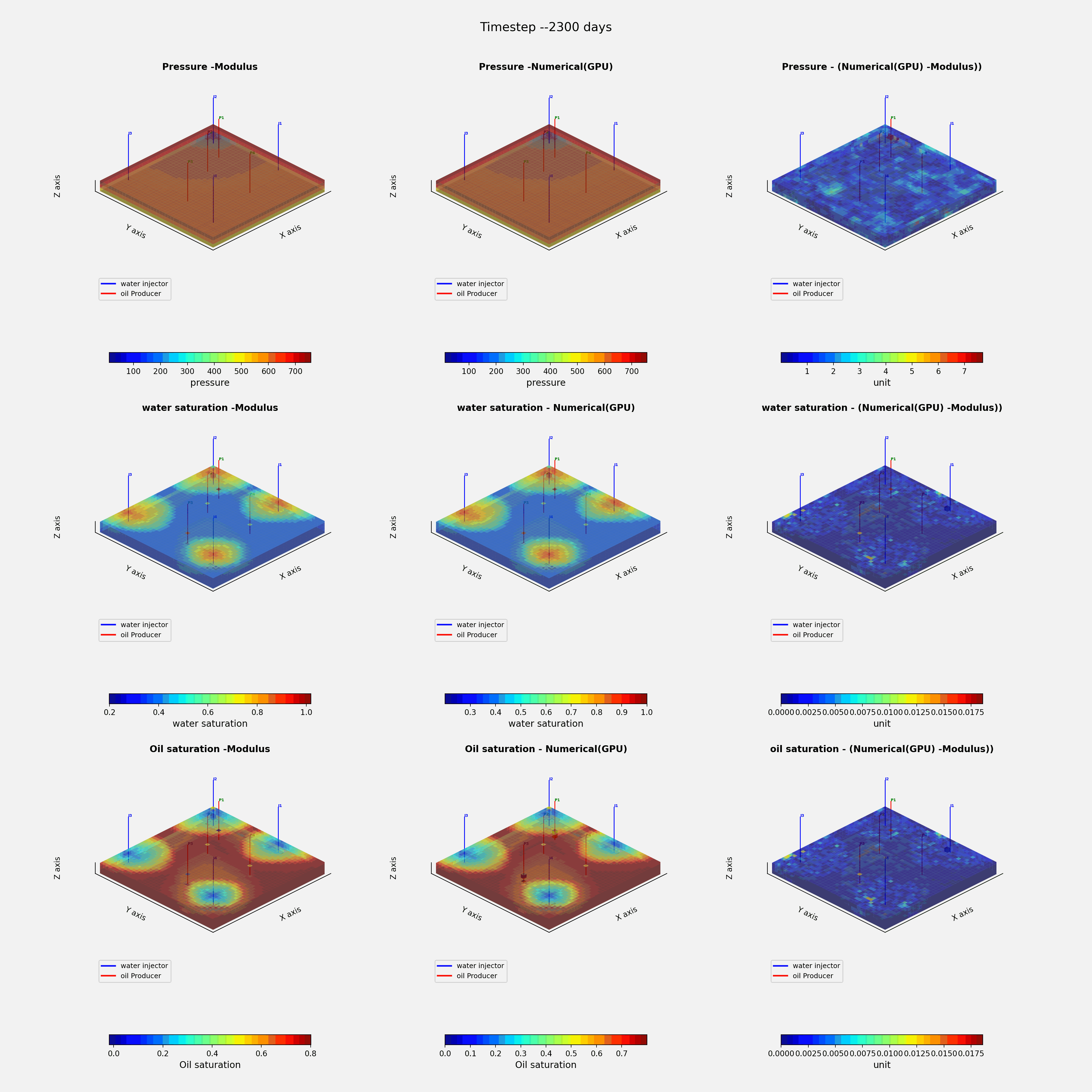}
    \caption{Comparison of pressure, water saturation, and oil saturation fields between the PINO surrogate (left column), the finite-volume solver (middle column), and the difference (right column), at the last time step. The results show strong agreement between the physics-informed neural operator and the traditional numerical simulator. Adapted from \cite{etienam_reservoir_readme}.}
    \label{fig:5:5:2:2:pino_3d_reservoir}
\end{figure}

\paragraph{Physics-Informed Neural Operator Formulation}
The goal of the surrogate modelling approach is to replace the traditional finite-volume simulator with a neural operator that predicts pressure, saturation, and flux fields given any input permeability and porosity field. Following the approach of \cite{etienam_reservoir_readme}, an additional vector variable, the flux $\mathbf{F}$, is introduced, which transforms the pressure equation into a first-order system:
\begin{equation}
\mathbf{u} = \nabla p, \qquad \mathbf{F} = T \nabla p, \qquad -\nabla \cdot \mathbf{F} = Q.
\end{equation}
The pressure equation loss is then formulated as a mixed residual loss:
\begin{equation}
\mathcal{L}_{\text{pressure}} = \frac{1}{n_s} \left( \|\mathbf{F} - T \odot \nabla \mathbf{u}\|_2^2 + \| -\nabla \cdot \mathbf{F} - Q \|_2^2 \right),
\end{equation}
while the saturation equation loss is given by:
\begin{equation}
\mathcal{L}_{\text{saturation}} = \frac{1}{n_s} \left\| \phi \frac{\partial S_w}{\partial t} - \nabla \cdot \left[ T_w \nabla \mathbf{u} \right] - Q_w \right\|_2^2.
\end{equation}
The total physics-informed loss combines both contributions:
\begin{equation}
\mathcal{L}_{\text{PINO}} = \mathcal{L}_{\text{pressure}} + \mathcal{L}_{\text{saturation}}.
\end{equation}

This mixed residual formulation enables the neural operator to learn the underlying physics while maintaining the ability to predict flux variables directly, which are essential for well-based production analysis.

\begin{figure}[t]
    \centering
    \begin{subfigure}[b]{0.48\textwidth}
        \centering
        \includegraphics[width=\textwidth]{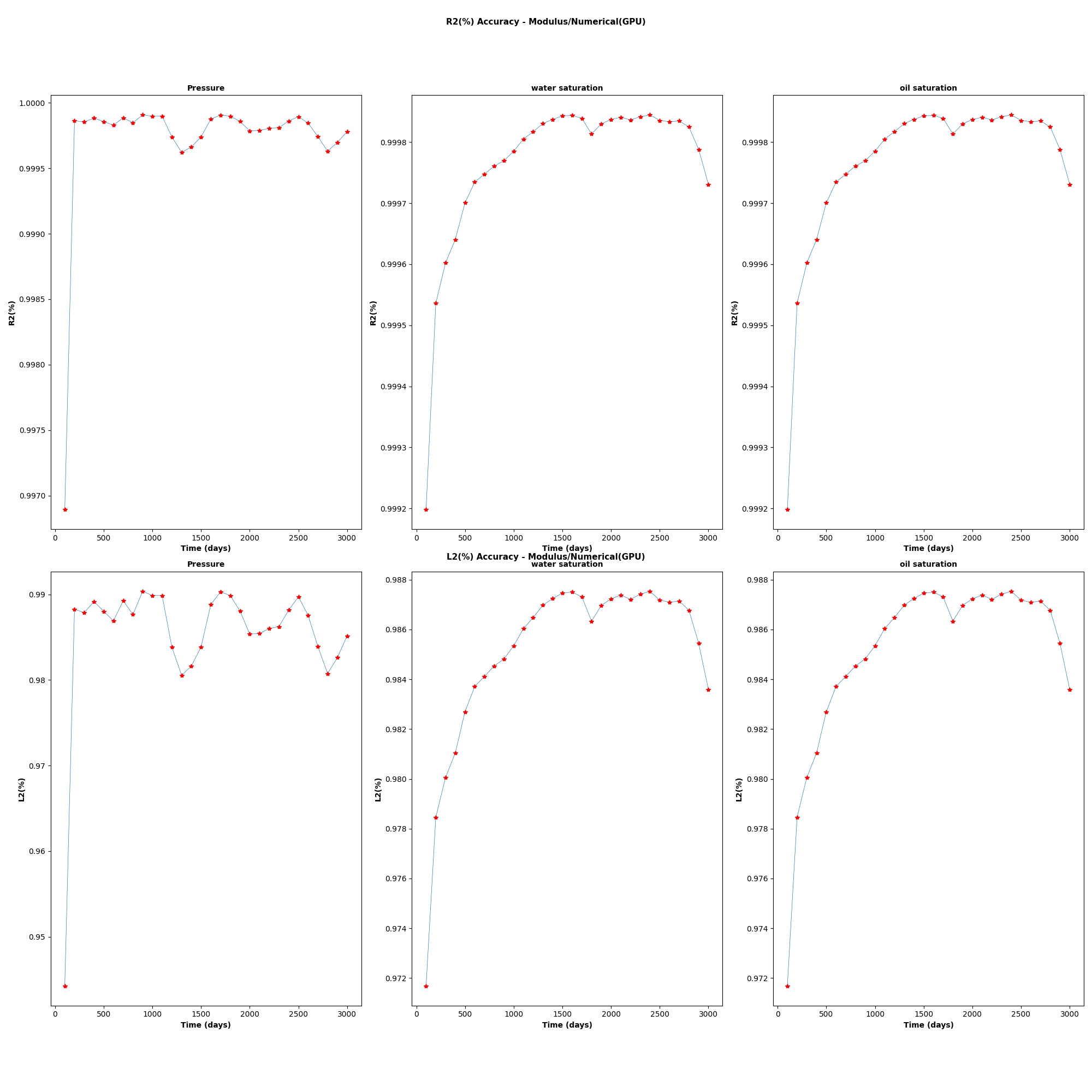}
        \caption{FNO: $R^2$ and L2 accuracy metrics for pressure and water saturation over time.}
    \end{subfigure}
    \hfill
    \begin{subfigure}[b]{0.48\textwidth}
        \centering
        \includegraphics[width=\textwidth]{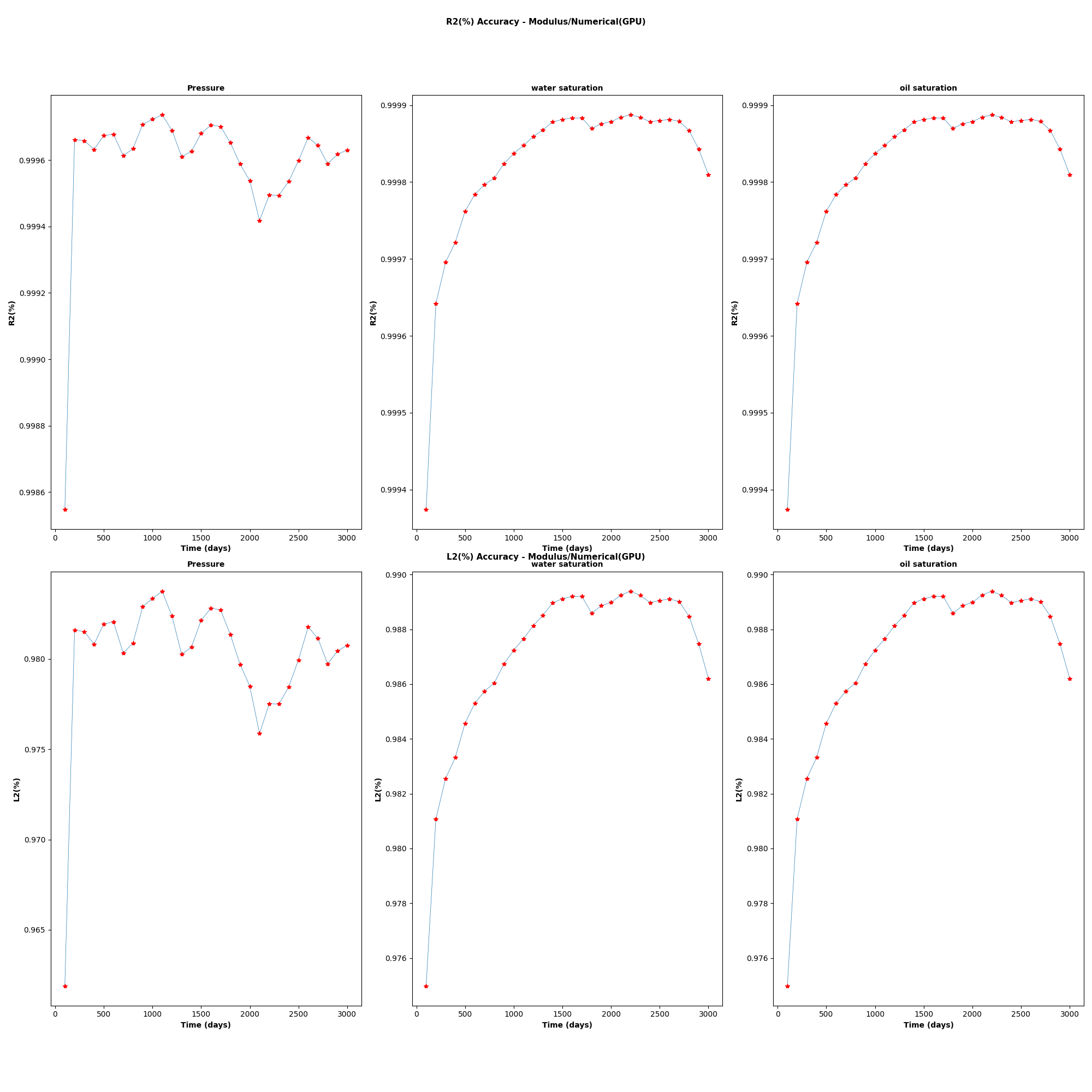}
        \caption{PINO: $R^2$ and L2 accuracy metrics for pressure and water saturation over time.}
    \end{subfigure}
    \caption{Accuracy metrics for the neural operator surrogates. Both FNO and PINO models achieve high $R^2$ scores (>0.95) for pressure and saturation predictions throughout the simulation period, demonstrating their capability to accurately surrogate the full-physics reservoir simulator. Adapted from \cite{etienam_reservoir_readme}.}
    \label{fig:5:5:2:2:accuracy_metrics}
\end{figure}

The reservoir model used in this example is a 3D channelized sandstone reservoir with dimensions $40 \times 40 \times 3$ grid cells, containing two lithofacies. The well configuration consists of 4 injectors and 4 producers arranged in an ``analogous 5-spot pattern''. A total of $2\,340$ days of simulation are performed, with the model predicting the evolution of pressure, water saturation, and oil saturation fields.

The neural operator is trained on 500 samples generated by the finite-volume simulator, using a combination of data loss and physics-informed loss as described above. \cref{fig:5:5:2:2:pino_3d_reservoir} shows the comparison between the PINO surrogate and the finite-volume reference. The left column displays the PINO predictions, the middle column shows the reference finite-volume solution, and the right column presents the difference between the two. For all panels, the first row corresponds to pressure, the second row to water saturation, and the third row to oil saturation.

\begin{figure}[t]
    \centering
    \includegraphics[width=0.95\textwidth]{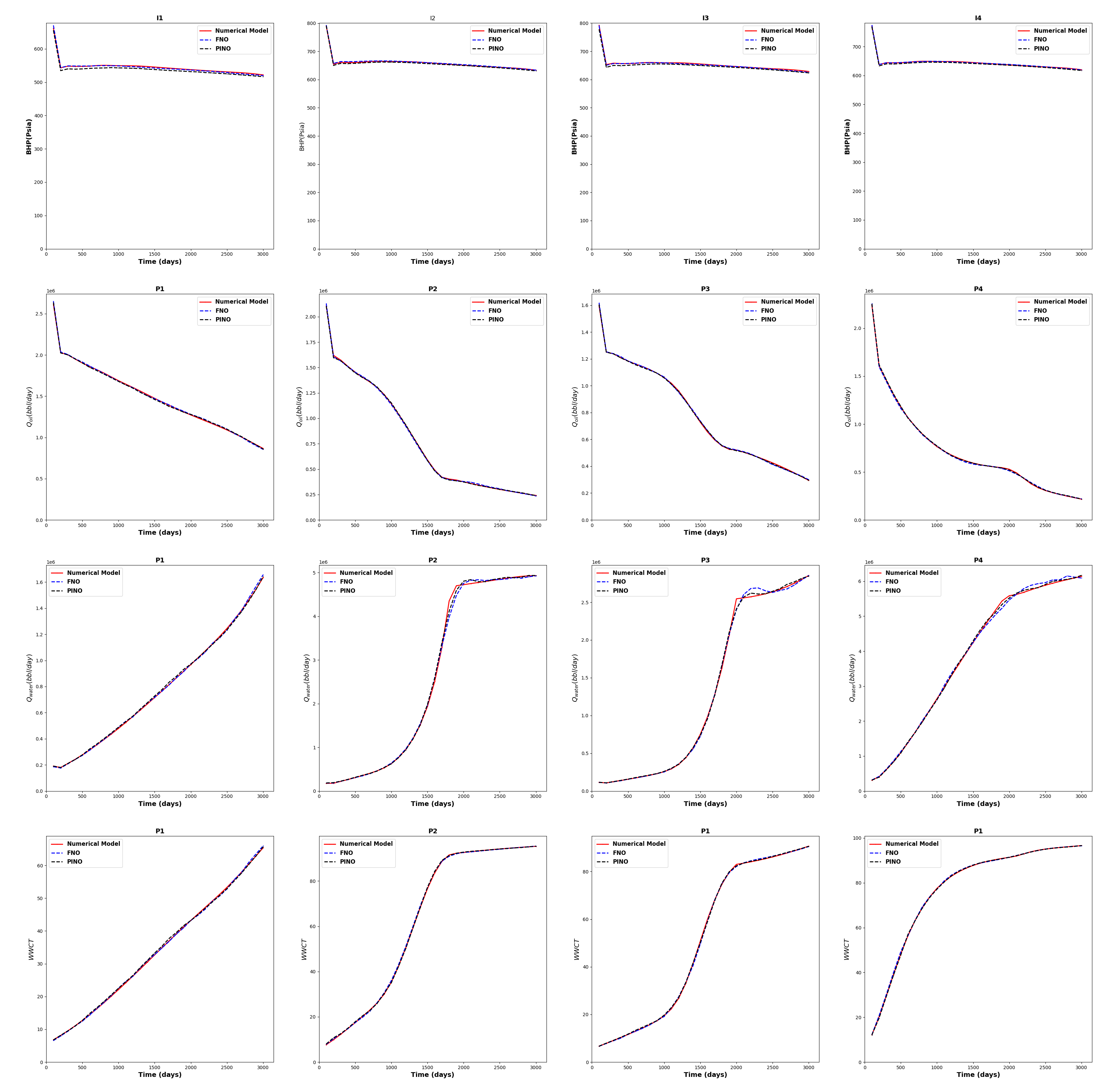}
    \caption{Production profile comparison between the true model (red) and the two surrogates (FNO and PINO). The panels show (first row) bottom-hole pressure for injectors I1–I4, (second row) oil production rate for producers P1–P4, (third row) water production rate for producers P1–P4, and (fourth row) water cut ratio for the four producers. The PINO and FNO surrogates closely match the reference production profiles. Adapted from \cite{etienam_reservoir_readme}.}
    \label{fig:5:5:2:production_profiles}
\end{figure}

The quantitative accuracy of the surrogates is assessed using $R^2$ and L2 metrics over the simulation time steps, as shown in \cref{fig:5:5:2:2:accuracy_metrics}. Both the FNO and PINO models achieve high $R^2$ scores (>0.95) for pressure and saturation predictions throughout the simulation period. \cref{fig:5:5:2:production_profiles} compares the production profiles between the reference finite-volume solver and the two surrogate models. The panels show bottom-hole pressure for the four injectors, oil production rate, water production rate, and water cut ratio for the four producers. The excellent agreement between the surrogates and the reference demonstrates that the neural operator models can effectively capture the complex dynamics of two-phase flow in porous media, including well responses critical for reservoir management decisions.

\subsection{Foundation Models for PDEs}
\subsubsection{MORPH}


The architectures discussed in \cref{subsec:5:4:1:FNO,subsec:5:4:2:AFNO,subsec:5:4:3:PINO} are most commonly trained as task-specific surrogate models for a prescribed PDE family and a specified data representation. A complementary line of research aims instead at \emph{foundation models for PDEs}: generalist models pretrained on heterogeneous collections of physical simulations and subsequently adapted to downstream tasks.

MORPH \cite{rautela2025morph} is one such model. It is a modality-agnostic autoregressive PDE foundation model based on a
convolutional Vision Transformer architecture, designed to process heterogeneous regular-grid datasets spanning 1D, 2D, and 3D physical systems.

A central challenge in building a foundation model for PDEs is the heterogeneity of scientific datasets: 1D time series from point probes, 2D weather maps, and 3D turbulence simulations all coexist, often with varying numbers of physical fields (pressure, velocity components, density, etc.). To address this, MORPH introduces the \emph{Unified Physics Tensor Format} (UPTF-7), which standardizes mini-batches into a common 7D shape:
\[
\text{UPTF-7} = (N, T, F, C, D, H, W),
\]
where \(N\) is the number of trajectories, \(T\) the number of time steps, \(F\) the number of physical fields, \(C\) the number of components per field (for vector-valued quantities), and \(D, H, W\) the spatial depth, height, and width.
This format avoids globally padding or reshaping entire heterogeneous datasets to a common spatial resolution. Instead, each dataset retains its native on-disk representation and individual mini-batches are converted on-the-fly to the common UPTF-7 convention. Some local component harmonisation is nevertheless required, including scalar-to-vector lifting and, where necessary, zero-padding of component channels. 
\cref{tab:5:5:3:1:morph_uptf} illustrates how datasets from popular benchmarks such as \textit{PDEBench} \cite{takamoto2022pdebench} and \textit{The Well} \cite{ohana2024well} map to UPTF-7.

\begin{table}[ht]
    \centering
    \resizebox{0.98\textwidth}{!}{%
    \begin{tabular}{l c c}
    \toprule
    \textbf{Dataset} & \textbf{Native Shape} & \textbf{UPTF-7 Mapping} \\
    \midrule
    1D Compressible Navier–Stokes & \((B, T, W)\) & \((B, T, 3, 1, 1, 1, 1024)\) \\
    2D Diffusion–Reaction & \((B, T, H, W, 2)\) & \((B, T, 2, 1, 1, 128, 128)\) \\
    3D Magnetohydrodynamics & \((B, T, D, H, W, 3, 3)\) & \((B, T, 3, 3, 64, 64, 64)\) \\
    \bottomrule
    \end{tabular}
    }
    \caption{Examples of datasets represented in the Unified Physics Tensor Format (UPTF-7).}
    \label{tab:5:5:3:1:morph_uptf}
\end{table}

\begin{figure}[t]
    \centering
    \includegraphics[width=0.99\textwidth]{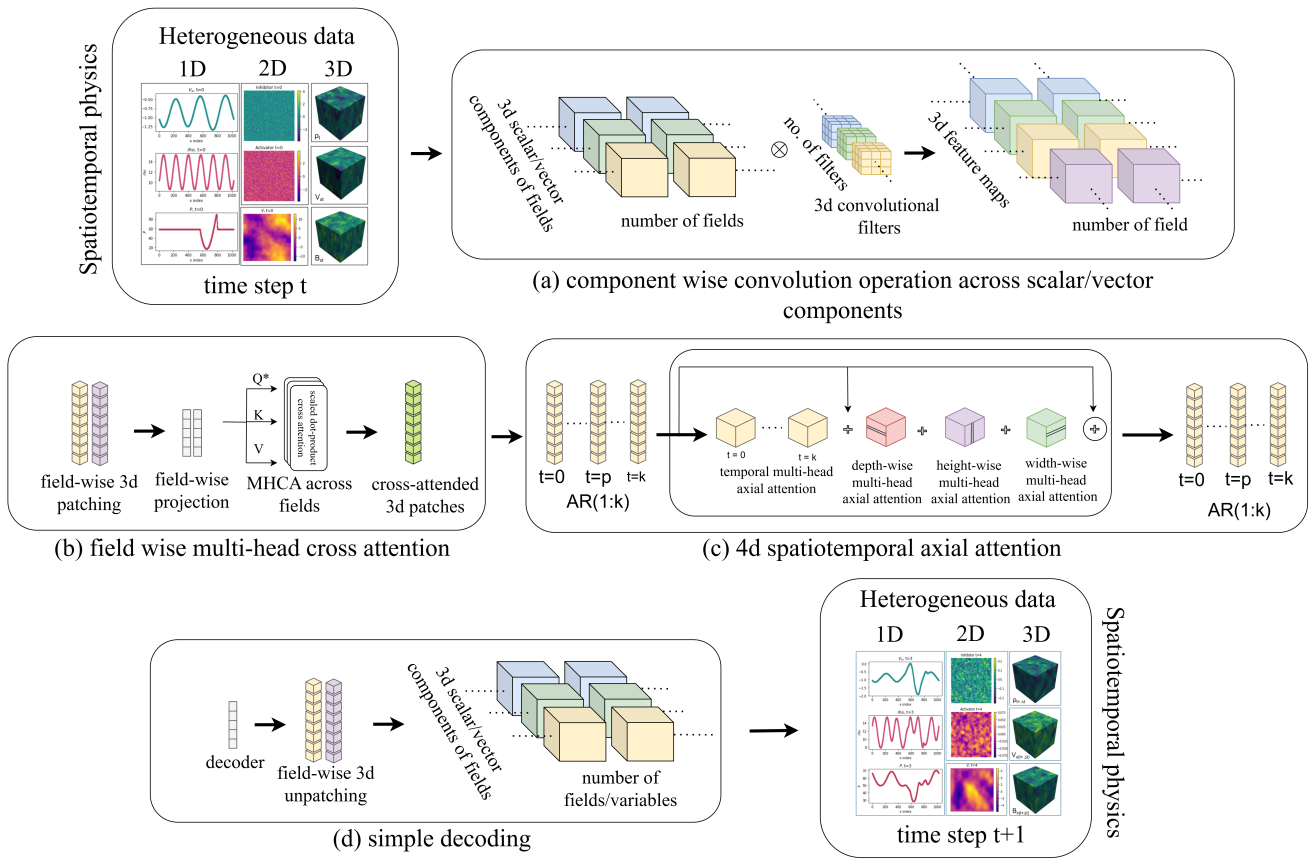}
    \caption{Schematic overview of the MORPH architecture. Input data in UPTF-7 format is processed by (a) component-wise convolutions that capture local interactions across scalar and vector channels, (b) inter-field cross-attention that selectively propagates information between physical fields, and (c) 4D axial attention that factorizes spatiotemporal self-attention along individual axes. Adapted from \cite{rautela2025morph}.}
    \label{fig:5:5:3:1:morph_architecture}
\end{figure}

MORPH's modality-agnostic capability stems from three key architectural mechanisms, which together enable it to learn from heterogeneous data while respecting the computational constraints of high-resolution 3D simulations. A schematic overview is shown in \cref{fig:5:5:3:1:morph_architecture}.

Unlike standard convolutional layers that operate on spatial dimensions, MORPH's first processing stage applies 3D convolutions along the component dimension \(C\) of the UPTF-7 tensor. This design jointly processes scalar and vector channels (e.g., the \(x\), \(y\), \(z\) components of a velocity field) to capture local interactions in a manner that is independent of the overall spatial dimensionality. This component-wise processing serves as a modality-agnostic alternative to the channel mixing operations in FNO (\cref{subsec:5:4:1:FNO}) and the block-diagonal MLP in AFNO (\cref{subsec:5:4:2:AFNO}), enabling the model to handle inputs ranging from 1D point measurements to 3D volumetric fields without architectural reconfiguration.

After the initial convolution, the model must reason about the relationships between different physical fields: for instance, how pressure and velocity co-evolve in a fluid system. MORPH employs a learned cross-attention mechanism that condenses the \(F\) fields into a single fused representation. Given a set of field embeddings \(\mathbf{X} \sim (F,  E)\), a learned query \(\mathbf{q} \sim (E,)\) attends over the fields:
\[
\begin{split}
    \alpha^{(h)} &= \mathrm{softmax}\left(\frac{(\mathbf{q}W_Q^{(h)})(\mathbf{X}W_K^{(h)})^T}{\sqrt{d_h}}\right), \\ 
\mathbf{z}^{(h)} &= \alpha^{(h)} (\mathbf{X}W_V^{(h)}), \\
\mathbf{z} &= \mathrm{Concat}_h(\mathbf{z}^{(h)})W_O.
\end{split}
\]

This operation is permutation-invariant to the ordering of fields and naturally accommodates a variable number of fields at inference time. This contrasts with the fixed field mixing in FNO and AFNO, which assume a fixed channel count.

The computational bottleneck in applying transformers to high-resolution PDE data is the quadratic cost of full spatio-temporal self-attention: \(\mathcal{O}((TDHW)^2)\). To overcome this, MORPH factorizes the 4D space-time tensor into separate 1D attention operations along the time (\(T\)), depth (\(D\)), height (\(H\)), and width (\(W\)) axes. For each axis, the remaining dimensions are folded into the batch, attention is applied along the axis length, and the outputs are summed residually:
\[
y = x + \tilde{x}^{(t)} + \tilde{x}^{(D)} + \tilde{x}^{(H)} + \tilde{x}^{(W)}.
\]

This factorization reduces the computational complexity to \(\mathcal{O}(T^2 + D^2 + H^2 + W^2)\), enabling MORPH to process large 3D grids (e.g., \(64^3\) voxels) that would be infeasible with full attention.

\paragraph{Pretraining and Scaling:} MORPH is pretrained on six heterogeneous datasets spanning 1D, 2D, and 3D domains, including compressible Navier–Stokes flows, magnetohydrodynamics (MHD), turbulent self-gravitating flows (TGC), and reaction–diffusion systems. The pretraining corpus comprises datasets from \textit{PDEBench} \cite{takamoto2022pdebench}, \textit{The Well} \cite{ohana2024well}, and \textit{PDEgym} \cite{herde2024poseidon}, totalling over $20\,000$ trajectories. Four model variants are released, scaling from 7 million to 480 million parameters (\cref{tab:5:5:3:1:morph_variants}).

\begin{table}[t]
    \centering
    \resizebox{0.99\textwidth}{!}{%
    \begin{tabular}{l c c c c c}
    \toprule
    \textbf{Variant} & \textbf{Conv filters} & \textbf{Attention dim} & \textbf{Heads} & \textbf{Depth} & \textbf{Parameters} \\
    \midrule
    MORPH-Ti (Tiny) & 8 & 256 & 4 & 4 & 7M \\
    MORPH-S (Small) & 8 & 512 & 8 & 4 & 30M \\
    MORPH-M (Medium) & 8 & 768 & 12 & 8 & 126M \\
    MORPH-L (Large) & 8 & 1024 & 16 & 16 & 480M \\
    \bottomrule
    \end{tabular}
    }
    \caption{MORPH model variants and their parameter counts.}
    \label{tab:5:5:3:1:morph_variants}
\end{table}

\paragraph{MORPH Results:}

\begin{figure}[t]
    \centering
    \includegraphics[width=0.7\textwidth]{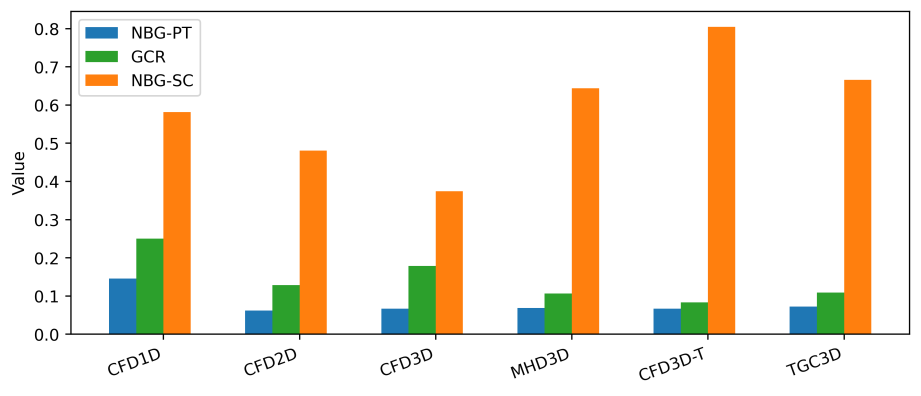}
    \caption{Zero-shot Gap-Closure Ratio (GCR) for MORPH pretrained on 2D incompressible Navier–Stokes and evaluated on six out-of-distribution targets. Positive GCR across all targets indicates successful transfer across modalities and physics. Adapted from \cite{rautela2025morph}.}
    \label{fig:5:5:3:1:morph_gcr}
\end{figure}

Perhaps the most striking evidence of MORPH's generalist capability comes from zero-shot transfer experiments. A MORPH model pretrained \emph{only} on 2D incompressible Navier–Stokes data was evaluated on six out-of-distribution targets without any fine-tuning. To quantify transfer, the authors introduced the \emph{Gap-Closure Ratio} (GCR):
\[
\mathrm{GCR} = \frac{1 - (E_{\mathrm{PT}} / E_{\mathrm{NB}})}{1 - (E_{\mathrm{SC}} / E_{\mathrm{NB}})},
\]
where \(E_{\mathrm{PT}}\) is the zero-shot error of the pretrained model, \(E_{\mathrm{SC}}\) is the error of a model trained from scratch on the target, and \(E_{\mathrm{NB}}\) is the error of a naive baseline (random initialization). A GCR value greater than zero indicates positive transfer. As shown in \cref{fig:5:5:3:1:morph_gcr}, MORPH achieves positive GCR across all six targets, including compressible flows in 1D, 2D, and 3D, as well as magnetohydrodynamics (MHD) and turbulent self-gravitating flows (TGC). This demonstrates that the model learns reusable operators that transfer across both spatial dimensionality and underlying physics.

\begin{figure}[t]
    \centering
    \begin{subfigure}[b]{0.48\textwidth}
        \centering
        \includegraphics[width=\textwidth]{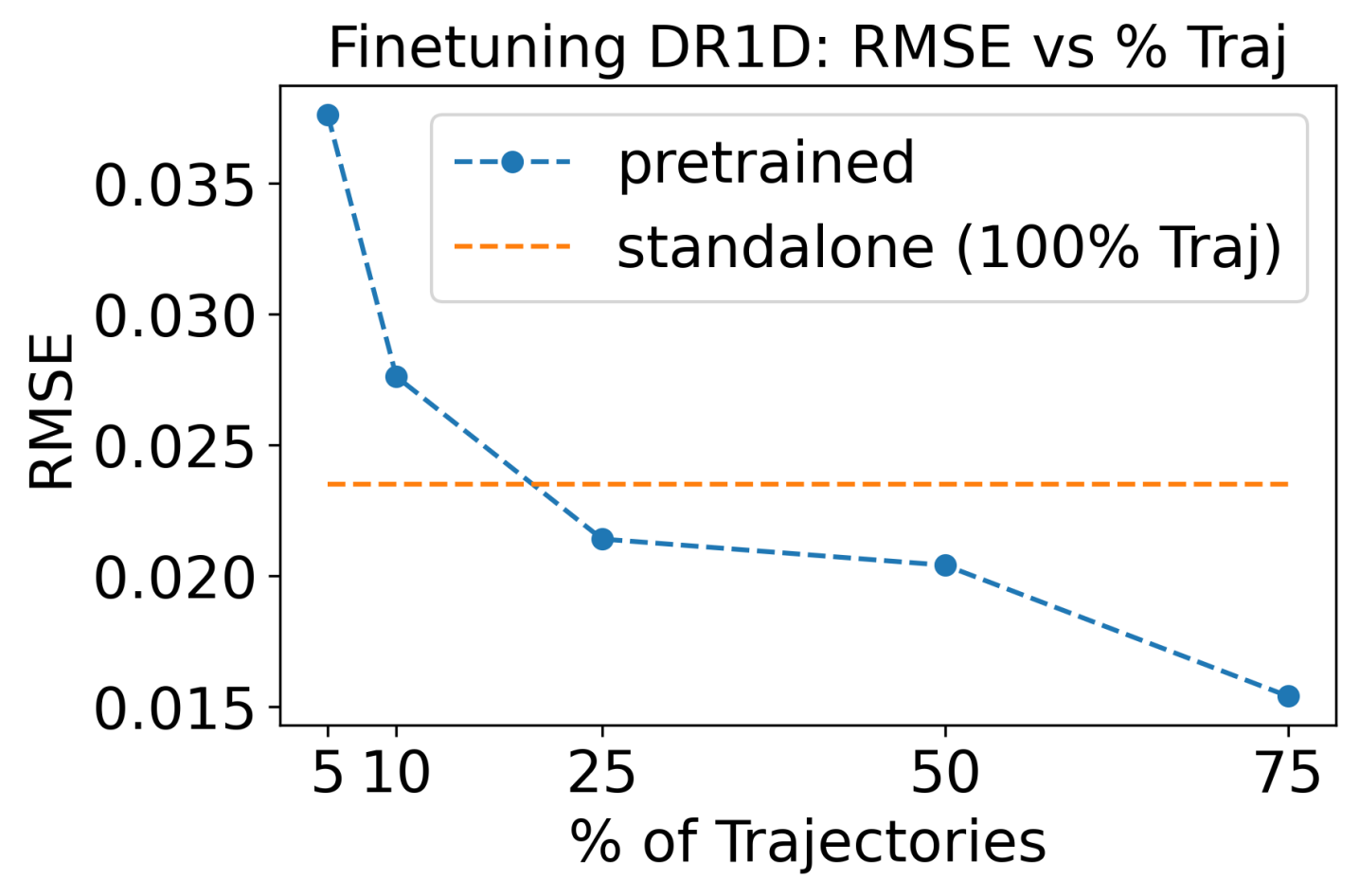}
        \caption{1D Diffusion–Reaction task.}
    \end{subfigure}
    \hfill
    \begin{subfigure}[b]{0.48\textwidth}
    \centering
        \includegraphics[width=\textwidth]{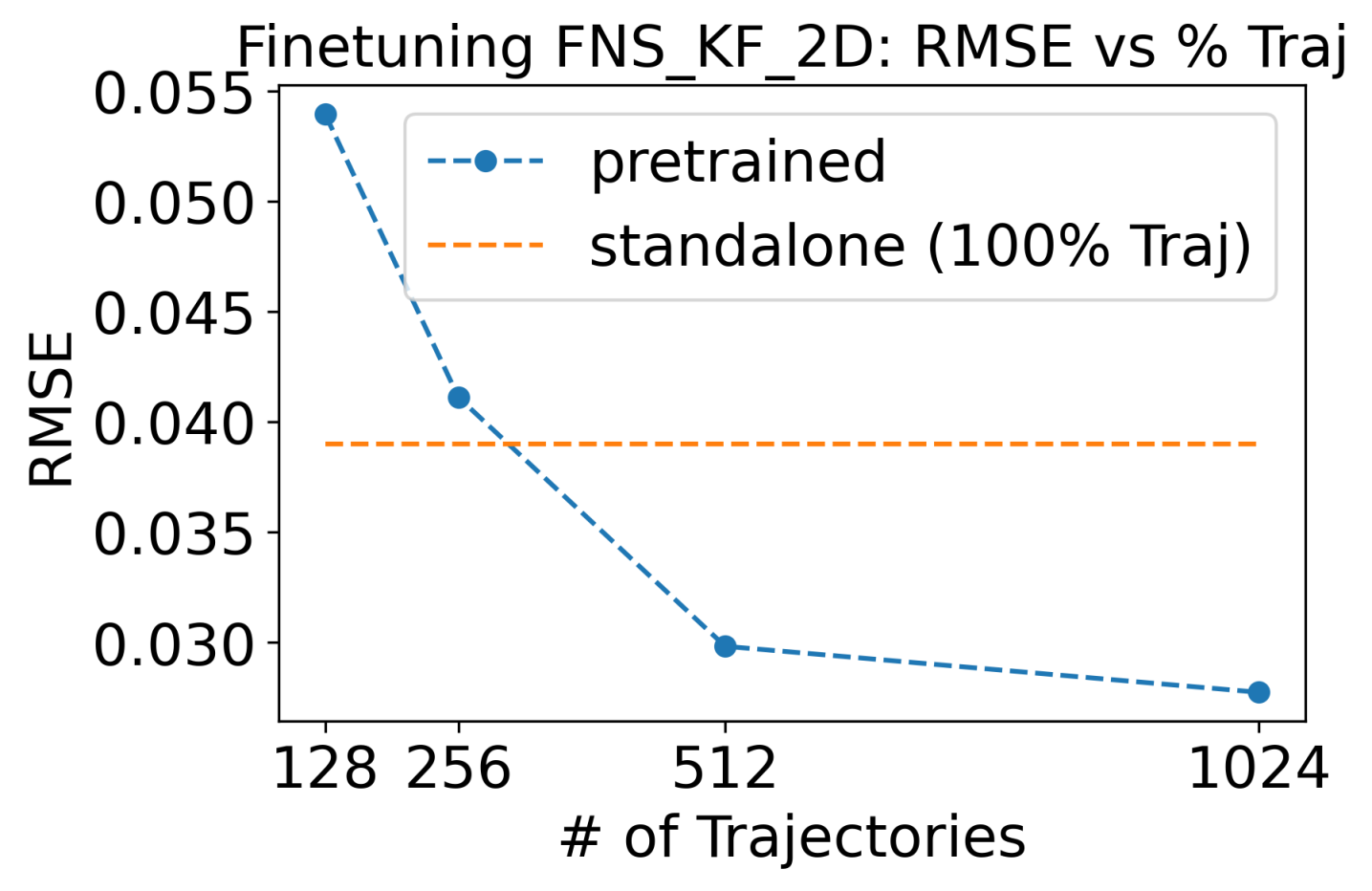}
        \caption{2D FNS-KF task.}
    \end{subfigure}
    \caption{Fine-tuning efficiency of MORPH. Pretrained models (MORPH-FM) achieve better performance with substantially less data than models trained from scratch (MORPH-SS). Adapted from \cite{rautela2025morph}.}
    \label{fig:5:5:3:1:morph_finetune}
\end{figure}

A foundation model's utility lies in its ability to adapt to new tasks with minimal data. \cref{fig:5:5:3:1:morph_finetune} compares MORPH's fine-tuning efficiency against a stand-alone MORPH model trained from scratch. On the 1D Diffusion–Reaction task, the pretrained MORPH-FM-Ti outperforms the stand-alone model using only 25\% of the training trajectories. On the 2D Forced Navier–Stokes Kolmogorov Flow (FNS-KF) task, the pretrained MORPH-FM-S surpasses the stand-alone model with fewer than 1\% of the trajectories (128 samples vs. $20\,000$). This data efficiency is a direct consequence of the diverse pretraining corpus, which teaches the model the fundamental patterns of PDE evolution.

\paragraph{Connections to Neural Operator Theory and Architecture:} 
MORPH's design reflects the theoretical foundations laid out in \cref{sec:5:3:NOsTheory}, where the kernel integral operator is introduced as the central building block for learning mappings between function spaces. Recall that the kernel operator is defined as:
\[
(\mathcal{K}(a))(x) = \int_{D} \kappa(x, y) \, a(y) \, dy, \qquad \forall x \in D,
\]
where \(\kappa(x, y)\) is a kernel function that captures the interaction between points \(x\) and \(y\) in the domain \(D\). In a discretized setting with \(L\) points, direct evaluation of this operator incurs a computational cost of \(\mathcal{O}(L^2)\), which becomes prohibitive for high-resolution problems where \(L\) can be in the millions.

The axial attention mechanism in MORPH provides a practical, factorised approximation to this kernel integral. Instead of learning a full \(L \times L\) kernel matrix, axial attention decomposes the interaction along the individual axes of the space-time domain. For a 4D domain with dimensions \((T, D, H, W)\) (time, depth, height, width), the total number of points is \(L = T \cdot D \cdot H \cdot W\). A factorised kernel can be expressed as:
\[
\kappa(x, y) \approx \kappa_t(t_x, t_y) + \kappa_d(d_x, d_y) + \kappa_h(h_x, h_y) + \kappa_w(w_x, w_y),
\]
where each \(\kappa_t, \kappa_d, \kappa_h, \kappa_w\) is a 1D kernel operating independently along its respective axis. This factorisation is implemented by applying multi-head self-attention separately along each axis. The resulting computational complexity becomes \(\mathcal{O}(T^2 + D^2 + H^2 + W^2)\), which is dramatically smaller than \(\mathcal{O}((TDHW)^2)\) for high-dimensional grids. For example, a \(64 \times 64 \times 64\) 3D grid would have \(L = 262{.}144\) points, making full attention intractable with \(\mathcal{O}(6.9 \times 10^{10})\) operations, while axial attention scales as \(\mathcal{O}(3 \times 64^2) = \mathcal{O}(12{.}288)\), a reduction of over six orders of magnitude.

This factorization aligns with the theoretical insight from \cref{sec:5:3:NOsTheory} that the kernel integral operator can be approximated by separable kernels without sacrificing the ability to model global correlations. By preserving interactions along each axis while keeping them independent, axial attention retains the expressive power to capture long-range dependencies in the data, while respecting the computational constraints of high-resolution scientific computing.

Compared to the Fourier Neural Operator (\cref{subsec:5:4:1:FNO}), which achieves global mixing through spectral convolution with complexity \(\mathcal{O}(L \log L)\), axial attention offers a complementary approach that handles non-periodic boundaries and anisotropic features differently. Relative to AFNO (\cref{subsec:5:4:2:AFNO}), which introduced block-diagonal channel mixing for vision tasks, MORPH extends the transformer backbone to arbitrary spatial dimensionalities and adds cross-field attention for multiphysics problems. And in contrast to PINO (\cref{subsec:5:4:3:PINO}), which incorporates physics-informed losses for data-efficient learning, MORPH explores a complementary route to data efficiency through large-scale pretraining and parameter-efficient fine-tuning with low-rank adapters (LoRA), achieving similar benefits without requiring explicit PDE residuals.

\subsection{Neural operators for 3D Scene Reconstruction: nPSR }

While the preceding applications focused on physics simulation, neural operators have also found a natural home in 3D computer vision, particularly for reconstructing geometric shapes from point cloud measurements. Classical approaches to 3D reconstruction often rely on solving the Poisson equation to recover surfaces from oriented points. \textit{nPSR} (Neural Poisson Surface Reconstruction) \cite{andrade2023poissonnet} replaces the traditional numerical solver with a Fourier Neural Operator (\cref{subsec:5:4:1:FNO}), learning the solution operator that maps point cloud measurements to the reconstructed 3D surface.

Given an oriented point cloud \(\mathcal{P} = \{(\mathbf{p}_i, \mathbf{n}_i)\}_{i=1}^N\) representing a 3D surface, where \(\mathbf{p}_i \in \mathbb{R}^3\) are point coordinates and \(\mathbf{n}_i \in \mathbb{R}^3\) are unit normals, the goal is to reconstruct a watertight mesh that approximates the underlying surface. Poisson surface reconstruction \cite{kazhdan2006poisson} formulates this as solving the Poisson equation:
\begin{equation}
\nabla^2 \phi = \nabla \cdot \mathbf{V}, \qquad \mathbf{V}(\mathbf{x}) = \sum_{i=1}^N \mathbf{n}_i \, \delta(\mathbf{x} - \mathbf{p}_i),
\end{equation}
where \(\mathbf{V}\) is a vector field derived from the oriented points (the normals extended to the volume), \(\phi\) is an implicit function whose level set \(\phi = \text{constant}\) defines the reconstructed surface, and \(\delta\) denotes the Dirac delta distribution. The solution \(\phi\) is then obtained by solving:
\begin{equation}
\phi(\mathbf{x}) = \int_{\mathbb{R}^3} G(\mathbf{x} - \mathbf{y}) \, (\nabla \cdot \mathbf{V})(\mathbf{y}) \, d\mathbf{y},
\end{equation}
where \(G(\mathbf{x}) = - 1/(4\pi\|\mathbf{x}\|)\) is the fundamental solution (Green's function) of the Laplace operator in \(\mathbb{R}^3\). This formulation highlights the operator nature of the problem: there exists an operator \(\mathcal{G}\) that maps the input vector field \(\mathbf{V}\) to the implicit function \(\phi\). Traditional methods discretize this equation on a uniform grid using finite differences, which limits resolution, requires careful parameter tuning, and does not easily generalize to irregular sampling patterns.


\begin{figure}[t]
    \centering
    \includegraphics[width=0.9\textwidth]{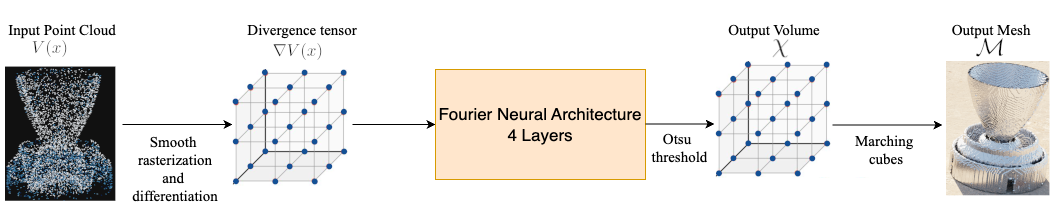}
    \caption{nPSR architecture: oriented point clouds are converted to a vector field, processed by Fourier Neural Operator layers in the spectral domain, and decoded to the implicit surface representation. The resolution-agnostic property enables training on coarse grids and inference at high resolution. Adapted from \cite{andrade2023poissonnet}.}
    \label{fig:5:5:4:poissonnet_arch}
\end{figure}

nPSR learns the operator \(\mathcal{G}: f=\nabla\cdot \mathbf{V} \mapsto \phi\) using a Fourier Neural Operator. Recall from Section~\ref{subsec:5:4:1:FNO} that the Fourier Neural Operator parameterises the kernel integral operator in the spectral domain:
\begin{equation}
(\mathcal{K}_\theta(\mathbf{V}))(\mathbf{x}) = \FFT^{-1}\left( R_\theta(\mathbf{k}) \cdot \FFT(\mathbf{V})(\mathbf{k}) \right)(\mathbf{x}),
\end{equation}
where \(\FFT\) denotes the Fourier transform, \(R_\theta\) is a learnable complex-valued weight function in frequency space, and \(\mathbf{k}\) denotes the frequency coordinates. For the 3D Poisson equation, the FNO architecture must learn the mapping from the 3D vector field \(\mathbf{V}\) (with 3 channels) to the scalar field \(\phi\) (1 channel). The architecture stacks multiple FNO layers; the final layer outputs the implicit function \(\phi\), from which the reconstructed surface is extracted via marching cubes.

As we have seen, key theoretical property of FNOs is their \textbf{discretization invariance}: they learn a mapping between function spaces that is independent of the discretization used during training. For Poisson surface reconstruction, this means:
\begin{itemize}
    \item A model trained on coarse grids (e.g., \(64^3\) resolution) can be evaluated at arbitrarily higher resolutions (e.g., \(256^3\) or \(512^3\)) without retraining
    \item The operator can be applied to point clouds with varying sampling densities
    \item The method naturally handles large-scale scenes by operating at the desired output resolution
\end{itemize}

\paragraph{Training Methodology:} nPSR  is trained in a supervised manner using synthetic data generated from known 3D shapes. The authors use the ShapeNet dataset \cite{chang2015shapenet}, which contains over $50\,000$ 3D models across 55 categories. For each shape:
\begin{enumerate}
    \item A watertight mesh is converted to an implicit function \(\phi_{\text{gt}}\) via \textit{signed distance function} (SDF) computation on a regular grid
    \item Oriented point clouds are generated by sampling points on the surface, with normals computed from the mesh
    \item The vector field \(\mathbf{V}_{\text{input}}\) is constructed by spreading the point normals to the volume using a Gaussian kernel
\end{enumerate}

\begin{figure}[t]
      \centering
      \includegraphics[width=\linewidth, trim=0 10mm 0 10mm, clip]{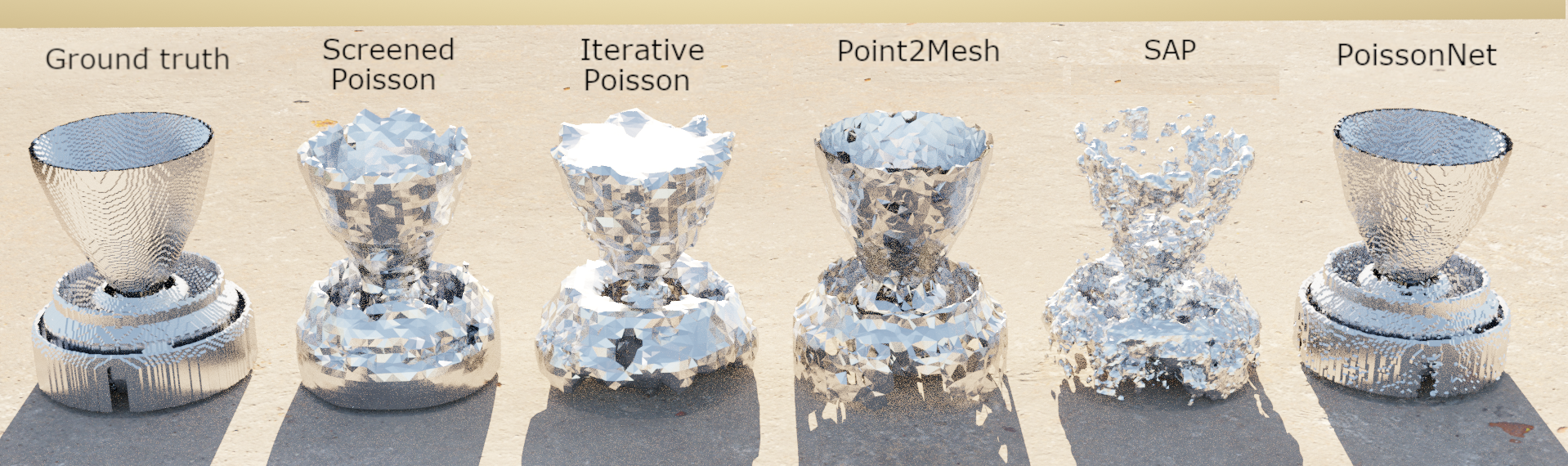}
        \includegraphics[width=\linewidth, trim=0 5mm 0 40mm, clip]{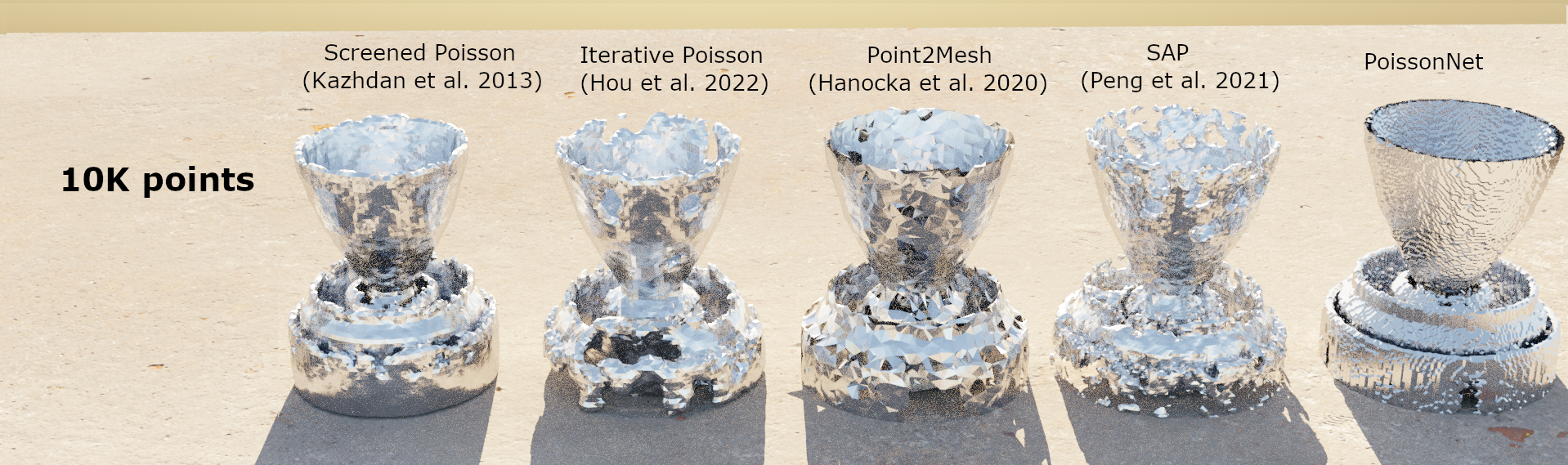}
        \includegraphics[width=\linewidth, trim=0 10mm 0 40mm, clip]{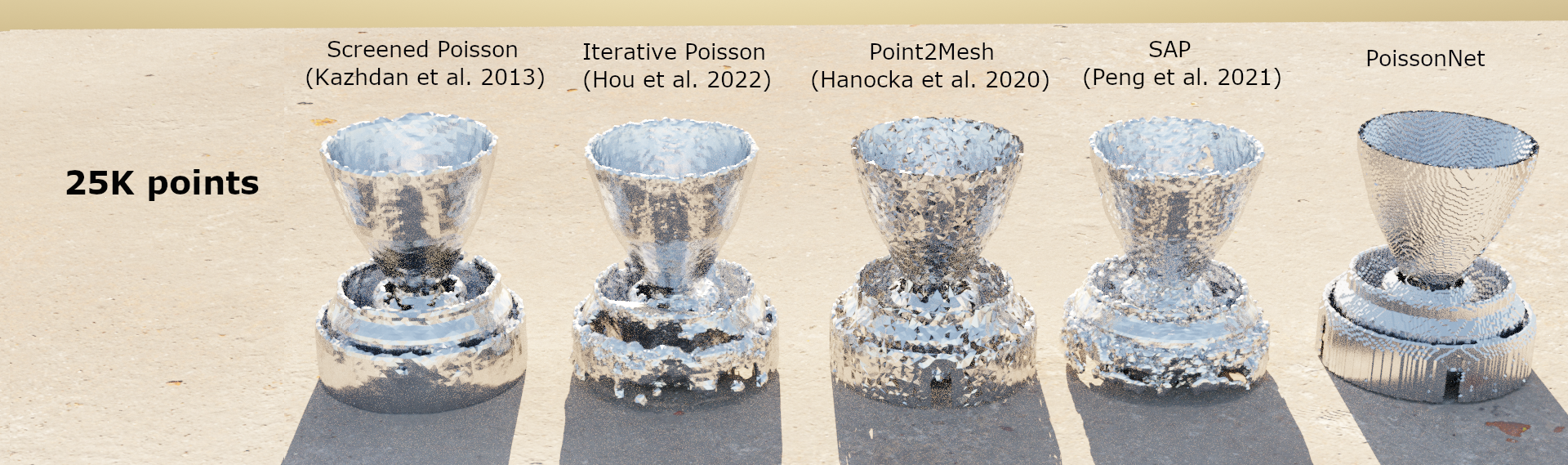}
        \includegraphics[width=\linewidth, trim=0 10mm 0 40mm, clip]{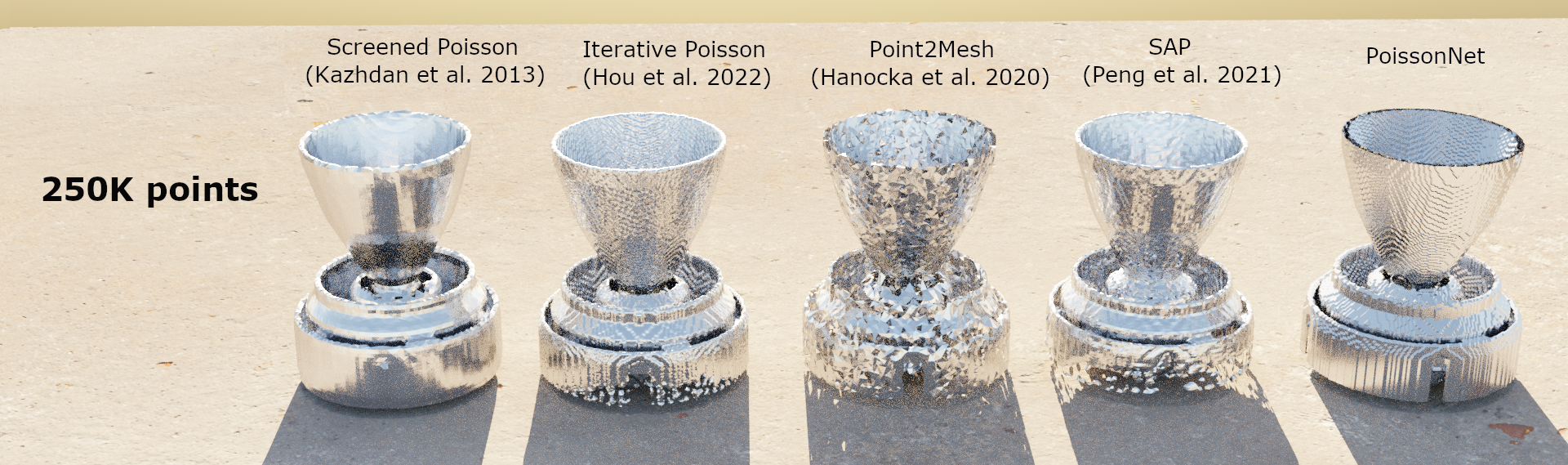}
\caption{Result on a ShapeNet example with different sampling rates, the first row corresponds to $3\,000$ points. From \cite{andrade2023poissonnet}.}
\label{fig:shapenet-3}
\end{figure}

The training data consists of pairs \((\mathbf{V}_{\text{input}}, \phi_{\text{gt}})\) at varying grid resolutions. The authors generate data at \(64^3\), \(128^3\), and \(256^3\) resolutions to evaluate resolution-agnostic properties.

For supervision, nPSR uses voxelised ShapeNet models as ground-truth occupancy grids. Synthetic oriented point clouds are generated from these shapes by extracting a mesh, sampling surface points, and assigning normals. 

The oriented point cloud is then converted to the FNO input by:
\begin{enumerate}
    \item rasterising points and normals onto a regular voxel grid;
    \item smoothing the resulting normal field with a Gaussian filter;
    \item computing its divergence using finite differences;
    \item smoothing the divergence field again and adding a small amount of Gaussian noise during training.
\end{enumerate}
The neural operator is trained with a voxel-wise mean-squared error,
\[
    \mathcal L_{\mathrm{MSE}}
    =
    \|o_{\mathrm{pred}}-o_{\mathrm{gt}}\|_2^2.
\]

Their FNO architecture has:
\begin{itemize}
    \item 4 3D FNO blocks;
    \item \(k_{\max}=20\) retained Fourier modes;
    \item GELU nonlinearities and dropout after the first three blocks;
    \item Adam with learning rate \(2.5\times10^{-3}\) and
          weight decay \(10^{-4}\);
    \item batch size \(20\);
    \item approximately \(10\,000\) optimisation steps.
\end{itemize}

In the reported experiments, the same nPSR parameters trained on \(64^3\) grids retain similar reconstruction quality when evaluated at \(128^3\), without retraining. This provides a concrete example of the cross-resolution capability enabled by the FNO parameterization, while the complete nPSR pipeline remains constrained by its regular-grid representation and finite computational resources.



    \backmatter         

    \printbibliography

\end{document}